\documentclass[lettersize,journal]{IEEEtran}
\usepackage{amsmath,amsfonts}
\usepackage{array}
\usepackage[caption=false,font=normalsize,labelfont=sf,textfont=sf]{subfig}
\usepackage{textcomp}
\usepackage{stfloats}
\usepackage{url}
\usepackage{verbatim}
\usepackage{graphicx}
\usepackage{bbm}
\def\BibTeX{{\rm B\kern-.05em{\sc i\kern-.025em b}\kern-.08em
    T\kern-.1667em\lower.7ex\hbox{E}\kern-.125emX}}
\usepackage{balance}
\usepackage{newclude}

\usepackage{xcolor}
\usepackage{booktabs}
\usepackage{multirow}
\usepackage{tabu}
\usepackage{multicol}
\usepackage{colortbl}
\usepackage{caption}
\usepackage[ruled,vlined]{algorithm2e}
\usepackage{algpseudocode}
\usepackage[hidelinks]{hyperref}
\usepackage{subcaption}
\usepackage{graphicx}

\usepackage{textcomp}
\newcommand{\degree}{\textdegree}

\definecolor{Green}{rgb}{0.0, 0.5, 0.0}
\definecolor{Blue}{rgb}{0.0, 0.0, 0.5}
\definecolor{Red}{rgb}{0.5, 0.0, 0.0}
\definecolor{Orchid}{rgb}{0.729, 0.333, 0.827}

\definecolor{darkgreen}{RGB}{0,200,0}

\definecolor{tabfirst}{rgb}{0.96, 0.77, 0.77} 
\definecolor{tabsecond}{rgb}{0.98 , 0.93, 0.77} 
\definecolor{tabthird}{rgb}{1, 1, 0.7} 

\newcommand{\myparagraph}[1]{\vspace{0.5em}\noindent\textbf{#1}}

\newcommand{\scfirst}[1]{\colorbox{tabfirst}{#1}}
\newcommand{\scsecond}[1]{\colorbox{tabsecond}{#1}}

\usepackage{mathtools}
\DeclarePairedDelimiter{\parens}{\lparen}{\rparen}

\newcommand{\vect}[1]{\ensuremath{\mathbf{#1}}}

\newcommand{\R}[1]{{%
    \textbf{%
        \ifstrequal{#1}{1}{\textcolor{red}{R#1}}{%
        \ifstrequal{#1}{2}{\textcolor{blue}{R#1}}{%
        \ifstrequal{#1}{3}{\textcolor{magenta}{R#1}}{%
        \ifstrequal{#1}{4}{\textcolor{teal}{R#1}}{%
                           \textcolor{cyan}{R#1}%
        }}}}%
    }%
}}

\newcounter{subsubsubsection}[subsubsection]

\begin{document}

\title{DiffusionLight-Turbo: Accelerated Light Probes for Free via Single-Pass Chrome Ball Inpainting}

\renewcommand{\thefootnote}{\fnsymbol{footnote}}
\author{
Worameth Chinchuthakun\mbox{$^{*\dagger}$},
Pakkapon Phongthawee\mbox{$^{*}$},
Amit Raj,
\\
Varun Jampani,
Pramook Khungurn, and
Supasorn Suwajanakorn \mbox{$^\ddagger$}
}
\markboth{Journal of \LaTeX\ Class Files,~Vol.~18, No.~9, September~2020}%
{How to Use the IEEEtran \LaTeX \ Templates}


\twocolumn[{%
\renewcommand\twocolumn[1][]{#1}%
\maketitle
\vspace{-3em}
\begin{center}
    \centering
    \captionsetup{type=figure}
    \includegraphics[width=0.98\textwidth]{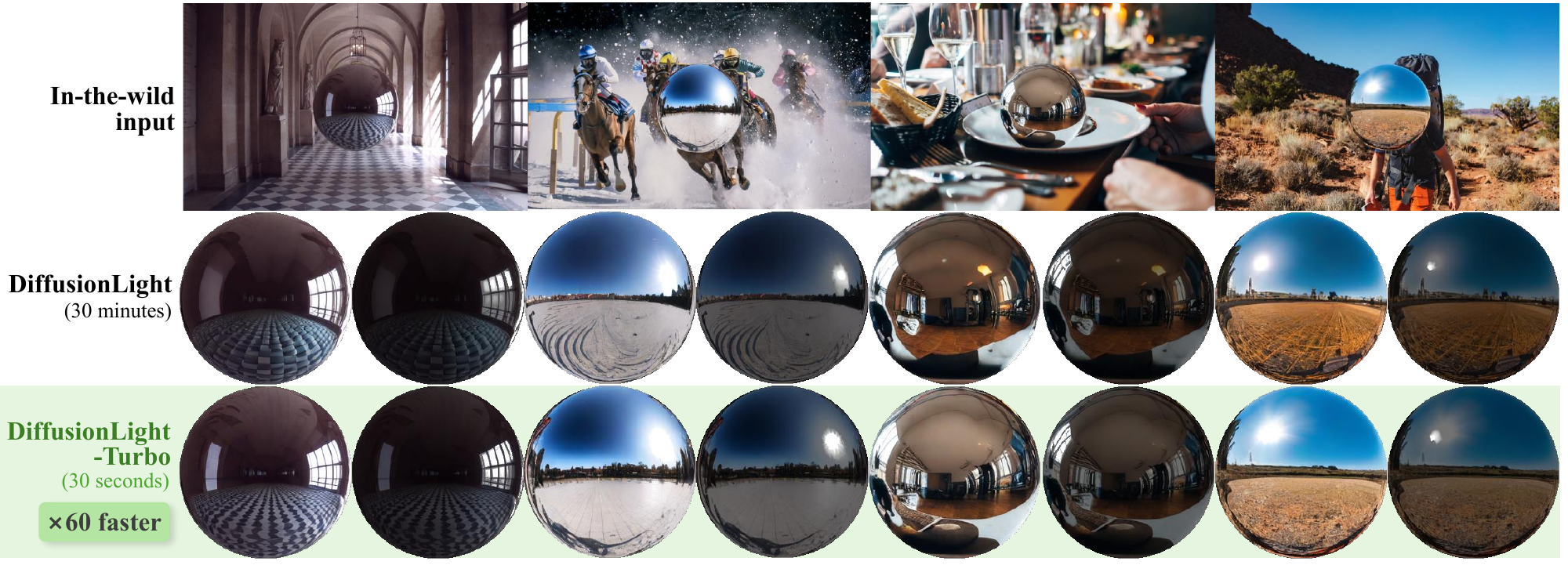}
    \captionof{figure}{\textbf{DiffusionLight-Turbo} leverages a pre-trained diffusion model (Stable Diffusion XL) for light estimation by synthesizing an HDR chrome ball. We reduce the runtime of DiffusionLight \cite{phongthawee2024diffusionlight}, our conference version, from approximately 30 \emph{minutes} to about 30 \emph{seconds} (on a single NVIDIA RTX 3090 Ti), with minimal quality degradation. In each scene, we show our normally exposed chrome ball on the left and our underexposed version, which reveals bright light sources, on the right.
    } \label{fig:teaser}
\end{center}%
}]


\begin{abstract}
We introduce a simple yet effective technique for estimating lighting from a single low-dynamic-range (LDR) image by reframing the task as a chrome ball inpainting problem. This approach leverages a pre-trained diffusion model, Stable Diffusion XL, to overcome the generalization failures of existing methods that rely on limited HDR panorama datasets.
While conceptually simple, the task remains challenging because diffusion models often insert incorrect or inconsistent content and cannot readily generate chrome balls in HDR format.
Our analysis reveals that the inpainting process is highly sensitive to the initial noise in the diffusion process, occasionally resulting in unrealistic outputs. 
To address this, we first introduce DiffusionLight~\cite{phongthawee2024diffusionlight}, which uses iterative inpainting to compute a median chrome ball from multiple outputs to serve as a stable, low-frequency lighting prior that guides the generation of a high-quality final result. 
To generate high-dynamic-range (HDR) light probes, an Exposure LoRA is fine-tuned to create LDR images at multiple exposure values, which are then merged.
While effective, DiffusionLight is time-intensive, requiring approximately 30 minutes per estimation. To reduce this overhead, we introduce DiffusionLight-Turbo, which reduces the runtime to about 30 seconds with minimal quality loss. This 60x speedup is achieved by training a Turbo LoRA to directly predict the averaged chrome balls from the iterative process. Inference is further streamlined into a single denoising pass using a LoRA swapping technique.
Experimental results that show our method produces convincing light estimates across diverse settings and demonstrates superior generalization to in-the-wild scenarios. Our code is available at \href{https://diffusionlight.github.io/turbo}{\textcolor{magenta}{https://diffusionlight.github.io/turbo}}.

\footnotetext[1]{Authors contributed equally to this work.}
\footnotetext[2]{Work done during research assistantship at VISTEC.}
\footnotetext[3]{Corresponding author.}

\end{abstract}
\renewcommand{\thefootnote}{\arabic{footnote}}

\begin{IEEEkeywords}
Light estimation, diffusion models, image inpainting
\end{IEEEkeywords}

\section{Introduction}
\label{sec:intro}

Accurate lighting estimation is crucial for applications in mixed reality, such as virtual object insertion and relighting. In many cases, only a single low-dynamic-range (LDR) image with a limited field of view (FOV) is available. This poses challenges for lighting estimation, as the target environment
map must have a high dynamic range (HDR) to accurately capture the true intensity of incoming light. It also needs to cover regions beyond the limited FOV of the input image. These challenges have led to numerous attempts to \emph{regress} HDR environment maps from LDR images.

A common strategy used in state-of-the-art techniques is to train a neural regressor with a dataset of HDR panoramas. For example, StyleLight \cite{wang2022stylelight} trains a GAN on thousands of panoramas and, at test time, uses GAN inversion to find a latent code that generates a full panorama whose cropped region matches the input image. Everlight \cite{dastjerdi2023everlight} trains a conditional GAN on 200k panoramas to directly predict an HDR map from an input image. However, these training panoramas offer limited scene variety, often featuring typical viewpoints like those from a room's center due to tripod use. Panoramas featuring elephants viewed from inside a safari jeep, for example, would be extremely rare (Figure \ref{fig:main_qualitative_wild}). 
So, how can we estimate lighting in the wild for any image, under any scenario?

\begin{figure}
    \centering

    \begin{tabu} to \textwidth {
        @{}
        c@{}
        c@{\hspace{0.5pt}}
        c@{\hspace{0.5pt}}
        c@{\hspace{0.5pt}}
        c@{\hspace{0.5pt}}
        c@{\hspace{0.5pt}}
        c@{}
    }
        
        \multicolumn{1}{l}{\rotatebox[origin=c]{90}{\shortstack[l]{\scriptsize Input \\ \scriptsize image}}} &
        \noindent\parbox[c]{0.083\textwidth}{\includegraphics[width=0.083\textwidth]{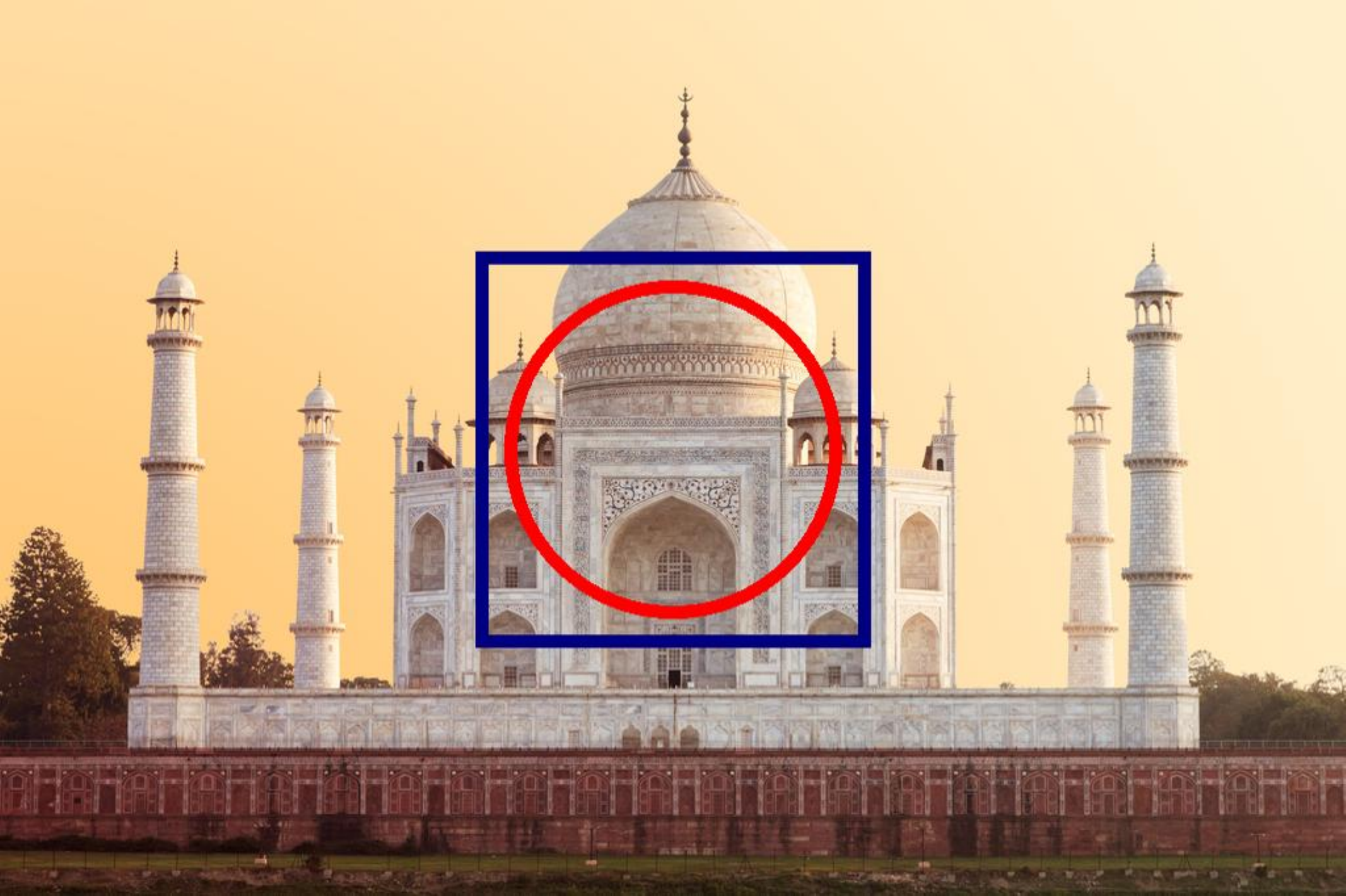}} & 
        \noindent\parbox[c]{0.083\textwidth}{\includegraphics[width=0.083\textwidth]{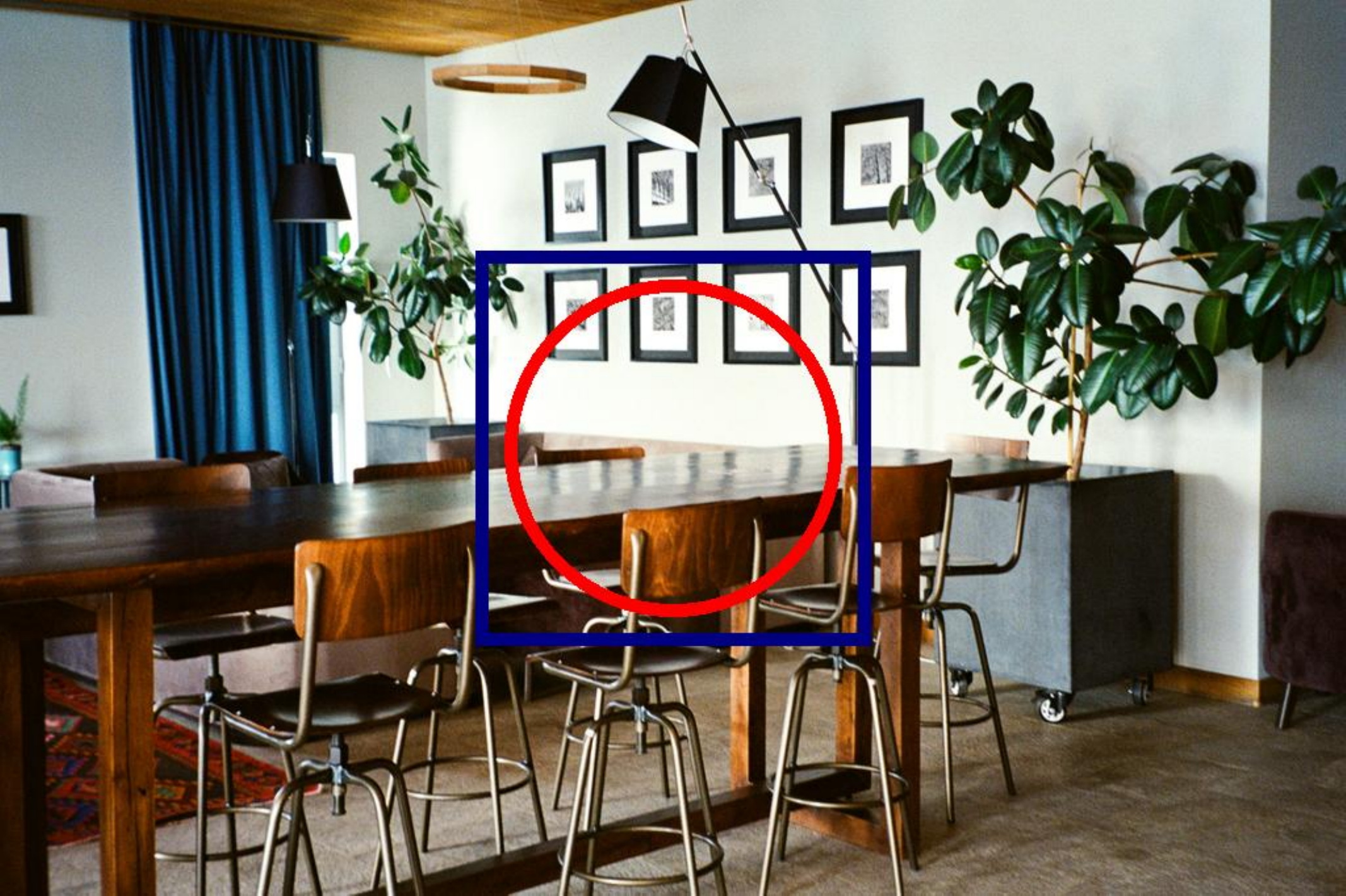}} & 
        \noindent\parbox[c]{0.083\textwidth}{\includegraphics[width=0.083\textwidth]{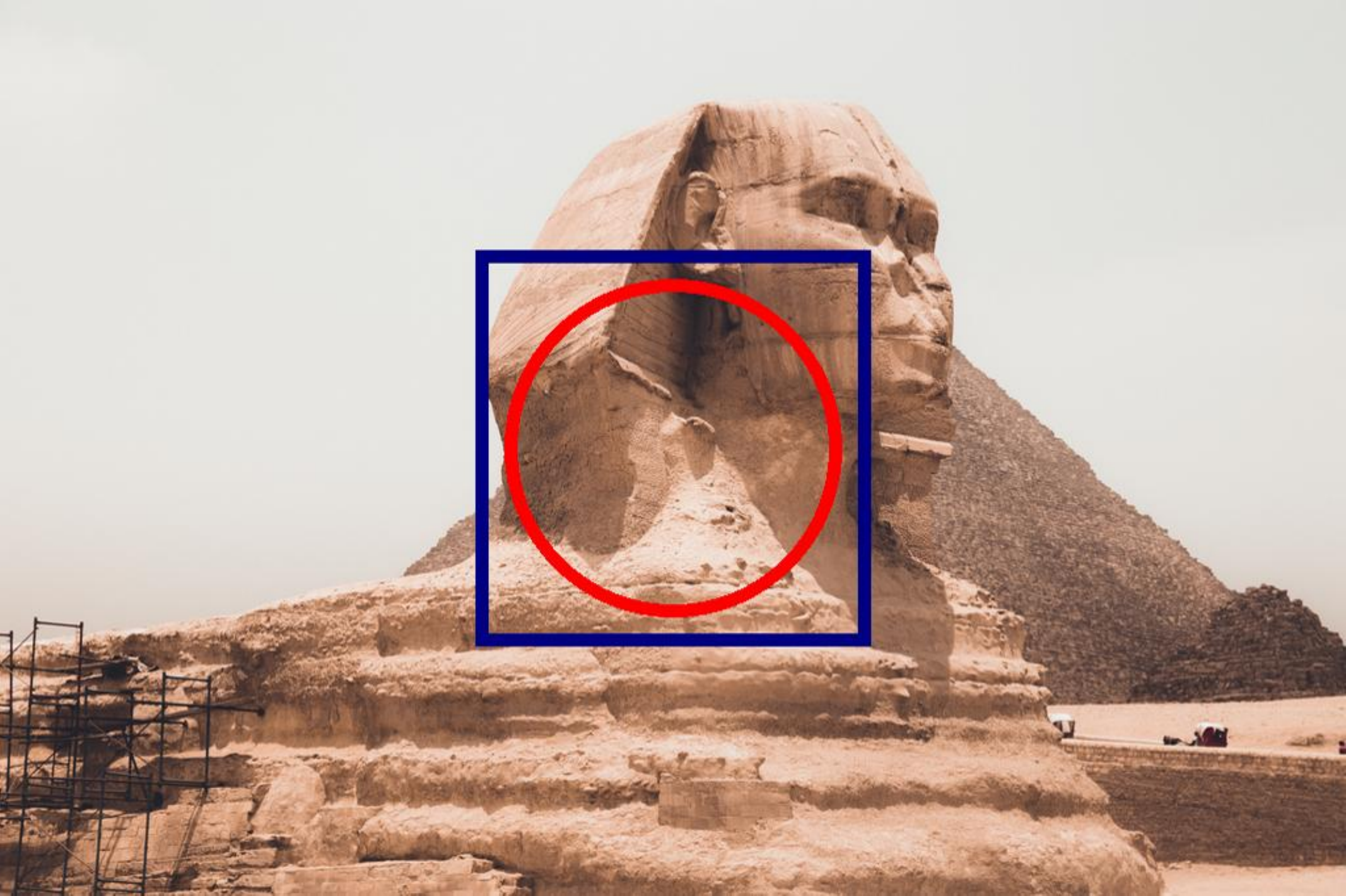}} & 
        \noindent\parbox[c]{0.083\textwidth}{\includegraphics[width=0.083\textwidth]{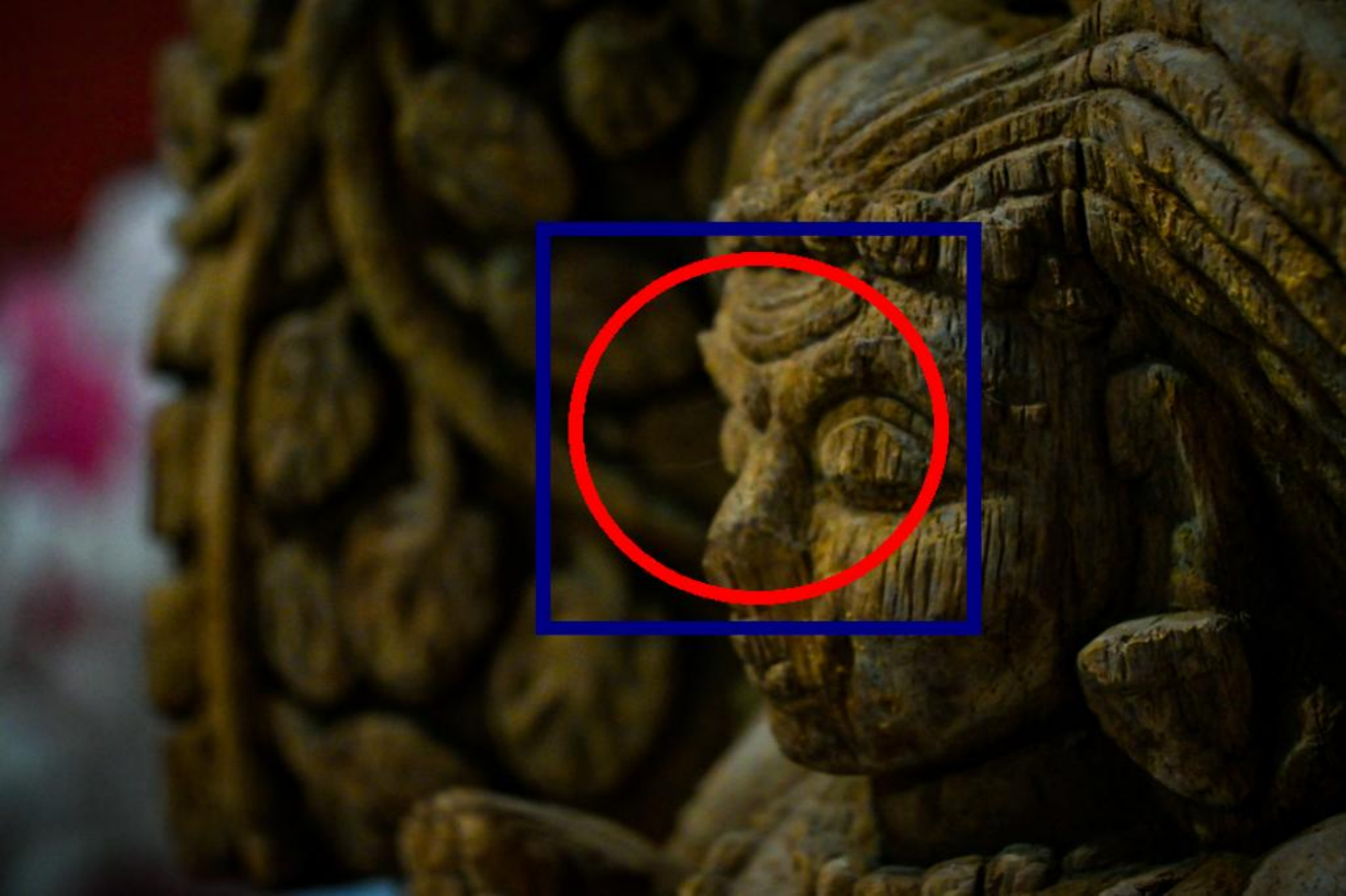}} & 
        \noindent\parbox[c]{0.083\textwidth}{\includegraphics[width=0.083\textwidth]{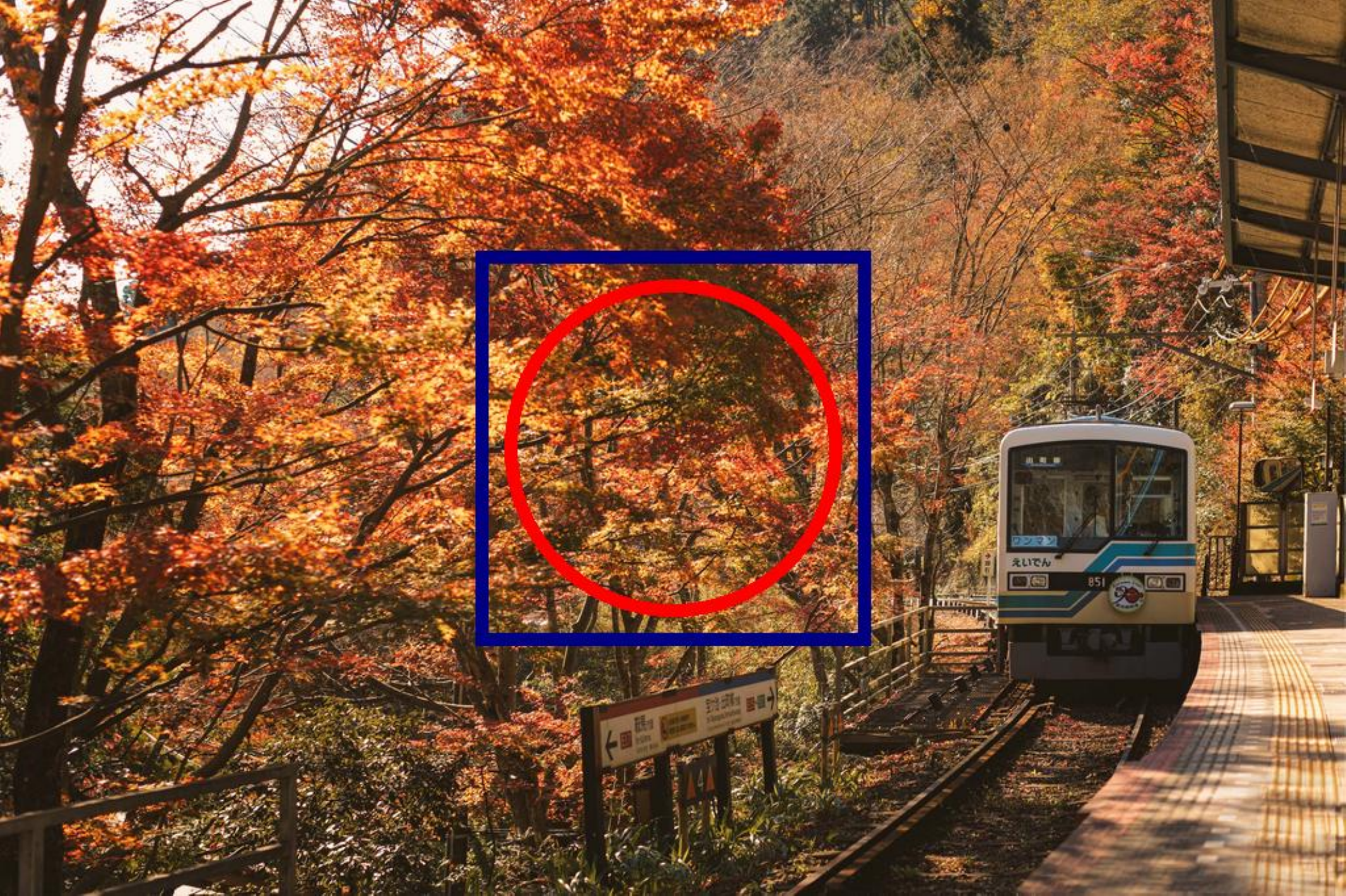}} &
        \\

        \multicolumn{1}{l}{\rotatebox[origin=c]{90}{\shortstack[l]{\scriptsize Blended Dif-\\ \scriptsize fusion \cite{avrahami2023blendedlatent, avrahami2022blendeddiffusion}}}} &
        \noindent\parbox[c]{0.083\textwidth}{\includegraphics[width=0.083\textwidth]{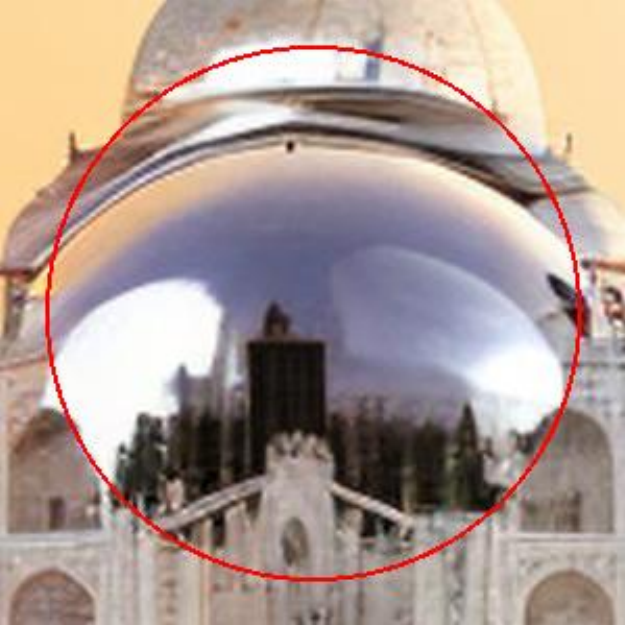}} & 
        \noindent\parbox[c]{0.083\textwidth}{\includegraphics[width=0.083\textwidth]{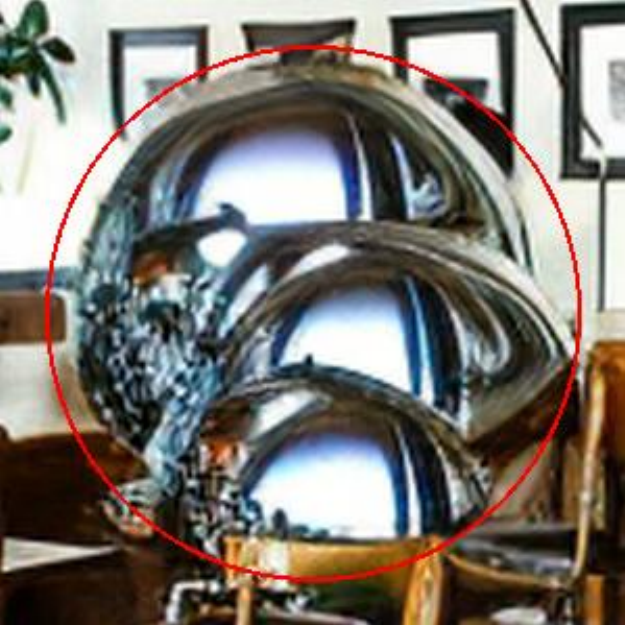}} & 
        \noindent\parbox[c]{0.083\textwidth}{\includegraphics[width=0.083\textwidth]{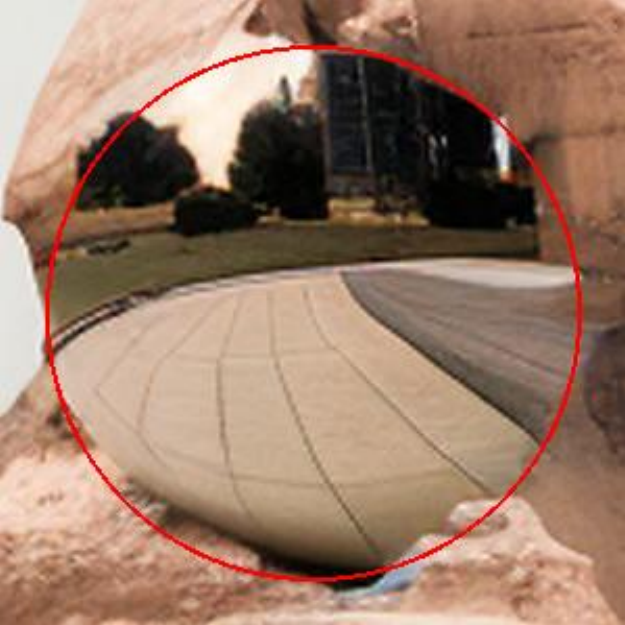}} & 
        \noindent\parbox[c]{0.083\textwidth}{\includegraphics[width=0.083\textwidth]{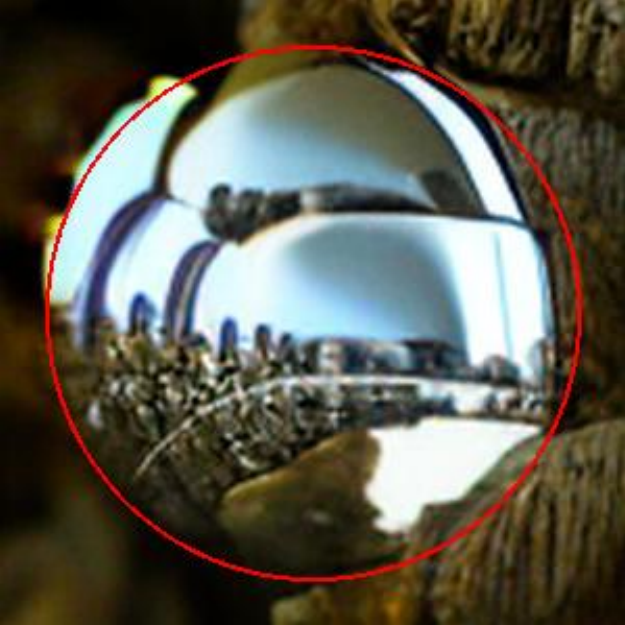}} & 
        \noindent\parbox[c]{0.083\textwidth}{\includegraphics[width=0.083\textwidth]{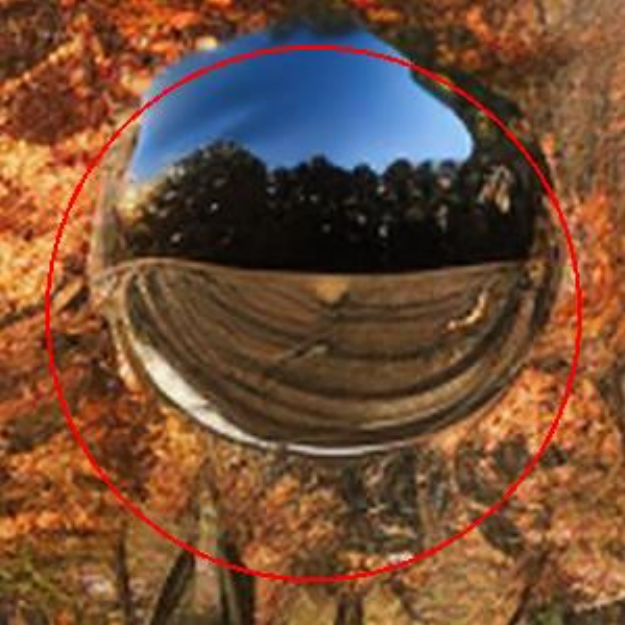}} &
        
        \\

        \multicolumn{1}{l}{\rotatebox[origin=c]{90}{\shortstack[l]{\scriptsize Paint-by-Ex\\ \scriptsize ample \cite{yang2023paint}}}} &
        \noindent\parbox[c]{0.083\textwidth}{\includegraphics[width=0.083\textwidth]{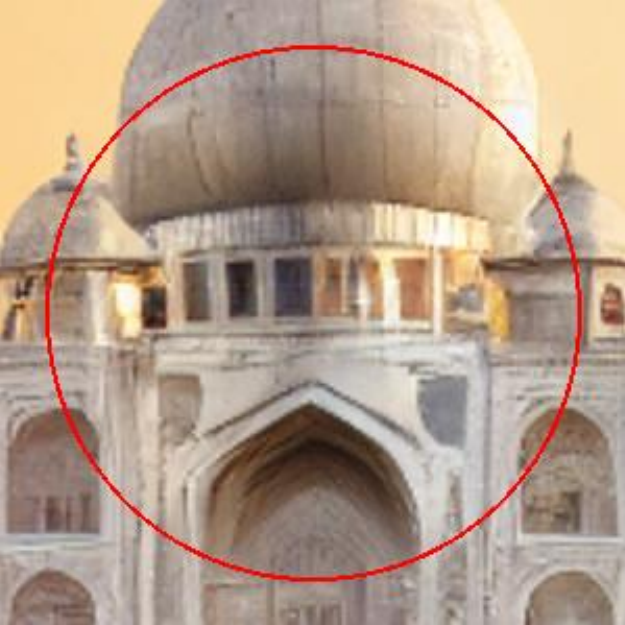}} & 
        \noindent\parbox[c]{0.083\textwidth}{\includegraphics[width=0.083\textwidth]{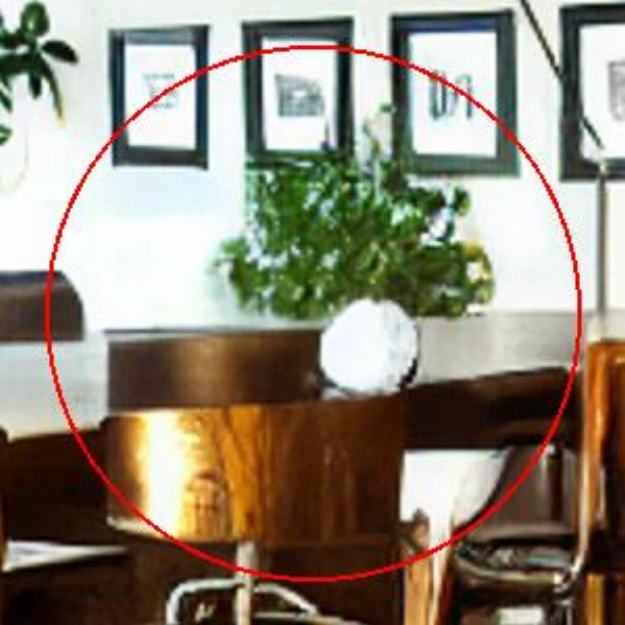}} & 
        \noindent\parbox[c]{0.083\textwidth}{\includegraphics[width=0.083\textwidth]{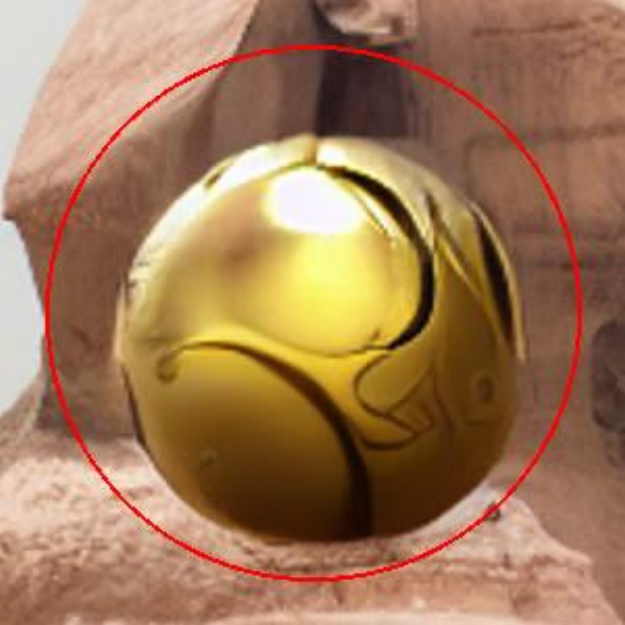}} & 
        \noindent\parbox[c]{0.083\textwidth}{\includegraphics[width=0.083\textwidth]{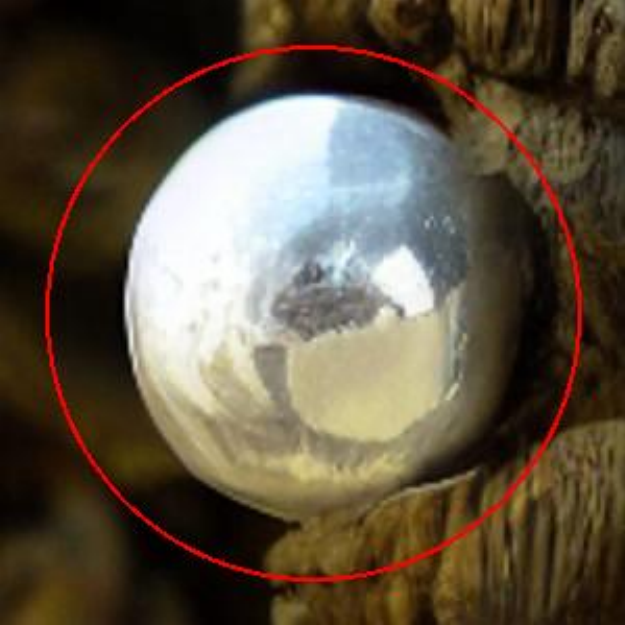}} & 
        \noindent\parbox[c]{0.083\textwidth}{\includegraphics[width=0.083\textwidth]{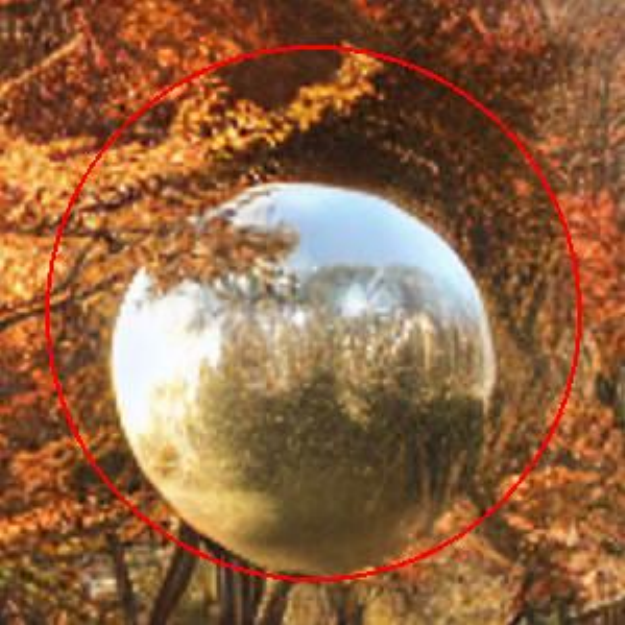}} &
        
        \\

        \multicolumn{1}{l}{\rotatebox[origin=c]{90}{\shortstack[l]{\scriptsize IP-Adapter\\ \scriptsize \cite{ye2023ip-adapter}}}} &
        \noindent\parbox[c]{0.083\textwidth}{\includegraphics[width=0.083\textwidth]{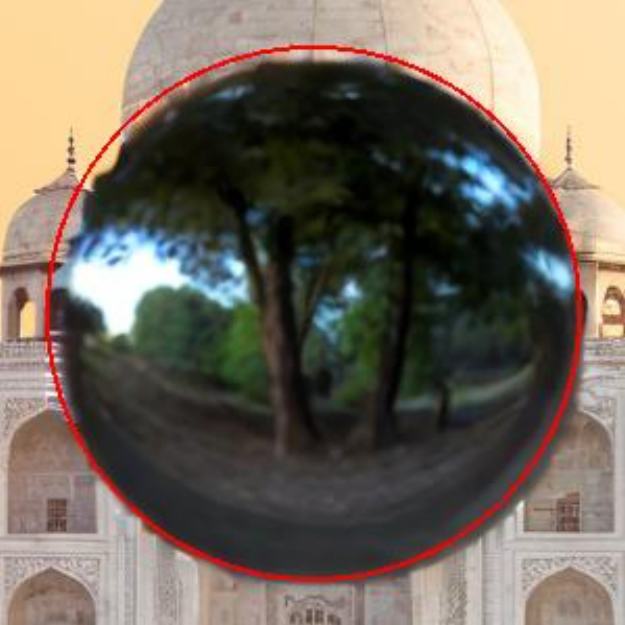}} & 
        \noindent\parbox[c]{0.083\textwidth}{\includegraphics[width=0.083\textwidth]{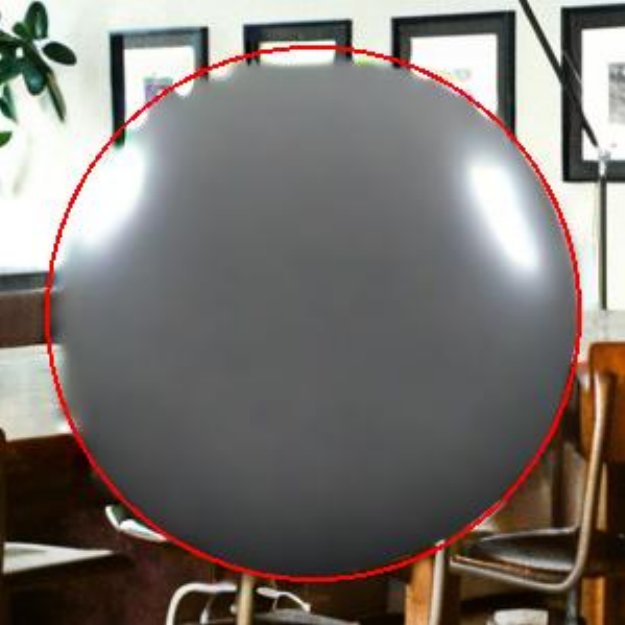}} & 
        \noindent\parbox[c]{0.083\textwidth}{\includegraphics[width=0.083\textwidth]{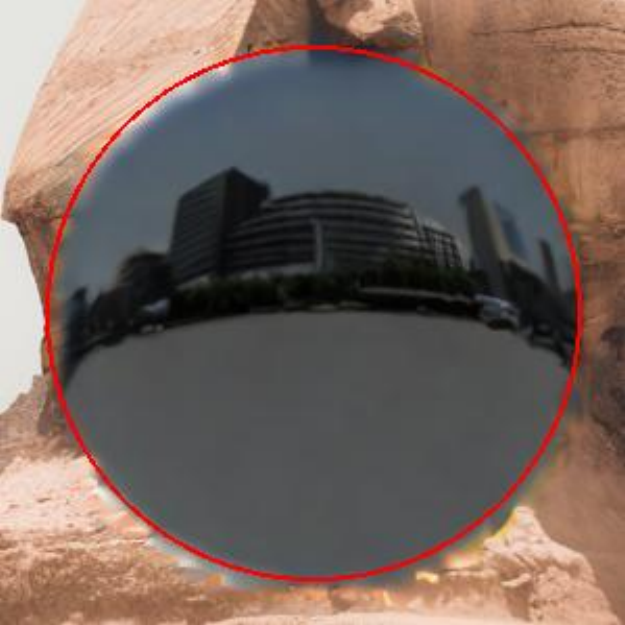}} & 
        \noindent\parbox[c]{0.083\textwidth}{\includegraphics[width=0.083\textwidth]{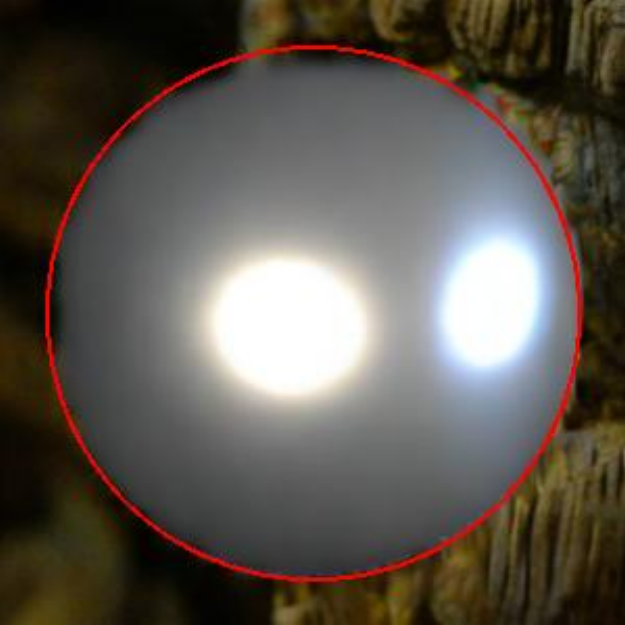}} & 
        \noindent\parbox[c]{0.083\textwidth}{\includegraphics[width=0.083\textwidth]{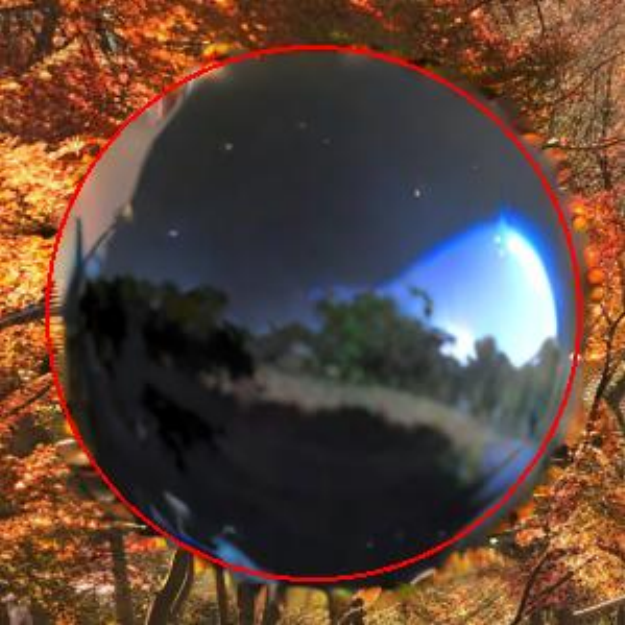}} &
        
        \\

        \multicolumn{1}{l}{\rotatebox[origin=c]{90}{\shortstack[l]{\scriptsize DALL·E2 \cite{dalle2}}}} &
        \noindent\parbox[c]{0.083\textwidth}{\includegraphics[width=0.083\textwidth]{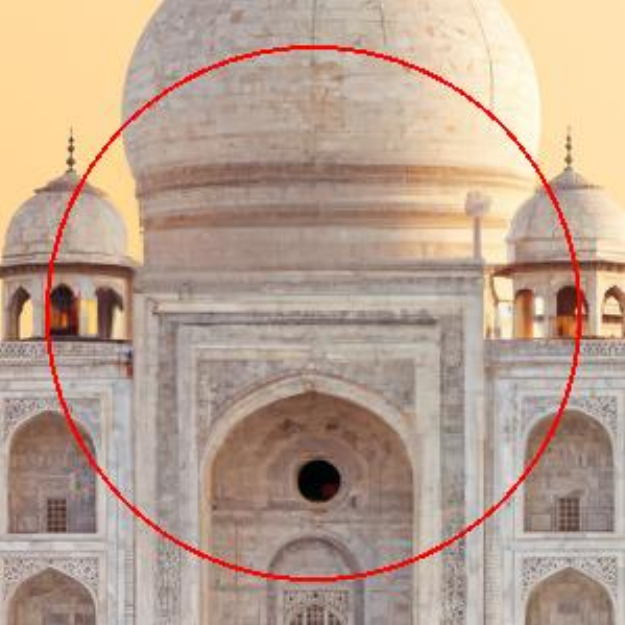}} & 
        \noindent\parbox[c]{0.083\textwidth}{\includegraphics[width=0.083\textwidth]{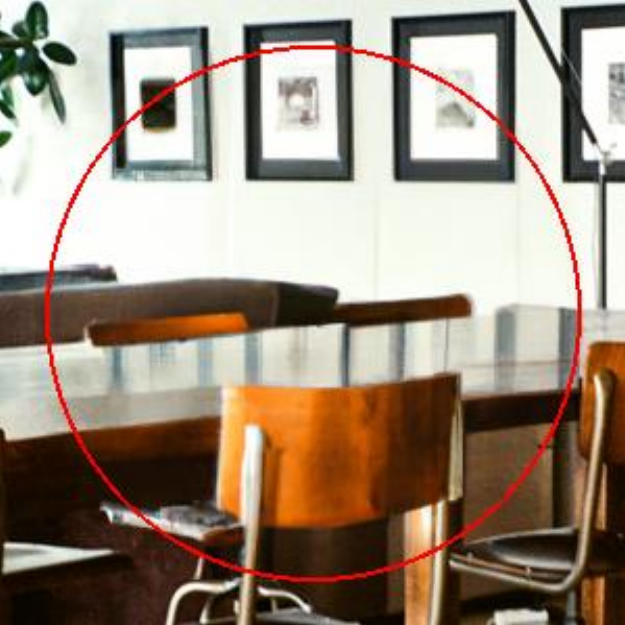}} & 
        \noindent\parbox[c]{0.083\textwidth}{\includegraphics[width=0.083\textwidth]{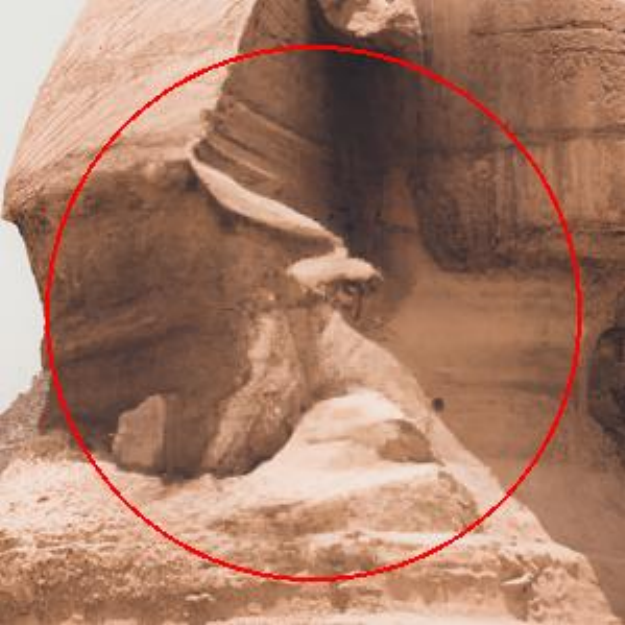}} & 
        \noindent\parbox[c]{0.083\textwidth}{\includegraphics[width=0.083\textwidth]{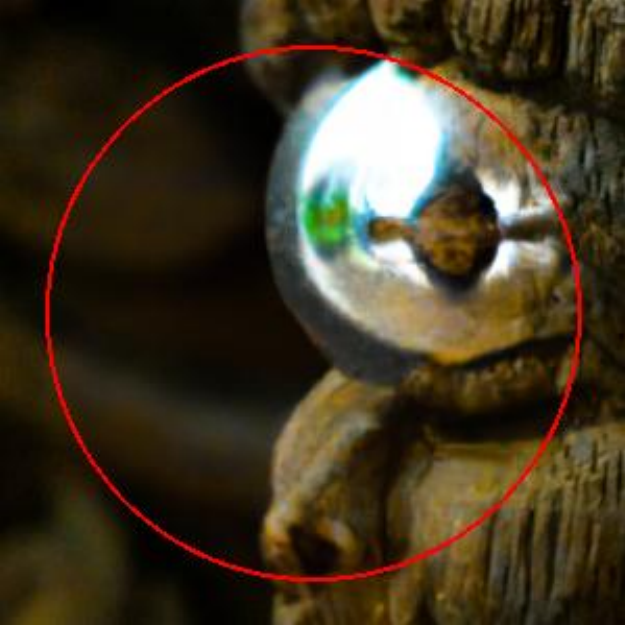}} & 
        \noindent\parbox[c]{0.083\textwidth}{\includegraphics[width=0.083\textwidth]{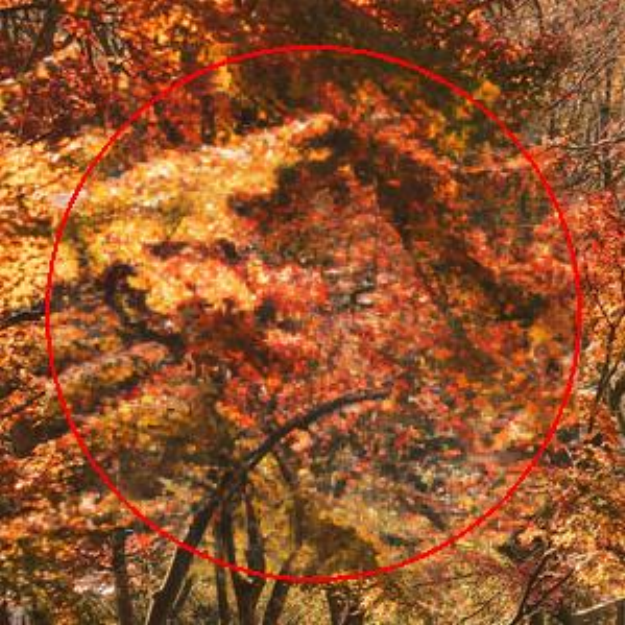}} &
        
        \\

        \multicolumn{1}{l}{\rotatebox[origin=c]{90}{\shortstack[l]{\scriptsize Adobe \\ \scriptsize Firefly \cite{adobefirefly}}}} &
        \noindent\parbox[c]{0.083\textwidth}{\includegraphics[width=0.083\textwidth]{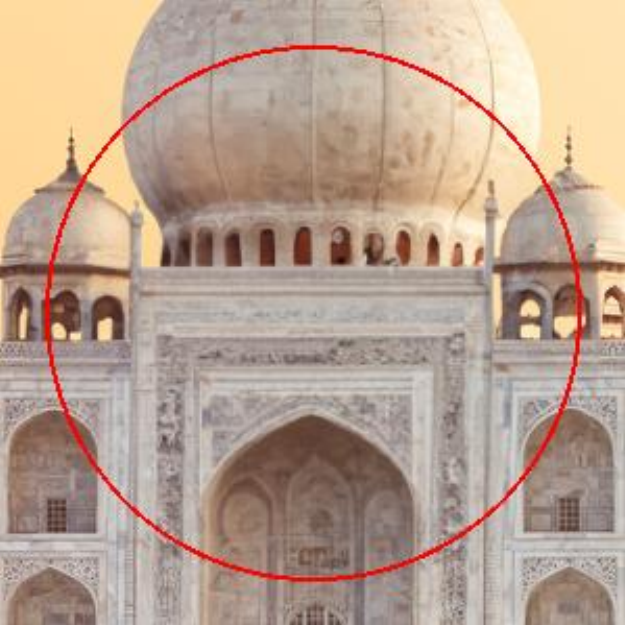}} & 
        \noindent\parbox[c]{0.083\textwidth}{\includegraphics[width=0.083\textwidth]{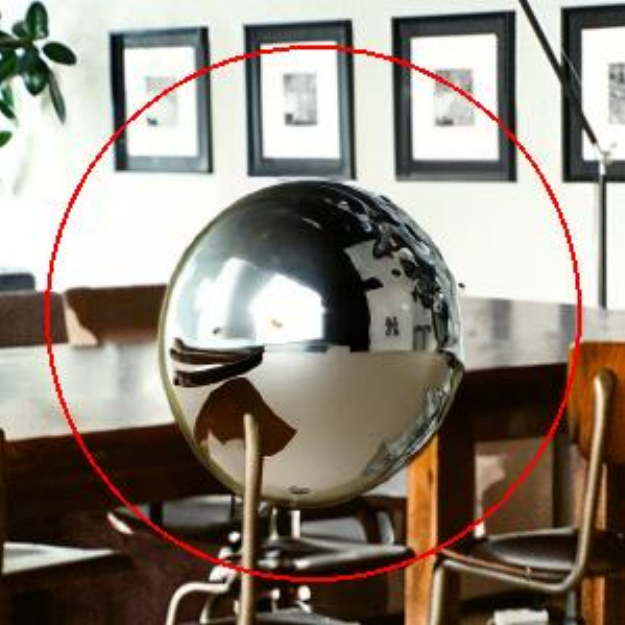}} & 
        \noindent\parbox[c]{0.083\textwidth}{\includegraphics[width=0.083\textwidth]{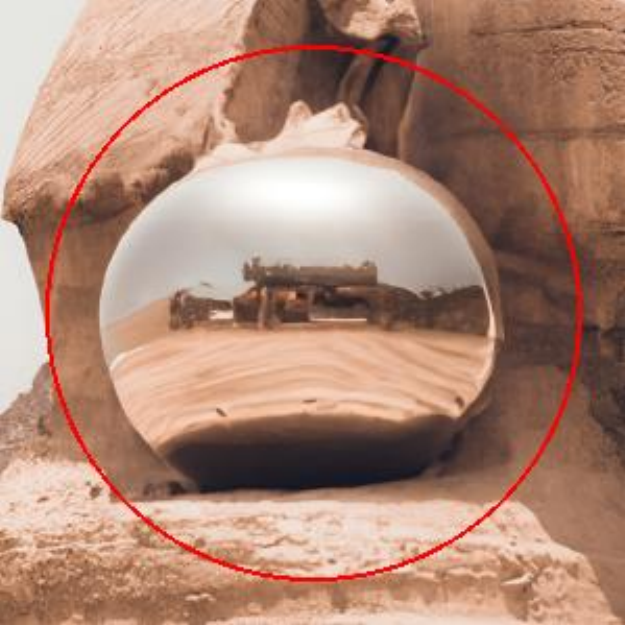}} & 
        \noindent\parbox[c]{0.083\textwidth}{\includegraphics[width=0.083\textwidth]{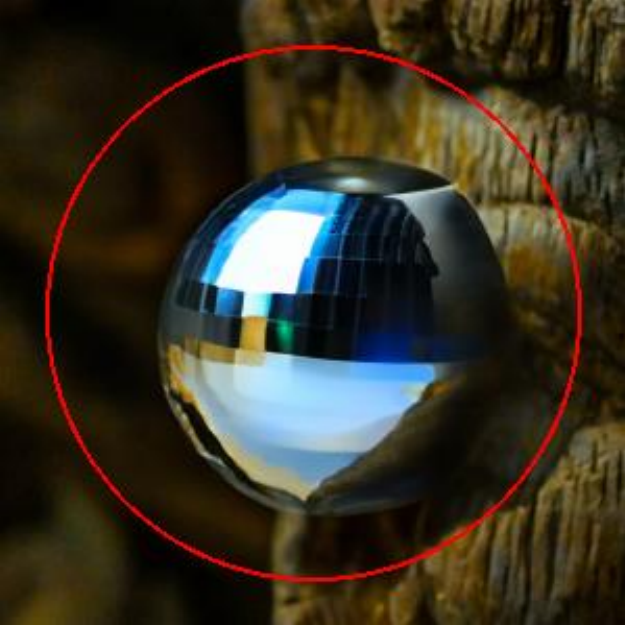}} & 
        \noindent\parbox[c]{0.083\textwidth}{\includegraphics[width=0.083\textwidth]{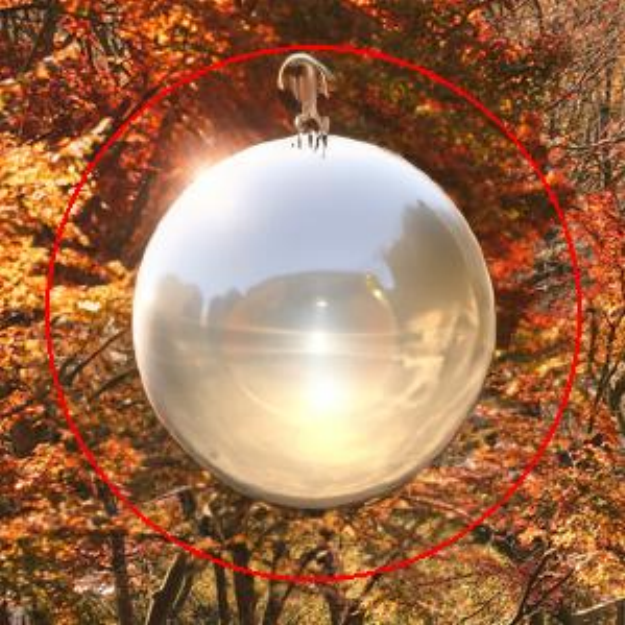}} &

        \\

        \multicolumn{1}{l}{\rotatebox[origin=c]{90}{\shortstack[l]{\scriptsize SDXL \cite{podell2023sdxl}}}} &
        \noindent\parbox[c]{0.083\textwidth}{\includegraphics[width=0.083\textwidth]{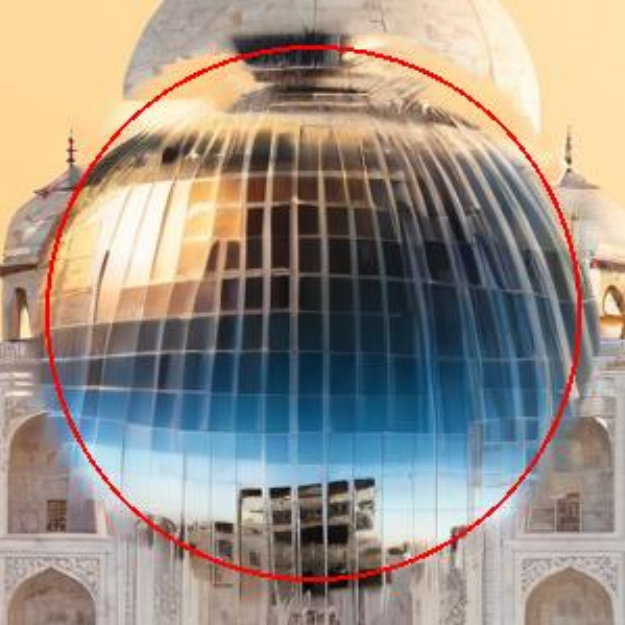}} & 
        \noindent\parbox[c]{0.083\textwidth}{\includegraphics[width=0.083\textwidth]{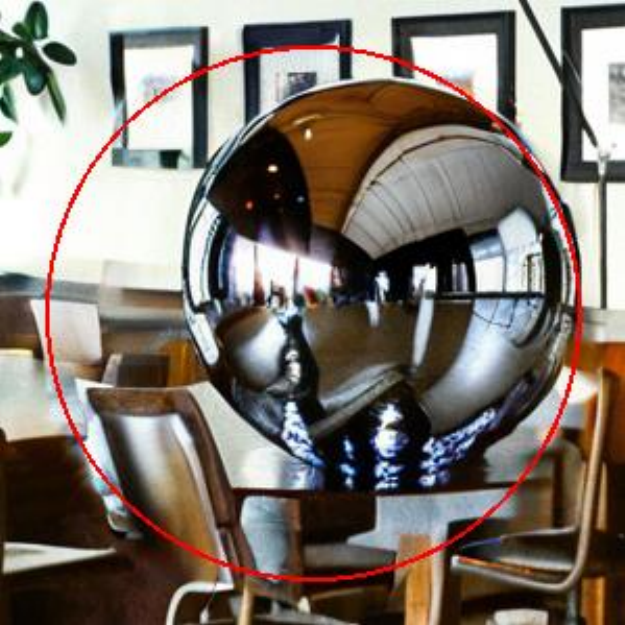}} & 
        \noindent\parbox[c]{0.083\textwidth}{\includegraphics[width=0.083\textwidth]{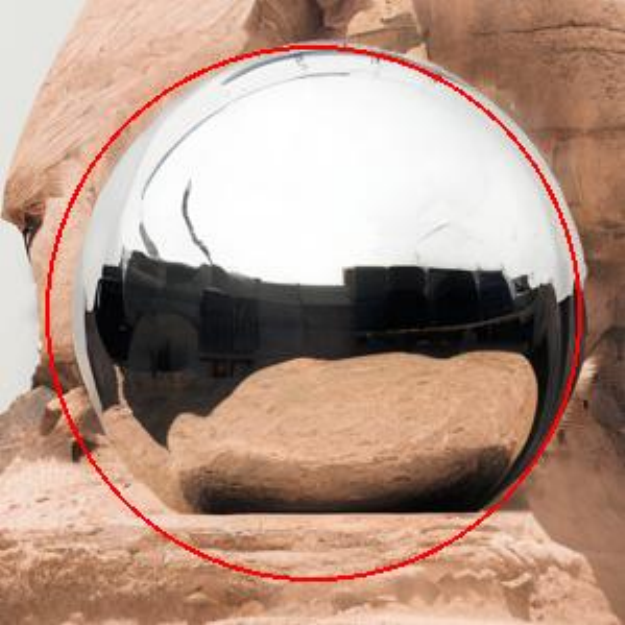}} & 
        \noindent\parbox[c]{0.083\textwidth}{\includegraphics[width=0.083\textwidth]{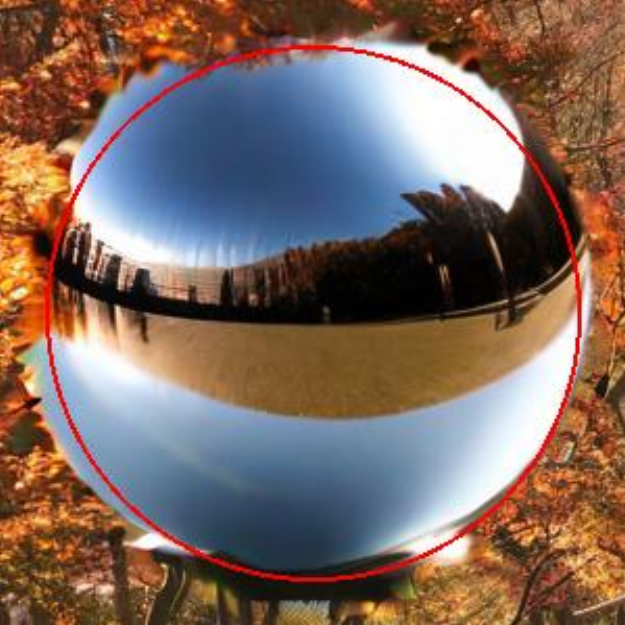}} & 
        \noindent\parbox[c]{0.083\textwidth}{\includegraphics[width=0.083\textwidth]{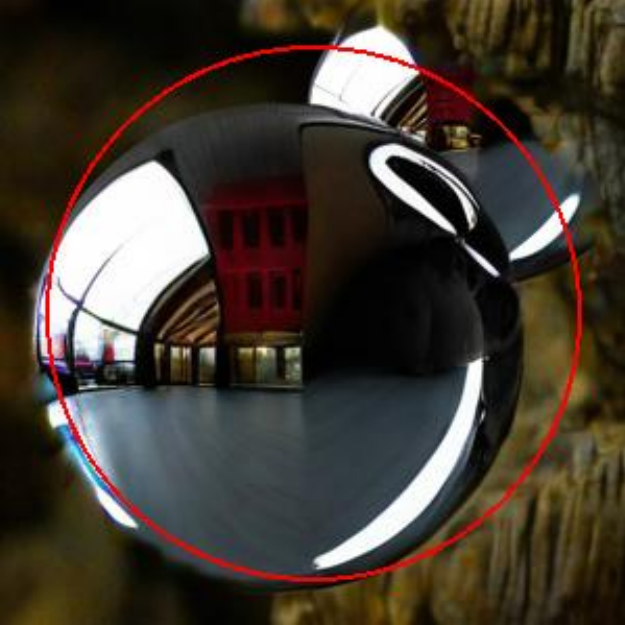}} &
        
        \\ \hline

        \multicolumn{1}{l}{\rotatebox[origin=c]{90}{\shortstack[l]{\scriptsize \textbf{DiffusionLight} \\ \scriptsize \hspace{0.35cm} \textbf{(Ours)} }}} &
        \noindent\parbox[c]{0.083\textwidth}{\includegraphics[width=0.083\textwidth]{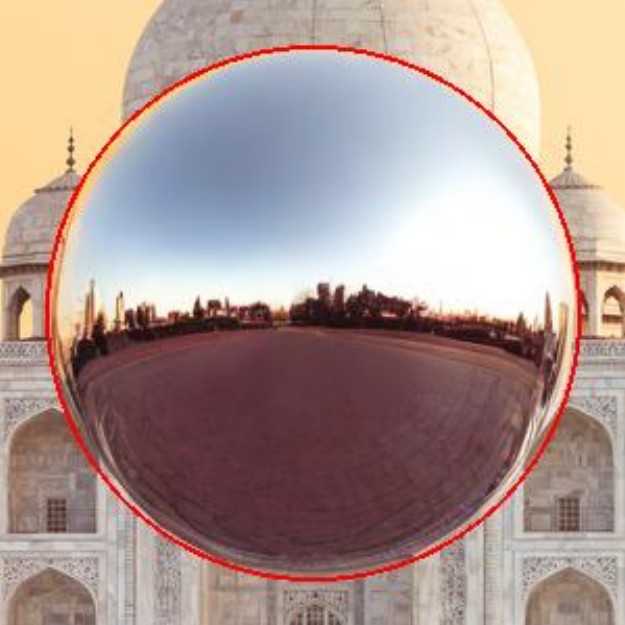}} & 
        \noindent\parbox[c]{0.083\textwidth}{\includegraphics[width=0.083\textwidth]{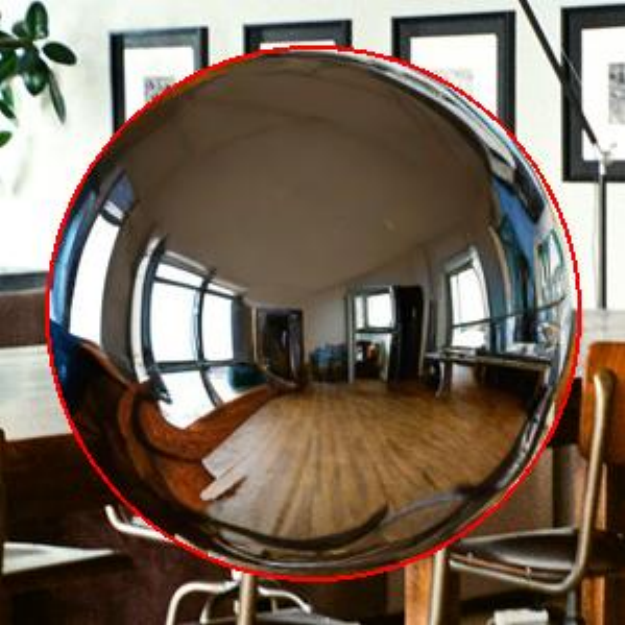}} & 
        \noindent\parbox[c]{0.083\textwidth}{\includegraphics[width=0.083\textwidth]{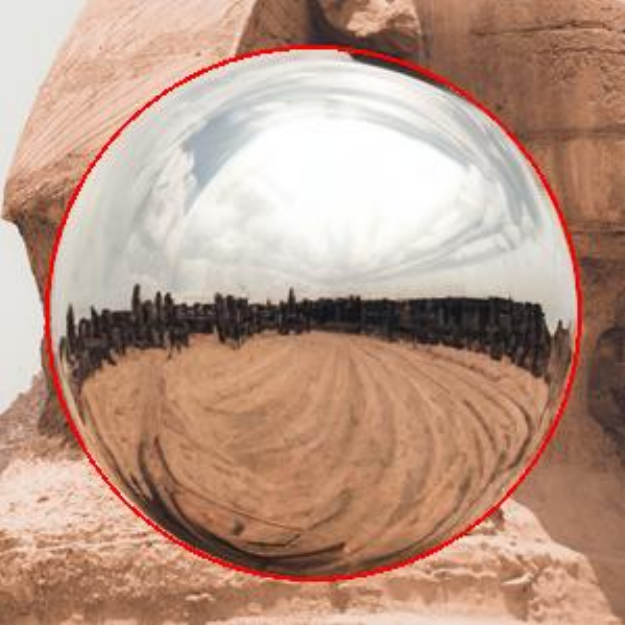}} & 
        \noindent\parbox[c]{0.083\textwidth}{\includegraphics[width=0.083\textwidth]{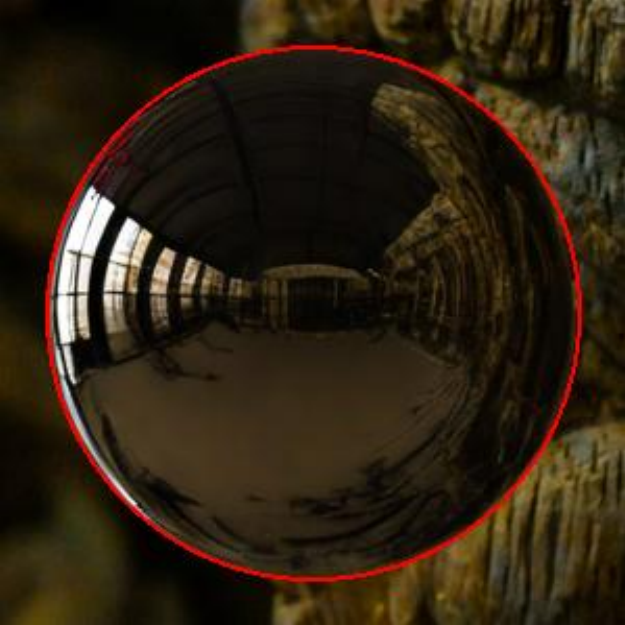}} & 
        \noindent\parbox[c]{0.083\textwidth}{\includegraphics[width=0.083\textwidth]{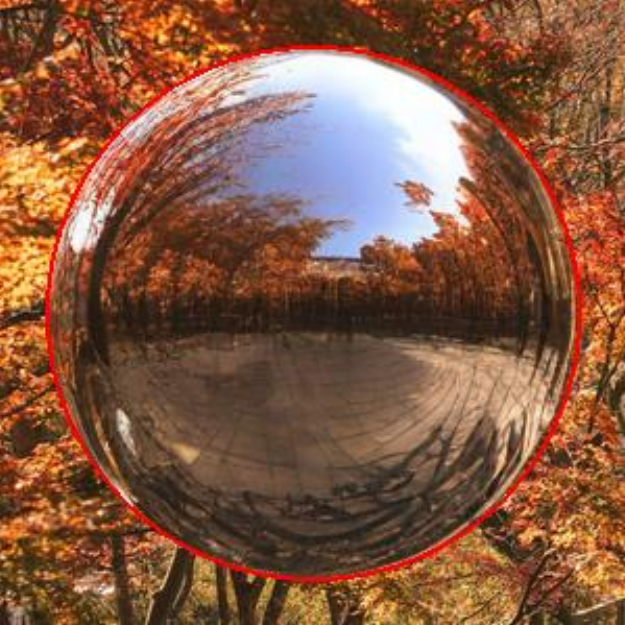}} & 
        
        \\
        
    \end{tabu}
    \caption{Chrome ball inpainting results from various methods. The red circle indicates the inpainted region, and we show a zoomed-in view of the blue crop. These baselines tend to produce distorted balls with undesirable textures or completely fail to produce a ball and instead just reconstruct the original content. Our method addresses all these issues and precisely follows the inpainting mask.}
    \vspace{-0.6cm}
    \label{fig:inpaint_sota}
\end{figure}

Inspired by traditional approaches in computer graphics \cite{devebechistory}, the conference version of this work, DiffusionLight \cite{phongthawee2024diffusionlight}, reframes single-view lighting estimation as the task of inpainting a chrome ball and leverages the image prior in a pre-trained text-to-image diffusion model (Stable Diffusion XL \cite{podell2023sdxl}) to solve it. While inpainting a chrome ball into an image may seem straightforward, state-of-the-art diffusion models with inpainting capabilities \cite{avrahami2023blendedlatent, avrahami2022blendeddiffusion, yang2023paint, ye2023ip-adapter} often struggle to consistently generate good chrome balls that convincingly reflect environmental lighting (see Figure \ref{fig:inpaint_sota}). Another key limitation is that these models were trained on LDR images and cannot readily produce HDR chrome balls.

In this paper, we explore how to efficiently leverage pretrained diffusion models for HDR chrome ball inpainting to address the challenges. Specifically: (1) we enhance the consistency and quality of inpainted LDR balls by conditioning the diffusion denoising process on depth information to precisely control the ball's position, and on a good lighting prior to mitigate spurious artifacts; (2) we synthesize HDR chrome balls by generating LDR images at multiple exposures using our \emph{Exposure LoRA}, then merging them in the luminance space to avoid ghosting artifacts; and (3) we accelerate inference by training a \emph{Turbo LoRA} to speed up the inpainting process and designing a strategy to best apply these LoRAs across different sampling timesteps.


First, to consistently produce high-quality chrome balls, our DiffusionLight~\cite{phongthawee2024diffusionlight} involves three key ideas. (1) We make inserting a ball reliable by precisely controlling its location using depth map conditioning via ControlNet \cite{zhang2023adding}. (2) To better mimic genuine chrome ball appearance, we fine-tune the model using LoRA \cite{hu2021lora} on a small set of synthetically generated chrome balls. (3) We start the diffusion sampling process from a good initial noise map, and we propose an iterative inpainting algorithm to find one. The last idea is based on surprising findings we discovered about diffusion model behavior and chrome ball appearance.


Second, to generate HDR chrome balls, we generate and combine multiple LDR chrome balls with varying exposure values, similar to exposure bracketing. A simple baseline that requires no additional training uses two text prompts: one for generating a standard chrome ball, and the other with ``black dark'' added to the text prompt. However, this approach offers limited control over exposure and often produces inconsistent results. Instead, we further fine-tune the LoRA used earlier to also map continuous interpolations of the two prompts to a target ball image with varying, known exposures (referred to as \textit{Exposure LoRA} hereinafter).
This enables specifying the exposure values of the generated balls at test time. While this fine-tuning requires a small number of panoramas for training, the core task of producing reflective chrome balls still relies on the model initial’s capability, which remains generalizable to a broad range of scenes.

Third, to accelerate the \emph{iterative inpainting algorithm} used to identify good initial noise maps in DiffusionLight \cite{phongthawee2024diffusionlight}, our extension, DiffusionLight-Turbo, directly infers these noise maps by training a new \emph{Turbo LoRA} to generate average chrome balls similar to those produced by DiffusionLight.
At test time, a straightforward way to use the Turbo LoRA is to follow DiffusionLight’s original pipeline: (1) inpaint an average chrome ball to find a good initial noise map, and (2) perform a second diffusion denoising pass with DiffusionLight’s Exposure LoRA. While this method yields high-quality results, it requires two full denoising processes. 
To reduce inference time further, we introduce a simple yet effective \emph{LoRA swapping technique}, which merges the two denoising passes into a single pass---halving the runtime on top of a 30x speed up. 
During denoising, we apply Turbo LoRA in the early stages (high-noise regions) to initialize the process with a good initial noise map, then switch to Exposure LoRA in the later stages (low-noise regions) to produce high-frequency details. This method is motivated by our observation that low-frequency components---such as lighting---are primarily established during the early stages. Experimental results show that this technique achieves quality comparable to SDEdit-based generation (see Table~\ref{tab:aba_diffusionlight_turbo} and Figures~\ref{fig:main_qualitative_wild} - \ref{fig:qualitative_benchmark}).

We evaluate our method against StyleLight \cite{wang2022stylelight}, EverLight \cite{dastjerdi2023everlight}, Weber et al. \cite{weber2022editableindoor}, and EMLight \cite{zhan2021emlight} on the standard benchmarks: Laval Indoor \cite{garder2017lavelindoor} and Poly Haven \cite{polyhaven} datasets.
DiffusionLight is competitive with StyleLight and achieves better performance on two out of three metrics across both datasets, while ranking second and third, when tested using a protocol in EverLight~\cite{dastjerdi2023everlight} on Laval Indoor. Note that the baselines were directly trained on the datasets; some specifically tailored to indoor scenes.
While DiffusionLight-Turbo yields slightly lower scores than DiffusionLight, it remains competitive with StyleLight while requiring only 1.67\% of DiffusionLight’s runtime.
When applied to more challenging, in-the-wild images beyond the benchmarks, our methods still produce convincing results, whereas the baselines often fail.

To summarize, the conference version of this paper~\cite{phongthawee2024diffusionlight} made three contributions:

\begin{itemize}
    \item A novel light estimation technique (Section~\ref{sec:diffusionlight}) that generalizes across diverse scenes based on a simple idea of inpainting a chrome ball using a pre-trained diffusion model, Stable Diffusion XL.
    \item An iterative inpainting algorithm (Section~\ref{sec:iterative}) that enhances quality and consistency based on our discovered relationship between the initial noise map and chrome ball appearances.
    \item A LoRA fine-tuning technique (Section~\ref{sec:hdr}) for generating chrome balls at arbitrary exposure levels to produce HDR chrome balls.
\end{itemize}

This paper extends the conference version with the following additions:
\begin{itemize}

    
    \item \emph{Turbo LoRA} (Section \ref{sec:turbo_lora}): Accelerates the iterative inpainting algorithm by training a \textit{Turbo LoRA} to directly generate average chrome balls at inference, eliminating the need to average multiple samples. 

    \item \emph{LoRA swapping} (Section \ref{sec:lora_swapping}): Further reduces runtime by half by using Turbo LoRA in the early steps and switching to Exposure LoRA in the remaining steps, with minimal quality loss compared to DiffusionLight's SDEdit.
    

    \item \emph{Additional experiments and ablation studies:}  We evaluate the design choices and runtime-quality trade-off of DiffusionLight-Turbo (Sections~\ref{sec:abla_diffusion_turbo} and ~\ref{sec:abla_running_time_tradeoff}), showing it achieves a 60× speedup over the original DiffusionLight with comparable output quality in a single denoising loop.
    
    \item \emph{Discussion of recent works:} We include additional discussion of works published after the conference version.
    
    
    
\end{itemize}

\section{Related Work}
\label{sec:related}



\tabulinesep=0.1pt



\subsubsection{Lighting estimation} 
Methods for estimating lighting in an image can be divided into two categories. The first category requires the presence of a certain \emph{probing} object in the captured image, such as a mirror ball \cite{Debevec1998}, a known 3D object \cite{weber2018learning, lombardi2015reflectance}, a reflective object \cite{park2020seeing, yu2023accidental, georgoulis2017around}, or naturally occurring elements, such as human faces \cite{calian2018faces, yi2018faces} or eyes \cite{nishino2004eyes}. However, most photos lack such probes and cannot be used with these methods. The second category does not require a probing object and can handle indoor \cite{gardner2019deepparam, garon2019fastspatialvary, weber2022editableindoor, zhan2021emlight}, outdoor scenes \cite{hold2017outdoor, hold2019laveloutdoor}, or both \cite{dastjerdi2023everlight, legendre2019deeplight}. Our method belongs to this category, which will be the focus of our review.

To estimate lighting without a specific probing object, earlier work requires initial lighting annotations \cite{karsch2011rendering}, analyzes physical cues from cast shadows or the sky \cite{lalonde2012estimating}, or jointly optimizes the scene geometry and lighting models that best reproduce an input image \cite{karsch2014automatic, karsch2011rendering}. However, the lighting models in these methods are highly limited---for example, discretized sun locations \cite{lalonde2012estimating}, directional light shafts \cite{karsch2011rendering}, or a best-fit illumination map retrieved from a dataset \cite{karsch2014automatic}. Later work utilizes more complex lighting models such as sky radiance models \cite{hosek2012analytic, hold2017outdoor}, spherical harmonic lighting \cite{garon2019fastspatialvary}, and 3D spherical or Gaussian functions \cite{gardner2019deepparam}. However, modern techniques have shifted toward predicting 360\degree HDR environment maps, which are essential for tasks such as virtual insertion of highly reflective objects.

A common strategy among these techniques involves regressing an LDR input with a limited field of view into an HDR map with neural networks. Gardner et al. \cite{garder2017lavelindoor} use a CNN classifier to locate lights using a large dataset of LDR panoramas and then fine-tune the CNN to predict HDR maps using a smaller HDR dataset. 
Hold-Geoffroy et al. \cite{hold2019laveloutdoor} first train an autoencoder for outdoor sky maps, then use another network to encode and decode an input image. Weber et al. \cite{weber2022editableindoor} predict LDR maps along with parametric light sources \cite{gardner2019deepparam}, also with a CNN. Somanath et al. \cite{somanath2021envmapnet} introduce two loss functions based on randomly masked L1 and cluster classification to enhance estimation accuracy. Zhan et al. \cite{zhan2021emlight, zhan2022gmlight} propose a two-step process involving a spherical light distribution predictor and an HDR map predictor. 
Xu et al. \cite{xu2022renderingaware} built upon Zhan et al. \cite{zhan2022gmlight} by leveraging self-attention mechanisms to better model global lighting distributions. SALENet \cite{zhao2024salenet} further enhances this framework by incorporating contrastive learning to improve lighting representation.
DSGLight \cite{bai2023deepgraph} incorporates depth information into the spherical Gaussian representation and explores Graph Convolutional Networks for predicting HDR maps.
These methods are typically tested in either indoor or outdoor settings due to the specific lighting models or the lack of sufficiently large and diverse HDR datasets.

To solve both indoor and outdoor settings, LeGendre et al. \cite{legendre2019deeplight} collected a new dataset of natural scenes captured using a mobile device with three reflective probing balls. Their method, DeepLight, regresses HDR lighting from an input image using a loss function that minimizes the difference between ground truth and rendered balls under the predicted lighting. Dastjerdi et al. \cite{dastjerdi2023everlight}'s EverLight combines multiple indoor and outdoor datasets and predicts an editable lighting representation, which then conditions a GAN to generate a full HDR map. Some GAN-based techniques focus on outpainting an input image to a 360$^{\circ}$ panorama \cite{akimoto2019360outpainting2stategan, dastjerdi2022immersegan, akimoto2022outpainting}, but they perform poorly in light estimation because the prediction is in LDR~\cite{dastjerdi2023everlight, wang2023360}. In contrast, StyleLight by Wang et al. \cite{wang2022stylelight} trains a two-branch StyleGAN network to predict LDR and HDR maps from noise and, at test time, uses GAN inversion to predict an HDR map from an input image. 

Despite many attempts to combine indoor and outdoor panoramas, the resulting datasets remain limited in size and diversity.
In contrast, we leverage diffusion models trained on billions of images, leading to better generalization.
Our work aims to develop a general framework applicable to a broader range of in-the-wild images. This stands in contrast to similar diffusion-based approaches that specifically target indoor scenes, such as CleAR \cite{zhao2024clear}, which fine-tunes ControlNets \cite{zhang2023adding} using prompts specifically crafted for indoor settings.

Since the publication of our conference version, two recent works have been proposed for indoor lighting estimation: IllumiDiff~\cite{shen2025illumidiff} and Zhao et al.~\cite{zhao2025asg}. Both leverage the idea of using Spherical Gaussians (SG) and/or Anisotropic Spherical Gaussians (ASG) as light representations, serving as conditional inputs to generative models for predicting environment maps. IllumiDiff \cite{shen2025illumidiff} employs a CNN to regress SG and ASG parameters, which are then used to guide a diffusion model to generate LDR panoramas, subsequently converted to HDR using an additional network. Zhao et al. \cite{zhao2025asg} use a transformer to estimate ASG parameters, which condition a GAN for HDR environment map generation. 
However, direct quantitative comparison with these methods is not straightforward: neither paper provides public code or access to their exact test split. Zhao et al. \cite{zhao2025asg} report better performance than DiffusionLight on their own split, except for si-RMSE, making definitive comparison difficult. 
Moreover, both methods are heavily reliant on the training set and do not generalize well to in-the-wild scenes, as noted in their paper, further highlighting the unique focus of our work.

\vspace{0.5em}
\subsubsection{Diffusion-based image inpainting} Our method relies on text-conditioned diffusion models to synthesize chrome balls. Numerous prior studies have addressed object insertion in images.
For instance, given a text prompt describing an arbitrary object and an inpainting mask, Blended Diffusion \cite{avrahami2022blendeddiffusion, avrahami2023blendedlatent} inserts the object using guided sampling \cite{dhariwal2021diffusion} based on the cosine distance between the CLIP embedding of the inpainted region and the prompt.
Alternatively, Paint-by-Example \cite{yang2023paint} directly conditions diffusion models using example images and their CLIP embeddings.
ControlNet \cite{zhang2023adding} and IP-Adapter \cite{ye2023ip-adapter} enable conditioning pre-trained diffusion models for image generation using both image and text prompts.
Commercial products like DALL·E2 \cite{dalle2} and Adobe Firefly \cite{adobefirefly} also offer similar inpainting capabilities with text prompts.

Unfortunately, these methods often struggle to consistently produce chrome balls or produce ones that do not convincingly reflect the environmental lighting (see Figure \ref{fig:inpaint_sota}).



\vspace{0.5em}
\subsubsection{Personalized text-to-image diffusion models} Our work leverages a pre-trained diffusion model to consistently generate known objects, a task closely related to \emph{personalized} image generation.
For this task, models are fine-tuned using single or a few reference object images while preserving their unique appearance. 
DreamBooth \cite{ruiz2022dreambooth} uses a special token during fine-tuning to represent the object while maintaining the pre-trained distribution.
Gal et al. \cite{gal2022image} introduce a learned word in the embedding space for representing referenced objects, and Voynov et al. \cite{voynov2023P+} adopt separate word embeddings per network layer. 
Additionally, several studies \cite{hu2021lora, han2023svdiff, oft2023} explore techniques to simplify the fine-tuning of diffusion models, with LoRA \cite{hu2021lora} being a popular choice that enforces low-rank weight changes.

These methods can be adapted to our task by providing chrome ball images for fine-tuning. In fact, a portion of our pipeline can be viewed as a variant of DreamBooth with LoRA, albeit without the prior preservation loss.


\section{Problem Setup}
\label{sec:method}


\begin{figure*}[!t]
    \centering
    \includegraphics[width=1.0\textwidth]{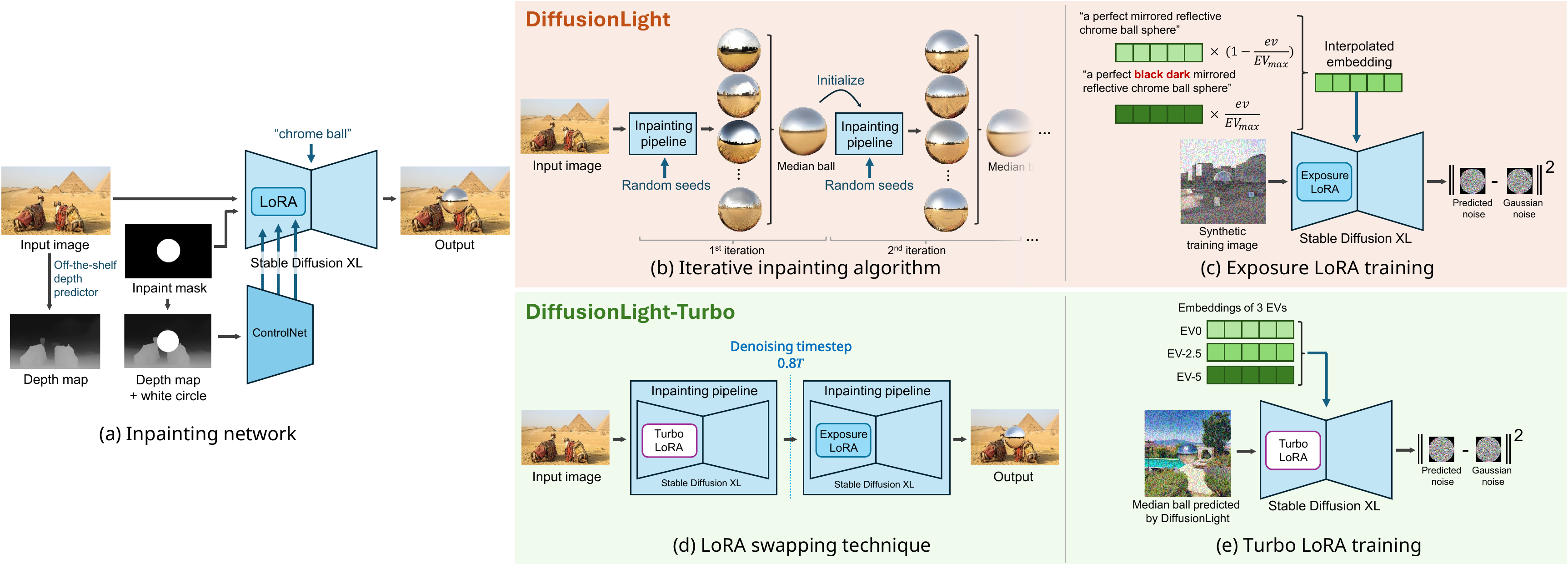}
    \caption{(a) We use Stable Diffusion XL \cite{podell2023sdxl} with depth-conditioned ControlNet \cite{zhang2023adding} to inpaint a chrome ball. (b) Our iterative inpainting algorithm enhances generation quality and consistency by constraining the initial noise through sample averaging. (c) We train Exposure LoRA to produce multiple LDR chrome balls at varying exposures for HDR merging. (d) We find a good initial noise map by denoising with Turbo LoRA until \textit{t=0.8T}, then switching to Exposure LoRA. (e) We train Turbo LoRA to predict the median of the predictions made by DiffusionLight. 
    } 

    \label{fig:pipeline_overview}
\end{figure*}



\tabulinesep=0.5pt
\begin{figure}
    \centering

    \begin{tabu} to \textwidth {
        @{}
        c@{}
        c@{\hspace{1.0pt}}
        c@{\hspace{0.5pt}}
        c@{\hspace{0.5pt}}
        c@{\hspace{0.5pt}}
        c@{\hspace{0.5pt}}
        c@{}
    }
        
        \multicolumn{1}{l}{\rotatebox[origin=c]{90}{\shortstack[l]{\scriptsize Input}}} &
        \noindent\parbox[c]{0.083\textwidth}{\includegraphics[width=0.083\textwidth]{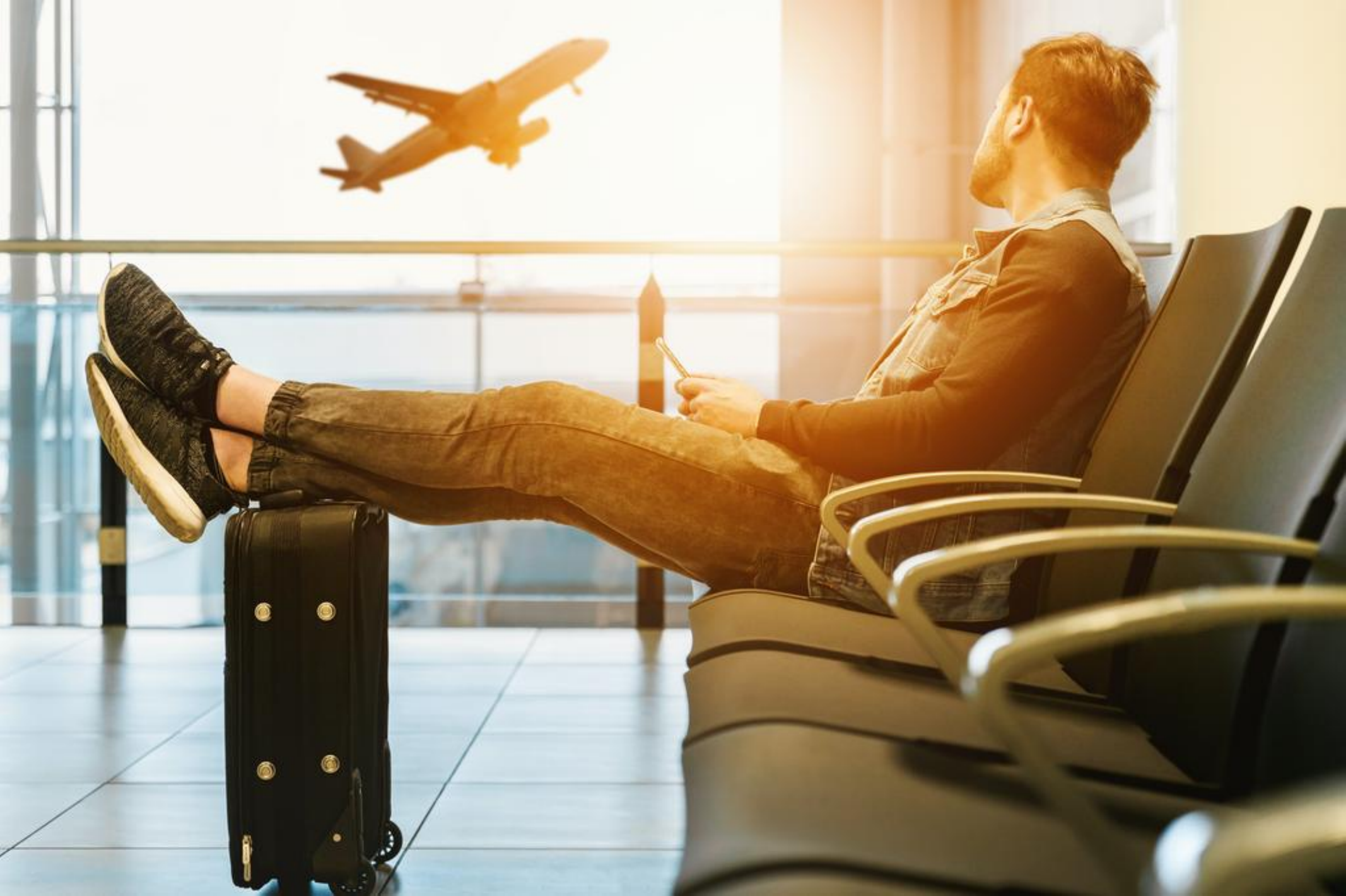}} & 
        \noindent\parbox[c]{0.083\textwidth}{\includegraphics[width=0.083\textwidth]{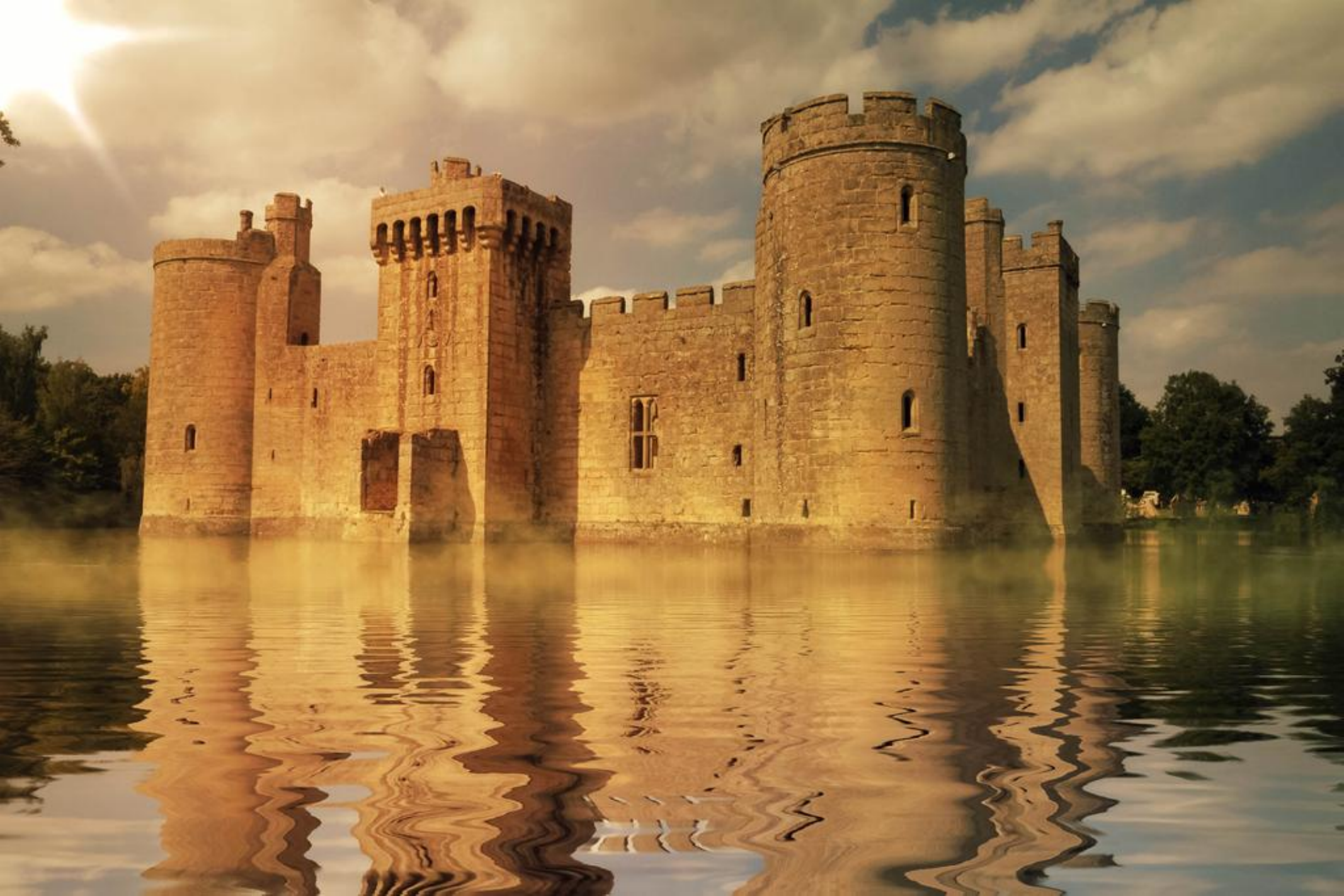}} & 
        \noindent\parbox[c]{0.083\textwidth}{\includegraphics[width=0.083\textwidth]{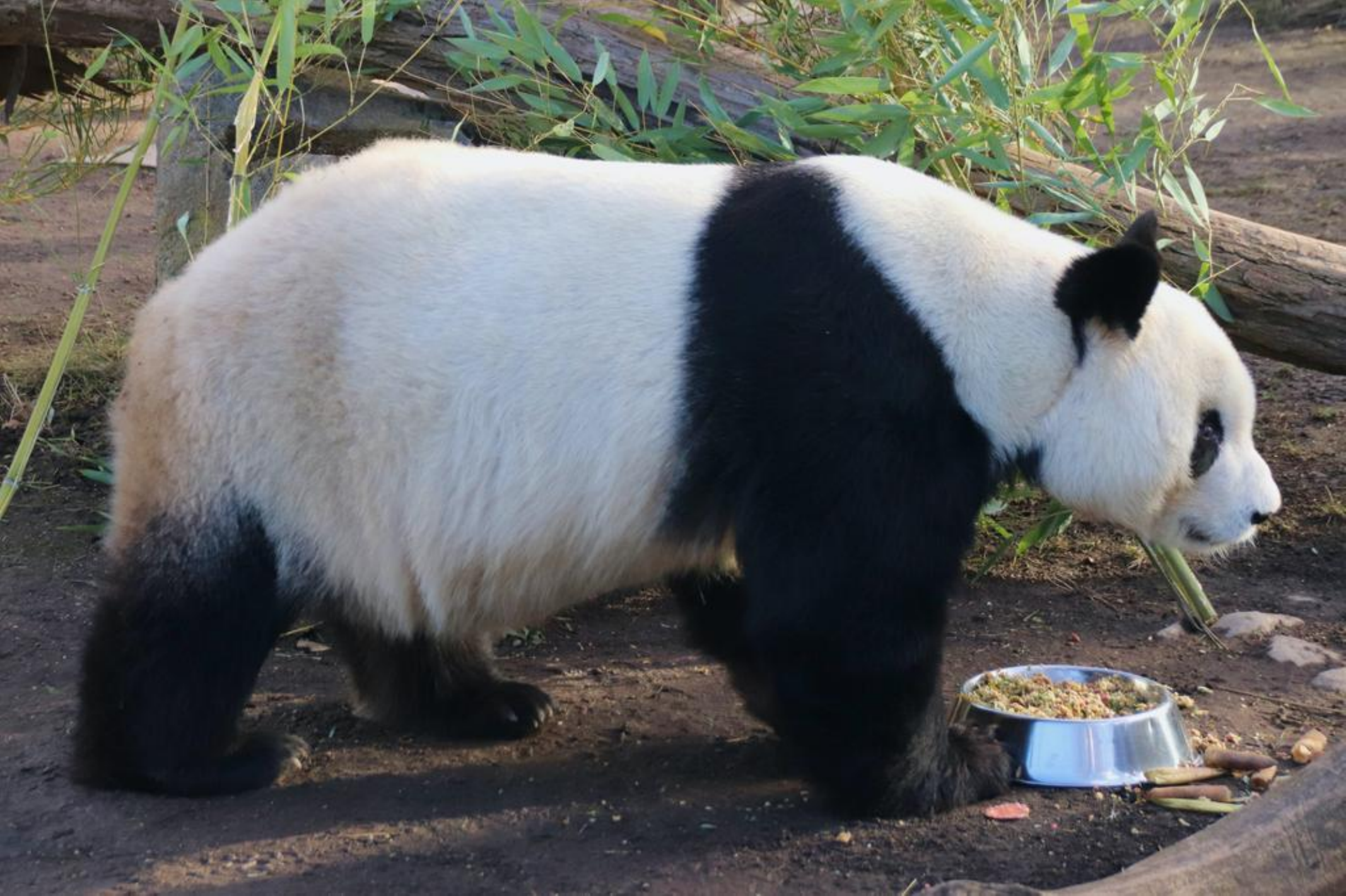}} & 
        \noindent\parbox[c]{0.083\textwidth}{\includegraphics[width=0.083\textwidth]{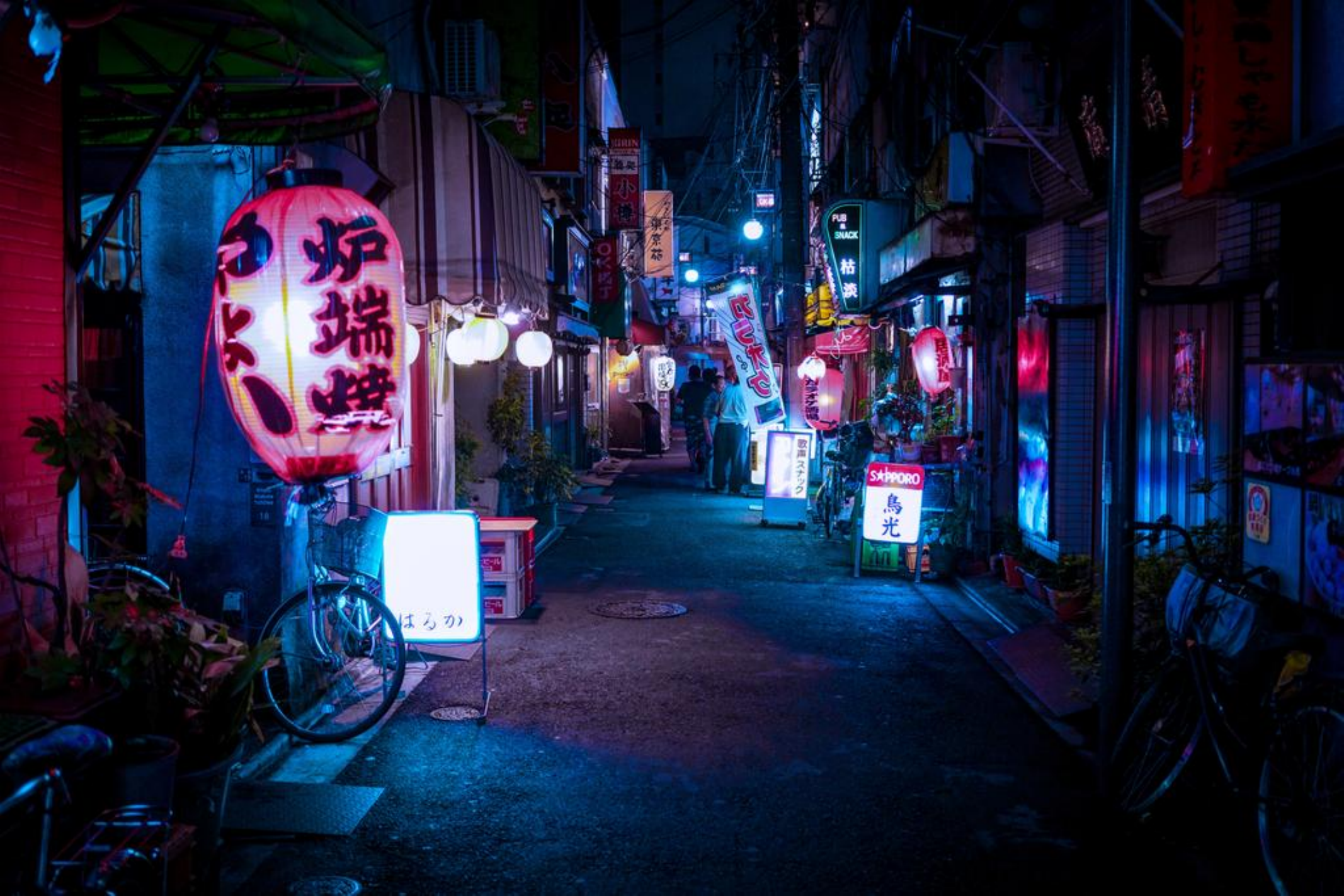}} & 
        \noindent\parbox[c]{0.083\textwidth}{\includegraphics[width=0.083\textwidth]{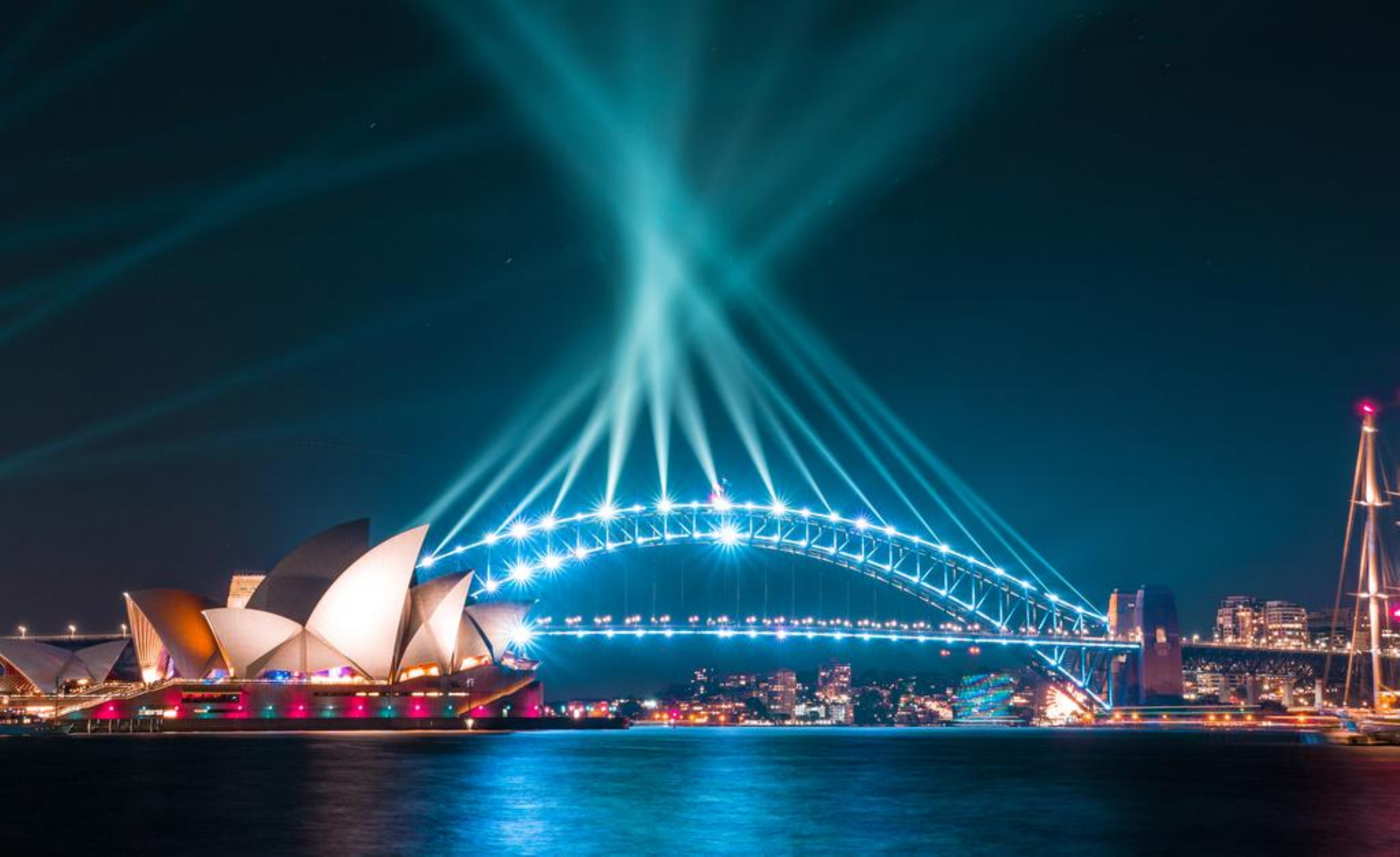}} & 

        \\

        \multicolumn{1}{l}{\rotatebox[origin=c]{90}{\shortstack[l]{\scriptsize Bad noise \#1}}} &
        \noindent\parbox[c]{0.083\textwidth}{\includegraphics[width=0.083\textwidth]{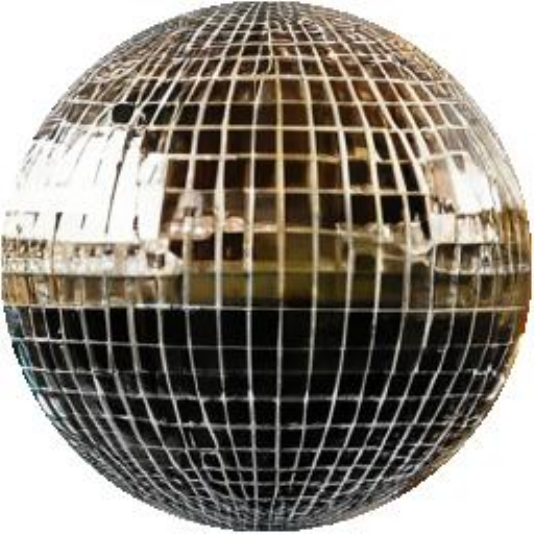}} & 
        \noindent\parbox[c]{0.083\textwidth}{\includegraphics[width=0.083\textwidth]{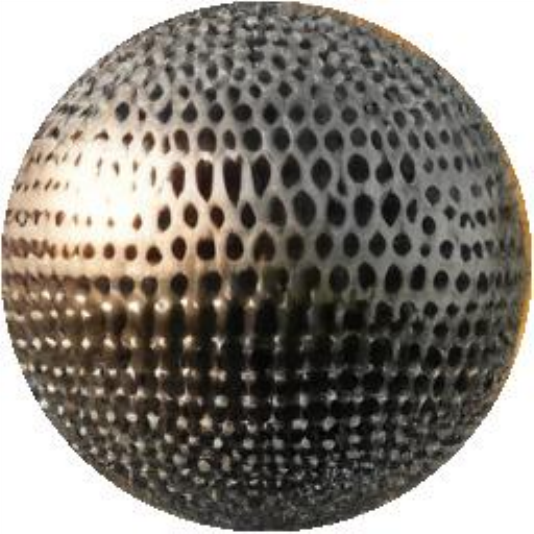}} & 
        \noindent\parbox[c]{0.083\textwidth}{\includegraphics[width=0.083\textwidth]{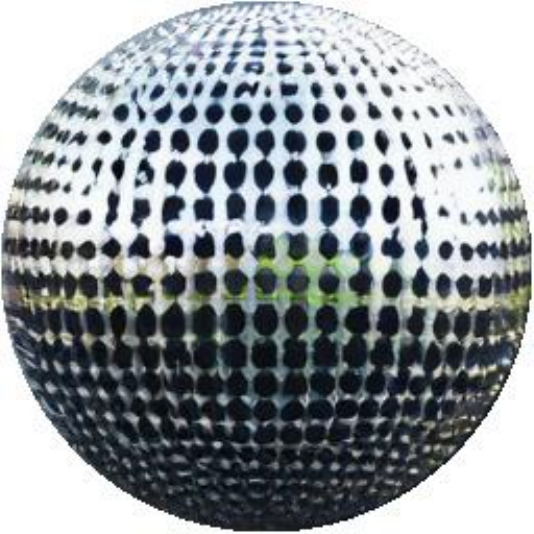}} & 
        \noindent\parbox[c]{0.083\textwidth}{\includegraphics[width=0.083\textwidth]{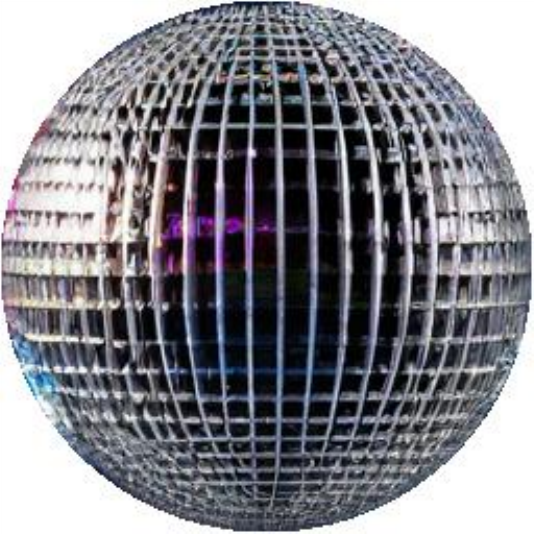}} & 
        \noindent\parbox[c]{0.083\textwidth}{\includegraphics[width=0.083\textwidth]{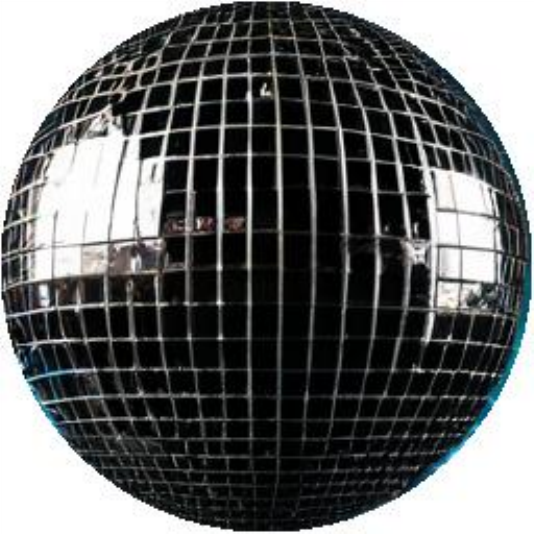}} & 
        
        \\

        \multicolumn{1}{l}{\rotatebox[origin=c]{90}{\shortstack[l]{\scriptsize Bad noise \#2}}} &
        \noindent\parbox[c]{0.083\textwidth}{\includegraphics[width=0.083\textwidth]{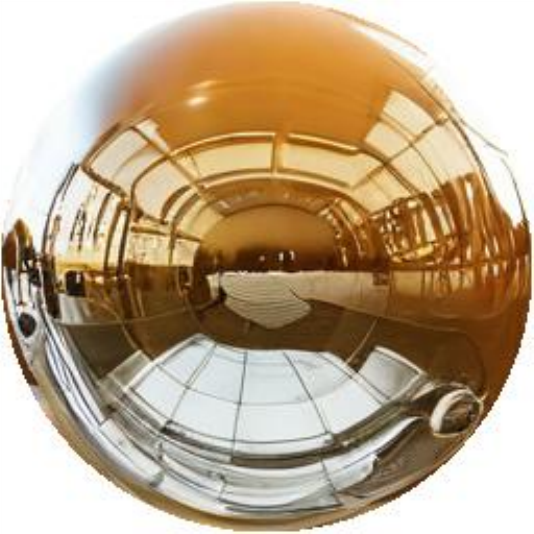}} & 
        \noindent\parbox[c]{0.083\textwidth}{\includegraphics[width=0.083\textwidth]{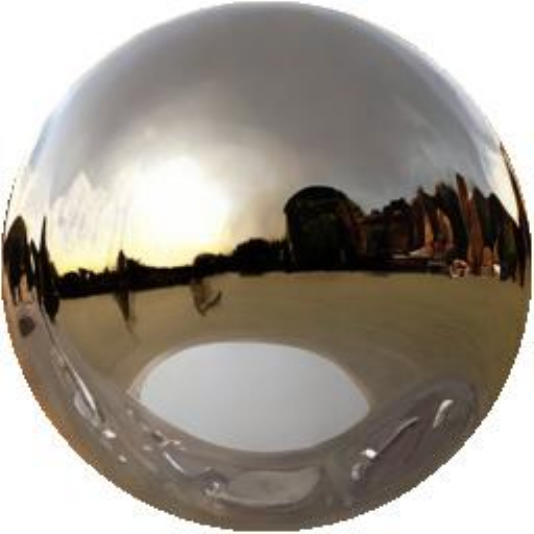}} & 
        \noindent\parbox[c]{0.083\textwidth}{\includegraphics[width=0.083\textwidth]{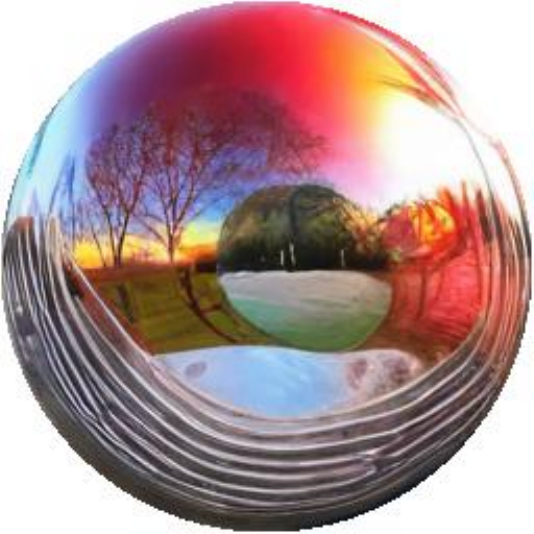}} & 
        \noindent\parbox[c]{0.083\textwidth}{\includegraphics[width=0.083\textwidth]{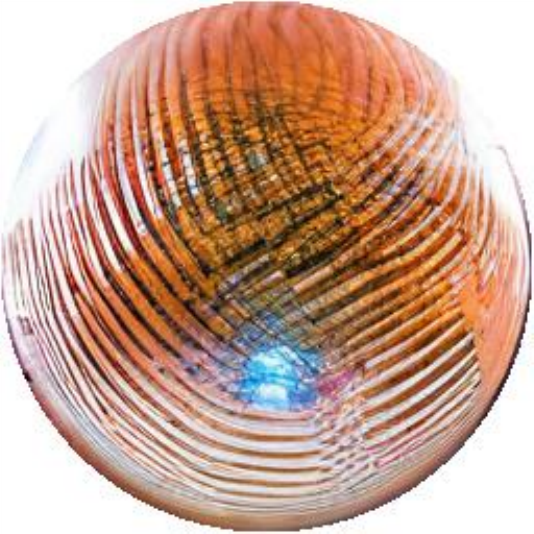}} & 
        \noindent\parbox[c]{0.083\textwidth}{\includegraphics[width=0.083\textwidth]{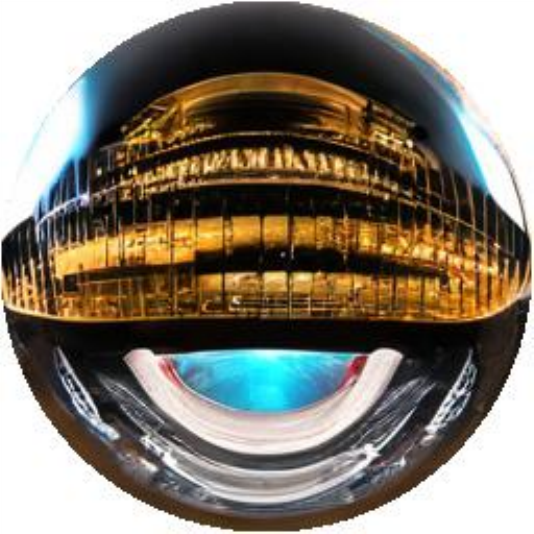}} & 
        
        \\
        
        \arrayrulecolor{red}\cline{2-6}

        \multicolumn{1}{l}{\rotatebox[origin=c]{90}{\shortstack[l]{\scriptsize \textbf{DiffusionLight}}}} &
        \multicolumn{1}{|c}{
        \noindent\parbox[c]{0.083\textwidth}{\includegraphics[width=0.083\textwidth]{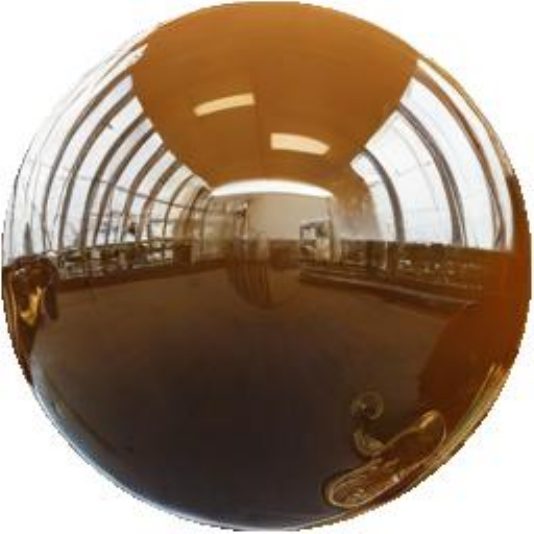}}} & 
        \noindent\parbox[c]{0.083\textwidth}{\includegraphics[width=0.083\textwidth]{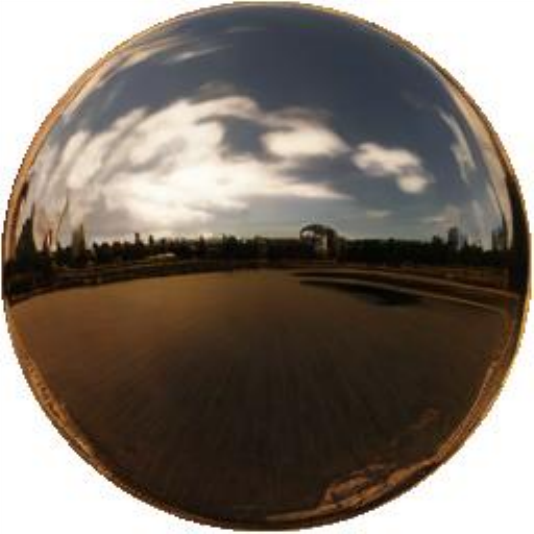}} & 
        \noindent\parbox[c]{0.083\textwidth}{\includegraphics[width=0.083\textwidth]{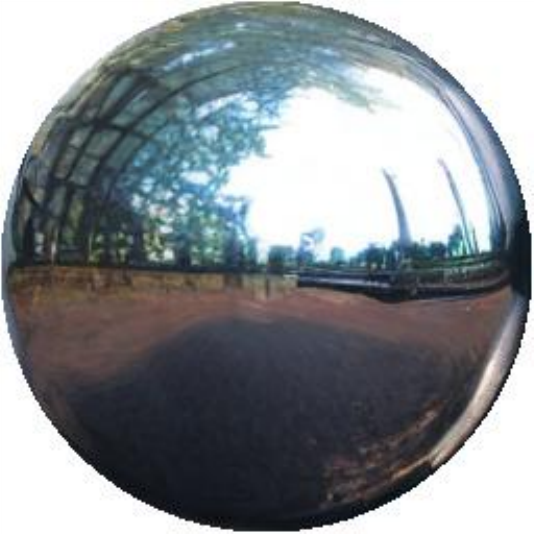}} & 
        \noindent\parbox[c]{0.083\textwidth}{\includegraphics[width=0.083\textwidth]{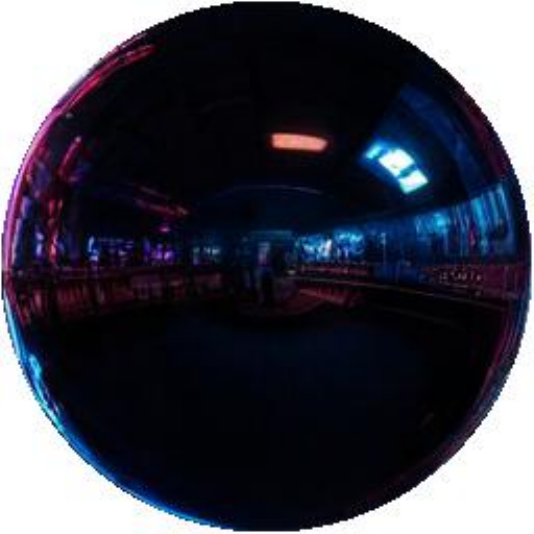}} & 
        \multicolumn{1}{c|}{\noindent\parbox[c]{0.083\textwidth}{\includegraphics[width=0.083\textwidth]{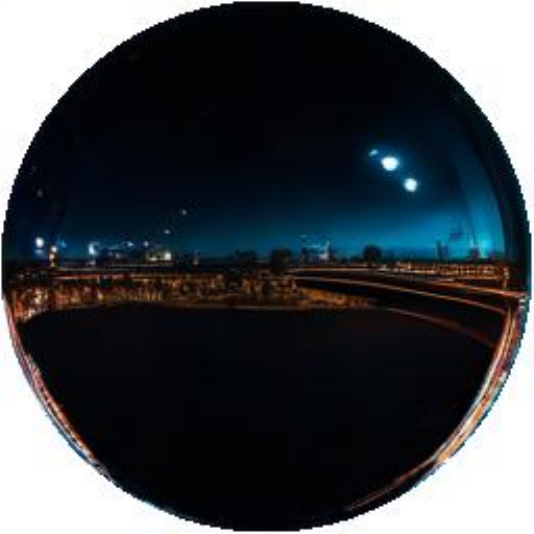}}} & 

        \\ \arrayrulecolor{red}\cline{2-6}
        \arrayrulecolor{black}
        
    \end{tabu}

    \caption{Observations: the initial noise map encodes some semantic patterns. We found certain noise maps to consistently produce a disco ball or bad patterns \emph{across} input images. In contrast, our method consistently produces clean chrome balls.
    }
    \label{fig:noise_appearance_relationship}
\end{figure}

Given a standard LDR input image, our goal is to estimate the scene's lighting as an HDR environment map. 
DiffusionLight is based on inserting a chrome ball into the image using a diffusion model, then unwraps it to an environment map. We tackle two key challenges: (1) how to consistently generate chrome balls and (2) how to generate HDR chrome balls using an LDR diffusion model.

To explain our method in detail, we first cover the background and standard notations of diffusion models. We then introduce DiffusionLight from our conference paper~\cite{phongthawee2024diffusionlight}, followed by DiffusionLight-Turbo (Section~\ref{sec:fast_diffusionlight}), an accelerated version proposed in this paper.

\section{Preliminaries} \label{sec:prelim}

\subsubsection{Diffusion models} \cite{ho2020denoising} form a family of generative models that can transform a noise distribution (Gaussian distribution) to a target data distribution $p_{\mathrm{data}}$ by learning to revert a Gaussian diffusion process. Following the convention in \cite{song2020denoising}, it is represented by a discrete-time stochastic process $\{ \vect{x}_t \}_{t=0}^T$ where $\vect{x}_0 \sim p_{\mathrm{data}}$, and $\mathbf{x}_t \sim \mathcal{N}(\vect{x}_{t-1}; \sqrt{\alpha_t / \alpha_{t-1} } \vect{x}_{t-1}, (1 - \alpha_t / \alpha_{t-1})\vect{I})$. The decreasing scalar function $\alpha_t$, with constraints that $\alpha_0 = 1$ and $\alpha_T \approx 0$, controls the noise level through time. It can be shown that
\begin{equation} \label{eq:add_noise}
    \vect{x}_t = \sqrt{\alpha_t} \vect{x}_0 + \sqrt{1 - \alpha_t} \vect{\epsilon}, ~ \text{where} ~ \vect{\epsilon} \sim \mathcal{N}(\vect{0}, \vect{I}).
\end{equation}

A diffusion model is a neural network $\epsilon_{\theta}$ trained to predict from $\vect{x}_t$ the noise $\epsilon$ that was used to generate it according to \eqref{eq:add_noise}. The commonly employed, simplified training loss is
\begin{equation} \label{eq:diffusion_objective}
    \mathcal{L} = \mathbb{E}_{\vect{x}_0, t, \vect{\epsilon}}  \left[ \lVert \epsilon_{\vect{\theta}} (\sqrt{\alpha_t}\vect{x}_0 + \sqrt{1 - \alpha_t}\epsilon, t, \vect{C}) - \vect{\epsilon} \rVert_2^2 \right],
\end{equation}
where $\vect{C}$ denotes conditioning signals such as text. The trained network can then be used to convert a Gaussian noise sample to a data sample through several sampling methods \cite{ho2020denoising, song2020denoising, zhao2023unipc}. In this paper, we use Stable Diffusion XL \cite{podell2023sdxl}, which operates on latent codes of images according to a variational autoencoder (VAE). As such, we use $\vect{x}_t$ to denote images and $\vect{z}_t$ to denote latent codes.

\vspace{0.5em}
\subsubsection{Low-Rank Adaptation (LoRA)} \cite{hu2021lora} is a parameter-efficient fine-tuning technique designed to adapt large pre-trained models with minimal computational overhead. Instead of updating the full weight matrix $\vect{W}_i \in \mathbb{R}^{m \times n}$ for each layer during fine-tuning---a process that can be memory-intensive and prone to overfitting---LoRA introduces a trainable low-rank decomposition that captures task-specific adjustments. Specifically, LoRA optimizes a residual matrix $\Delta \vect{W}_i = \vect{A}_i\vect{B}_i$, where $\vect{A}_i \in \mathbb{R}^{m \times d}$ and $\vect{B}_i \in \mathbb{R}^{d \times n}$ with $d \ll m,n$, significantly reducing the number of learnable parameters. This residual is then scaled by a factor $\alpha$, known as the "LoRA scale," and added to the original frozen weights to produce the final adapted weights: $\vect{W}_i' = \vect{W}_i + \alpha \Delta \vect{W}_i$. 


\section{DiffusionLight} \label{sec:diffusionlight}
DiffusionLight consists of three main components: (A) a base inpainting pipeline, (B) an iterative inpainting technique for producing consistently high-quality chrome balls, and (C) an Exposure LoRA for generating multiple LDR images for exposure bracketing.

\subsection{Base inpainting pipeline} 
The inpainting pipeline is based on Stable Diffusion XL \cite{podell2023sdxl} with depth-conditioned ControlNet \cite{zhang2023adding}, as shown in Figure \ref{fig:pipeline_overview}.
We first predict a depth map from the input image using an off-the-shelf model \cite{ranftl2020towards, ranftl2021vision}. Then, we paint a circle in the center of the depth map using the depth value closest to the camera (visualized as white) and paint the same white circle to the inpaint mask. 
We feed them along with the input image, the prompt ``\emph{a perfect mirrored reflective chrome ball sphere},'' and the negative prompt ``\emph{matte, diffuse, flat, dull}'' to the diffusion model.

While this simple pipeline could reliably \emph{insert} a chrome ball instead of merely recovering the masked out content, the chrome balls often contain undesirable patterns and fail to convincingly reflect environment lighting (Figure \ref{fig:noise_appearance_relationship}).

\subsection{Iterative inpainting} \label{sec:iterative} Iterative inpainting is based on two key observations.
First, the same initial noise map leads to the generation of semantically similar inpainted areas \emph{regardless} of the input image. 
For instance, there exists a ``disco'' noise map that consistently produces a disco ball \emph{across} different input images, while a good noise map almost always produces a reflective chrome ball (Figure \ref{fig:noise_appearance_relationship}). When searching for images of a ``chrome ball'' on the Internet, not all results match the specific reflective chrome ball we want. So, the encoded semantics found within the noise map are perhaps understandable, as text prompts alone cannot encode all visual appearances of ``chrome ball.'' Here, adding ``disco'' to the negative prompt may fix this specific instance, but many other failure modes are not as easy to describe and exclude via text prompts. 

Second, averaging multiple ball samples can approximate the overall lighting reasonably well, but the average ball itself is too blurry and not as useful.

Based on these insights, we propose a simple algorithm to automatically locate a good noise neighborhood by sample averaging.
Specifically, we first inpaint $N$ chrome balls onto an input image using different random seeds. Then, we calculate a pixel-wise \textit{median} ball and composite it back to the input image. Let us denote this composited image by $\vect{B}_1$. To sample a better chrome ball, we apply SDEdit \cite{meng2022sdedit} by adding noise to $\vect{B}_1$ to simulate the diffusion process up to time step $t$: $\vect{B}_1' = \sqrt{\alpha_t} \vect{B}_1 + \sqrt{1 - \alpha_t}\vect{\epsilon}$, where $\vect{\epsilon} \sim \mathcal{N}(\vect{0}, \vect{I})$, $t < T$, and $T$ denotes the maximum timestep. Then, we continue denoising $\vect{B}_1'$ starting from $t$ to 0. In our implementation, $t = 0.8T$.

\begin{figure}
    \centering

    \begin{tabu} to \textwidth {
        @{}
        c@{\hspace{1pt}}
        c@{\hspace{1pt}}
        @{\hspace{8pt}}
        c@{\hspace{1pt}}
        c@{\hspace{1pt}}
        c@{}
    } 
        



        
        
        

        \noindent\parbox[c]{0.08\textwidth}{\includegraphics[width=0.08\textwidth]{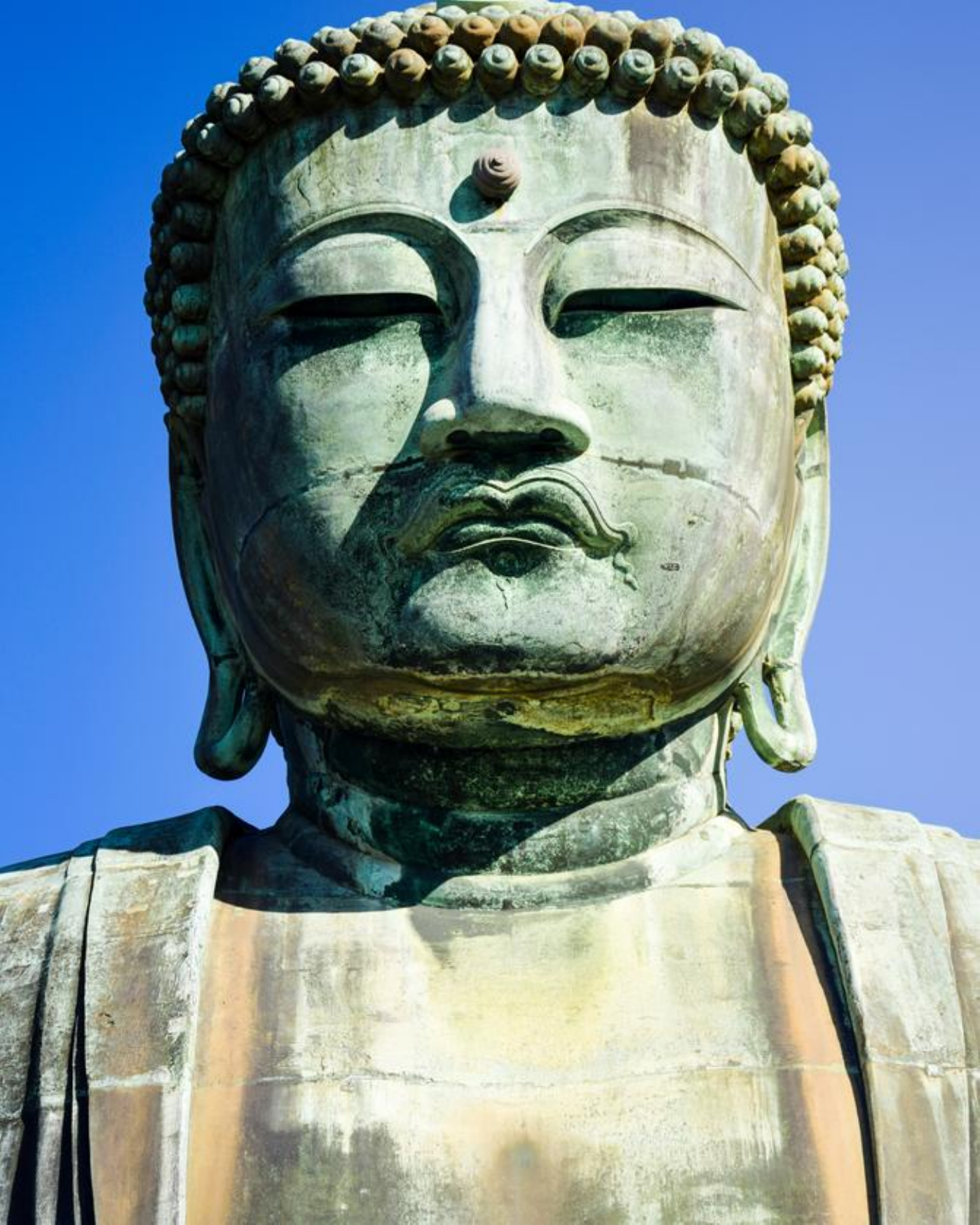}} 
        &
        \noindent\parbox[c]{0.14\textwidth}{\includegraphics[width=0.14\textwidth]{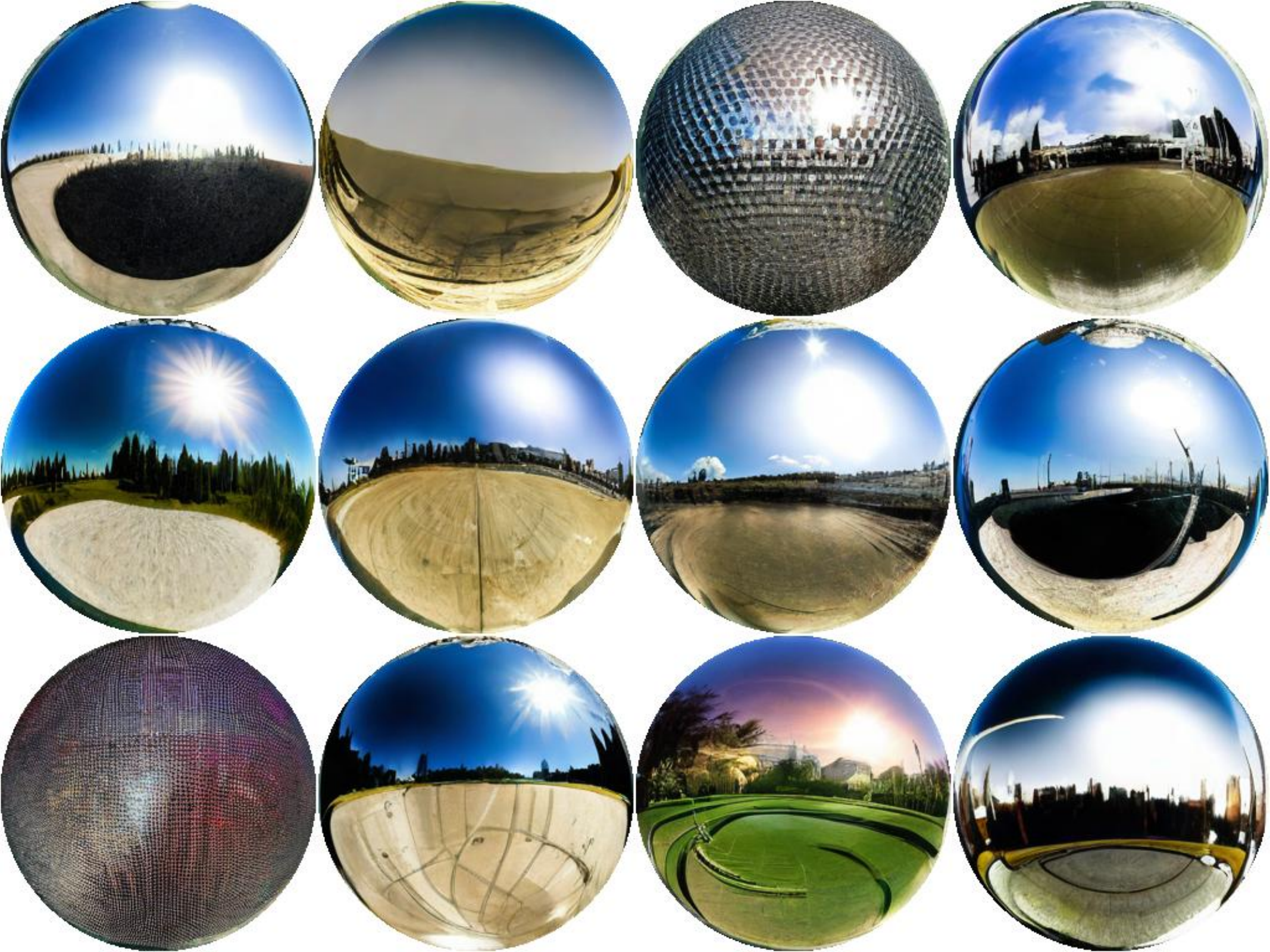}} 

        
        &
        \noindent\parbox[c]{0.08\textwidth}{\shortstack{\tiny Median ball \\ \includegraphics[width=0.08\textwidth]{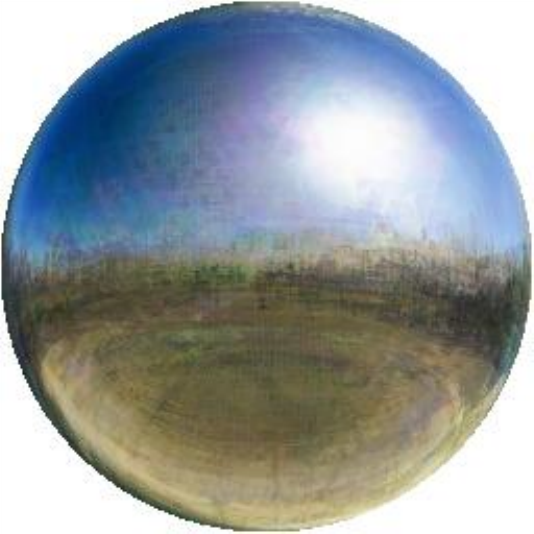}}}
        
        &
        \noindent\parbox[c]{0.14\textwidth}{\includegraphics[width=0.14\textwidth]{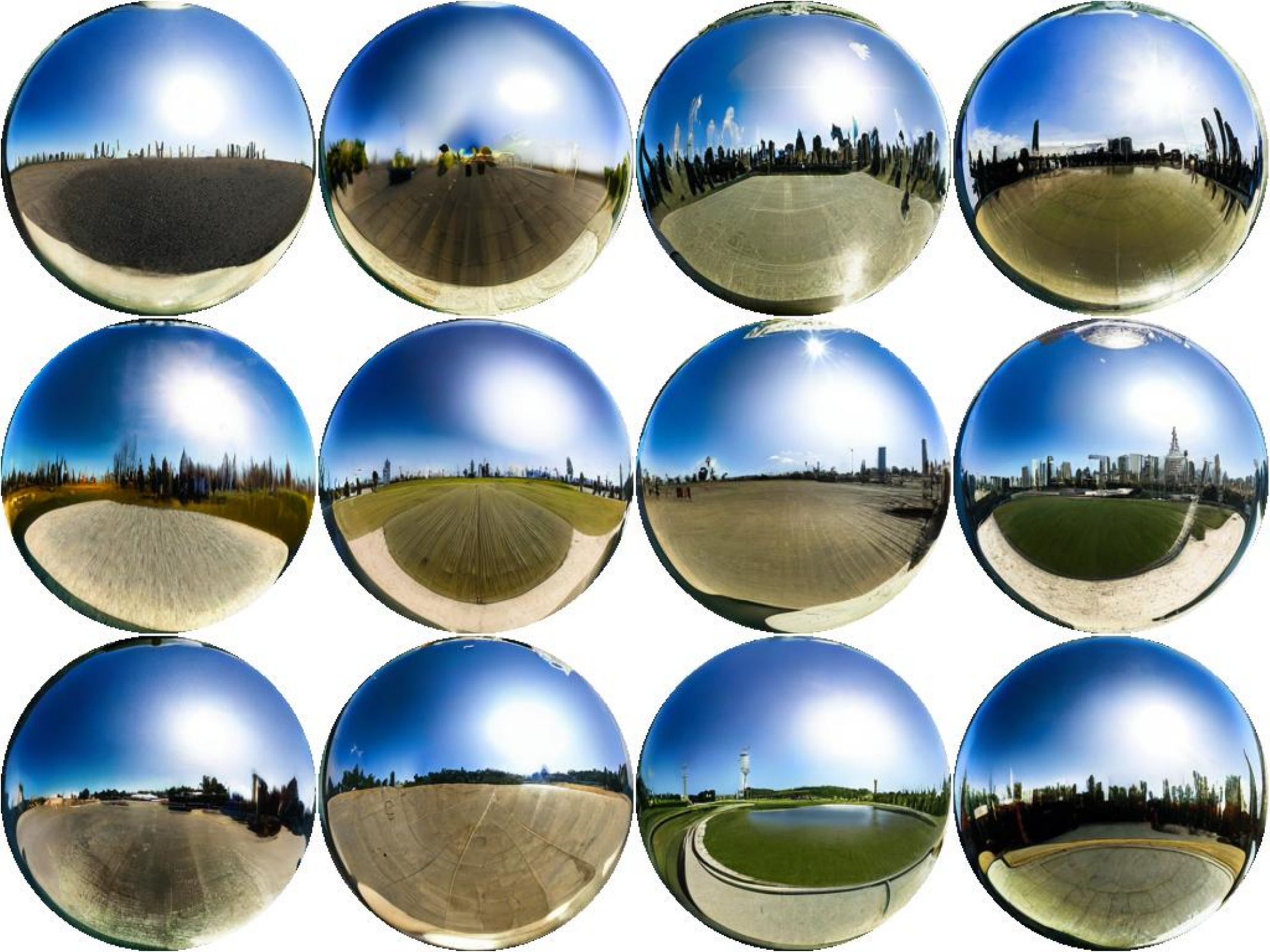}} 
        
    \end{tabu}
    \caption{Balls before (left) and after (right) iterative inpainting.}
    \label{fig:compare_median_distribution}
\end{figure}

We can repeat the process by using SDEdit to generate another set of $N$ chrome balls from the output and recompute another median chrome ball $\vect{B}_2$.
This repetition not only minimizes artifacts and spurious patterns from incorrect ball types but also enhances the consistency of light estimation 
(see Figure \ref{fig:compare_median_distribution} and Appendix \ref{appendix:aba_median}).
In our implementation, we perform $k = 2$ iterations of median computation, with the final output generated by applying one last SDEdit to $\vect{B}_2$.

\subsection{Predicting HDR chrome balls}\label{sec:hdr_algo} 
\begin{figure}
    \centering

    \begin{tabu} to \textwidth {
        @{}
        c@{\hspace{1pt}}
        c@{\hspace{1pt}}
        c@{\hspace{1pt}}
        c@{\hspace{1pt}}
        c@{\hspace{1pt}}
        c@{\hspace{1pt}}
        c@{}
    }
        
        \noindent\parbox[c]{0.090\textwidth}{\includegraphics[width=0.090\textwidth]{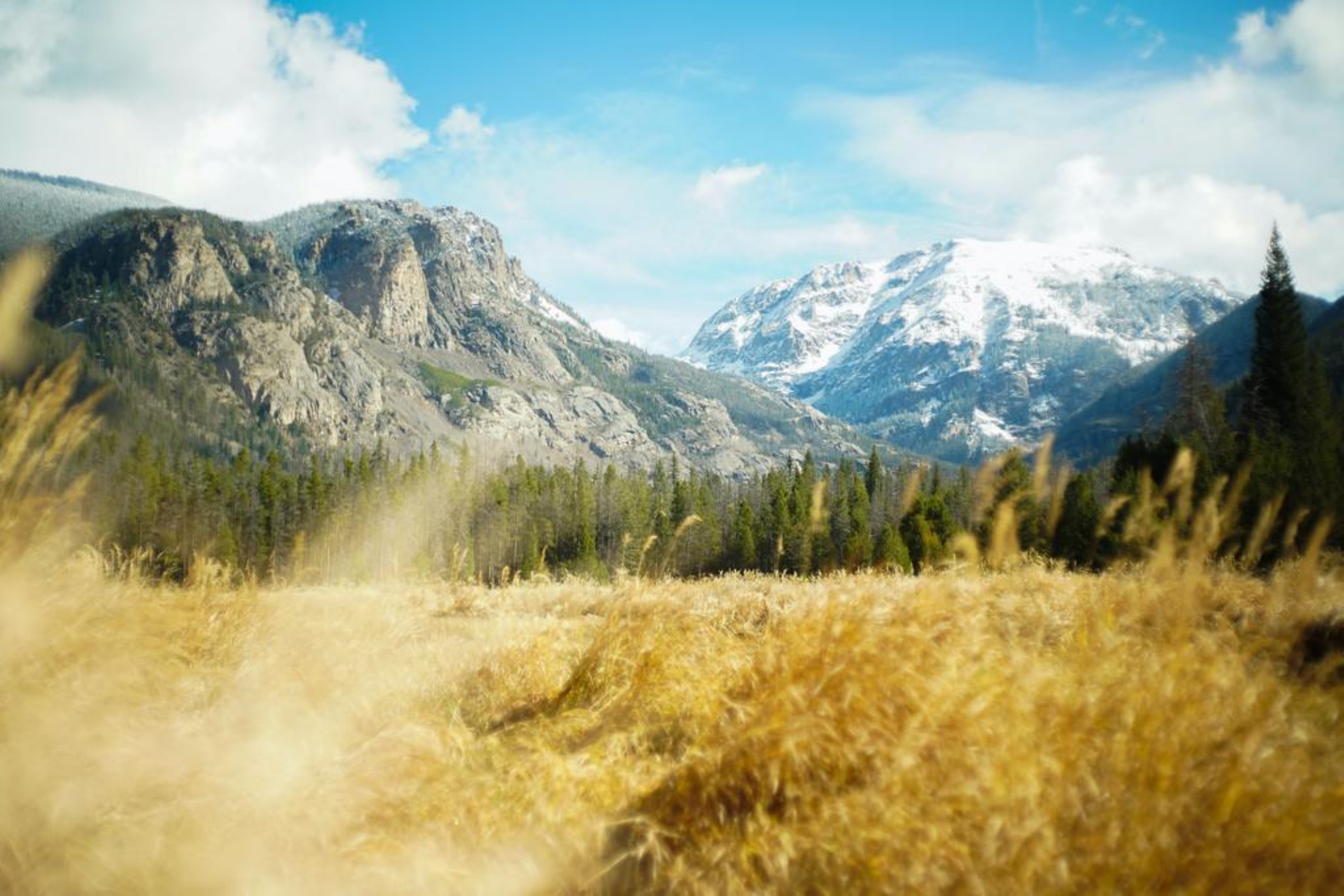}} &
        \noindent\parbox[c]{0.071\textwidth}{\includegraphics[width=0.071\textwidth]{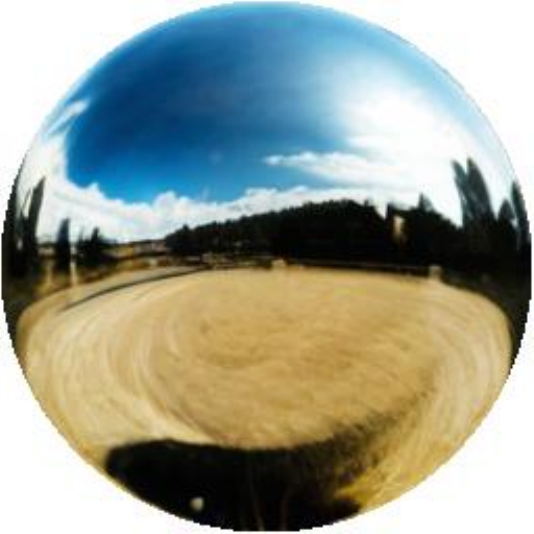}} & 
        \noindent\parbox[c]{0.071\textwidth}{\includegraphics[width=0.071\textwidth]{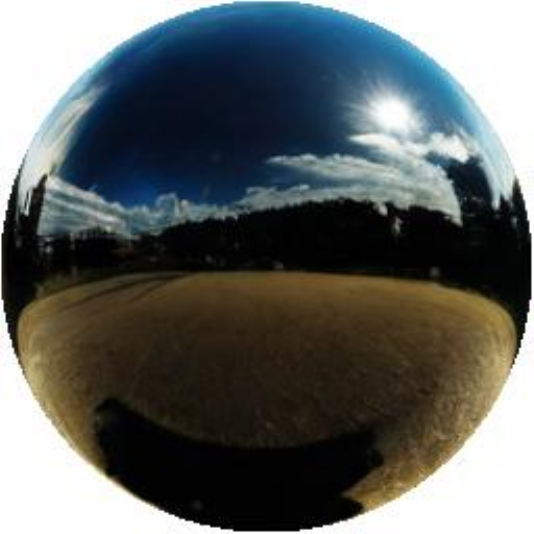}} & 
        \noindent\parbox[c]{0.090\textwidth}{\includegraphics[width=0.090\textwidth]{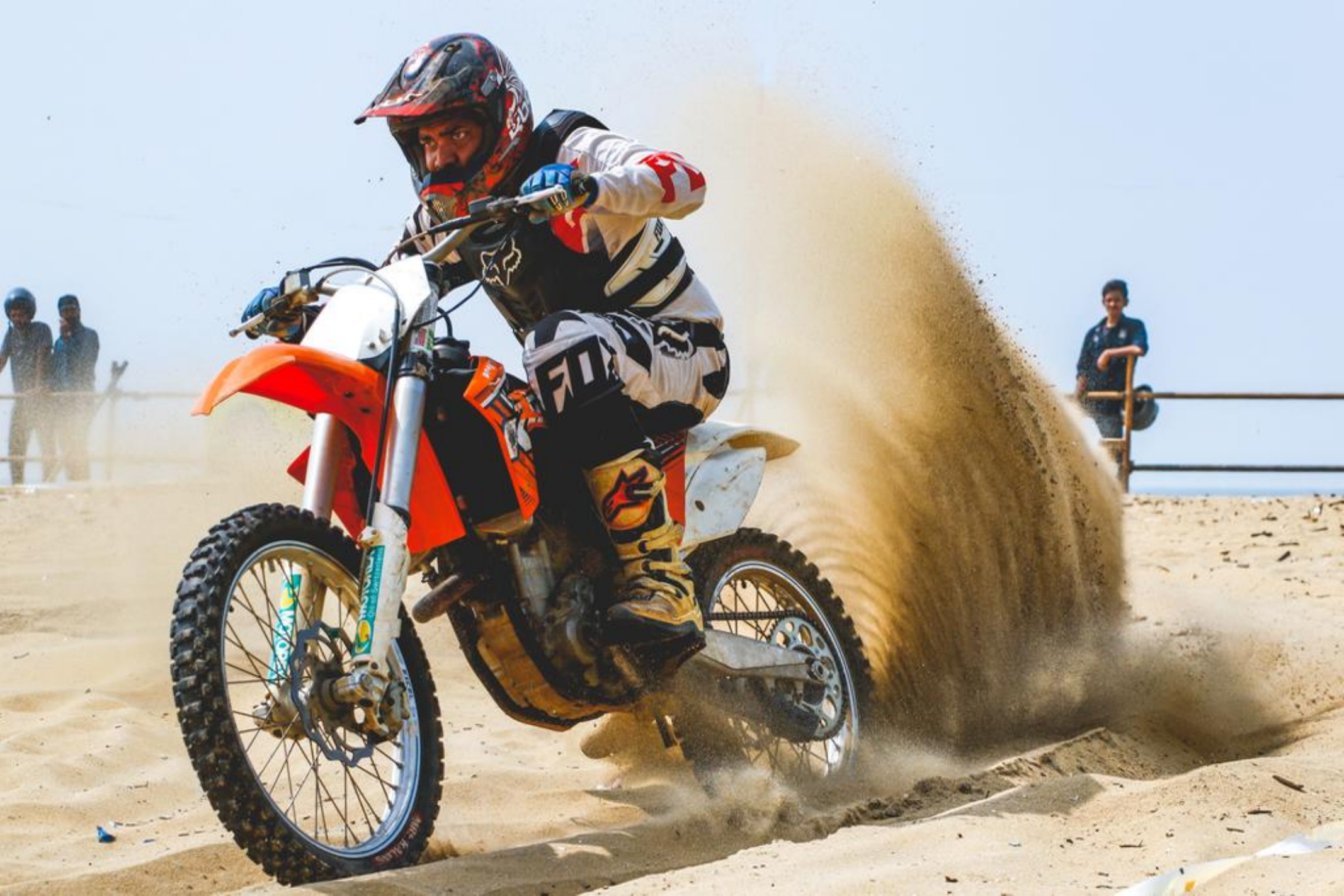}} &
        \noindent\parbox[c]{0.071\textwidth}{\includegraphics[width=0.071\textwidth]{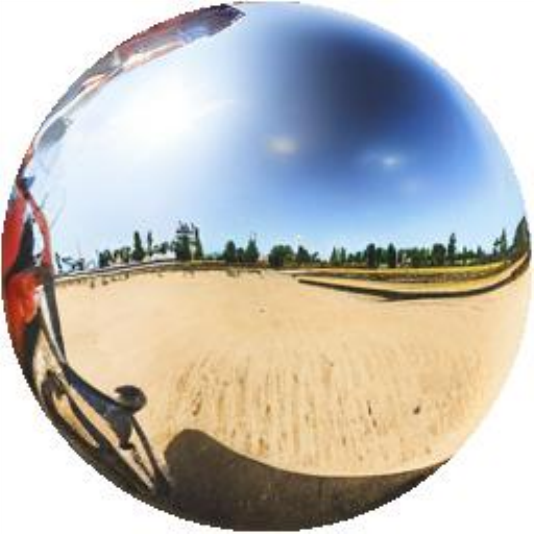}} & 
        \noindent\parbox[c]{0.071\textwidth}{\includegraphics[width=0.071\textwidth]{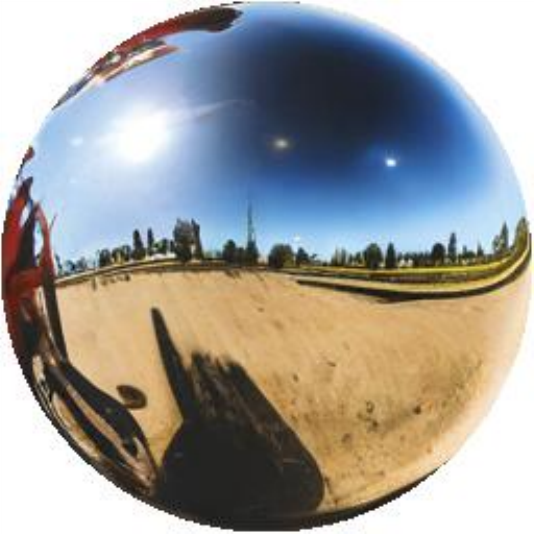}} & 
    \end{tabu}

    \caption{Simply adding ``\emph{black dark}'' to the prompt allows the overexposed sun to appear. However, we need a way to specify the target EV, which is addressed by our Exposure LoRA.}
    \label{fig:add_black_dark}
    
\end{figure}

The main issue with using pre-trained diffusion models for HDR prediction is that they have never been exposed to HDR images during training.
Nonetheless, these models can still indirectly learn about HDR and the wide range of luminance through examples of under and overexposed images in their training sets. This ability is evident in our experiment where adding `black dark' to our text prompt can reduce the overexposed white sky, allowing the round sun to emerge in outdoor scenes (see Figure \ref{fig:add_black_dark}).
To leverage this ability, we propose to use the exposure bracketing technique by inpainting multiple chrome balls with different exposure values and combining them to produce a linearized HDR output.
Our idea is to train a LoRA to steer the sampling process such that the output conforms to the appearances associated with specific exposure compensation values (EVs). 

\subsubsection{Training set} We construct a training set for our \emph{Exposure} LoRA using HDR panoramas synthetically generated from Text2Light \cite{chen2022text2light} to avoid direct access to scenes in benchmark datasets.
As illustrated in Figure \ref{fig:pipeline_overview}, each training pair consists of a random EV, denoted by $ev$, and a panorama crop with a chrome ball rendered with EV=$ev$ at the center.
This crop is constructed by projecting a full HDR panorama to a small field-of-view image and then tone-mapping to LDR \emph{without} exposure compensation (EV0). The chrome ball is rendered using the panorama as the environment map in Blender \cite{blender}, but its luminance is scaled by $2^{ev}$ before being tone-mapped to LDR.
Following \cite{wang2022stylelight}, we use a simple $\gamma$-2.4 tone-mapping function and map the 99$^\text{th}$ percentile intensity to 0.9. 

Here, we assume that the typical output images from the diffusion model have a mean EV of zero. For light estimation purposes, our focus is on recovering high-intensity light sources, which are crucial for relighting and are captured in underexposed or negative EV images. Therefore, we randomly sample the EV values from [$EV_\text{min}$, 0].



\subsubsection{Training} To generate a chrome ball with a specific EV, we condition the model on a continuous exposure signal. Rather than training a conditioning mechanism from scratch, we leverage the observation that appending ``black dark'' to the original prompt naturally induces underexposure. By interpolating between the original prompt and this augmented version according to the target $ev$ value, we provide the model with a starting point that already approximates the desired visual behavior, making training more efficient.

We define $\vect{\xi}_o$ as the text embedding of the original prompt and  $\vect{\xi}_d$ as the embedding of the prompt with ``black dark'' added. For a target  $ev$, the conditioned embedding $\vect{\xi}^{ev}$ is calculated via linear interpolation:
\begin{align}
    \vect{\xi}^{ev} = \vect{\xi}_o + (ev / EV_{\text{min}}) (\vect{\xi}_d - \vect{\xi}_o).
\end{align}
We train our LoRA with a masked version of the standard L$_2$ loss function computed only on the chrome ball pixels given by a mask $\vect{M}$:
\begin{equation} \label{eq:lora_loss_fn}
    \mathcal{L} = \mathbb{E}_{\vect{z}_0, t, \vect{\epsilon}, ev}\left[ \lVert \vect{M} \odot \parens*{\epsilon_{\vect{\theta}} (\vect{z}_t^{ev}, t, \vect{\xi}^{ev}) - \epsilon} \rVert_2^2 \right],
\end{equation}
where $\vect{z}_t^{ev}$ is computed using Equation \eqref{eq:add_noise} from our training image with EV=$ev$. We choose to train a single LoRA as opposed to multiple LoRAs for individual EVs because it helps preserve the overall scene structure across exposures due to weight sharing. Refer to Appendix \ref{appendix:aba_lora} for details.

\subsubsection{LDR balls generation and HDR merging}\label{sec:hdr} We generate chrome balls with multiple EVs $ =\{-5, -2.5, 0\}$, each using their own median ball computation. While our LoRA can  maintain the overall scene structure across exposures, some details do become altered. As a result, using standard HDR merging algorithms can lead to ghosting artifacts when details in each LDR are not fully aligned. As our primary goal is to gather high-intensity light sources from underexposed images to construct a useful light map, we can merge the luminances while retaining the chroma from the normally exposed EV0 image to reduce ghosting.

In particular, we first identify overexposed regions in each LDR image with a simple threshold of 0.9, assuming the pixel range is between 0 and 1. Then, the luminance values in these regions are replaced by the exposure-corrected luminances from lower EV images. This luminance replacement is performed in pairs, starting from the lowest EV to the normal EV0 image, detailed in Algorithm \ref{algo:ldr2hdr} in Appendix \ref{appendix:implement}. 

\tabulinesep=0.5pt


\tabulinesep=2pt

\section{DiffusionLight-Turbo}
\label{sec:fast_diffusionlight}

While DiffusionLight effectively leverages diffusion's rich image priors for light estimation, it requires inpainting and averaging multiple chrome balls for each exposure value via SDEdit \cite{meng2022sdedit}. Our new method, DiffusionLight-Turbo, accelerates this iterative inpainting algorithm by using another LoRA called \emph{Turbo LoRA} to directly predict an average chrome ball that provides a reliable overall lighting estimate. 
To further speed up inference, we explore different strategies for integrating the Turbo LoRA with the Exposure LoRA in DiffusionLight, and propose a simple yet effective \emph{LoRA swapping} that applies different LoRAs at specific timesteps within the same denoising process.

\subsection{Predicting Average Balls} \label{sec:turbo_lora}

Given a set of input images and corresponding chrome balls (e.g., predicted by DiffusionLight), one can fine-tune a diffusion model using the standard diffusion loss (Equation \ref{eq:add_noise}) to ensure it consistently produces high-quality chrome balls \textit{regardless of the initial noise maps}. 
However, this straightforward approach encourages the model to overfit to specific image details, which is counterproductive and particularly problematic when using a small dataset compared to the one used to train Stable Diffusion.
Instead, we fine-tune the model to predict median chrome balls that primarily capture low-frequency information, such as overall lighting and color. These median balls can then be used to condition the denoising process via SDEdit \cite{meng2022sdedit}, similar to the iterative inpainting approach.
Interestingly, we also observed that training converges faster when using median balls as targets compared to standard chrome balls.
A possible explanation is that the reduced variance in median balls leads to faster convergence during optimization \cite{xu2023stabletargetfieldreduced}.


\subsubsection{Training set} 
We create our Turbo LoRA training set using inpainting results from DiffusionLight. Similar to Exposure LoRA, each training pair consists of a random exposure value (EV), denoted as $ev \sim U(\{0, -2.5, -5\})$, and an LDR image containing a chrome ball at the center with EV=$ev$. To minimize bias during training, we ensure that the LDR images are evenly split between indoor and outdoor scenes by combining images from the Flickr2K dataset \cite{Timofte2017} and the InteriorVerse dataset \cite{zhu2022learning}, as approximately 80\% of images in the former are indoor scenes according to our analysis using LLaVA \cite{liu2023llava}. When applying DiffusionLight to these images, we collect LDR average balls from the final iteration of iterative inpainting and merge them into HDR average balls using the HDR merging algorithm (Section \ref{sec:hdr_algo}). Finally, we render the average balls for training pairs by scaling the luminance of the HDR balls by $2^{ev}$ before tone-mapping them using a $\gamma$-2.4 function. This step helps prevent the Turbo LoRA from learning subtle differences in details between different EVs, introduced by the Exposure LoRA.

\subsubsection{Training} We use the same LoRA fine-tuning approach as DiffusionLight, applying the masked $\text{L}_2$ loss and conditioning the model by interpolating text embeddings between the original prompt and the \emph{black dark} prompt. However, we sample the exposure value only from the set used by DiffusionLight: $\{-5.0, -2.5, 0\}$.


\begin{figure}[ht]
    \centering
    \begin{tabular}{c} 
        \begin{tabular}{c} 
                \includegraphics[width=0.46\textwidth]{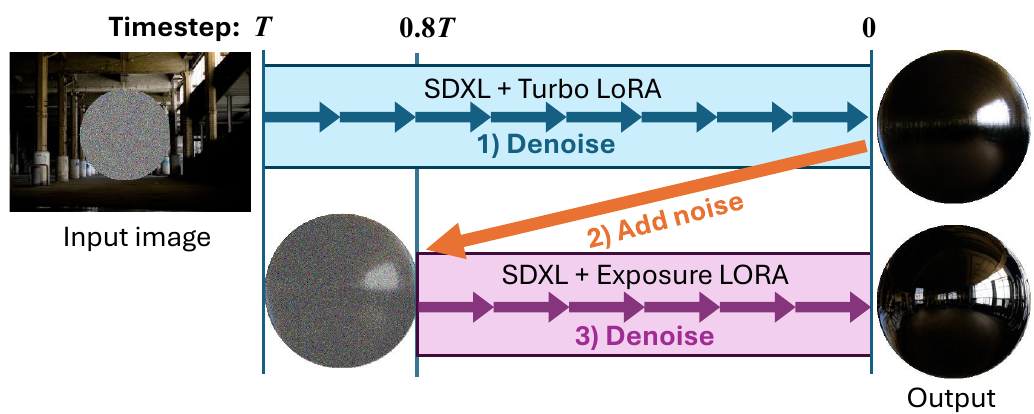} \vspace{-1em}\\
            \footnotesize{
            (a) Turbo-SDEdit}
            \vspace{0.2em}
        \end{tabular}
    \end{tabular}
    \vspace{1em} 
     \begin{tabular}{c}
        \begin{tabular}{c} 
            \includegraphics[width=0.46\textwidth]{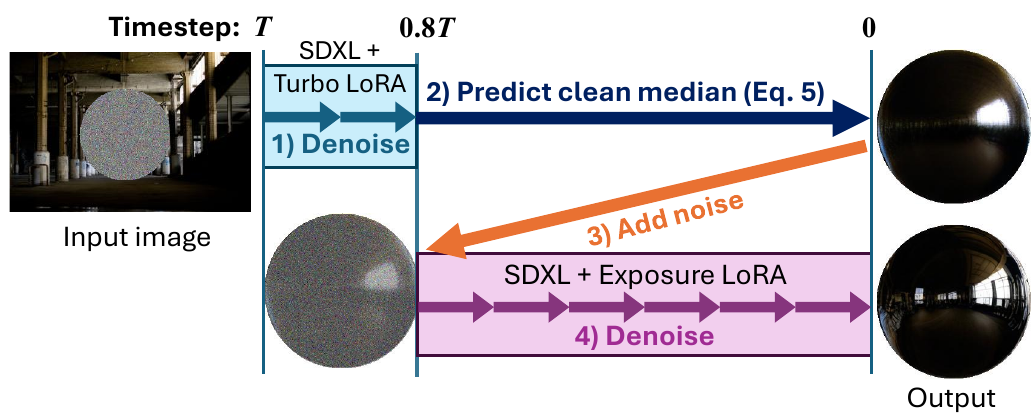} \vspace{-1em} \\ 
            \footnotesize{(b) Turbo-Pred}
            \vspace{-0.6em}
        \end{tabular}
    \end{tabular}
     \begin{tabular}{c}
        \begin{tabular}{c} 
                \includegraphics[width=0.46\textwidth]{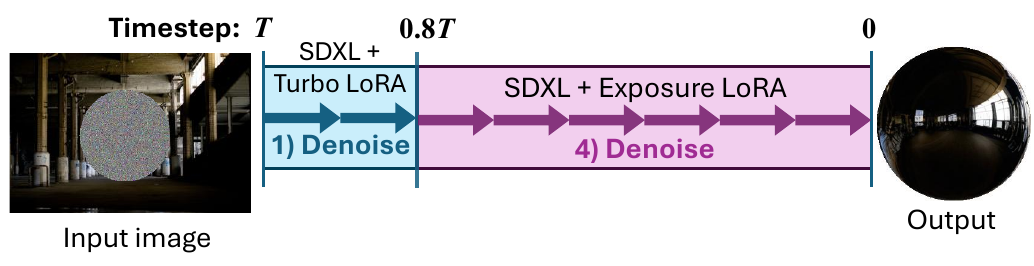} \vspace{-1em}\\ 
            \footnotesize{(c) Turbo-Swapping (ours)}
            \vspace{0.2em}
        \end{tabular}
    \end{tabular}
    \caption{Alternative inference strategies: (a) \textbf{Turbo-SDEdit}: inpaint a median ball using Turbo LoRA, then generate the final output with Exposure LoRA via SDEdit \cite{meng2022sdedit}; (b) \textbf{Turbo-Pred}: skip some denoising steps when generating the median ball with Turbo LoRA using Equation \ref{eq:predx0}; (c) \textbf{Turbo-Swapping}: apply Turbo LoRA during the early denoising steps, then switch to Exposure LoRA for the remaining steps. We adopt the last approach.}
    \label{fig:lora_swapping}
\end{figure}



\tabulinesep=0.5pt
\begin{figure}[ht]
    \centering

        \begin{tabu} to \textwidth {
        @{}
        c@{\hspace{1pt}}
        c@{\hspace{1pt}}
        c@{\hspace{1pt}}
        c@{\hspace{1pt}}
        c@{\hspace{1pt}}
        c@{\hspace{1pt}}
        c@{\hspace{1pt}}
    }


        \multicolumn{1}{c}{\shortstack{\scriptsize Input}} & 
        \multicolumn{1}{c}{\shortstack{\scriptsize \!$t = 0.8T$\!}} &
        \multicolumn{1}{c}{\shortstack{\scriptsize \!$t = 0.6T$\!}} & 
        \multicolumn{1}{c}{\shortstack{\scriptsize \!$t = 0.4T$\!}} & 
        \multicolumn{1}{c}{\shortstack{\scriptsize \!$t = 0.2T$\!}} &
        \multicolumn{1}{c}{\shortstack{\scriptsize \!$t = 0$\!}} &
        \\

        \noindent\parbox[c]{0.102\textwidth}{\includegraphics[width=0.102\textwidth]{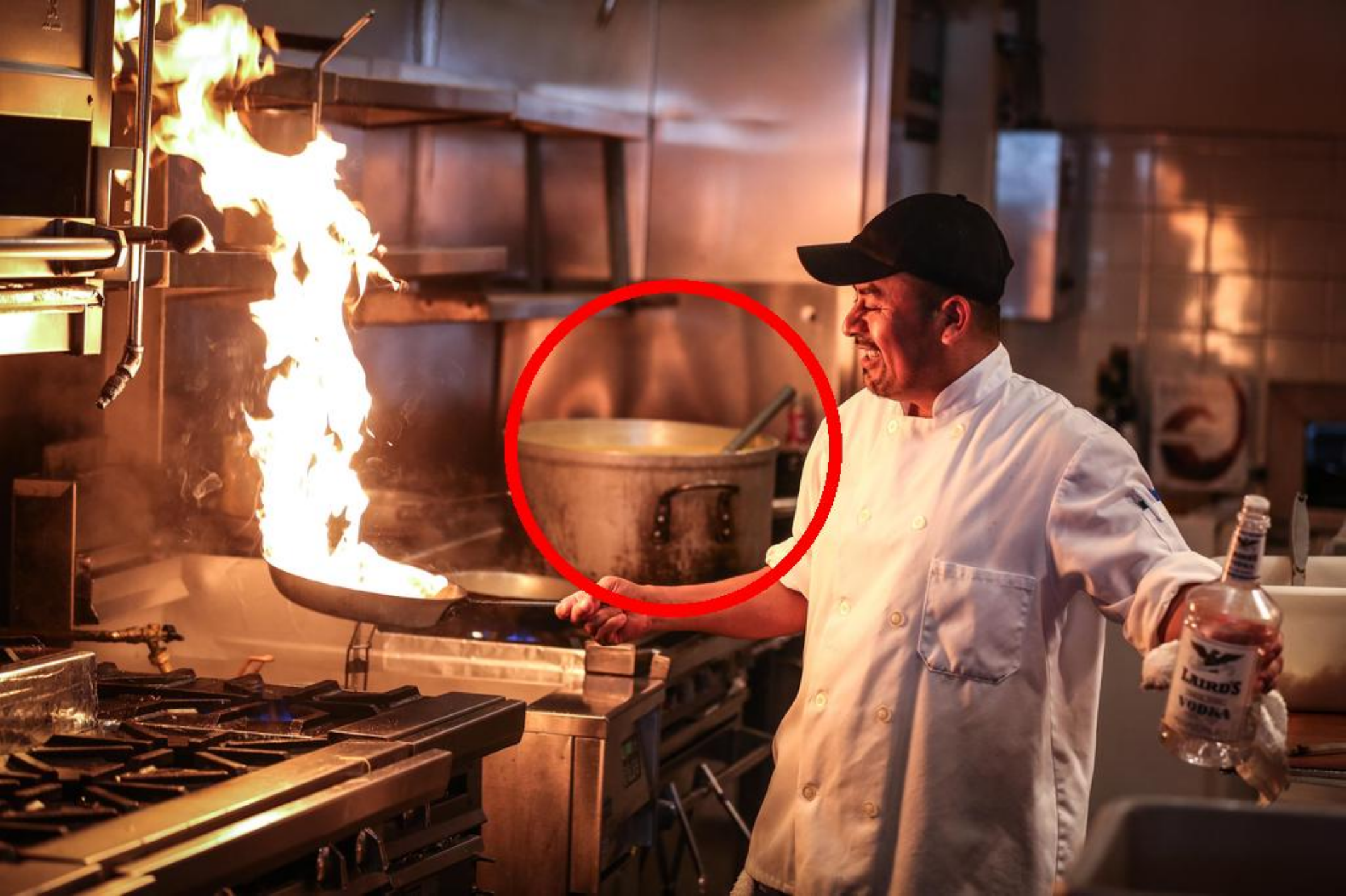}} & 
        \noindent\parbox[c]{0.069\textwidth}{\includegraphics[width=0.069\textwidth]{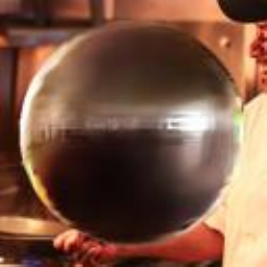}} &  
        \noindent\parbox[c]{0.069\textwidth}{\includegraphics[width=0.069\textwidth]{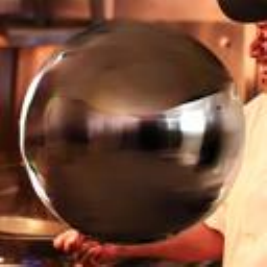}} &
        \noindent\parbox[c]{0.069\textwidth}{\includegraphics[width=0.069\textwidth]{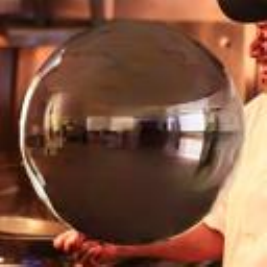}} &
        \noindent\parbox[c]{0.069\textwidth}{\includegraphics[width=0.069\textwidth]{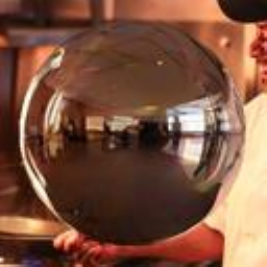}} &
        \noindent\parbox[c]{0.069\textwidth}{\includegraphics[width=0.069\textwidth]{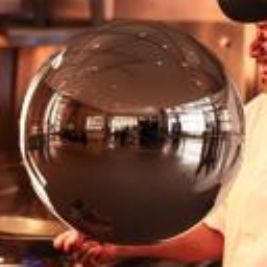}} &
        \\

        \noindent\parbox[c]{0.102\textwidth}{\includegraphics[width=0.102\textwidth]{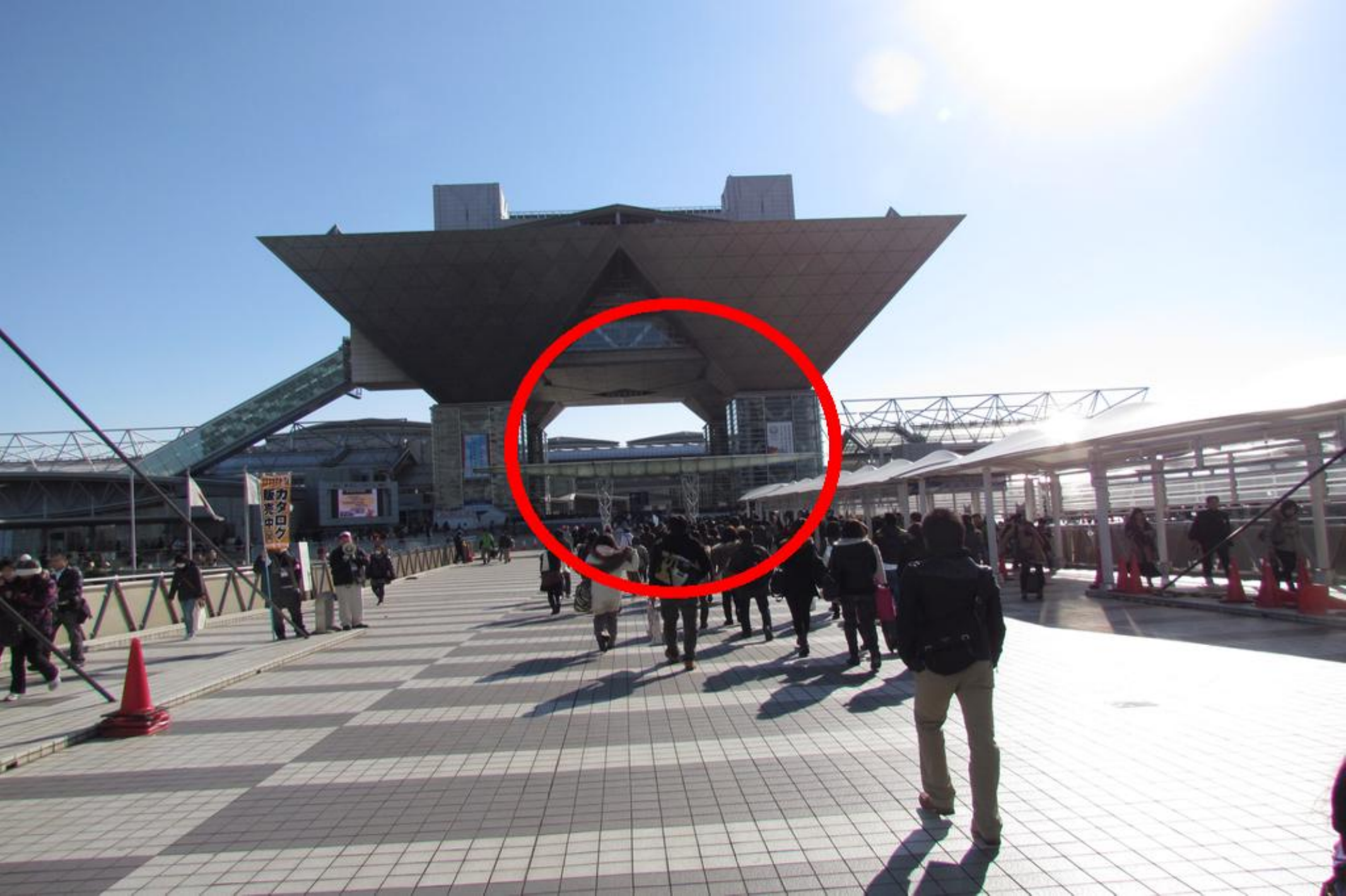}} & 
        \noindent\parbox[c]{0.069\textwidth}{\includegraphics[width=0.069\textwidth]{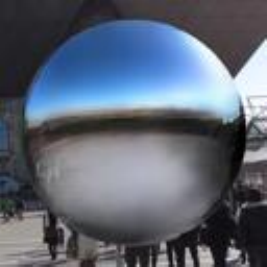}} &  
        \noindent\parbox[c]{0.069\textwidth}{\includegraphics[width=0.069\textwidth]{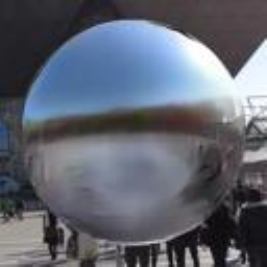}} &
        \noindent\parbox[c]{0.069\textwidth}{\includegraphics[width=0.069\textwidth]{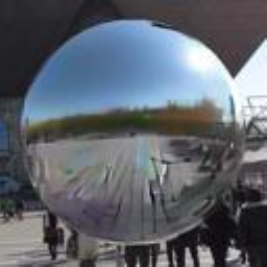}} &
        \noindent\parbox[c]{0.069\textwidth}{\includegraphics[width=0.069\textwidth]{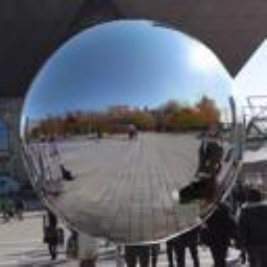}} &
        \noindent\parbox[c]{0.069\textwidth}{\includegraphics[width=0.069\textwidth]{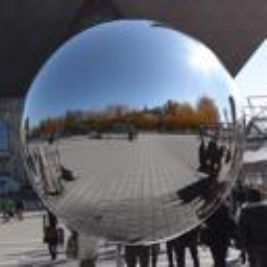}} &
        \\

        \end{tabu}
    \caption{Overall lighting is primarily determined during the early steps of sampling. Here, we visualize intermediate predictions at various timesteps $t$ during 30-step sampling with UniPC \cite{zhao2023unipc}. These intermediate predictions (i.e., predicted $\vect{z}_0$) can be computed from $\vect{z}_t$ at any timestep $t$ using Equation \ref{eq:add_noise}. 
    }
    \label{fig:main_lightinfo_early_stage}
\end{figure}

\subsection{LoRA Swapping at Inference Time} \label{sec:lora_swapping}

As shown in Figure \ref{fig:lora_swapping}(a), by replacing the iterative inpainting with Turbo LoRA, the inference pipeline is simplified into three stages: (1) generating an average ball $\vect{B}^*$, (2) simulating the diffusion process up to time $t=cT$, where $0 < c < 1$, using SDEdit \cite{song2020denoising}: $\vect{B}^*_t = \sqrt{\alpha_t} \vect{B}^* + \sqrt{1 - \alpha_t} \epsilon$ where $\epsilon \sim \mathcal{N}(\vect{0}, \vect{I})$, and (3) denoising $\vect{B}^*_t$ from $t=cT$ to 0. While this pipeline (referred to as \emph{Turbo-SDEdit}) allows us to inpaint high-quality chrome balls using two denoising processes, we can accelerate it even further.

Specifically, we observed that the predicted clean median ball at timestep $t$,
\begin{equation} \label{eq:predx0}
    \hat{\vect{z}}_0(t) = \frac{\vect{z}_t - \sqrt{1 - \alpha_t} \epsilon_{\theta}(\vect{z}_t, t)}{\sqrt{\alpha_t}},
\end{equation}
already contains the overall lighting information even at $t=0.8T$  (see Figure \ref{fig:main_lightinfo_early_stage}).
This suggests that there is no need to continue to denoise from $t=0.8T$ down to $t=0$ because the subsequent SEDdit operation will add noise to get back to $t=0.8T$ again. Instead, once we reach $t=0.8T$, we can simply predict the clean ball $\hat{\vect{z}}_0(t)$ and apply SDEdit to it. We call this process the \emph{Turbo-Pred} pipeline and show it in Figure \ref{fig:lora_swapping}(b).


A potential drawback of predicting the clean sample $\hat{\vect{z}}_0(t)$ is that it may introduce additional errors, leading to greater deviation from the result of the Turbo-SDEdit pipeline. To avoid this, we propose simply removing the second stage of the \emph{Turbo-SDEdit} pipeline. Specifically, we first denoise the sample using Turbo LoRA during $t \in [T, 0.8T]$ to inject a good lighting prior, then switch to the Exposure LoRA during $t \in (0.8T, 0]$ to add high-frequency details. We refer to this simplified pipeline as \emph{Turbo-Swapping}, as illustrated in Figure \ref{fig:lora_swapping}(c).
In our implementation, we use the Turbo-Swapping pipeline as it offers the best quality-runtime tradeoff (see Table \ref{tab:aba_diffusionlight_turbo}).

\section{Experiments}
\label{sec:experiment}

\begin{table*}[ht]
\centering
\small
\caption{Scores on three-sphere evaluation protocol. The \colorbox{tabfirst}{best} and \colorbox{tabsecond}{second-best} are color-coded. All running times are measured on a single RTX 3090 Ti GPU over 10 random samples.}
\label{tab:indoor_stylelight}

\resizebox{1\textwidth}{!}{%
\setlength{\tabcolsep}{1pt}
\begin{tabular}{
    l@{\hspace{9pt}}
    l@{\hspace{9pt}}
    c@{\hspace{9pt}}
    c@{\hspace{9pt}}
    c@{\hspace{9pt}}
    c@{\hspace{9pt}}
    c@{\hspace{9pt}}
    c@{\hspace{9pt}}
    c@{\hspace{9pt}}
    c@{\hspace{9pt}}
    c@{\hspace{9pt}}
    c@{\hspace{9pt}}
    c
}
\toprule
\multirow{2}{*}{\textbf{Dataset}} & \multirow{2}{*}{\textbf{Method}} & \textbf{Time} & \multicolumn{3}{c}{\textbf{Scale-invariant RMSE} $\downarrow$} & \multicolumn{3}{c}{\textbf{Angular Error} $\downarrow$}  & \multicolumn{3}{c}{\textbf{Normalized RMSE} $\downarrow$}                                                            \\
 & & \textbf{(mins)} & Diffuse           & Matte           & Mirror           & Diffuse           & Matte           & Mirror   & Diffuse           & Matte           & Mirror      \\ 
\midrule

\multirow{6}{*}{\shortstack[l]{Laval \\ Indoor \cite{garder2017lavelindoor}}} & StyleLight {\footnotesize(reported by the paper)} & - & 0.11 & 0.29 & 0.55 & 2.41 & 2.96 & 4.30 & - & - & - \\
\cline{2-12}\\[-2.1ex]

& StyleLight {\footnotesize (reproduced using official code)} & 0.607 & \scsecond{0.135} & \scfirst{0.315} & \scfirst{0.552} & 4.238 & 4.742 & 6.781 & 0.234 & 0.404 & 0.511 \\
& SDXL$^+$ & \scfirst{0.523} & 0.196 & 0.453 & 0.715 & 5.867 & 6.196 & 8.279 & 0.320 & 0.475 & 0.510 \\
& \textbf{DiffusionLight} & 31.889 & 0.140 & \scsecond{0.325} & \scsecond{0.597} & \scfirst{2.139} & \scfirst{3.421} & \scfirst{5.936} & \scfirst{0.201} & \scfirst{0.361} & \scfirst{0.431} \\
& \textbf{DiffusionLight-Turbo} & \scsecond{0.535} & \scfirst{0.130} & 0.349 & 0.620 & \scsecond{2.983} & \scsecond{3.687} & \scsecond{6.114} & \scsecond{0.232} & \scsecond{0.370} & \scsecond{0.433} \\[0.3ex]
\hline\\[-2.2ex]
\multirow{5}{*}{\shortstack[l]{Poly \\ Haven \cite{polyhaven}}} & StyleLight {\footnotesize (reproduced using official code)} & 0.607 & 0.166 & \scfirst{0.441} & \scfirst{0.639} & 3.533 & 4.438 & 7.122 & 0.225 & 0.415 & 0.488 \\
& SDXL$^+$ & \scfirst{0.523} & 0.218 & 0.585 & 0.804 & 2.983 & 4.252 & 5.738 & 0.296 & 0.504 & 0.543 \\
& \textbf{DiffusionLight} & 31.889 & \scfirst{0.140} & \scsecond{0.446} & \scsecond{0.664} & \scfirst{2.139} & \scfirst{3.595} & \scfirst{4.286} & \scfirst{0.201} & \scfirst{0.388} & \scfirst{0.428} \\
& \textbf{DiffusionLight-Turbo} & \scsecond{0.535} & \scsecond{0.161} & 0.472 & 0.680 & \scsecond{2.348} & \scsecond{3.696} & \scsecond{4.504} & \scsecond{0.222} & \scsecond{0.402} & \scsecond{0.439} \\

\bottomrule
\addlinespace[0.2em]
\multicolumn{12}{l}{\textit{SDXL$^+$ denotes a simple inpainting pipeline with SDXL \cite{podell2023sdxl} and depth-conditioned ControlNet \cite{zhang2023adding}}.} \\
\end{tabular}}
\end{table*}

\subsection{Implementation Details}
\subsubsection{DiffusionLight} We fine-tuned SDXL \cite{podell2023sdxl} for multi-exposure generation using a rank-4 LoRA \cite{hu2021lora}. We trained our Exposure LoRA on 1,412 HDR panoramas synthetically generated by Text2Light \cite{chen2022text2light} for 2,500 steps with a learning rate of $10^{-5}$ and a batch size of 4. During training, we sampled timestep $t$ from $U(900, 999)$ to help speed up training as we found that the light information is determined in the early timesteps of the denoising process.


When applying iterative inpainting, we generated $N = 30$ chrome balls per each median computation iteration. We use UniPC \cite{zhao2023unipc} sampler with 30 sampling steps, a guidance scale of 5.0, and a LoRA scale of $0.75$.


\subsubsection{DiffusionLight-Turbo} 
We trained a rank-4 Turbo LoRA to predict average chrome balls using 4,650 LDR images---2,650 from the Flickr2K dataset and 2,000 from the InteriorVerse dataset---along with reference median balls obtained from DiffusionLight.
The LoRA was trained for 230,000 steps with a learning rate of $4 \times 10^{-4}$ and a batch size of 4. During training, we also sampled timestep $t$ from $U(500, 999)$ to better learn low-frequency details, which are dominant in median chrome balls. During inference, we use Turbo LoRA (Section \ref{sec:fast_diffusionlight}) during timestep $t \in [999, 800]$ and switch to Exposure LoRA (Section \ref{sec:diffusionlight}) for $t \in (800, 0]$.




\subsection{Evaluation Details}

\subsubsection{Datasets} We evaluated our approach on two standard benchmarks: Laval Indoor HDR \cite{garder2017lavelindoor} and Poly Haven \cite{polyhaven}. The latter covers both indoor and outdoor settings.

\subsubsection{Evaluation metrics}
Following previous works \cite{wang2022stylelight, zhan2021emlight}, we used three scale-invariant metrics: scale-invariant Root Mean Square Error (si-RMSE) \cite{grosse2009groundtruth}, Angular Error \cite{legendre2019deeplight}, and normalized RMSE. The normalization for the last metric is done by mapping the 0.1st and 99.9th percentiles to 0 and 1, following \cite{marnerides2019expandnet}. We chose these metrics instead of standard RMSE because each benchmark dataset has its own specific range and statistics of light intensity, but our method was not trained on any of them. 

\begin{table*}[!h]
\centering
\small
\caption{Ablations of DiffusionLight on three-sphere evaluation protocol.}
\label{tab:aba_diffusionlight}

\resizebox{1\textwidth}{!}{%
\setlength{\tabcolsep}{1.2pt}
\begin{tabular}{
    l@{\hspace{9pt}}
    l@{\hspace{9pt}}
    c@{\hspace{9pt}}
    c@{\hspace{9pt}}
    c@{\hspace{9pt}}
    c@{\hspace{9pt}}
    c@{\hspace{9pt}}
    c@{\hspace{9pt}}
    c@{\hspace{9pt}}
    c@{\hspace{9pt}}
    c@{\hspace{9pt}}
    c
}
\toprule
\multirow{2}{*}{\textbf{Dataset}} & \multirow{2}{*}{\textbf{Method}} & \multicolumn{3}{c}{\textbf{Scale-invariant RMSE} $\downarrow$} & \multicolumn{3}{c}{\textbf{Angular Error} $\downarrow$}  & \multicolumn{3}{c}{\textbf{Normalized RMSE} $\downarrow$}                                                            \\
 & & Diffuse           & Matte           & Mirror           & Diffuse           & Matte           & Mirror   & Diffuse           & Matte           & Mirror      \\ 
\midrule

\multirow{3}{*}{\shortstack[l]{Laval \\ Indoor \cite{garder2017lavelindoor}}} & DiffusionLight w/o LoRA  & \scsecond{0.146} & 0.388 & 0.669 & 3.581 & 4.551 & 7.053 & \scsecond{0.246} & 0.408 & 0.467 \\
& DiffusionLight w/o iterative  & 0.147 & \scsecond{0.380} & \scsecond{0.651} & \scsecond{3.470} & \scsecond{3.864} & \scsecond{6.146} & 0.255 & \scsecond{0.397} & \scsecond{0.451} \\
& \textbf{DiffusionLight (ours)}  & \scfirst{0.140} & \scfirst{0.325} & \scfirst{0.597} & \scfirst{2.139} & \scfirst{3.421} & \scfirst{5.936} & \scfirst{0.201} & \scfirst{0.361} & \scfirst{0.431} \\[0.3ex]

\hline\\[-2.2ex]

\multirow{3}{*}{\shortstack[l]{Poly \\ Haven \cite{polyhaven}}} & DiffusionLight w/o LoRA  & \scsecond{0.161} & 0.500 & 0.735 & 2.574 & 4.183 & 5.306 & \scsecond{0.221} & 0.443 & 0.495 \\
& DiffusionLight w/o iterative  & 0.165 & \scsecond{0.499} & \scsecond{0.718} & \scsecond{2.352} & \scfirst{3.560} & \scsecond{4.433} & 0.235 & \scsecond{0.430} & \scsecond{0.470} \\
& \textbf{DiffusionLight (ours)}  & \scfirst{0.140} & \scfirst{0.446} & \scfirst{0.664} & \scfirst{2.139} & \scsecond{3.595} & \scfirst{4.286} & \scfirst{0.201} & \scfirst{0.388} & \scfirst{0.428} \\

\bottomrule
\end{tabular}
}
\end{table*}
\begin{table*}[!h]
\centering
\small
\caption{Ablations of DiffusionLight-Turbo on three-sphere evaluation protocol.}
\label{tab:aba_diffusionlight_turbo}

\resizebox{1\textwidth}{!}{%
\setlength{\tabcolsep}{1pt}
\begin{tabular}{
    l@{\hspace{9pt}}
    l@{\hspace{9pt}}
    c@{\hspace{9pt}}
    c@{\hspace{9pt}}
    c@{\hspace{9pt}}
    c@{\hspace{9pt}}
    c@{\hspace{9pt}}
    c@{\hspace{9pt}}
    c@{\hspace{9pt}}
    c@{\hspace{9pt}}
    c@{\hspace{9pt}}
    c@{\hspace{9pt}}
    c
}
\toprule
\multirow{2}{*}{\textbf{Dataset}} & \multirow{2}{*}{\textbf{Method}} & \textbf{Time} & \multicolumn{3}{c}{\textbf{Scale-invariant RMSE} $\downarrow$} & \multicolumn{3}{c}{\textbf{Angular Error} $\downarrow$}  & \multicolumn{3}{c}{\textbf{Normalized RMSE} $\downarrow$}                                                            \\
 & & \textbf{(mins)} & Diffuse           & Matte           & Mirror           & Diffuse           & Matte           & Mirror   & Diffuse           & Matte           & Mirror      \\ 
\midrule

\multirow{3}{*}{\shortstack[l]{Laval \\ Indoor \cite{garder2017lavelindoor}}} & Turbo-SDEdit & 1.229 & \scsecond{0.133} & \scsecond{0.353} & 0.630 & \scfirst{2.938} & \scsecond{3.708} & \scsecond{6.255} & \scsecond{0.238} & \scsecond{0.372} & \scsecond{0.438} \\
& Turbo-Pred & \scsecond{0.825} & 0.140 & 0.355 & \scsecond{0.621} & 3.002 & 3.951 & 6.449 & 0.239 & 0.379 & 0.441 \\
& \textbf{Turbo-Swapping (ours)} & \scfirst{0.535} & \scfirst{0.130} & \scfirst{0.349} & \scfirst{0.620} & \scsecond{2.983} & \scfirst{3.687} & \scfirst{6.114} & \scfirst{0.232} & \scfirst{0.370} & \scfirst{0.433} \\ [0.3ex]

\hline\\[-2.2ex]

\multirow{3}{*}{\shortstack[l]{Poly \\ Haven \cite{polyhaven}}}  & Turbo-SDEdit & 1.229 & \scfirst{0.154} & \scfirst{0.459} & \scfirst{0.671} & \scsecond{2.369} & \scsecond{3.785} & \scsecond{4.600} & \scfirst{0.213} & \scfirst{0.391} & \scfirst{0.435} \\
& Turbo-Pred & \scsecond{0.825} & 0.166 & \scsecond{0.467} & \scsecond{0.678} & 2.475 & 4.135 & 4.926 & \scsecond{0.220} & \scsecond{0.396} & 0.440 &  \\
& \textbf{Turbo-Swapping (ours)} & \scfirst{0.535} & \scsecond{0.161} & 0.472 & 0.680 & \scfirst{2.348} & \scfirst{3.696} & \scfirst{4.504} & 0.221 & 0.402 & \scsecond{0.439} &  \\


\bottomrule
\end{tabular}}
\end{table*}
\begin{table}[!h]
\centering
\caption{Scores on indoor array-of-spheres protocol
}
\label{tab:indoor_everlight}
\resizebox{0.95\columnwidth}{!}{%
\setlength{\tabcolsep}{5pt}
\begin{tabular}{lcc
}
\toprule
\textbf{Method} & \textbf{si-RMSE} $\downarrow$ & \textbf{Angular Error} $\downarrow$ \\
\midrule
EverLight \cite{dastjerdi2023everlight} & 0.091 & 6.36 \\
StyleLight \cite{wang2022stylelight} & 0.123 & 7.09 \\
Weber et al. \cite{weber2022editableindoor} & \scfirst{0.081} & \scsecond{4.13} \\
EMLight \cite{zhan2021emlight} & 0.099 & \scfirst{3.99} \\
\textbf{DiffusionLight} & \scsecond{0.090} & 5.25 \\
\textbf{DiffusionLight-Turbo} & 0.103 & 7.35 \\

\bottomrule
\end{tabular}}
\end{table}
\begin{table}[]
\centering
\small
\caption{Scores on the random-camera protocol.
}
\label{tab:randfov}
\begin{tabular}{
    l@{\hspace{5pt}}
    l@{\hspace{3pt}}
    c@{\hspace{3pt}}
    c@{\hspace{3pt}}
    c
}
\toprule
\textbf{Sphere} & \textbf{Method} & \textbf{si-RMSE} $\downarrow$ & \begin{tabular}[c]{@{}c@{}}\textbf{Angular}\\ \textbf{Error}\end{tabular} $\downarrow$ & \begin{tabular}[c]{@{}c@{}}\textbf{Norm.}\\ \textbf{RMSE}\end{tabular} $\downarrow$ \\
\midrule

Diffuse & StyleLight & \colorbox{tabsecond}{0.143} & 3.741 & \colorbox{tabsecond}{0.236}\\
        & \textbf{DiffusionLight} & \colorbox{tabfirst}{0.135} & \colorbox{tabfirst}{2.337} & \colorbox{tabfirst}{0.219} \\
        & \textbf{DiffusionLight-Turbo} & 0.149 & \colorbox{tabsecond}{3.060} & 0.245 \\[0.3ex]
\hline\\[-2.2ex]
Matte   & StyleLight & \colorbox{tabfirst}{0.347} & 4.492 & 0.429 \\
        & \textbf{DiffusionLight} & \colorbox{tabsecond}{0.359} & \colorbox{tabfirst}{3.483} & \colorbox{tabfirst}{0.369} \\
        & \textbf{DiffusionLight-Turbo} & 0.382 & \colorbox{tabsecond}{3.793} & \colorbox{tabsecond}{0.385} \\[0.3ex]
  \hline\\[-2.2ex]
Mirror & StyleLight & \colorbox{tabfirst}{0.606} & 7.655 & 0.544\\
 & \textbf{DiffusionLight} & \colorbox{tabsecond}{0.644} & \colorbox{tabfirst}{5.988} & \colorbox{tabfirst}{0.438} \\
 & \textbf{DiffusionLight-Turbo} & 0.661 & \colorbox{tabsecond}{6.182} & \colorbox{tabsecond}{0.446} \\

\bottomrule
\end{tabular}
\end{table}

\subsubsection{Evaluation protocols} We adopt two different evaluation protocols used in the literature: from each input LDR image, we generate an HDR panorama of size $128 \times 256$ pixels and use it to render (1) three spheres with different materials (gray-diffuse, silver-matte, and silver-mirror spheres) \cite{wang2022stylelight, gardner2019deepparam, garder2017lavelindoor} or (2) an array of diffuse spheres \cite{dastjerdi2023everlight, weber2022editableindoor}. Then, we computed the evaluation metrics on these renderings. Many studies do not publish their source code and use only one of the protocols, resulting in missing baselines' scores in some experiments.


\subsection{Results}

\subsubsection{Evaluation on three spheres} We compared our method to StyleLight \cite{wang2022stylelight} on (1) 289 panoramas from the Laval Indoor dataset and (2) 500 panoramas from Poly Haven dataset. 
As StyleLight was trained on the Laval \emph{Indoor} dataset, its scores on Poly Haven are provided solely as a reference to demonstrate how existing methods perform in out-of-distribution scenarios.
Following StyleLight's protocol, we created one input image from each panorama by cropping it to a size of $192 \times 256$ with a vertical FOV of $60^\circ$ and then applying tone-mapping, setting the 99th percentile to 0.9 and using $\gamma = 2.4$. In only our pipeline, we upscaled the image while keeping the aspect ratio for SDXL.

Table \ref{tab:indoor_stylelight} shows that DiffusionLight outperforms StyleLight in terms of Angular Error and Normalized RMSE on Laval indoor dataset, with significantly lower Angular Error: 49.5\% (diffuse), 27.8\% (matte), and 12.4\% (mirror). It is also effective on Poly Haven outdoor scenes, while StyleLight's performance drops with a large 39.7\% gap in Angular Error for mirror spheres.
After applying our Turbo LoRA, DiffusionLight-Turbo achieves a 59.9x speedup, with only a slight reduction in accuracy. Nonetheless, it still outperforms StyleLight in both Angular Error and Normalized RMSE across both datasets.

\begin{figure}
    \centering

    \begin{tabu} to \textwidth {
        @{}
        c@{\hspace{1pt}}
        c@{\hspace{1pt}}
        c@{\hspace{1pt}}
        c@{\hspace{1pt}}
        c@{\hspace{1pt}}
    }
        \noindent\parbox[c]{0.085\textwidth}{\centering \hspace{0.01\textwidth} \tiny Input image} &
        \noindent\parbox[c]{0.08\textwidth}{\centering \tiny  GT} &
        \noindent\parbox[c]{0.08\textwidth}{\centering \tiny  Stylelight} &        
        \noindent\parbox[c]{0.08\textwidth}{\centering \tiny DiffusionLight (Ours)} &
        \noindent\parbox[c]{0.08\textwidth}{\centering \tiny DiffusiobLight-Turbo (Ours)} 
        \\
        
        \multicolumn{5}{c}{
        \noindent\parbox[c]{0.4\textwidth}{\includegraphics[width=0.4\textwidth]{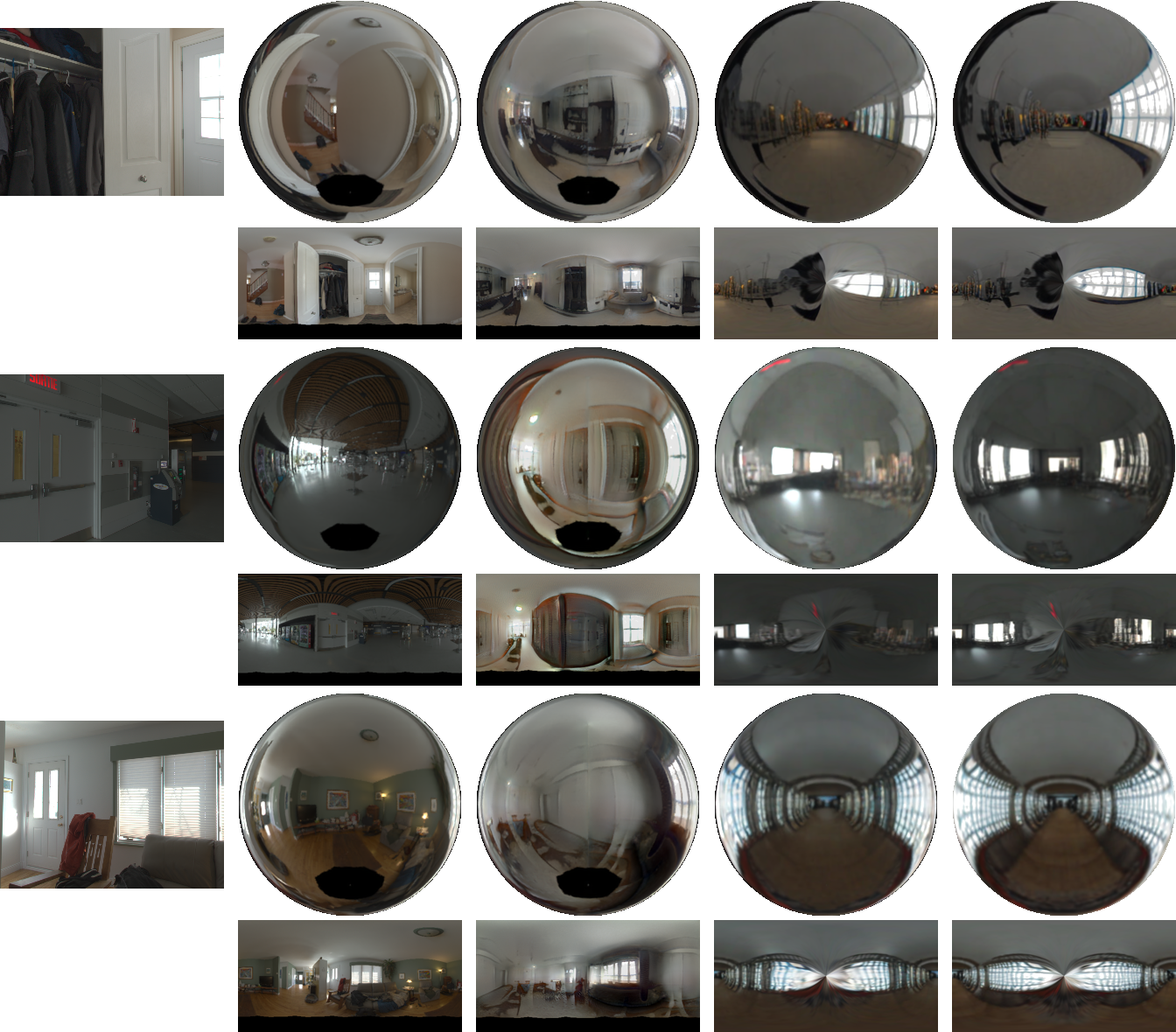}
         \\ \centering \scriptsize
         (a) Laval Indoor \cite{garder2017lavelindoor}
        }
        } \\

        \multicolumn{5}{c}{
        \noindent\parbox[c]{0.4\textwidth}{\includegraphics[width=0.4\textwidth]{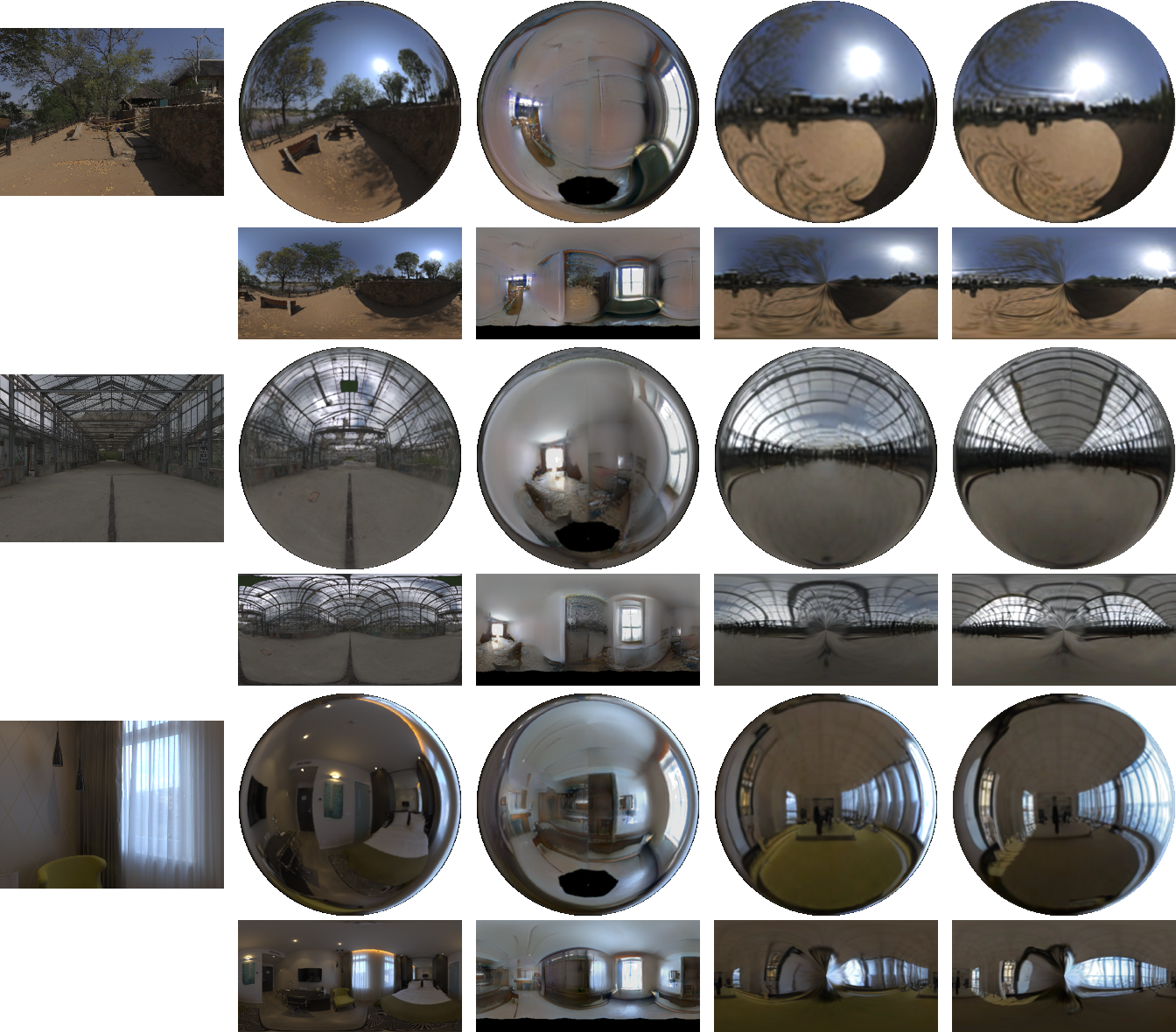}
         \\ \centering \scriptsize
         (b) PolyHaven \cite{polyhaven}
        }
        }
        
    \end{tabu}
    \caption{Qualitative results on benchmark datasets: (a) Laval Indoor \cite{garder2017lavelindoor} and (b) Poly Haven \cite{polyhaven}. For each input image, we show the rendered chrome ball (1\textsuperscript{st} row) and the corresponding environment map (2\textsuperscript{nd} row) from each method.}
    \label{fig:qualitative_benchmark}
\end{figure}

We show qualitative results in Figure \ref{fig:qualitative_benchmark} and Appendix \ref{appendix:more_result_benchmark}.
Note that we used StyleLight's official code to produce these scores; however, discrepancies exist with those reported in the paper. (See Appendix \ref{appendix:stylelight-score-diff} 
for details and our discussion with StyleLight's authors on this issue).


\subsubsection{Evaluation on array of spheres} 
We compared our approach with StyleLight, EverLight \cite{dastjerdi2023everlight}, EMLight \cite{zhan2021emlight}, and Weber et al. \cite{weber2022editableindoor} using 224 panoramas from the Laval Indoor dataset---the same set used in EverLight's evaluations. For each panorama, we generated 10 input LDR images by centering the panorama at certain azimuthal angles and cropping to a $50^\circ$ field of view, following Weber et al. \cite{weber2022editableindoor}. This results in 2,240 input-output pairs for evaluation.

\begin{figure}
    \centering

    \begin{tabu} to \textwidth {
        @{}
        c@{\hspace{1pt}}
        c@{\hspace{1pt}}
        c@{\hspace{1pt}}
        c@{\hspace{1pt}}
        c@{\hspace{1pt}}
        c@{\hspace{1pt}}
        c@{}
    }
        \noindent\parbox[c]{0.090\textwidth}{\centering \tiny Input image} &
        \noindent\parbox[c]{0.071\textwidth}{\centering \tiny  DiffusionLight} &
        \noindent\parbox[c]{0.071\textwidth}{\centering \tiny  DiffusionLight-Turbo} &
        \noindent\parbox[c]{0.090\textwidth}{\centering \tiny Input image} &
        \noindent\parbox[c]{0.071\textwidth}{\centering \tiny DiffusionLight} &
        \noindent\parbox[c]{0.071\textwidth}{\centering \tiny DiffusiobLight-Turbo} 
        \\
        
        \noindent\parbox[c]{0.090\textwidth}{\includegraphics[width=0.090\textwidth]{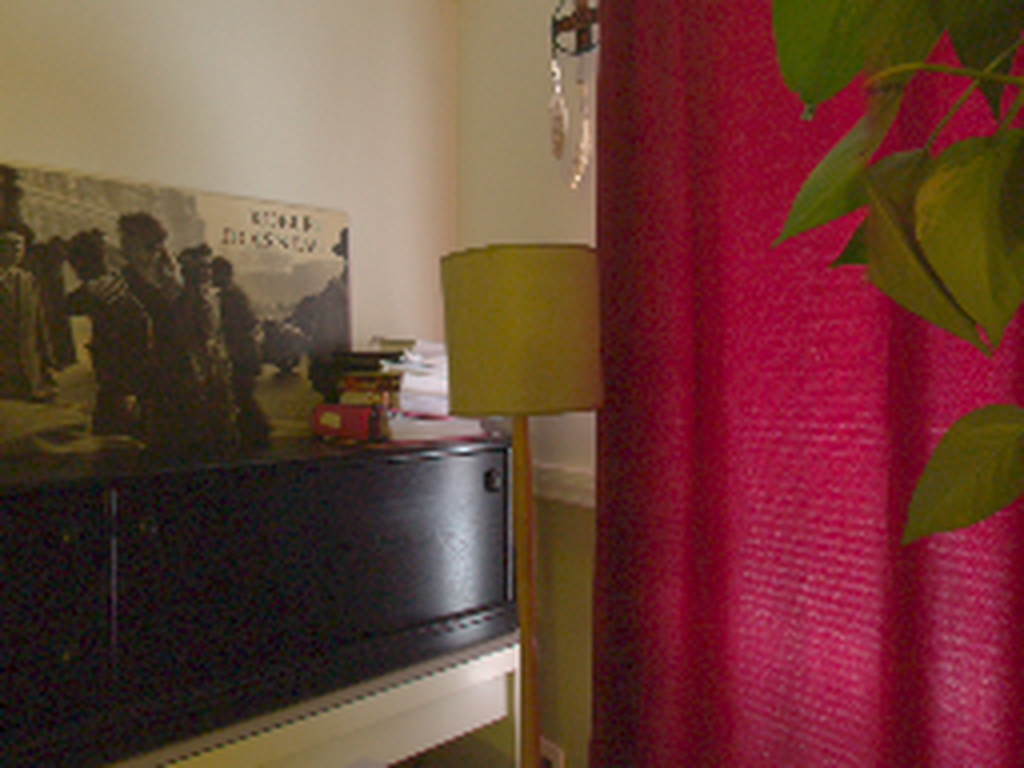}} &
        \noindent\parbox[c]{0.071\textwidth}{\includegraphics[width=0.071\textwidth]{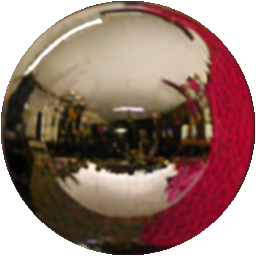}} & 
        \noindent\parbox[c]{0.071\textwidth}{\includegraphics[width=0.071\textwidth]{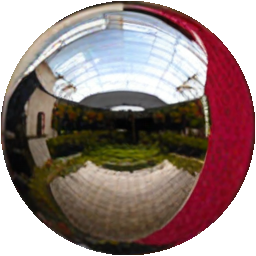}} & 
        \noindent\parbox[c]{0.090\textwidth}{\includegraphics[width=0.090\textwidth]{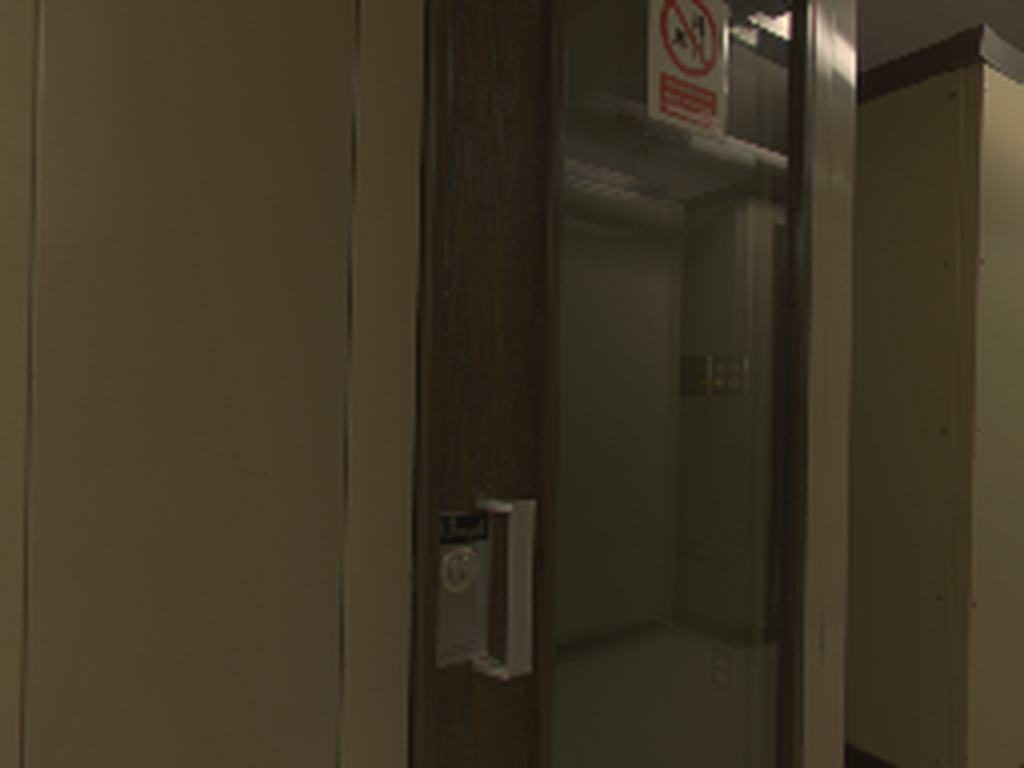}} &
        \noindent\parbox[c]{0.071\textwidth}{\includegraphics[width=0.071\textwidth]{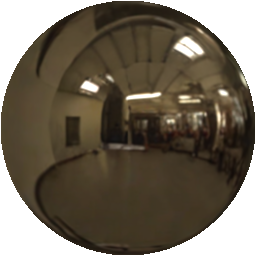}} & 
        \noindent\parbox[c]{0.071\textwidth}{\includegraphics[width=0.071\textwidth]{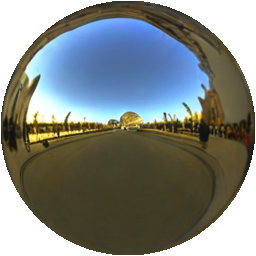}} & 
        \\
        \noindent\parbox[c]{0.090\textwidth}{\includegraphics[width=0.090\textwidth]{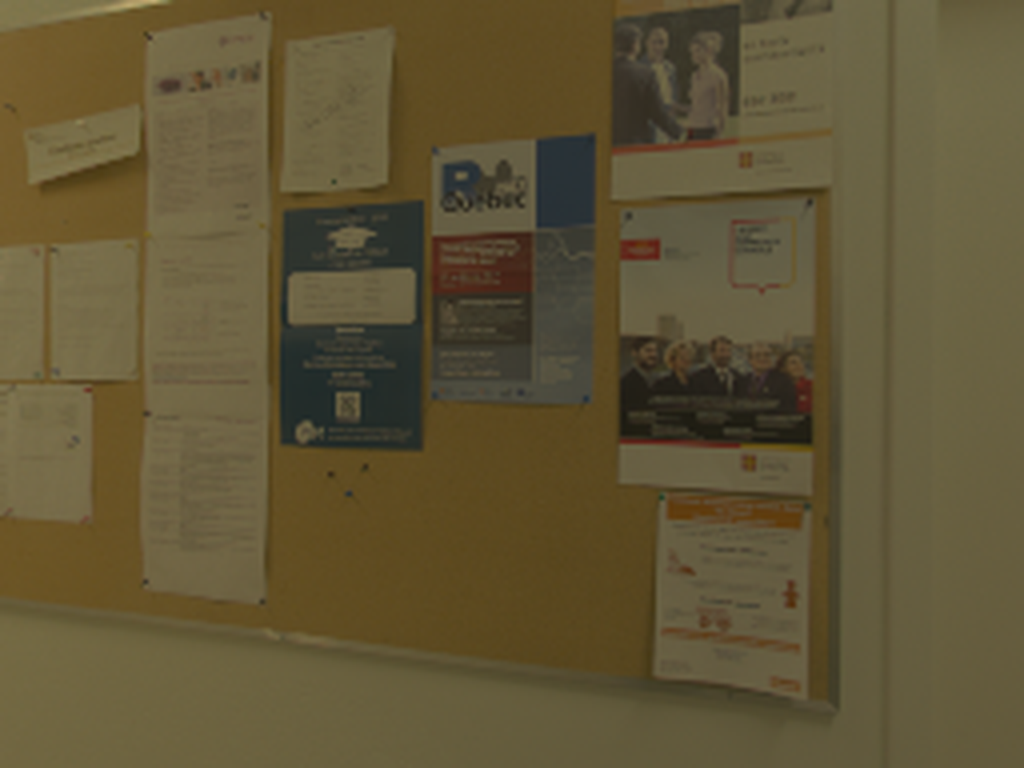}} &
        \noindent\parbox[c]{0.071\textwidth}{\includegraphics[width=0.071\textwidth]{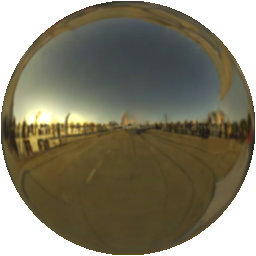}} & 
        \noindent\parbox[c]{0.071\textwidth}{\includegraphics[width=0.071\textwidth]{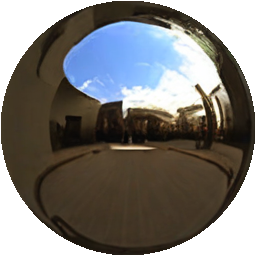}} & 
        \noindent\parbox[c]{0.090\textwidth}{\includegraphics[width=0.090\textwidth]{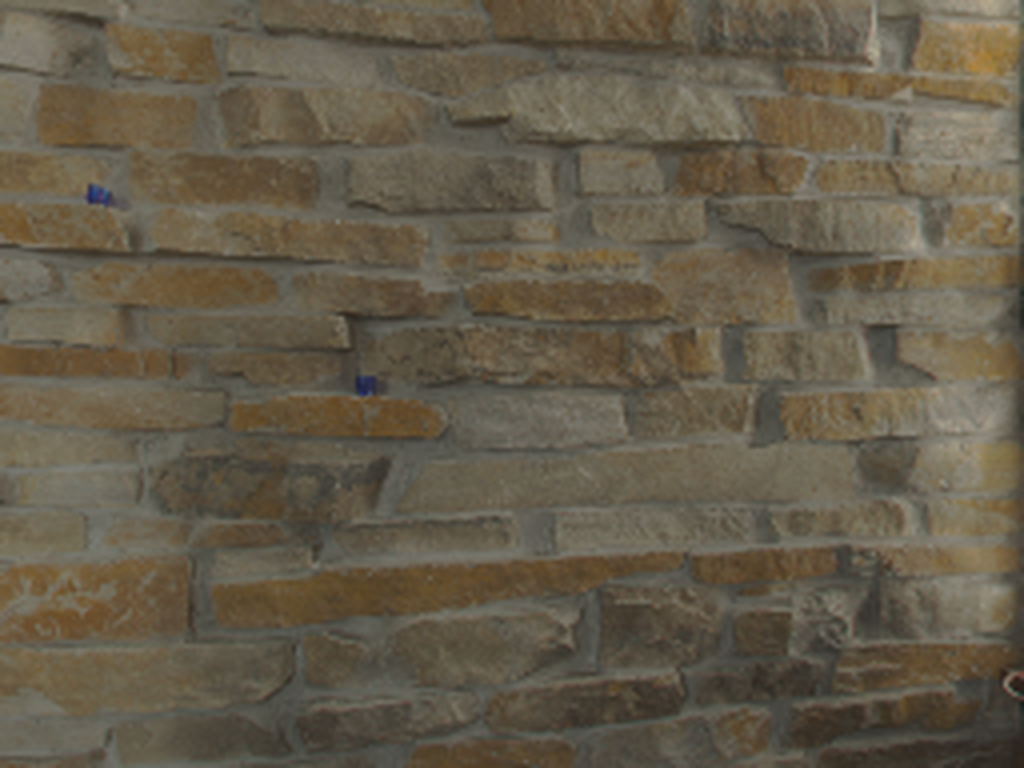}} &
        \noindent\parbox[c]{0.071\textwidth}{\includegraphics[width=0.071\textwidth]{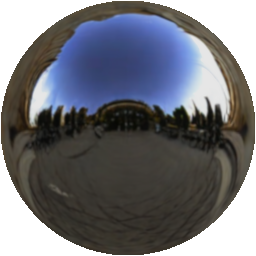}} & 
        \noindent\parbox[c]{0.071\textwidth}{\includegraphics[width=0.071\textwidth]{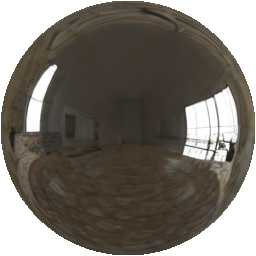}} 
        
    \end{tabu}
    \caption{Our methods occasionally hallucinate outdoor scenes (e.g., inpainting a chrome ball reflecting the sky) on indoor images, particularly when visual cues are limited. This issue is more pronounced in DiffusionLight-Turbo, but fine-tuning Turbo LoRA on more scenes with limited visual cues can help mitigate it.
    }
    \label{fig:sky_artifact}
\end{figure}

As shown in Table \ref{tab:indoor_everlight}, DiffusionLight ranks behind Weber et al. and EMLight but outperforms both EverLight and StyleLight, despite not being explicitly trained on the dataset. 
DiffusionLight-Turbo shows a slight drop in si-RMSE compared to DiffusionLight but remains competitive with EMLight \cite{zhan2021emlight}. We observe that this performance drop is partly due to DiffusionLight-Turbo’s tendency to hallucinate outdoor reflections in indoor scenes, especially when the input images lack strong visual cues (see Figure \ref{fig:sky_artifact}).


\begin{figure*}[ht]
    \centering
    \begin{tabu} to \textwidth {
        @{}
        c@{\hspace{1pt}}
        c@{\hspace{1pt}}
        c@{\hspace{1pt}}
        c@{\hspace{1pt}}
        c@{\hspace{1pt}}
        c@{\hspace{1pt}}
        c@{\hspace{1pt}}
        c@{}
    }   
        \noindent\parbox[c]{0.15\textwidth}{\centering \scriptsize \textbf{StyleLight}} &
        \multicolumn{3}{c}{
             \centering \scriptsize \textbf{DiffusionLight (Ours)}
        } &
        \multicolumn{3}{c}{
             \centering \scriptsize \textbf{DiffusionLight-Turbo (Ours)}
        } &
        \\
        \noindent\parbox[c]{0.15\textwidth}{\centering \scriptsize EV0} &
        \noindent\parbox[c]{0.16\textwidth}{\centering \scriptsize  Inpainted Image} &
        \noindent\parbox[c]{0.14\textwidth}{\centering \scriptsize  EV0} &
        \noindent\parbox[c]{0.1\textwidth}{\centering \scriptsize Underexposed} &
        \noindent\parbox[c]{0.21\textwidth}{\centering \scriptsize Inpainted Image} &
        \noindent\parbox[c]{0.072\textwidth}{\centering \scriptsize  EV0} &
        \noindent\parbox[c]{0.072\textwidth}{\centering \scriptsize Underexposed} &
        \\
        
        \multicolumn{7}{c}{
        \noindent\parbox[c]{0.98\textwidth}{\includegraphics[width=0.98\textwidth]{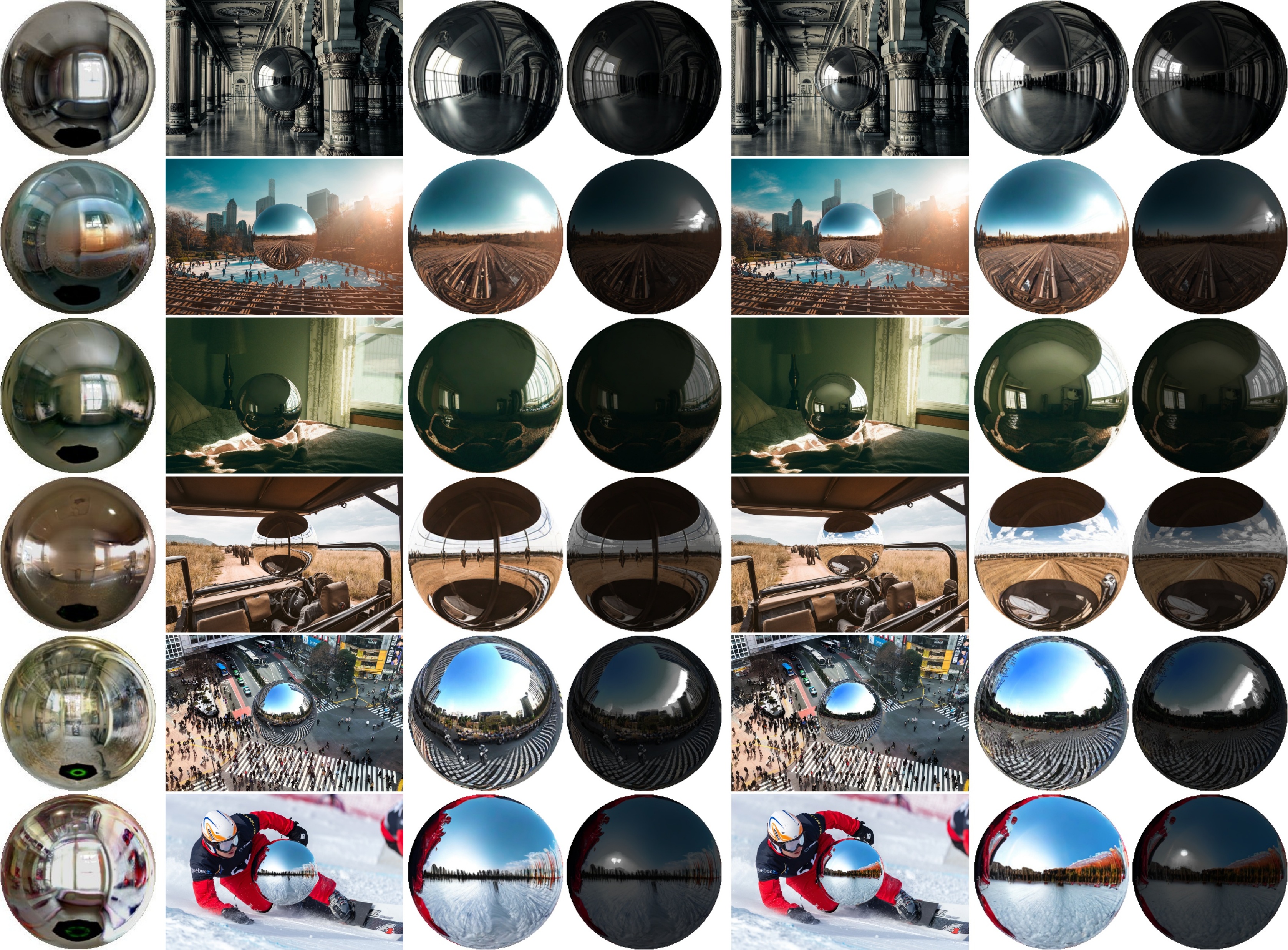}}    
        }
        
    \end{tabu}
        
        \caption{Qualitative results for in-the-wild scenes. In each scene, we show chrome balls with normal exposure generated by our pipeline, along with their underexposed versions, rendered from HDR chrome balls. DiffusionLight-Turbo produces results comparable to DiffusionLight with a 60x speedup. }
    \label{fig:main_qualitative_wild}
\end{figure*}

\tabulinesep=0.5pt


\begin{figure}[h!]
    \centering
    \begin{tabular}{@{}c@{\hspace{0.5em}}c@{}} 
        \begin{tabular}{@{}c@{}} 
            \includegraphics[width=0.11\textwidth]{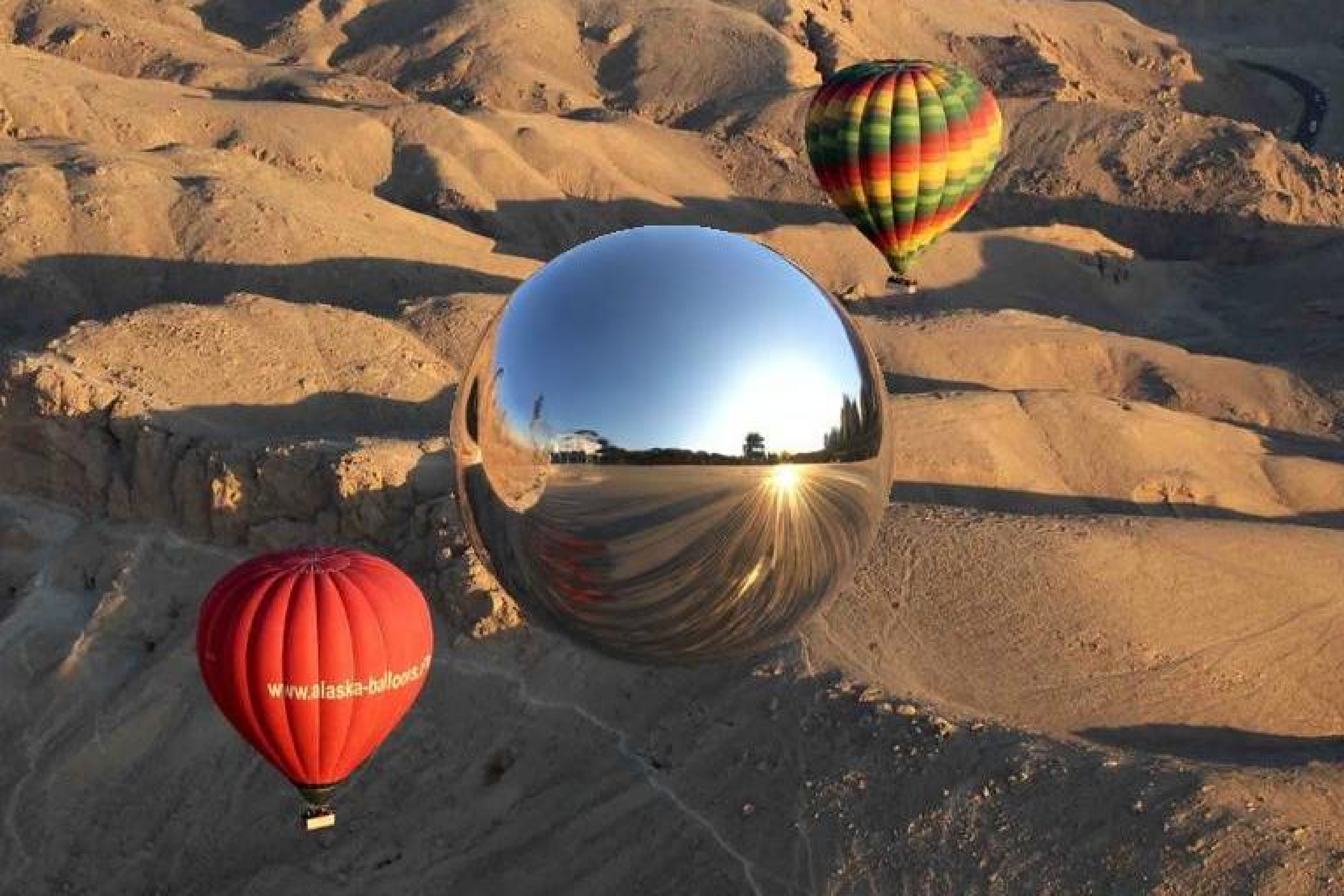} 
            \includegraphics[width=0.11\textwidth]{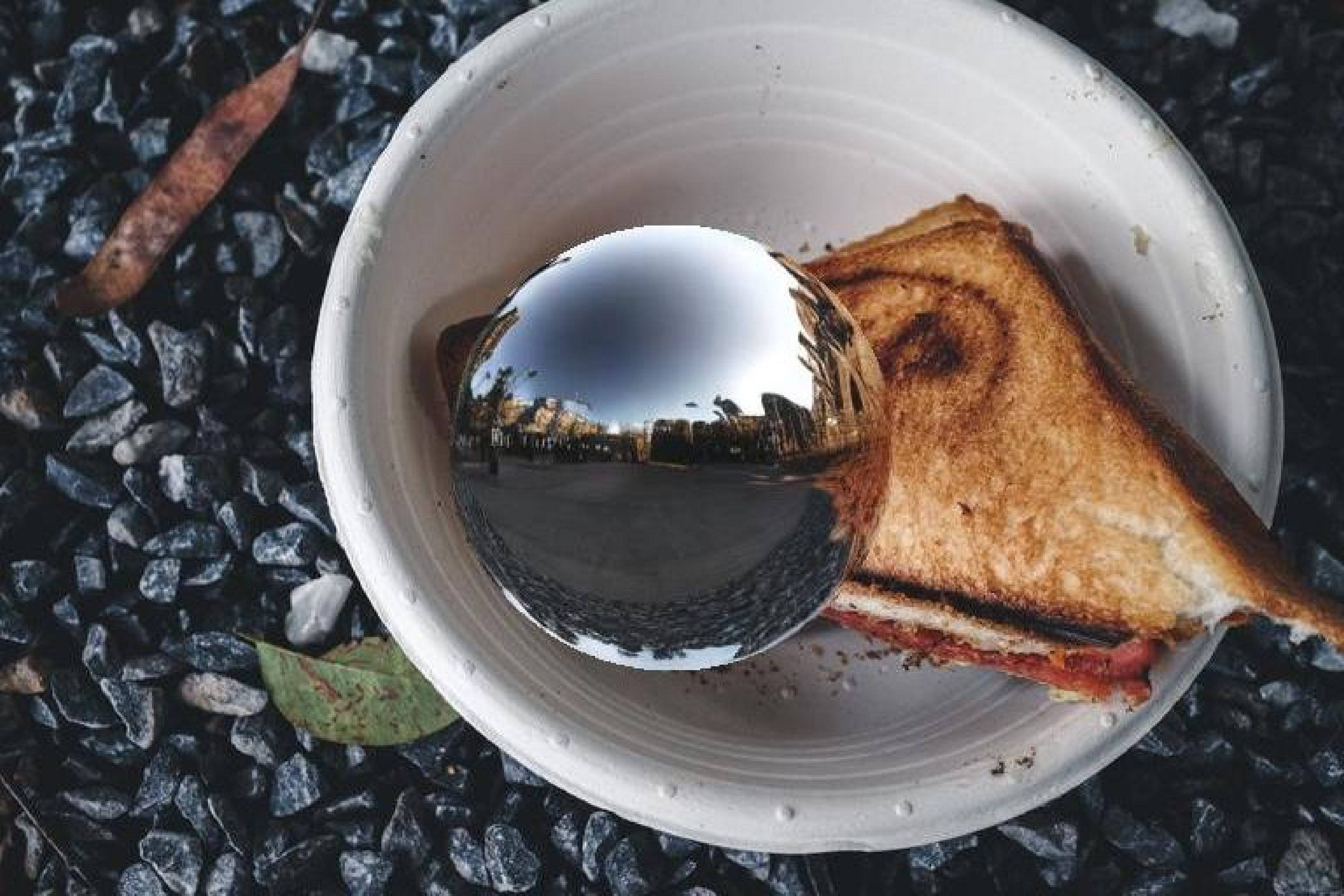} \\
            \noindent\parbox[c]{0.22\textwidth}{\centering \scriptsize (a) Overhead and bird’s-eye view images}             
        \end{tabular}
        & 
        \begin{tabular}{@{}c@{}} 
            \includegraphics[width=0.11\textwidth]{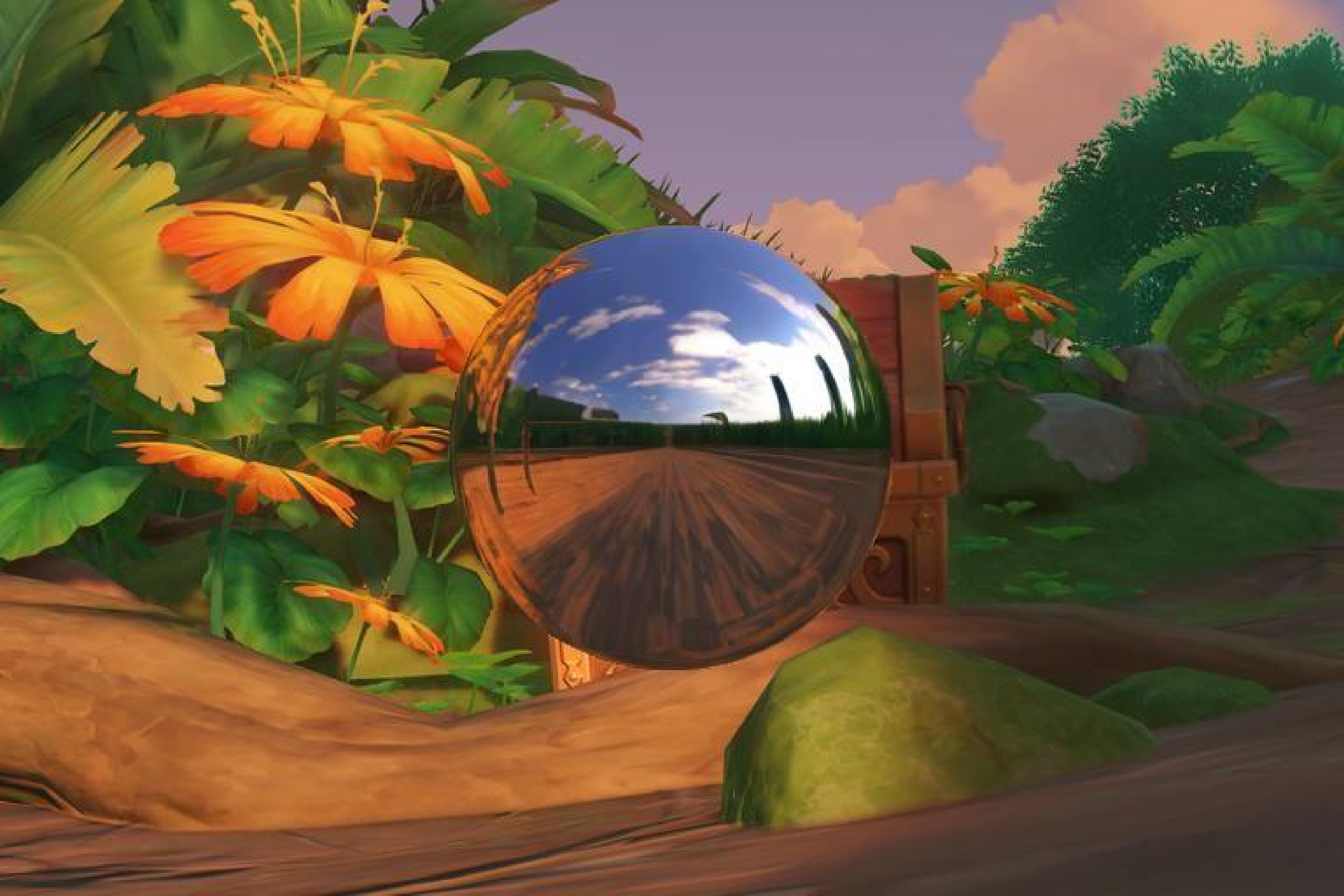}
            \includegraphics[width=0.11\textwidth]{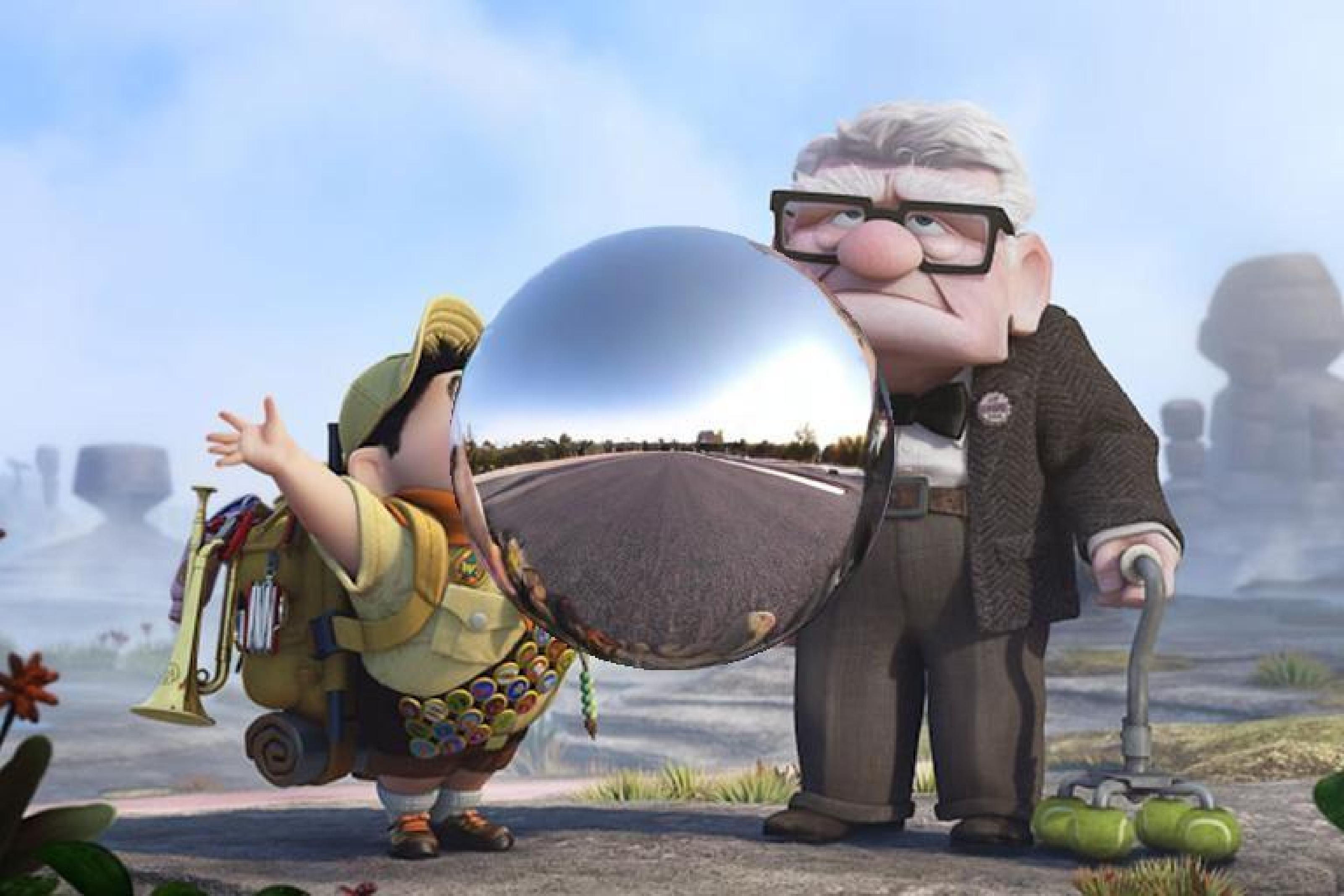} \\
            \noindent\parbox[c]{0.22\textwidth}{\centering \scriptsize (b) Images with significant style difference from natural photos} 
            
        \end{tabular}
    \end{tabular}

    \vspace{0.5em} 

    \begin{tabular}{@{}c@{\hspace{0.5em}}c@{}}
        \begin{tabular}{@{}c@{}}
            \includegraphics[width=0.11\textwidth]{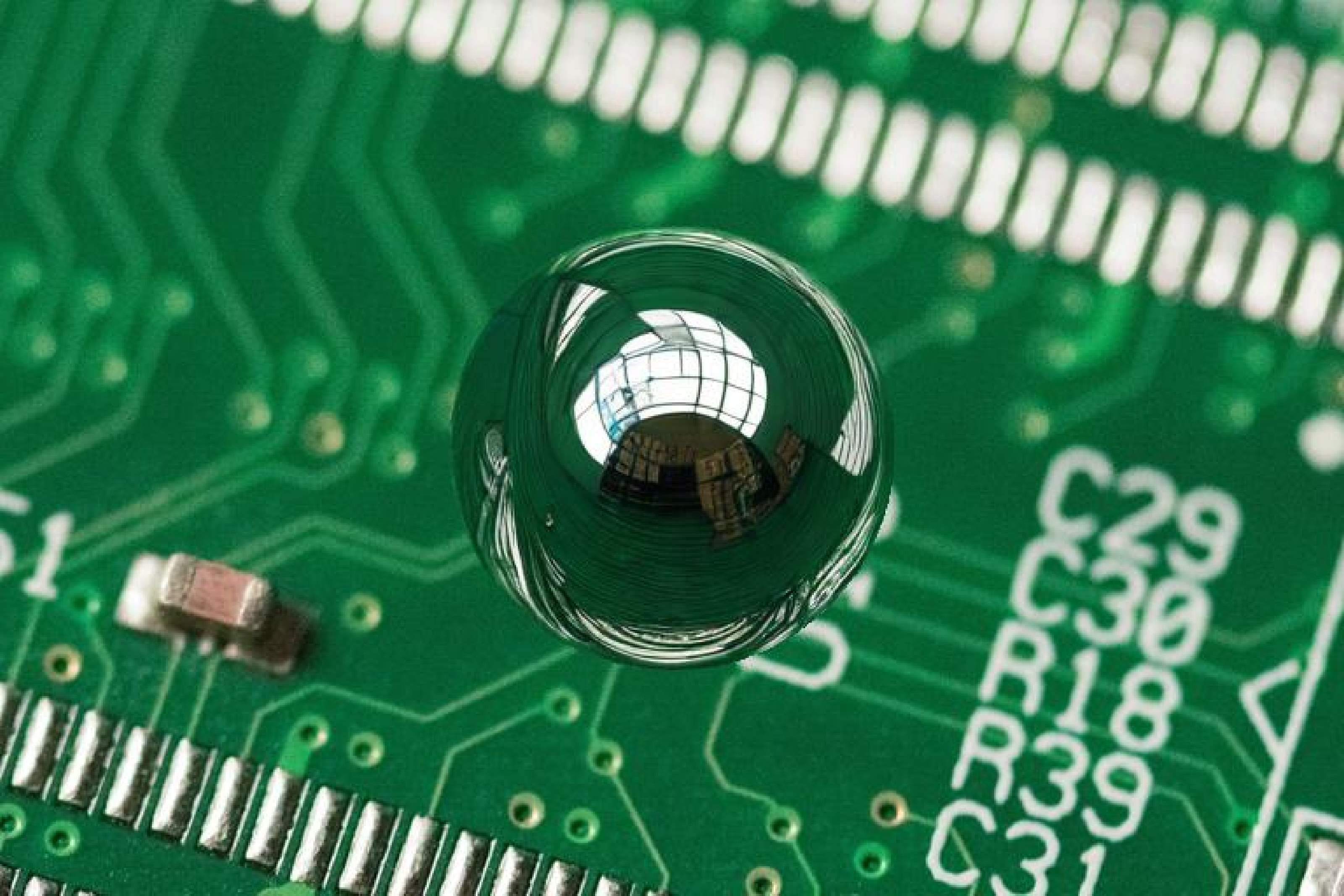}
            \includegraphics[width=0.11\textwidth]{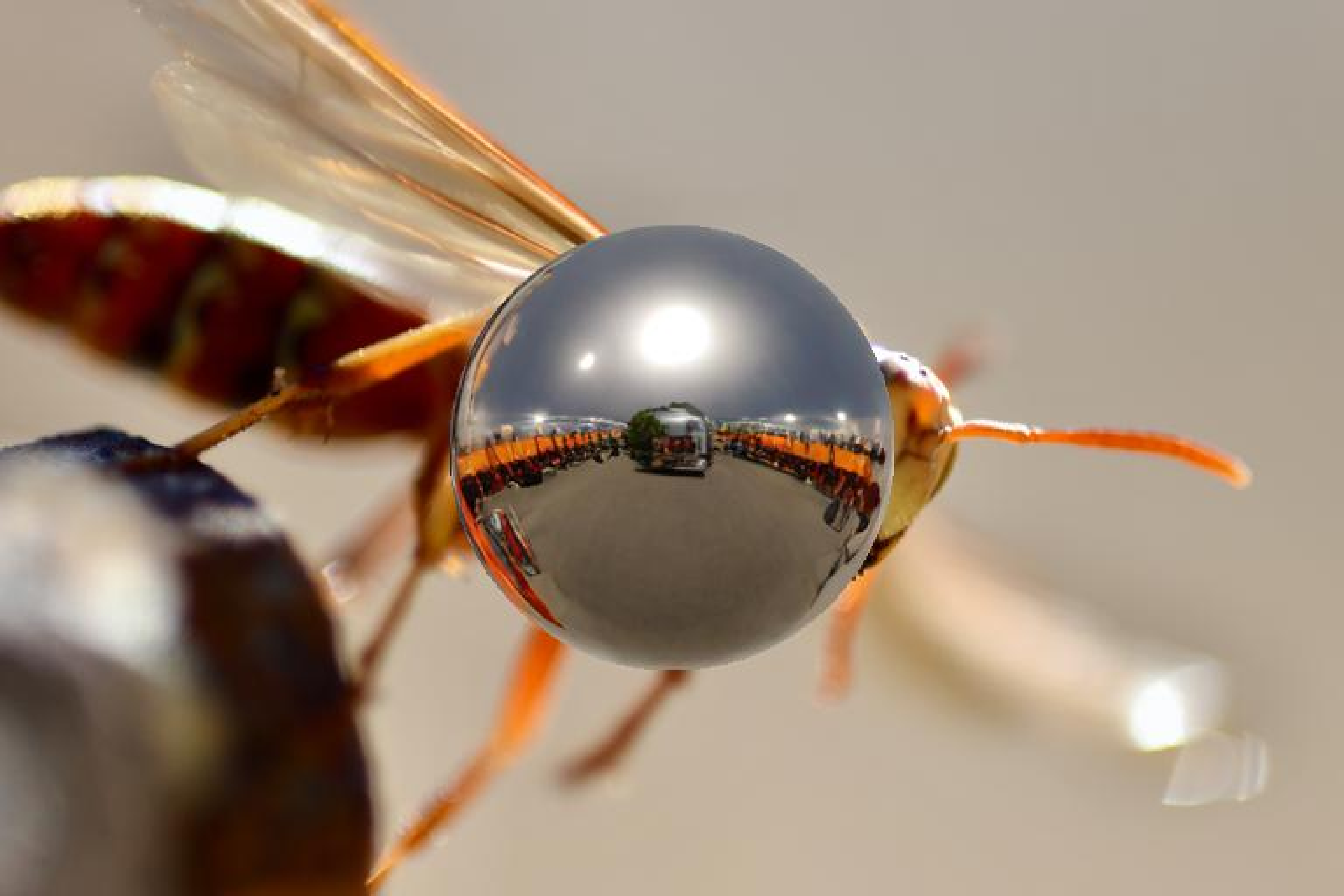} \\
            \noindent\parbox[c]{0.22\textwidth}{\centering \scriptsize (c) Macro and close-up images}
        \end{tabular}
        & 
        \begin{tabular}{@{}c@{}}
            \includegraphics[width=0.11\textwidth]{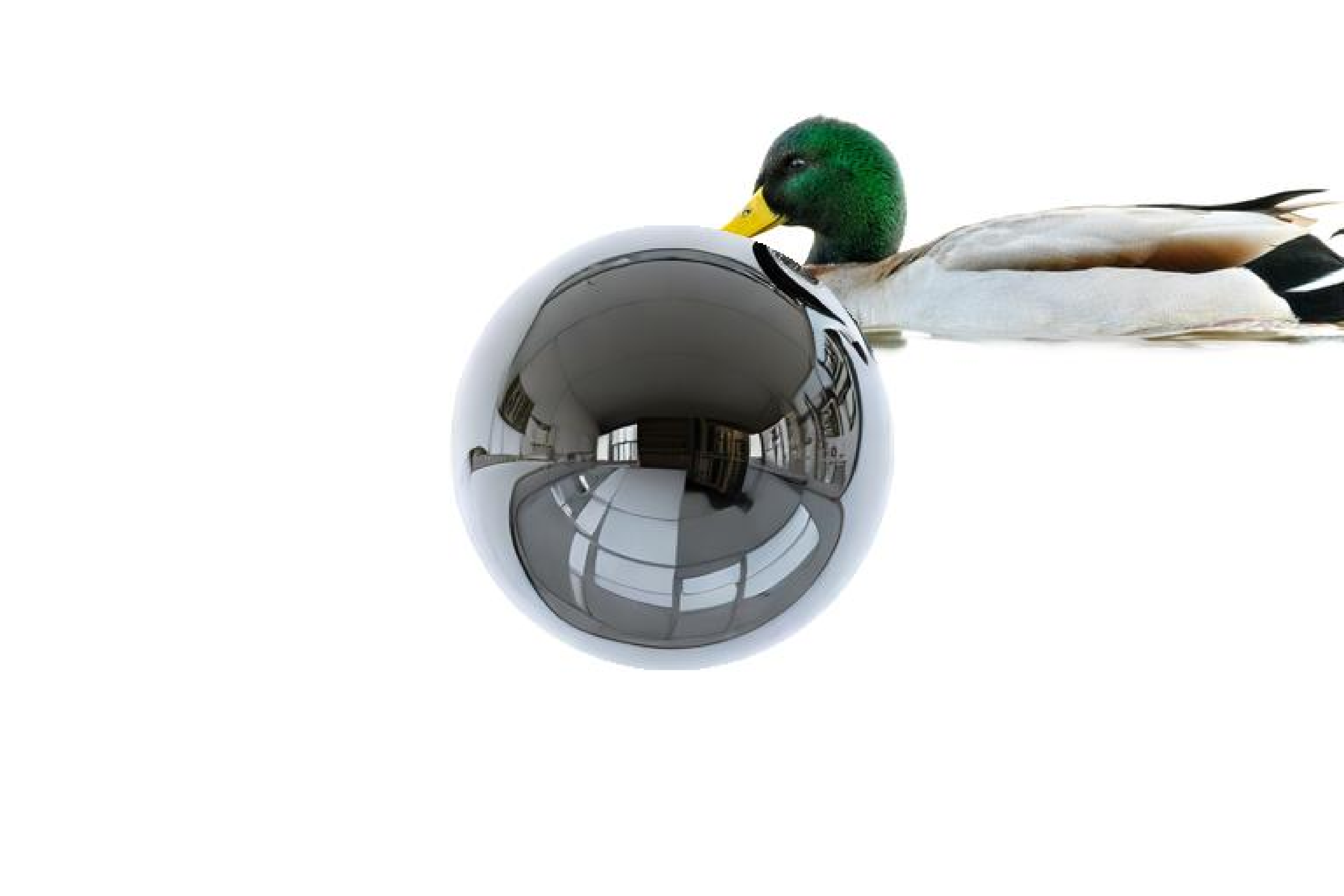}
            \includegraphics[width=0.11\textwidth]{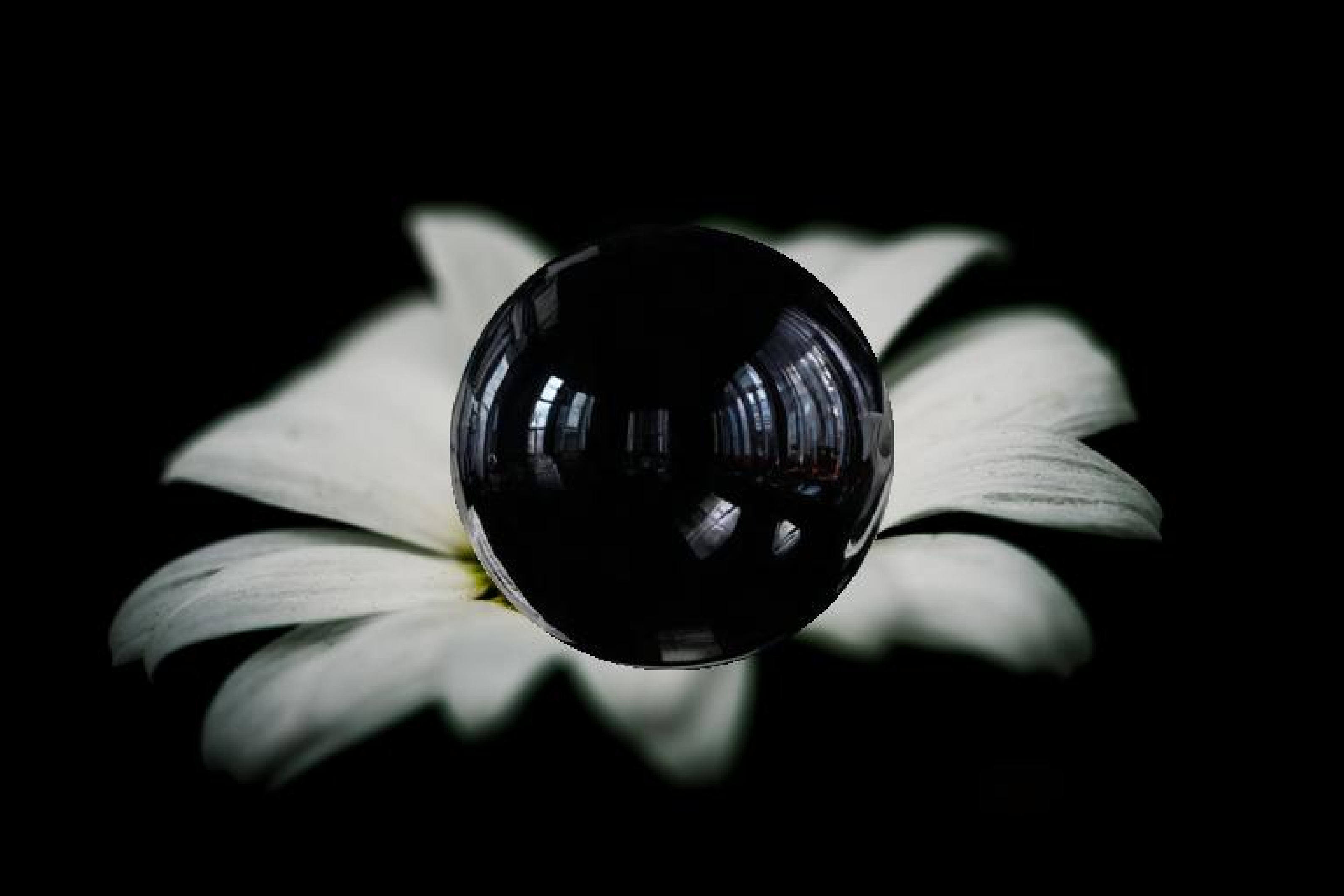} \\
            \noindent\parbox[c]{0.22\textwidth}{\centering \scriptsize (d) High-key and low-key images with solid backgrounds}
        \end{tabular}
    \end{tabular}
    \caption{\textbf{Failure modes}: (a) Our method may produce unrealistic chrome balls from overhead and bird's-eye view images, distorting the horizon line's curvature. (b) The chrome balls may not harmonize well with inputs whose styles differ significantly from natural photos. (c) Macro and close-up images may lack sufficient shading cues, leading to less convincing estimates. (d) Images with empty or solid backgrounds often cause our method to hallucinate some details onto the balls, and the balls may appear too dark, especially on a white background. \ }
    \label{fig:failure_modes}
\end{figure}



\tabulinesep=0.5pt


\begin{figure}[!h]
    \centering
    \includegraphics[width=0.32\columnwidth]{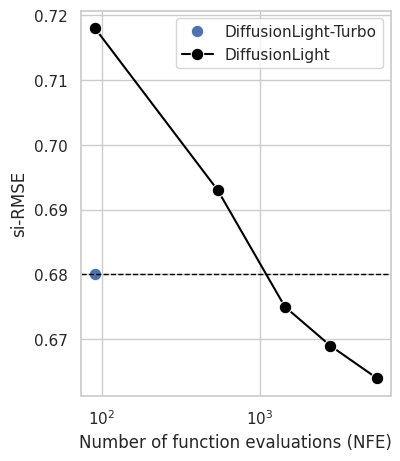}
    \includegraphics[width=0.32\columnwidth]{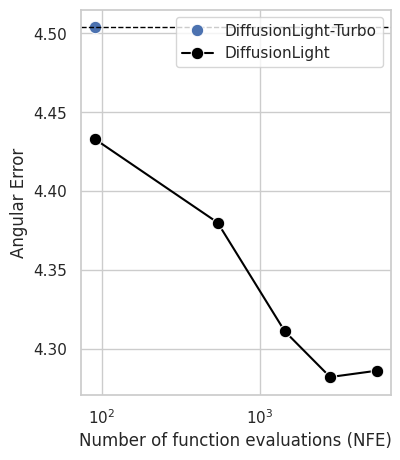}
    \includegraphics[width=0.32\columnwidth]{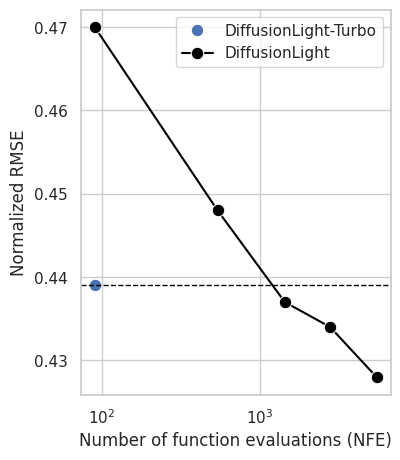}
    \caption{DiffusionLight-Turbo provides a better trade-off between performance and runtime compared to DiffusionLight on si-RMSE and Normalized RMSE.}

    \label{fig:tradeoff}
\end{figure}



\subsubsection{Evaluation on unknown camera parameters} Evaluation protocols in the last section crop HDR panoramas at fixed camera angles and FOVs. In real-world settings, however, we often do not know the camera parameters of a photograph. This evaluation considers more challenging scenarios that reflect this situation better. In particular, to generate an input LDR image, we randomly sample the FOV from the interval $[30^\circ, 150^\circ]$, the elevation from $[-45^\circ, 45^\circ]$, and the azimuth from all $360^\circ$. We then crop an HDR panorama accordingly. We generate one LDR image from 289 HDR panoramas of the Laval Indoor dataset and compare our method with StyleLight using the generated input-output pairs and the three-sphere protocol. Table \ref{tab:randfov} shows that both DiffusionLight and DiffusionLight-Turbo outperform StyleLight in Angular Error and Normalized RMSE and remain competitive in si-RMSE. 

\subsection{Qualitative results for in-the-wild scenes}
We present additional qualitative results on diverse in-the-wild scenes sourced from Unsplash (www.unsplash.com) and other websites under a CC4.0 license in Figure \ref{fig:main_qualitative_wild} and Appendix \ref{appendix:more_result_wild}.
Compared to existing techniques, our methods produce chrome balls that ``reflect'' elements present in the input image, such as the car's ceiling, the zebra crossing, and the red garment of the snowboarder, as well as recover overexposed details like window frames and the sun (see also Figure \ref{fig:teaser}). 




\begin{figure*}[t]
    \centering

        \begin{tabu} to \columnwidth {
        @{}
        c@{}
        c@{}
        c@{}
        c@{}
    }
\noindent\parbox[c]{0.22\textwidth}{\centering Image} &
\noindent\parbox[c]{0.22\textwidth}{\centering No Lighting} &
\noindent\parbox[c]{0.22\textwidth}{\centering DiffusionLight} &
\noindent\parbox[c]{0.22\textwidth}{\centering DiffusionLight-Turbo} \\
    \noindent\parbox[c]{0.22\textwidth}{\includegraphics[width=0.22\textwidth]{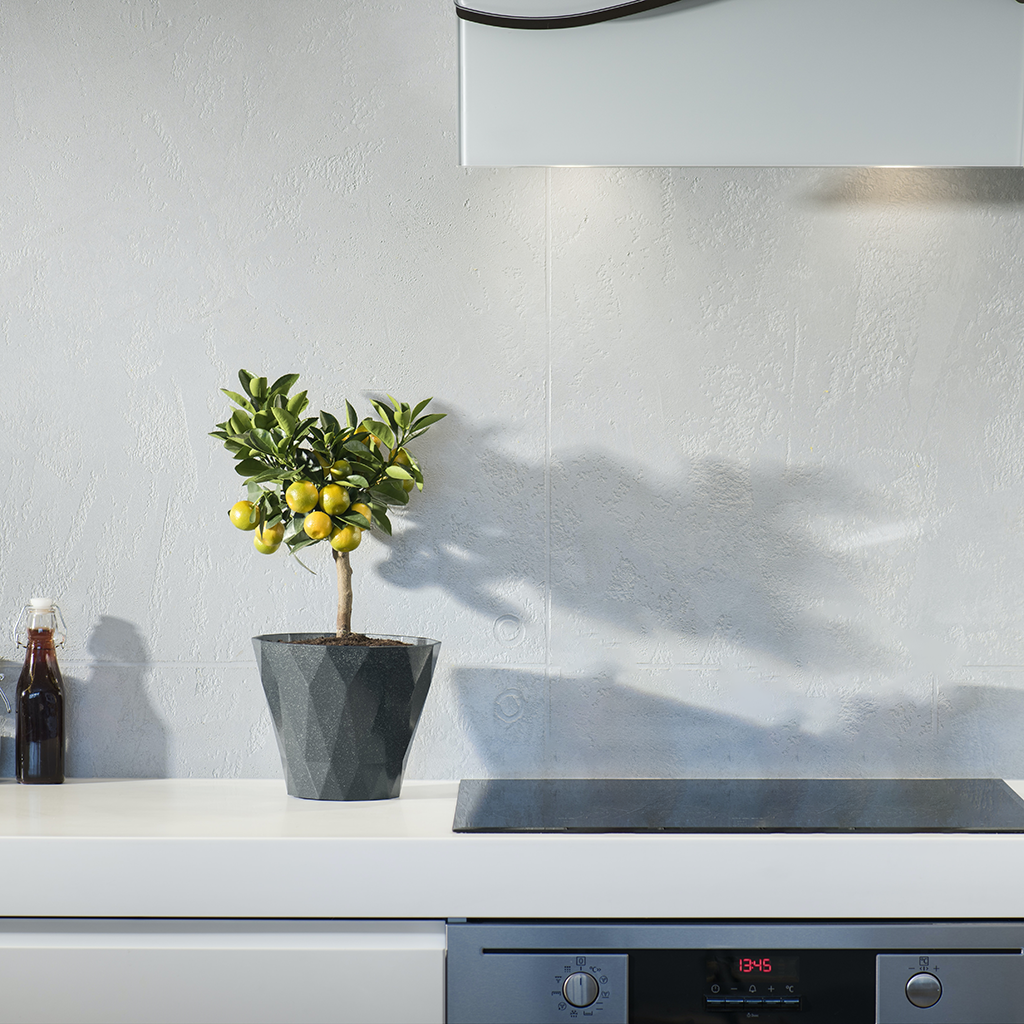}} &
    \noindent\parbox[c]{0.22\textwidth}{\includegraphics[width=0.22\textwidth]{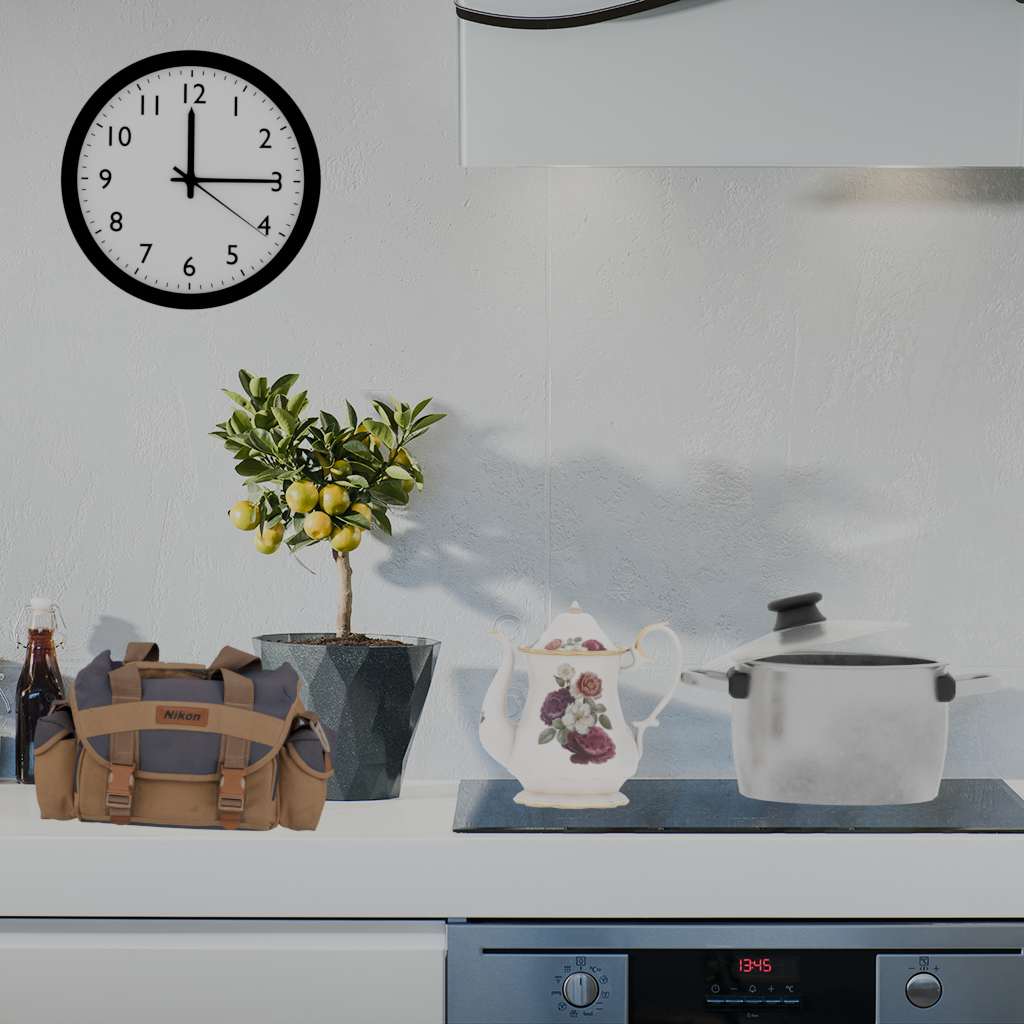}} &
    \noindent\parbox[c]{0.22\textwidth}{\includegraphics[width=0.22\textwidth]{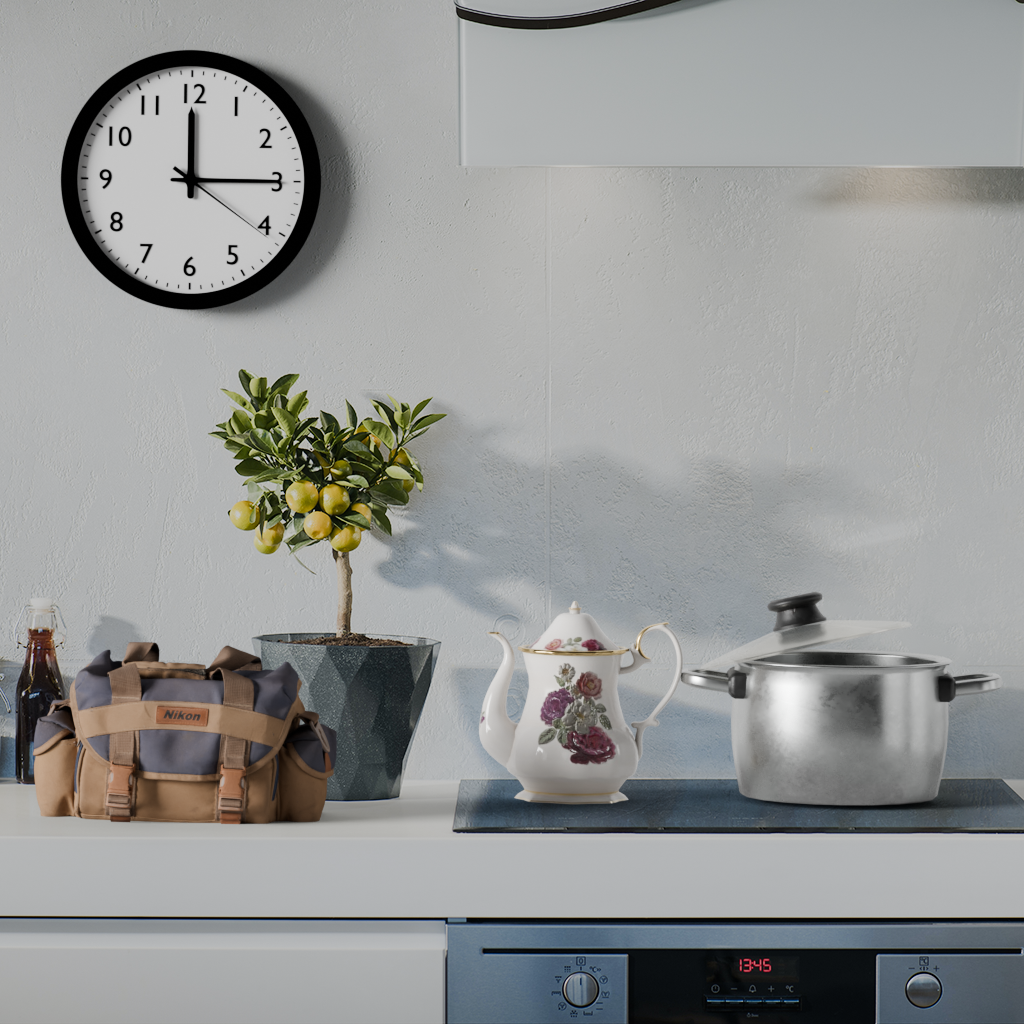}} &
    \noindent\parbox[c]{0.22\textwidth}{\includegraphics[width=0.22\textwidth]{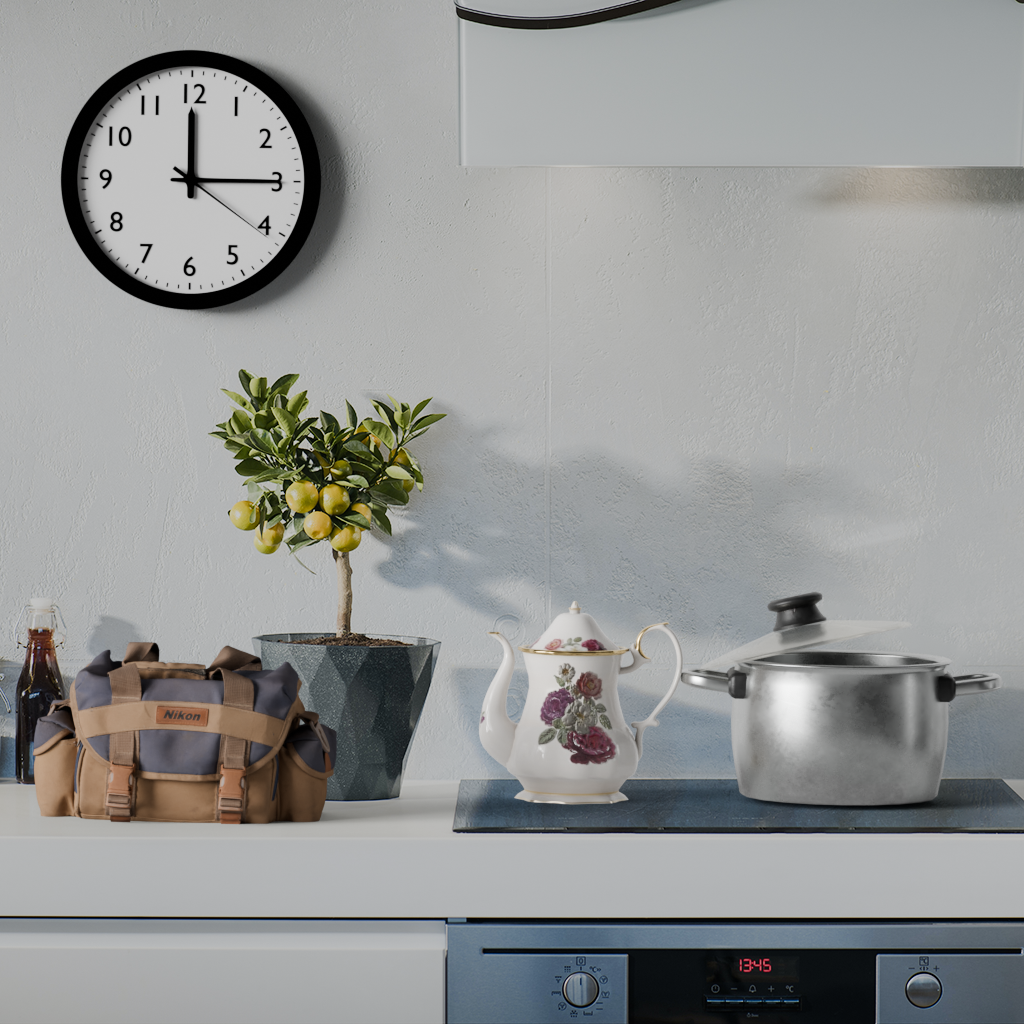}}
    \end{tabu}
    \caption{
    We synthetically render each 3D object into input images using our estimated lighting.
    }
    \label{fig:additional_object_insertion}
\end{figure*}

\subsection{Ablation studies}
\subsubsection{DiffusionLight} We perform ablation studies on our iterative inpainting and LoRA using Laval Indoor and Poly Haven datasets.
Table \ref{tab:indoor_stylelight} shows that our full method
surpasses all ablated versions on all metrics except for Angular Error on matte balls in Poly Haven. Ablations on the size and number of balls are in Appendix \ref{appendix:aba_median}. 
Ablations on LoRA scale and timesteps are in Appendix \ref{appendix:aba_lora}.

\subsubsection{DiffusionLight-Turbo} \label{sec:abla_diffusion_turbo} We conduct an ablation study on our LoRA swapping technique (Section \ref{sec:fast_diffusionlight}) using the Laval Indoor and Poly Haven datasets. As shown in Table \ref{tab:aba_diffusionlight_turbo}, swapping LoRA at inference time (Turbo-Swapping) yields performance competitive with predicting the median then performing SDEdit (Turbo-SDEdit) while using 2.30x less runtime. An attempt to accelerate Turbo-SDEdit by directly predicting clean samples (Turbo-Pred) performs the worst, as it is prone to introducing additional errors when computing Equation \ref{eq:predx0}. Additional ablations on Turbo LoRA training are provided in Appendix \ref{appendix:aba_turbo_lora}.

\subsubsection{Trade-off between runtime and quality.} \label{sec:abla_running_time_tradeoff} 
\begin{figure}[!h]
    \centering
    \begin{tabu} to \textwidth {
        @{\hspace{3.0pt}}
        c@{\hspace{0.2pt}} 
        c@{\hspace{0.6pt}}
        c@{\hspace{0.2pt}}
        c@{\hspace{0.6pt}}
        c@{\hspace{0.2pt}}
        c@{\hspace{0.2pt}}
        @{}
    }
    \multicolumn{1}{c}{\footnotesize $k$=1} &
    \multicolumn{1}{c}{\footnotesize $k$=2} &
    \multicolumn{1}{c}{\footnotesize $k$=1} &
    \multicolumn{1}{c}{\footnotesize $k$=2} &
    \multicolumn{1}{c}{\footnotesize $k$=1} &
    \multicolumn{1}{c}{\footnotesize $k$=2}
    \\
    
    \noindent\parbox[c]{0.07\textwidth}{\includegraphics[width=0.070\textwidth]{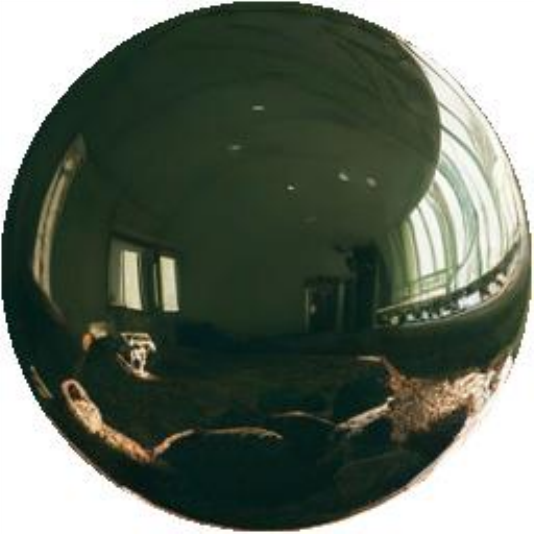}} & 
    \noindent\parbox[c]{0.07\textwidth}{\includegraphics[width=0.070\textwidth]{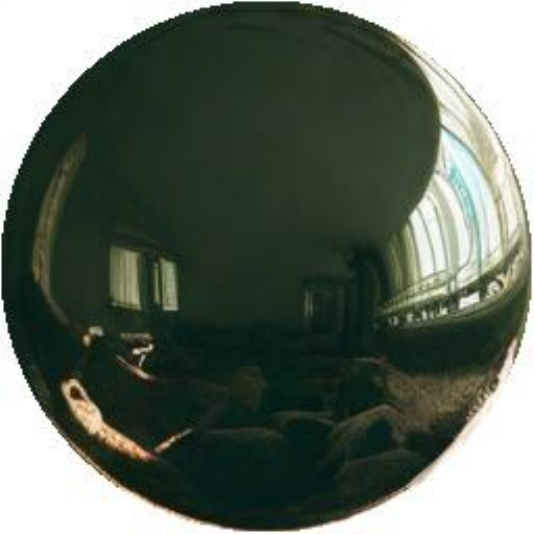}} & 
    \noindent\parbox[c]{0.07\textwidth}{\includegraphics[width=0.070\textwidth]{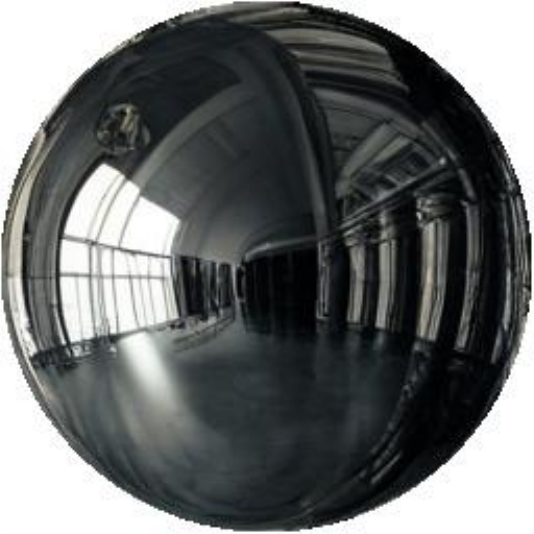}} & 
    \noindent\parbox[c]{0.07\textwidth}{\includegraphics[width=0.070\textwidth]{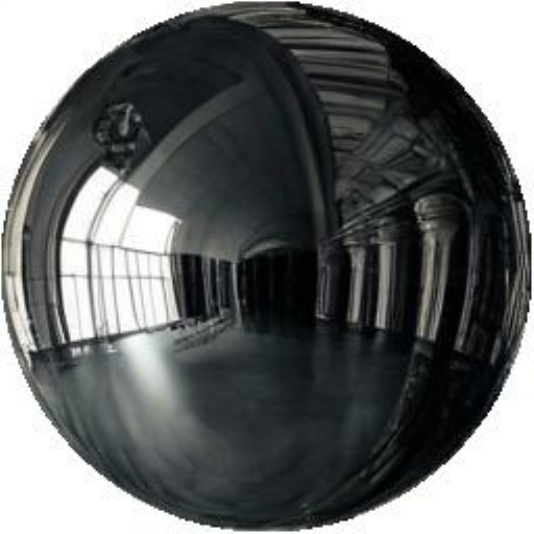}} &
    \noindent\parbox[c]{0.07\textwidth}{\includegraphics[width=0.070\textwidth]{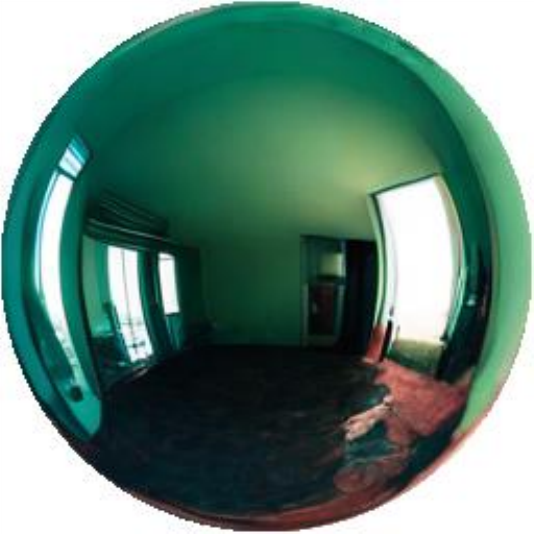}} & 
    \noindent\parbox[c]{0.07\textwidth}{\includegraphics[width=0.070\textwidth]{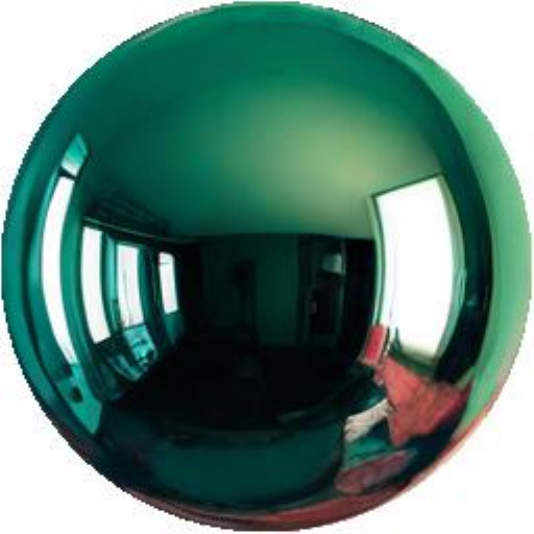}}
    \\
    
    \multicolumn{6}{c}{\footnotesize Input \#1 ($N$=30) \ \ \ \ \ \ \ Input \#2  ($N$=30) \ \ \ \ \ \ \ Input \#3  ($N$=30)} 
    \\
    
    \end{tabu}
    
    \caption{Qualitative results for the ablation study in Table \ref{tab:plot_time_accuracy}. We show results for $(k=1, N=30)$ and $(k=2, N=30)$. We can halve the runtime with minimal quality degradation by reducing the number of iterations to one. } 

    \label{fig:ablation_inpainting_ball_iteration}
\end{figure}

We experimented with various numbers of balls $N$ and iterations $k$ in the iterative inpainting algorithm in DiffusionLight on the Poly Haven dataset \cite{polyhaven}. As shown in Table \ref{tab:plot_time_accuracy}, increasing either parameter generally improves accuracy. Notably, while two iterations of iterative inpainting ($k=2$) deliver the best performance, reducing to one iteration ($k=1$) halves the runtime with minimal quality degradation as shown in Figure \ref{fig:ablation_inpainting_ball_iteration}. 
Compared to DiffusionLight, DiffusionLight-Turbo offers a better quality-runtime trade-off, achieving a competitive performance in terms of si-RMSE and Normalized RMSE while requiring over 10x fewer NFEs.

\begin{table}[!h]
\centering
\small

\caption{Ablation study on the number of iterations $k$ and balls $N$. We report number of function evaluations (NFE) and scores on mirror balls in the Poly Haven dataset \cite{polyhaven}. See Figure \ref{fig:ablation_inpainting_ball_iteration} for qualitative results. } 
 \label{tab:plot_time_accuracy}
\begin{tabular}{
    c@{\hspace{4pt}}
    c@{\hspace{4pt}}
    c@{\hspace{4pt}}
    c@{\hspace{4pt}}
    c@{\hspace{4pt}}
    c
}
\toprule
\textbf{\begin{tabular}[c]{@{}c@{}}\textbf{\#Iterations}\\ \textbf{($k$)}\end{tabular}} & \textbf{\begin{tabular}[c]{@{}c@{}}\textbf{\#Balls}\\ \textbf{($N$)}\end{tabular}} & \textbf{\begin{tabular}[c]{@{}c@{}}\textbf{NFE}\end{tabular}} & \textbf{si-RMSE} $\downarrow$ & \begin{tabular}[c]{@{}c@{}}\textbf{Angular}\\ \textbf{Error}\end{tabular} $\downarrow$ & \begin{tabular}[c]{@{}c@{}}\textbf{Normalized}\\ \textbf{RMSE}\end{tabular} $\downarrow$ \\
\midrule
1 & 5  & 540 & 0.693 & 4.380 & 0.448 \\
  & 15 & 1440 & 0.675 & 4.311 & 0.437 \\
  & 30 & 2790  & 0.669 & 4.282 & 0.434 \\
\midrule
2 & 5  & 990  & 0.694 & 4.435 & 0.448 \\
  & 15 & 2790 & 0.672 & 4.325 & 0.432 \\
  & 30 & 5490 & 0.664 & 4.286 & 0.428 \\
\bottomrule
\end{tabular}
\end{table}


\tabulinesep=2pt




\section{Limitations and Discussion}
\label{sec:limitation}

Given the absence of focal length or FOV information, we assume an orthographic projection when converting a chrome ball to an environment map. However, this may not accurately reflect the projection model used by the diffusion model. Additionally, our reliance on an off-the-shelf depth predictor may lead to inaccurate lighting estimation in cases of incorrect depth on reflective, textureless, or overexposed surfaces.
While our methods generalize well to diverse in-the-wild scenes, they may struggle with certain input images, including: (1) overhead and bird's-eye view images, (2) macro and close-up images, (3) high-key or low-key images with solid backgrounds, and (4) images with styles that differ significantly from natural photos, such as cartoons or Japanese-style animations. Figure \ref{fig:failure_modes} illustrates these failure cases. 
Switching to a fine-tuned model, such as AnimagineXL\footnote{\url{https://huggingface.co/Linaqruf/animagine-xl}}, to leverage a more specialized image prior can improve performance on specific image styles.

While DiffusionLight-Turbo accelerates the chrome ball inpainting by reducing it to a single denoising process, it occasionally hallucinates outdoor scenes on indoor inputs when visual cues are limited (see Figure \ref{fig:sky_artifact}). Fine-tuning the Turbo LoRA on a larger and more diverse in-the-wild dataset could further improve its performance.

Despite these limitations, our method is applicable to downstream tasks such as virtual object insertion, which requires light estimation to seamlessly render virtual objects into the scene.
In Figure \ref{fig:additional_object_insertion}, we present qualitative results for four objects, rendered using HDR environment maps obtained through both DiffusionLight and DiffusionLight-Turbo.

\section{Conclusion}
\label{sec:conclusion}


This paper presents a novel HDR light estimation approach that inpaints a chrome ball into the scene using a pretrained LDR diffusion model. To consistently render high-quality chrome balls, we propose an iterative algorithm to locate a suitable initial noise neighborhood and apply continuous LoRA fine-tuning across different exposures for bracketing and HDR chrome ball generation. 
Additionally, we introduce a second LoRA fine-tuning step that replicates the behavior of the iterative algorithm in a single denoising process, significantly accelerating chrome ball generation with minimal performance loss.
Our method performs competitively with the state of the art in both indoor and outdoor settings and marks the first work that achieves strong generalization to in-the-wild images.


\bibliographystyle{IEEEtran}
\bibliography{references}

\begin{IEEEbiographynophoto}{Worameth Chinchuthakun}
received the BEng and
MEng degrees from Tokyo Institute of Technology, Japan,
in 2022 and 2024, respectively. He is currently a Data Scientist at Siam Commercial Bank, Thailand. His research
interests are generative modeling and computer vision.
\end{IEEEbiographynophoto}

\begin{IEEEbiographynophoto}{Pakkapon Phongthawee}
received the BSc
degree from Silpakorn University, Thailand, in
2019. He is currently working toward the PhD
degree with the School of Information Science
and Technology, VISTEC, Thailand. His research
interests are view synthesis, neural rendering, light estimation and relighting.
\end{IEEEbiographynophoto}

\begin{IEEEbiographynophoto}{Amit Raj}
is a Research Scientist at Google DeepMind. He primarily works at the intersection of Computer Vision, Graphics and Machine Learning with a focus on diffusion models, video and 3D content authoring. He obtained his PhD in Machine Learning from Georgia Institute of Technology working with Dr. James Hays. He obtained his BTech from National Institute of Technology, Surathkal .
\end{IEEEbiographynophoto}

\begin{IEEEbiographynophoto}{Varun Jampani}
is Chief AI Officer at Arcade.AI. Previously, he was VP research at Stability AI and also held researcher positions at Google and Nvidia. He works in the areas of machine learning and computer vision and his main research interests include image, video and 3D generation. He obtained his PhD with highest honors at Max Planck Institute for Intelligent Systems (MPI) and University of Tübingen in Germany. He obtained his BTech and MS from International Institute of Information Technology, Hyderabad (IIIT-H), India, where he was a gold medalist. He actively contributes to the research community and regularly serves as area chair and reviewer for major computer vision and machine learning conferences. His works have received ‘Best Paper Honorable Mention’ award at CVPR’18 and 'Best Student Paper Honorable Mention' award at CVPR’23
\end{IEEEbiographynophoto}

\begin{IEEEbiographynophoto}{Pramook Khungurn} is a research follow at pixiv, Inc. He received undergraduate and master degrees from Massachusetts Institute of Technology and a PhD degree in computer science from Cornell University. He does research on artificial intelligence and its application to computer graphics.
\end{IEEEbiographynophoto}

\begin{IEEEbiographynophoto}{Supasorn Suwajanakorn}
received the BEng
degree from Cornell University, in 2011 and the PhD
degree from the University of Washington, in 2017.
He is currently a lecturer with the School of Information Science and Technology, VISTEC, Thailand. His
research interests lie in the intersection of computer
vision, deep learning, and computer graphics.
\end{IEEEbiographynophoto}


\pagebreak
\clearpage
\section*{Appendix} \label{appendix:implement}

\subsection{Inpainting algorithm}
The pseudocode of the iterative inpainting algorithm described in Section~\ref{sec:diffusionlight} is given in Algorithm \ref{algo:median_algo}. Our implementation uses $\eta = 0.8$, $k = 2$, and $N = 30$. The algorithm repeatedly invokes the $\textsc{Inpaint}$ function, which stands for an inpainting algorithm based on SDEdit \cite{meng2022sdedit} as implemented in the Diffusers library \cite{diffusers}. For completeness, we include its pseudocode in Algorithm \ref{algo:inpainting_algo}.
This algorithm requires no modification to the diffusion model and resembles standard diffusion sampling except the ``imputing'' step in Line 16. 


\DecMargin{1em}
\SetKwInput{Input}{Input}
\SetKwInput{Output}{Output}
\SetKwFunction{FUpdate}{Update}
\setcounter{algocf}{-1}
\begin{algorithm}[h!]
  
  \caption{Inpainting using Diffusion Model}\label{algo:inpainting_algo}
  \begin{algorithmic}[1]
    \Function{Update}{$\vect{z}, t, \vect{\epsilon}$} \\
        \quad \Return $\sqrt{\alpha_{t-1}} \parens*{\frac{\vect{z} - \sqrt{1 - \alpha_t} \vect{\epsilon}}{\sqrt{\alpha}}} + \sqrt{1 - \alpha_{t-1}} \vect{\epsilon}$
    \EndFunction \\
    {\color{gray}
    \\
     \small // \textbf{input:} latent code of input image $\vect{z}_I$, \\ // initial latent code $\vect{z}$, // inpainting mask $M$ \\ // conditioning signal (e.g. text, depth map) $\vect{C}$ \\ // timestep to start denoising ($\text{denoising-start}$).\\
    \small // \textbf{output:} Input image with an inpainted chrome ball.
    }
    \Function{Inpaint}{$\vect{z}_I$, $\vect{z}, \vect{M}, \vect{C}, \text{denoising-start} = T$} \\
    \quad $\vect{\epsilon} \sim \mathcal{N}(\vect{0}, \vect{I})$ \\
    \quad \textbf{for} $i \in \{\text{denoising-start}, \ldots, 1\}$ \textbf{do} \\
    \quad \quad $\vect{\epsilon} \leftarrow \epsilon_{\vect{\theta}}(\vect{z}, t, \vect{C})$ \\
    \quad \quad $\vect{z} \leftarrow \Call{Update}{\vect{z}, t, \vect{\epsilon}}$ \\
    
    \quad \quad $\vect{z}_I' \leftarrow \sqrt{\alpha_t}\vect{z}_I + \sqrt{1-\alpha_t} \vect{\epsilon}$ \\
    \quad \quad $\vect{z} \leftarrow (1 - \vect{M}) \odot \vect{z}_I' + \vect{M} \odot \vect{z}$ \\
    \quad \textbf{end for} \\
    \quad $\vect{\epsilon} \leftarrow \epsilon_{\theta}(\vect{z}, 0, \vect{C})$ \\
    \quad $\vect{z} \leftarrow \Call{Update}{\vect{z}, 0, \vect{\epsilon}}$ \\
    \quad \Return $\Call{Decode}{\vect{z}}$
    \EndFunction
  \end{algorithmic}  
\end{algorithm}

\SetKwInput{Input}{Input}
\SetKwInput{Output}{Output}
\begin{algorithm}[htbp]
  \caption{Iterative Inpainting Algorithm}\label{algo:median_algo}
  \begin{algorithmic}[1]
    \Function{InpaintBall}{$\vect{I}, \vect{M}, \vect{C}, \eta, T=1000$} \\
    \quad $\vect{z} \leftarrow \Call{Encode}{\vect{I}}$ \\
    \quad $\vect{\epsilon} \sim \mathcal{N}(\vect{0}, \vect{I})$ \\
    \quad $\vect{z}_{\eta T} \leftarrow \sqrt{\alpha_{\eta T}}\vect{z} + \sqrt{1-\alpha_{\eta T}} \vect{\epsilon}$ \\
    \quad \Return $\Call{Inpaint}{\vect{z}, \vect{z}_{\eta T}, \vect{M}, \vect{C}, \text{denoising-start}=\eta T}$
    \EndFunction \\ 
    { \color{gray} \\
    {\small // \textbf{input:} Input image $\vect{I}$, binary inpainting mask $\vect{M}$, \\ // Conditioning signal (e.g. text, depth map) $\vect{C}$ \\ // denoising strength $\eta$, number of balls for median $N$, \\ // number of median iterations $k$.}\\
    {\small // \textbf{output:} Input image with an inpainted chrome ball.}
    }
    \Function{IterativeInpaint}{$\vect{I}, \vect{M}, \vect{C}, \eta, k, N$} \\
    \quad \textbf{for} $i \in \{1, \ldots, k\}$ \textbf{do} \\
    \quad \quad \textbf{for} $j \in \{1, \ldots, N\}$ \textbf{do} \\
    \quad \quad \quad $\eta' \leftarrow \eta$ \textbf{if} $i > 1$ \textbf{else} $1.0$ \\ 
    \quad \quad \quad $\vect{B}_j \leftarrow \Call{InpaintBall}{\vect{I}, \vect{M}, \vect{C}, \eta'}$ \\
    \quad \quad \textbf{end for}\\ 
    \quad \quad $\vect{B} \leftarrow \Call{PixelwiseMedian}{\vect{B}_1, \ldots, \vect{B}_N}$ \\
    \quad \quad $\vect{I} \leftarrow (1 - \vect{M}) \odot \vect{I} + \vect{M} \odot \vect{B}$ \\
    \quad \textbf{end for} \\
    \quad \Return $\Call{InpaintBall}{\vect{I}, \vect{M}, \eta}$
    \EndFunction
  \end{algorithmic}
\end{algorithm}

\subsection{HDR merging algorithm}

Algorithm \ref{algo:ldr2hdr} merges a bracket of LDR images to create an HDR image. As mentioned in the main paper, we merge in the luminance space to avoid ghosting artifacts. Our luminance conversion assumes sRGB \cite{nguyen2017luminance} and gamma value of 2.4, following StyleLight~\cite{wang2022stylelight}.

\begin{algorithm}[htbp]
  \caption{HDR Merging Algorithm}\label{algo:ldr2hdr}
  \begin{algorithmic}[1]
    \Function{Luminance}{$\vect{I}, ev, \gamma=2.4$} \\
    \quad \Return $\vect{I}^\gamma \cdot [0.21267, 0.71516, 0.07217]^\top (2^{-ev})$
    \EndFunction \\ 
    {\color{gray} \\
    {\small // \textbf{input:} LDR images and a list of exposure values in \\ // descending order, where $ev_0=0$. E.g., [0, -2.5, -5]}\\
    {\small // \textbf{output:} A linearized HDR image.}
    }
    \Function{MergeLDRs}{$\vect{I}_0, ..., \vect{I}_{n-1}, ev_0, ..., ev_{n-1}$} \\
    \quad $\vect{L} \leftarrow  \Call{Luminance}{\vect{I}_{n-1}, ev_{n-1}}$ \\
    \quad \textbf{for} $i \in \{n-2, n-1, ..., 0\}$ \textbf{do} \\ \quad \quad 
        $\vect{L}_{i} \leftarrow \Call{Luminance}{\vect{I}_{i} , ev_i}$ \\ \quad \quad 
        $\vect{M} \leftarrow \Call{clip}{\frac{(2^{ev_i} \vect{L}_i)-0.9}{0.1}, 0, 1} \odot \mathbbm{1}(\vect{L} > \vect{L}_i)$ \\ \quad \quad
        $\vect{L} \leftarrow (1 - \vect{M}) \odot \vect{L}_i + \vect{M} \odot \vect{L}$ \\ \quad
        \textbf{end for}\\ 
    \quad \Return $\vect{I}_0^\gamma \odot \left(\frac{\vect{L}}{\vect{L}_0}\right)$
    \EndFunction
  \end{algorithmic}
\end{algorithm}

\section{Ablation on DiffusionLight}

\subsection{Inpainting ball size}
\begin{figure}[ht]
    \centering
    \begin{tabu} to \textwidth {
        @{}
        c@{}
        c@{\hspace{1.0pt}}
        c@{\hspace{0.5pt}}
        c@{\hspace{0.5pt}}
        c@{\hspace{0.5pt}}
        c@{\hspace{0.5pt}}
        c@{}
    }
    
        &
        \multicolumn{1}{c}{\tiny Ball radius:} &
        \multicolumn{1}{c}{\tiny 128} &
        \multicolumn{1}{c}{\tiny 256} &
        \multicolumn{1}{c}{\tiny 384} &
        \multicolumn{1}{c}{\tiny 512} &
        \\
        
        \multicolumn{1}{l}{\rotatebox[origin=c]{90}{\shortstack[l]{\tiny Input \#1}}} &
        \noindent\parbox[c]{0.098\textwidth}{\includegraphics[width=0.098\textwidth]{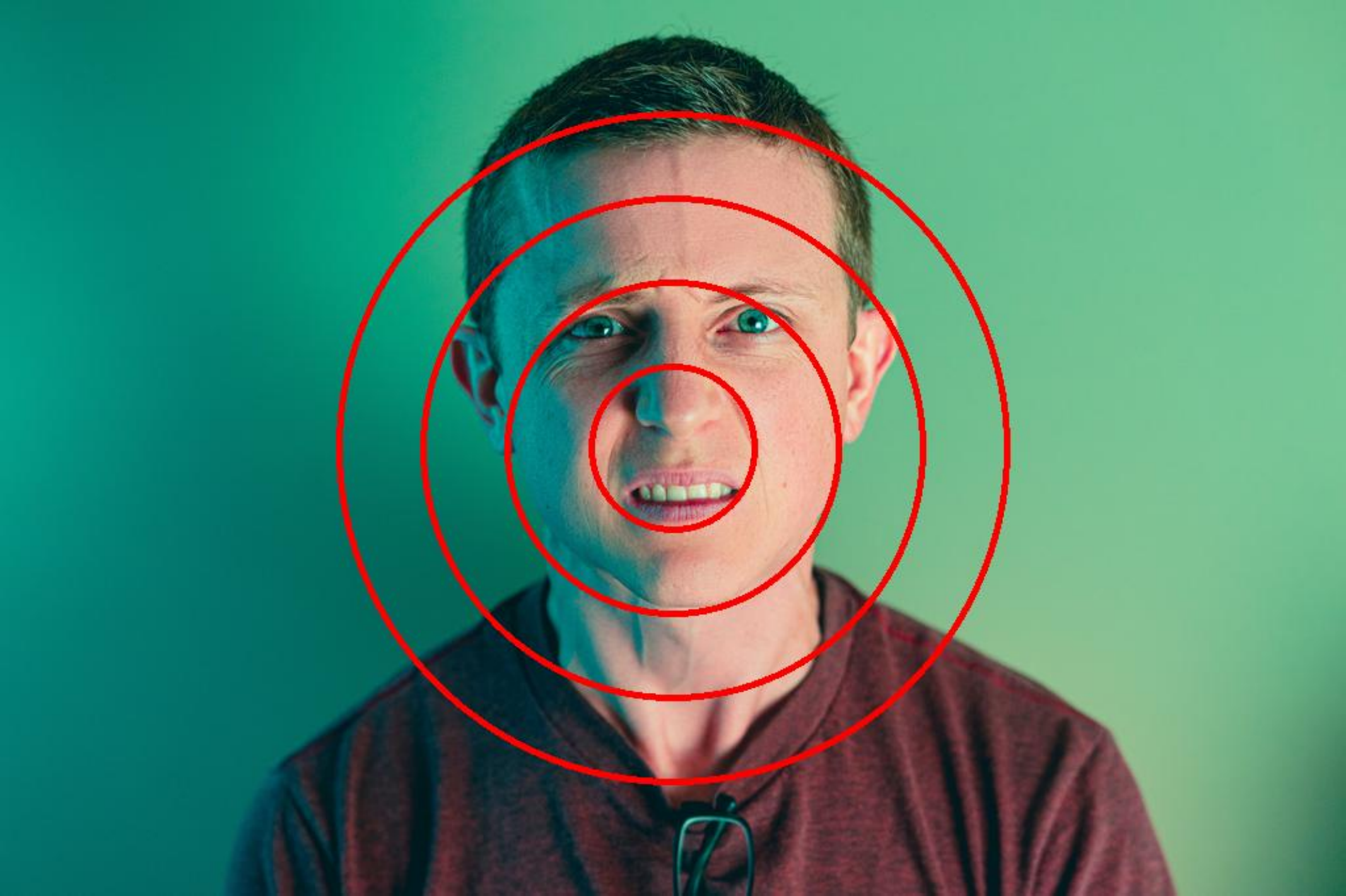}} & 
        \noindent\parbox[c]{0.065\textwidth}{\includegraphics[width=0.065\textwidth]{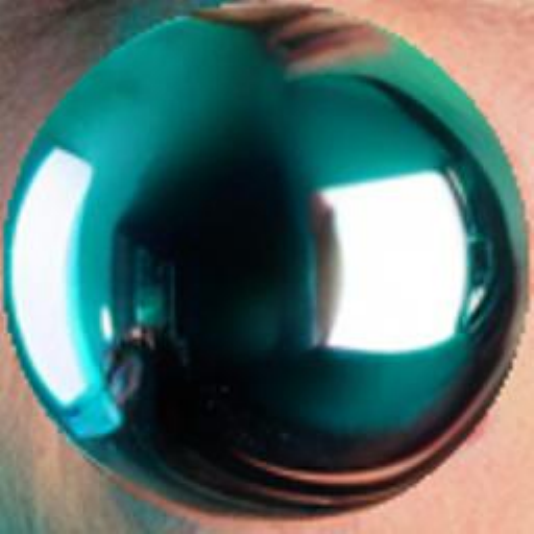}} & 
        \noindent\parbox[c]{0.065\textwidth}{\includegraphics[width=0.065\textwidth]{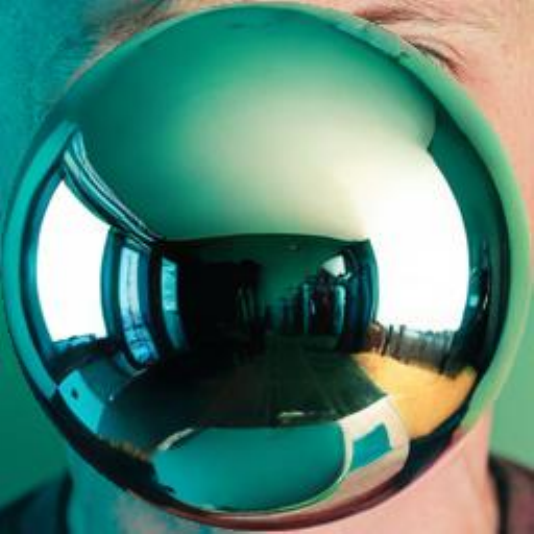}} & 
        \noindent\parbox[c]{0.065\textwidth}{\includegraphics[width=0.065\textwidth]{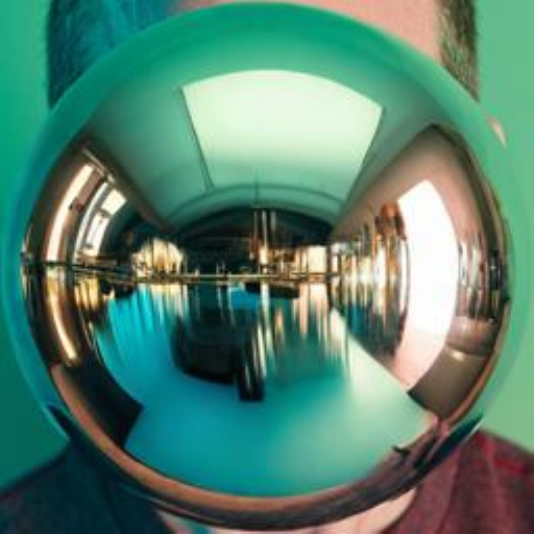}} & 
        \noindent\parbox[c]{0.065\textwidth}{\includegraphics[width=0.065\textwidth]{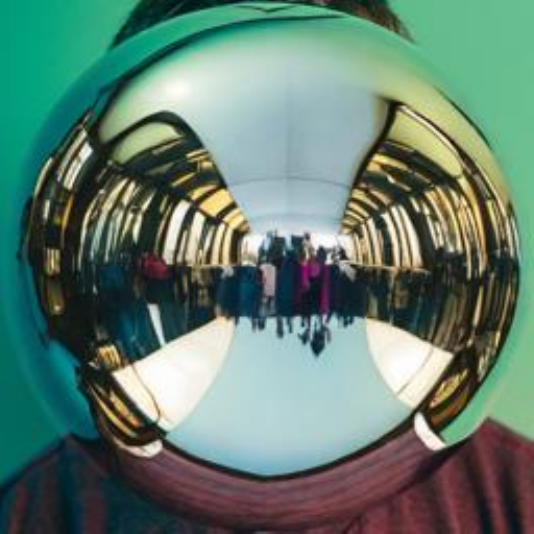}} & 
        \\

        \multicolumn{1}{l}{\rotatebox[origin=c]{90}{\shortstack[l]{\tiny Input \#2}}} &
        \noindent\parbox[c]{0.098\textwidth}{\includegraphics[width=0.098\textwidth]{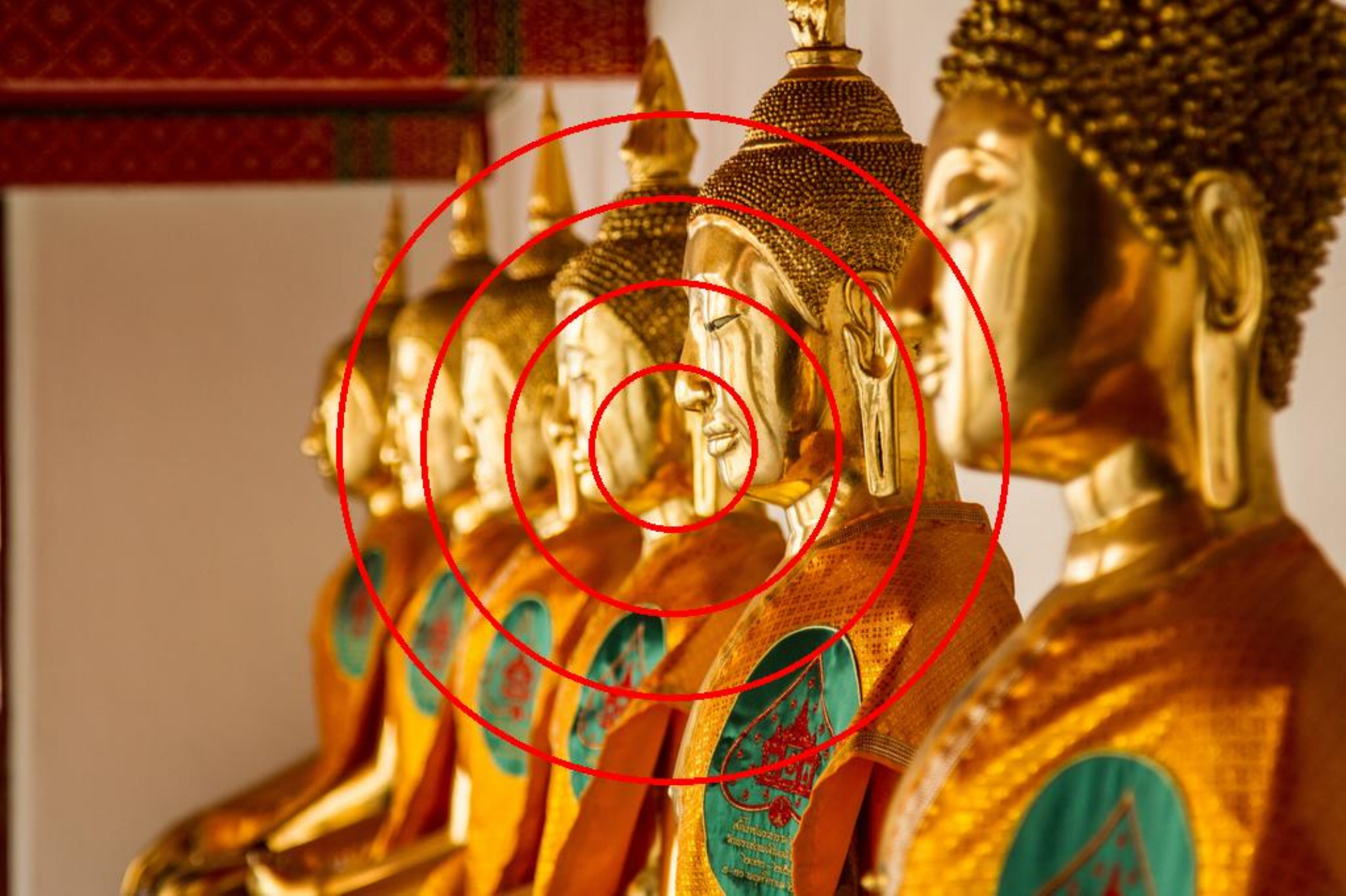}} & 
        \noindent\parbox[c]{0.065\textwidth}{\includegraphics[width=0.065\textwidth]{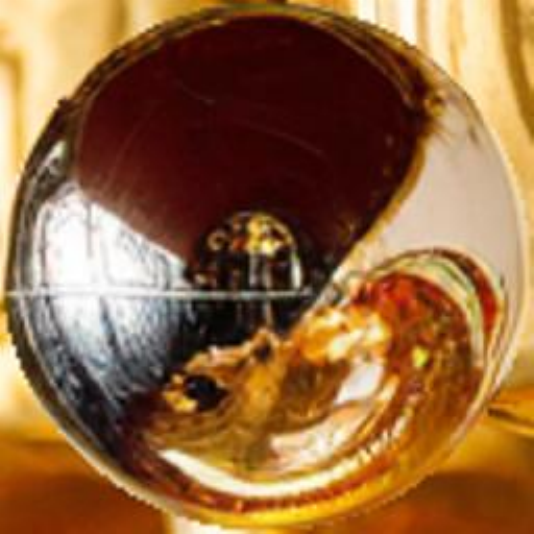}} & 
        \noindent\parbox[c]{0.065\textwidth}{\includegraphics[width=0.065\textwidth]{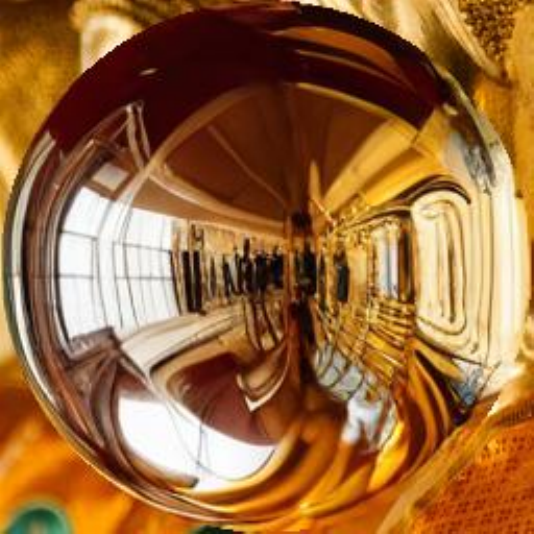}} & 
        \noindent\parbox[c]{0.065\textwidth}{\includegraphics[width=0.065\textwidth]{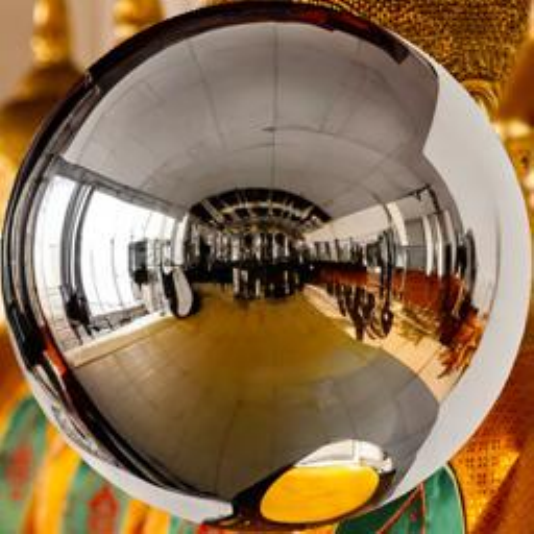}} & 
        \noindent\parbox[c]{0.065\textwidth}{\includegraphics[width=0.065\textwidth]{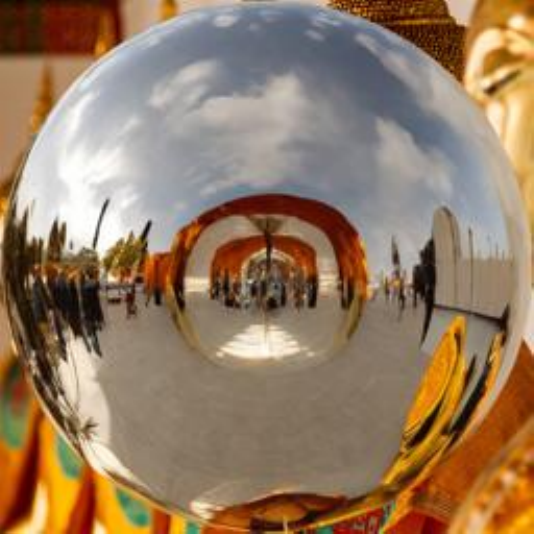}} & 
        \\

        \multicolumn{1}{l}{\rotatebox[origin=c]{90}{\shortstack[l]{\tiny Input \#2}}} &
        \noindent\parbox[c]{0.098\textwidth}{\includegraphics[width=0.098\textwidth]{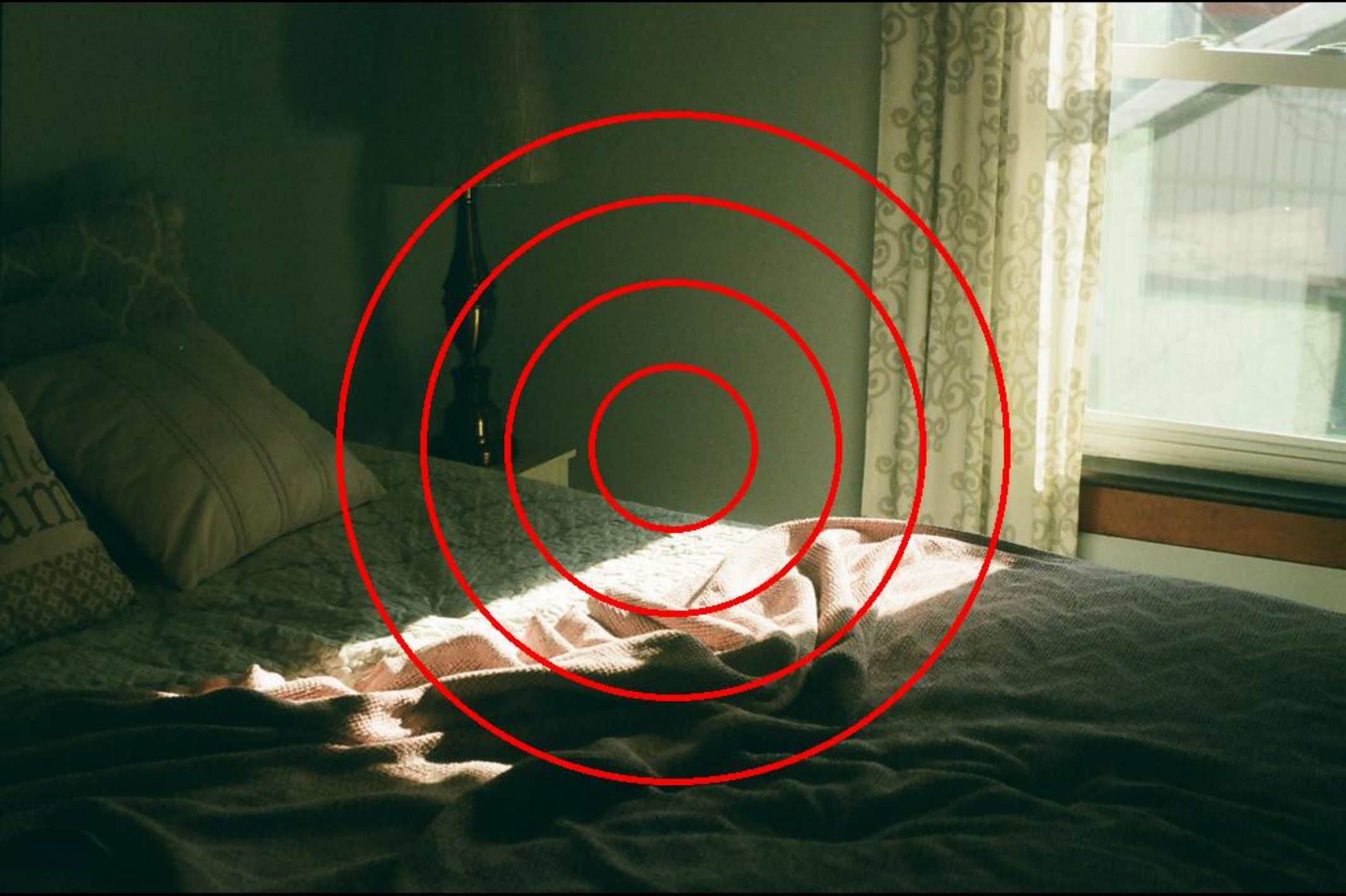}} & 
        \noindent\parbox[c]{0.065\textwidth}{\includegraphics[width=0.065\textwidth]{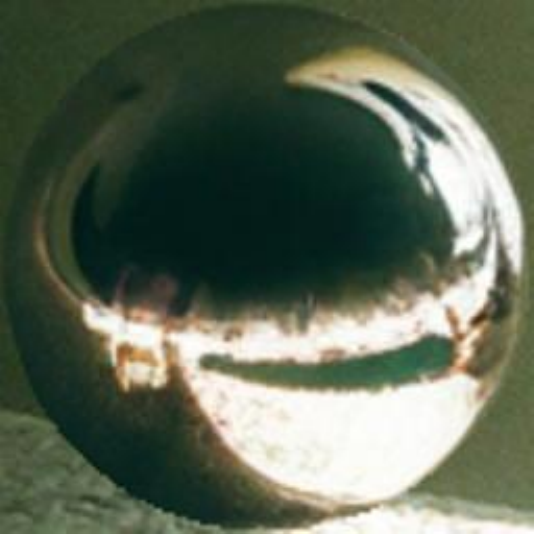}} & 
        \noindent\parbox[c]{0.065\textwidth}{\includegraphics[width=0.065\textwidth]{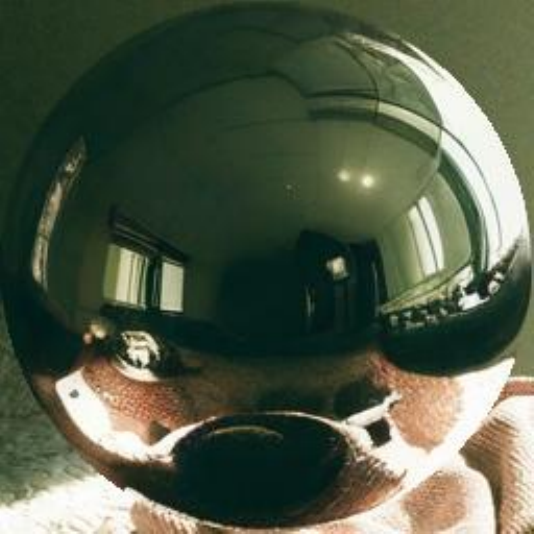}} & 
        \noindent\parbox[c]{0.065\textwidth}{\includegraphics[width=0.065\textwidth]{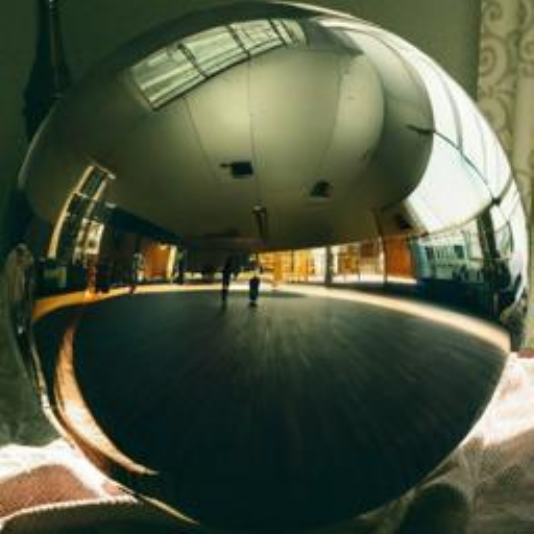}} & 
        \noindent\parbox[c]{0.065\textwidth}{\includegraphics[width=0.065\textwidth]{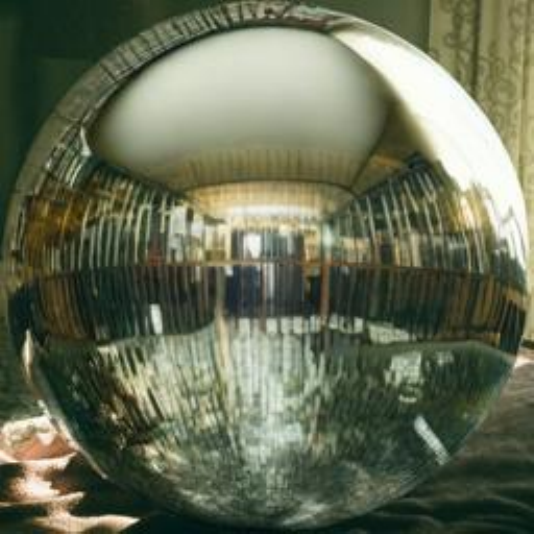}} & 
        \\

        \multicolumn{1}{l}{\rotatebox[origin=c]{90}{\shortstack[l]{\tiny Input \#2}}} &
        \noindent\parbox[c]{0.098\textwidth}{\includegraphics[width=0.098\textwidth]{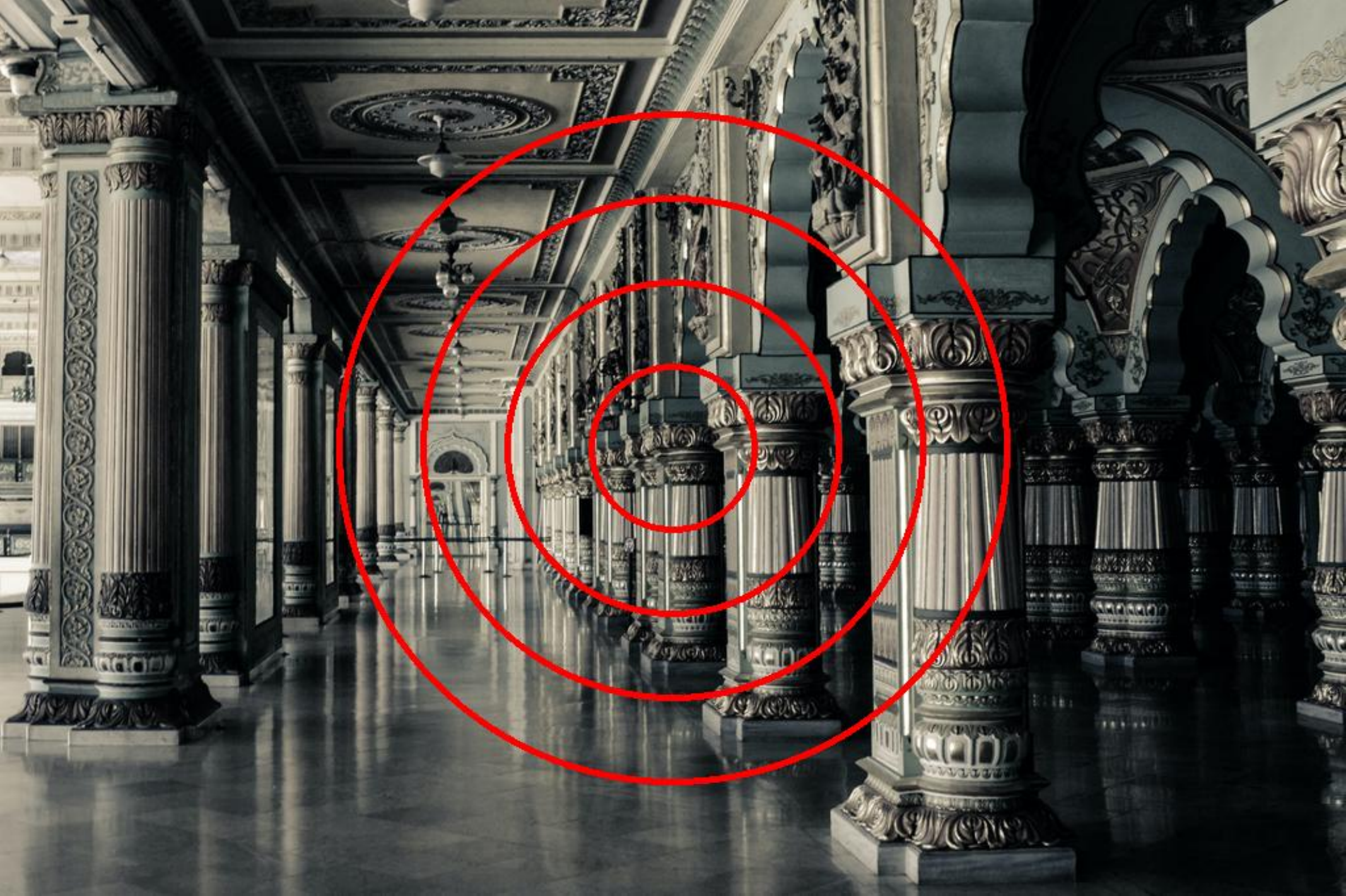}} & 
        \noindent\parbox[c]{0.065\textwidth}{\includegraphics[width=0.065\textwidth]{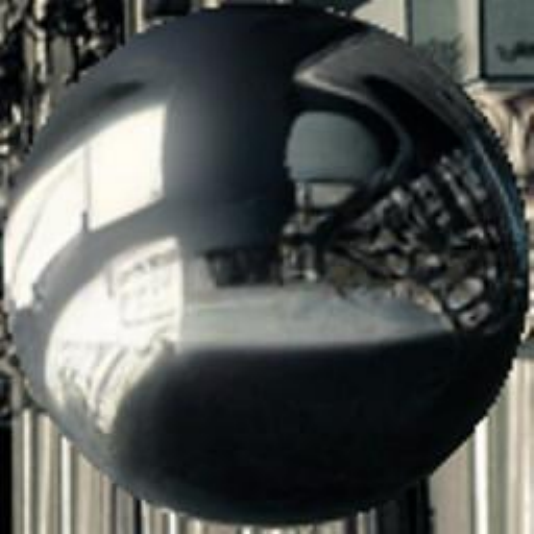}} & 
        \noindent\parbox[c]{0.065\textwidth}{\includegraphics[width=0.065\textwidth]{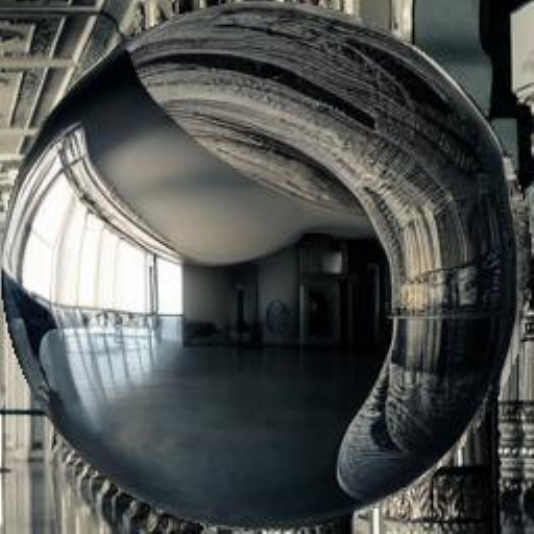}} & 
        \noindent\parbox[c]{0.065\textwidth}{\includegraphics[width=0.065\textwidth]{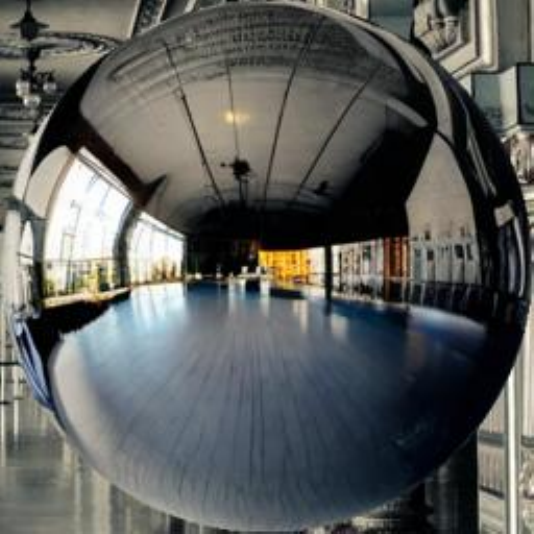}} & 
        \noindent\parbox[c]{0.065\textwidth}{\includegraphics[width=0.065\textwidth]{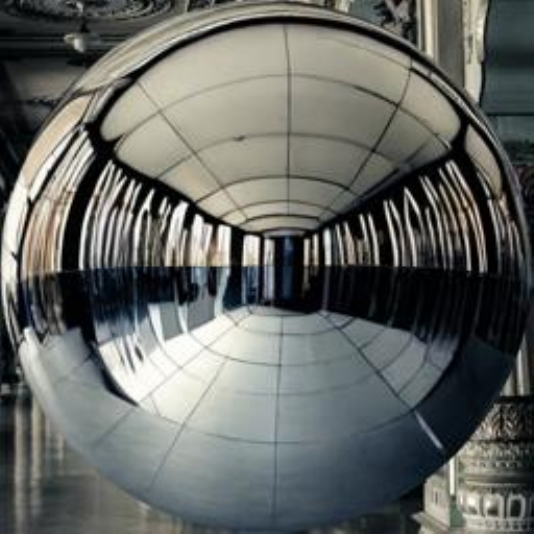}} & 
        \\
        
    \end{tabu}
    \vskip-3pt
    \begin{tabu} to \textwidth {
        @{}
        c@{}
        c@{\hspace{0.2pt}}
        c@{\hspace{0.2pt}}
        c@{\hspace{0.2pt}}
        c@{\hspace{2pt}}
        c@{\hspace{0.2pt}}
        c@{\hspace{0.2pt}}
        c@{\hspace{0.2pt}}
        c@{\hspace{0.2pt}}
        c@{}
    }

        
        &
        \multicolumn{4}{c}{\tiny Median balls for input \#1} &
        \multicolumn{4}{c}{\tiny Median balls for input \#2} &
        \\

        \multicolumn{1}{l}{\vspace{-1pt} \tiny Ball radius:} &
        \multicolumn{1}{c}{\tiny 128} &
        \multicolumn{1}{c}{\tiny 256} &
        \multicolumn{1}{c}{\tiny 384} &
        \multicolumn{1}{c}{\tiny 512} &
        \multicolumn{1}{c}{\tiny 128} &
        \multicolumn{1}{c}{\tiny 256} &
        \multicolumn{1}{c}{\tiny 384} &
        \multicolumn{1}{c}{\tiny 512} &
        \\

        \multicolumn{1}{l}{\shortstack[l]{\tiny Last \\ \tiny iteration}} &
        \noindent\parbox[c]{0.050\textwidth}{\includegraphics[width=0.050\textwidth]{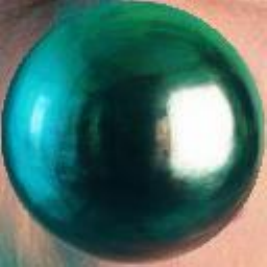}} & 
        \noindent\parbox[c]{0.050\textwidth}{\includegraphics[width=0.050\textwidth]{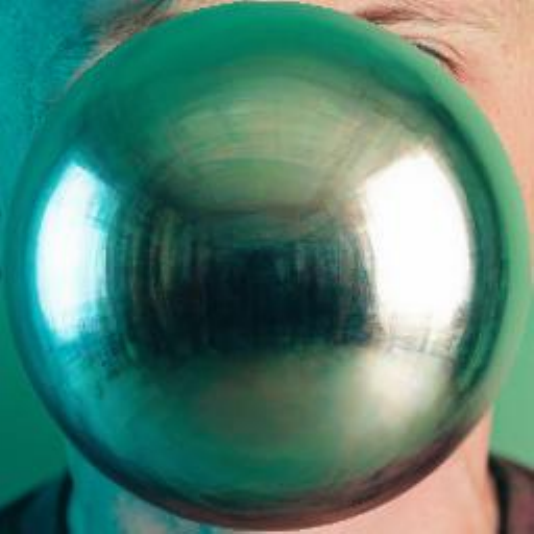}} & 
        \noindent\parbox[c]{0.050\textwidth}{\includegraphics[width=0.050\textwidth]{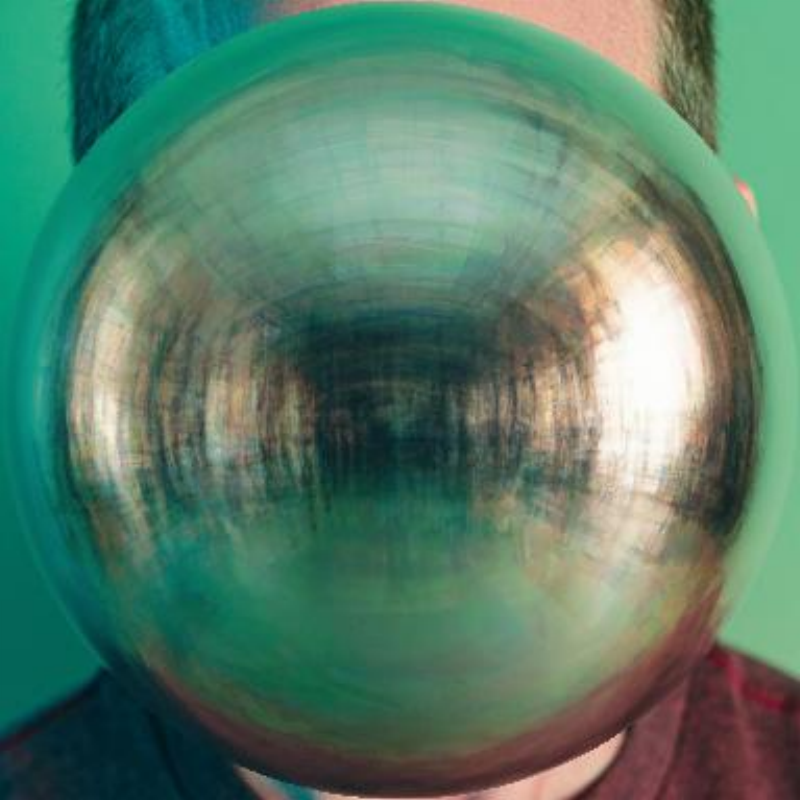}} & 
        \noindent\parbox[c]{0.050\textwidth}{\includegraphics[width=0.050\textwidth]{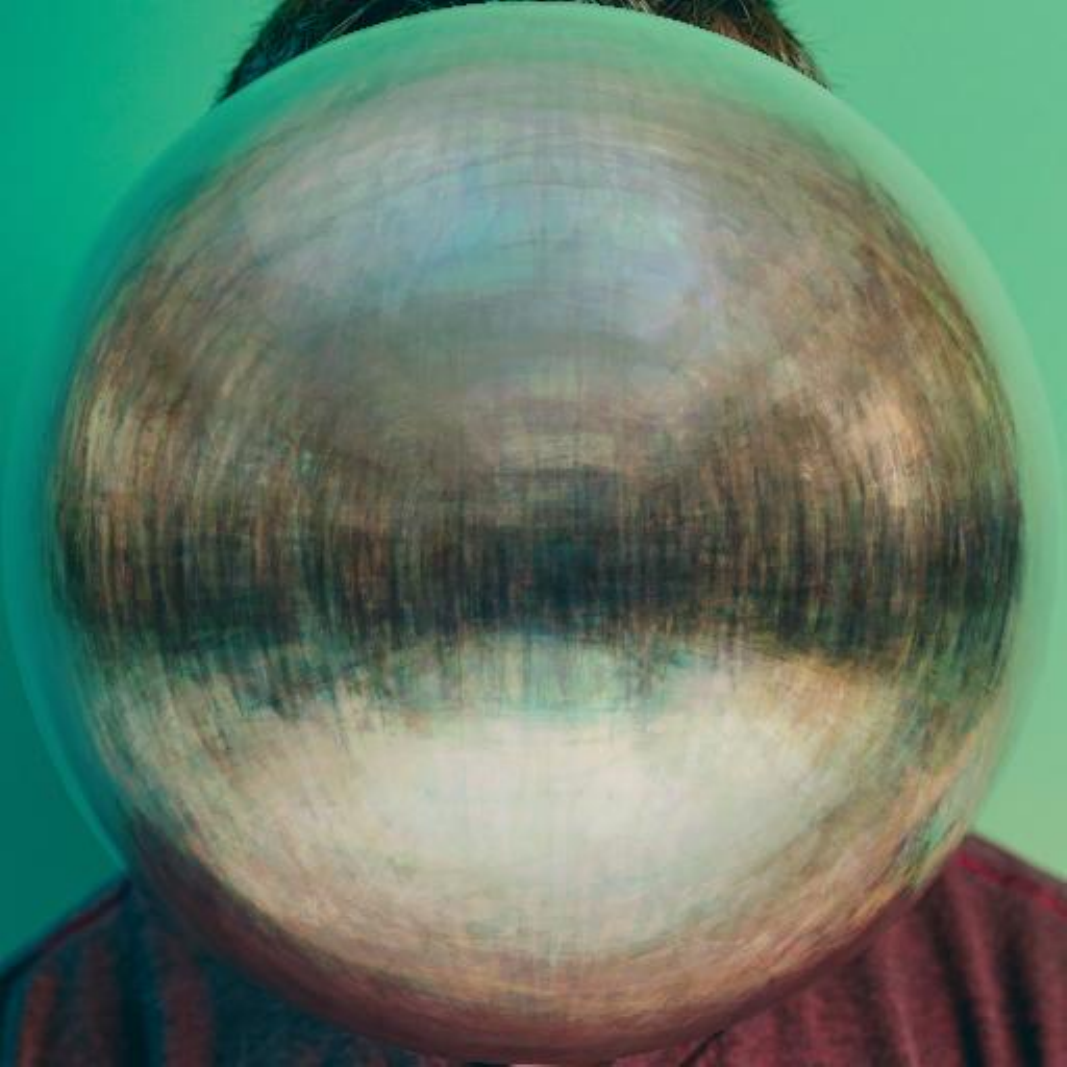}} & 

        \noindent\parbox[c]{0.050\textwidth}{\includegraphics[width=0.050\textwidth]{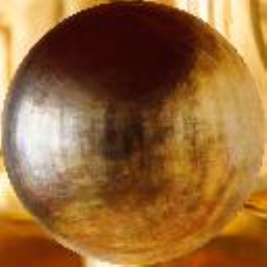}} & 
        \noindent\parbox[c]{0.050\textwidth}{\includegraphics[width=0.050\textwidth]{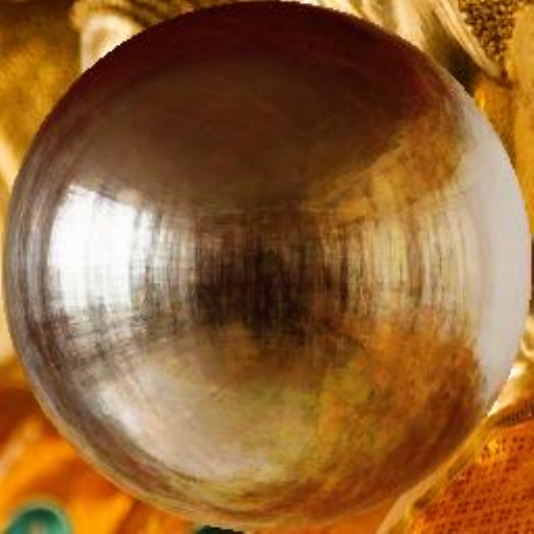}} & 
        \noindent\parbox[c]{0.050\textwidth}{\includegraphics[width=0.050\textwidth]{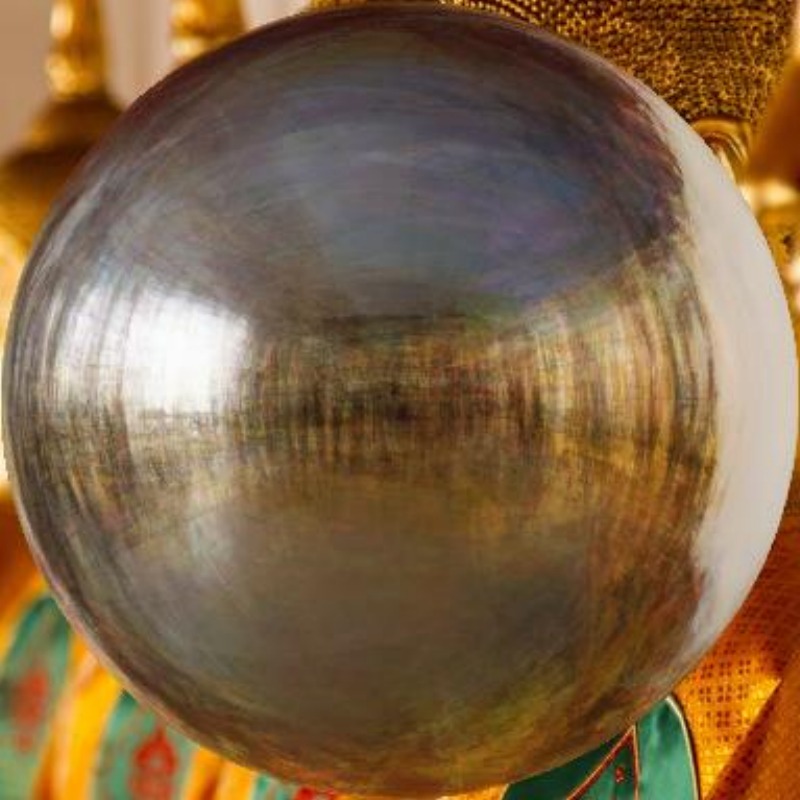}} & 
        \noindent\parbox[c]{0.050\textwidth}{\includegraphics[width=0.050\textwidth]{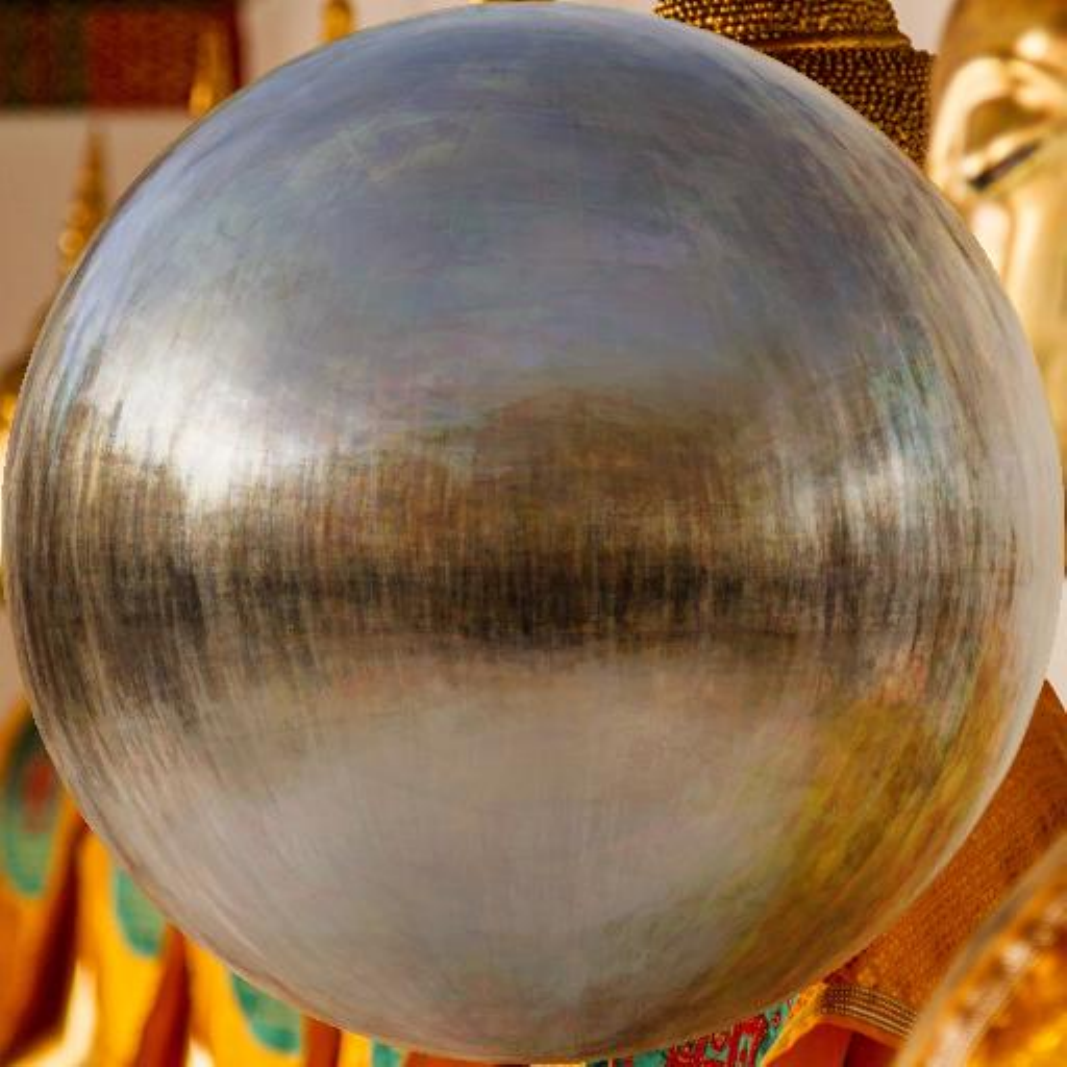}} & 
        \\

        \multicolumn{1}{l}{\shortstack[l]{\tiny Last \\ \tiny iteration}} &
        \noindent\parbox[c]{0.050\textwidth}{\includegraphics[width=0.050\textwidth]{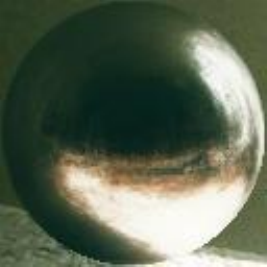}} & 
        \noindent\parbox[c]{0.050\textwidth}{\includegraphics[width=0.050\textwidth]{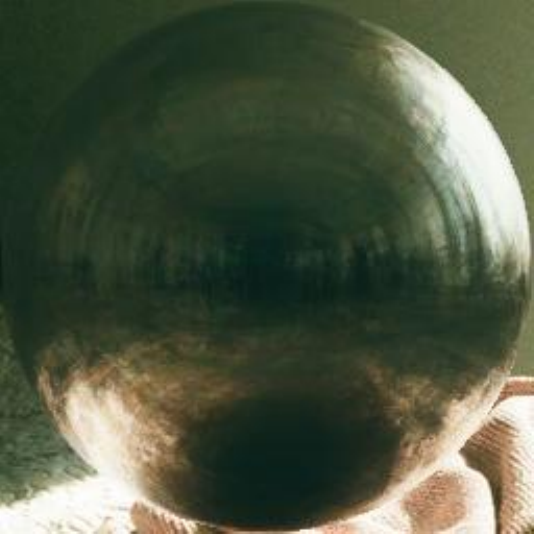}} & 
        \noindent\parbox[c]{0.050\textwidth}{\includegraphics[width=0.050\textwidth]{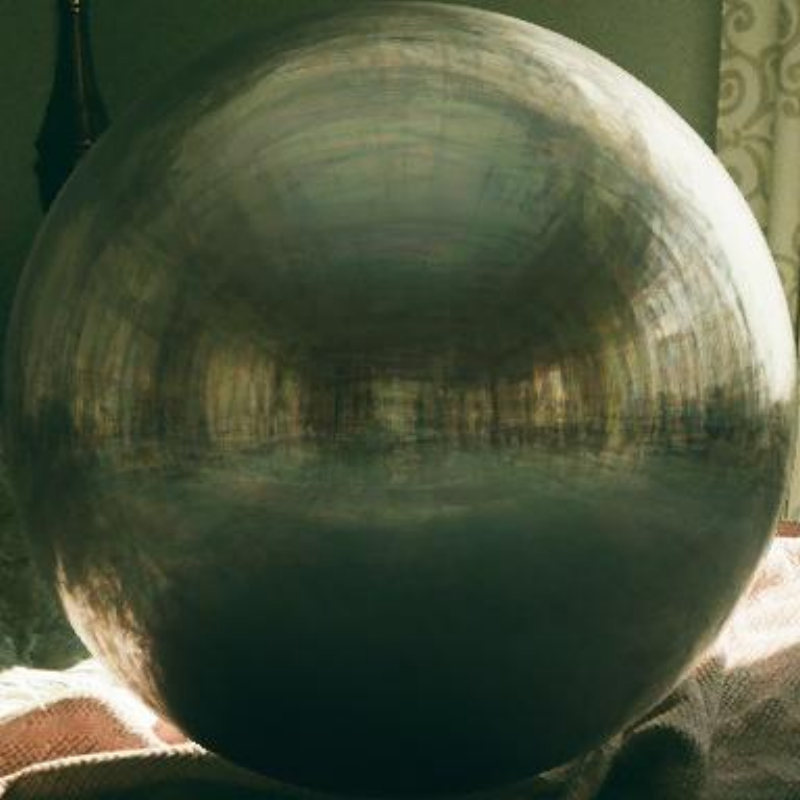}} & 
        \noindent\parbox[c]{0.050\textwidth}{\includegraphics[width=0.050\textwidth]{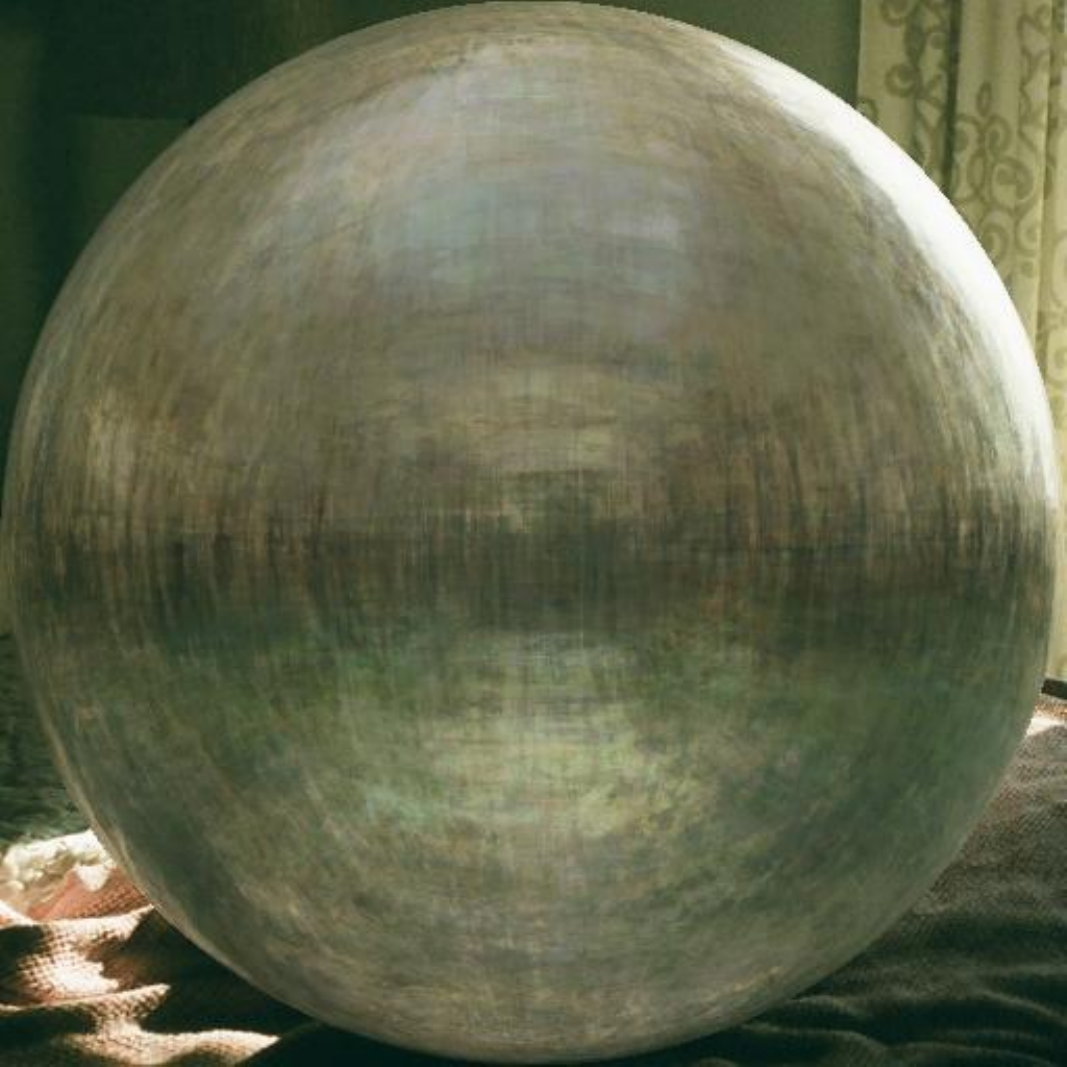}} & 

        \noindent\parbox[c]{0.050\textwidth}{\includegraphics[width=0.050\textwidth]{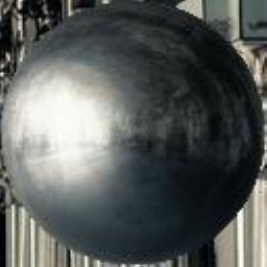}} & 
        \noindent\parbox[c]{0.050\textwidth}{\includegraphics[width=0.050\textwidth]{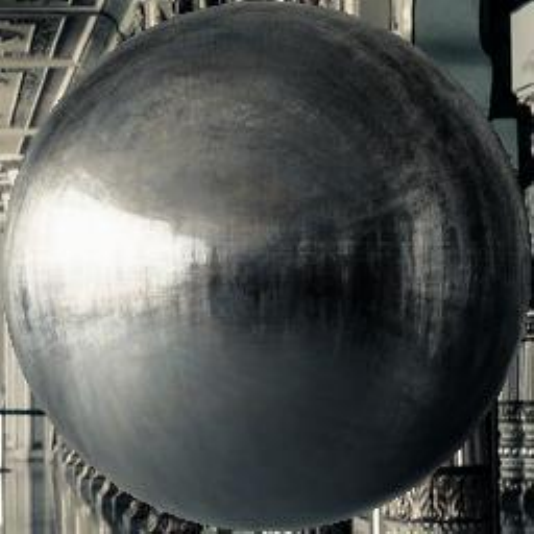}} & 
        \noindent\parbox[c]{0.050\textwidth}{\includegraphics[width=0.050\textwidth]{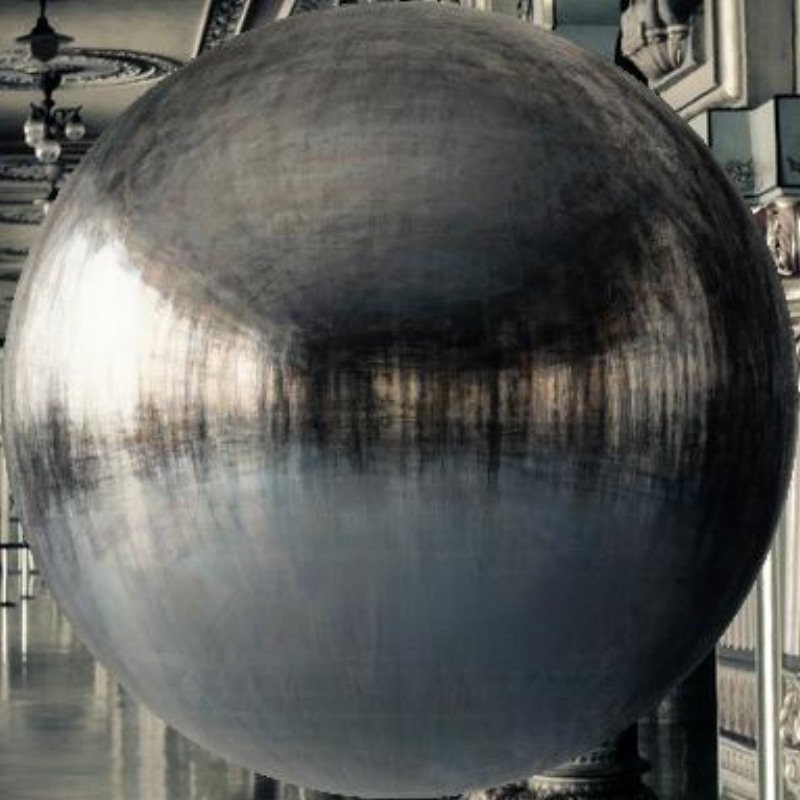}} & 
        \noindent\parbox[c]{0.050\textwidth}{\includegraphics[width=0.050\textwidth]{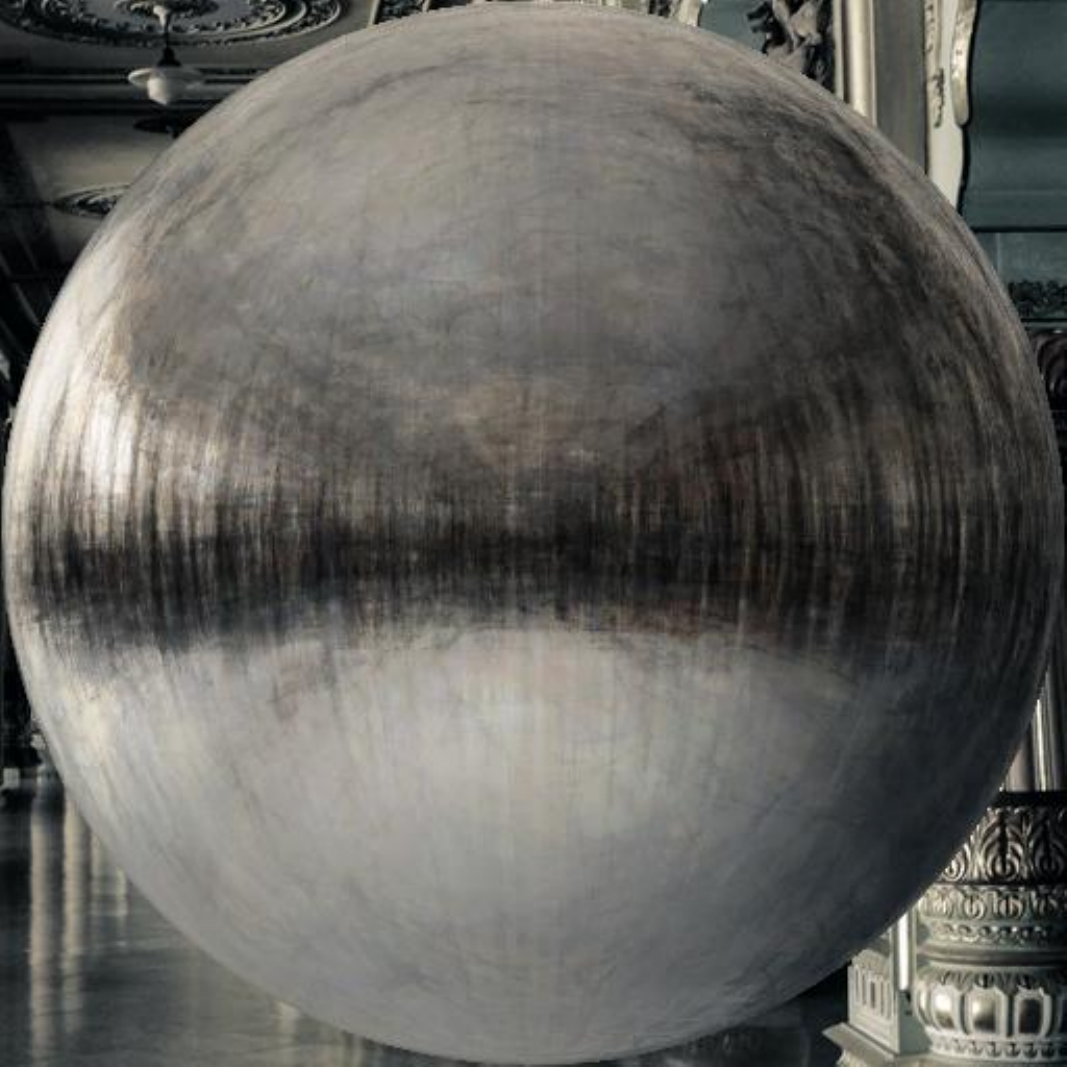}} & 
        \\
        
    \end{tabu}
    \caption{Results when varying the ball size (without LoRA).}
    \label{fig:ball_size}
\end{figure}

We investigated the effects of ball diameter (i.e., white circle) on the depth maps used by ControlNet. Specifically, we analyze and compare the efficacy of various ball sizes, ranging from 128 to 512 pixels in diameter, as illustrated in Figure \ref{fig:ball_size}. The result shows that increasing the ball size from 256 to 384 or 512 still results in realistic balls, but they do not reflect the environment as convincingly. This is likely because the original input content seen by the model is reduced. On the other hand, smaller balls can capture the lighting well but are less detailed and not as useful.


\tabulinesep=0.1pt

\subsection{Ablation on Iterative Inpainting Algorithm} \label{appendix:aba_median}

\begin{figure}
    \centering

    \begin{tabu} to \textwidth {
        @{}
        c@{\hspace{1pt}}
        c@{\hspace{1pt}}
        @{\hspace{8pt}}
        c@{\hspace{1pt}}
        c@{\hspace{1pt}}
        c@{}
    }
        
        \noindent\parbox[c]{0.08\textwidth}{\includegraphics[width=0.08\textwidth]{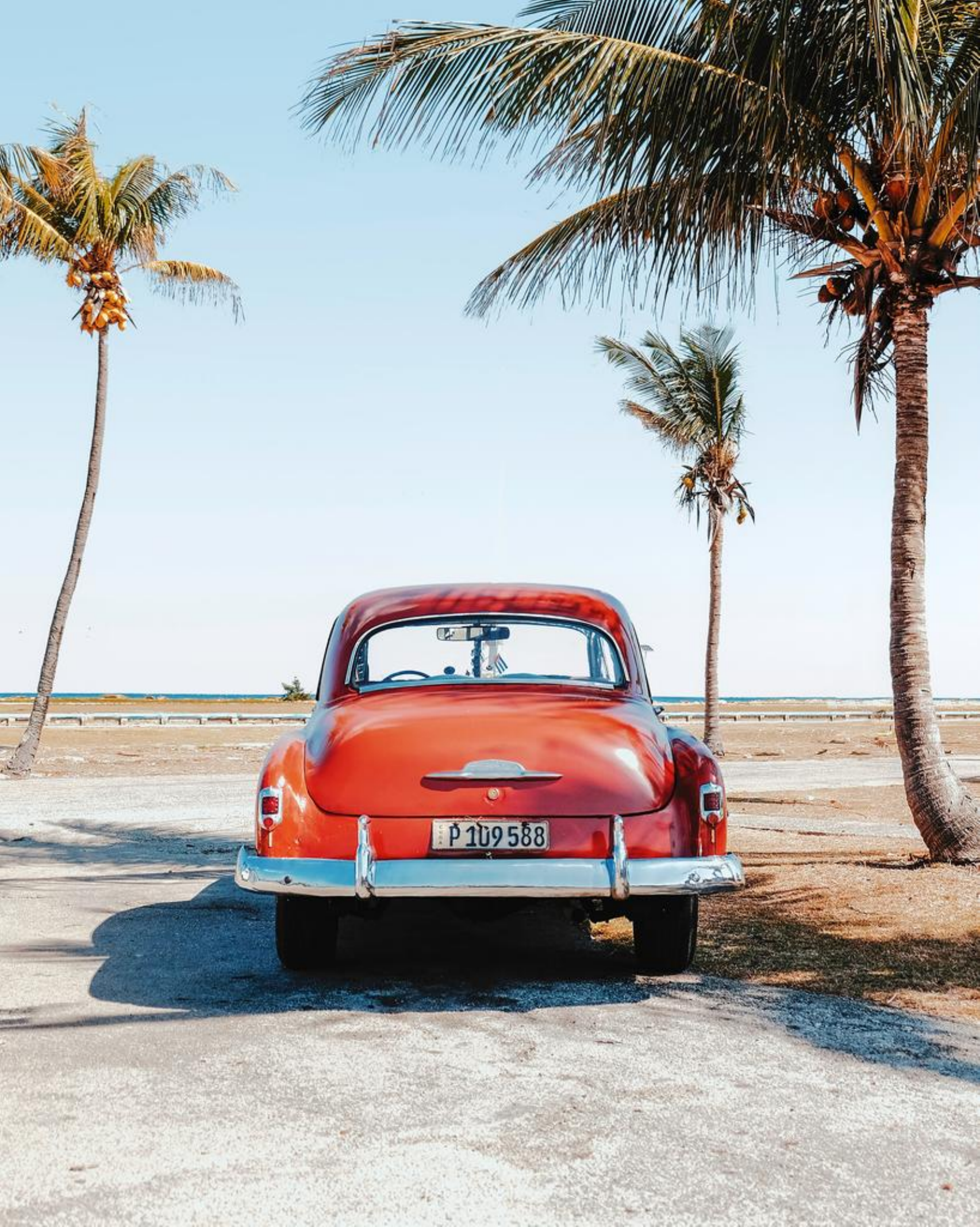}} 
        &
        \noindent\parbox[c]{0.14\textwidth}{\includegraphics[width=0.14\textwidth]{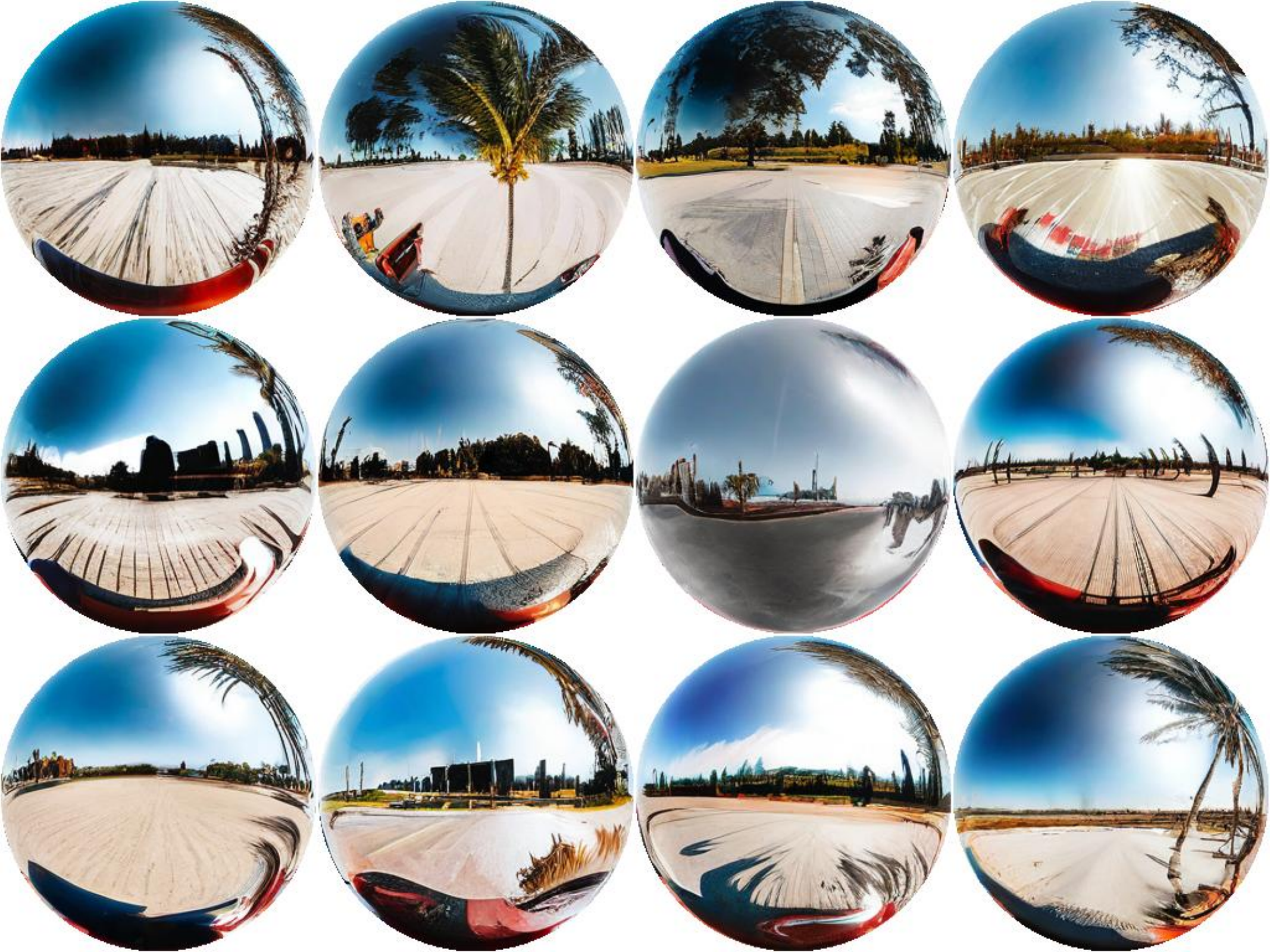}} 

        
        &
        \noindent\parbox[c]{0.08\textwidth}{\shortstack{\tiny Median ball \\ \includegraphics[width=0.08\textwidth]{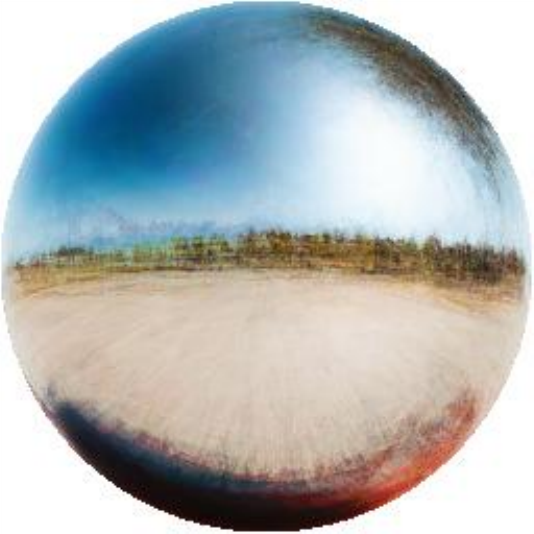}}}
        
        &
        \noindent\parbox[c]{0.14\textwidth}{\includegraphics[width=0.14\textwidth]{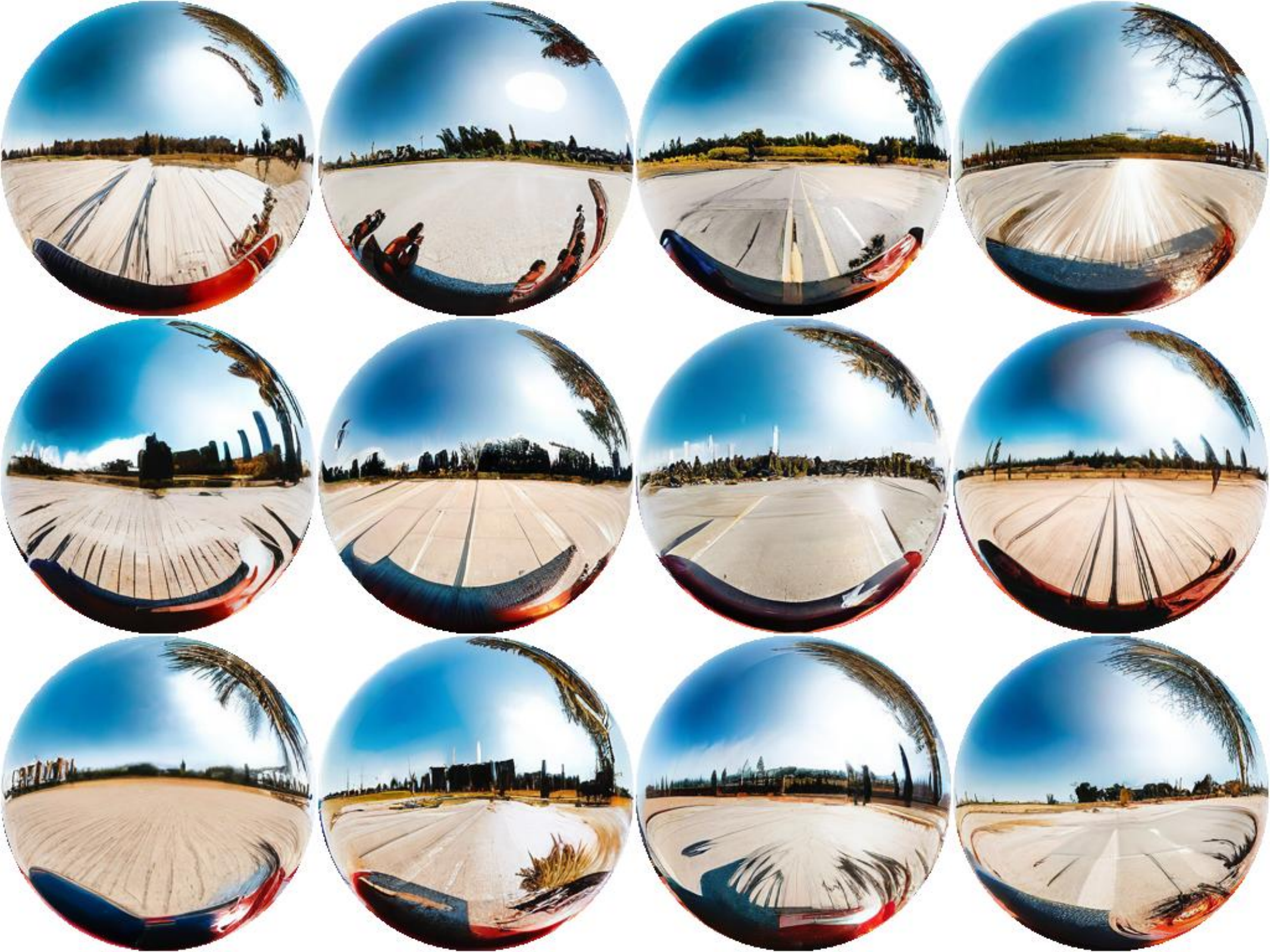}} 
        
        \\

        \hline


        
        
        

        \noindent\parbox[c]{0.08\textwidth}{\includegraphics[width=0.08\textwidth]{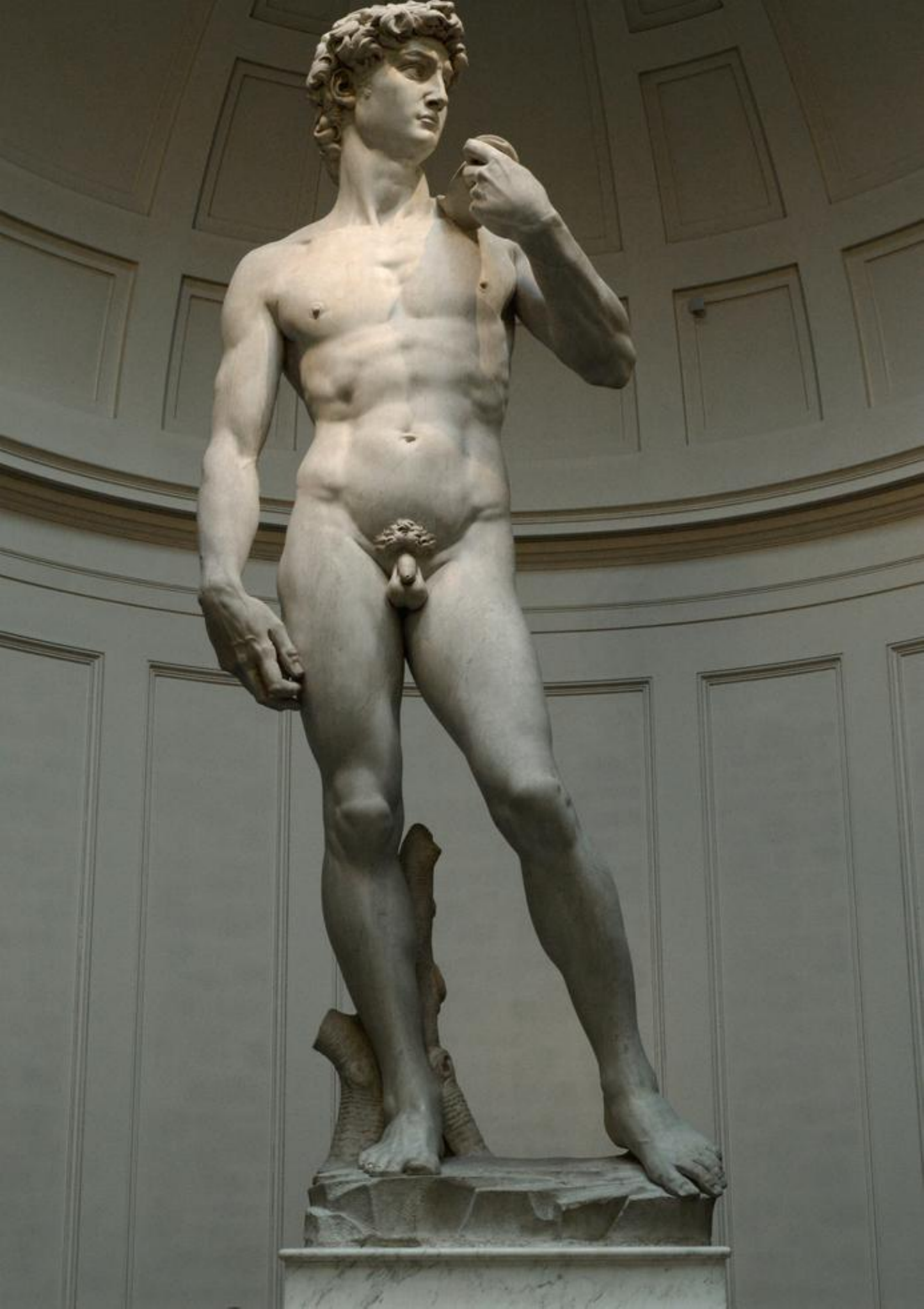}} 
        &
        \noindent\parbox[c]{0.14\textwidth}{\includegraphics[width=0.14\textwidth]{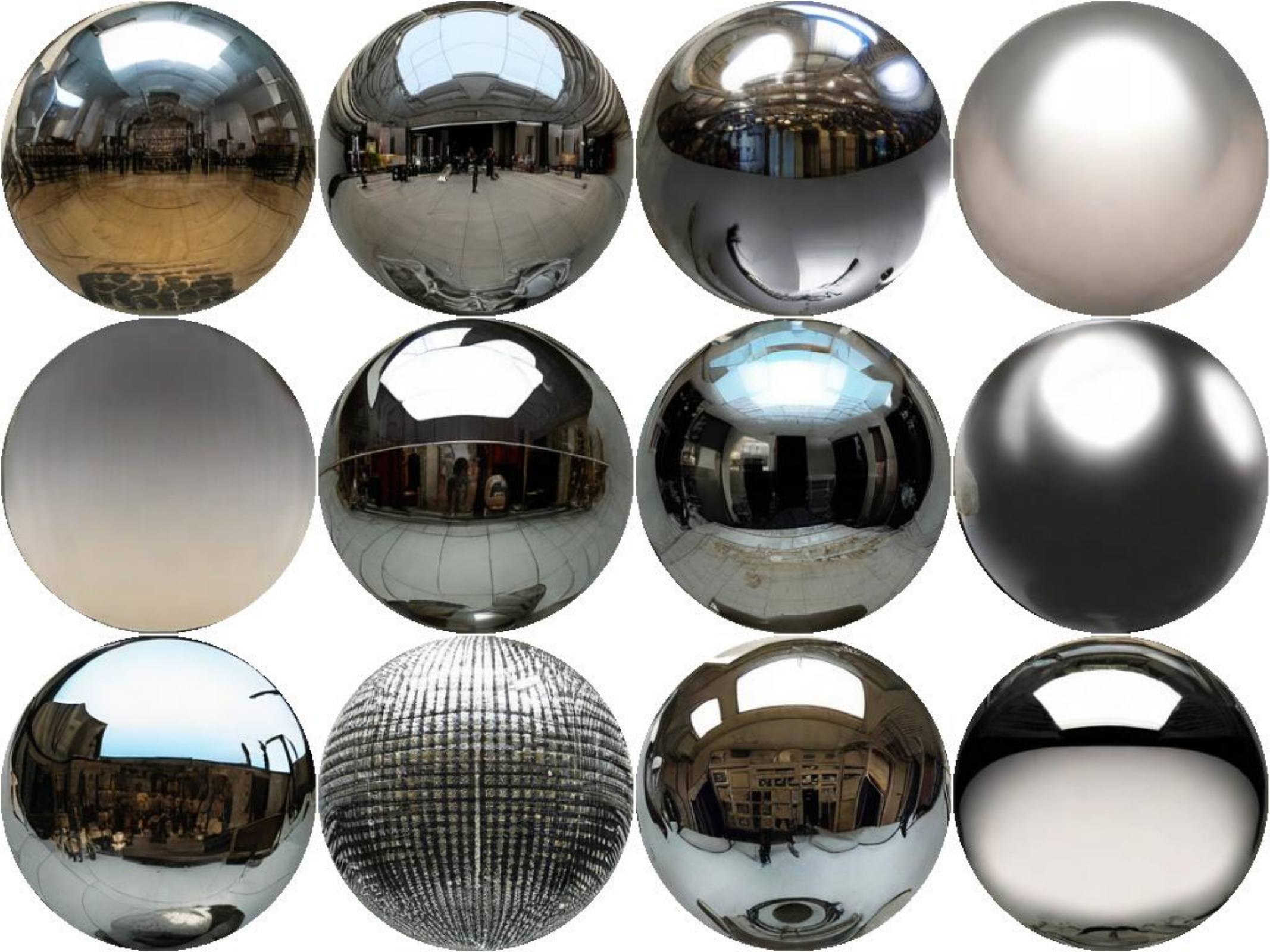}}

        &
        \noindent\parbox[c]{0.08\textwidth}{\shortstack{\tiny Median ball \\ \includegraphics[width=0.08\textwidth]{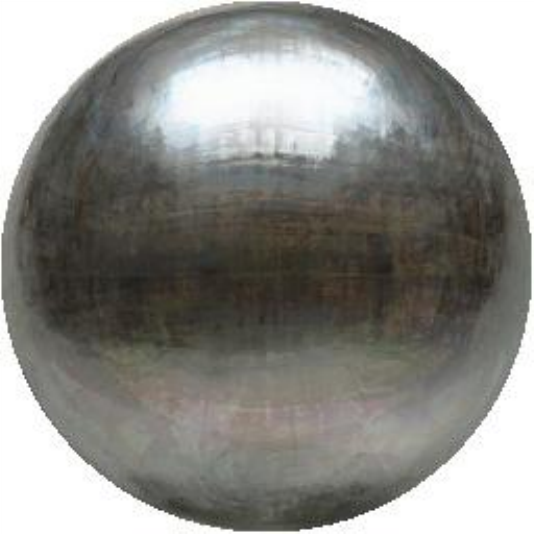}}}
        
        &
        \noindent\parbox[c]{0.14\textwidth}{\includegraphics[width=0.14\textwidth]{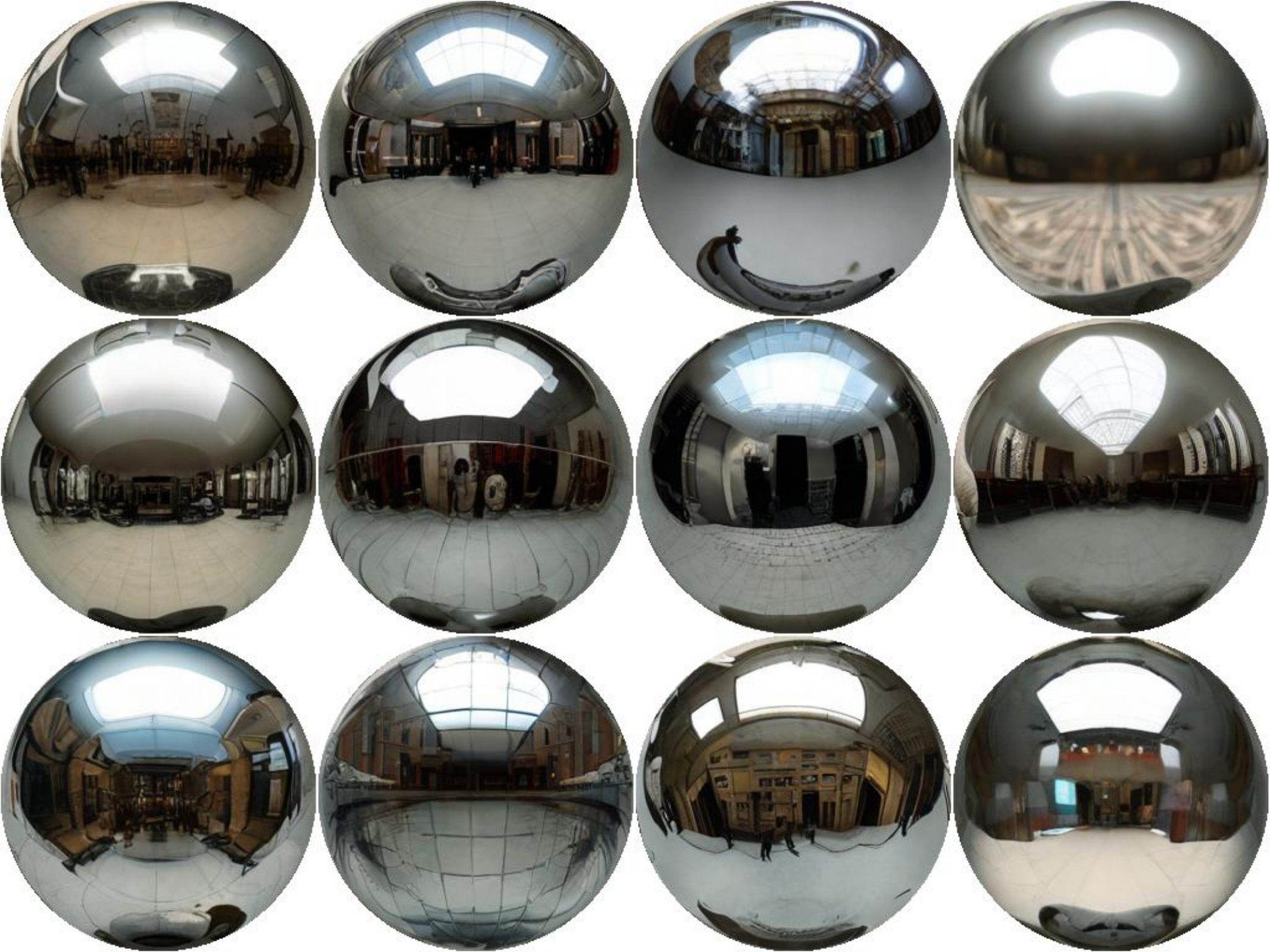}}
        
        
    \end{tabu}
    \caption{Chrome balls before (left) and after (right) one iteration of our iterative inpainting algorithm. Notice how poor chrome balls are fixed and the light estimation becomes more consistent.}
    \label{fig:compare_median_distribution_aba}
\end{figure}
\begin{figure*}
    \centering
        \begin{tabu} to \textwidth {
        c@{}|@{\hspace{0pt}}
        c@{}
        c@{\hspace{0.3pt}}
        c@{\hspace{0.3pt}}
        c@{\hspace{0.3pt}}
        c@{\hspace{0.3pt}}
        c@{\hspace{0.3pt}}
        c@{\hspace{0.3pt}}
        c@{\hspace{0.3pt}}
        c@{\hspace{0.3pt}}
        c@{\hspace{0.3pt}}|@{\hspace{0.3pt}}
        c@{}
    }
        \multicolumn{1}{c}{\shortstack{\scriptsize \hspace{0.3pt} Input image}} &
        &
        \multicolumn{1}{c}{\shortstack{\scriptsize Pred\#1}} & 
        \multicolumn{1}{c}{\shortstack{\scriptsize Pred\#2}} & 
        \multicolumn{1}{c}{\shortstack{\scriptsize Pred\#3}} & 
        \multicolumn{1}{c}{\shortstack{\scriptsize Pred\#4}} & 
        \multicolumn{1}{c}{\shortstack{\scriptsize Pred\#5}} & 
        \multicolumn{1}{c}{\shortstack{\scriptsize Pred\#6}} & 
        \multicolumn{1}{c}{\shortstack{\scriptsize Pred\#7}} & 
        \multicolumn{1}{c}{\shortstack{\scriptsize Pred\#8}} & 
        \multicolumn{1}{c}{\shortstack{\scriptsize \hspace{-5pt} Pred\#9}} & 
        \multicolumn{1}{c}{\shortstack{\scriptsize Median Ball}}
        \\
        
        \multirow{2}{*}{\noindent\parbox[c]{0.112\textwidth}{\includegraphics[width=0.112\textwidth]{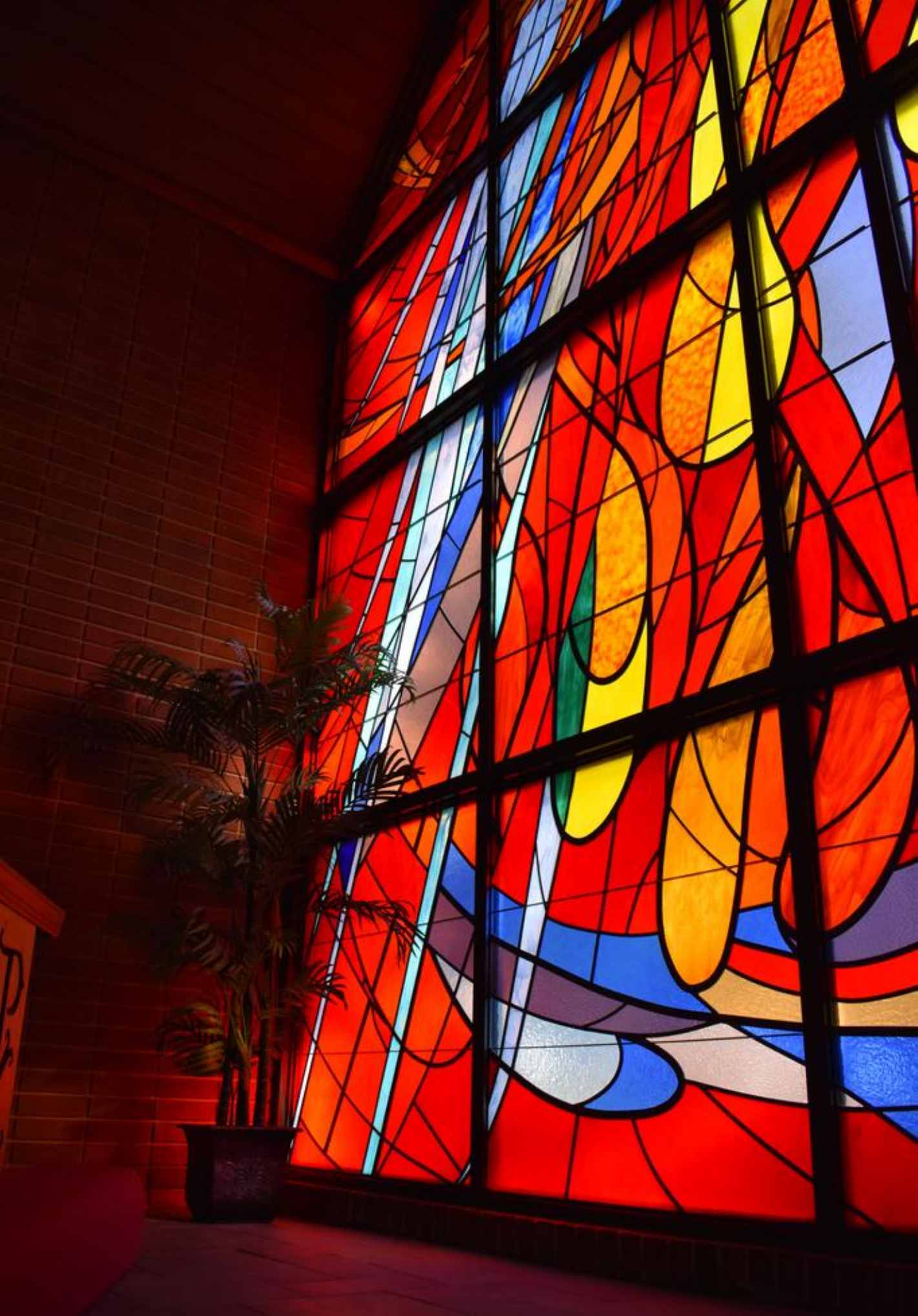}}} &
        \multicolumn{1}{l}{\rotatebox[origin=c]{90}{\shortstack[l]{\tiny 1\textsuperscript{st} iteration}}} &
        \noindent\parbox[c]{0.082\textwidth}{\includegraphics[width=0.082\textwidth]{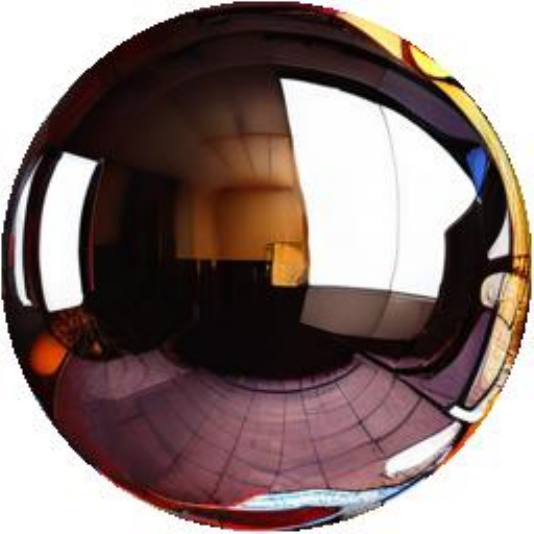}} & 
        \noindent\parbox[c]{0.082\textwidth}{\includegraphics[width=0.082\textwidth]{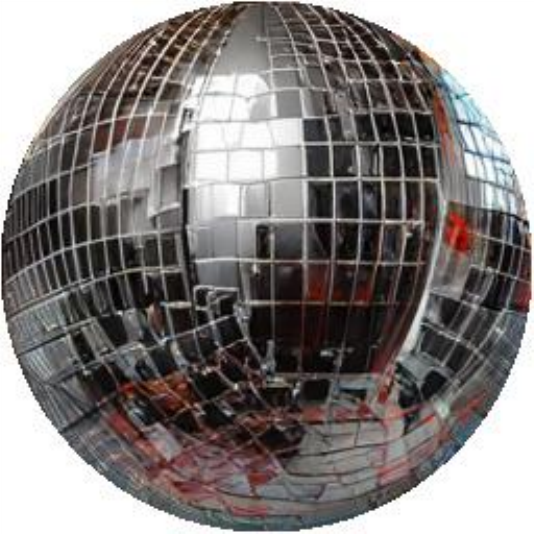}} & 
        \noindent\parbox[c]{0.082\textwidth}{\includegraphics[width=0.082\textwidth]{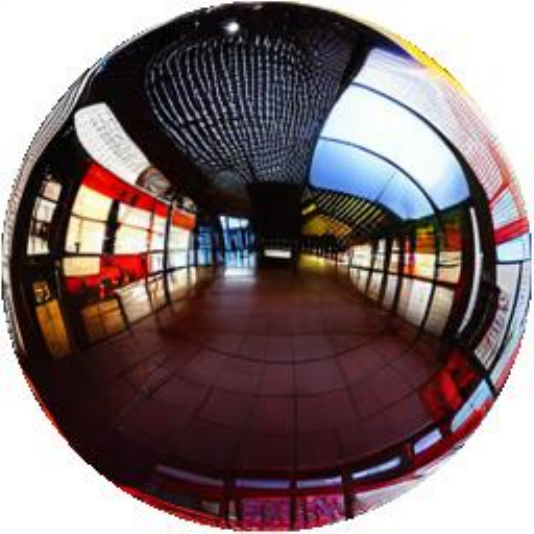}} & 
        \noindent\parbox[c]{0.082\textwidth}{\includegraphics[width=0.082\textwidth]{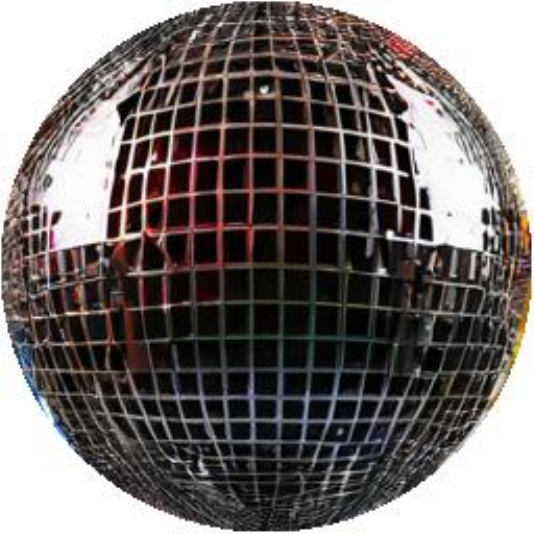}} & 
        \noindent\parbox[c]{0.082\textwidth}{\includegraphics[width=0.082\textwidth]{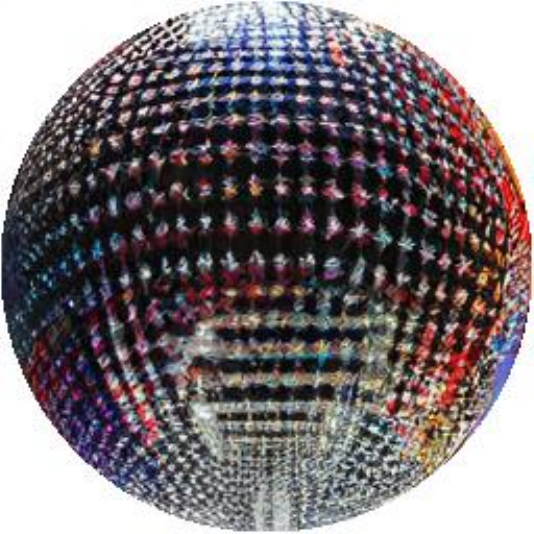}} & 
        \noindent\parbox[c]{0.082\textwidth}{\includegraphics[width=0.082\textwidth]{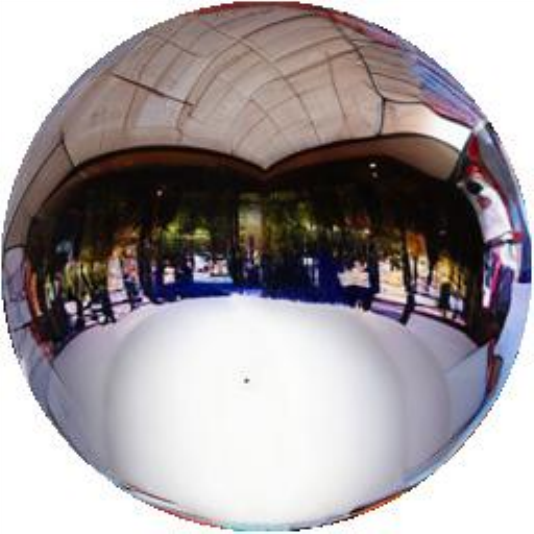}} & 
        \noindent\parbox[c]{0.082\textwidth}{\includegraphics[width=0.082\textwidth]{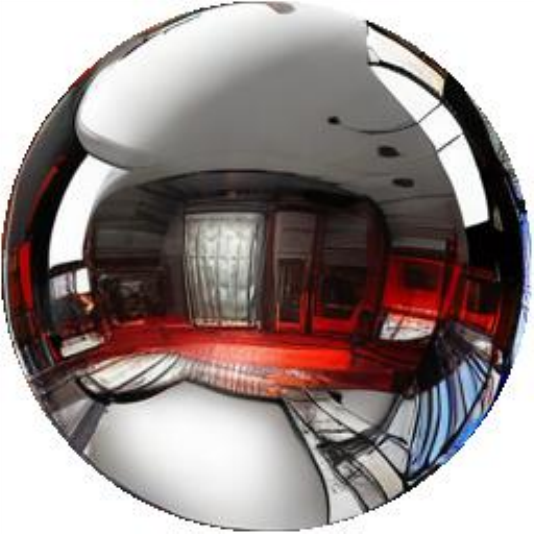}} & 
        \noindent\parbox[c]{0.082\textwidth}{\includegraphics[width=0.082\textwidth]{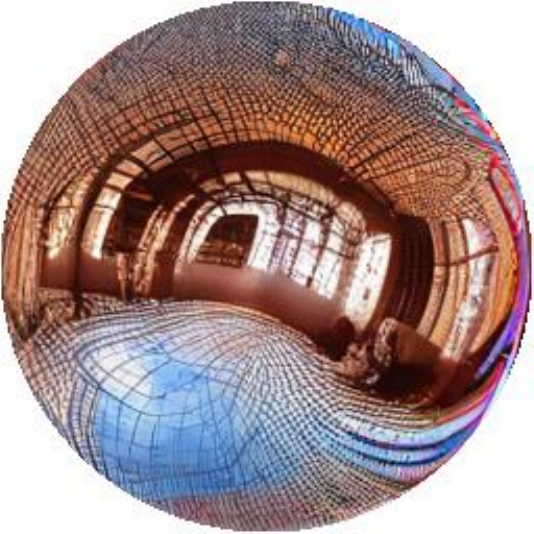}} & 
        \noindent\parbox[c]{0.082\textwidth}{\includegraphics[width=0.082\textwidth]{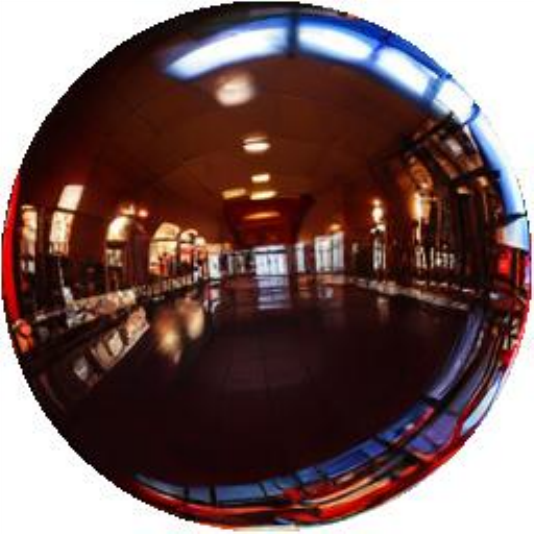}} & 
        \noindent\parbox[c]{0.082\textwidth}{\includegraphics[width=0.082\textwidth]{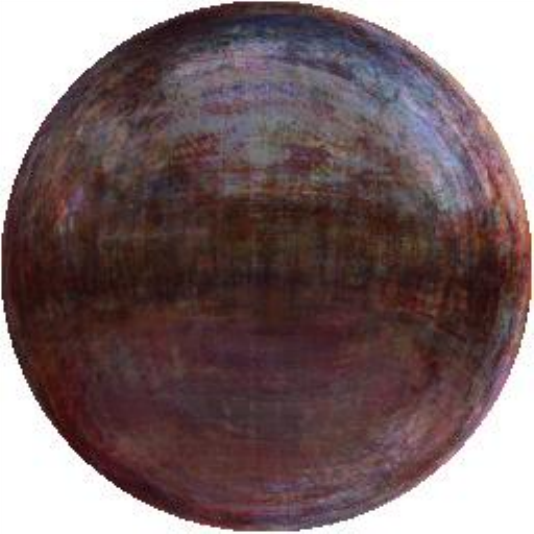}}

        \\ \cline{2-12}

        & 
        \multicolumn{1}{l}{\rotatebox[origin=c]{90}{\shortstack[l]{\tiny 2\textsuperscript{nd} iteration}}} &
        \noindent\parbox[c]{0.082\textwidth}{\includegraphics[width=0.082\textwidth]{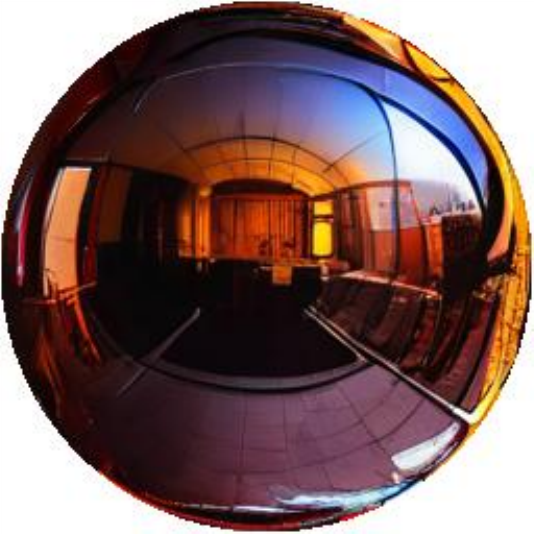}} & 
        \noindent\parbox[c]{0.082\textwidth}{\includegraphics[width=0.082\textwidth]{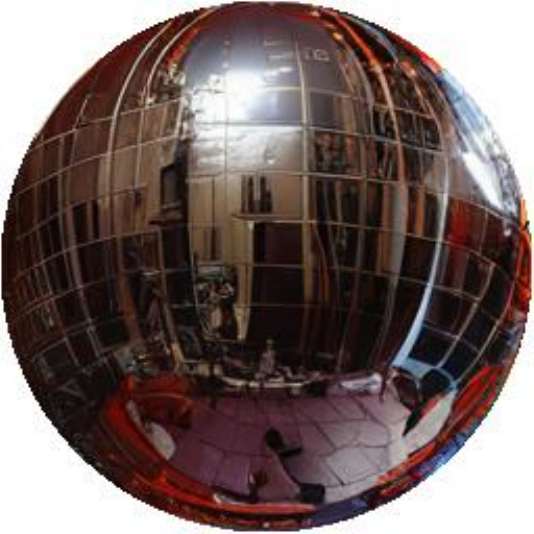}} & 
        \noindent\parbox[c]{0.082\textwidth}{\includegraphics[width=0.082\textwidth]{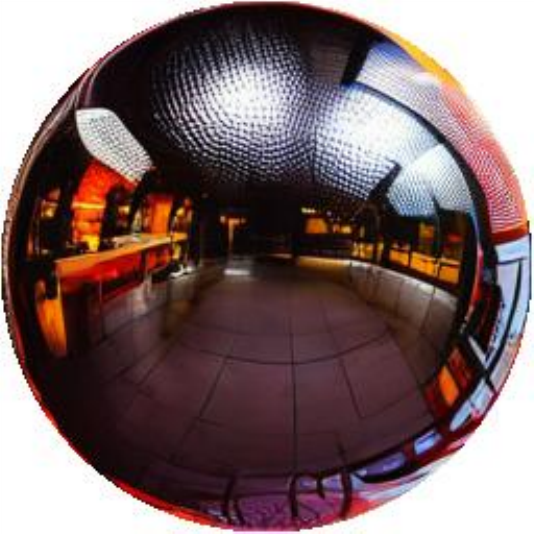}} & 
        \noindent\parbox[c]{0.082\textwidth}{\includegraphics[width=0.082\textwidth]{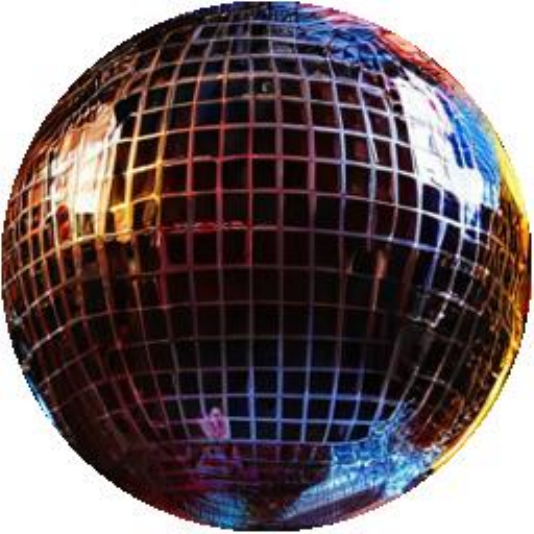}} & 
        \noindent\parbox[c]{0.082\textwidth}{\includegraphics[width=0.082\textwidth]{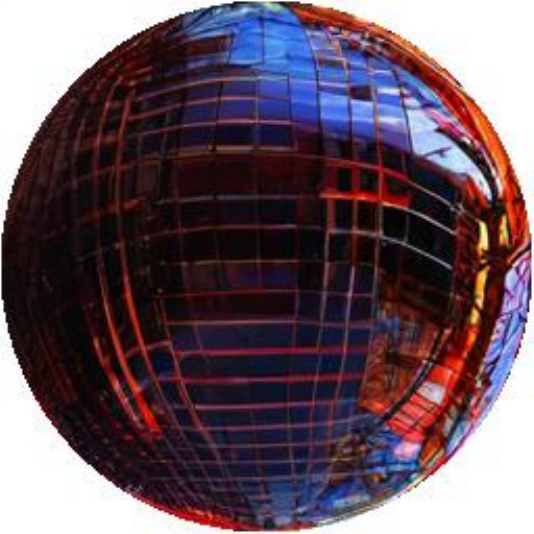}} & 
        \noindent\parbox[c]{0.082\textwidth}{\includegraphics[width=0.082\textwidth]{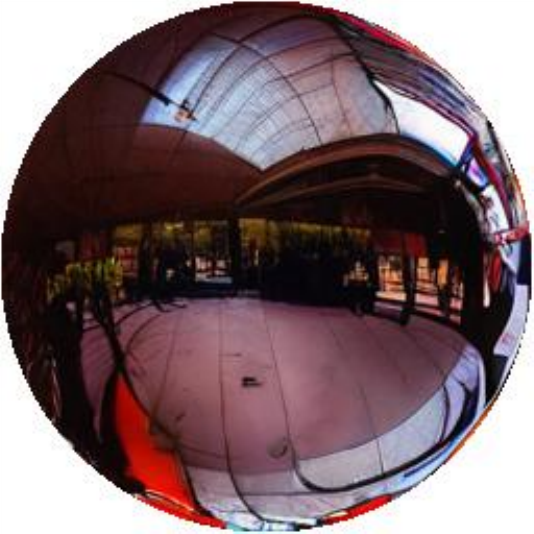}} & 
        \noindent\parbox[c]{0.082\textwidth}{\includegraphics[width=0.082\textwidth]{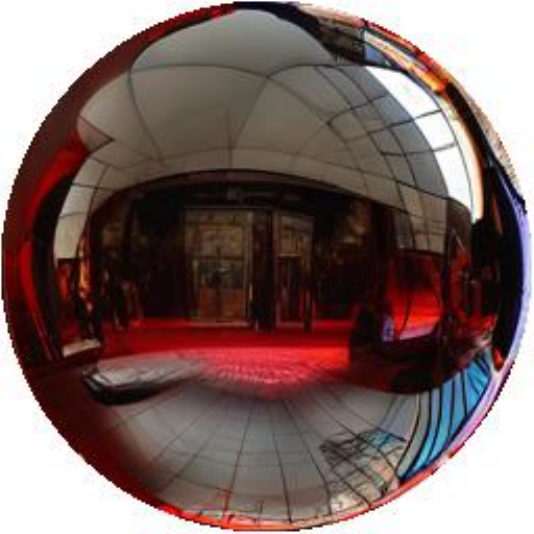}} & 
        \noindent\parbox[c]{0.082\textwidth}{\includegraphics[width=0.082\textwidth]{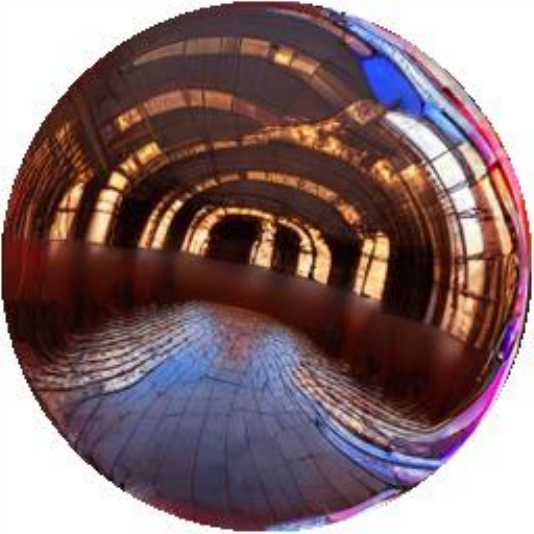}} & 
        \noindent\parbox[c]{0.082\textwidth}{\includegraphics[width=0.082\textwidth]{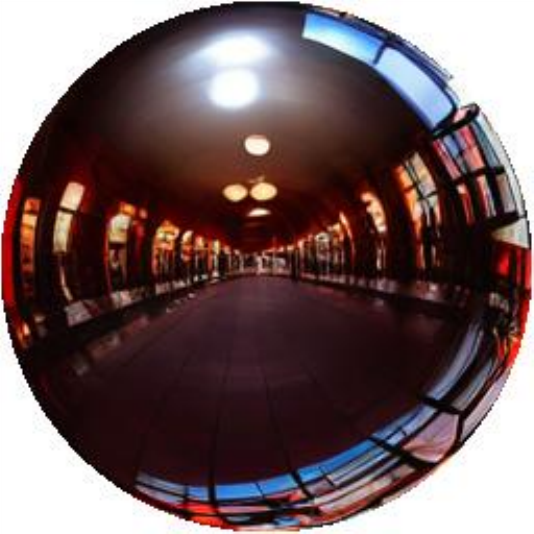}} & 
        \noindent\parbox[c]{0.082\textwidth}{\includegraphics[width=0.082\textwidth]{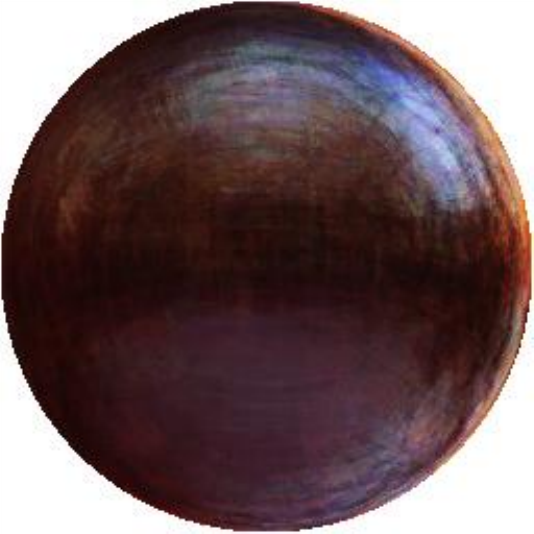}}

        \\

         &
        \multicolumn{1}{l}{\rotatebox[origin=c]{90}{\shortstack[l]{\tiny 3\textsuperscript{rd} iteration}}} &
        \noindent\parbox[c]{0.082\textwidth}{\includegraphics[width=0.082\textwidth]{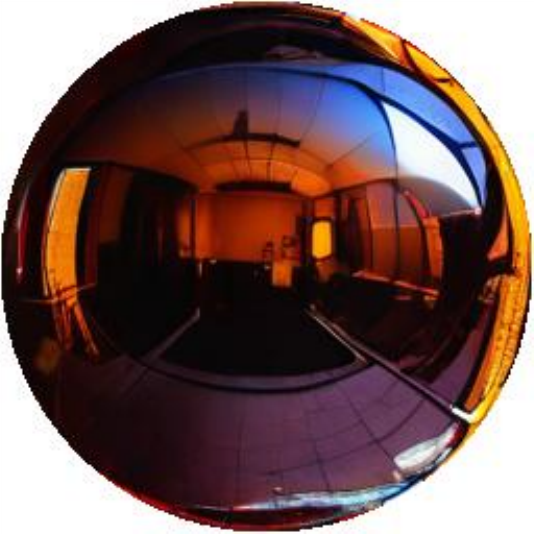}} & 
        \noindent\parbox[c]{0.082\textwidth}{\includegraphics[width=0.082\textwidth]{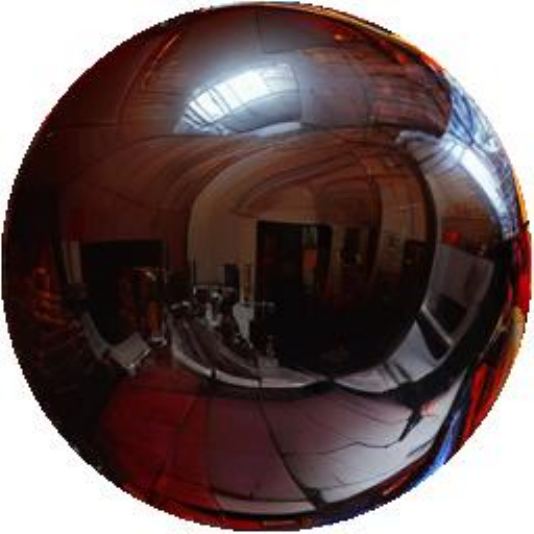}} & 
        \noindent\parbox[c]{0.082\textwidth}{\includegraphics[width=0.082\textwidth]{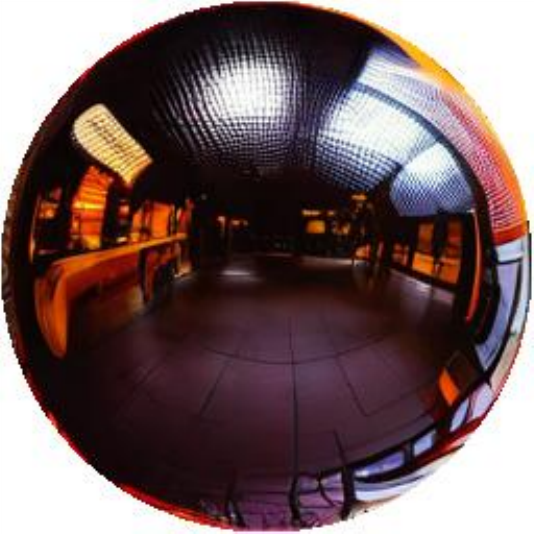}} & 
        \noindent\parbox[c]{0.082\textwidth}{\includegraphics[width=0.082\textwidth]{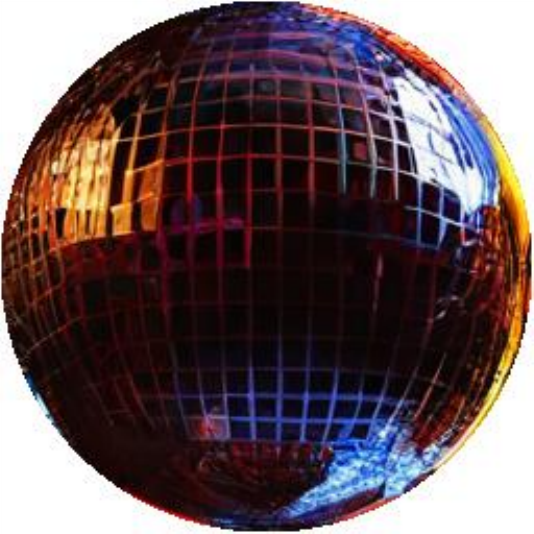}} & 
        \noindent\parbox[c]{0.082\textwidth}{\includegraphics[width=0.082\textwidth]{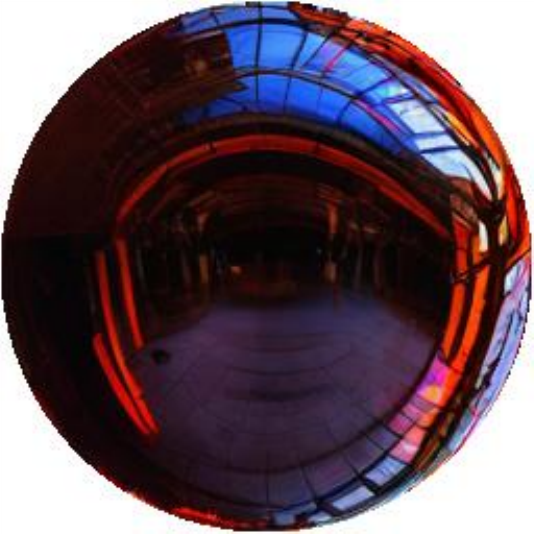}} & 
        \noindent\parbox[c]{0.082\textwidth}{\includegraphics[width=0.082\textwidth]{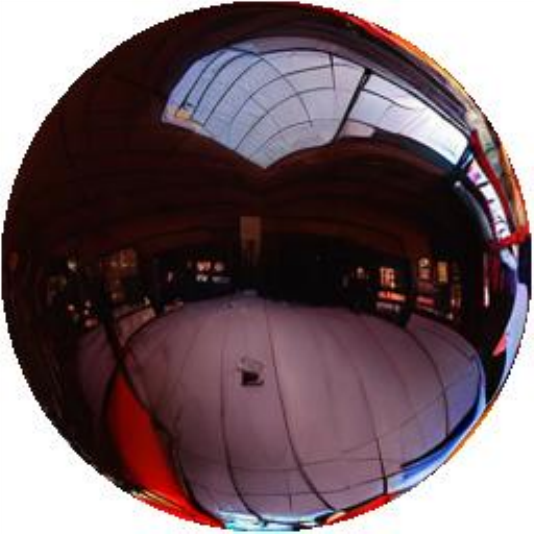}} & 
        \noindent\parbox[c]{0.082\textwidth}{\includegraphics[width=0.082\textwidth]{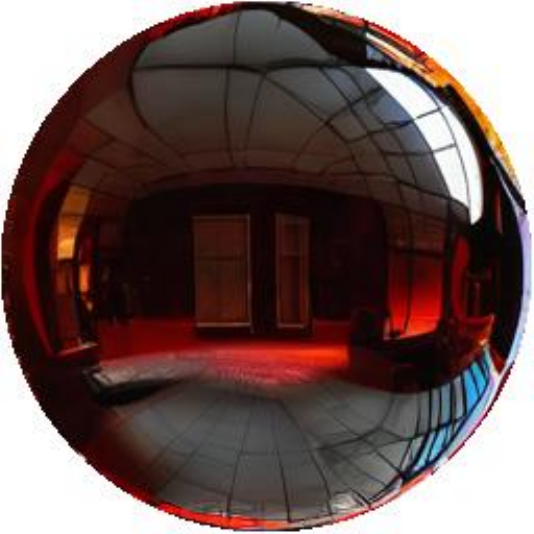}} & 
        \noindent\parbox[c]{0.082\textwidth}{\includegraphics[width=0.082\textwidth]{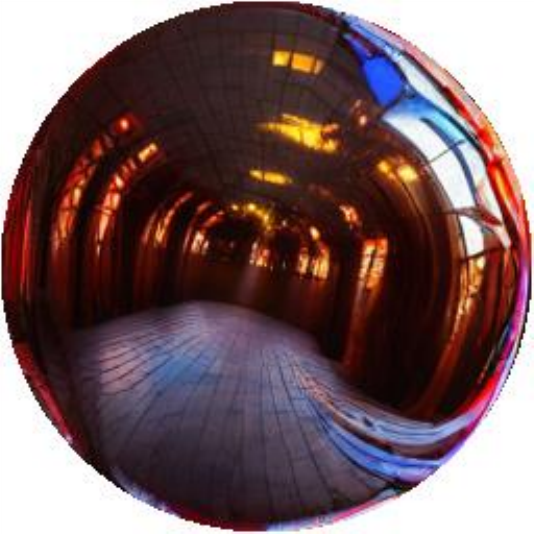}} & 
        \noindent\parbox[c]{0.082\textwidth}{\includegraphics[width=0.082\textwidth]{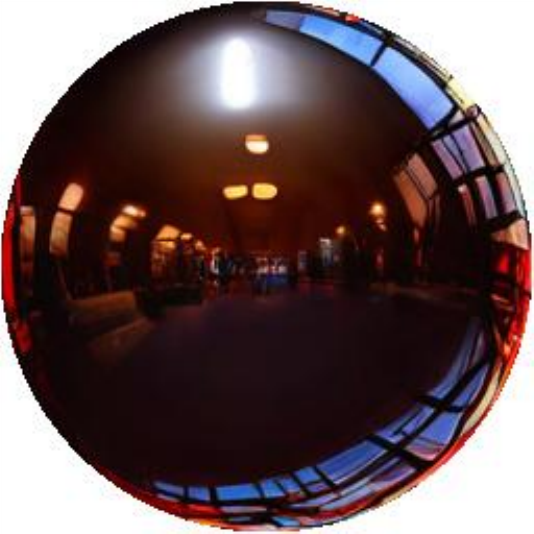}} & 
        \noindent\parbox[c]{0.082\textwidth}{\includegraphics[width=0.082\textwidth]{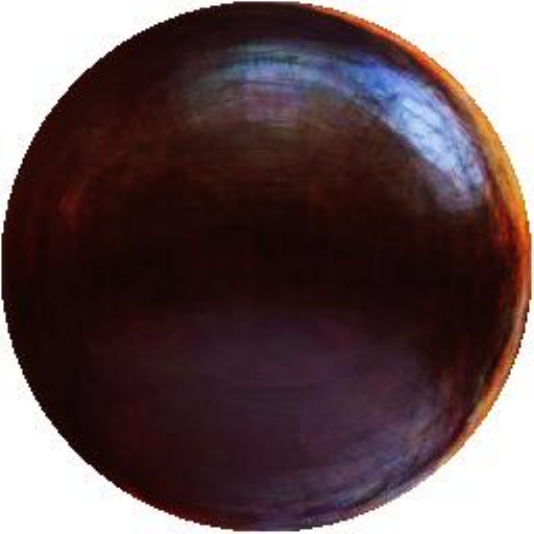}} 

        \\

         &
        \multicolumn{1}{l}{\rotatebox[origin=c]{90}{\shortstack[l]{\tiny 4\textsuperscript{th} iteration}}} &
        \noindent\parbox[c]{0.082\textwidth}{\includegraphics[width=0.082\textwidth]{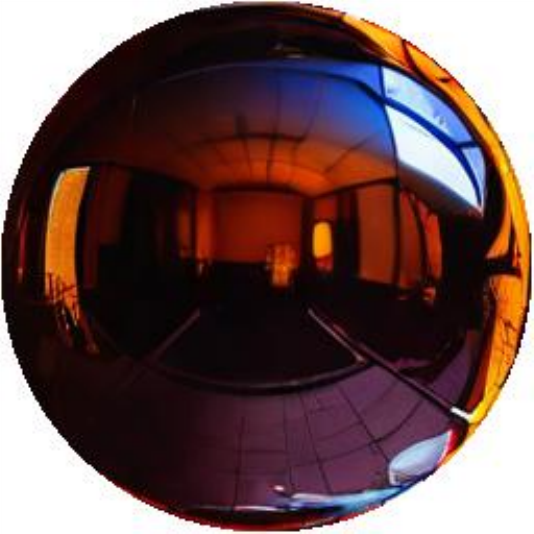}} & 
        \noindent\parbox[c]{0.082\textwidth}{\includegraphics[width=0.082\textwidth]{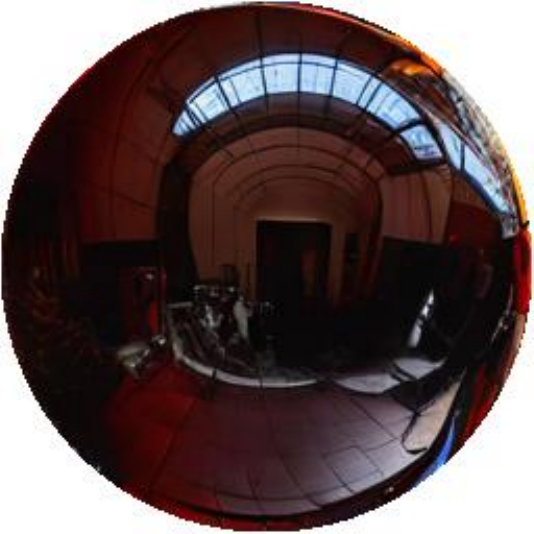}} & 
        \noindent\parbox[c]{0.082\textwidth}{\includegraphics[width=0.082\textwidth]{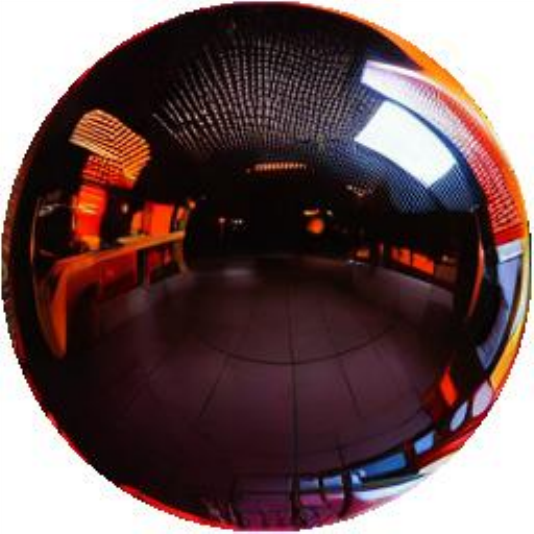}} & 
        \noindent\parbox[c]{0.082\textwidth}{\includegraphics[width=0.082\textwidth]{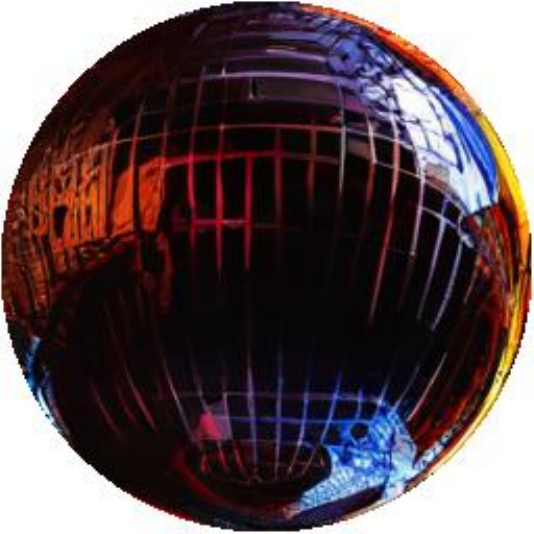}} & 
        \noindent\parbox[c]{0.082\textwidth}{\includegraphics[width=0.082\textwidth]{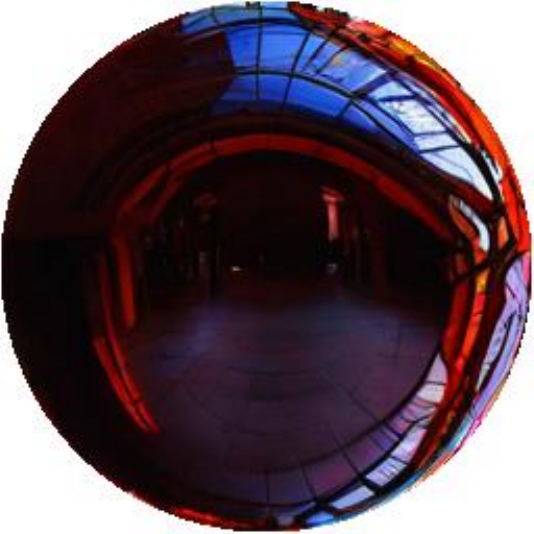}} & 
        \noindent\parbox[c]{0.082\textwidth}{\includegraphics[width=0.082\textwidth]{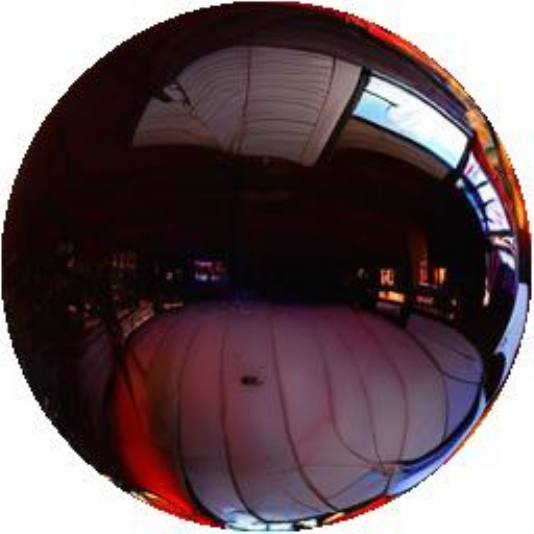}} & 
        \noindent\parbox[c]{0.082\textwidth}{\includegraphics[width=0.082\textwidth]{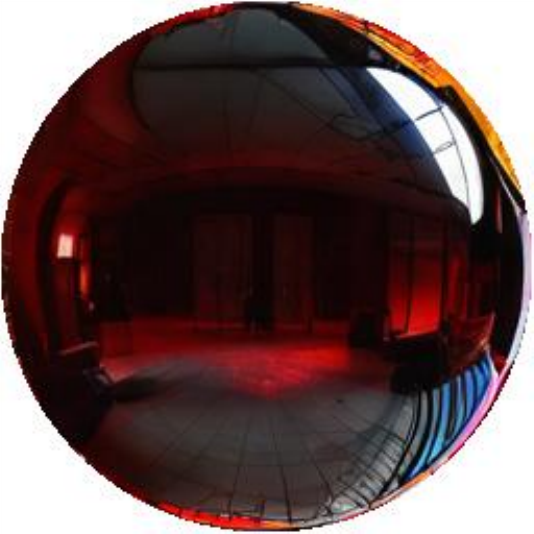}} & 
        \noindent\parbox[c]{0.082\textwidth}{\includegraphics[width=0.082\textwidth]{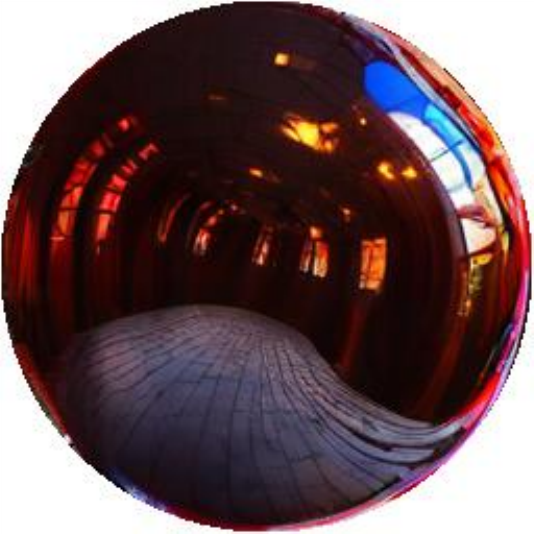}} & 
        \noindent\parbox[c]{0.082\textwidth}{\includegraphics[width=0.082\textwidth]{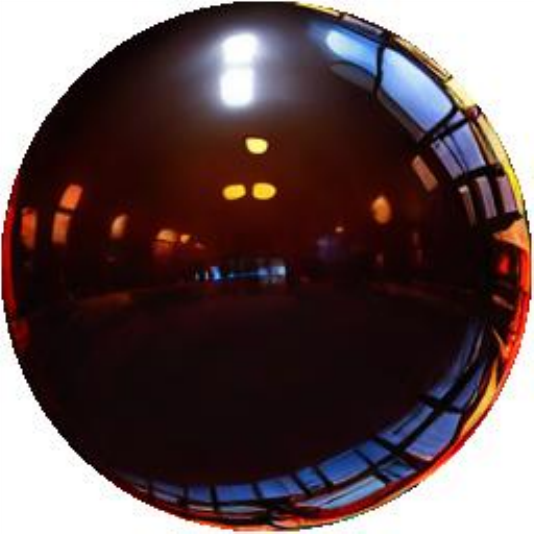}} & 
        \noindent\parbox[c]{0.082\textwidth}{\includegraphics[width=0.082\textwidth]{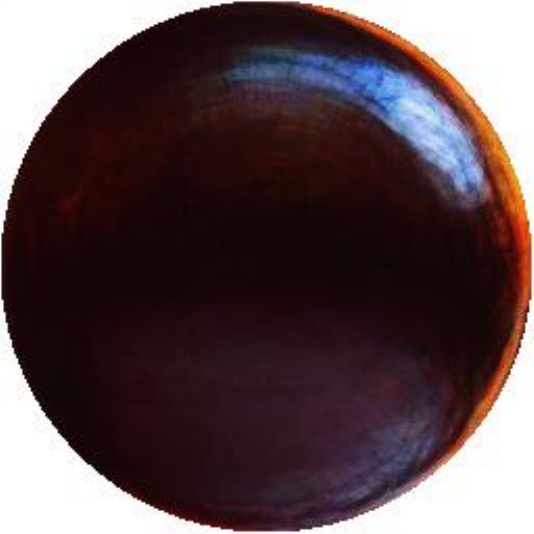}}

        \\

         &
        \multicolumn{1}{l}{\rotatebox[origin=c]{90}{\shortstack[l]{\tiny 5\textsuperscript{th} iteration}}} &
        \noindent\parbox[c]{0.082\textwidth}{\includegraphics[width=0.082\textwidth]{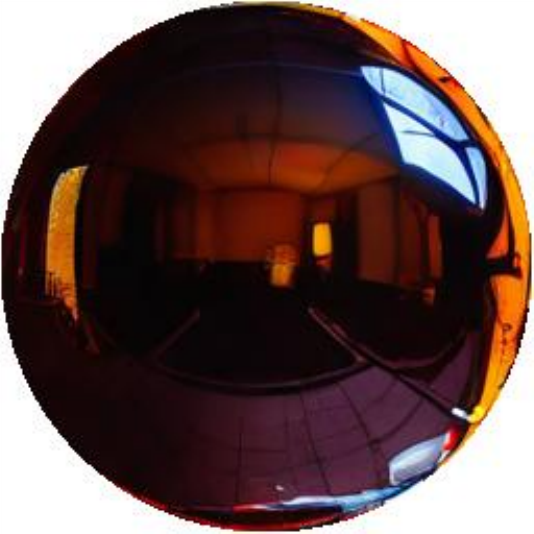}} & 
        \noindent\parbox[c]{0.082\textwidth}{\includegraphics[width=0.082\textwidth]{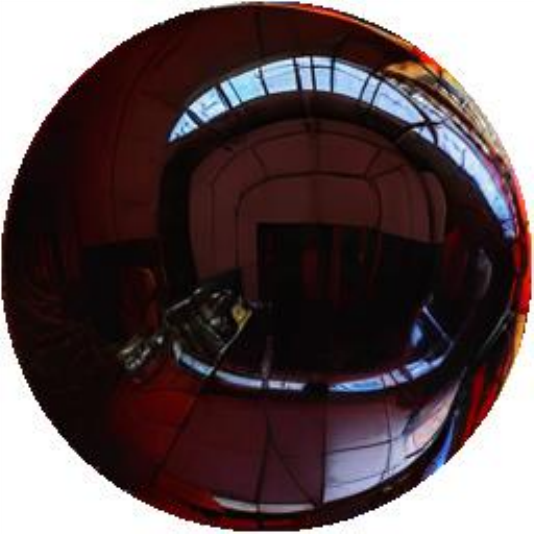}} & 
        \noindent\parbox[c]{0.082\textwidth}{\includegraphics[width=0.082\textwidth]{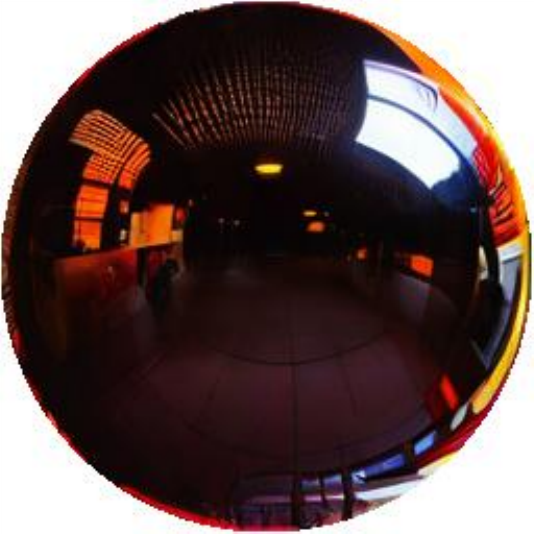}} & 
        \noindent\parbox[c]{0.082\textwidth}{\includegraphics[width=0.082\textwidth]{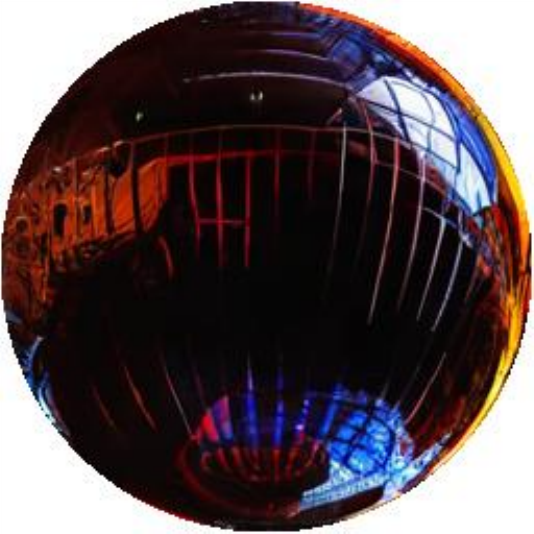}} & 
        \noindent\parbox[c]{0.082\textwidth}{\includegraphics[width=0.082\textwidth]{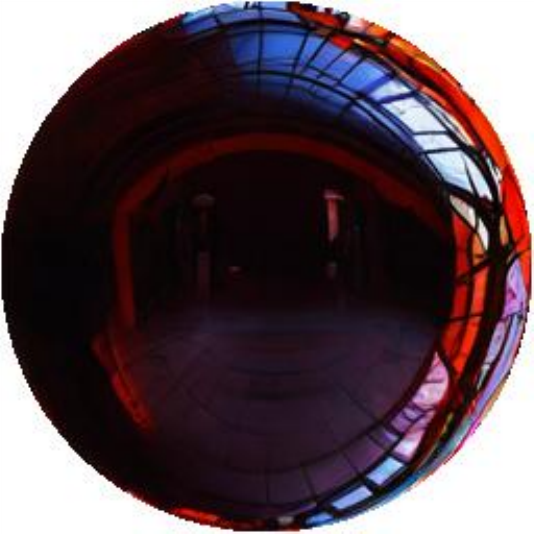}} & 
        \noindent\parbox[c]{0.082\textwidth}{\includegraphics[width=0.082\textwidth]{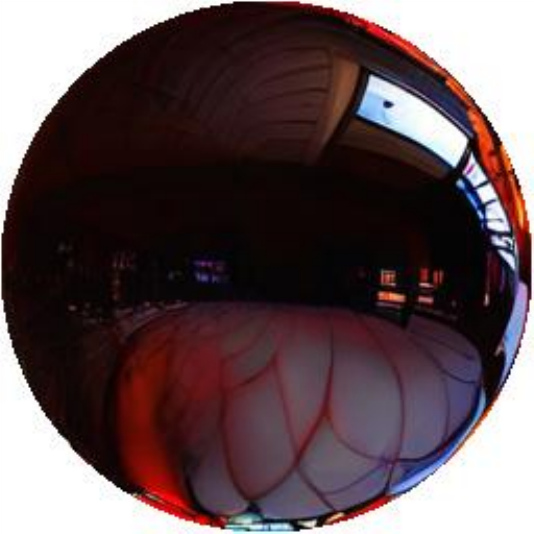}} & 
        \noindent\parbox[c]{0.082\textwidth}{\includegraphics[width=0.082\textwidth]{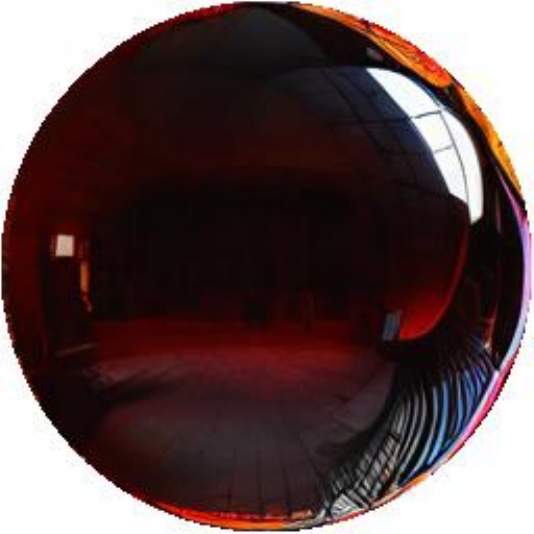}} & 
        \noindent\parbox[c]{0.082\textwidth}{\includegraphics[width=0.082\textwidth]{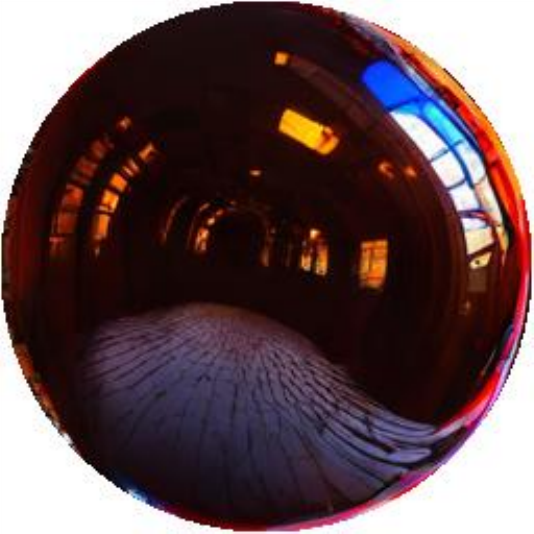}} & 
        \noindent\parbox[c]{0.082\textwidth}{\includegraphics[width=0.082\textwidth]{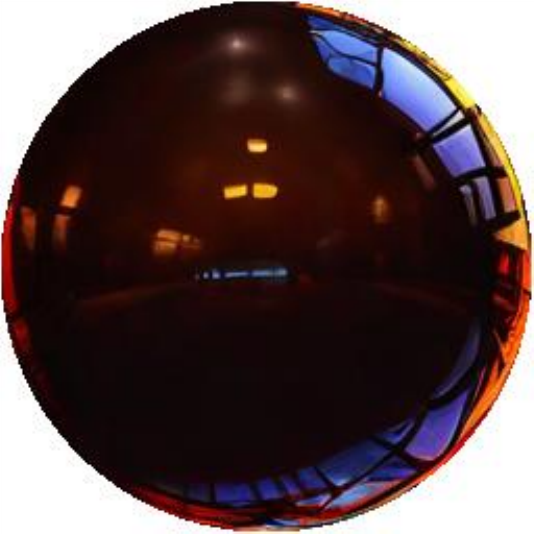}} & 
        \noindent\parbox[c]{0.082\textwidth}{\includegraphics[width=0.082\textwidth]{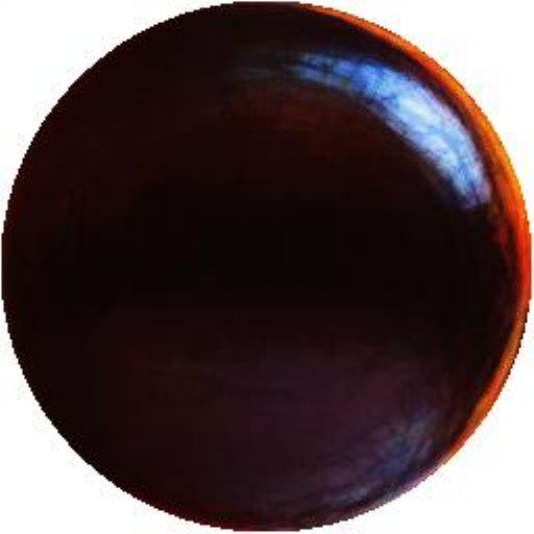}}

        \\

         &
        \multicolumn{1}{l}{\rotatebox[origin=c]{90}{\shortstack[l]{\tiny 6\textsuperscript{th} iteration}}} &
        \noindent\parbox[c]{0.082\textwidth}{\includegraphics[width=0.082\textwidth]{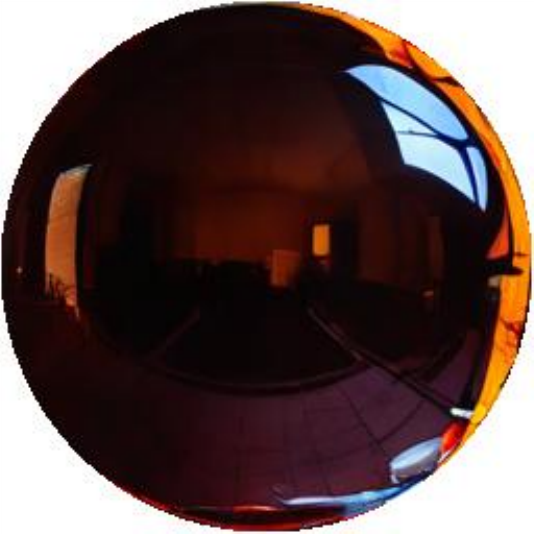}} & 
        \noindent\parbox[c]{0.082\textwidth}{\includegraphics[width=0.082\textwidth]{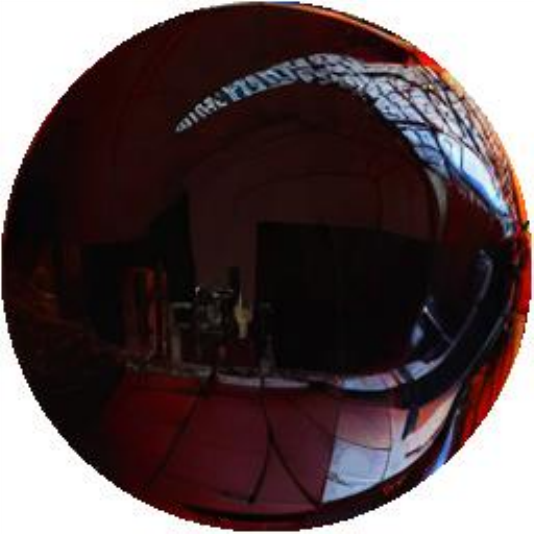}} & 
        \noindent\parbox[c]{0.082\textwidth}{\includegraphics[width=0.082\textwidth]{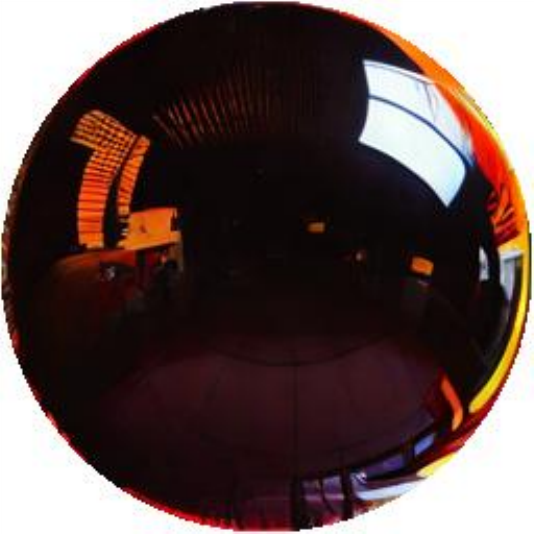}} & 
        \noindent\parbox[c]{0.082\textwidth}{\includegraphics[width=0.082\textwidth]{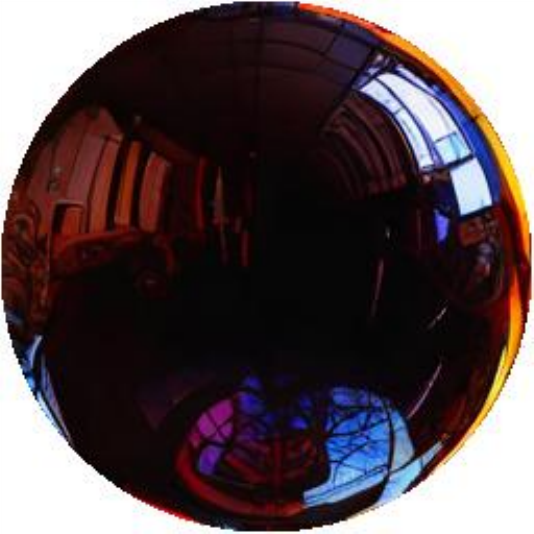}} & 
        \noindent\parbox[c]{0.082\textwidth}{\includegraphics[width=0.082\textwidth]{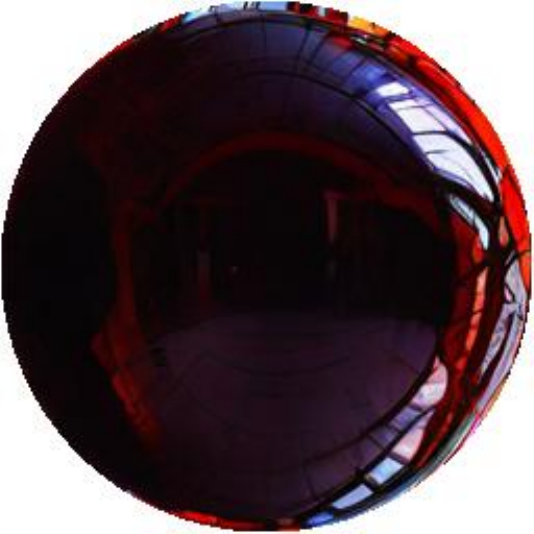}} & 
        \noindent\parbox[c]{0.082\textwidth}{\includegraphics[width=0.082\textwidth]{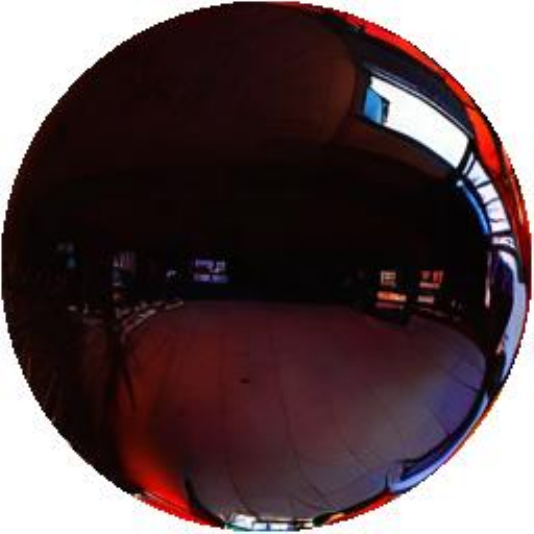}} & 
        \noindent\parbox[c]{0.082\textwidth}{\includegraphics[width=0.082\textwidth]{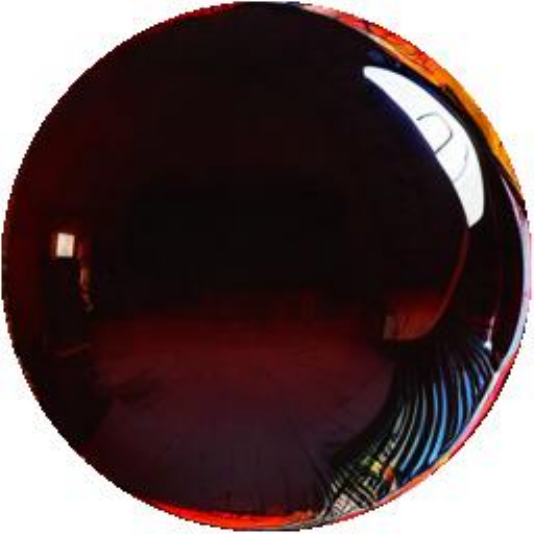}} & 
        \noindent\parbox[c]{0.082\textwidth}{\includegraphics[width=0.082\textwidth]{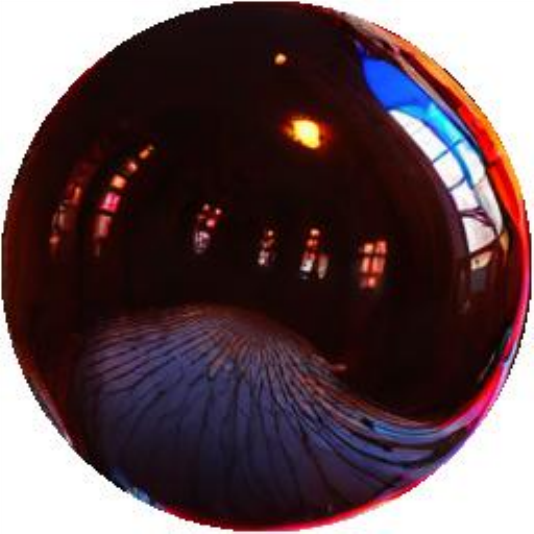}} & 
        \noindent\parbox[c]{0.082\textwidth}{\includegraphics[width=0.082\textwidth]{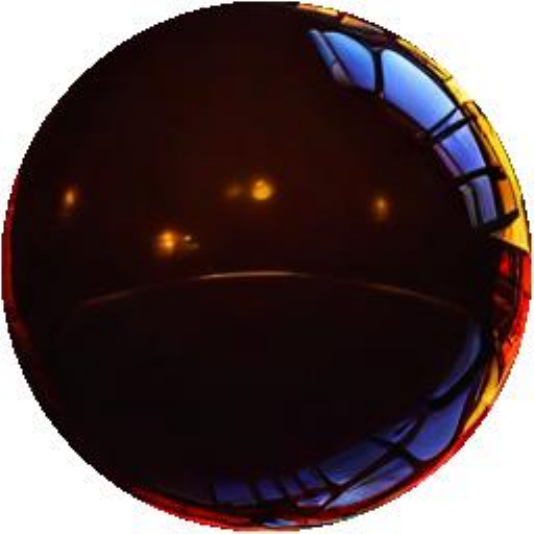}} & 
        \noindent\parbox[c]{0.082\textwidth}{\includegraphics[width=0.082\textwidth]{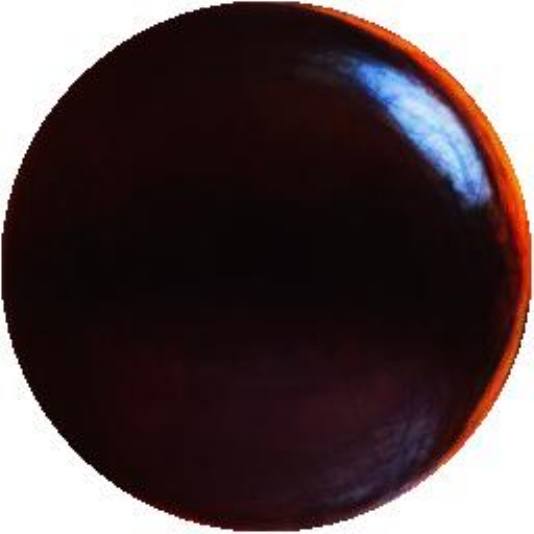}}

        \\

        \end{tabu}
    \caption{
    Repeatedly applying our iterative inpainting algorithm gradually produces chrome balls with better light estimation and fix degenerate balls such as Pred\#2, Pred\#4, and Pred\#5.
    }
    \label{fig:compare_median_distribution_aba_infinite}
\end{figure*}

Figure \ref{fig:compare_median_distribution_aba} presents additional results demonstrating how our iterative inpainting algorithm improves the consistency and quality of the generated chrome balls after one iteration. Figure \ref{fig:compare_median_distribution_aba_infinite} presents results with more iterations. Note that the experiments on DiffusionLight in our main paper were limited to two iterations due to resource constraints. 


\tabulinesep=2pt
\tabulinesep=0.5pt

\subsection{Ablation on Exposure LoRA Training} \label{appendix:aba_lora}

\subsubsection{Additional details on training set generation.} To generate training panoramas from text prompts as mentioned in Section \ref{sec:diffusionlight}, we use Text2Light \cite{chen2022text2light}, using prompts from its official GitHub repository\footnote{\url{https://github.com/FrozenBurning/Text2Light}} and additional prompts generated by Chat-GPT 3.5\footnote{\url{https://chat.openai.com/}} from short instructions and examples.

To eliminate near-duplicate samples, we use the perceptual hashing algorithm implemented in the \texttt{ imagededup} package \footnote{\url{https://github.com/idealo/imagededup}}. This process yielded a dataset of 1,412 unique HDR panoramas at resolution of $2048 \times 4196$ pixels. We used orthographic projection and a 60$^\circ$ field of view.

\subsubsection{Range of timesteps for LoRA training.} We sampled timesteps $t \sim U(900,999)$ as we observed that the overall lighting information is determined earlier in the sampling process (see Figure \ref{fig:main_lightinfo_early_stage}). This choice helped speed up training. In Table \ref{tab:ablation_lora_t}, we compare this choice to training with $t \sim U(0,999)$ and $t \sim U(500,999)$ given the same number of training iterations and report scores on the validation comprising 100 scenes from the Laval Indoor dataset \cite{garder2017lavelindoor} and 100 scenes from the Poly Haven dataset \cite{polyhaven}. Our choice of 900-999 yielded the best performance across all three metrics.


\begin{table}[]
\centering
\small
\caption{Ablation study on sampled timesteps for exposure LoRA training.
}
\label{tab:ablation_lora_t}
\begin{tabular}{
    l@{\hspace{5pt}}
    l@{\hspace{5pt}}
    c@{\hspace{5pt}}
    c@{\hspace{5pt}}
    c
}
\toprule
\textbf{Sphere} & \begin{tabular}[c]{@{}c@{}}\textbf{Denoising}\\ \textbf{step}\end{tabular} & \textbf{si-RMSE} $\downarrow$ & \begin{tabular}[c]{@{}c@{}}\textbf{Angular}\\ \textbf{Error}\end{tabular} $\downarrow$ & \begin{tabular}[c]{@{}c@{}}\textbf{Normalized}\\ \textbf{RMSE}\end{tabular} $\downarrow$ \\
\midrule

Diffuse & $\vect{x}_0:\vect{x}_{999}$ & 0.194 & 3.322 & 0.292\\
        & $\vect{x}_{500}:\vect{x}_{999}$ & 0.188 & 3.260 & 0.284\\
        & $\vect{x}_{900}:\vect{x}_{999}$ & \colorbox{tabfirst}{0.156} & \colorbox{tabfirst}{2.956} & \colorbox{tabfirst}{0.246} \\[0.3ex]
\hline\\[-2.2ex]
Matte & $\vect{x}_0:\vect{x}_{999}$ & 0.449 & 4.121 & 0.436\\
        & $\vect{x}_{500}:\vect{x}_{999}$ & 0.452 & 4.008 & 0.438\\
        & $\vect{x}_{900}:\vect{x}_{999}$ & \colorbox{tabfirst}{0.385} & \colorbox{tabfirst}{3.575} & \colorbox{tabfirst}{0.371} \\[0.3ex]
  \hline\\[-2.2ex]
Mirror & $\vect{x}_0:\vect{x}_{999}$ & 0.727 & 6.292 & 0.483\\
        & $\vect{x}_{500}:\vect{x}_{999}$ & 0.730 & 6.277 & 0.479\\
        & $\vect{x}_{900}:\vect{x}_{999}$ & \colorbox{tabfirst}{0.656} & \colorbox{tabfirst}{5.464} & \colorbox{tabfirst}{0.431} \\
\bottomrule
\end{tabular}
\end{table}

\subsubsection{LoRA scale.} We conducted an experiment to assess the effect of using different LoRA scales. Here, the LoRA scale refers to the $\alpha$ value in the weight update equation: $\vect{W}^{\prime} = \vect{W} + \alpha \Delta \vect{W}$, where $\vect{W}^{\prime}$ is the new weight for inference, $\vect{W}$ is the original weight of SDXL, $\Delta \vect{W}$ is the weight update from LoRA. (See Section \ref{sec:prelim} for a brief background on LoRA.) In Table \ref{tab:ldr-lorascale}, we report scores computed on EV0 LDR chrome balls evaluated on scenes in Poly Haven, which were never part of Text2Light's training set. In our implementation, we use the LoRA scale of 0.75, which has the best si-RMSE and angular error scores. 

\begin{table}[!h]
\centering
\small
\caption{Ablation study on exposure LoRA scales}
\label{tab:ldr-lorascale}
\begin{tabular}{
    l@{\hspace{20pt}}
    c@{\hspace{5pt}}
    c@{\hspace{5pt}}
    c
}
\toprule
\textbf{LoRA scale} & \textbf{RMSE} $\downarrow$ & \textbf{si-RMSE} $\downarrow$ & \begin{tabular}[c]{@{}c@{}}\textbf{Angular}\\ \textbf{Error}\end{tabular} $\downarrow$\\
\midrule

 0.00 & 0.232 & 0.327 & 6.189\\
 0.25 & 0.220 & 0.307 & 6.287\\
 0.50 & 0.211 & 0.303 & 6.180\\
 \textbf{0.75} & 0.204 & \colorbox{tabfirst}{0.300} & \colorbox{tabfirst}{6.109}\\
 1.00 & \colorbox{tabfirst}{0.199} & 0.303 & 6.267\\

\bottomrule
\end{tabular}
\end{table}

\subsubsection{Effectiveness of continuous LoRA.} As described in Section \ref{sec:diffusionlight}, we train a single \textit{continuous} LoRA for multiple EVs by conditioning it on an interpolated text prompt embedding instead of training multiple LoRAs for individual EVs. This strategy helps preserve the overall scene structure across exposures due to weight sharing. 

\begin{figure*}
    \centering
    \begin{tabu} to \textwidth {
        @{}
        c@{}
        *{10}{c}
        @{}
    }
        
        \multicolumn{1}{l}{\rotatebox[origin=c]{90}{\shortstack[l]{\scriptsize Input}}} &
        \multicolumn{2}{c}{
            \noindent\parbox[c]{0.19\textwidth}{\includegraphics[width=0.19\textwidth]{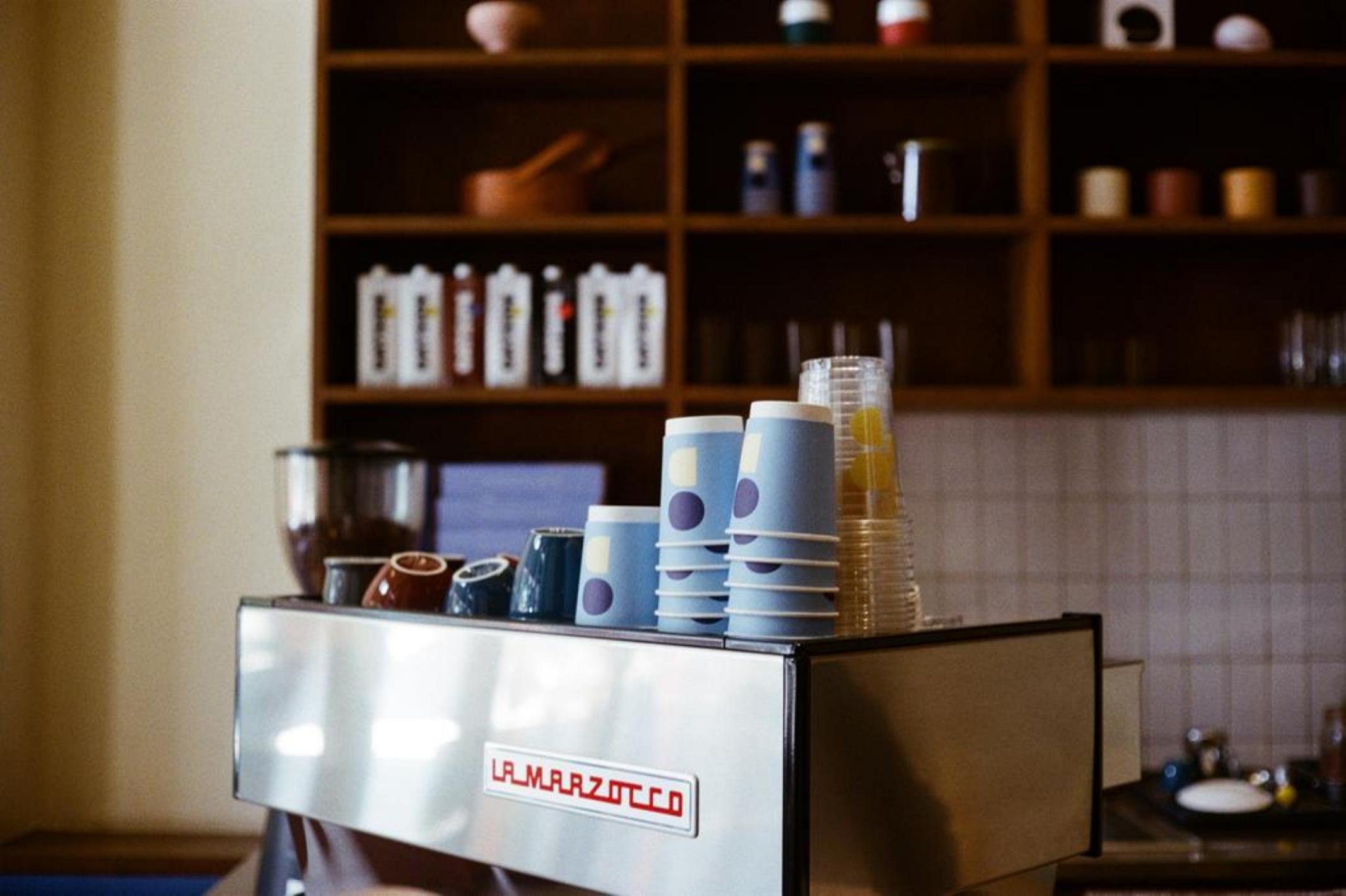}}
        } & 
        \multicolumn{2}{c}{
            \noindent\parbox[c]{0.19\textwidth}{\includegraphics[width=0.19\textwidth]{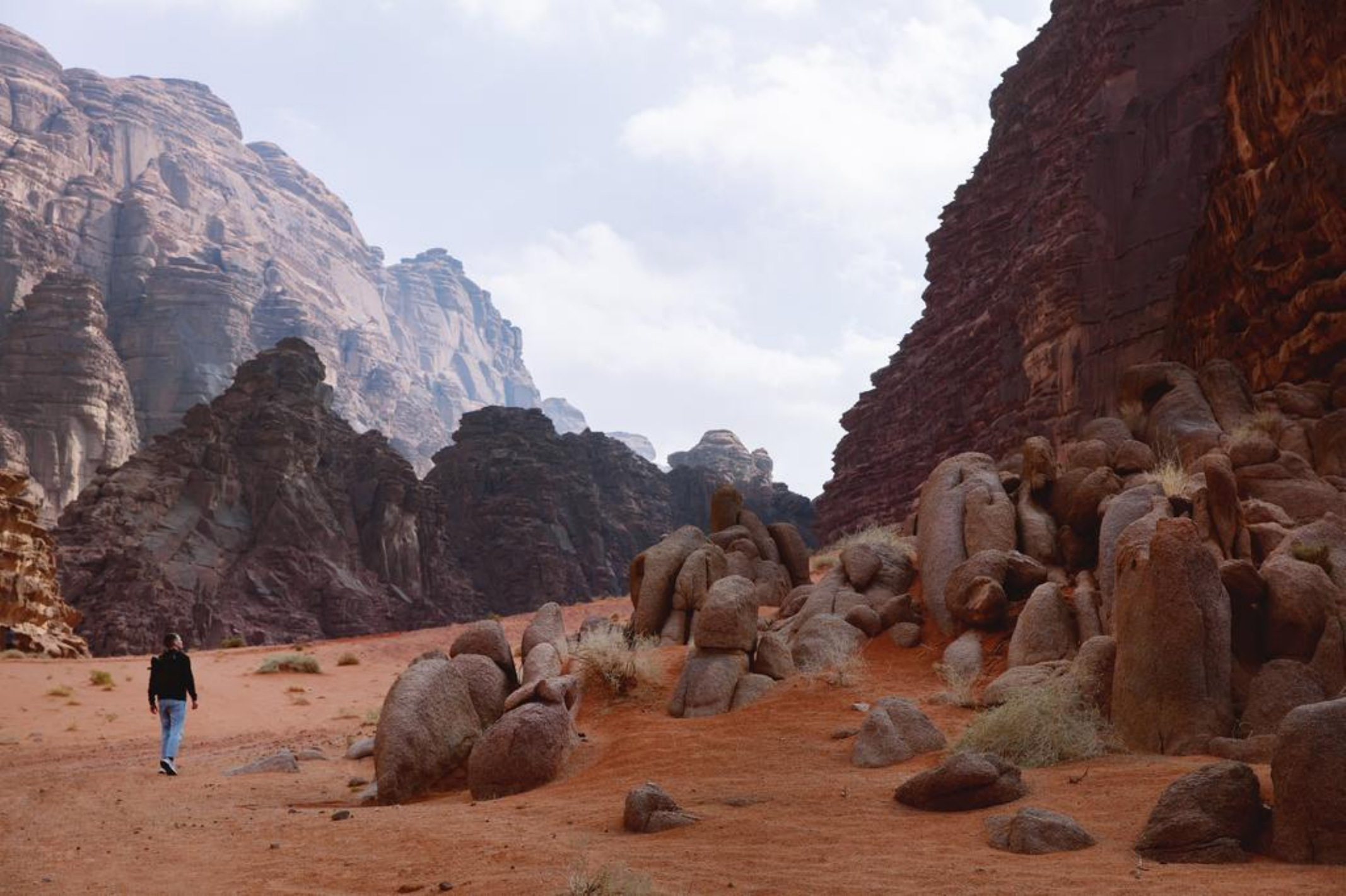}}
        } & 
        \multicolumn{2}{c}{
            \noindent\parbox[c]{0.19\textwidth}{\includegraphics[width=0.19\textwidth]{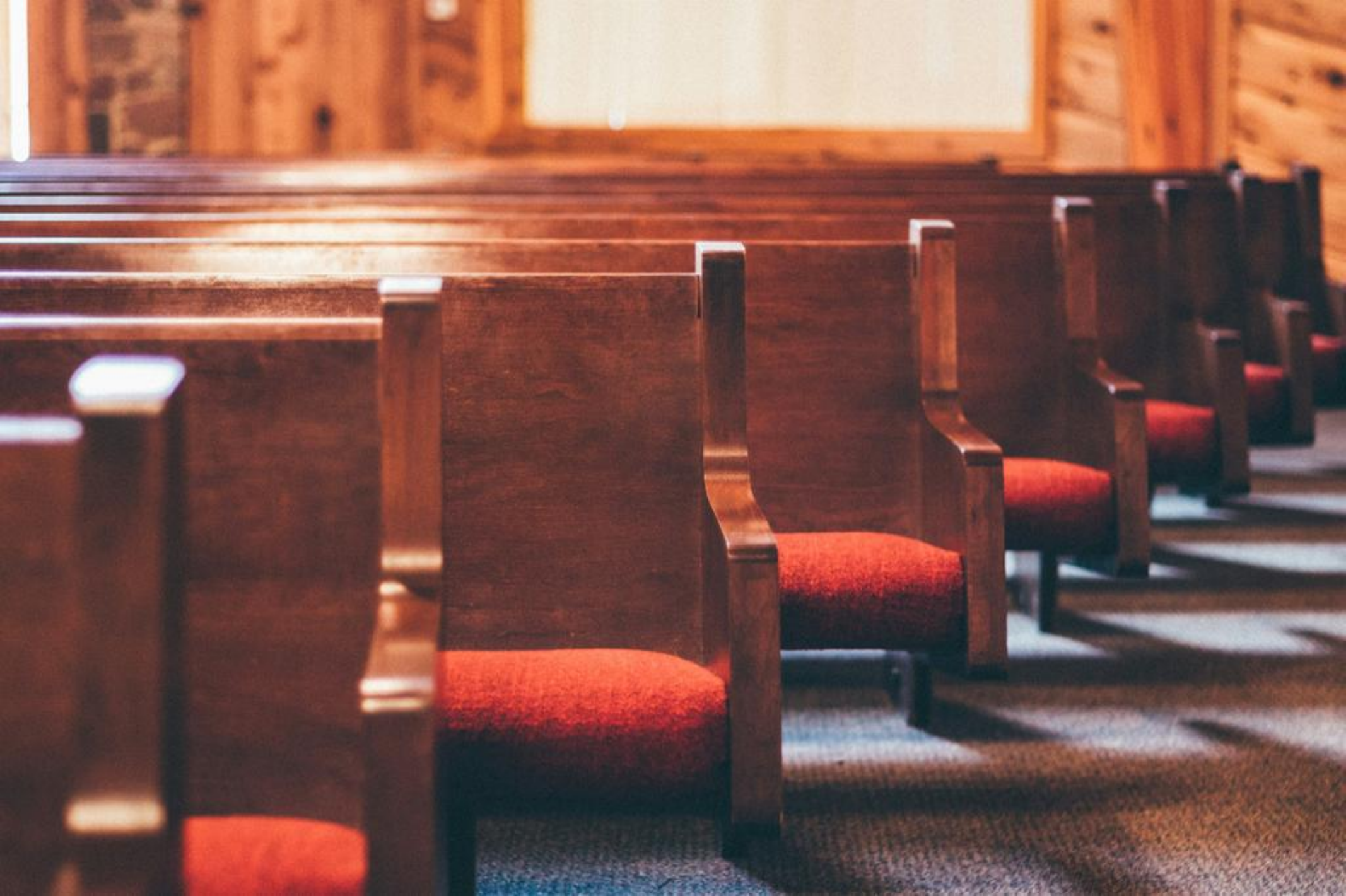}}
        } & 
        \multicolumn{2}{c}{
            \noindent\parbox[c]{0.19\textwidth}{\includegraphics[width=0.19\textwidth]{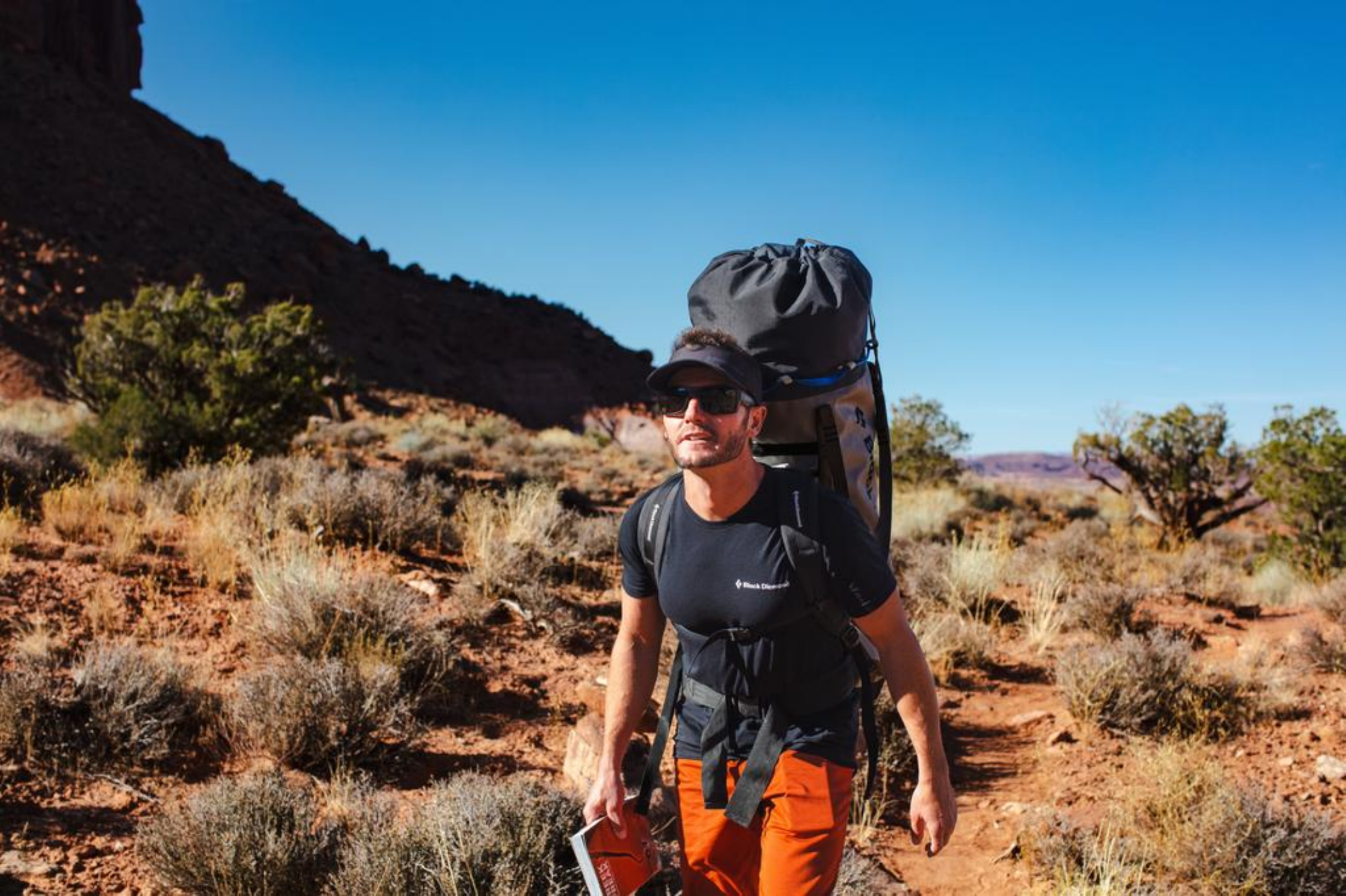}}
        } & 
        \multicolumn{2}{c}{
            \noindent\parbox[c]{0.19\textwidth}{\includegraphics[width=0.19\textwidth]{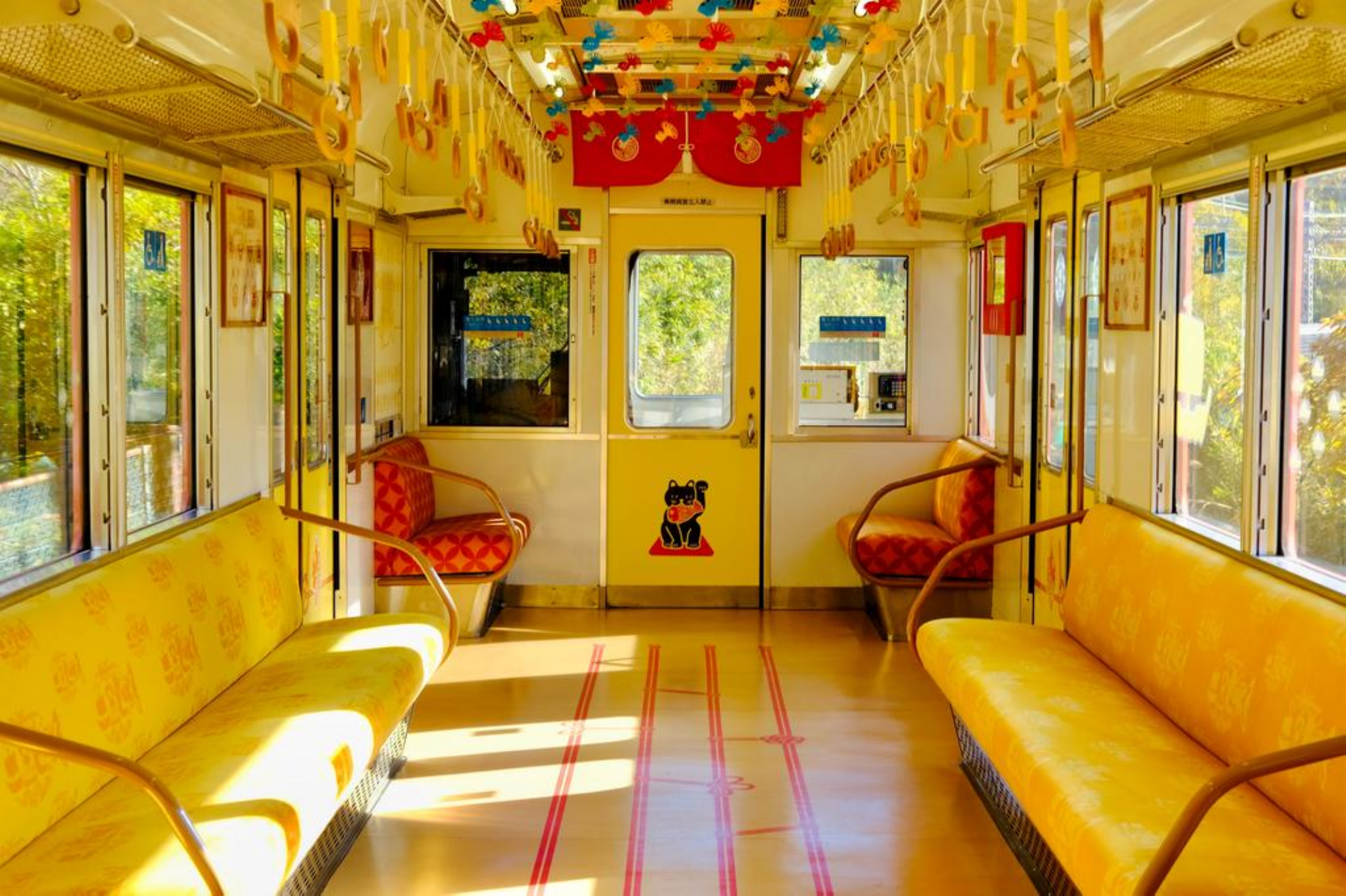}}
        }

        \\

        &
        \multicolumn{1}{c}{\shortstack{\scriptsize 3 LoRAs}} & 
        \multicolumn{1}{c}{\shortstack{\scriptsize Cont. LoRA}} &
        \multicolumn{1}{c}{\shortstack{\scriptsize 3 LoRAs}} & 
        \multicolumn{1}{c}{\shortstack{\scriptsize Cont. LoRA}} &
        \multicolumn{1}{c}{\shortstack{\scriptsize 3 LoRAs}} & 
        \multicolumn{1}{c}{\shortstack{\scriptsize Cont. LoRA}} &
        \multicolumn{1}{c}{\shortstack{\scriptsize 3 LoRAs}} & 
        \multicolumn{1}{c}{\shortstack{\scriptsize Cont. LoRA}} &
        \multicolumn{1}{c}{\shortstack{\scriptsize 3 LoRAs}} & 
        \multicolumn{1}{c}{\shortstack{\scriptsize Cont. LoRA}}
        \\

        \multicolumn{1}{l}{\rotatebox[origin=c]{90}{\shortstack[l]{\scriptsize EV0}}} &
        \noindent\parbox[c]{0.095\textwidth}{\includegraphics[width=0.095\textwidth]{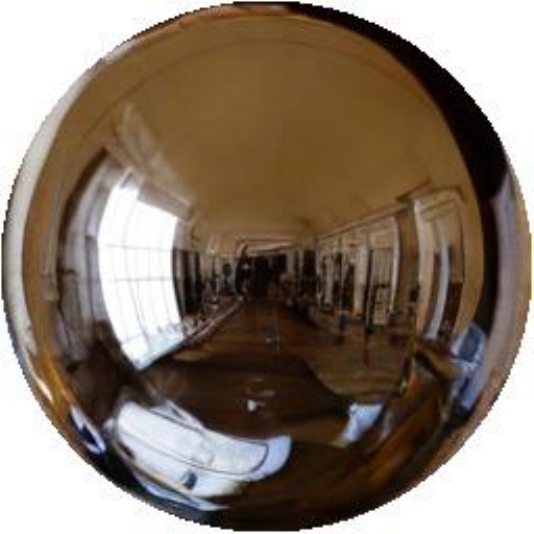}} & 
        \noindent\parbox[c]{0.095\textwidth}{\includegraphics[width=0.095\textwidth]{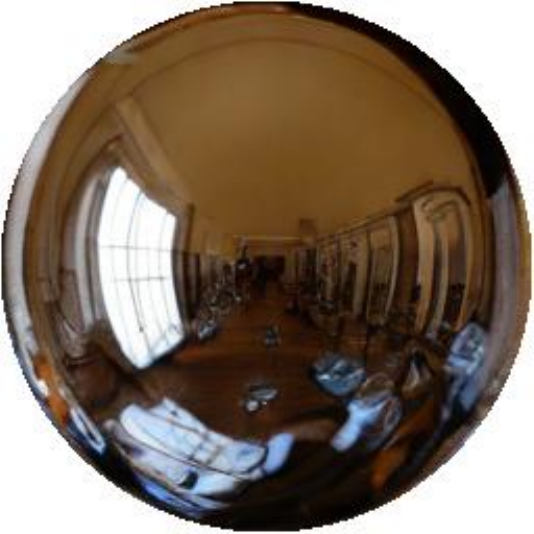}} & 
        \noindent\parbox[c]{0.095\textwidth}{\includegraphics[width=0.095\textwidth]{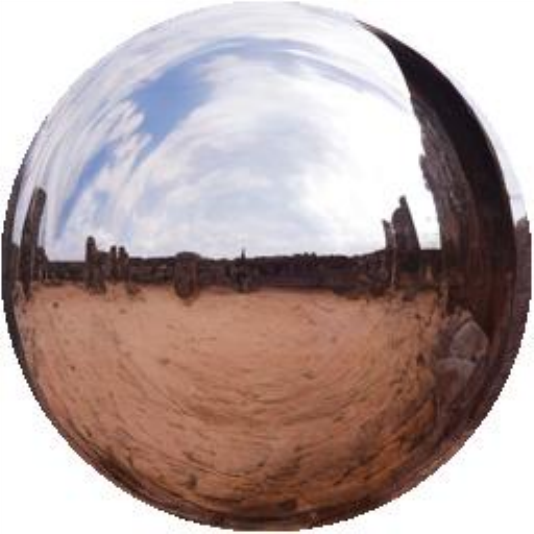}} & 
        \noindent\parbox[c]{0.095\textwidth}{\includegraphics[width=0.095\textwidth]{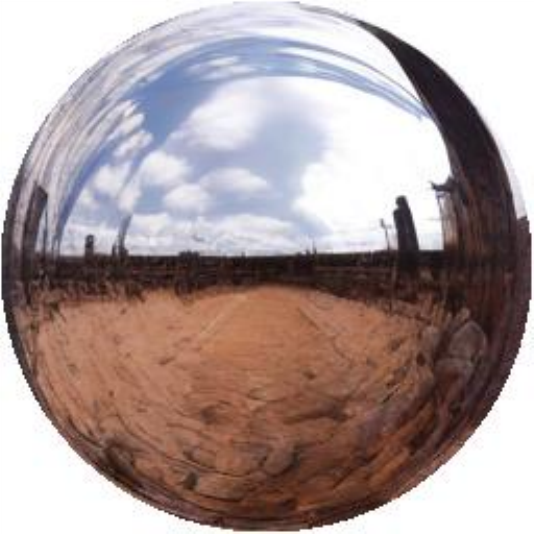}} & 
        \noindent\parbox[c]{0.095\textwidth}{\includegraphics[width=0.095\textwidth]{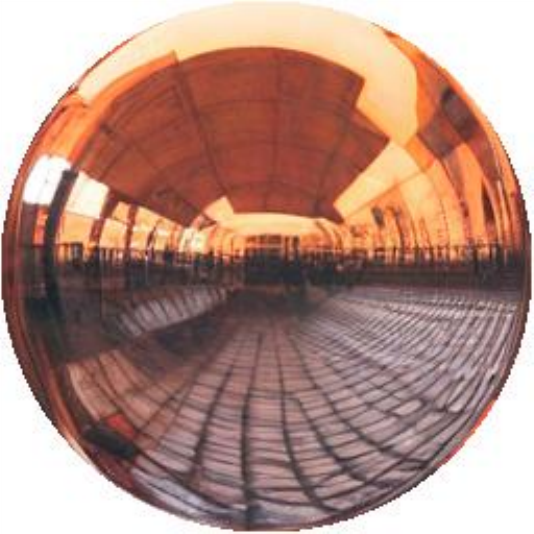}} & 
        \noindent\parbox[c]{0.095\textwidth}{\includegraphics[width=0.095\textwidth]{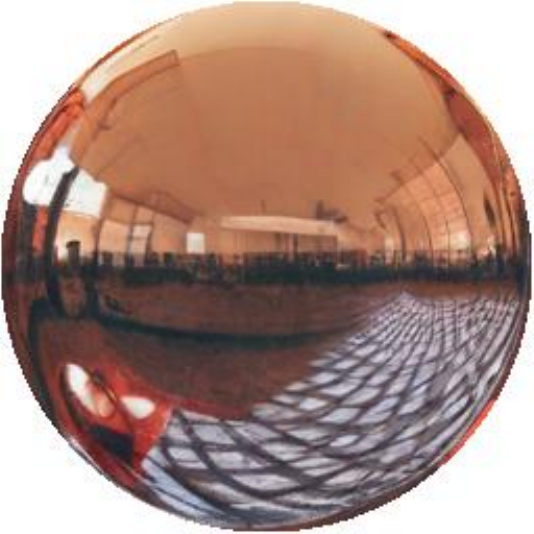}} & 
        \noindent\parbox[c]{0.095\textwidth}{\includegraphics[width=0.095\textwidth]{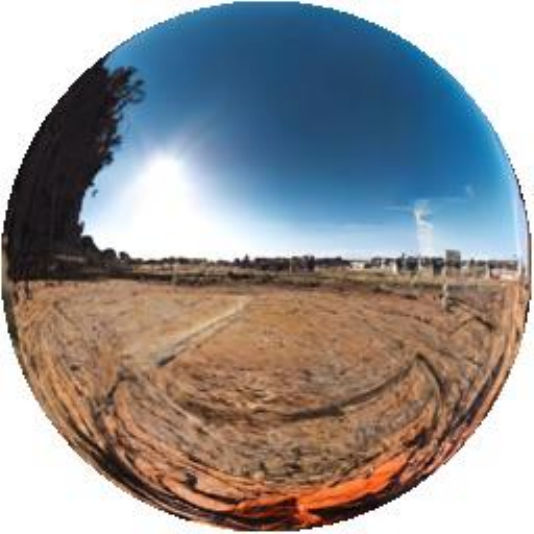}} & 
        \noindent\parbox[c]{0.095\textwidth}{\includegraphics[width=0.095\textwidth]{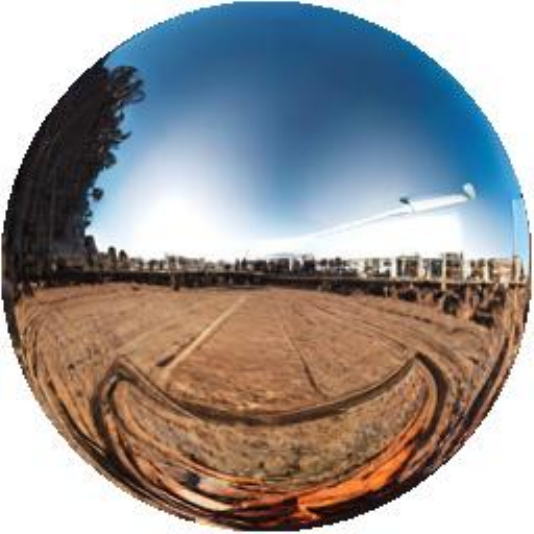}} & 
        \noindent\parbox[c]{0.095\textwidth}{\includegraphics[width=0.095\textwidth]{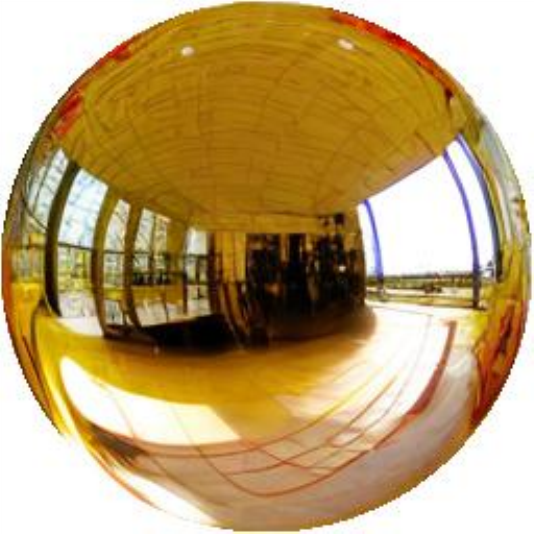}} & 
        \noindent\parbox[c]{0.095\textwidth}{\includegraphics[width=0.095\textwidth]{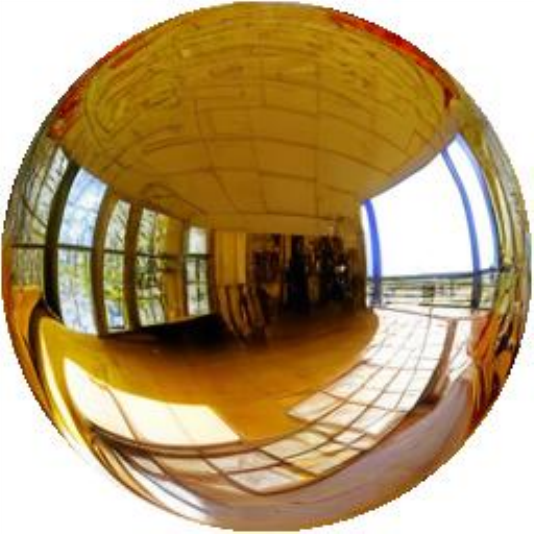}}
        
        \\

        \multicolumn{1}{l}{\rotatebox[origin=c]{90}{\shortstack[l]{\scriptsize EV-2.5}}} &
        \noindent\parbox[c]{0.095\textwidth}{\includegraphics[width=0.095\textwidth]{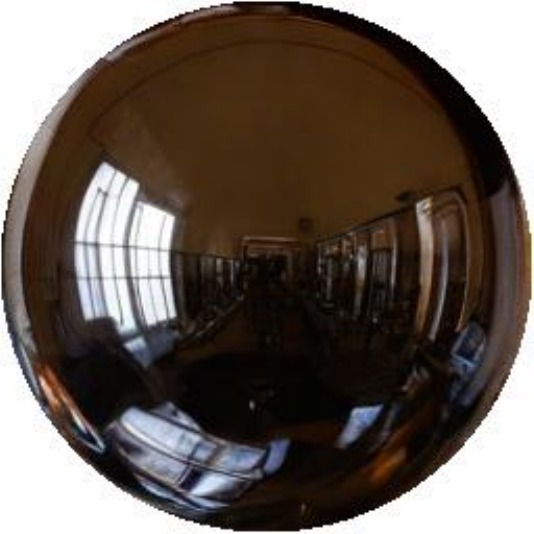}} & 
        \noindent\parbox[c]{0.095\textwidth}{\includegraphics[width=0.095\textwidth]{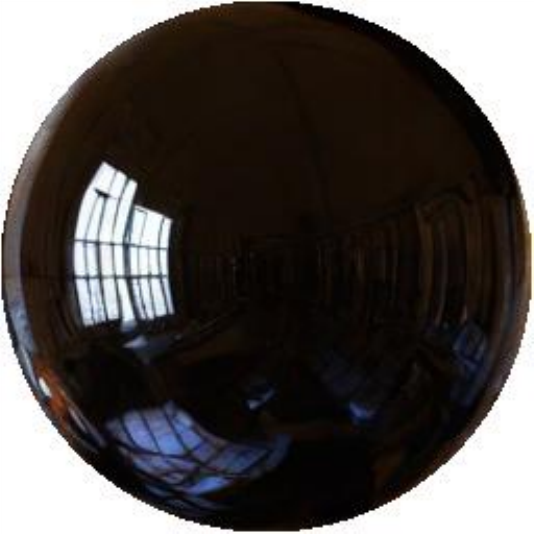}} & 
        \noindent\parbox[c]{0.095\textwidth}{\includegraphics[width=0.095\textwidth]{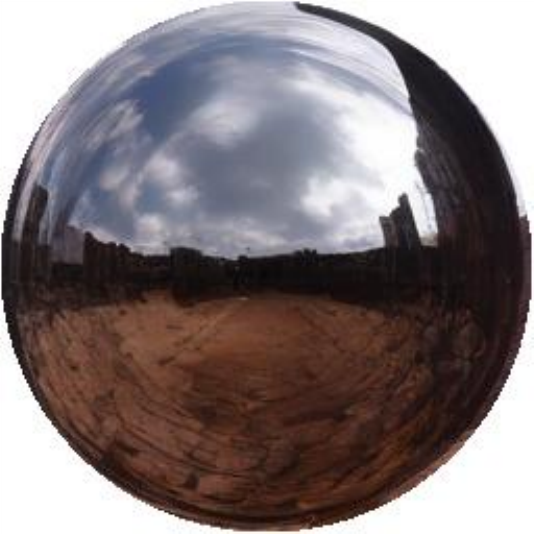}} & 
        \noindent\parbox[c]{0.095\textwidth}{\includegraphics[width=0.095\textwidth]{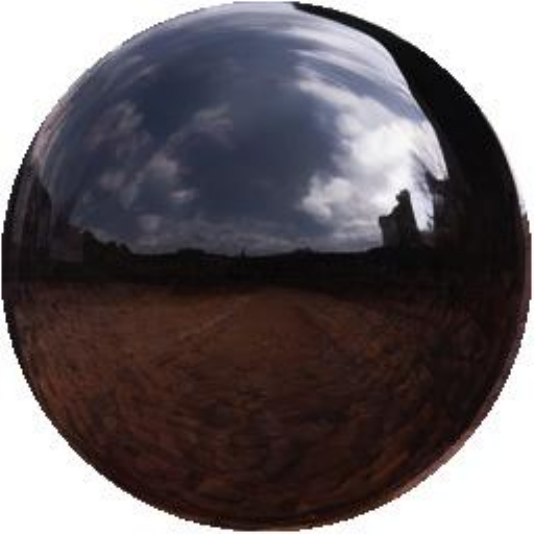}} & 
        \noindent\parbox[c]{0.095\textwidth}{\includegraphics[width=0.095\textwidth]{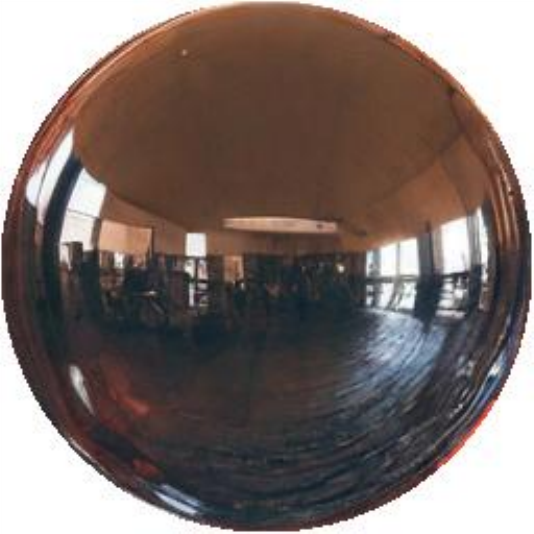}} & 
        \noindent\parbox[c]{0.095\textwidth}{\includegraphics[width=0.095\textwidth]{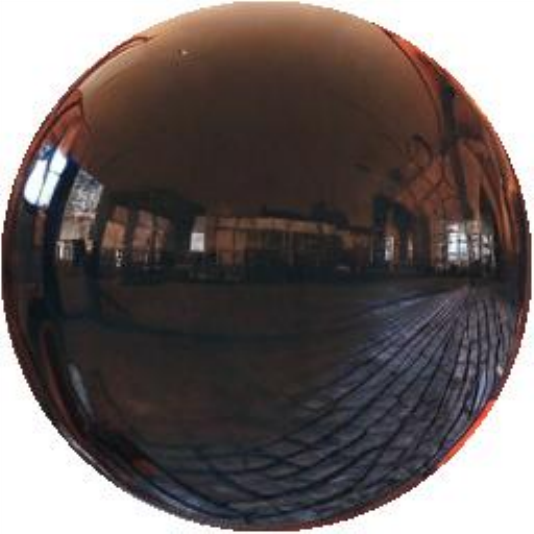}} & 
        \noindent\parbox[c]{0.095\textwidth}{\includegraphics[width=0.095\textwidth]{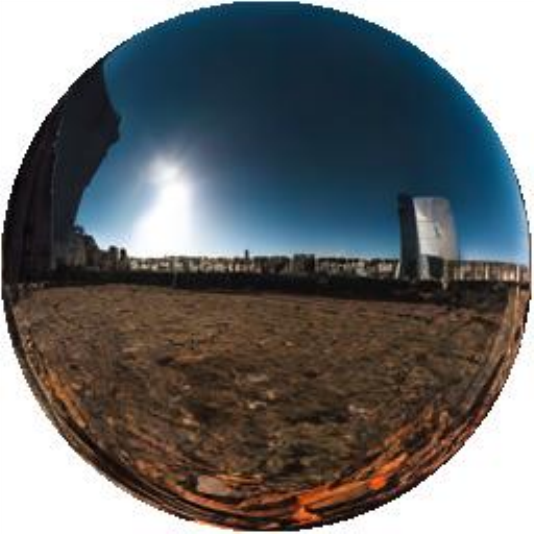}} & 
        \noindent\parbox[c]{0.095\textwidth}{\includegraphics[width=0.095\textwidth]{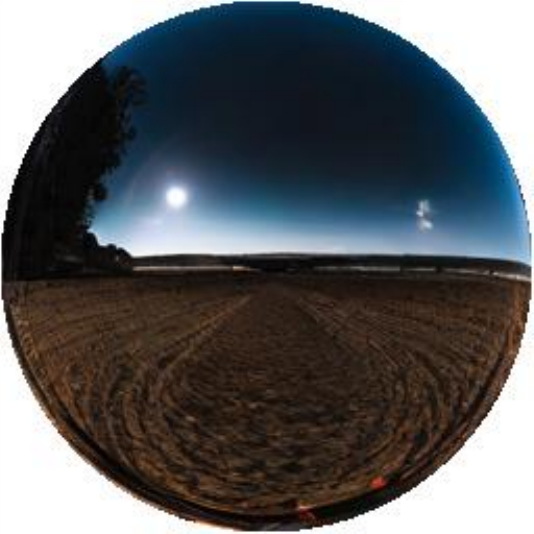}} & 
        \noindent\parbox[c]{0.095\textwidth}{\includegraphics[width=0.095\textwidth]{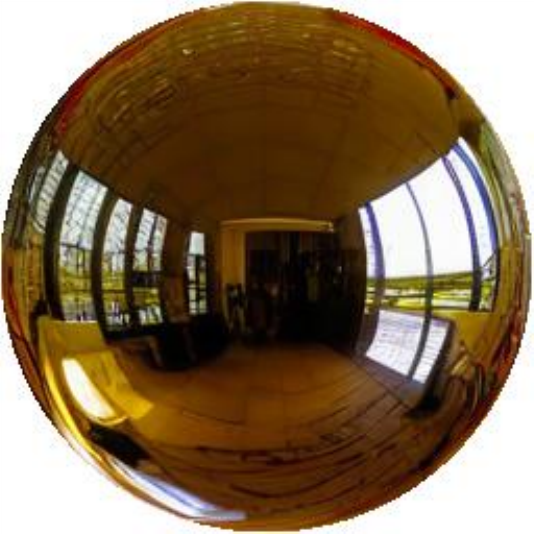}} & 
        \noindent\parbox[c]{0.095\textwidth}{\includegraphics[width=0.095\textwidth]{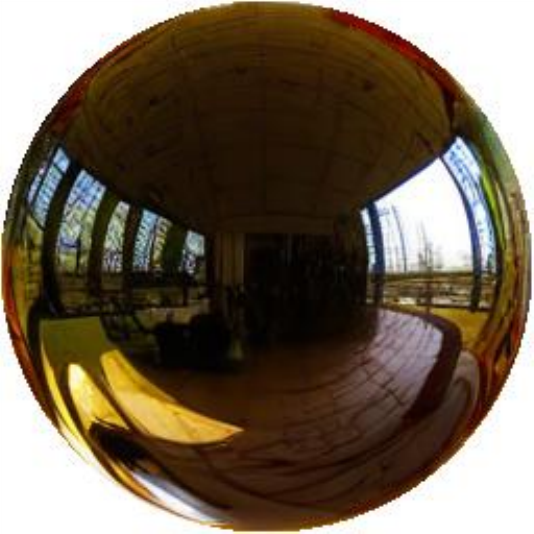}}
        
        \\

        \multicolumn{1}{l}{\rotatebox[origin=c]{90}{\shortstack[l]{\scriptsize EV-5.0}}} &
        \noindent\parbox[c]{0.095\textwidth}{\includegraphics[width=0.095\textwidth]{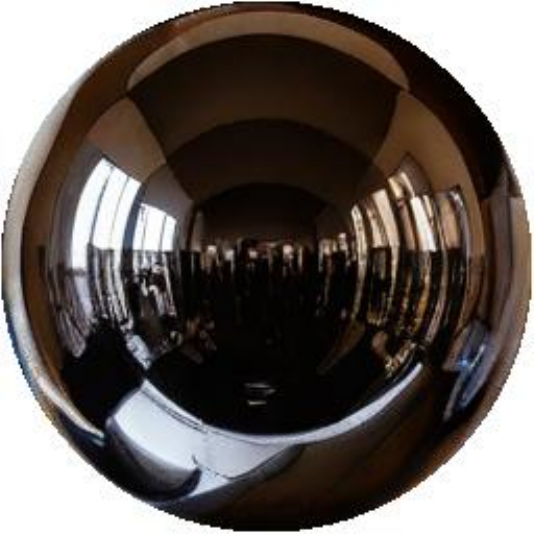}} & 
        \noindent\parbox[c]{0.095\textwidth}{\includegraphics[width=0.095\textwidth]{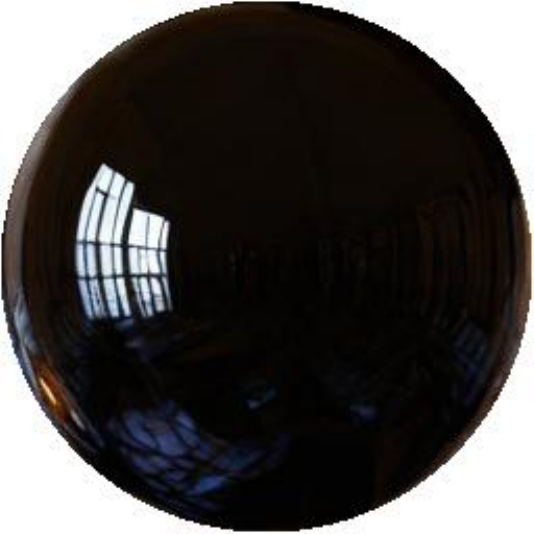}} & 
        \noindent\parbox[c]{0.095\textwidth}{\includegraphics[width=0.095\textwidth]{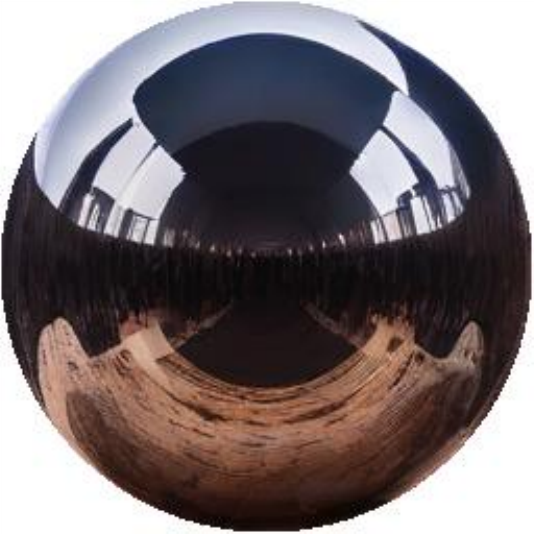}} & 
        \noindent\parbox[c]{0.095\textwidth}{\includegraphics[width=0.095\textwidth]{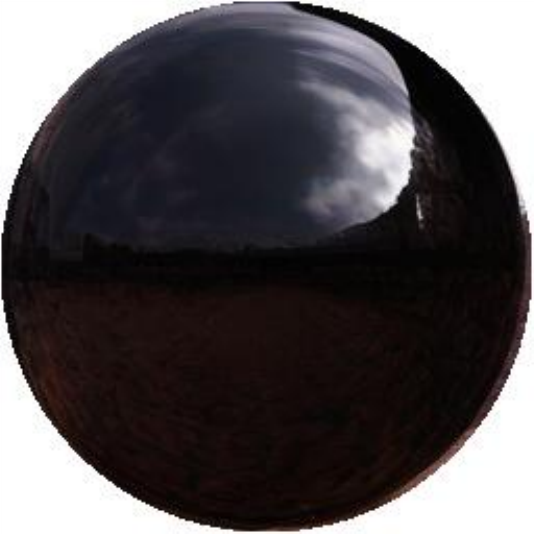}} & 
        \noindent\parbox[c]{0.095\textwidth}{\includegraphics[width=0.095\textwidth]{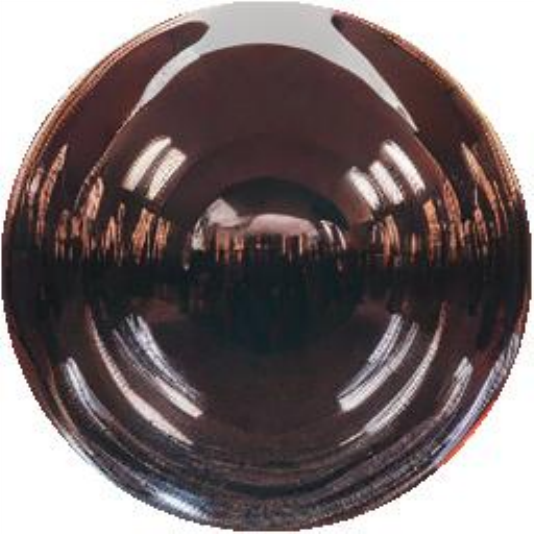}} & 
        \noindent\parbox[c]{0.095\textwidth}{\includegraphics[width=0.095\textwidth]{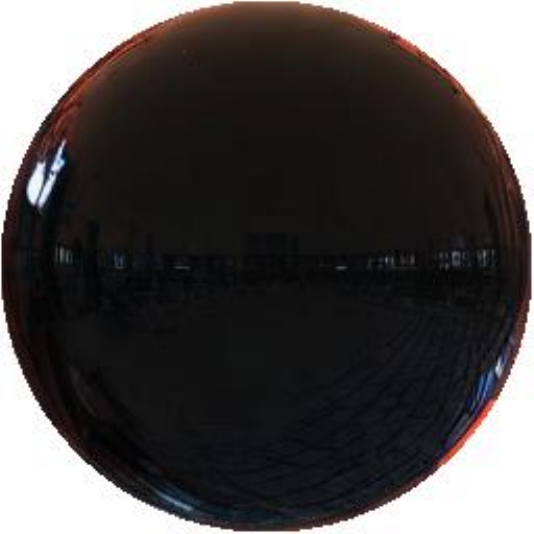}} & 
        \noindent\parbox[c]{0.095\textwidth}{\includegraphics[width=0.095\textwidth]{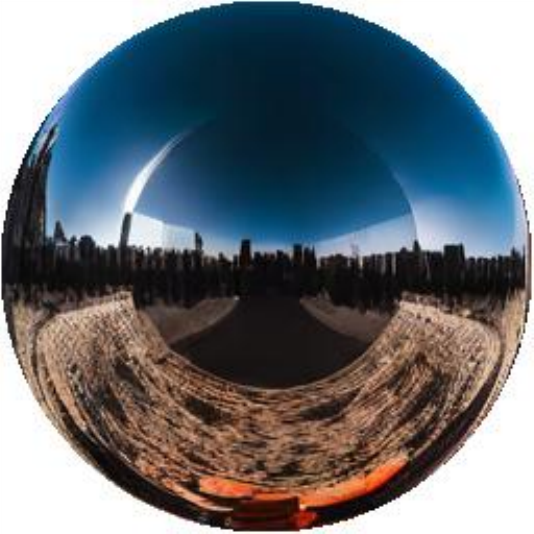}} & 
        \noindent\parbox[c]{0.095\textwidth}{\includegraphics[width=0.095\textwidth]{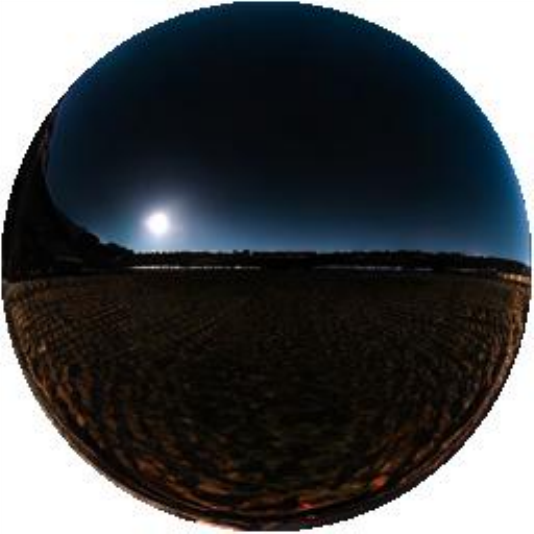}} & 
        \noindent\parbox[c]{0.095\textwidth}{\includegraphics[width=0.095\textwidth]{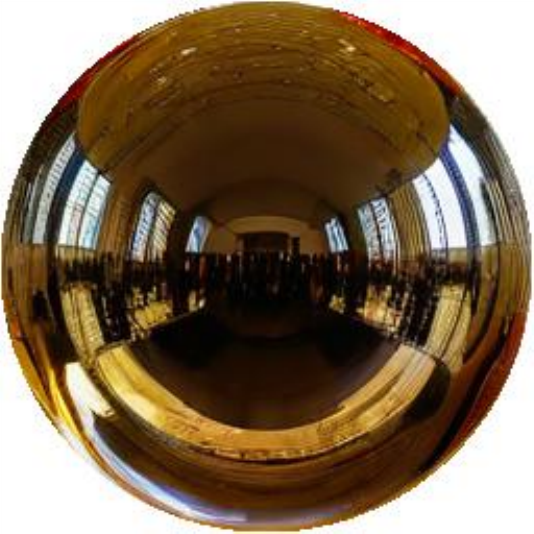}} & 
        \noindent\parbox[c]{0.095\textwidth}{\includegraphics[width=0.095\textwidth]{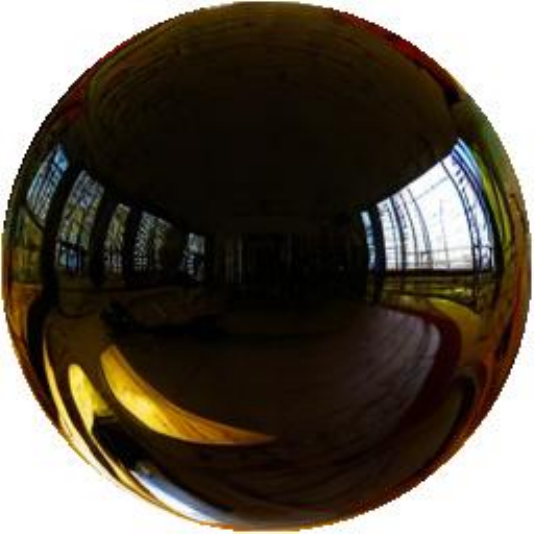}}
        
        \\

        \end{tabu}
    \caption{
    Our proposed continuous LoRA training (Cont. LoRA) yields chrome balls with higher detail consistency across different EVs than results obtained from using three separate LoRAs (3 LoRAs).}
    
    \label{fig:lora_multiple_consistency}
\end{figure*}

To show this, we conducted an experiment comparing results from our LoRA and three separately trained LoRAs at EVs 0, -2.5, and -5.0. Following the commonly adopted training pipeline \cite{ruiz2022dreambooth} implemented in the Diffusers library \cite{diffusers}, our three LoRAs are trained with the prompt containing the `\textit{sks}' token: ``a perfect sks mirrored reflective chrome ball sphere.'' We use the same hyperparameters, random seeds, and HDR panoramas during training. We present results without our iterative algorithm to isolate the effect of LoRA in Figure \ref{fig:lora_multiple_consistency}. Note that we use a lora scale of 1.0 to apply the same weight residual matrices obtained from the training. Our LoRA produces chrome balls with better structural consistency, particularly at EV-5.0.

\tabulinesep=0.5pt
\setlength{\tabcolsep}{0.5pt}


\section{Ablation on DiffusionLight-Turbo}

\subsection{Ablation on Turbo LoRA Training} \label{appendix:aba_turbo_lora}

\subsubsection{Additional details on training set generation}
Following DiffusionLight \cite{phongthawee2024diffusionlight}, we generate median chrome balls using iterative inpainting with $k = 2$ and $N = 30$. The input images are sourced from two datasets: (1) the Flickr2K dataset \cite{Timofte2017}, containing 2,650 images, and (2) 2,000 unique scenes from the InteriorVerse dataset \cite{zhu2022learning}, which have an $85^\circ$ field of view and randomized camera positions and angles. This results in a total of 4,650 HDR average chrome balls, which are then tone-mapped to exposure values of 0, -2.5, and -5.0, yielding 13,950 unique input-ball pairs.


\subsubsection{Range of timesteps for LoRA training.} Unlike Exposure LoRA, Turbo Lora aims to learn a mapping function from an input image to a plausible median chrome ball generated by DiffusionLight. Since these median balls primarily contain low-frequency details, we sample timesteps $t$ from $U(500, 999)$ to focus training on the high-noise region, where such details are typically formed. In Table \ref{tab:ablation_turbo_lora_t}, we compare this setting with two alternatives: $t \sim U(0, 999)$ and $t \sim U(800, 999)$, using the same number of training steps. Evaluation is conducted on the same validation set described in Appendix \ref{appendix:aba_lora}. Our chosen range of 500–999 consistently achieves the best performance across all three metrics.

\begin{table}[]
\centering
\small
\caption{
Ablation study on sampled timesteps for Turbo LoRA training. Models are evaluated at 90,000 training steps.
}
\label{tab:ablation_turbo_lora_t}

\resizebox{0.5\textwidth}{!}{%
\setlength{\tabcolsep}{9pt}
\begin{tabular}{
    l@{\hspace{5pt}}
    l@{\hspace{5pt}}
    c@{\hspace{5pt}}
    c@{\hspace{5pt}}
    c
}
\toprule
\textbf{Sphere} & \begin{tabular}[c]{@{}c@{}}\textbf{Denoising}\\ \textbf{step}\end{tabular} & \textbf{si-RMSE} $\downarrow$ & \begin{tabular}[c]{@{}c@{}}\textbf{Angular}\\ \textbf{Error}\end{tabular} $\downarrow$ & \begin{tabular}[c]{@{}c@{}}\textbf{Normalized}\\ \textbf{RMSE}\end{tabular} $\downarrow$ \\
\midrule

Diffuse & $\vect{x}_0:\vect{x}_{999}$ & 0.153 & 3.007 & 0.236 \\
        & $\vect{x}_{500}:\vect{x}_{999}$ & \scfirst{0.143} & \scfirst{2.593} & \scfirst{0.224} \\
        & $\vect{x}_{800}:\vect{x}_{999}$ & 0.149 & 2.796 & 0.233 \\[0.3ex]
\hline\\[-2.2ex]
Matte & $\vect{x}_0:\vect{x}_{999}$ & 0.372 & 3.901 & 0.352 \\
        & $\vect{x}_{500}:\vect{x}_{999}$ & \scfirst{0.354} & \scfirst{3.509} & \scfirst{0.337} \\
        & $\vect{x}_{800}:\vect{x}_{999}$ & 0.361 & 3.718 & 0.345 \\[0.3ex]
  \hline\\[-2.2ex]
Mirror & $\vect{x}_0:\vect{x}_{999}$ & 0.635 & 5.891 & 0.422 \\
        & $\vect{x}_{500}:\vect{x}_{999}$ & \scfirst{0.620} & \scfirst{5.539} & \scfirst{0.411} \\
        & $\vect{x}_{800}:\vect{x}_{999}$ & 0.628 & 5.775 & 0.418 \\
\bottomrule
\end{tabular}
}
\end{table}

\section{More Comparison with SOTA Inpainting Techniques} \label{appendix:compare_sota}

In Figure \ref{fig:inpaint_sota} in Section \ref{sec:related}, we provide a qualitative comparison between our approach and existing SOTA diffusion-based inpainting methods: Blended Diffusion \cite{avrahami2023blendedlatent, avrahami2022blendeddiffusion}, Paint-by-Example \cite{yang2023paint}, IP-Adapter \cite{ye2023ip-adapter}, DALL·E2 \cite{dalle2}, Adobe Firefly \cite{adobefirefly}, and SDXL \cite{podell2023sdxl}. In this section, we describe the experimental settings for these methods. Additionally, we investigate the behavior of each under different random seeds.

\subsection{Experimental setups}
\begin{figure}
    \centering
    \includegraphics[width=0.49\textwidth]{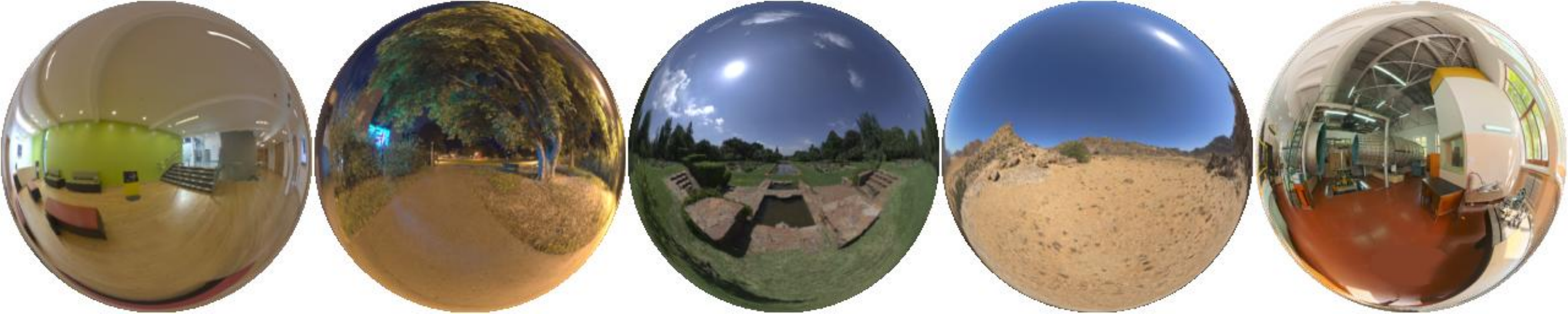}
    \caption{Chrome balls used as inputs for Paint-by-Example \cite{yang2023paint} and IP-Adapter \cite{ye2023ip-adapter}. We generate them from five randomly selected HDR environment maps from Poly Haven dataset \cite{polyhaven}.}
    \label{fig:text2light_reference_ball}
\end{figure}

Blended Diffusion, IP-Adapter, and SDXL shared the same text prompt: ``a perfect mirrored reflective chrome ball sphere.''. We used negative prompt ``matte, diffuse, flat, dull'' when executing methods that can accept one: IP-Adapter and SDXL. We provided Paint-by-Example and IP-Adapter with reference chrome balls from five randomly selected HDR environment maps in Poly Haven dataset \cite{polyhaven} as shown in Figure \ref{fig:text2light_reference_ball}. We used the official OpenAI API \footnote{\url{https://platform.openai.com/docs/guides/images/edits-dall-e-2-only}} for DALL·E2, and we used the Generative Fill feature in Photoshop for Adobe Firefly. We followed the default configurations in the methods' official implementations as described in Table \ref{tab:sota_default_config}.

\setlength{\tabcolsep}{2pt}
\begin{table}
    \centering
    
    \caption{Sampler, number of sampling steps, and classifier-guidance scale (CFG) used in different SOTA methods.}
    \label{tab:sota_default_config}
\resizebox{0.5\textwidth}{!}{%
\setlength{\tabcolsep}{11pt}
    \begin{tabu}{lccc}
    \toprule
      \multicolumn{1}{c}{\textbf{Method}}   &  \textbf{Sampler} & \textbf{\#Steps} & \textbf{CFG} \\
      \midrule
      Blended Diffusion   &  DDIM \cite{ho2020denoising} & 50 & 7.5 \\
      Paint-by-Example   &  PLMS \cite{liu2022pseudo} & 50 & 5.0 \\
      IP-Adapter   &  UniPC \cite{zhao2023unipc} & 30 & 5.0 \\
      SDXL & UniPC \cite{zhao2023unipc} & 30 & 5.0 \\
      \bottomrule
    \end{tabu}
    }
\end{table}

\subsection{Behavior under different random seeds}
\begin{figure*}
    \centering

    \begin{tabu} to \textwidth {
        @{}
        l@{}
        l@{\hspace{0.5pt}}
    }
        \multicolumn{1}{l}{\rotatebox[origin=c]{90}{\shortstack[l]{\scriptsize Input \\ \scriptsize image}}} &
        \noindent\parbox[c]{0.5\textwidth}{\includegraphics[width=0.5\textwidth]{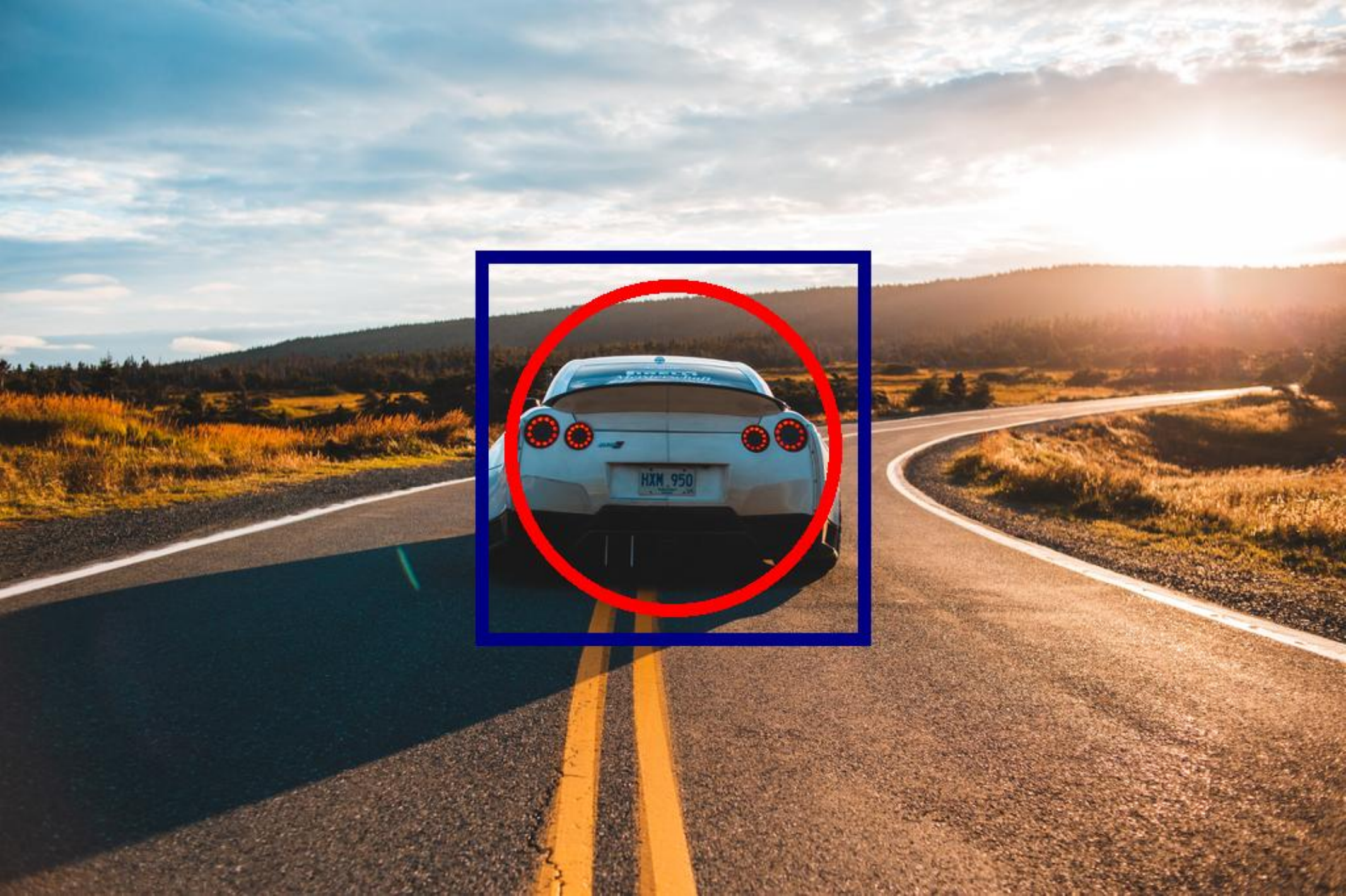}}  \\
    \end{tabu}
    \begin{tabu} to \textwidth {
        @{}
        c@{}
        c@{\hspace{0.5pt}}
        c@{\hspace{0.5pt}}
        c@{\hspace{0.5pt}}
        c@{\hspace{0.5pt}}
        c@{\hspace{0.5pt}}
        c@{\hspace{0.5pt}}
        c@{\hspace{0.5pt}}
        c@{\hspace{0.5pt}}
        c@{\hspace{0.5pt}}
        c@{\hspace{0.5pt}}
        c@{}
    }
        

        \multicolumn{1}{l}{\rotatebox[origin=c]{90}{\shortstack[l]{\scriptsize Blended Dif-\\ \scriptsize fusion \cite{avrahami2023blendedlatent, avrahami2022blendeddiffusion}}}} &
        \noindent\parbox[c]{0.081\textwidth}{\includegraphics[width=0.081\textwidth]{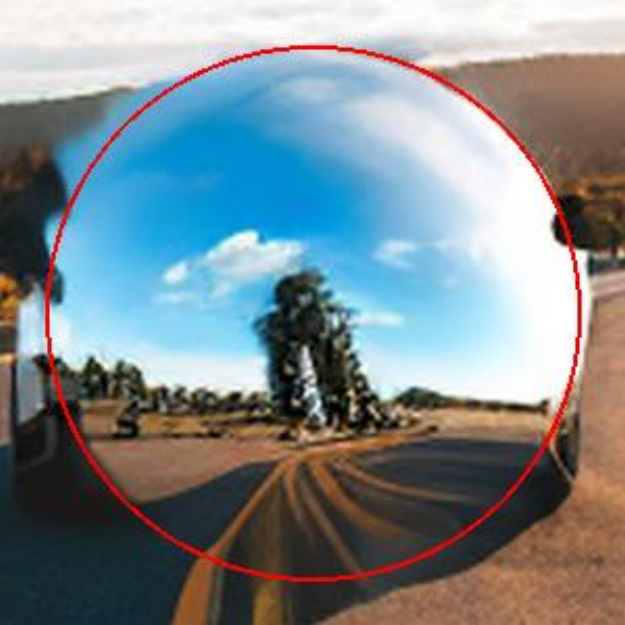}} & 
        \noindent\parbox[c]{0.081\textwidth}{\includegraphics[width=0.081\textwidth]{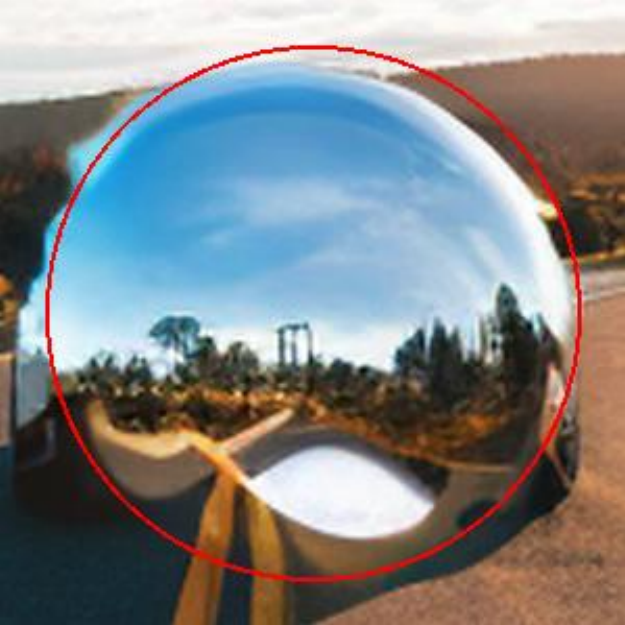}} &  
        \noindent\parbox[c]{0.081\textwidth}{\includegraphics[width=0.081\textwidth]{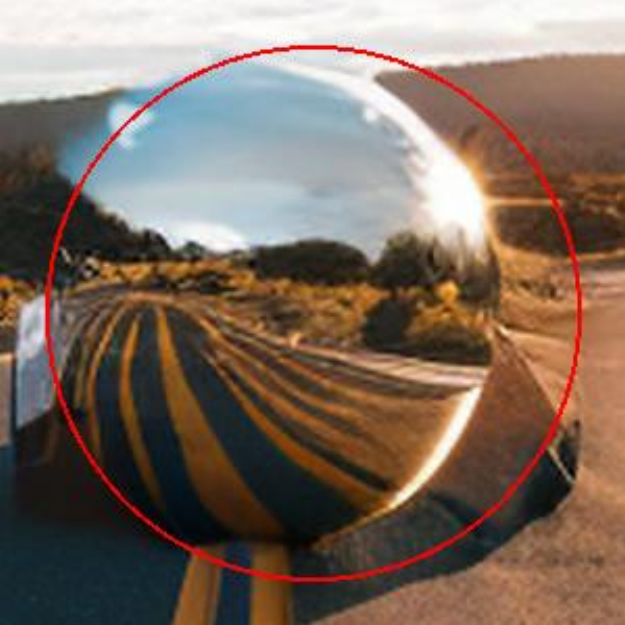}} & 
        \noindent\parbox[c]{0.081\textwidth}{\includegraphics[width=0.081\textwidth]{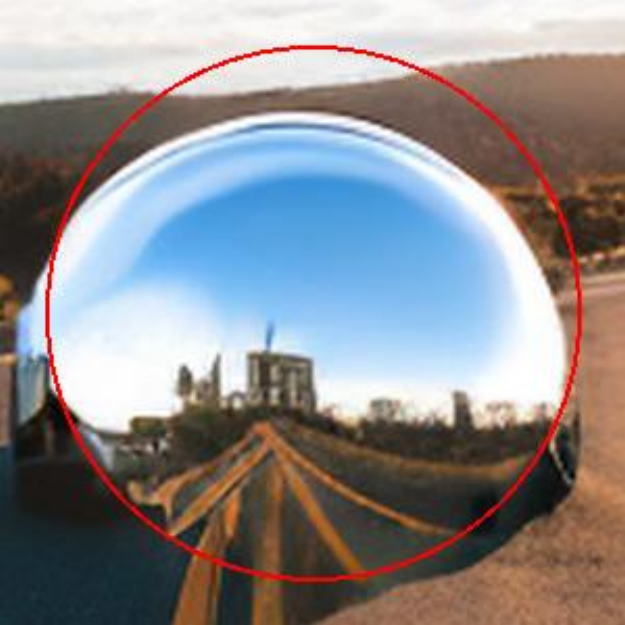}} & 
        \noindent\parbox[c]{0.081\textwidth}{\includegraphics[width=0.081\textwidth]{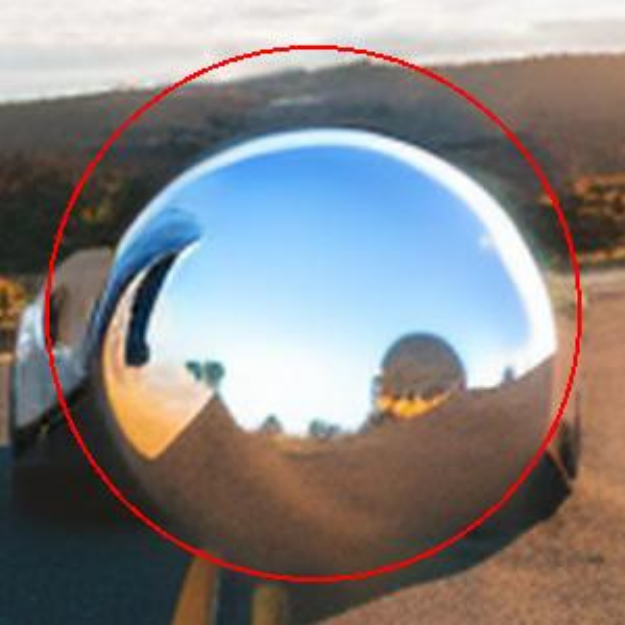}} & 
        \noindent\parbox[c]{0.081\textwidth}{\includegraphics[width=0.081\textwidth]{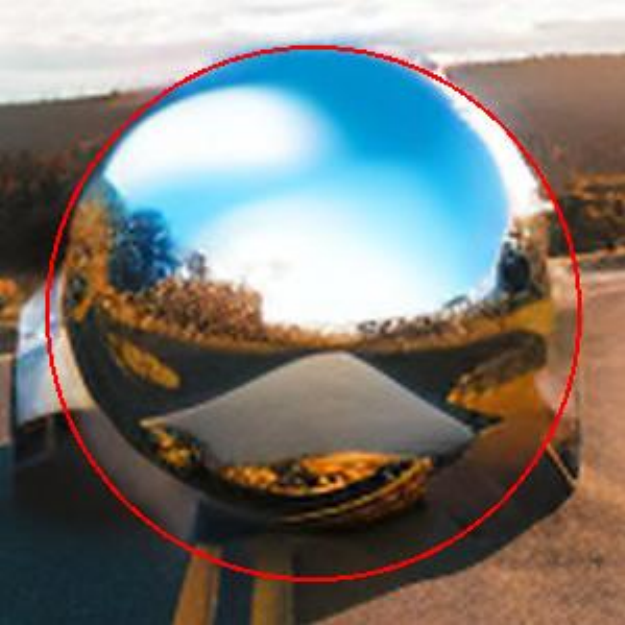}} & 
        \noindent\parbox[c]{0.081\textwidth}{\includegraphics[width=0.081\textwidth]{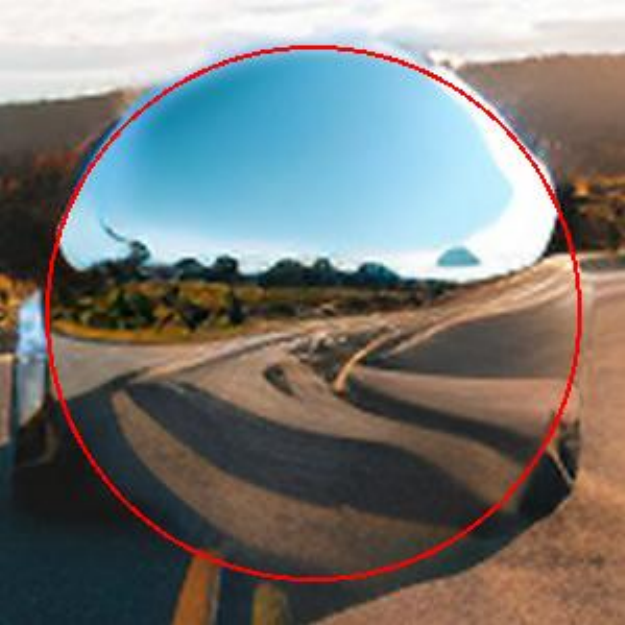}} & 
        \noindent\parbox[c]{0.081\textwidth}{\includegraphics[width=0.081\textwidth]{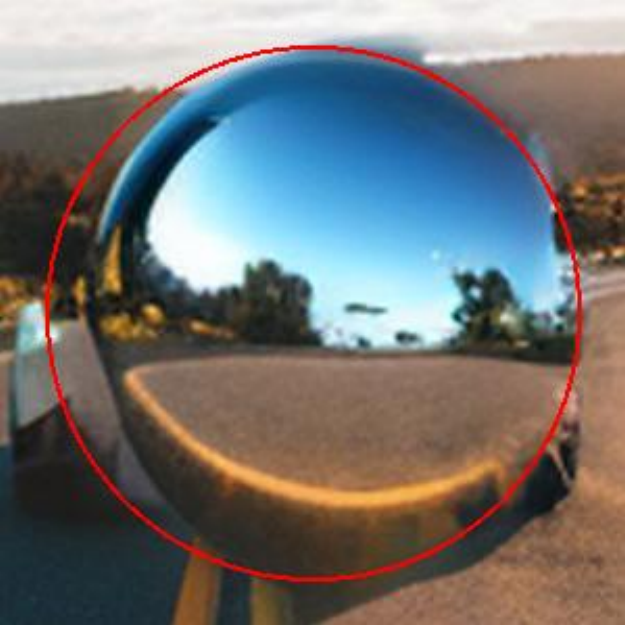}} & 
        \noindent\parbox[c]{0.081\textwidth}{\includegraphics[width=0.081\textwidth]{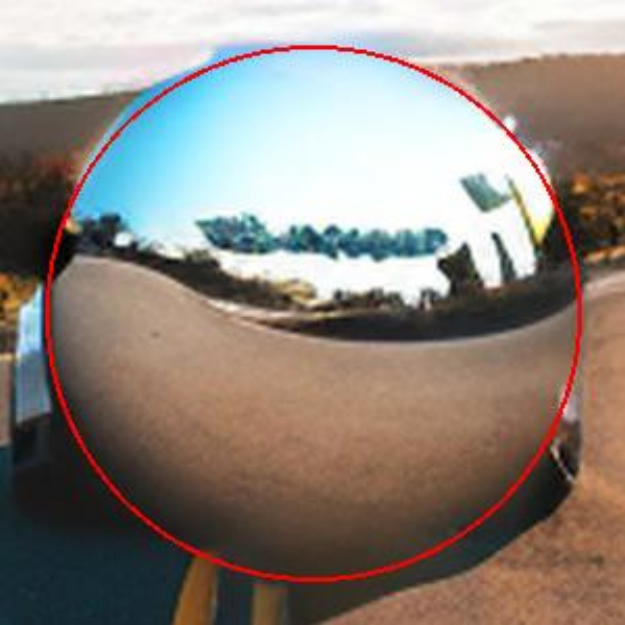}} & 
        \noindent\parbox[c]{0.081\textwidth}{\includegraphics[width=0.081\textwidth]{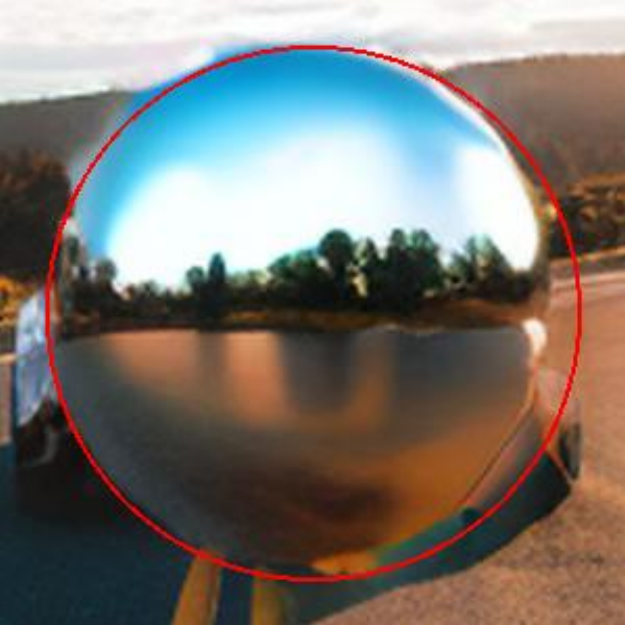}} & 
        
        \\

        \multicolumn{1}{l}{\rotatebox[origin=c]{90}{\shortstack[l]{\scriptsize Paint-by-Ex\\ \scriptsize ample \cite{yang2023paint}}}} &
        \noindent\parbox[c]{0.081\textwidth}{\includegraphics[width=0.081\textwidth]{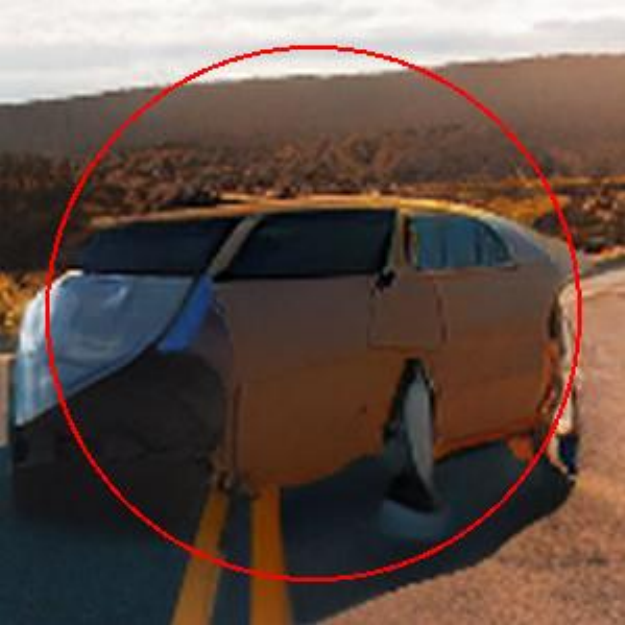}} & 
        \noindent\parbox[c]{0.081\textwidth}{\includegraphics[width=0.081\textwidth]{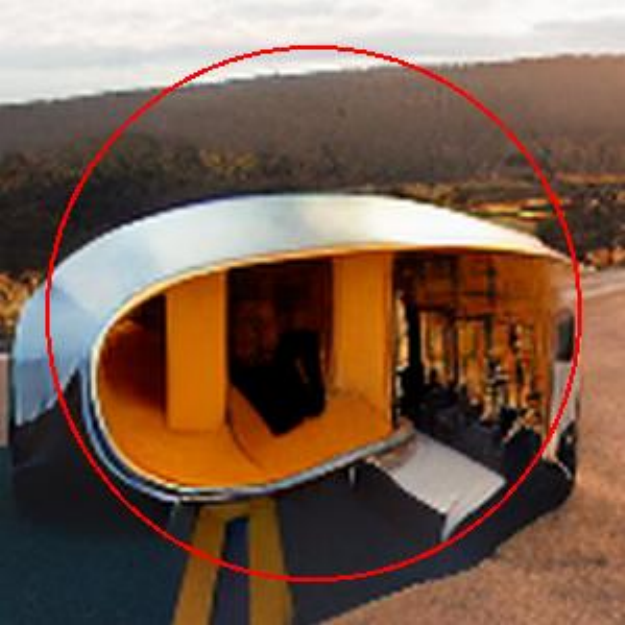}} &  
        \noindent\parbox[c]{0.081\textwidth}{\includegraphics[width=0.081\textwidth]{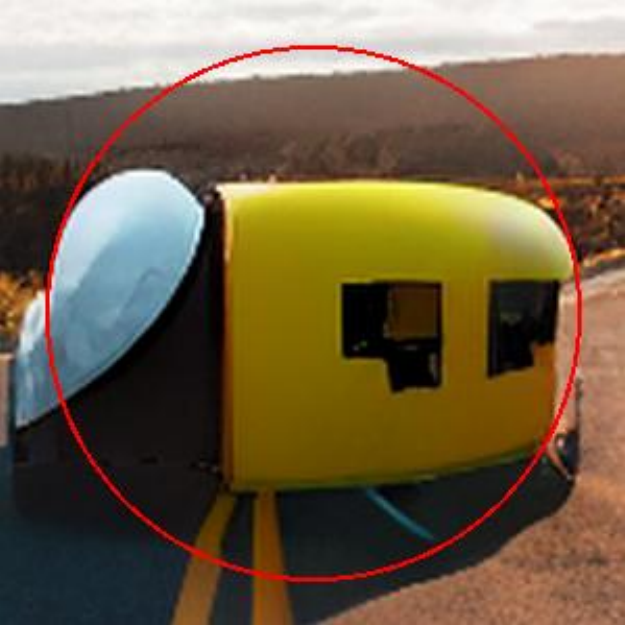}} & 
        \noindent\parbox[c]{0.081\textwidth}{\includegraphics[width=0.081\textwidth]{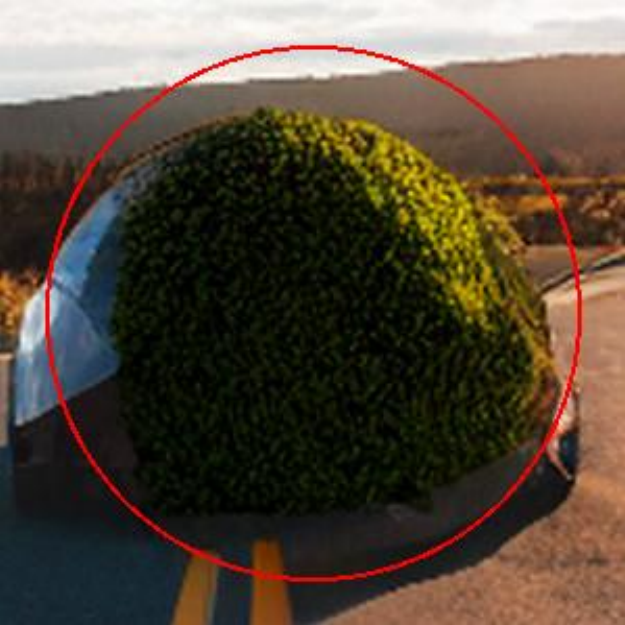}} & 
        \noindent\parbox[c]{0.081\textwidth}{\includegraphics[width=0.081\textwidth]{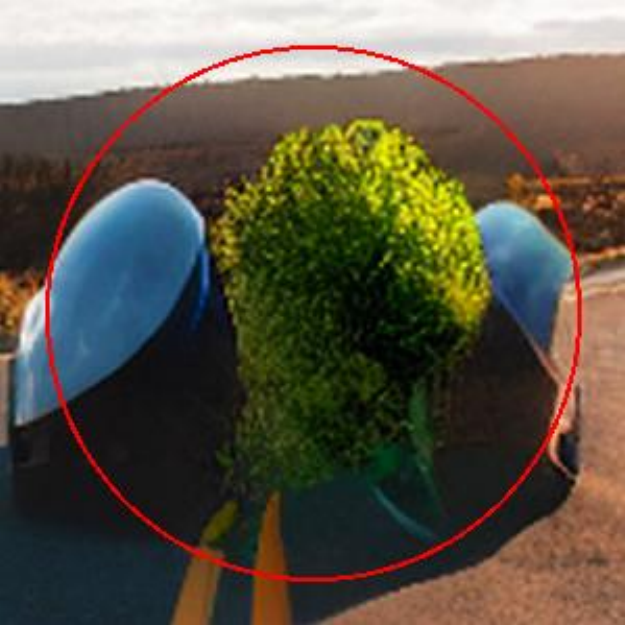}} & 
        \noindent\parbox[c]{0.081\textwidth}{\includegraphics[width=0.081\textwidth]{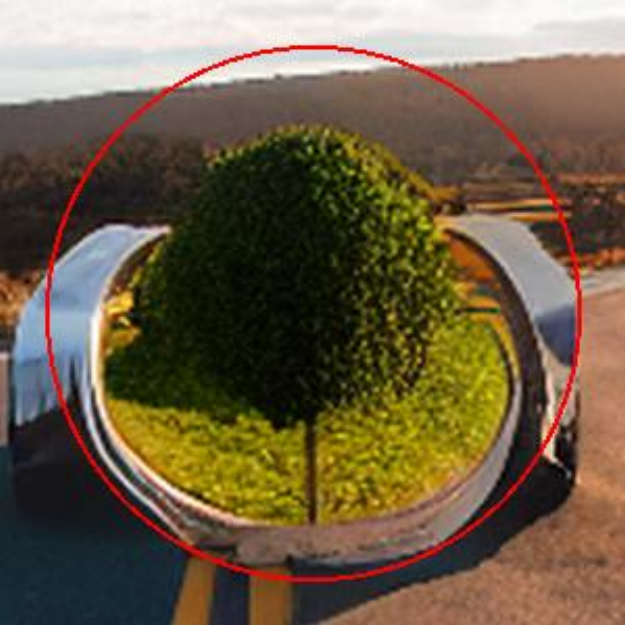}} & 
        \noindent\parbox[c]{0.081\textwidth}{\includegraphics[width=0.081\textwidth]{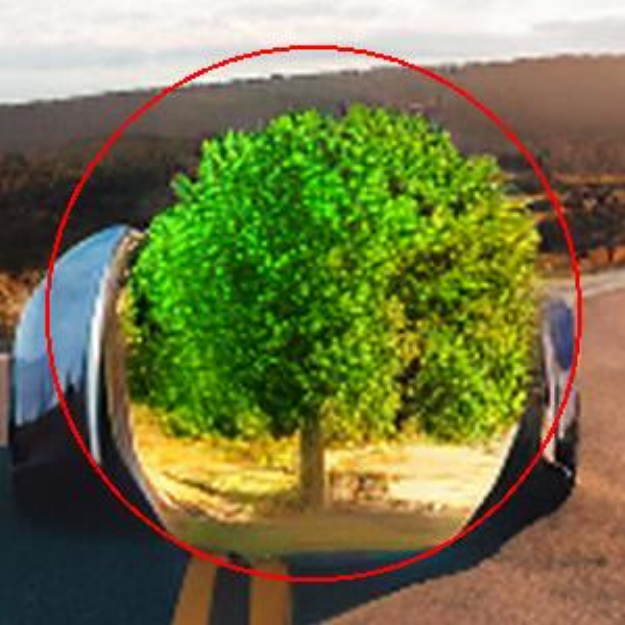}} & 
        \noindent\parbox[c]{0.081\textwidth}{\includegraphics[width=0.081\textwidth]{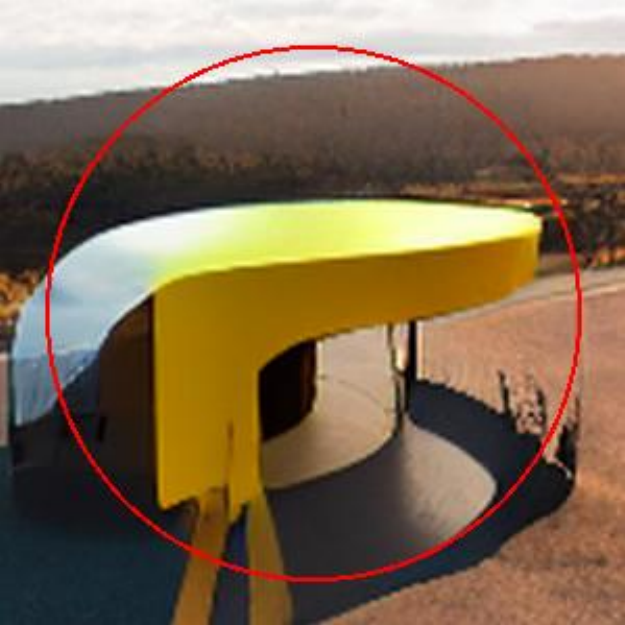}} & 
        \noindent\parbox[c]{0.081\textwidth}{\includegraphics[width=0.081\textwidth]{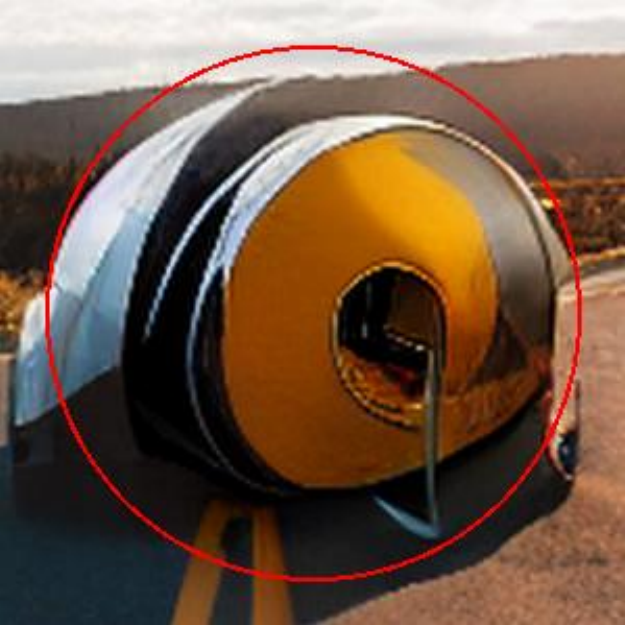}} & 
        \noindent\parbox[c]{0.081\textwidth}{\includegraphics[width=0.081\textwidth]{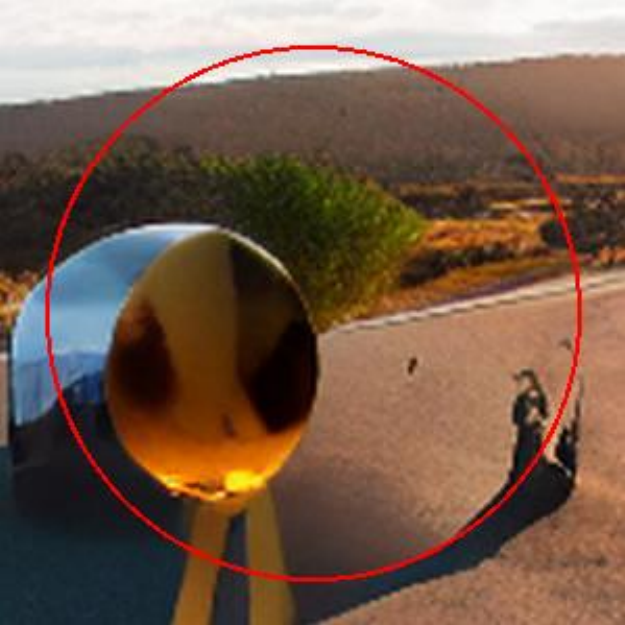}} & 
        
        \\

        \multicolumn{1}{l}{\rotatebox[origin=c]{90}{\shortstack[l]{\scriptsize IP-Adapter\\ \scriptsize \cite{ye2023ip-adapter}}}} &
        \noindent\parbox[c]{0.081\textwidth}{\includegraphics[width=0.081\textwidth]{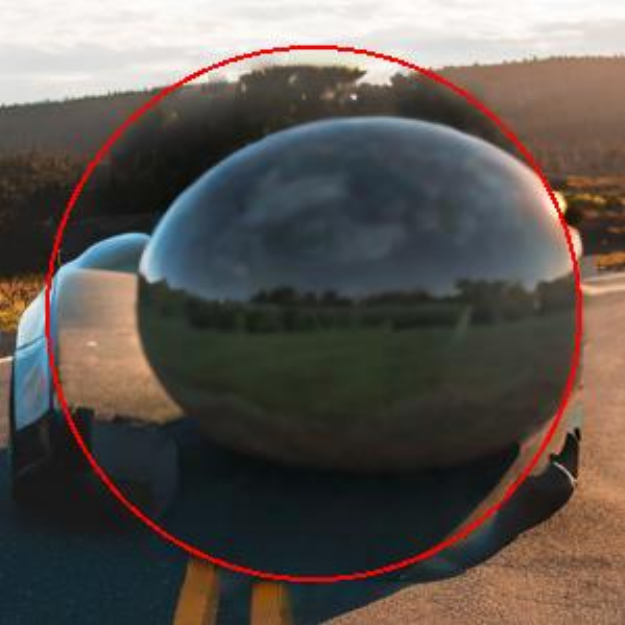}} & 
        \noindent\parbox[c]{0.081\textwidth}{\includegraphics[width=0.081\textwidth]{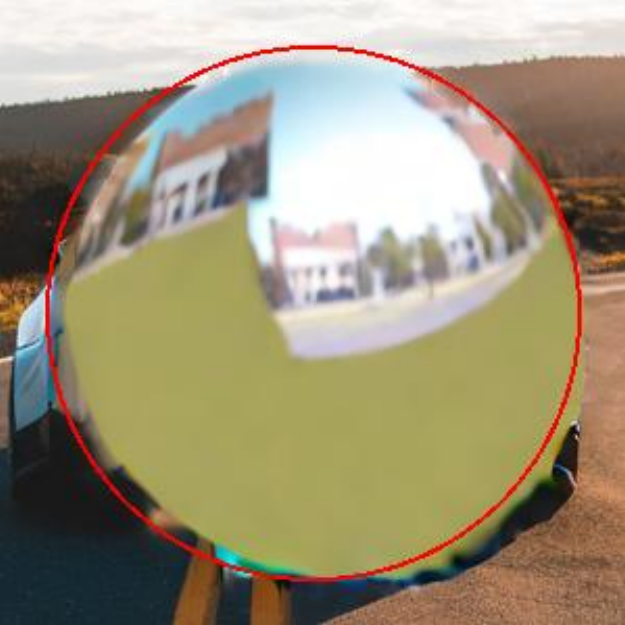}} &  
        \noindent\parbox[c]{0.081\textwidth}{\includegraphics[width=0.081\textwidth]{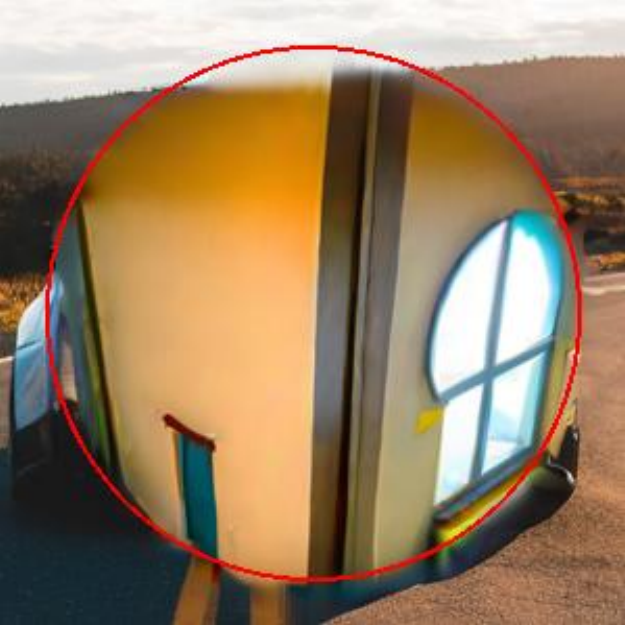}} & 
        \noindent\parbox[c]{0.081\textwidth}{\includegraphics[width=0.081\textwidth]{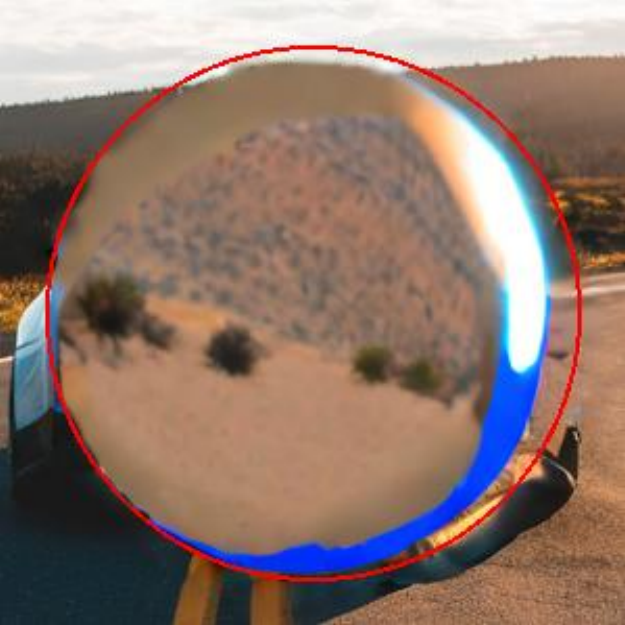}} & 
        \noindent\parbox[c]{0.081\textwidth}{\includegraphics[width=0.081\textwidth]{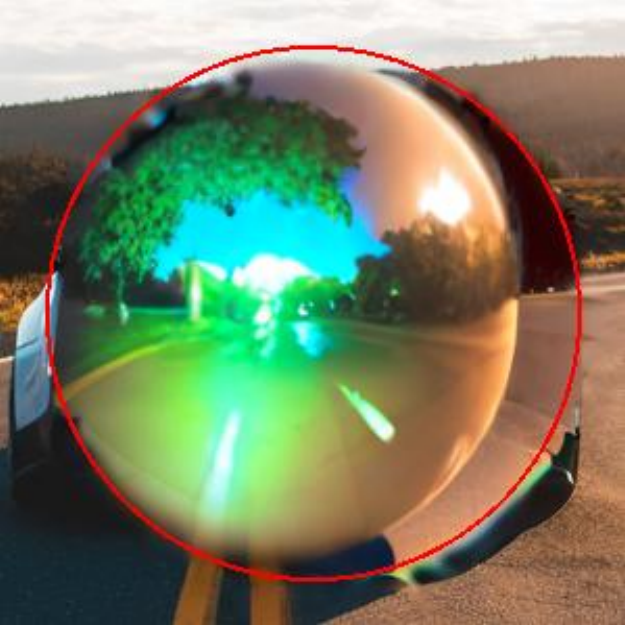}} & 
        \noindent\parbox[c]{0.081\textwidth}{\includegraphics[width=0.081\textwidth]{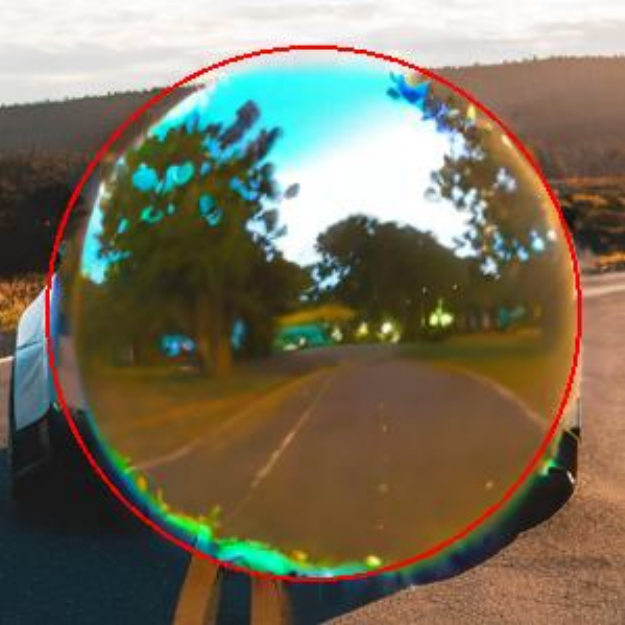}} & 
        \noindent\parbox[c]{0.081\textwidth}{\includegraphics[width=0.081\textwidth]{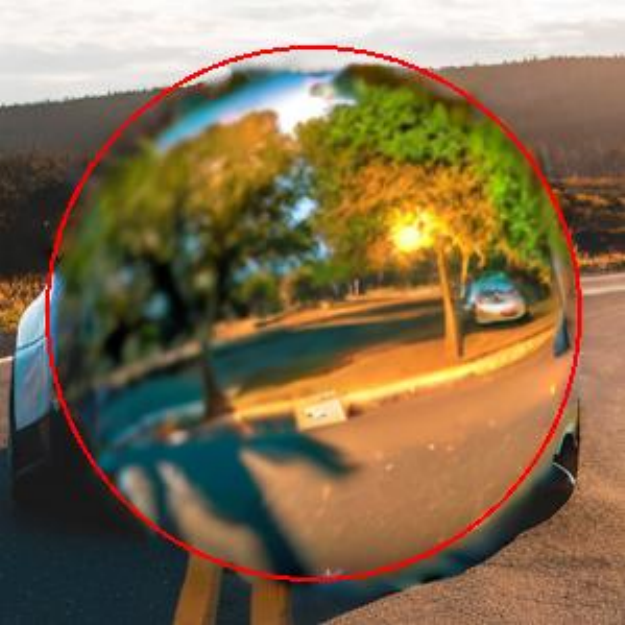}} & 
        \noindent\parbox[c]{0.081\textwidth}{\includegraphics[width=0.081\textwidth]{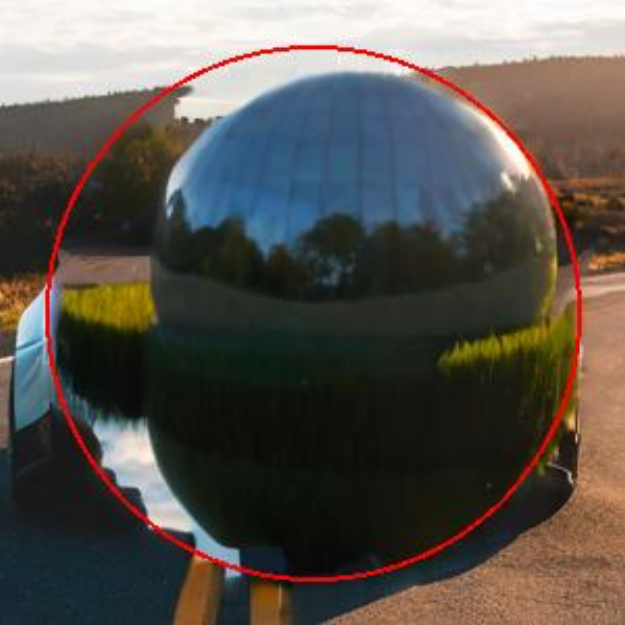}} & 
        \noindent\parbox[c]{0.081\textwidth}{\includegraphics[width=0.081\textwidth]{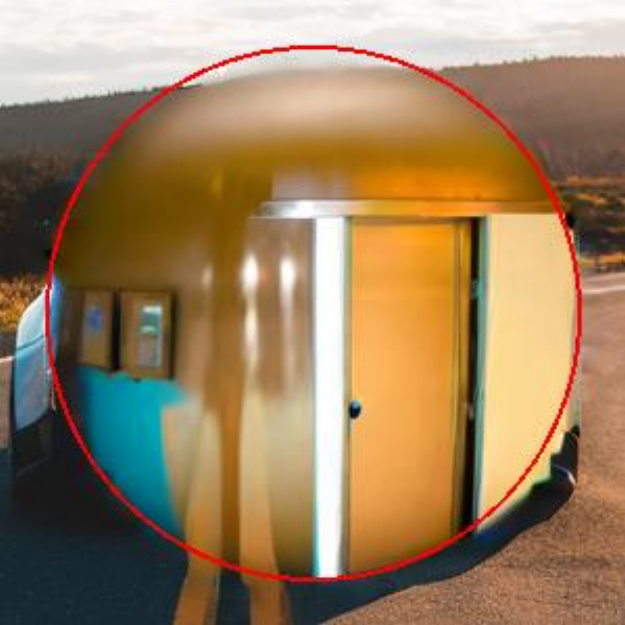}} & 
        \noindent\parbox[c]{0.081\textwidth}{\includegraphics[width=0.081\textwidth]{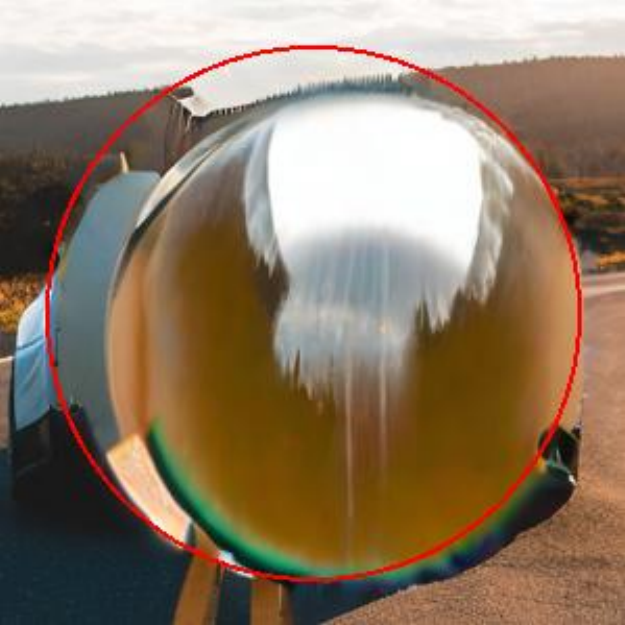}} & 
        
        \\

        \multicolumn{1}{l}{\rotatebox[origin=c]{90}{\shortstack[l]{\scriptsize DALL·E2 \cite{dalle2}}}} &
        \noindent\parbox[c]{0.081\textwidth}{\includegraphics[width=0.081\textwidth]{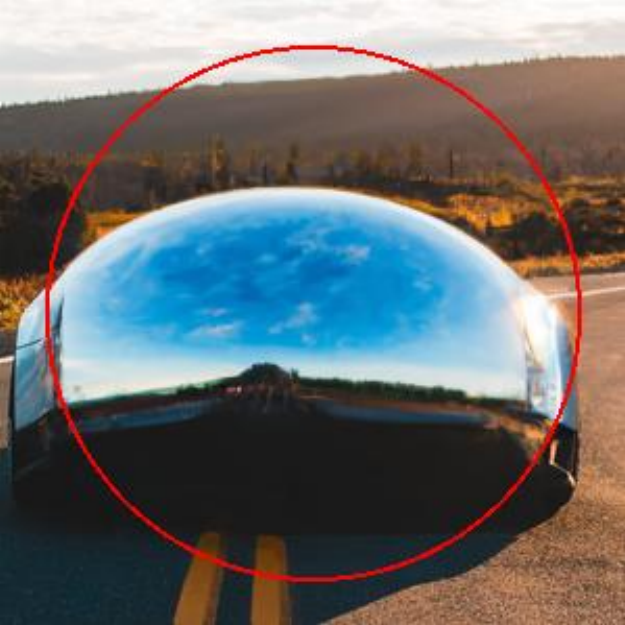}} & 
        \noindent\parbox[c]{0.081\textwidth}{\includegraphics[width=0.081\textwidth]{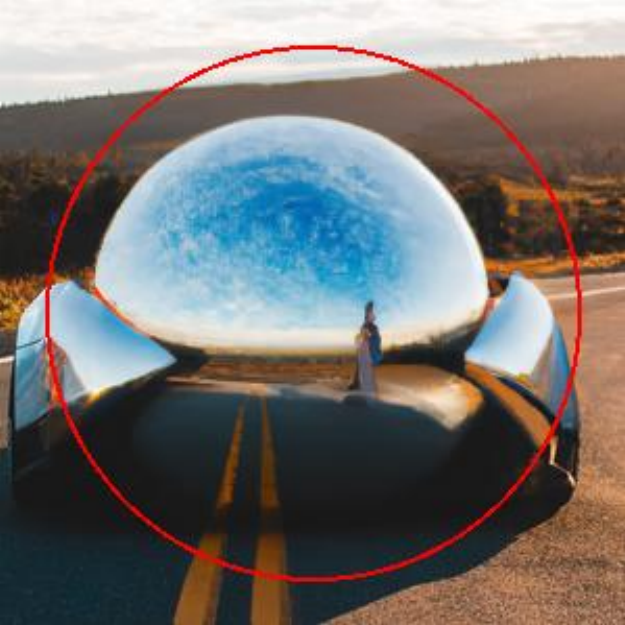}} &  
        \noindent\parbox[c]{0.081\textwidth}{\includegraphics[width=0.081\textwidth]{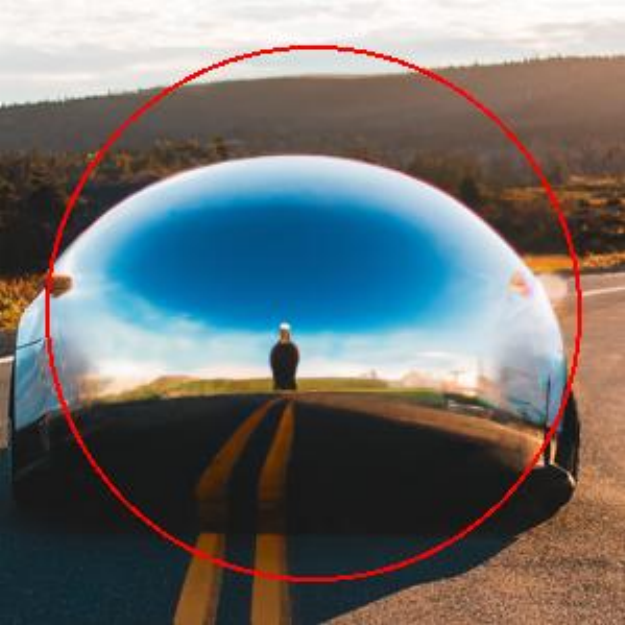}} & 
        \noindent\parbox[c]{0.081\textwidth}{\includegraphics[width=0.081\textwidth]{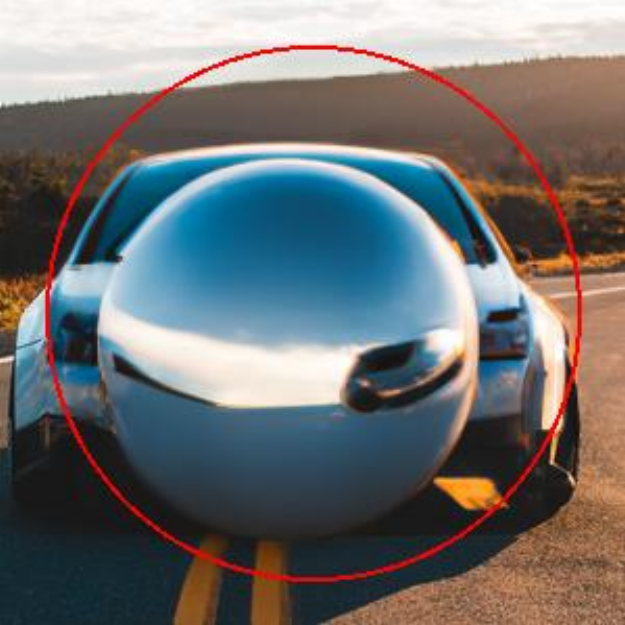}} & 
        \noindent\parbox[c]{0.081\textwidth}{\includegraphics[width=0.081\textwidth]{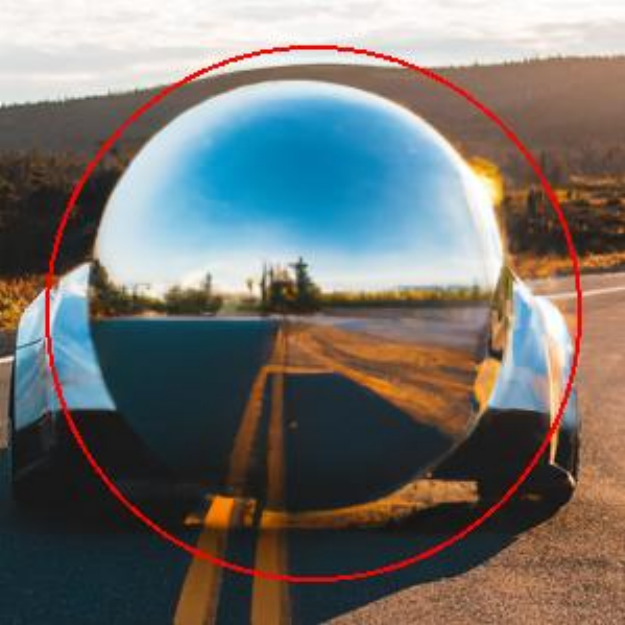}} & 
        \noindent\parbox[c]{0.081\textwidth}{\includegraphics[width=0.081\textwidth]{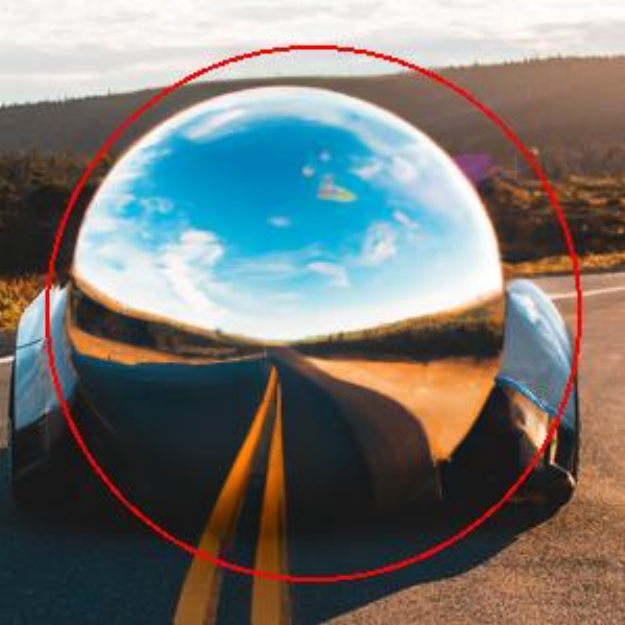}} & 
        \noindent\parbox[c]{0.081\textwidth}{\includegraphics[width=0.081\textwidth]{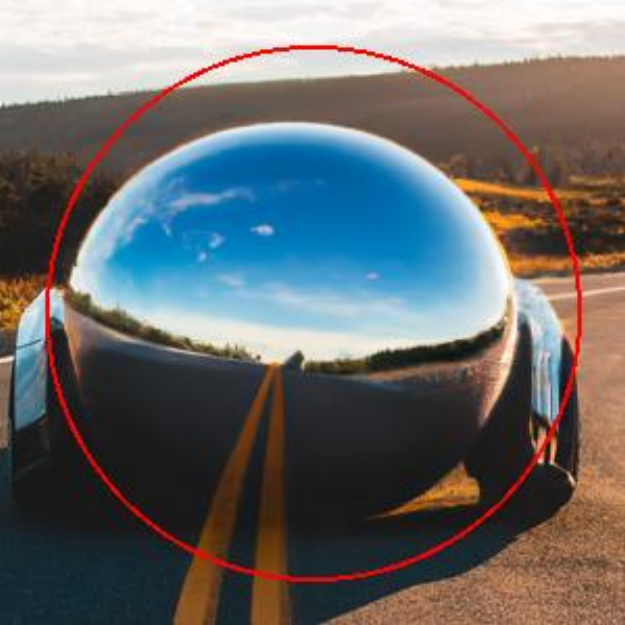}} & 
        \noindent\parbox[c]{0.081\textwidth}{\includegraphics[width=0.081\textwidth]{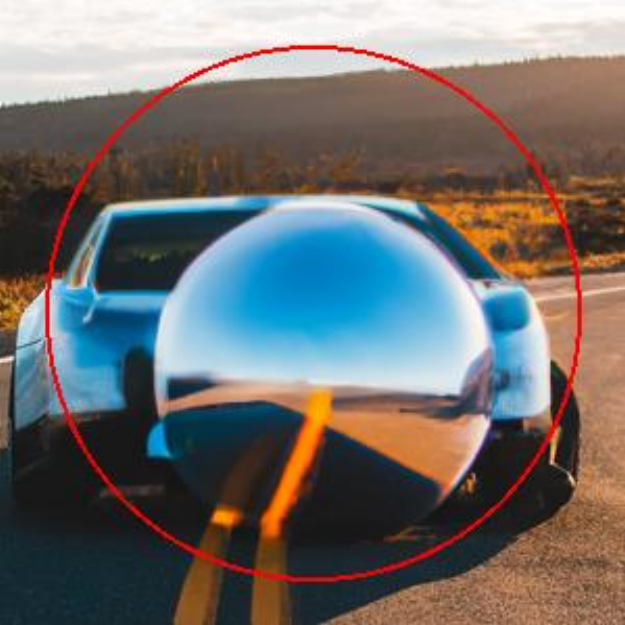}} & 
        \noindent\parbox[c]{0.081\textwidth}{\includegraphics[width=0.081\textwidth]{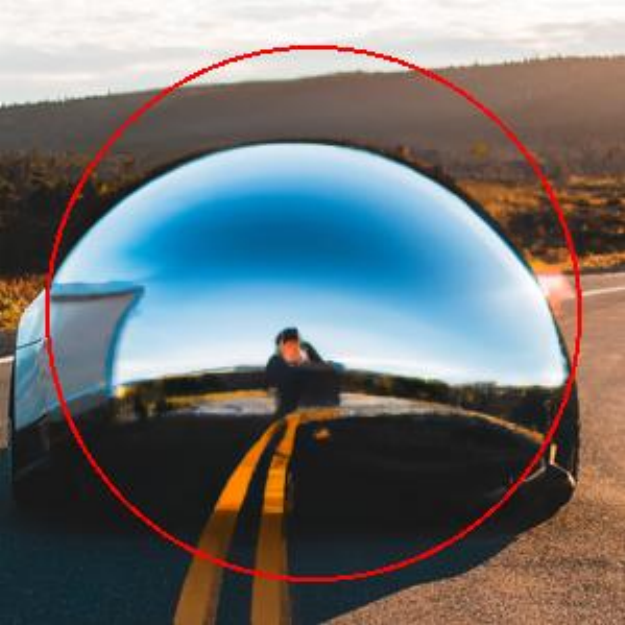}} & 
        \noindent\parbox[c]{0.081\textwidth}{\includegraphics[width=0.081\textwidth]{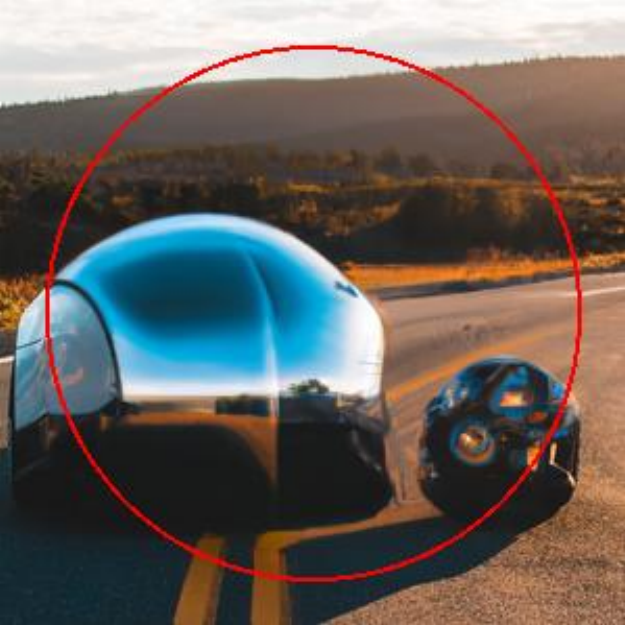}} & 
        
        \\

        \multicolumn{1}{l}{\rotatebox[origin=c]{90}{\shortstack[l]{\scriptsize Adobe \\ \scriptsize Firefly \cite{adobefirefly}}}} &
        \noindent\parbox[c]{0.081\textwidth}{\includegraphics[width=0.081\textwidth]{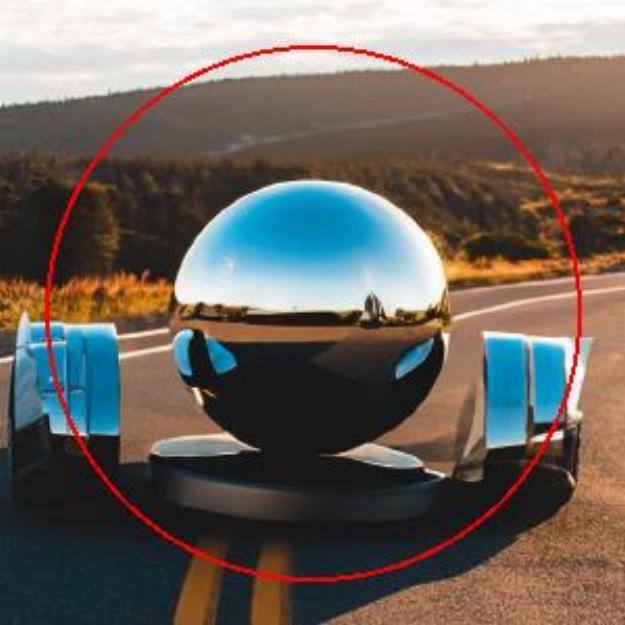}} & 
        \noindent\parbox[c]{0.081\textwidth}{\includegraphics[width=0.081\textwidth]{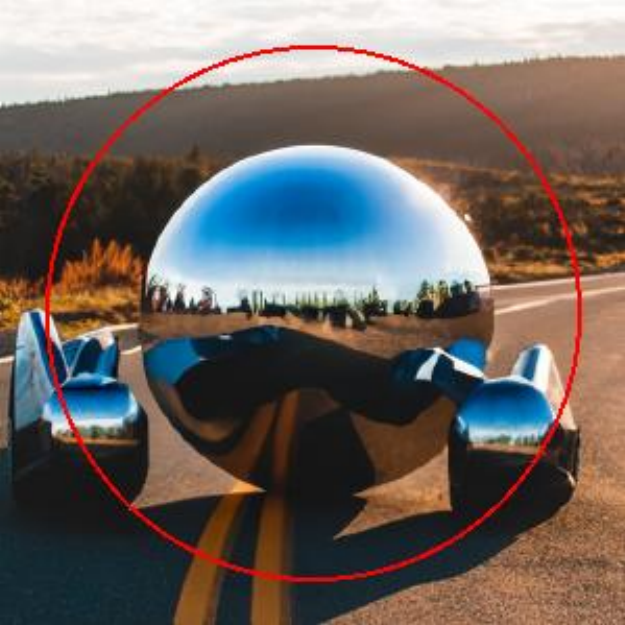}} &  
        \noindent\parbox[c]{0.081\textwidth}{\includegraphics[width=0.081\textwidth]{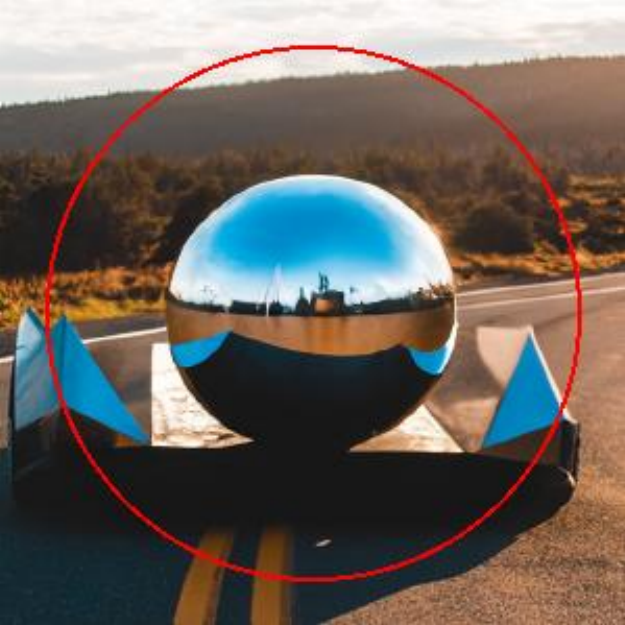}} & 
        \noindent\parbox[c]{0.081\textwidth}{\includegraphics[width=0.081\textwidth]{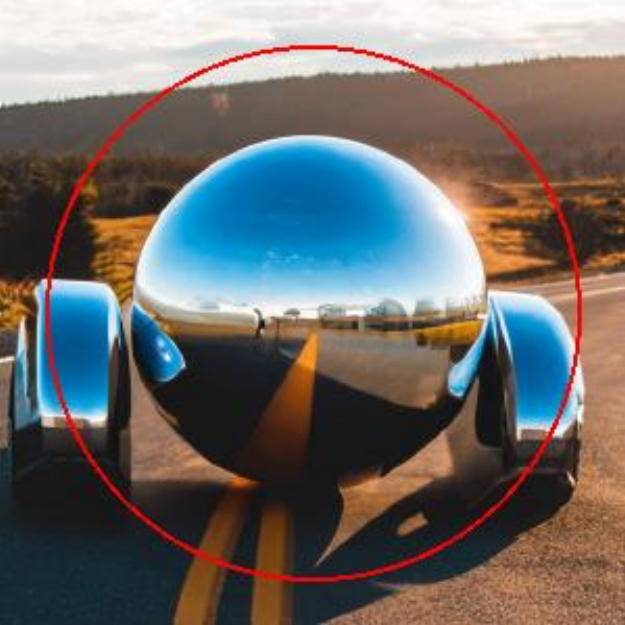}} & 
        \noindent\parbox[c]{0.081\textwidth}{\includegraphics[width=0.081\textwidth]{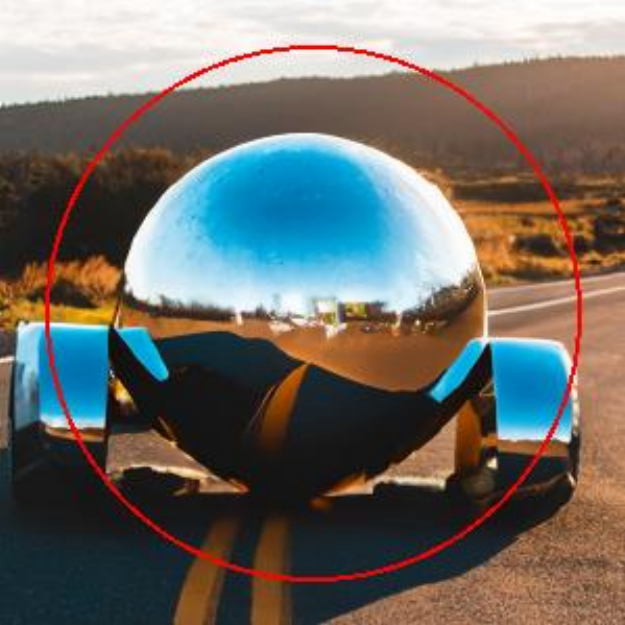}} & 
        \noindent\parbox[c]{0.081\textwidth}{\includegraphics[width=0.081\textwidth]{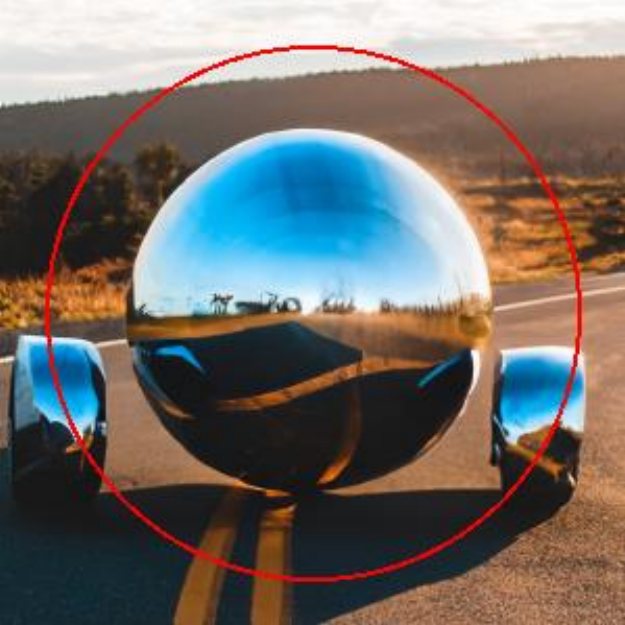}} & 
        \noindent\parbox[c]{0.081\textwidth}{\includegraphics[width=0.081\textwidth]{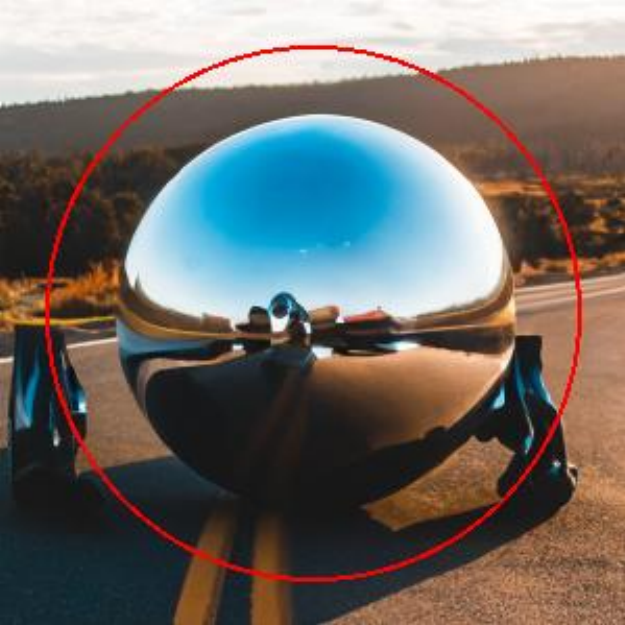}} & 
        \noindent\parbox[c]{0.081\textwidth}{\includegraphics[width=0.081\textwidth]{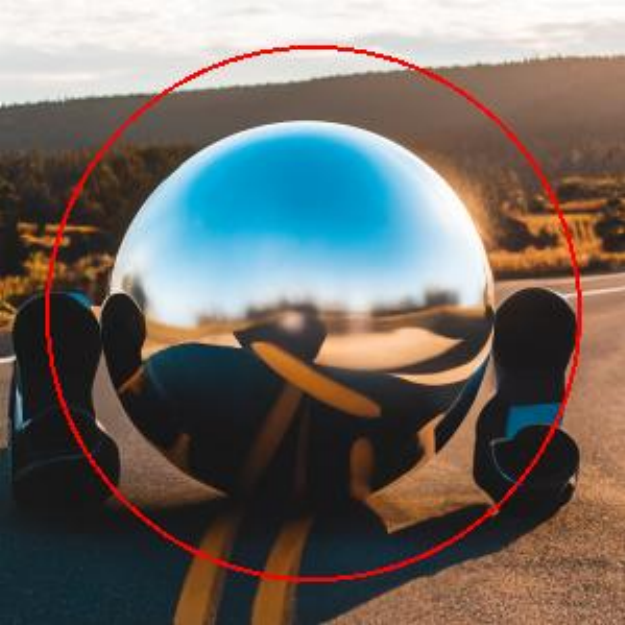}} & 
        \noindent\parbox[c]{0.081\textwidth}{\includegraphics[width=0.081\textwidth]{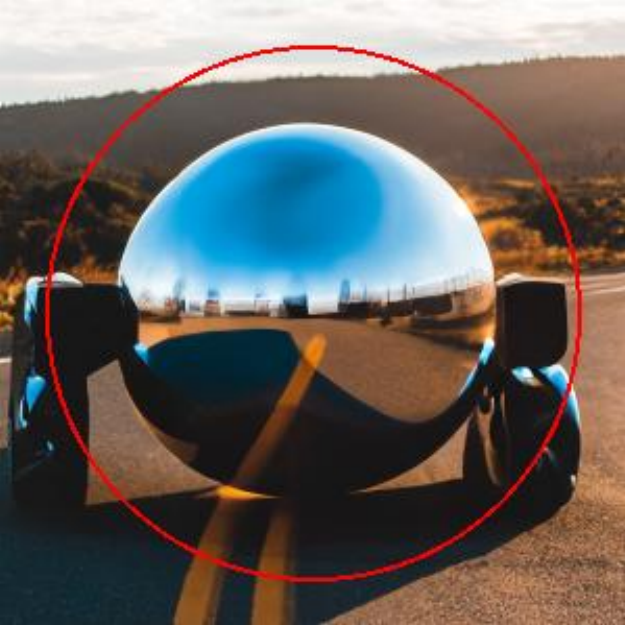}} & 
        \noindent\parbox[c]{0.081\textwidth}{\includegraphics[width=0.081\textwidth]{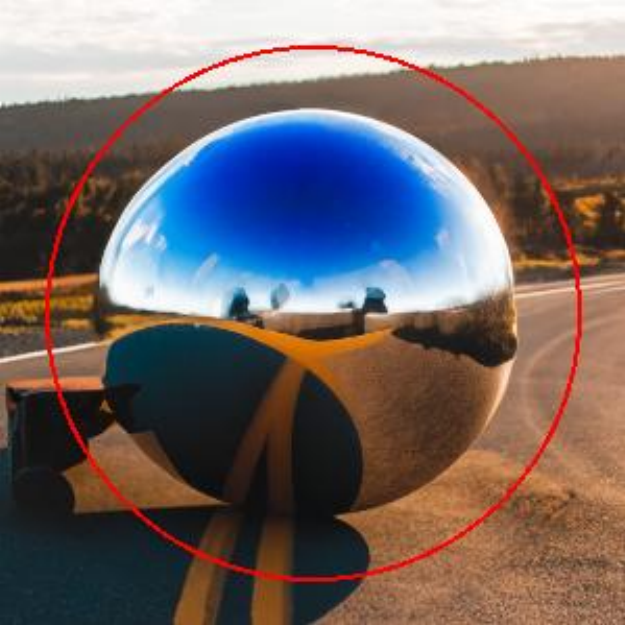}} & 

        \\

        \multicolumn{1}{l}{\rotatebox[origin=c]{90}{\shortstack[l]{\scriptsize SDXL \cite{podell2023sdxl}}}} &
        \noindent\parbox[c]{0.081\textwidth}{\includegraphics[width=0.081\textwidth]{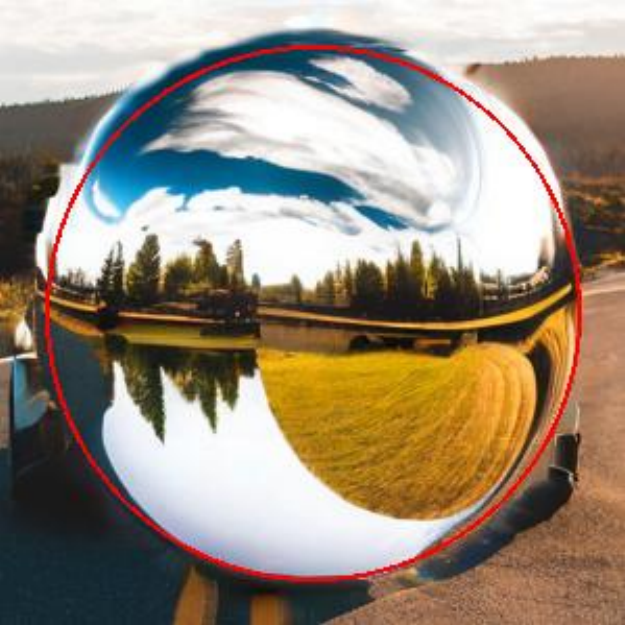}} & 
        \noindent\parbox[c]{0.081\textwidth}{\includegraphics[width=0.081\textwidth]{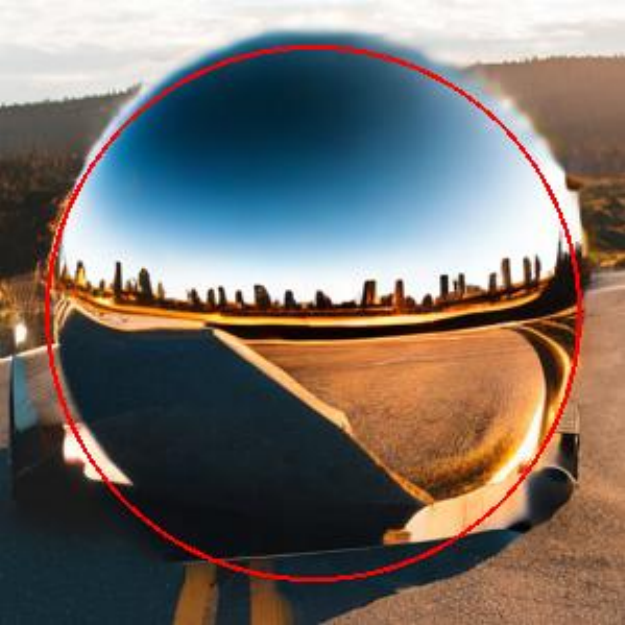}} &  
        \noindent\parbox[c]{0.081\textwidth}{\includegraphics[width=0.081\textwidth]{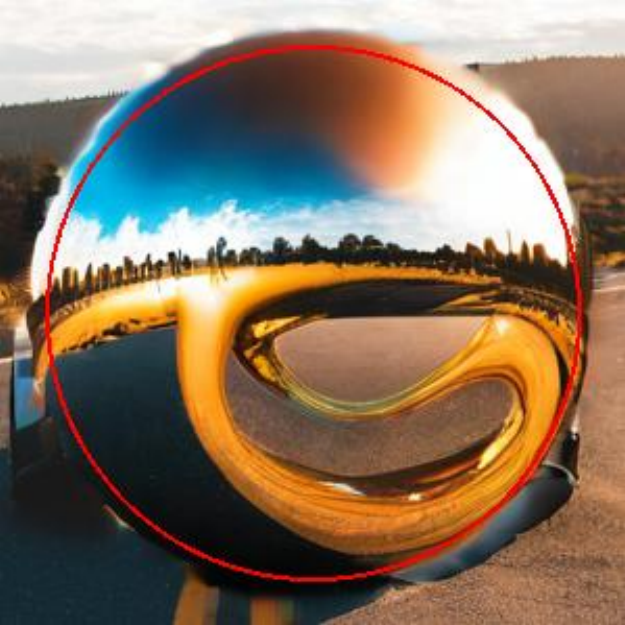}} & 
        \noindent\parbox[c]{0.081\textwidth}{\includegraphics[width=0.081\textwidth]{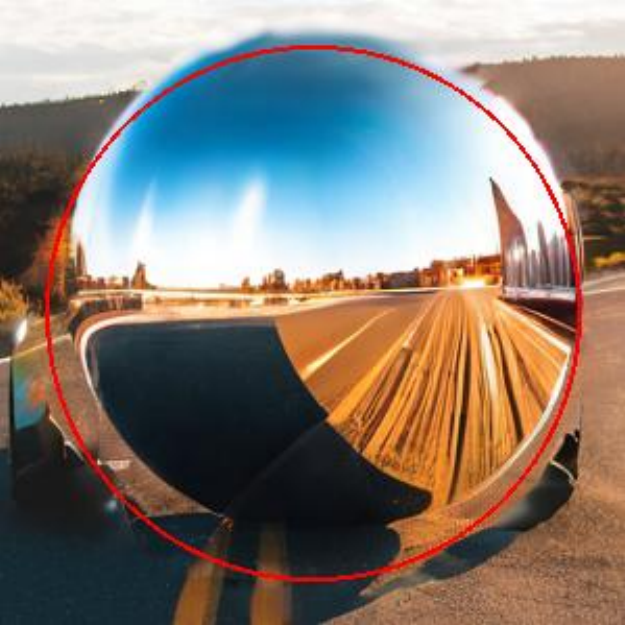}} & 
        \noindent\parbox[c]{0.081\textwidth}{\includegraphics[width=0.081\textwidth]{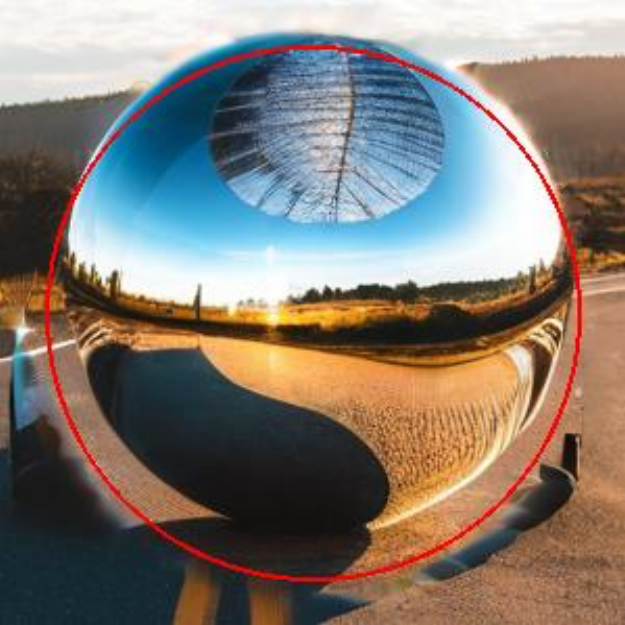}} & 
        \noindent\parbox[c]{0.081\textwidth}{\includegraphics[width=0.081\textwidth]{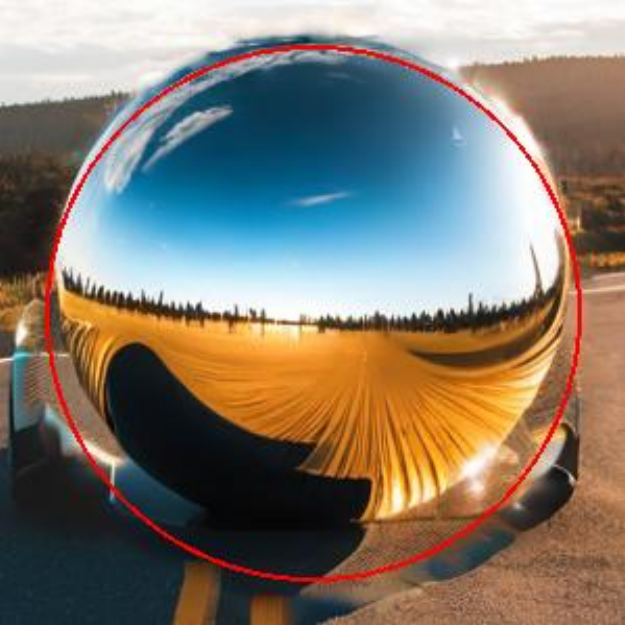}} & 
        \noindent\parbox[c]{0.081\textwidth}{\includegraphics[width=0.081\textwidth]{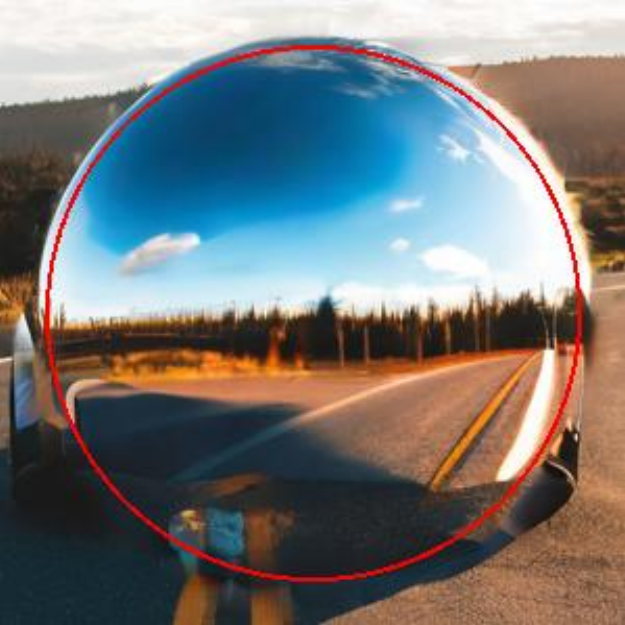}} & 
        \noindent\parbox[c]{0.081\textwidth}{\includegraphics[width=0.081\textwidth]{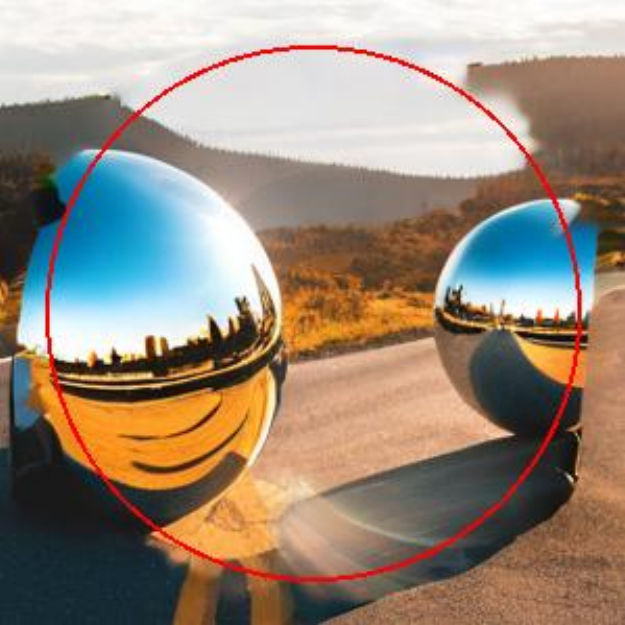}} & 
        \noindent\parbox[c]{0.081\textwidth}{\includegraphics[width=0.081\textwidth]{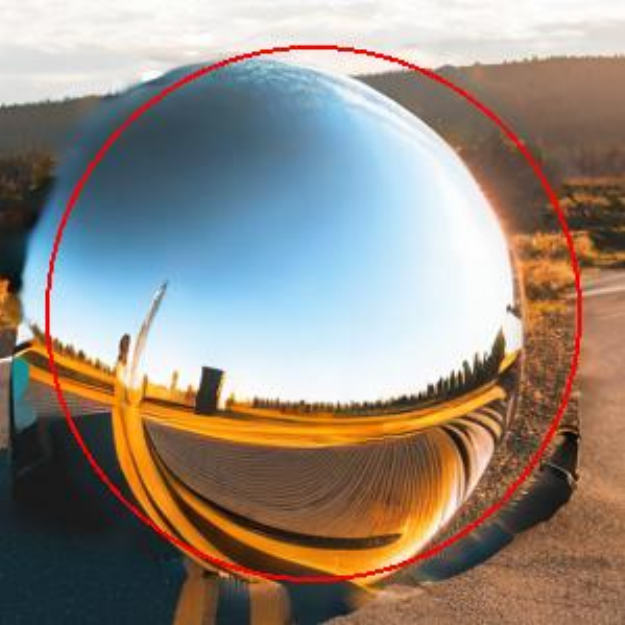}} & 
        \noindent\parbox[c]{0.081\textwidth}{\includegraphics[width=0.081\textwidth]{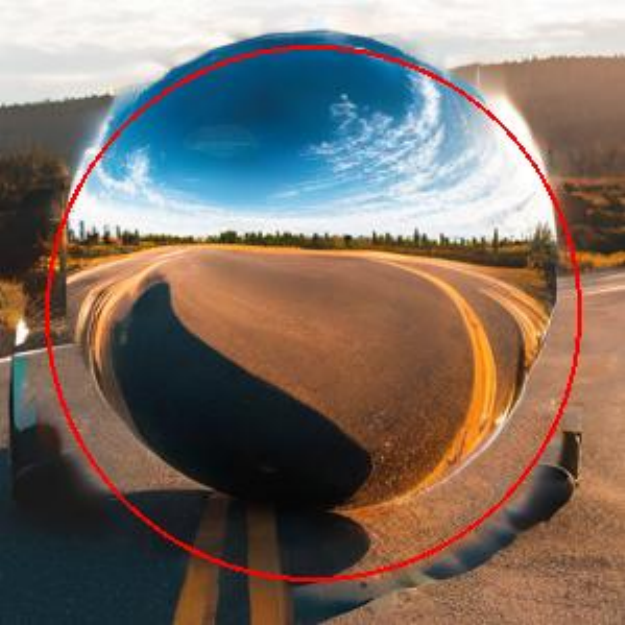}} & 
        
        \\ \hline

        \multicolumn{1}{l}{\rotatebox[origin=c]{90}{\shortstack[l]{\scriptsize \textbf{Ours}}}} &
        \noindent\parbox[c]{0.081\textwidth}{\includegraphics[width=0.081\textwidth]{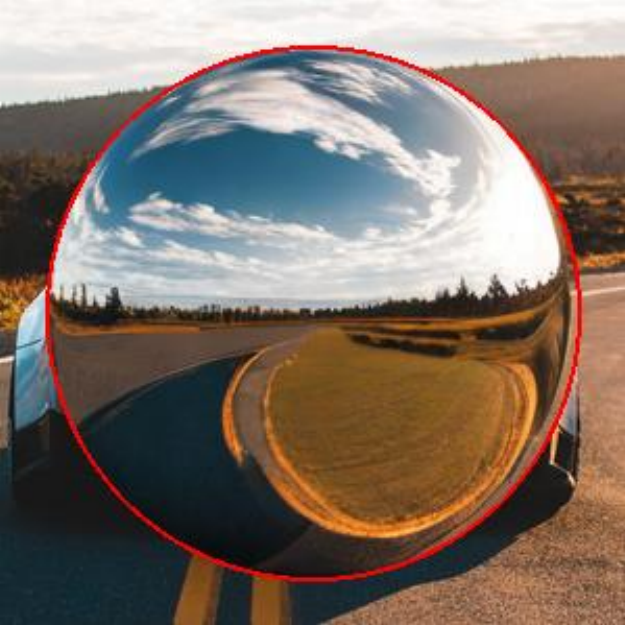}} & 
        \noindent\parbox[c]{0.081\textwidth}{\includegraphics[width=0.081\textwidth]{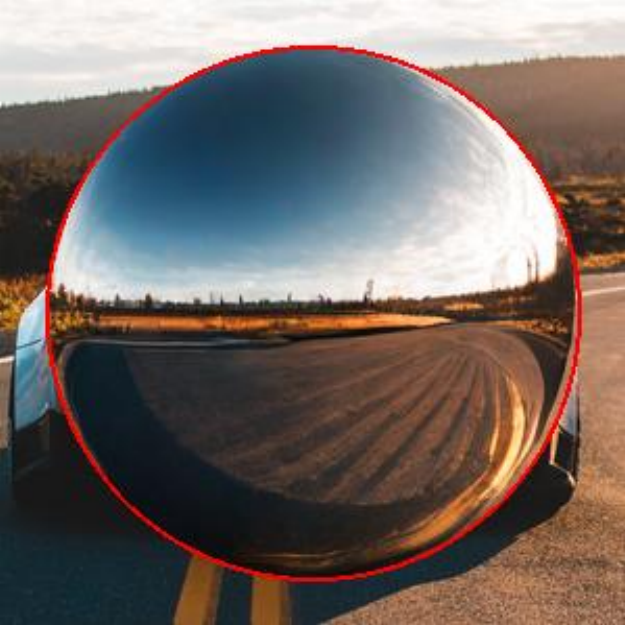}} &  
        \noindent\parbox[c]{0.081\textwidth}{\includegraphics[width=0.081\textwidth]{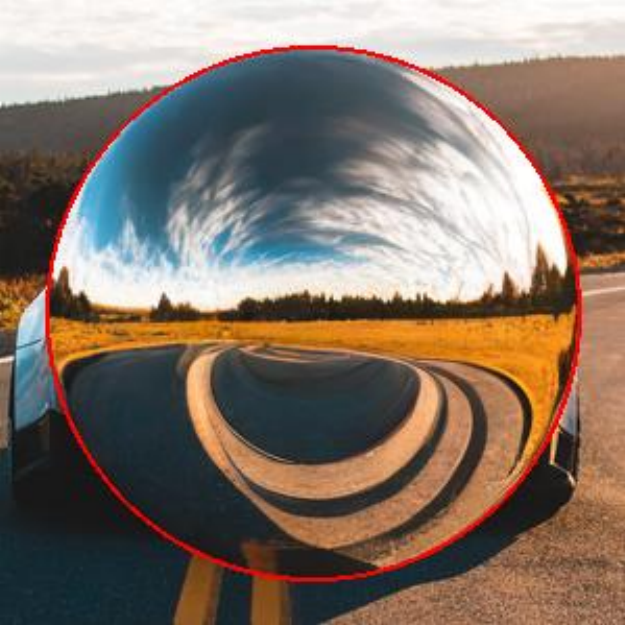}} & 
        \noindent\parbox[c]{0.081\textwidth}{\includegraphics[width=0.081\textwidth]{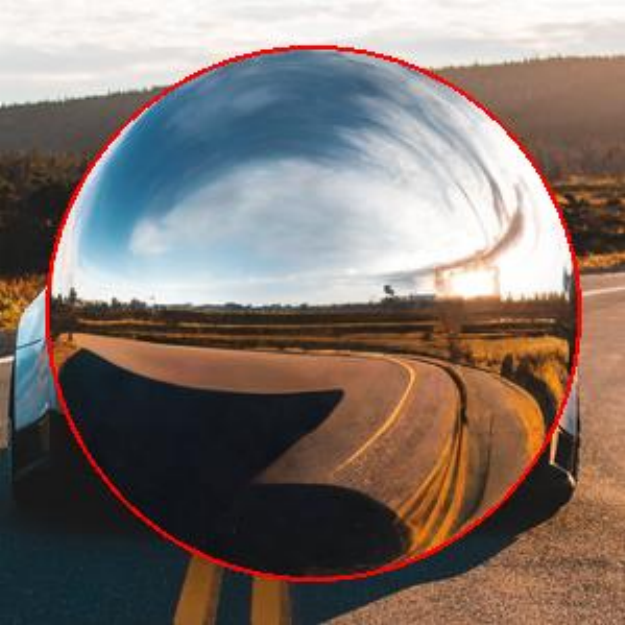}} & 
        \noindent\parbox[c]{0.081\textwidth}{\includegraphics[width=0.081\textwidth]{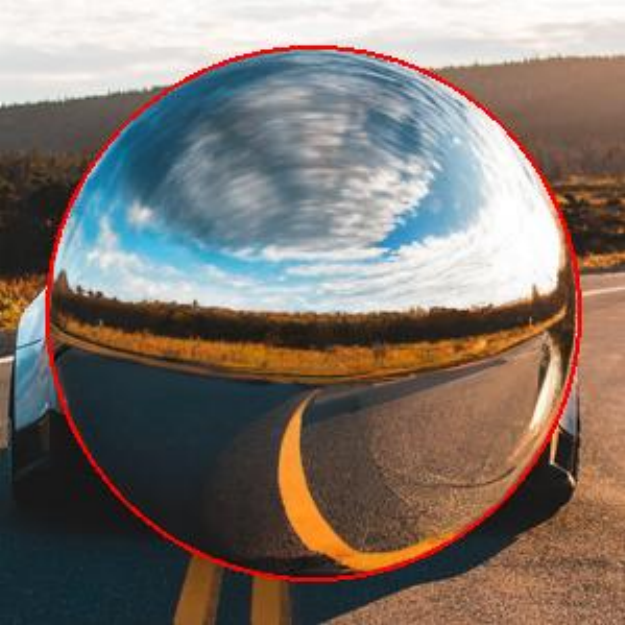}} & 
        \noindent\parbox[c]{0.081\textwidth}{\includegraphics[width=0.081\textwidth]{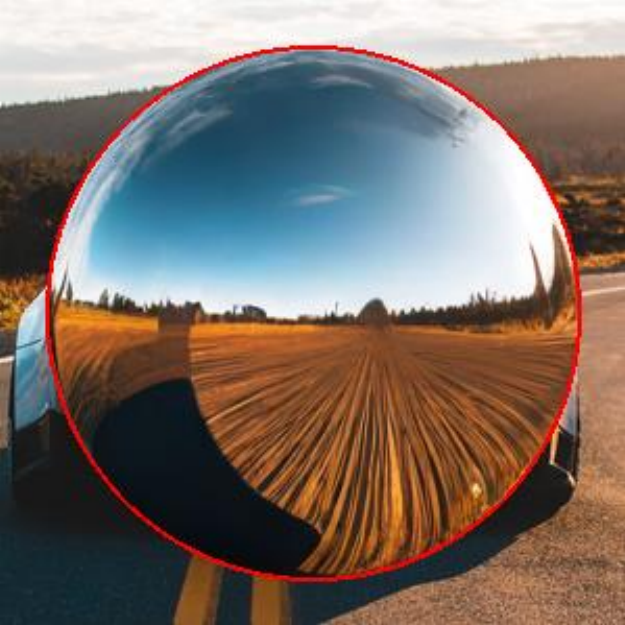}} & 
        \noindent\parbox[c]{0.081\textwidth}{\includegraphics[width=0.081\textwidth]{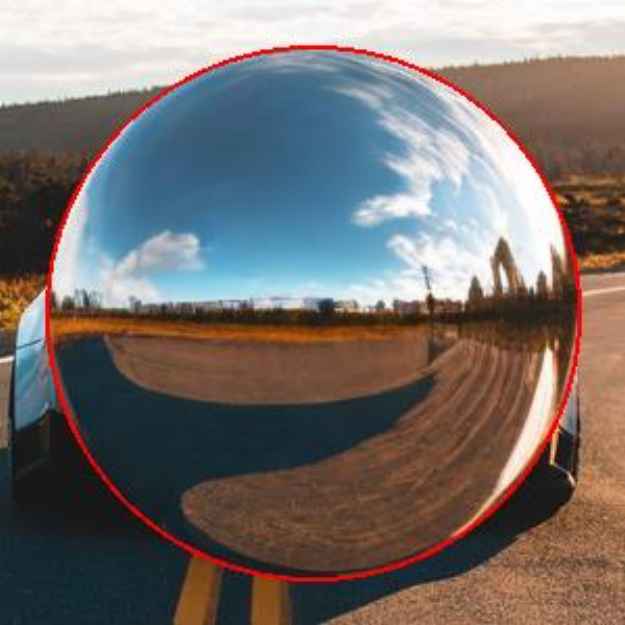}} & 
        \noindent\parbox[c]{0.081\textwidth}{\includegraphics[width=0.081\textwidth]{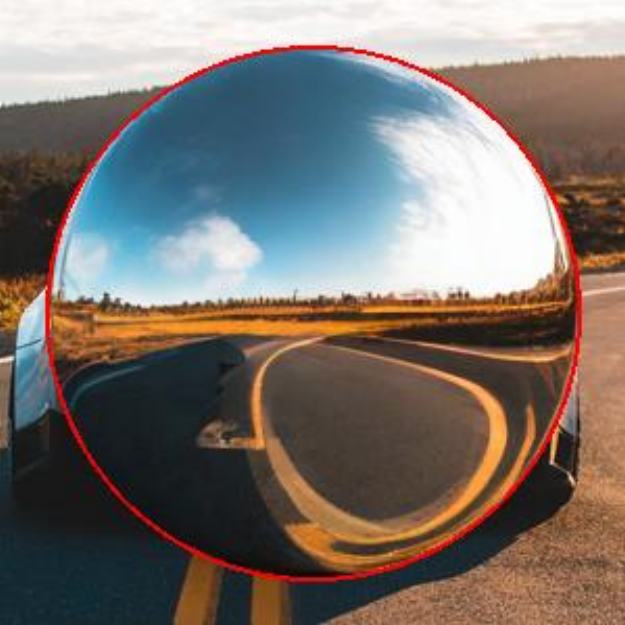}} & 
        \noindent\parbox[c]{0.081\textwidth}{\includegraphics[width=0.081\textwidth]{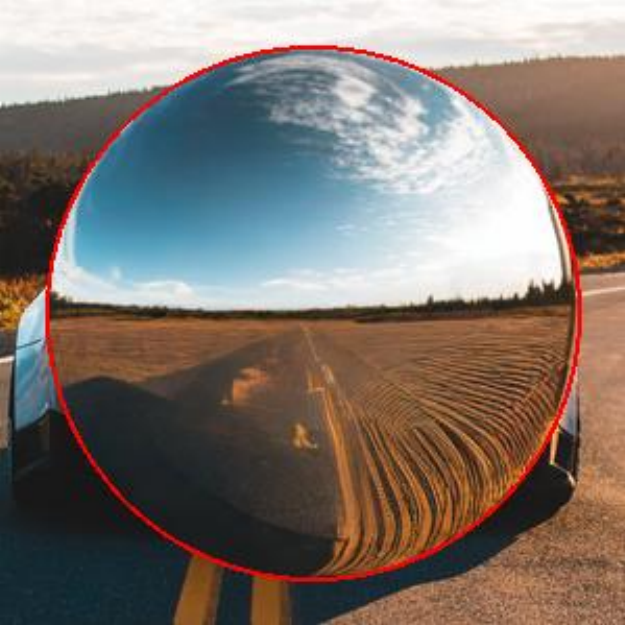}} & 
        \noindent\parbox[c]{0.081\textwidth}{\includegraphics[width=0.081\textwidth]{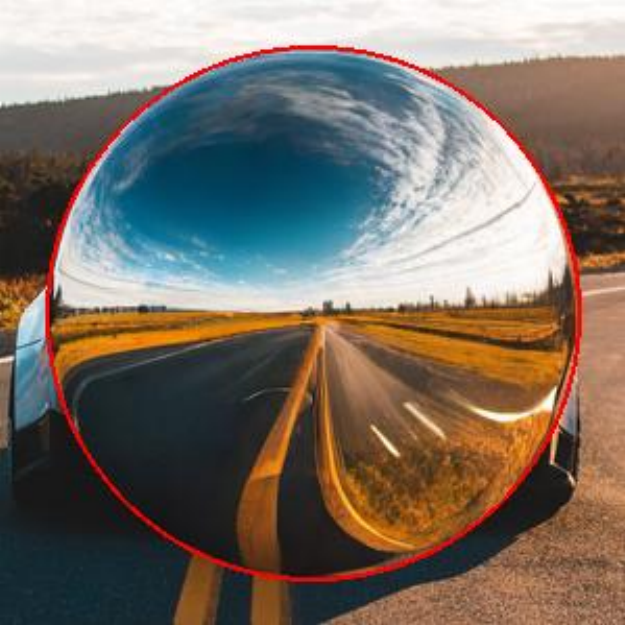}}
        
        \\
        
    \end{tabu}

    \caption{Chrome ball inpainting results from various methods. The
    red circle indicates the inpainted region, and we show a zoomed-in
    view of the blue crop. Each row contains results from ten different random seeds.}
    \label{fig:aba_seed-cherry}
\end{figure*}

\begin{figure*}
    \centering

    \begin{tabu} to \textwidth {
        @{}
        l@{}
        l@{\hspace{0.5pt}}
    }
        \multicolumn{1}{l}{\rotatebox[origin=c]{90}{\shortstack[l]{\scriptsize Input \\ \scriptsize image}}} &
        \noindent\parbox[c]{0.5\textwidth}{\includegraphics[width=0.5\textwidth]{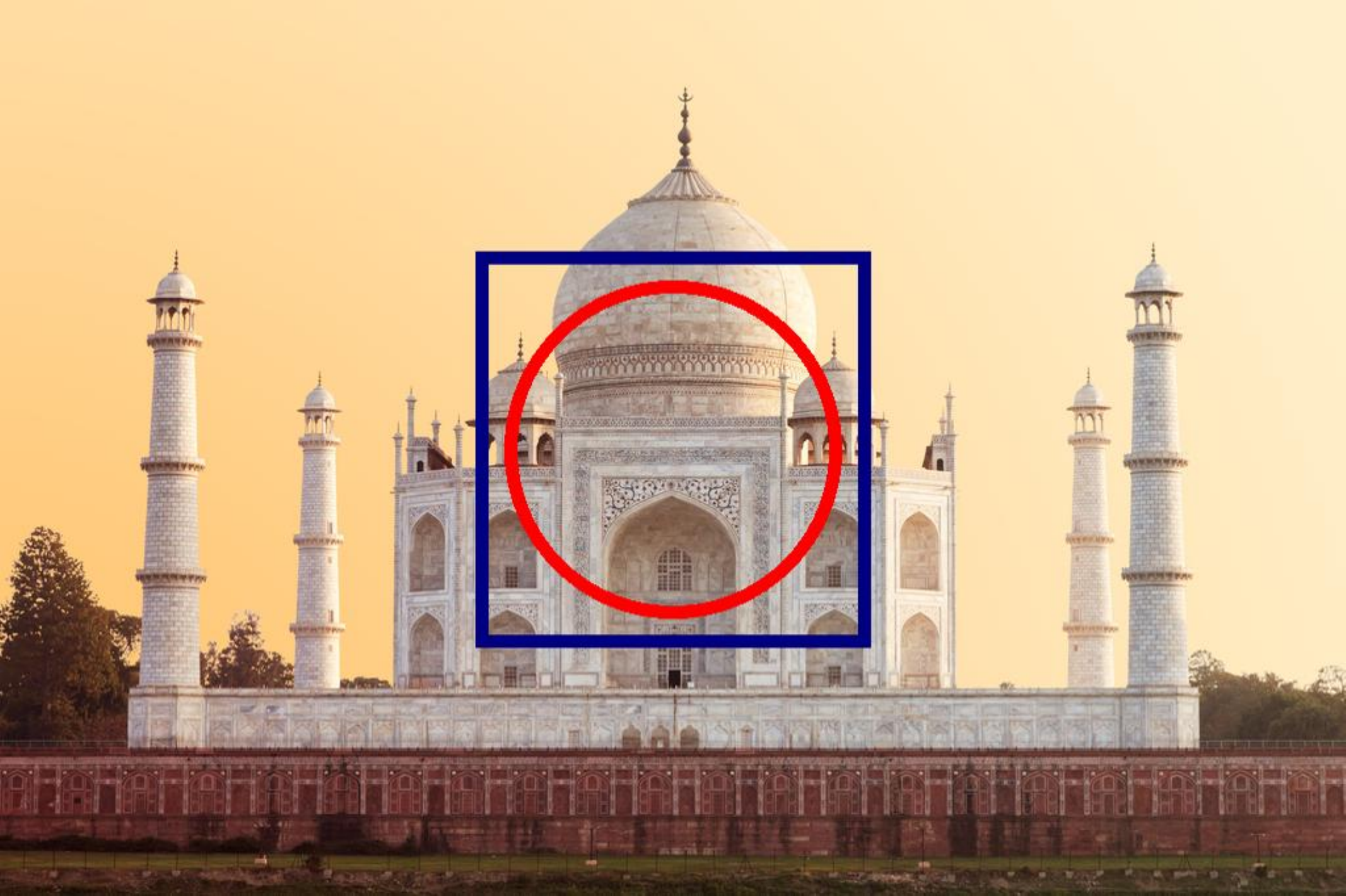}}  \\
    \end{tabu}
    
    \begin{tabu} to \textwidth {
        @{}
        c@{}
        c@{\hspace{0.5pt}}
        c@{\hspace{0.5pt}}
        c@{\hspace{0.5pt}}
        c@{\hspace{0.5pt}}
        c@{\hspace{0.5pt}}
        c@{\hspace{0.5pt}}
        c@{\hspace{0.5pt}}
        c@{\hspace{0.5pt}}
        c@{\hspace{0.5pt}}
        c@{\hspace{0.5pt}}
        c@{}
    }
        

        \multicolumn{1}{l}{\rotatebox[origin=c]{90}{\shortstack[l]{\scriptsize Blended Dif-\\ \scriptsize fusion \cite{avrahami2023blendedlatent, avrahami2022blendeddiffusion}}}} &
        \noindent\parbox[c]{0.081\textwidth}{\includegraphics[width=0.081\textwidth]{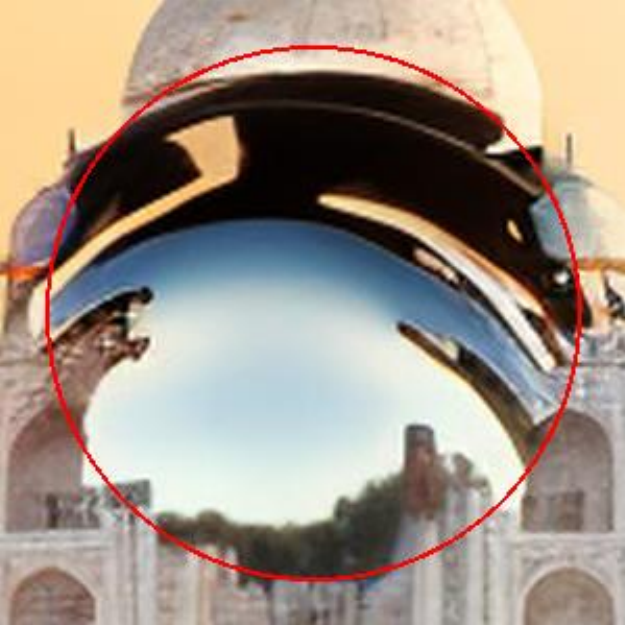}} & 
        \noindent\parbox[c]{0.081\textwidth}{\includegraphics[width=0.081\textwidth]{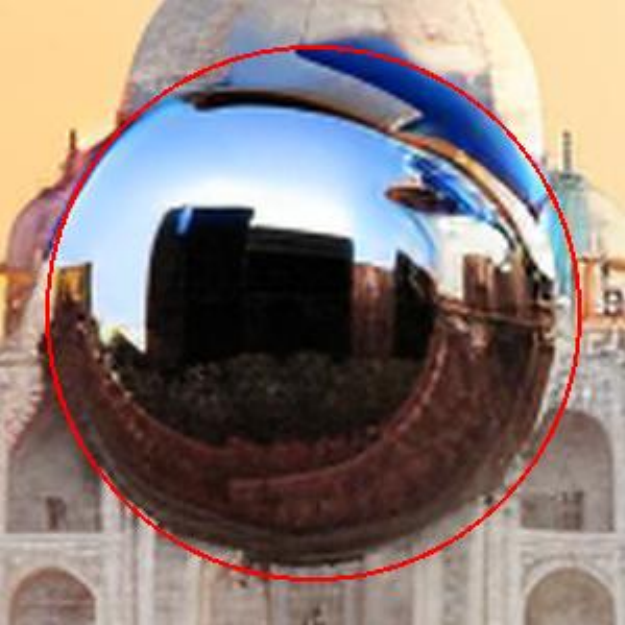}} &  
        \noindent\parbox[c]{0.081\textwidth}{\includegraphics[width=0.081\textwidth]{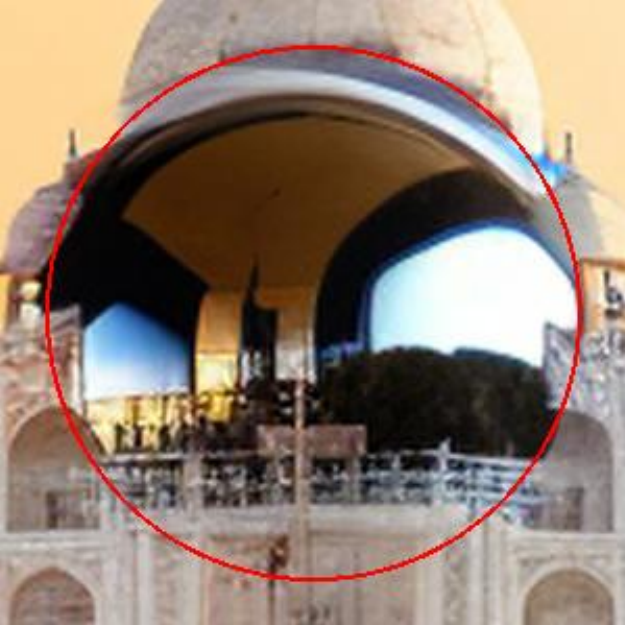}} & 
        \noindent\parbox[c]{0.081\textwidth}{\includegraphics[width=0.081\textwidth]{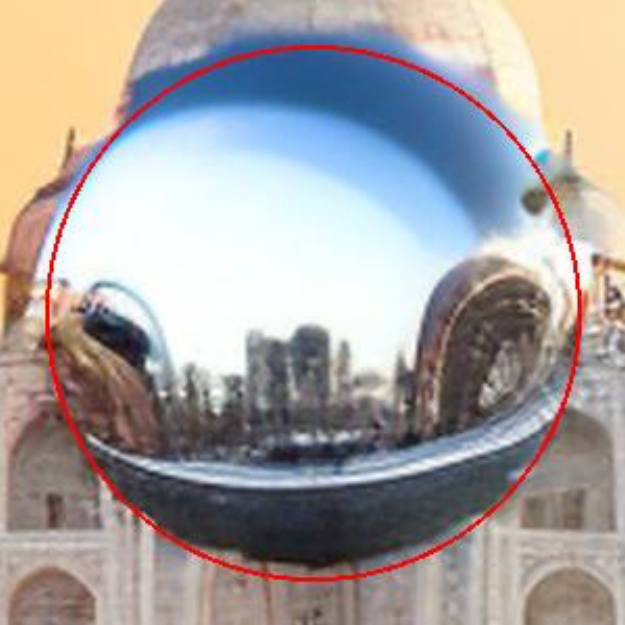}} & 
        \noindent\parbox[c]{0.081\textwidth}{\includegraphics[width=0.081\textwidth]{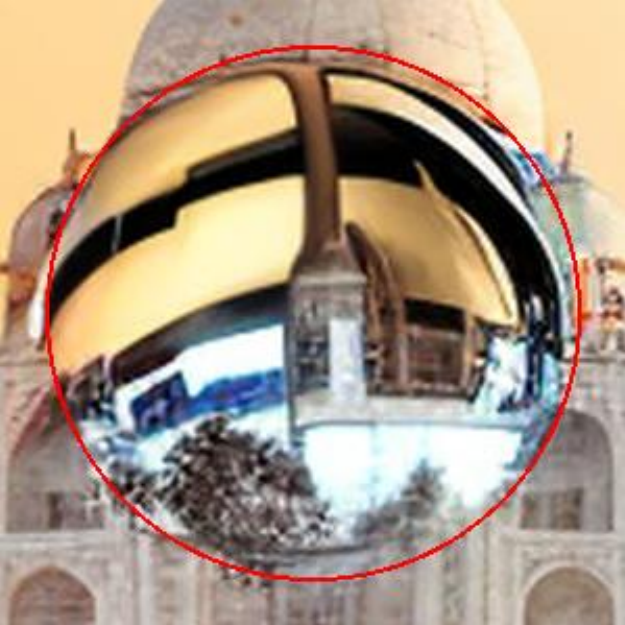}} & 
        \noindent\parbox[c]{0.081\textwidth}{\includegraphics[width=0.081\textwidth]{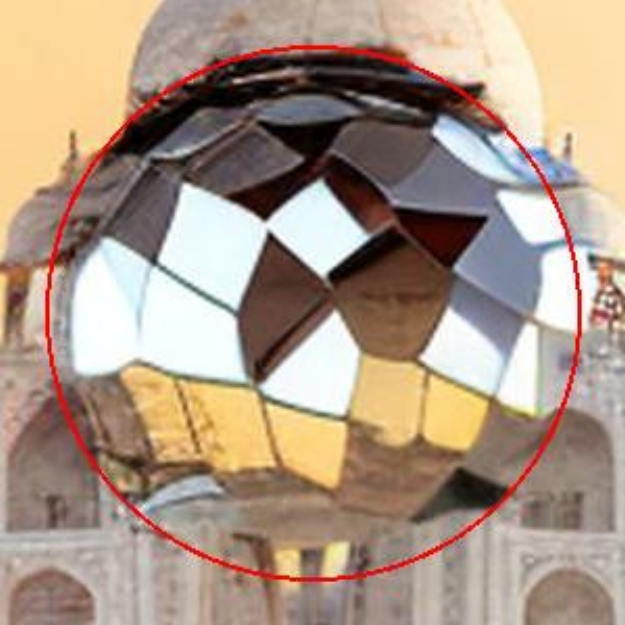}} & 
        \noindent\parbox[c]{0.081\textwidth}{\includegraphics[width=0.081\textwidth]{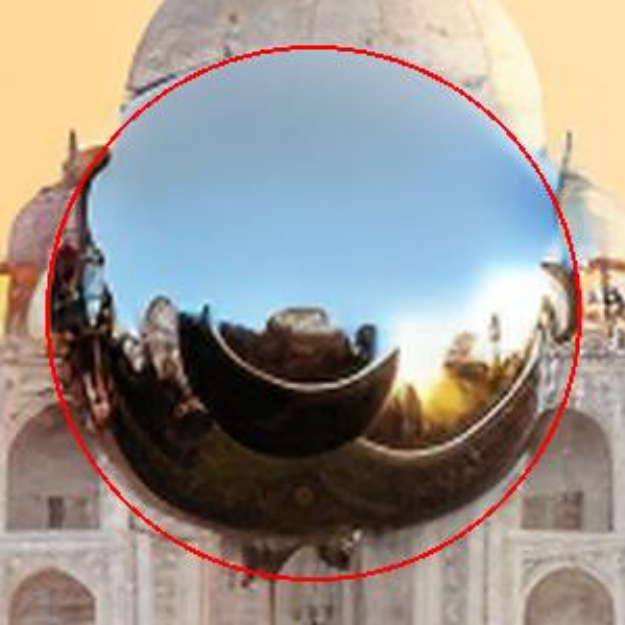}} & 
        \noindent\parbox[c]{0.081\textwidth}{\includegraphics[width=0.081\textwidth]{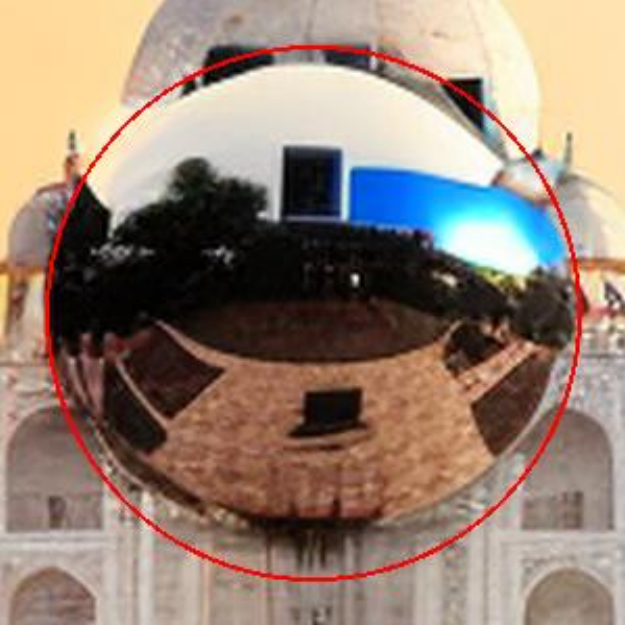}} & 
        \noindent\parbox[c]{0.081\textwidth}{\includegraphics[width=0.081\textwidth]{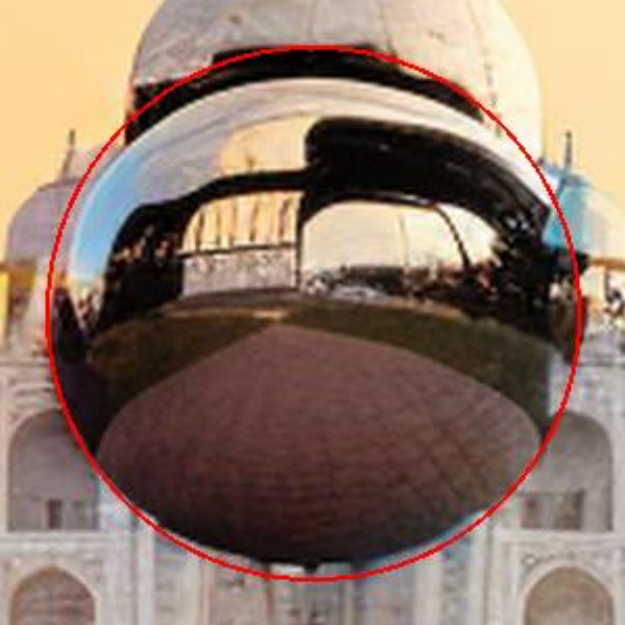}} & 
        \noindent\parbox[c]{0.081\textwidth}{\includegraphics[width=0.081\textwidth]{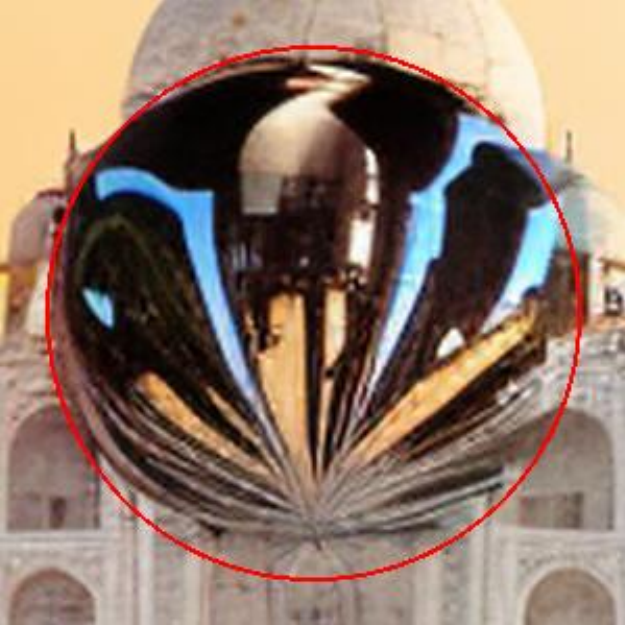}} & 
        
        \\

        \multicolumn{1}{l}{\rotatebox[origin=c]{90}{\shortstack[l]{\scriptsize Paint-by-Ex\\ \scriptsize ample \cite{yang2023paint}}}} &
        \noindent\parbox[c]{0.081\textwidth}{\includegraphics[width=0.081\textwidth]{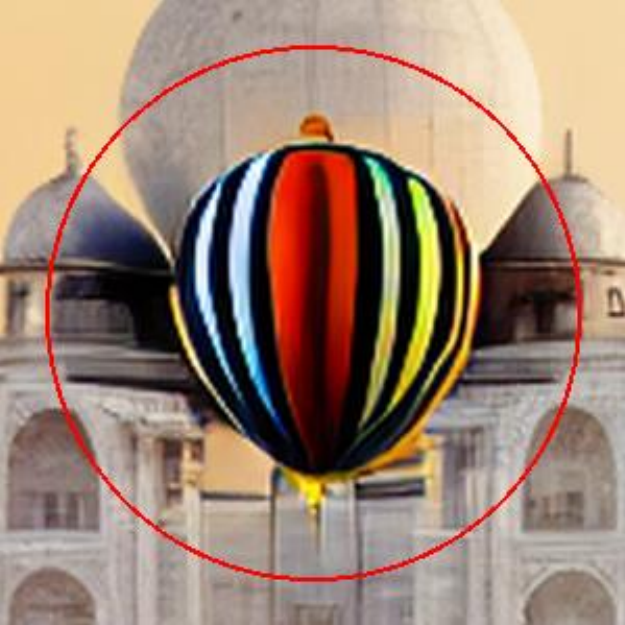}} & 
        \noindent\parbox[c]{0.081\textwidth}{\includegraphics[width=0.081\textwidth]{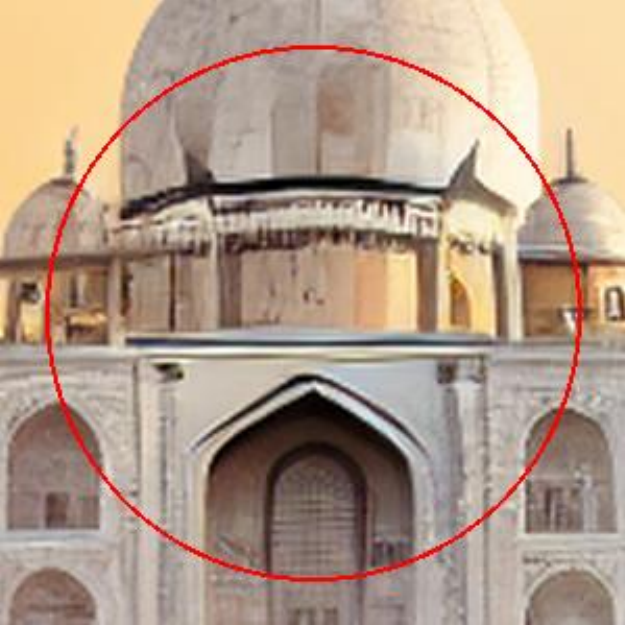}} &  
        \noindent\parbox[c]{0.081\textwidth}{\includegraphics[width=0.081\textwidth]{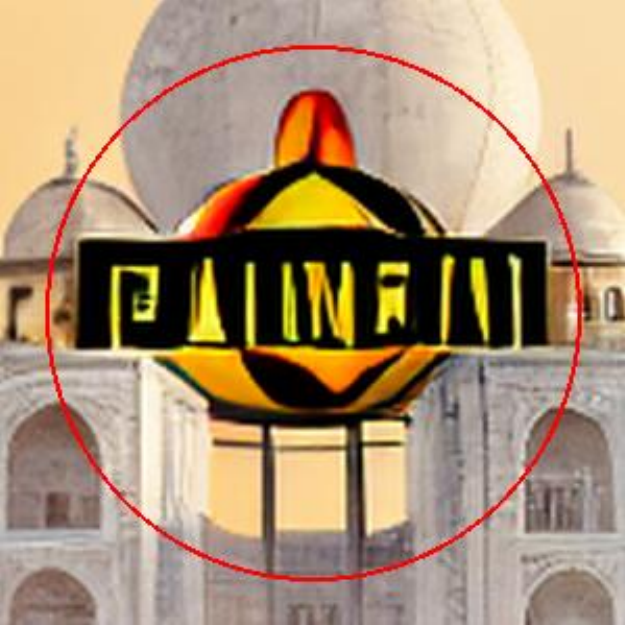}} & 
        \noindent\parbox[c]{0.081\textwidth}{\includegraphics[width=0.081\textwidth]{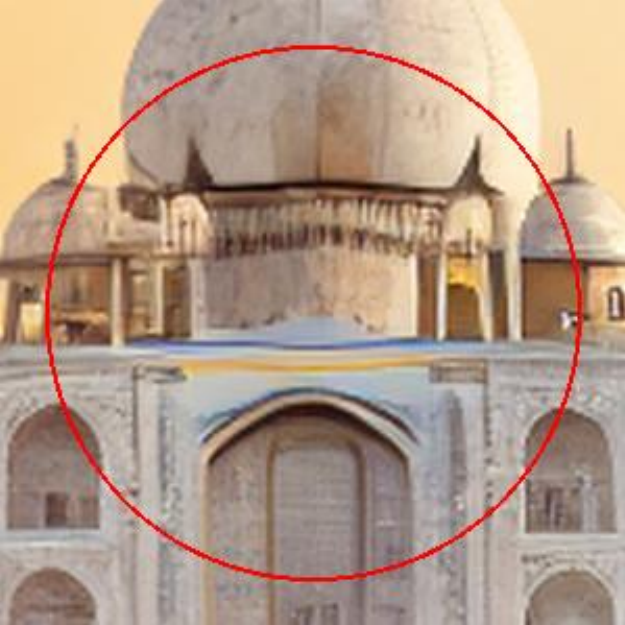}} & 
        \noindent\parbox[c]{0.081\textwidth}{\includegraphics[width=0.081\textwidth]{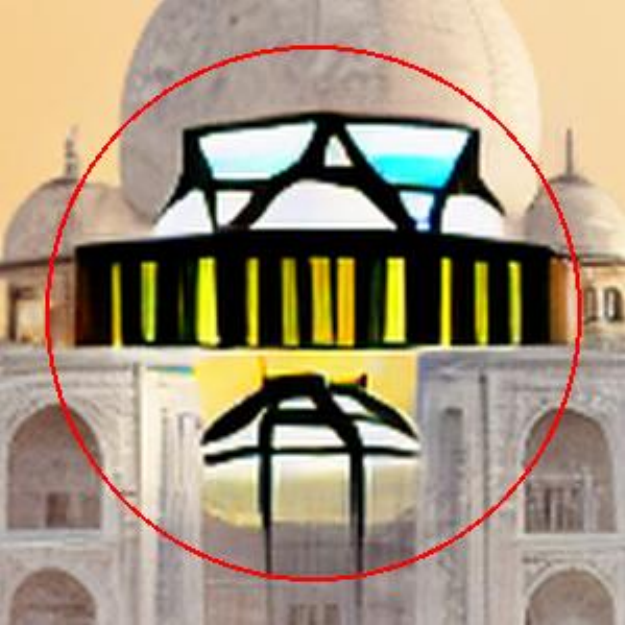}} & 
        \noindent\parbox[c]{0.081\textwidth}{\includegraphics[width=0.081\textwidth]{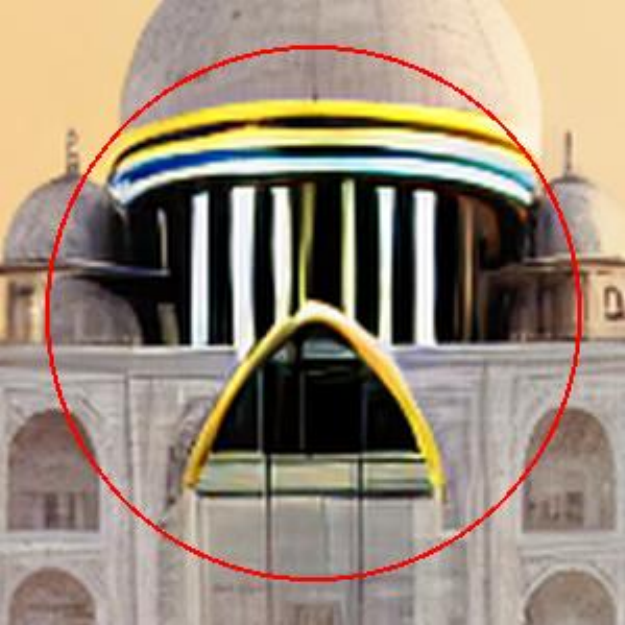}} & 
        \noindent\parbox[c]{0.081\textwidth}{\includegraphics[width=0.081\textwidth]{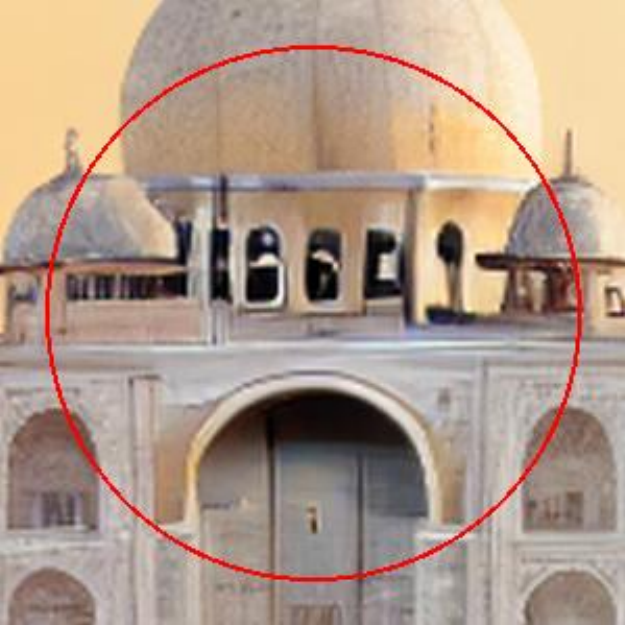}} & 
        \noindent\parbox[c]{0.081\textwidth}{\includegraphics[width=0.081\textwidth]{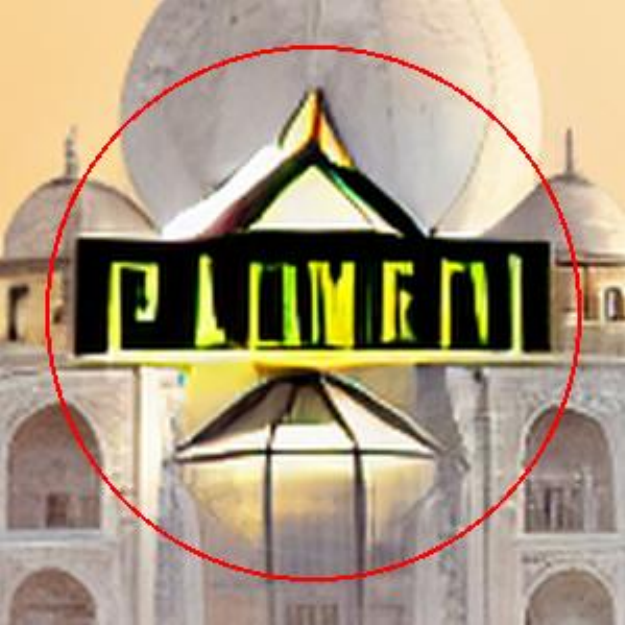}} & 
        \noindent\parbox[c]{0.081\textwidth}{\includegraphics[width=0.081\textwidth]{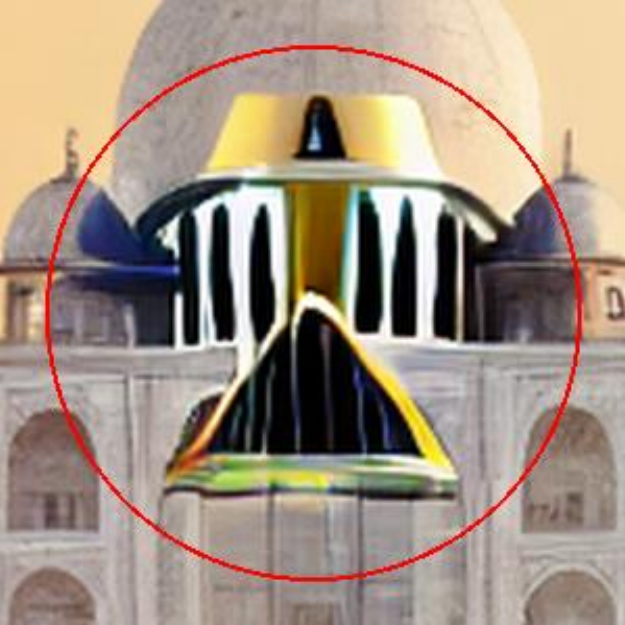}} & 
        \noindent\parbox[c]{0.081\textwidth}{\includegraphics[width=0.081\textwidth]{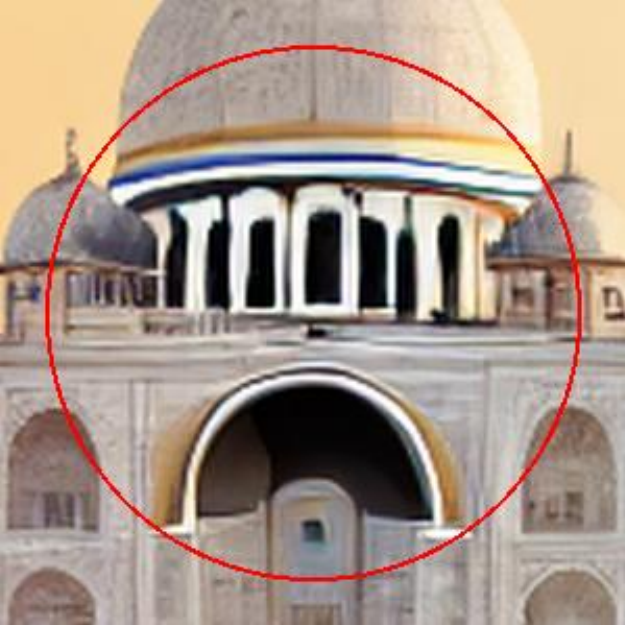}} & 
        
        \\

        \multicolumn{1}{l}{\rotatebox[origin=c]{90}{\shortstack[l]{\scriptsize IP-Adapter\\ \scriptsize \cite{ye2023ip-adapter}}}} &
        \noindent\parbox[c]{0.081\textwidth}{\includegraphics[width=0.081\textwidth]{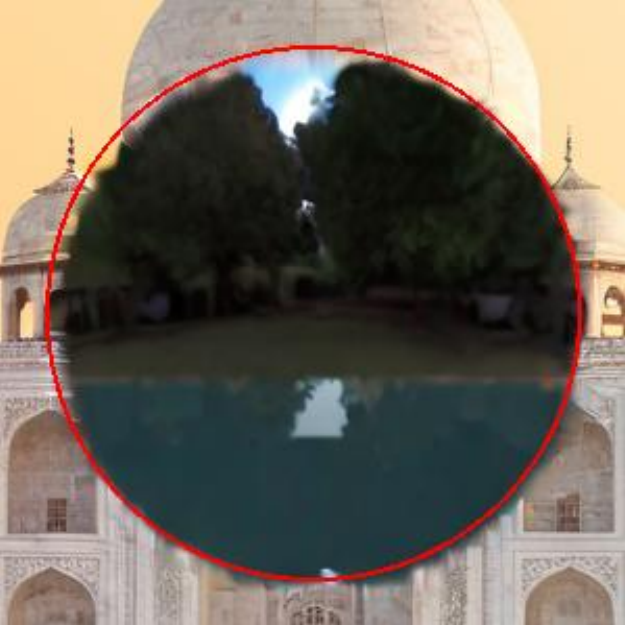}} & 
        \noindent\parbox[c]{0.081\textwidth}{\includegraphics[width=0.081\textwidth]{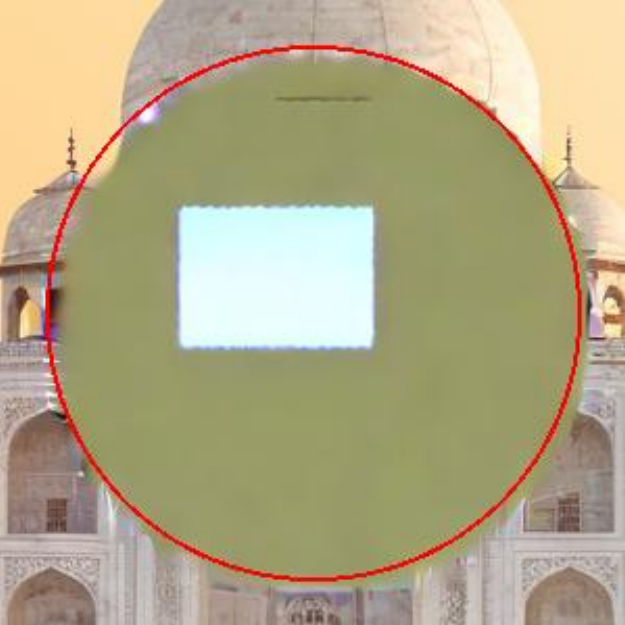}} &  
        \noindent\parbox[c]{0.081\textwidth}{\includegraphics[width=0.081\textwidth]{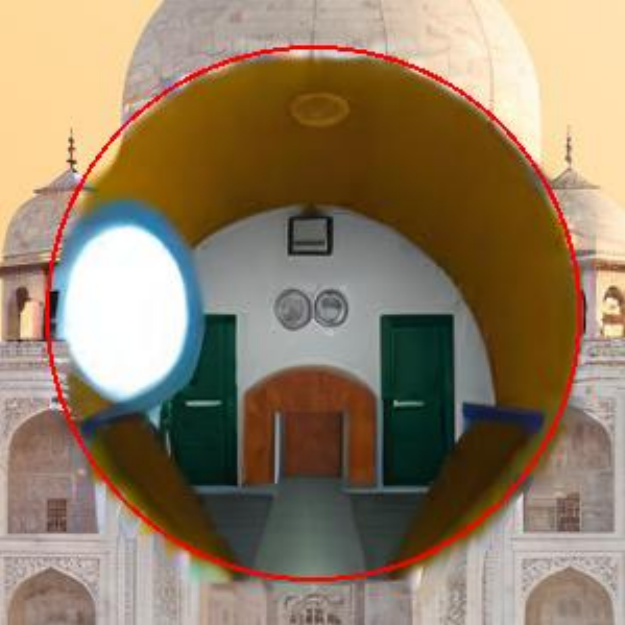}} & 
        \noindent\parbox[c]{0.081\textwidth}{\includegraphics[width=0.081\textwidth]{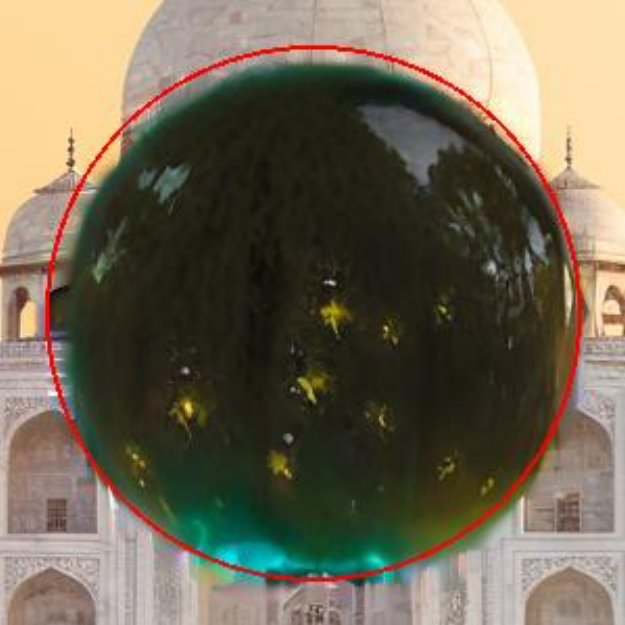}} & 
        \noindent\parbox[c]{0.081\textwidth}{\includegraphics[width=0.081\textwidth]{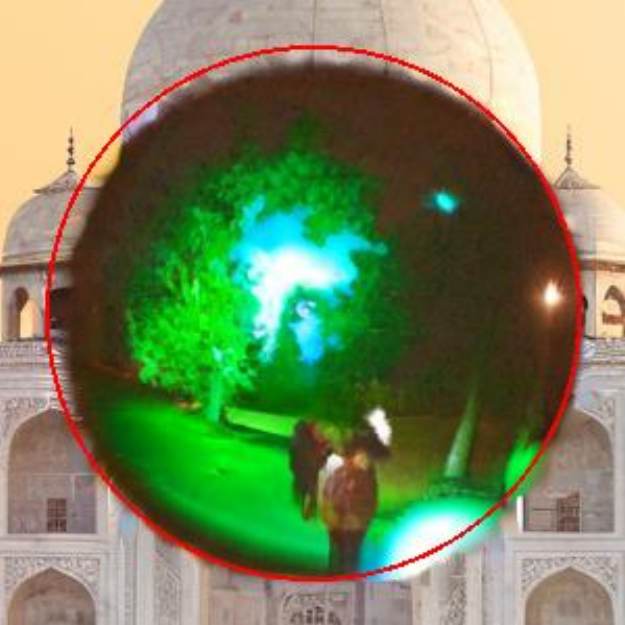}} & 
        \noindent\parbox[c]{0.081\textwidth}{\includegraphics[width=0.081\textwidth]{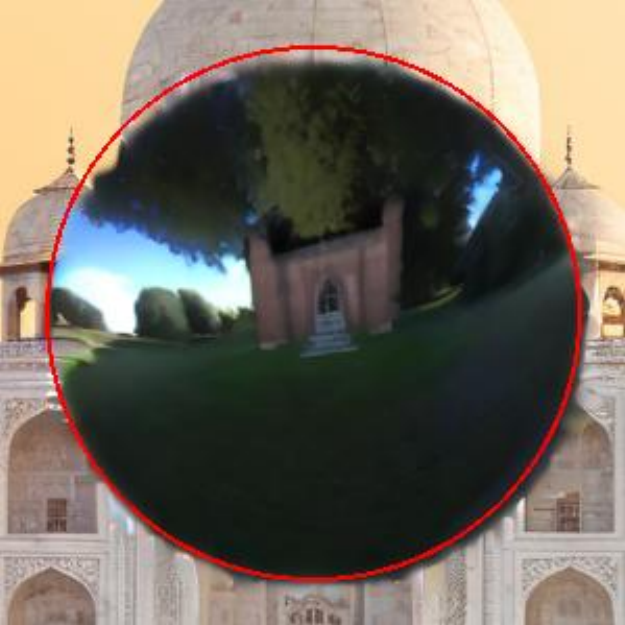}} & 
        \noindent\parbox[c]{0.081\textwidth}{\includegraphics[width=0.081\textwidth]{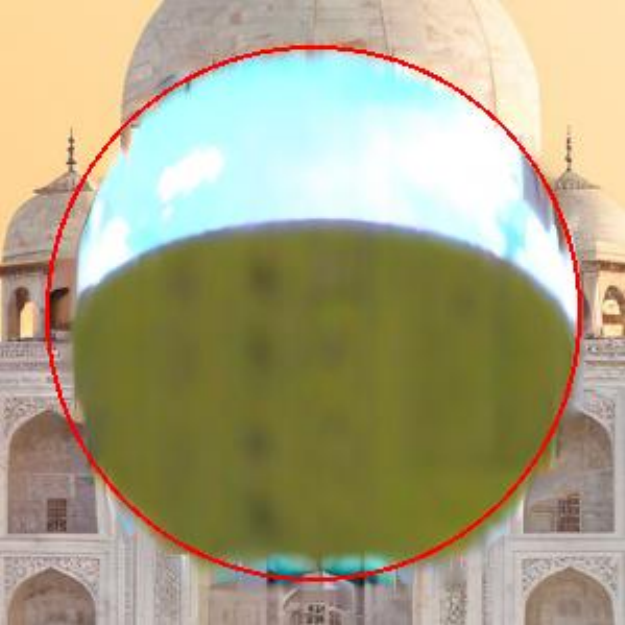}} & 
        \noindent\parbox[c]{0.081\textwidth}{\includegraphics[width=0.081\textwidth]{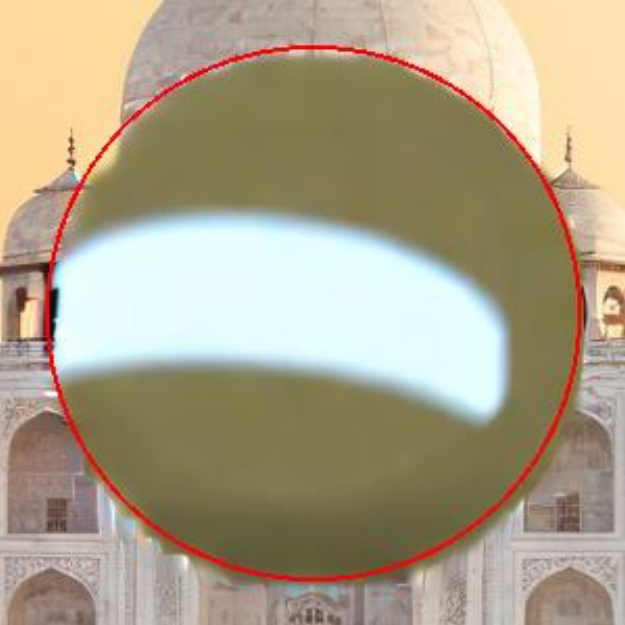}} & 
        \noindent\parbox[c]{0.081\textwidth}{\includegraphics[width=0.081\textwidth]{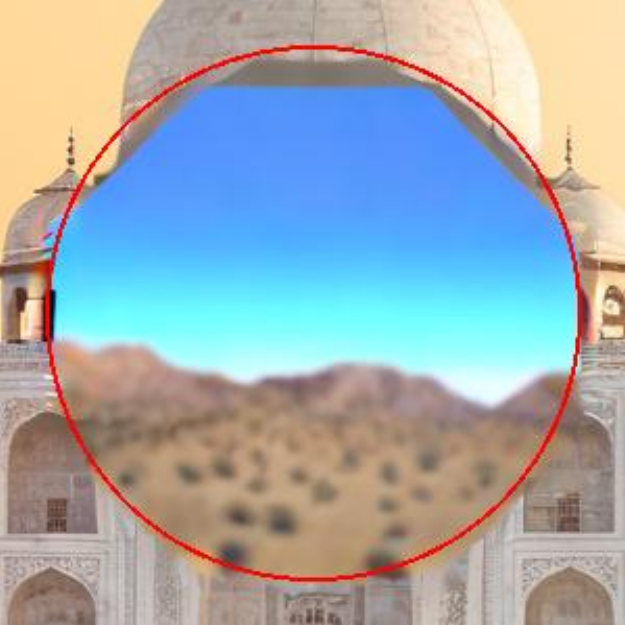}} & 
        \noindent\parbox[c]{0.081\textwidth}{\includegraphics[width=0.081\textwidth]{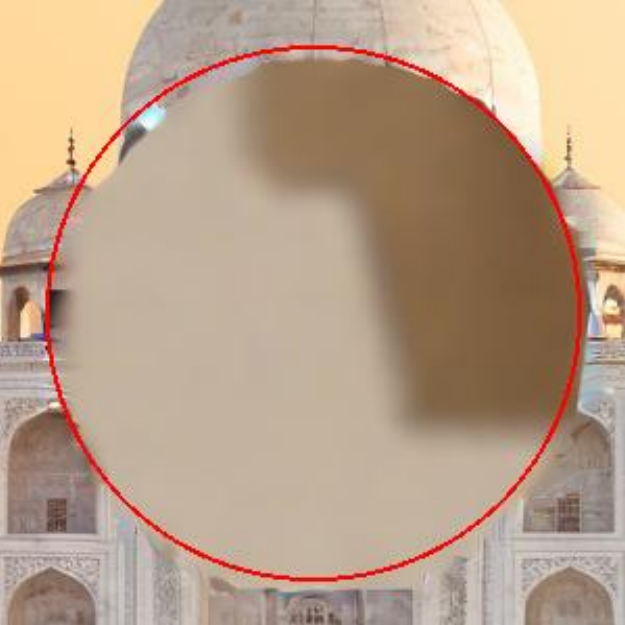}} & 
        
        \\

        \multicolumn{1}{l}{\rotatebox[origin=c]{90}{\shortstack[l]{\scriptsize DALL·E2 \cite{dalle2}}}} &
        \noindent\parbox[c]{0.081\textwidth}{\includegraphics[width=0.081\textwidth]{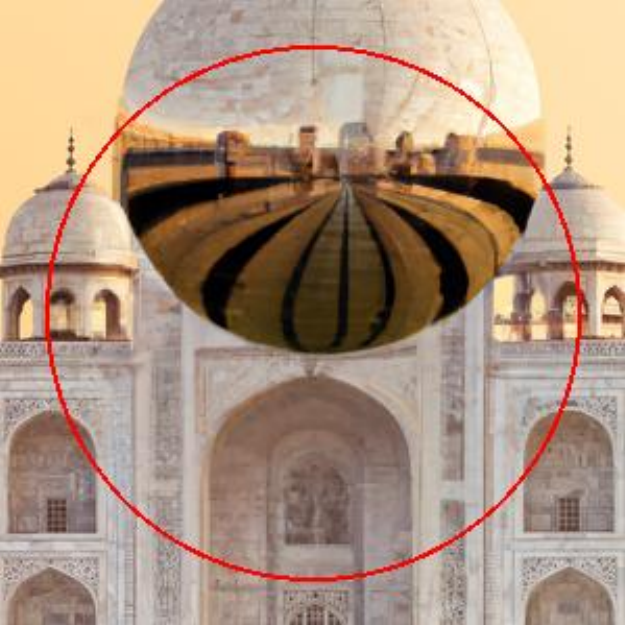}} & 
        \noindent\parbox[c]{0.081\textwidth}{\includegraphics[width=0.081\textwidth]{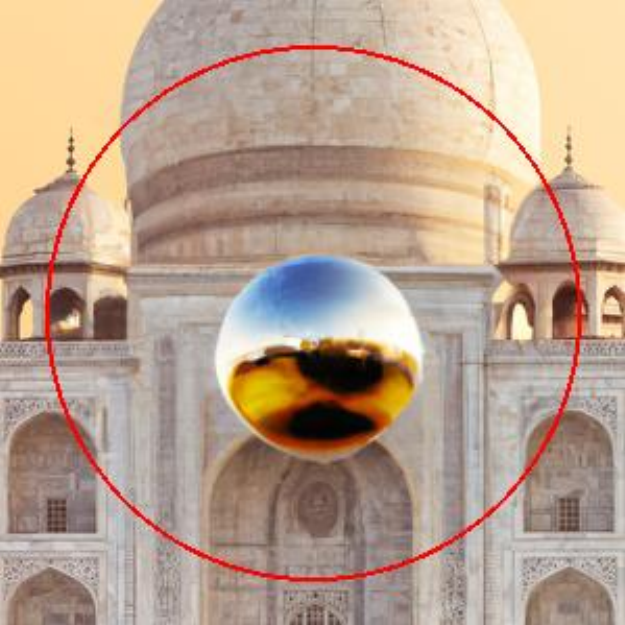}} &  
        \noindent\parbox[c]{0.081\textwidth}{\includegraphics[width=0.081\textwidth]{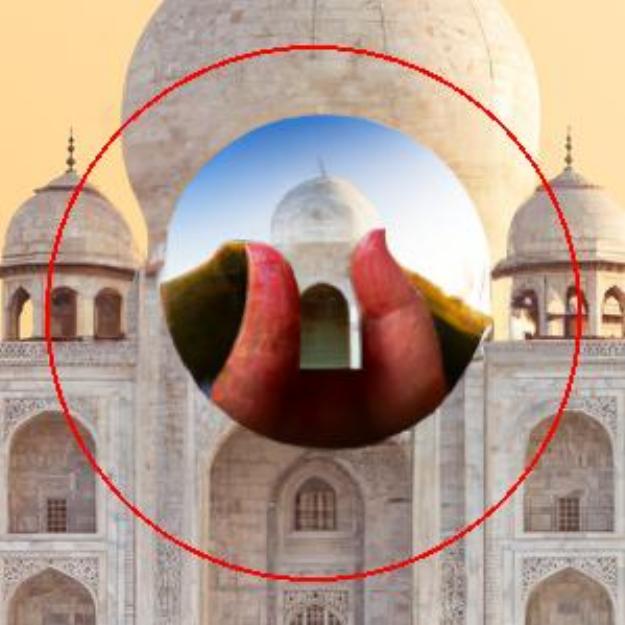}} & 
        \noindent\parbox[c]{0.081\textwidth}{\includegraphics[width=0.081\textwidth]{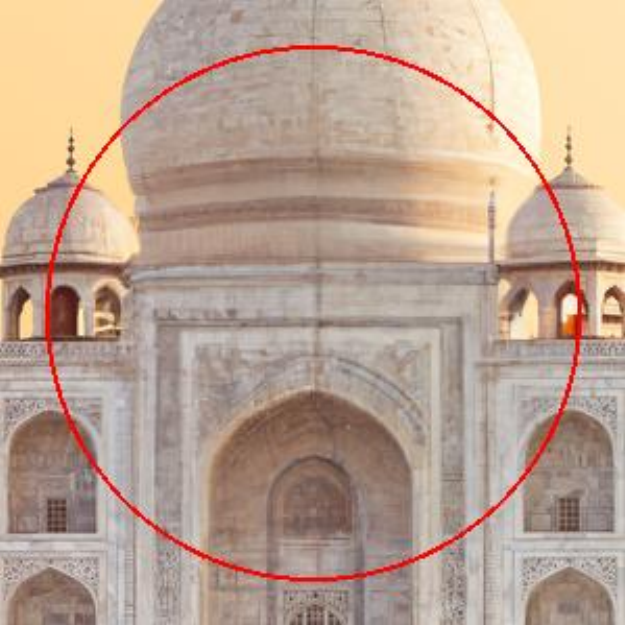}} & 
        \noindent\parbox[c]{0.081\textwidth}{\includegraphics[width=0.081\textwidth]{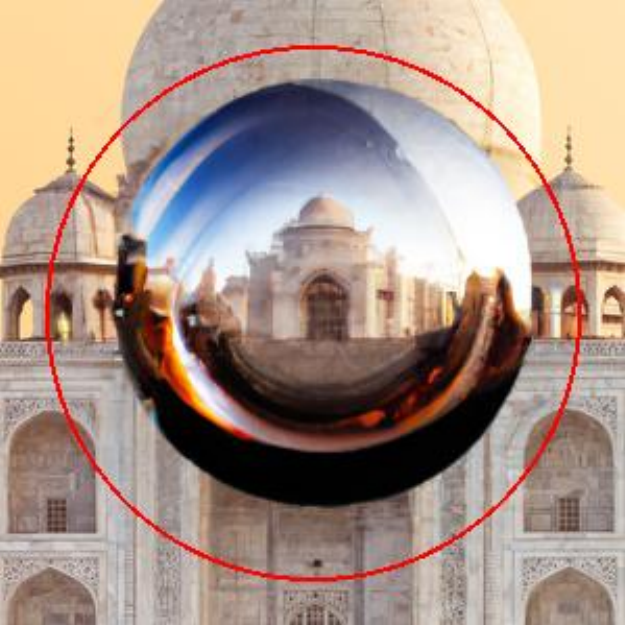}} & 
        \noindent\parbox[c]{0.081\textwidth}{\includegraphics[width=0.081\textwidth]{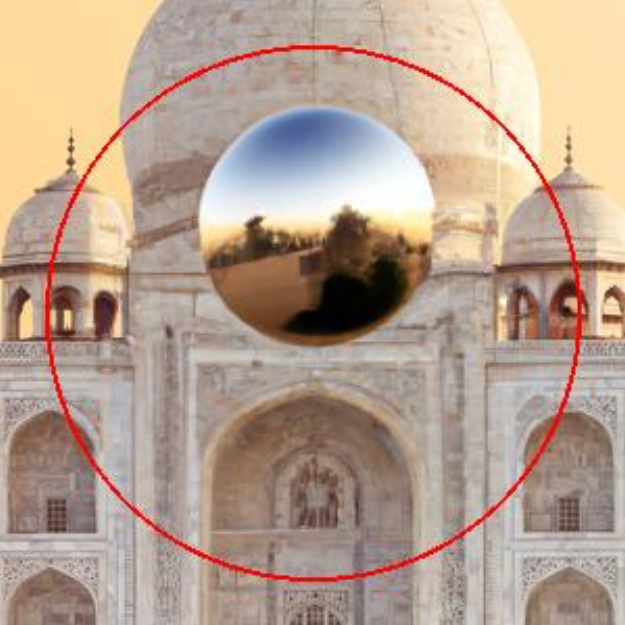}} & 
        \noindent\parbox[c]{0.081\textwidth}{\includegraphics[width=0.081\textwidth]{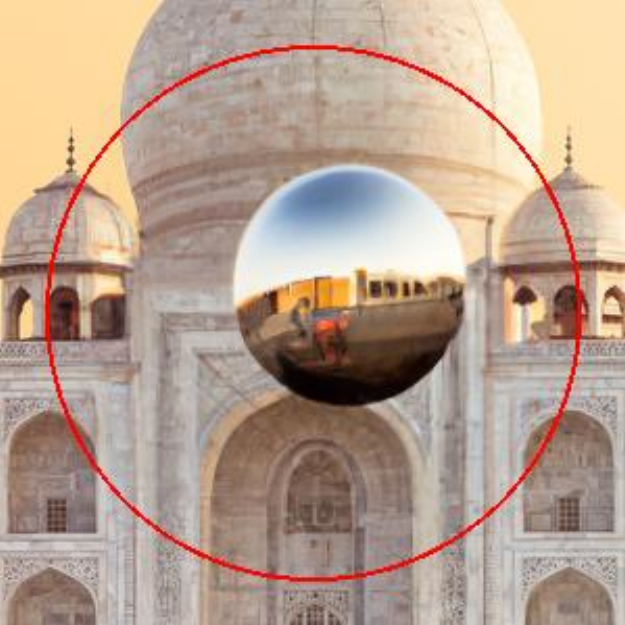}} & 
        \noindent\parbox[c]{0.081\textwidth}{\includegraphics[width=0.081\textwidth]{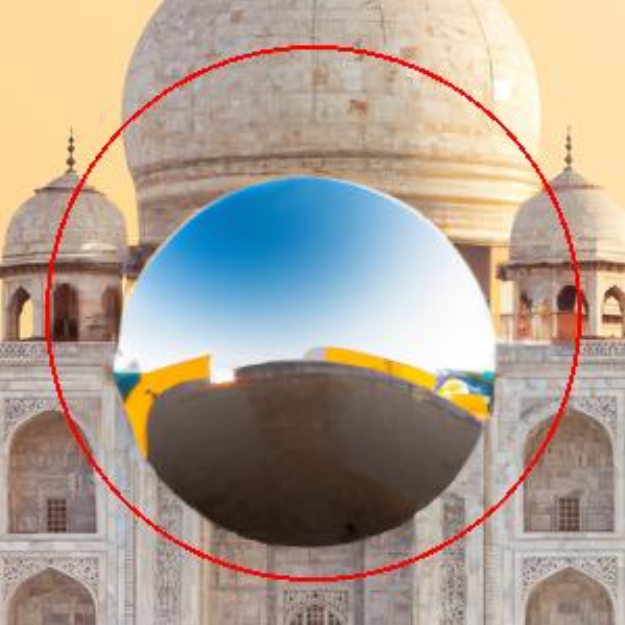}} & 
        \noindent\parbox[c]{0.081\textwidth}{\includegraphics[width=0.081\textwidth]{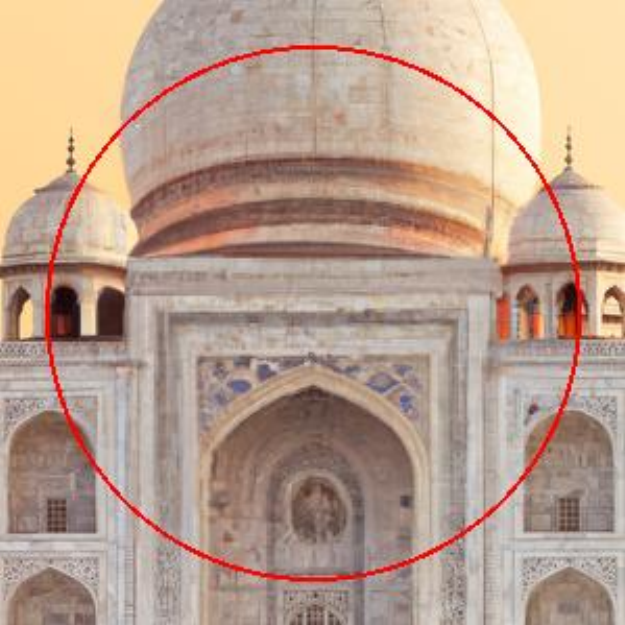}} & 
        \noindent\parbox[c]{0.081\textwidth}{\includegraphics[width=0.081\textwidth]{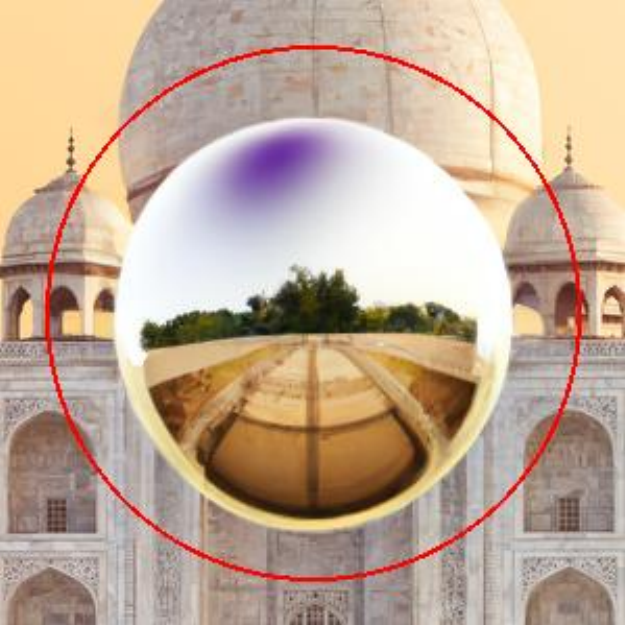}} & 
        
        \\

        \multicolumn{1}{l}{\rotatebox[origin=c]{90}{\shortstack[l]{\scriptsize Adobe \\ \scriptsize Firefly \cite{adobefirefly}}}} &
        \noindent\parbox[c]{0.081\textwidth}{\includegraphics[width=0.081\textwidth]{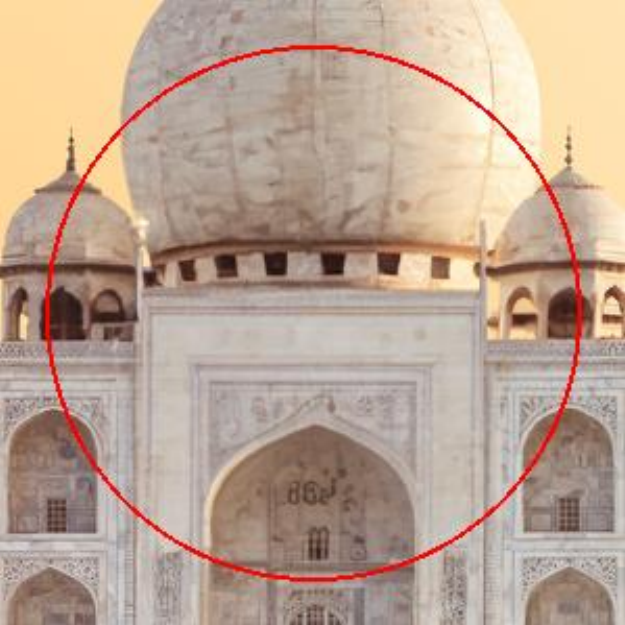}} & 
        \noindent\parbox[c]{0.081\textwidth}{\includegraphics[width=0.081\textwidth]{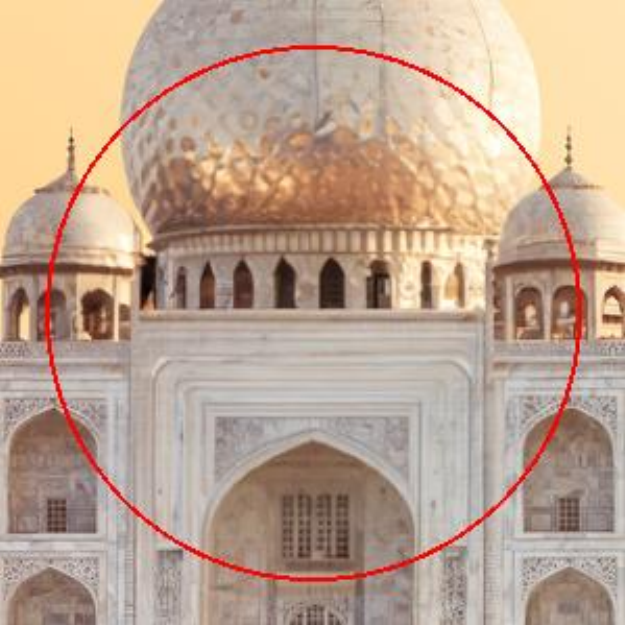}} &  
        \noindent\parbox[c]{0.081\textwidth}{\includegraphics[width=0.081\textwidth]{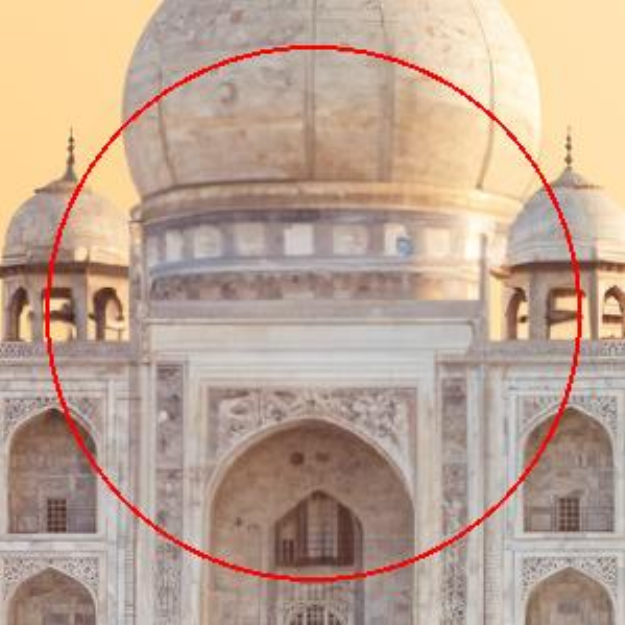}} & 
        \noindent\parbox[c]{0.081\textwidth}{\includegraphics[width=0.081\textwidth]{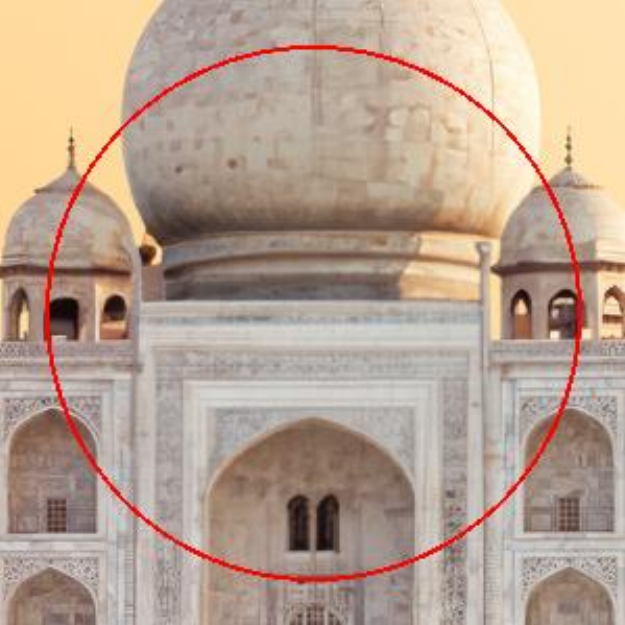}} & 
        \noindent\parbox[c]{0.081\textwidth}{\includegraphics[width=0.081\textwidth]{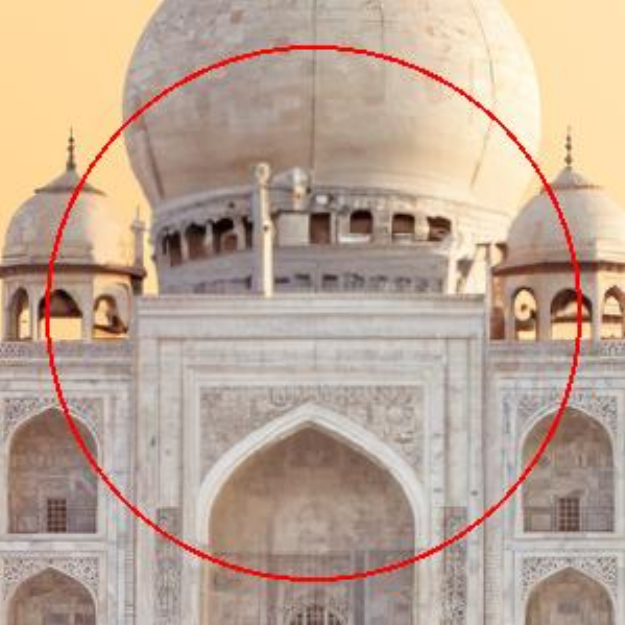}} & 
        \noindent\parbox[c]{0.081\textwidth}{\includegraphics[width=0.081\textwidth]{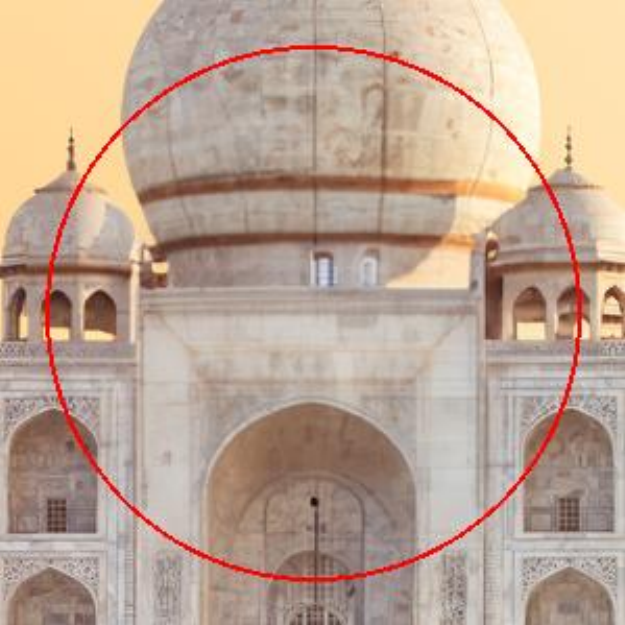}} & 
        \noindent\parbox[c]{0.081\textwidth}{\includegraphics[width=0.081\textwidth]{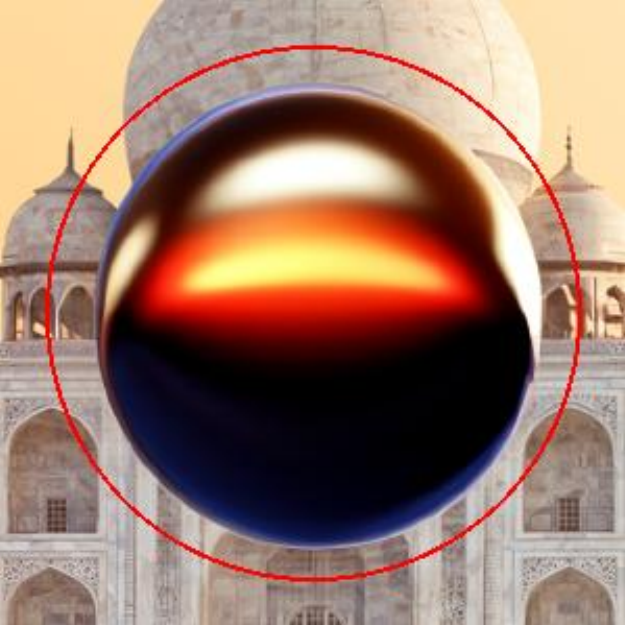}} & 
        \noindent\parbox[c]{0.081\textwidth}{\includegraphics[width=0.081\textwidth]{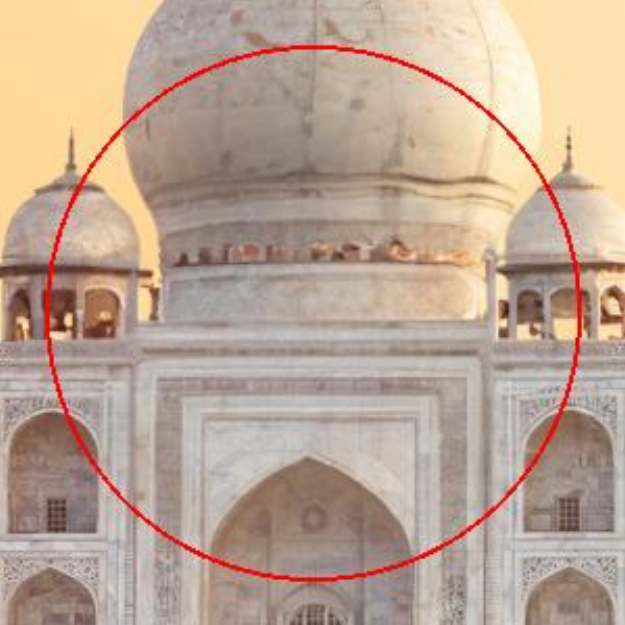}} & 
        \noindent\parbox[c]{0.081\textwidth}{\includegraphics[width=0.081\textwidth]{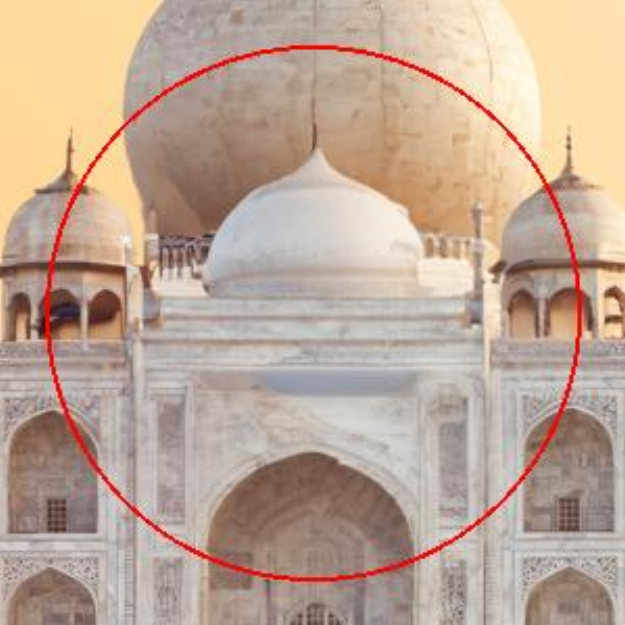}} & 
        \noindent\parbox[c]{0.081\textwidth}{\includegraphics[width=0.081\textwidth]{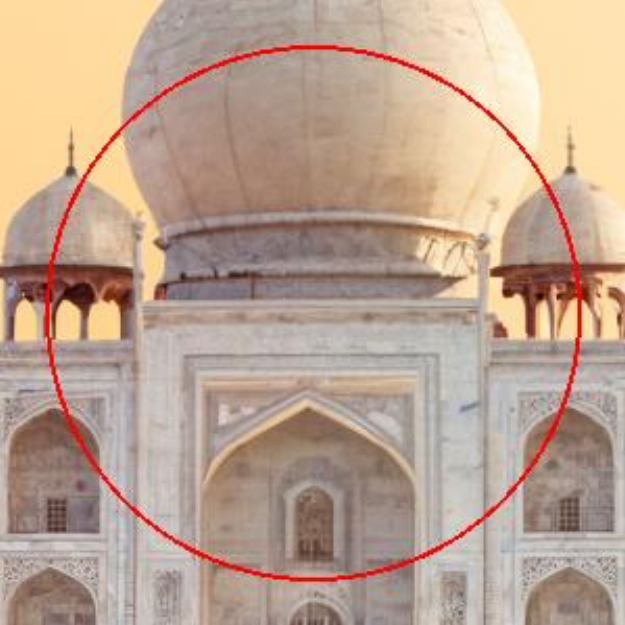}} & 

        \\

        \multicolumn{1}{l}{\rotatebox[origin=c]{90}{\shortstack[l]{\scriptsize SDXL \cite{podell2023sdxl}}}} &
        \noindent\parbox[c]{0.081\textwidth}{\includegraphics[width=0.081\textwidth]{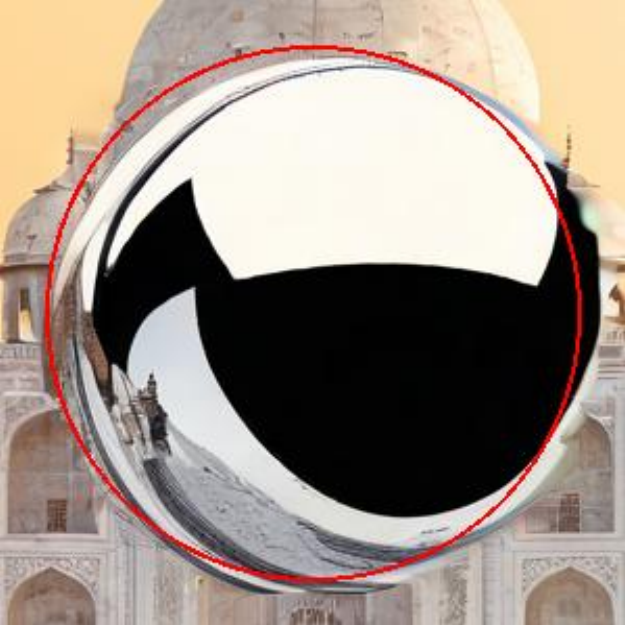}} & 
        \noindent\parbox[c]{0.081\textwidth}{\includegraphics[width=0.081\textwidth]{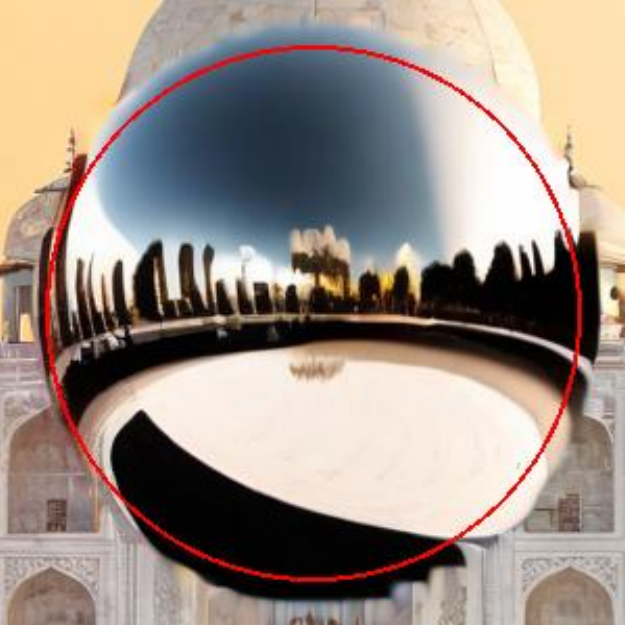}} &  
        \noindent\parbox[c]{0.081\textwidth}{\includegraphics[width=0.081\textwidth]{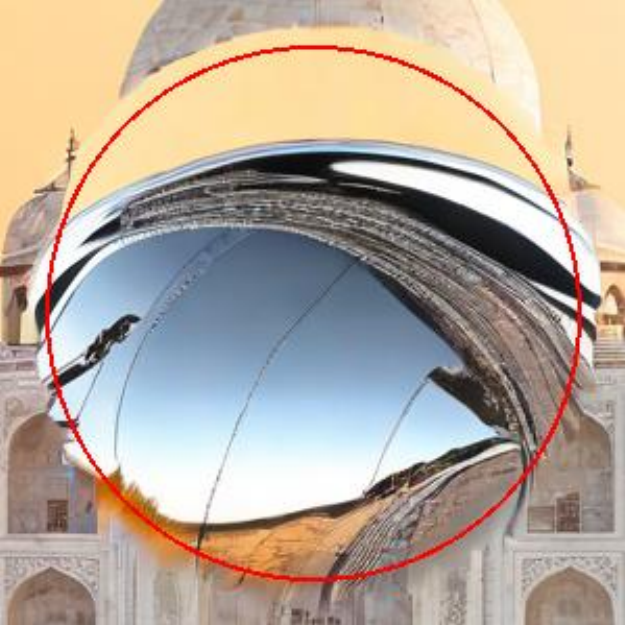}} & 
        \noindent\parbox[c]{0.081\textwidth}{\includegraphics[width=0.081\textwidth]{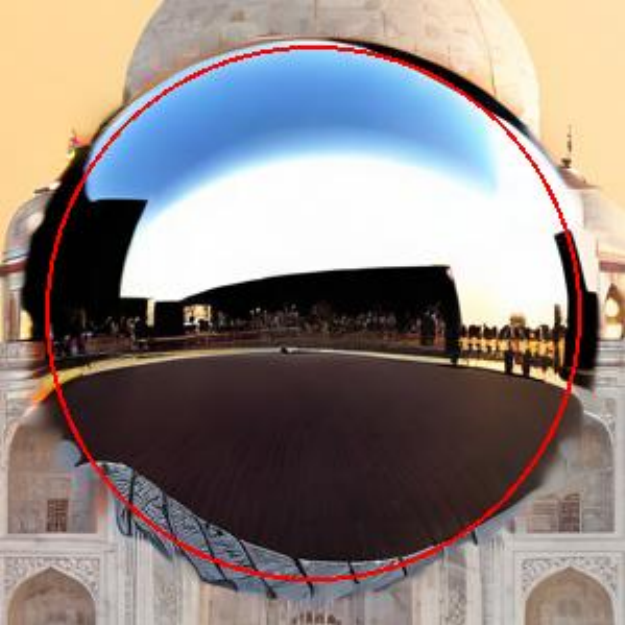}} & 
        \noindent\parbox[c]{0.081\textwidth}{\includegraphics[width=0.081\textwidth]{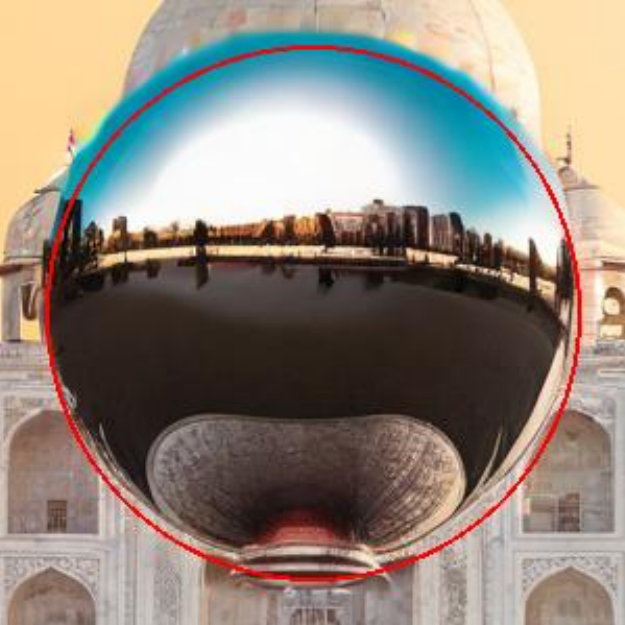}} & 
        \noindent\parbox[c]{0.081\textwidth}{\includegraphics[width=0.081\textwidth]{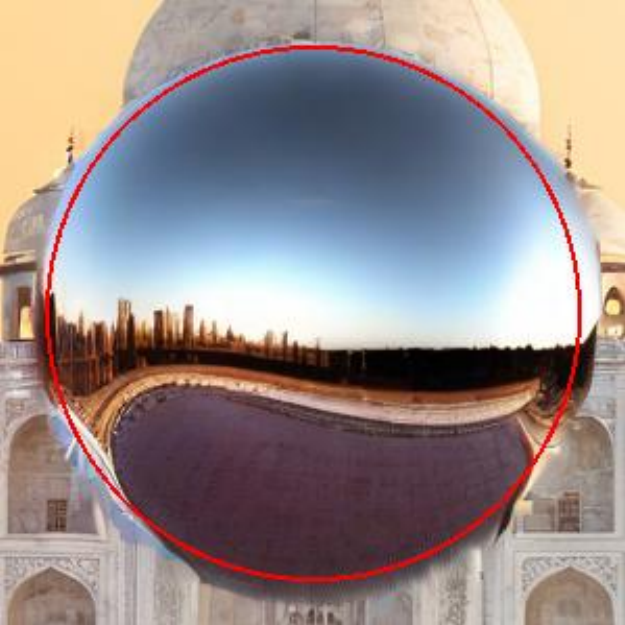}} & 
        \noindent\parbox[c]{0.081\textwidth}{\includegraphics[width=0.081\textwidth]{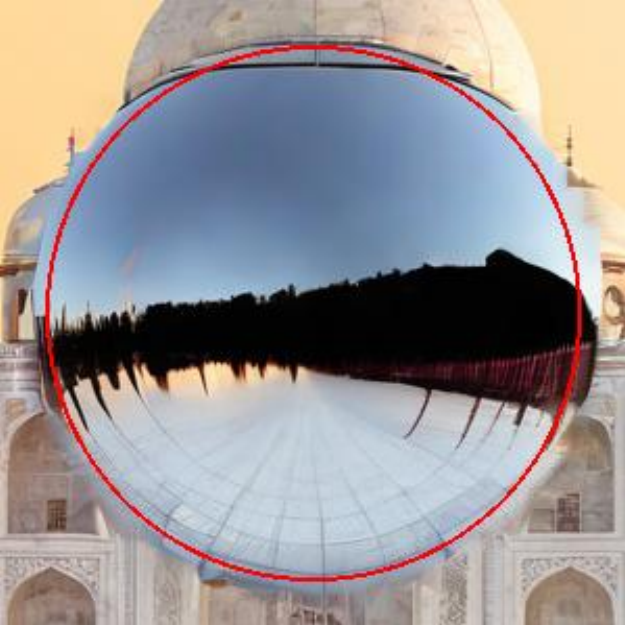}} & 
        \noindent\parbox[c]{0.081\textwidth}{\includegraphics[width=0.081\textwidth]{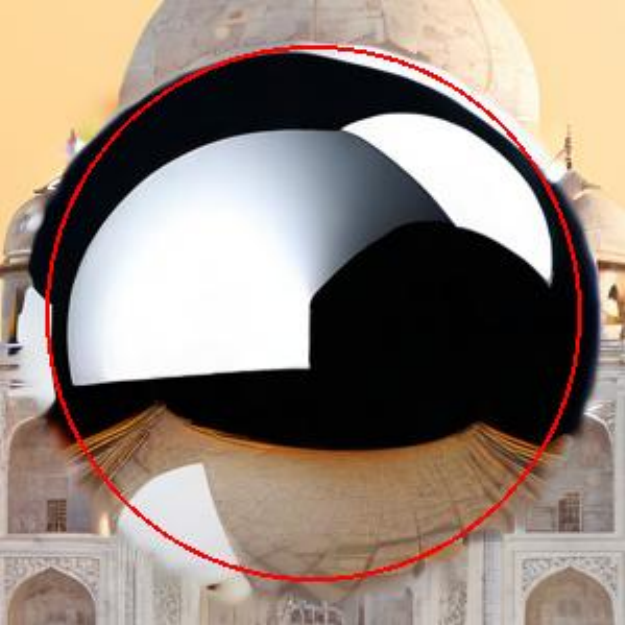}} & 
        \noindent\parbox[c]{0.081\textwidth}{\includegraphics[width=0.081\textwidth]{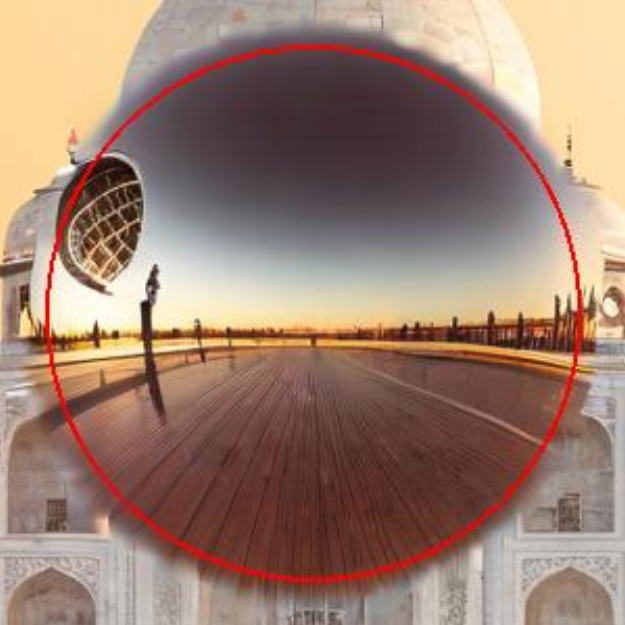}} & 
        \noindent\parbox[c]{0.081\textwidth}{\includegraphics[width=0.081\textwidth]{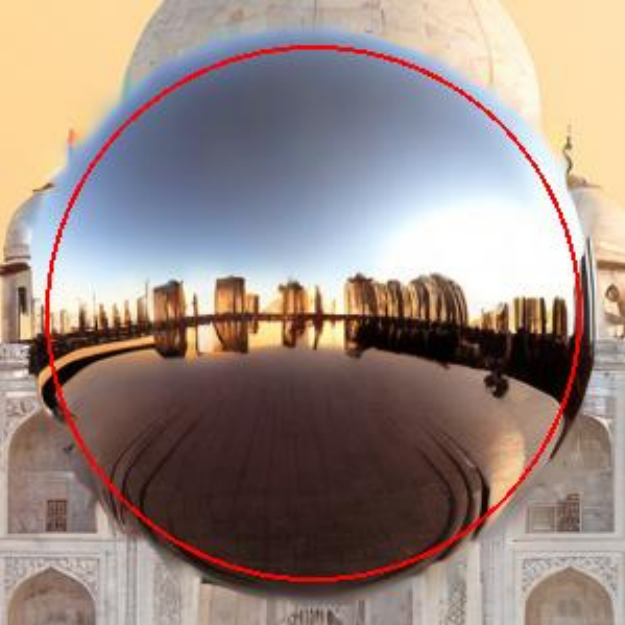}} & 
        
        \\ \hline

        \multicolumn{1}{l}{\rotatebox[origin=c]{90}{\shortstack[l]{\scriptsize \textbf{Ours}}}} &
        \noindent\parbox[c]{0.081\textwidth}{\includegraphics[width=0.081\textwidth]{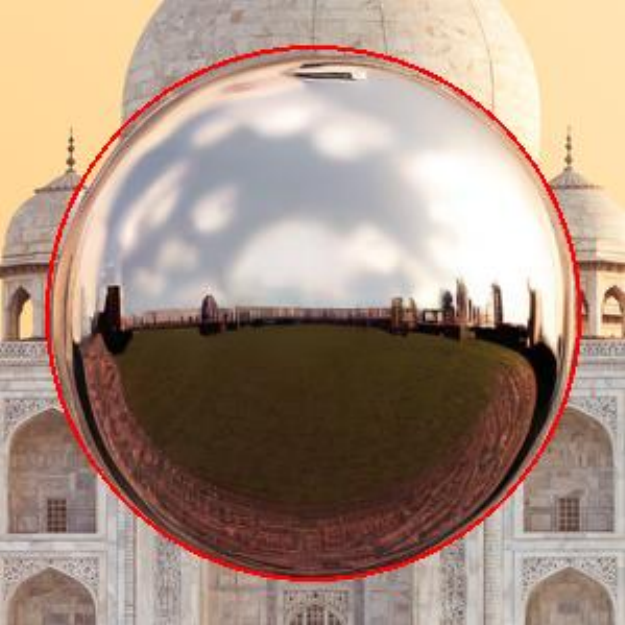}} & 
        \noindent\parbox[c]{0.081\textwidth}{\includegraphics[width=0.081\textwidth]{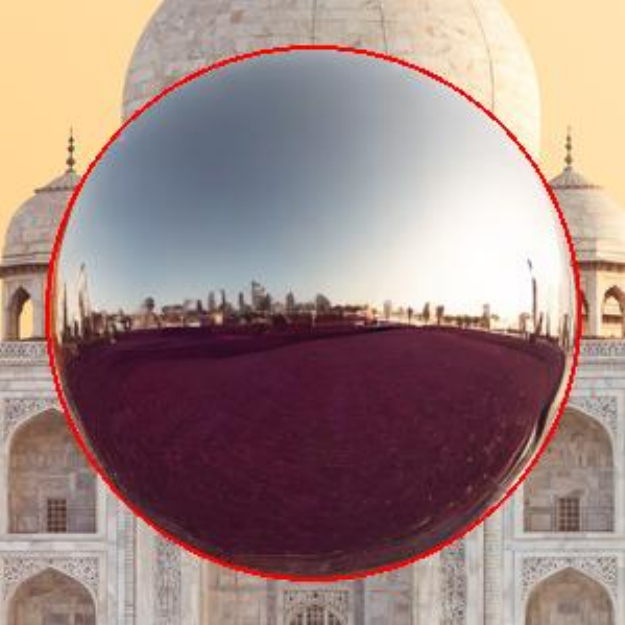}} &  
        \noindent\parbox[c]{0.081\textwidth}{\includegraphics[width=0.081\textwidth]{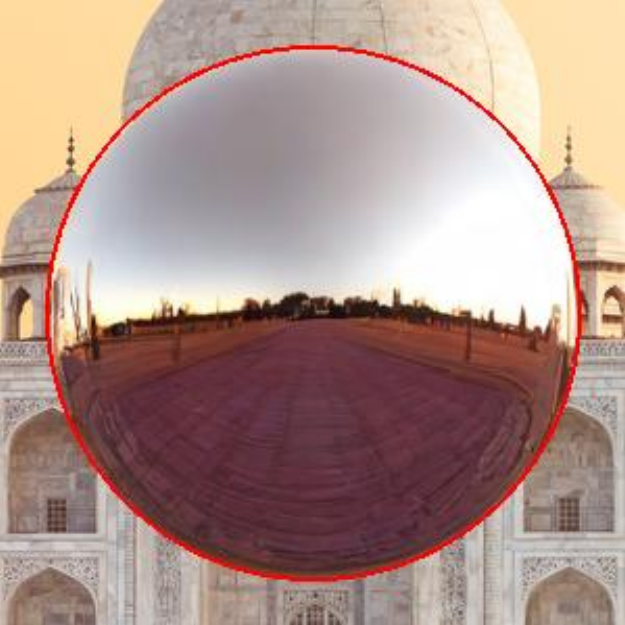}} & 
        \noindent\parbox[c]{0.081\textwidth}{\includegraphics[width=0.081\textwidth]{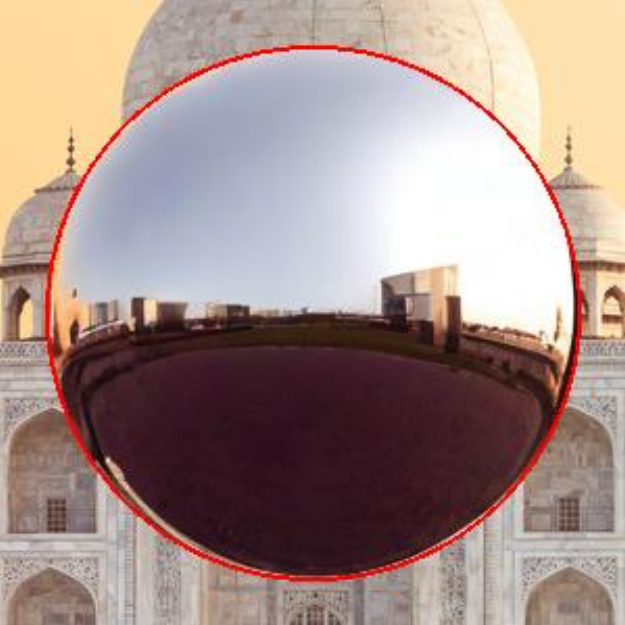}} & 
        \noindent\parbox[c]{0.081\textwidth}{\includegraphics[width=0.081\textwidth]{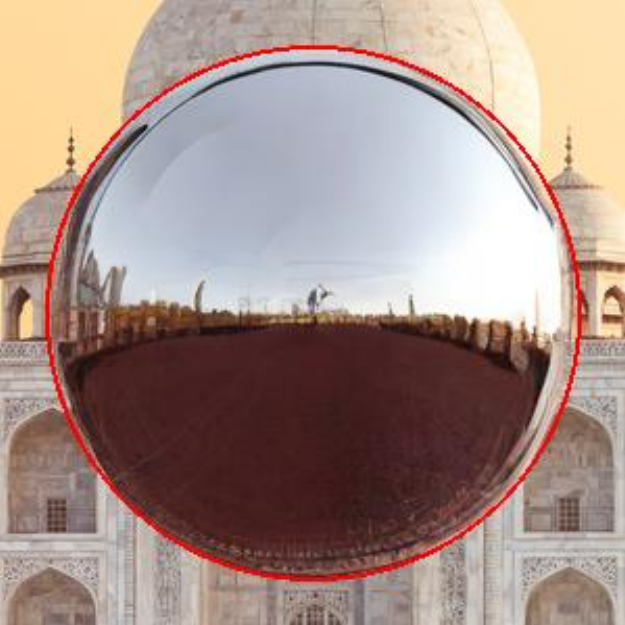}} & 
        \noindent\parbox[c]{0.081\textwidth}{\includegraphics[width=0.081\textwidth]{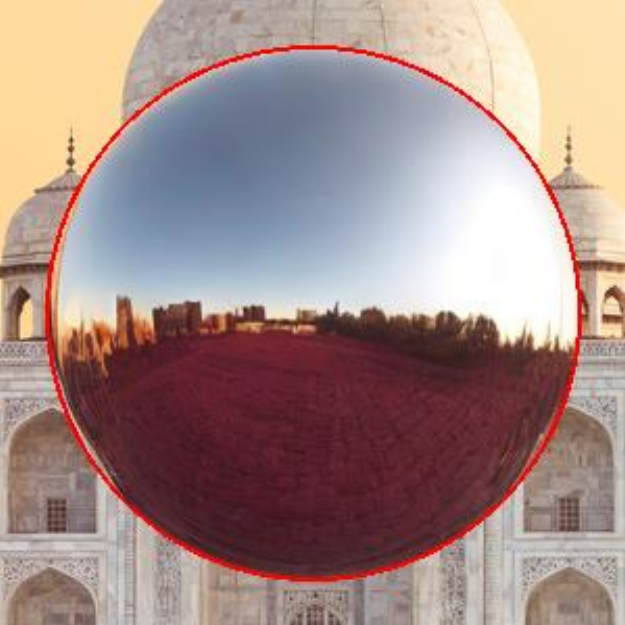}} & 
        \noindent\parbox[c]{0.081\textwidth}{\includegraphics[width=0.081\textwidth]{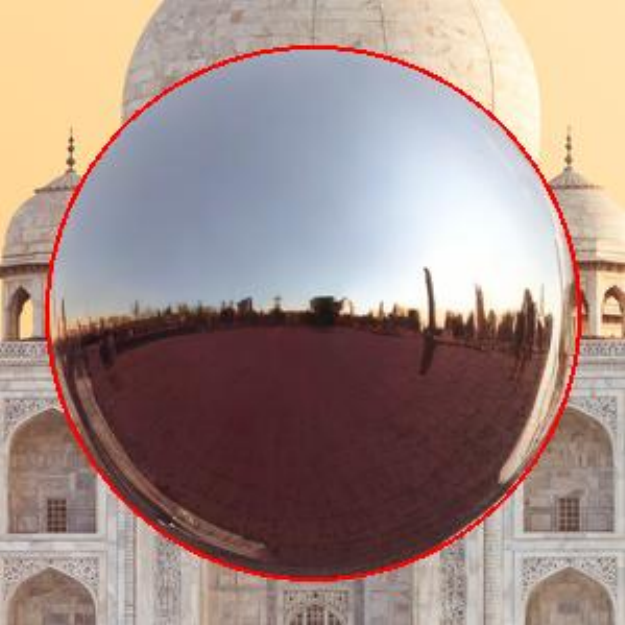}} & 
        \noindent\parbox[c]{0.081\textwidth}{\includegraphics[width=0.081\textwidth]{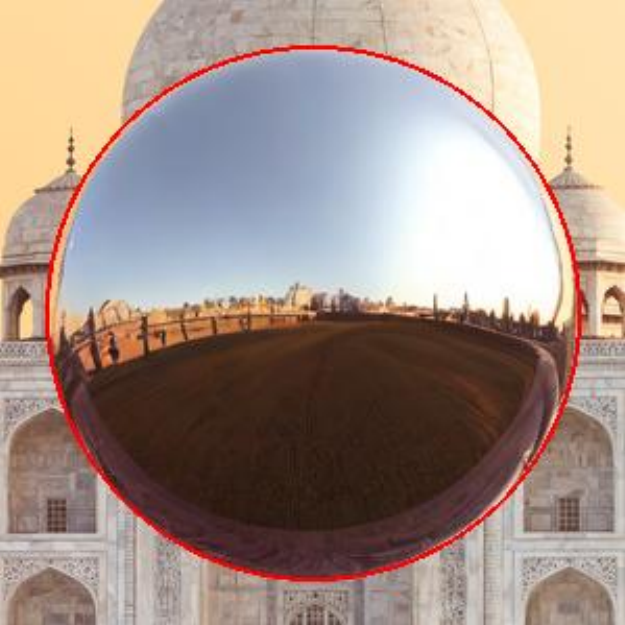}} & 
        \noindent\parbox[c]{0.081\textwidth}{\includegraphics[width=0.081\textwidth]{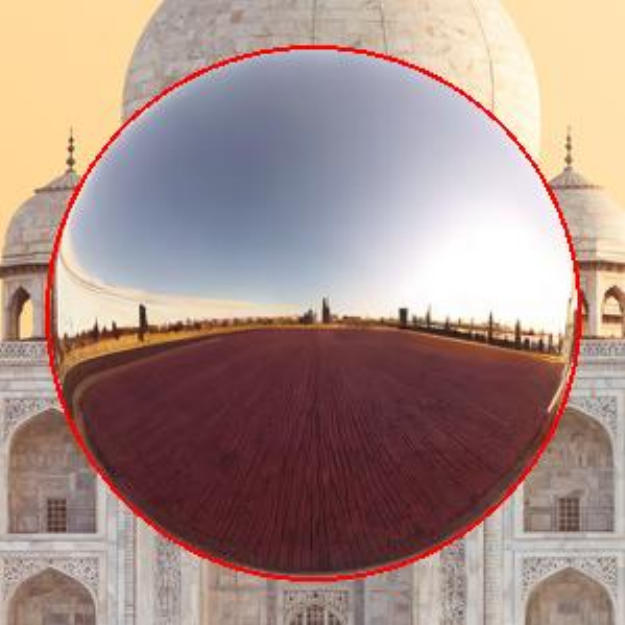}} & 
        \noindent\parbox[c]{0.081\textwidth}{\includegraphics[width=0.081\textwidth]{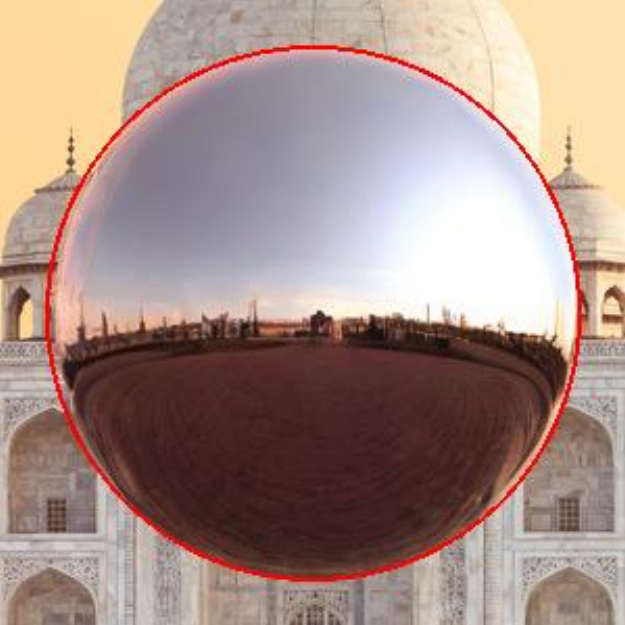}} & 
        
        \\
        
    \end{tabu}

    \caption{Chrome ball inpainting results from various methods. The
    red circle indicates the inpainted region, and we show a zoomed-in
    view of the blue crop. Each row contains results from ten different random seeds.}
    \label{fig:aba_seed-cherry2}
\end{figure*}

We show inpainting results of our method and other baselines using different random seeds in Figure \ref{fig:aba_seed-cherry} and Figure \ref{fig:aba_seed-cherry2}.

What we observed in general was that Blend Diffusion \cite{avrahami2022blendeddiffusion, avrahami2023blendedlatent} produced distorted balls. Paint-by-Example \cite{yang2023paint} failed to reproduce mirrored chrome balls altogether. IP-Adapter \cite{ye2023ip-adapter} replicated textures and details of the example chrome balls, making it unsuitable for light estimation. DALL·E2 \cite{dalle2} often simply reconstructed most of the masked-out content. Adobe Firefly \cite{adobefirefly} had a similar problem with DALL·E2 \cite{dalle2}, albeit more severe (see Figure \ref{fig:aba_seed-cherry2}). Moreover, none of these techniques precisely followed the inpainting mask. Our proposed method can address all these issues and consistently inpaint high-quality chrome balls.

\section{Additional Qualitative Results} \label{appendix:more_result_all}

\subsection{Benchmark datasets} \label{appendix:more_result_benchmark}

This section provides qualitative results for the experiments in Section~\ref{sec:experiment} in the main paper.

\subsubsection{Evaluation on three spheres.} We show spheres with three material types---mirror, matte, and diffuse---rendered using the inferred environment maps from the following methods: 
\begin{enumerate}
    \item Ground truth
    \item StyleLight \cite{wang2022stylelight}
    \item Stable Diffusion XL \cite{podell2023sdxl} with depth-conditioned ControlNet \cite{zhang2023adding} (SDXL$^\dagger$)
    \item DiffusionLight:
    \begin{enumerate}
        \item Without our Exposure LoRA (ablated)
        \item Without our iterative inpainting (ablated)
        \item Full method (ours)
    \end{enumerate}
    \item DiffusionLight-Turbo:
    \begin{enumerate}
        \item Without our LoRA swapping technique (ablated)
        \item Full method (ours)
    \end{enumerate}
\end{enumerate}

\begin{figure*}
    \centering
    \includegraphics[width=1.0\textwidth, trim=0 7.5cm 0 7.5cm, clip]{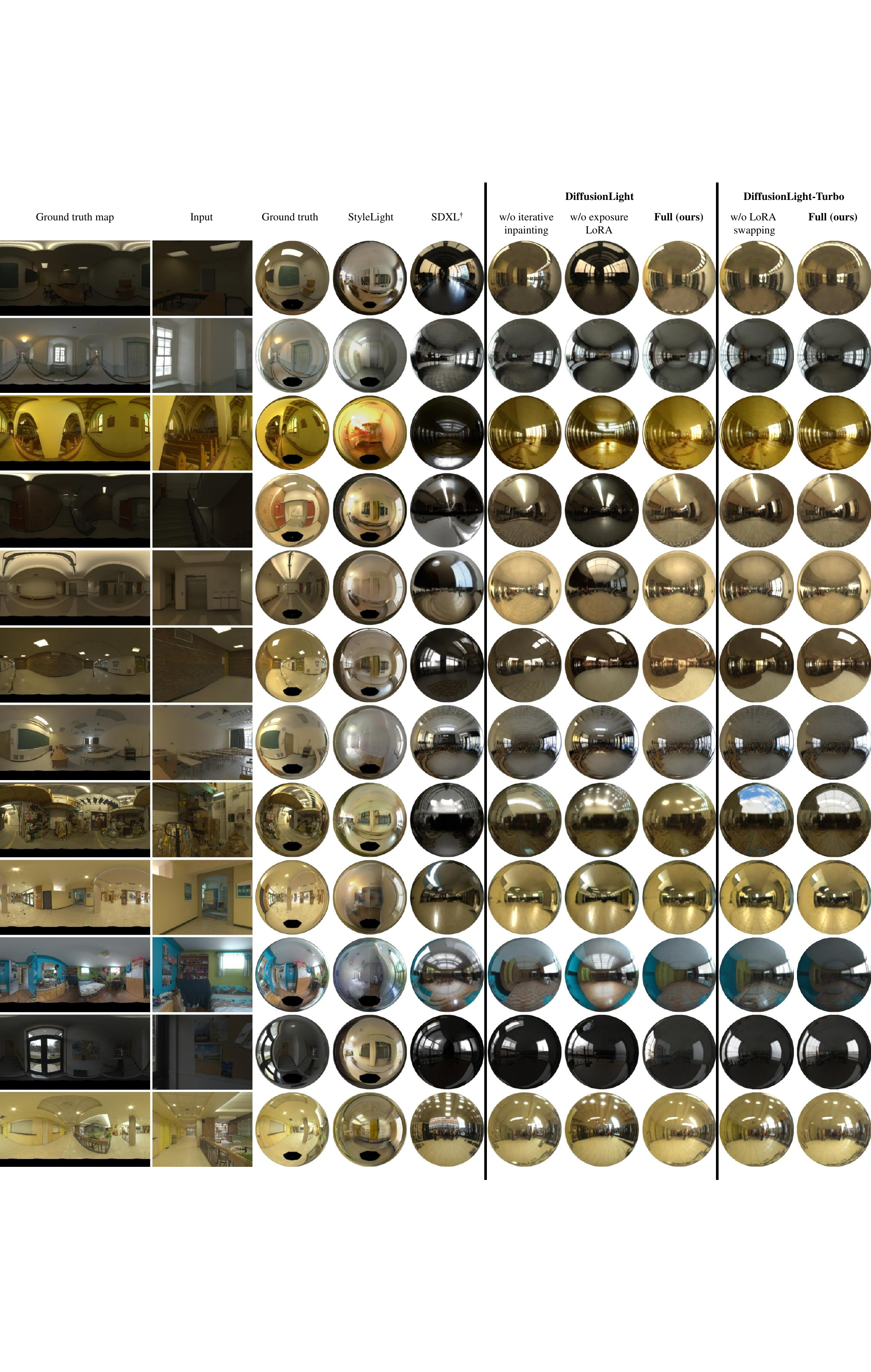}
    \caption{Qualitative results on the Laval Indoor dataset using mirror balls. Our DiffusionLight-Turbo generates plausible chrome balls with visual quality comparable to DiffusionLight, while requiring significantly fewer computational resources.}
    \label{fig:additional_indoor_mirror}
\end{figure*}

\begin{figure*}
    \centering
    \includegraphics[width=1.0\textwidth, trim=0 7.5cm 0 7.5cm, clip]{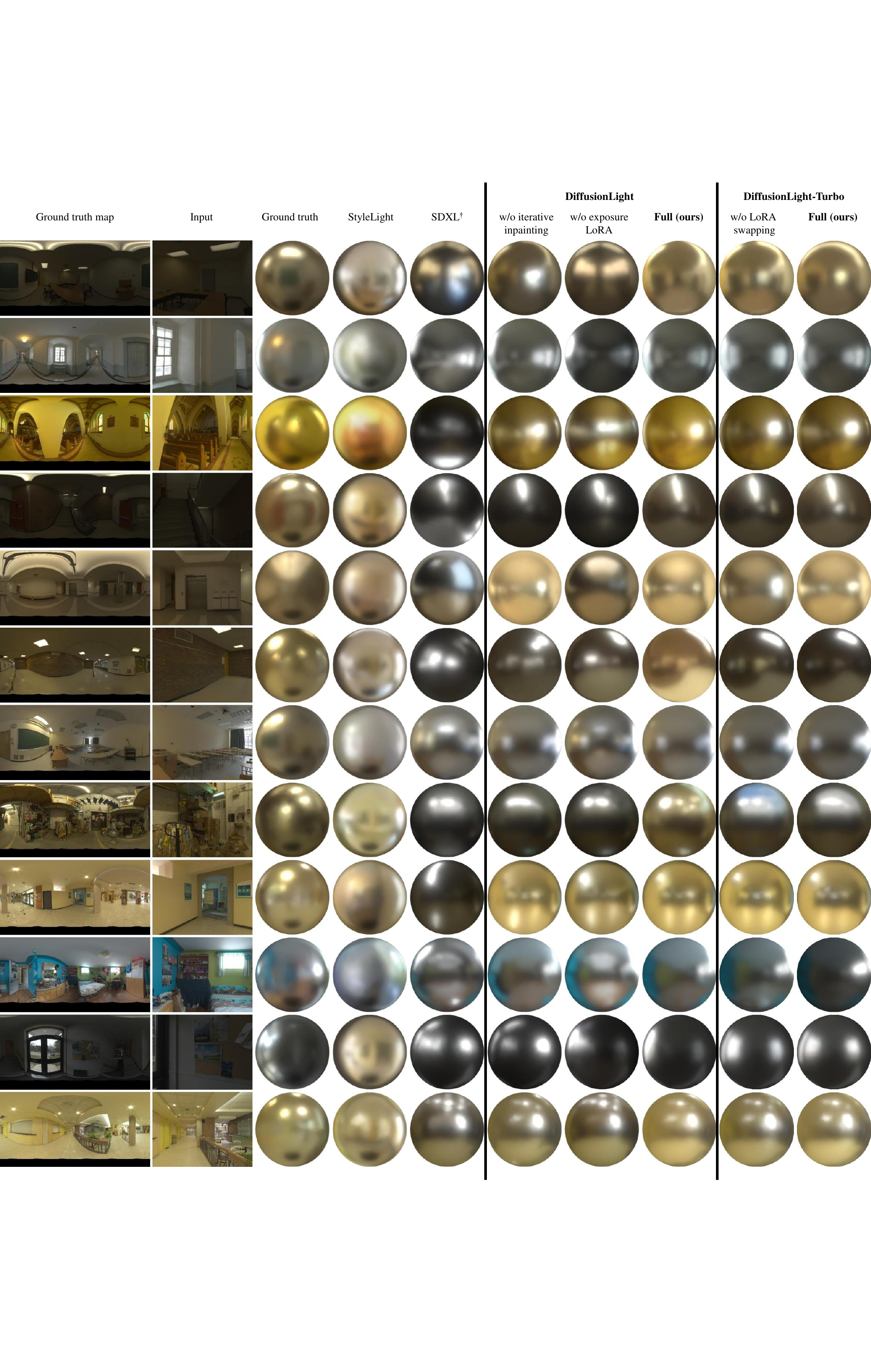}
    \caption{Qualitative results on the Laval Indoor dataset using matte silver balls. Our DiffusionLight-Turbo generates plausible chrome balls with visual quality comparable to DiffusionLight, while requiring significantly fewer computational resources.}
    \label{fig:additional_indoor_matte}
\end{figure*}

\begin{figure*}
    \centering
    \includegraphics[width=1.0\textwidth, trim=0 7.5cm 0 7.5cm, clip]{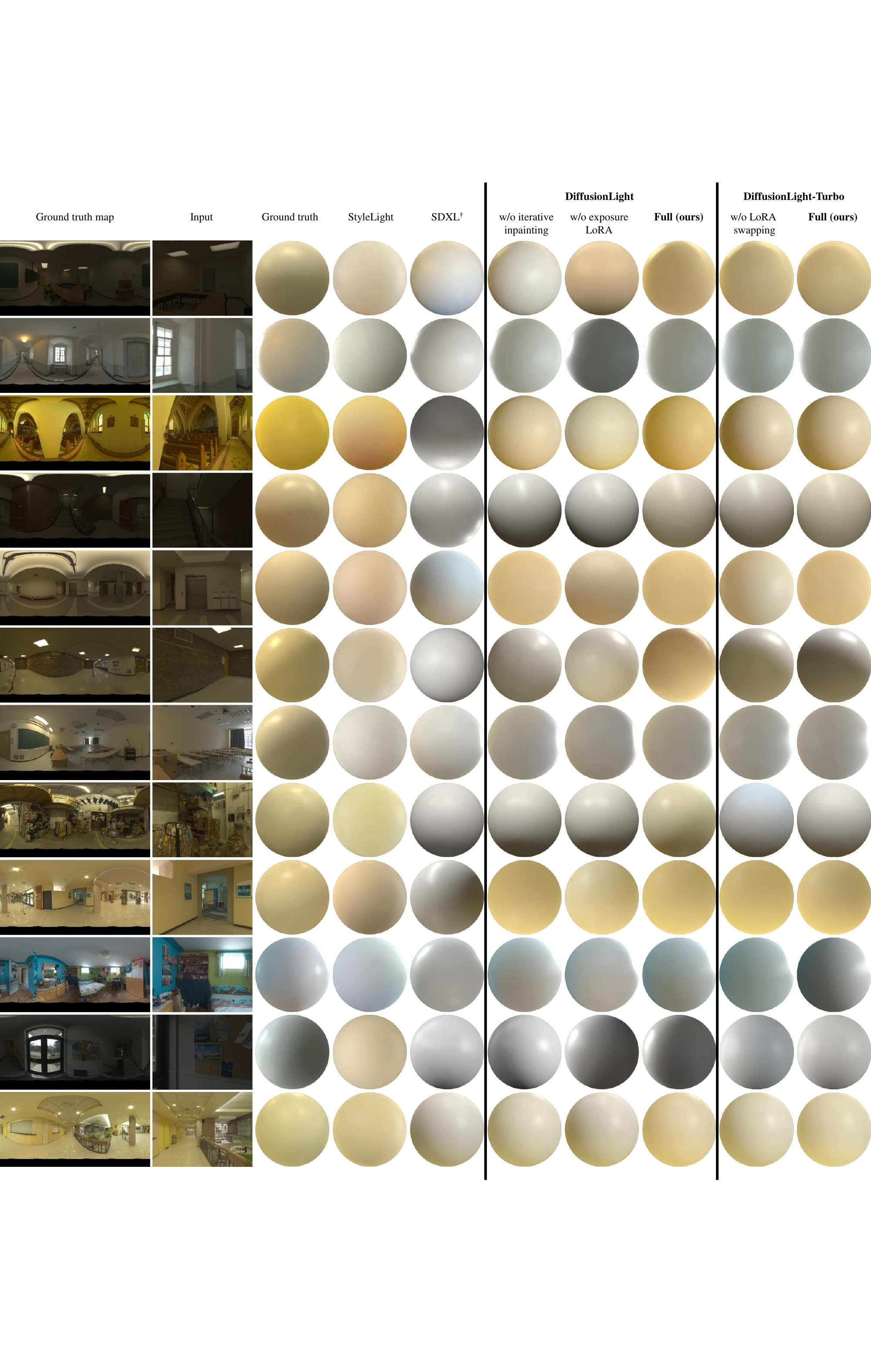}
    \caption{Qualitative results on the Laval Indoor dataset using diffuse balls. Our DiffusionLight-Turbo generates plausible chrome balls with visual quality comparable to DiffusionLight, while requiring significantly fewer computational resources.}
    \label{fig:additional_indoor_diffuse}
\end{figure*}

\begin{figure*}
    \centering
    \includegraphics[width=0.95\textwidth, trim=0 3.5cm 0 5cm, clip]{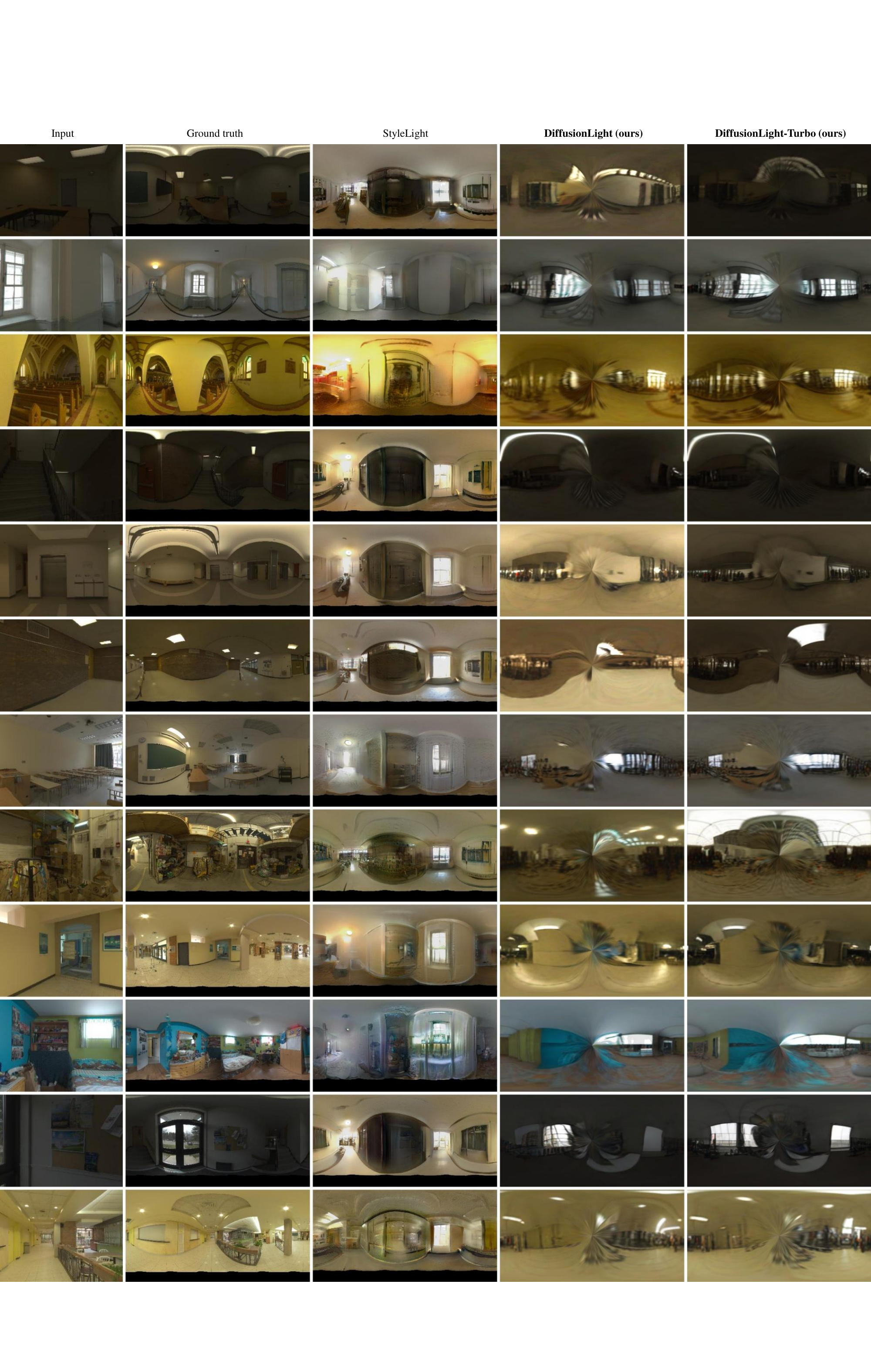}
    \caption{Unwarped equirectangular maps for the Laval Indoor dataset.}
    \label{fig:additional_indoor_envmap}
\end{figure*}

\begin{figure*}
    \centering
    \includegraphics[width=1.0\textwidth, trim=0 7.5cm 0 7.5cm, clip]{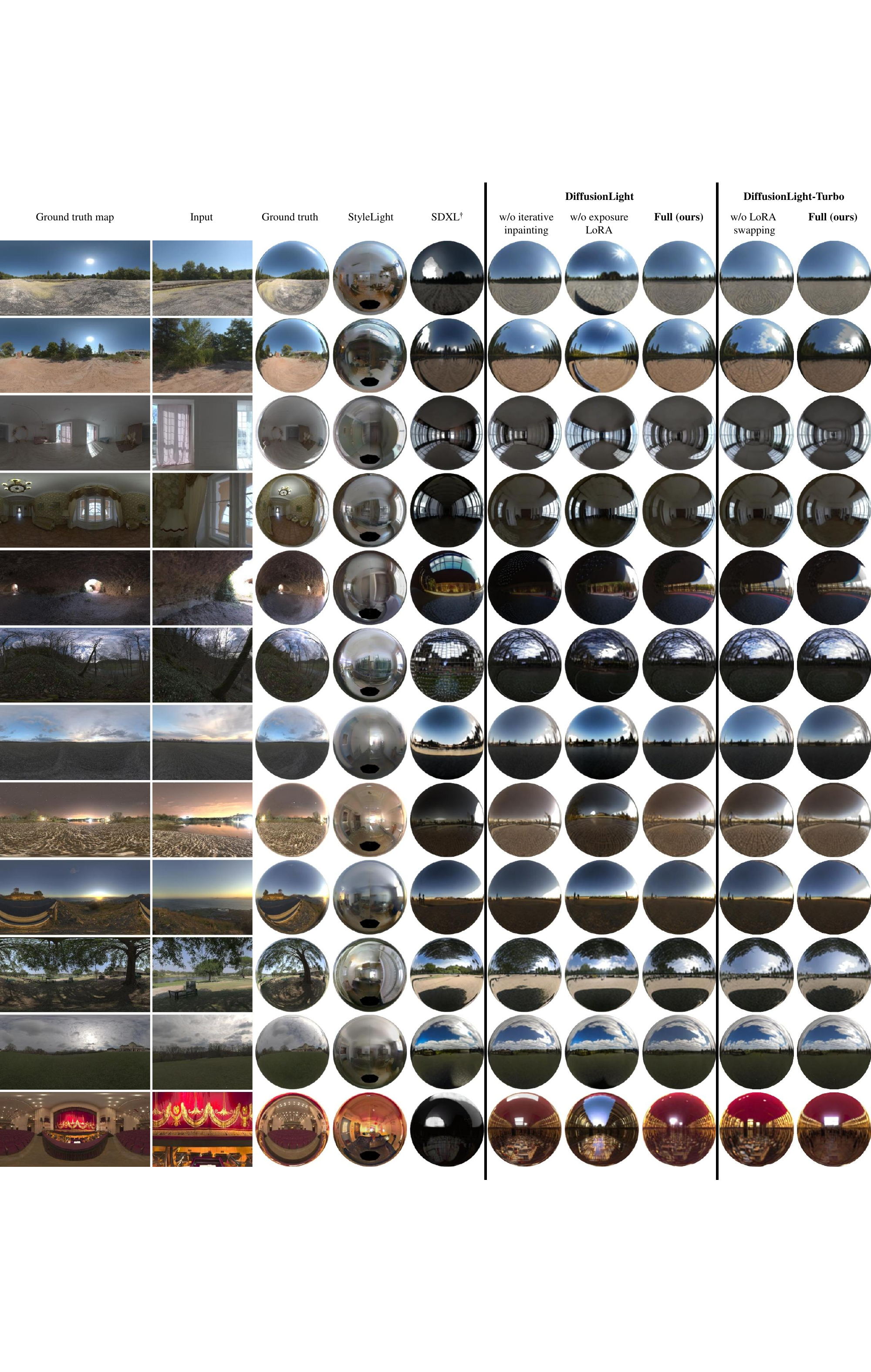}
    \caption{Qualitative results on the Poly Haven dataset using mirror balls. Our DiffusionLight-Turbo generates plausible chrome balls with visual quality comparable to DiffusionLight, while requiring significantly fewer computational resources.}
    \label{fig:additional_polyhaven_mirror}
\end{figure*}

\begin{figure*}
    \centering
    \includegraphics[width=1.0\textwidth, trim=0 7.5cm 0 7.5cm, clip]{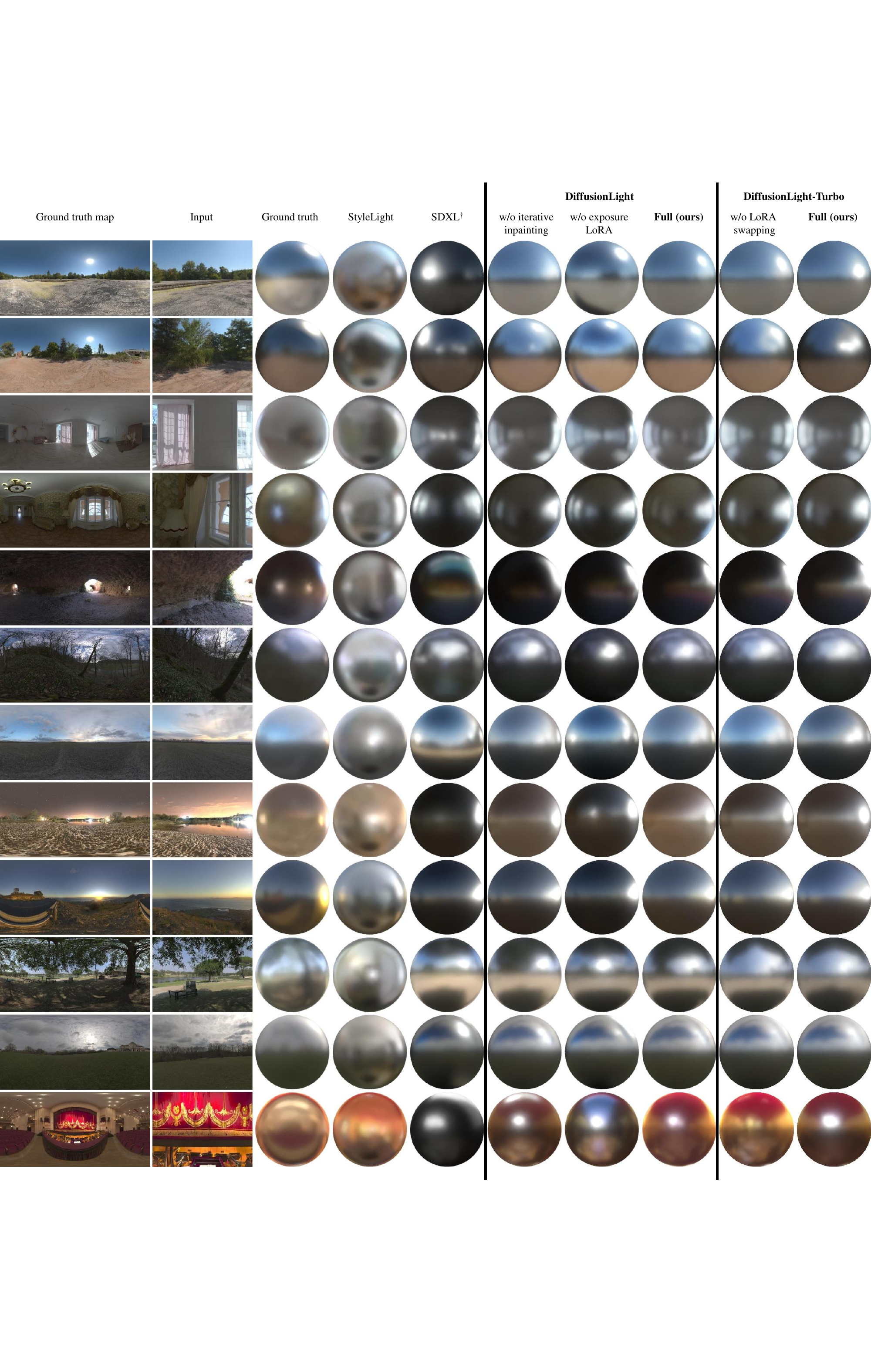}
    \caption{Qualitative results on the Poly Haven dataset using matte silver balls. Our DiffusionLight-Turbo generates plausible chrome balls with visual quality comparable to DiffusionLight, while requiring significantly fewer computational resources.}
    \label{fig:additional_polyhaven_matte}
\end{figure*}

\begin{figure*}
    \centering
    \includegraphics[width=1.0\textwidth, trim=0 7.5cm 0 7.5cm, clip]{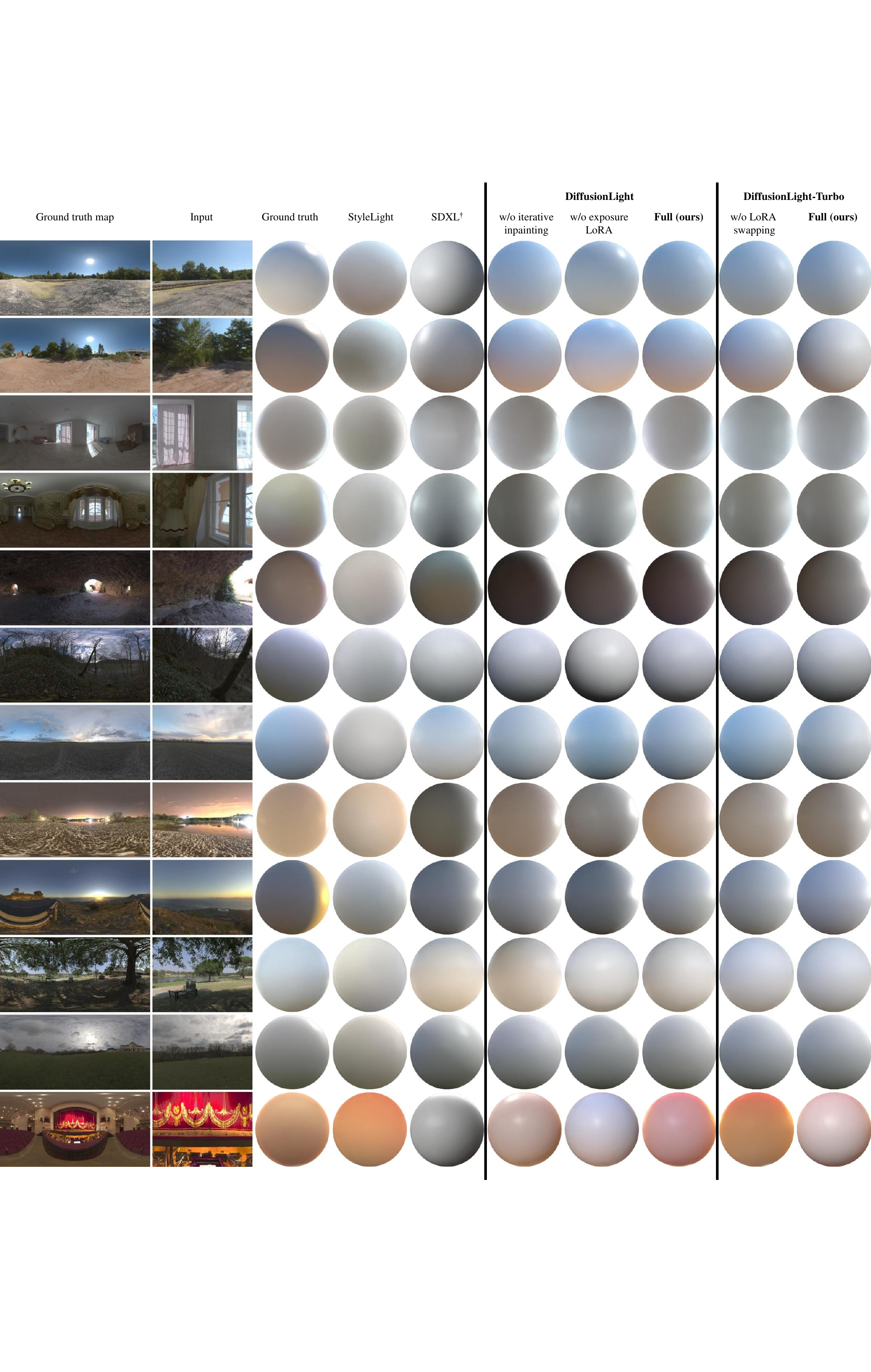}
    \caption{Qualitative results on the Poly Haven dataset using diffuse balls. Our DiffusionLight-Turbo generates plausible chrome balls with visual quality comparable to DiffusionLight, while requiring significantly fewer computational resources.}
    \label{fig:additional_polyhaven_diffuse}
\end{figure*}

\begin{figure*}
    \centering
    \includegraphics[width=0.95\textwidth, trim=0 3.5cm 0 5cm, clip]{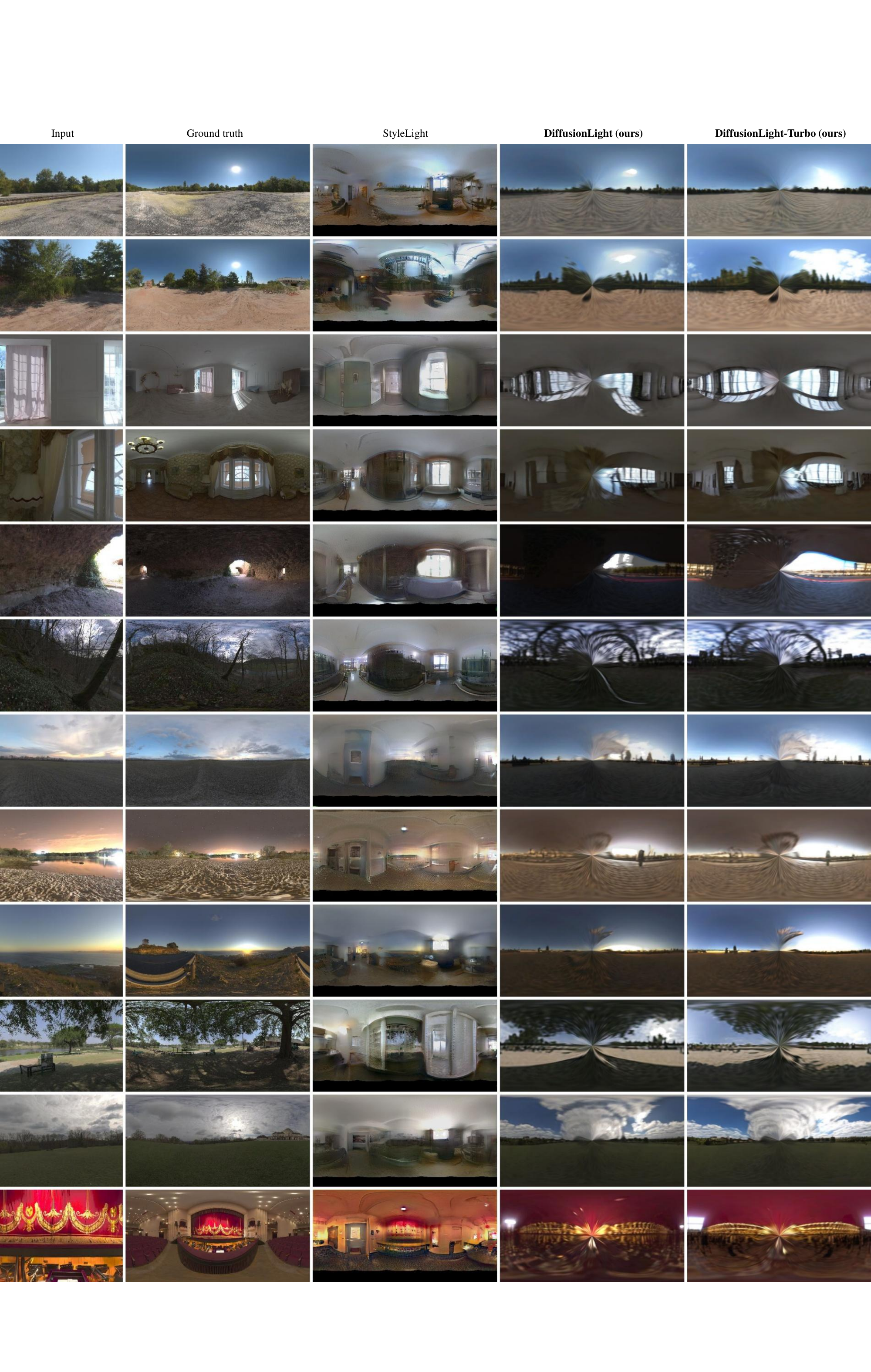}
    \caption{Unwarped equirectangular maps for the Poly Haven dataset.}
    \label{fig:additional_polyhaven_envmap}
\end{figure*}

Qualitative results for the Laval Indoor dataset are in Figures \ref{fig:additional_indoor_mirror}-\ref{fig:additional_indoor_envmap}, and for Poly Haven in Figures \ref{fig:additional_polyhaven_mirror}-\ref{fig:additional_polyhaven_envmap}. 
As discussed in the main paper, we start with SDXL$^\dagger$ and incrementally add our Exposure LoRA and iterative inpainting to obtain our DiffusionLight. 
Then, from DiffusionLight, we replace iterative inpainting with our Turbo LoRA and apply the LoRA swapping technique during inference to obtain DiffusionLight-Turbo.

\subsubsection{Evaluation on an array of spheres.} 
\begin{figure*}
    \centering
    \includegraphics[width=1.0\textwidth, trim=0 13cm 0 5.7cm, clip]{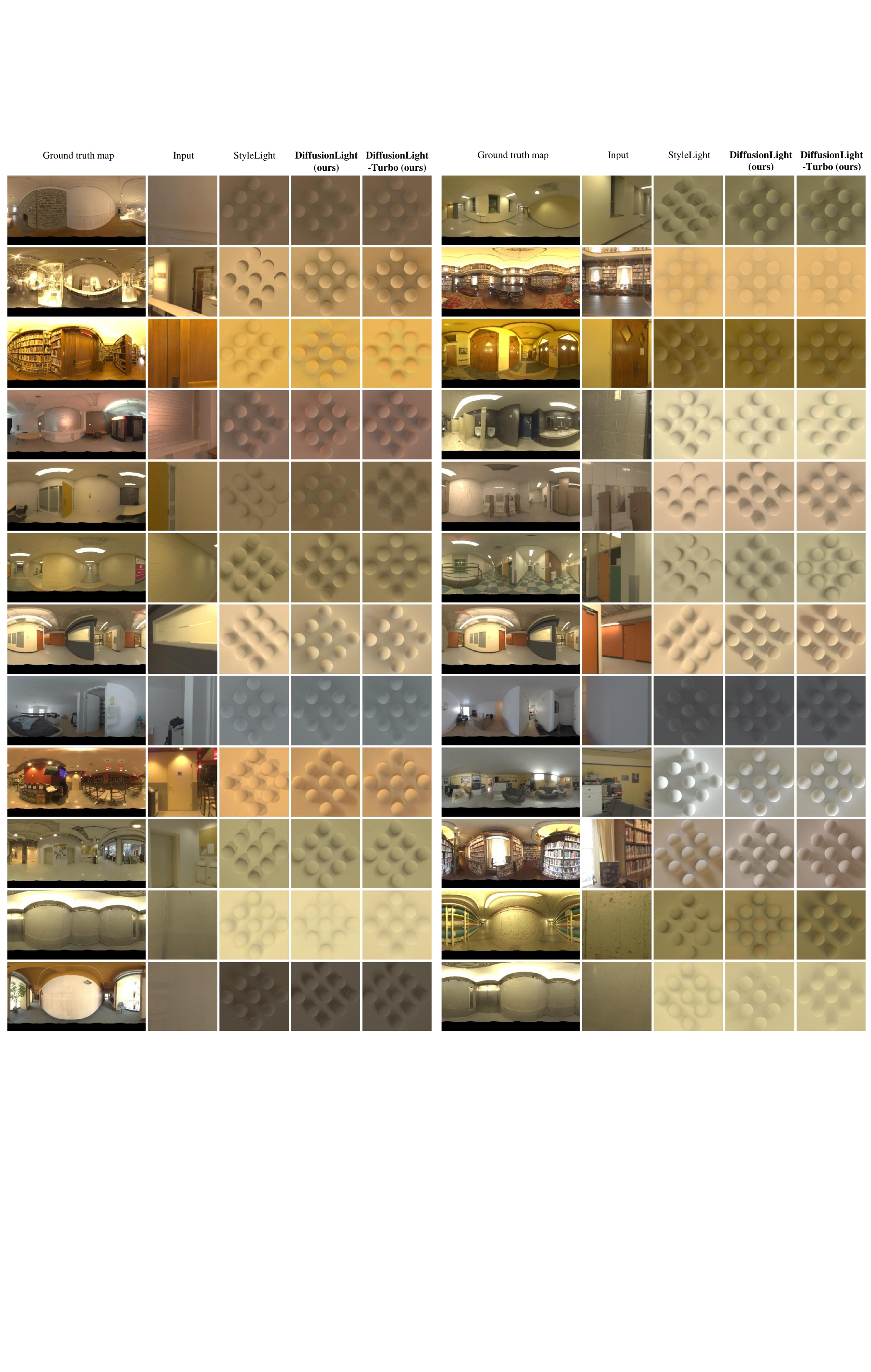}
    \caption{Qualitative results for the Laval indoor dataset using an array of spheres protocol.}
    \label{fig:additional_everlight}
\end{figure*}

We show renderings of an array of spheres on a plane using our estimated lighting, following the evaluation protocol from Weber et al.  \cite{weber2022editableindoor}. We display 24 random test images from a total of 2,240 test images of the Laval Indoor dataset in Figure \ref{fig:additional_everlight}. Because the scale of the HDR images our methods generate and that of the test set are different, for visualization purposes, we scale each output image so that its 0.1st and 99.9th percentiles of pixel intensity match those of the ground truth. Note that this scaling does not affect the quantitative scores reported in the main paper since the metrics are already scale-invariant.
The last row shows challenging test scenes featuring only plain, solid backgrounds without any shaded objects. Estimating lighting from such input images is highly ill-posed and multi-modal. As a result, visual assessment or evaluations using pixel-based metrics, as used in this protocol, may not be meaningful for such cases.

\subsection{In-the-wild images} \label{appendix:more_result_wild}
\begin{figure*}
    \centering
    \renewcommand{\arraystretch}{0.5} 
    \begin{tabular}{c}
        \includegraphics[width=1.0\textwidth]{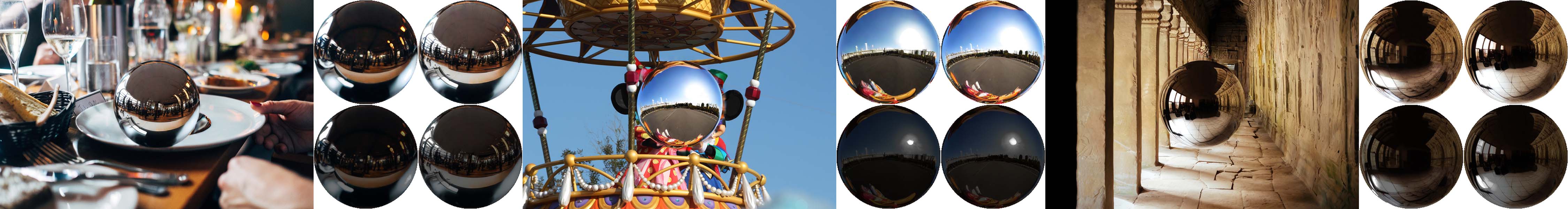} \\
        \includegraphics[width=1.0\textwidth]{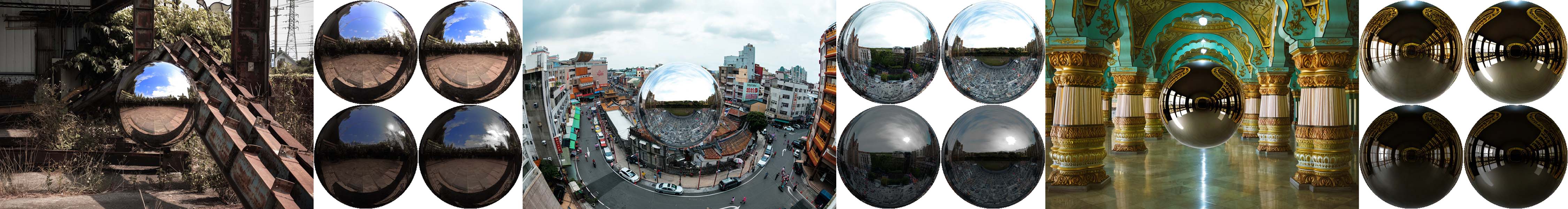} \\
        \includegraphics[width=1.0\textwidth]{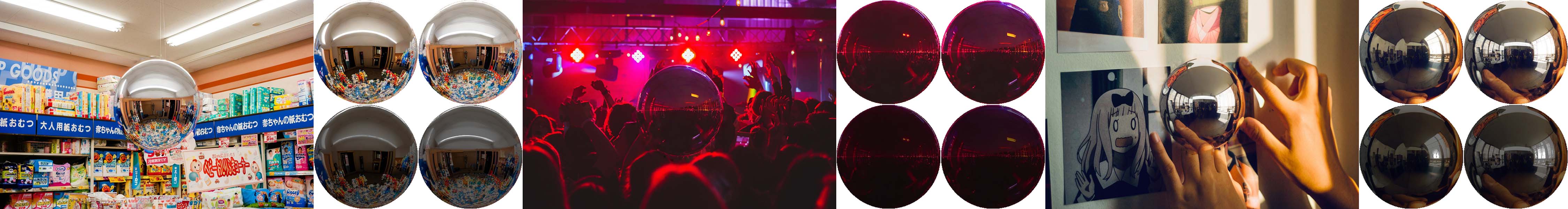} \\
        \includegraphics[width=1.0\textwidth]{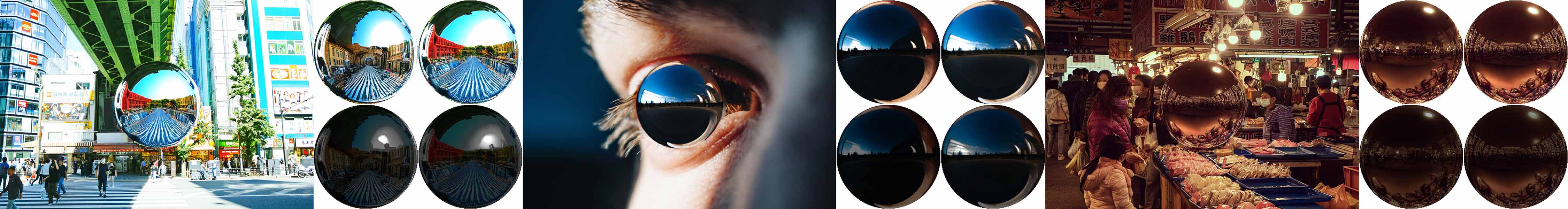} \\
        \includegraphics[width=1.0\textwidth]{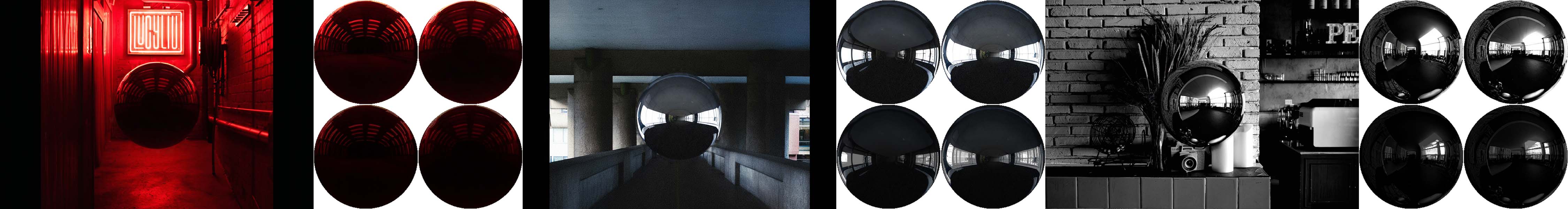} \\
        \includegraphics[width=1.0\textwidth]{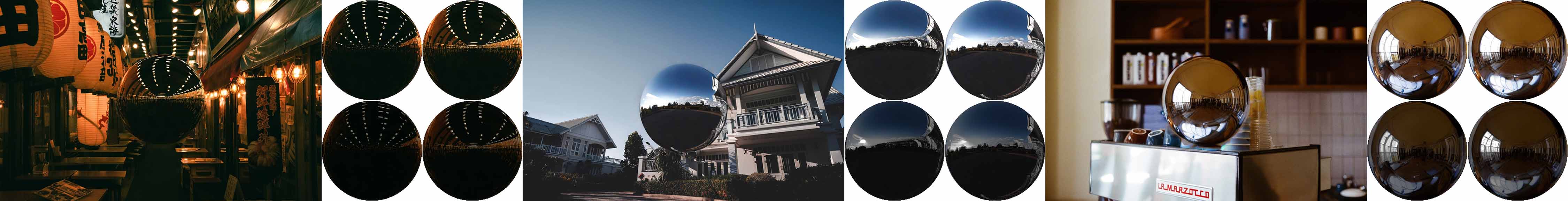} \\
        \includegraphics[width=1.0\textwidth]{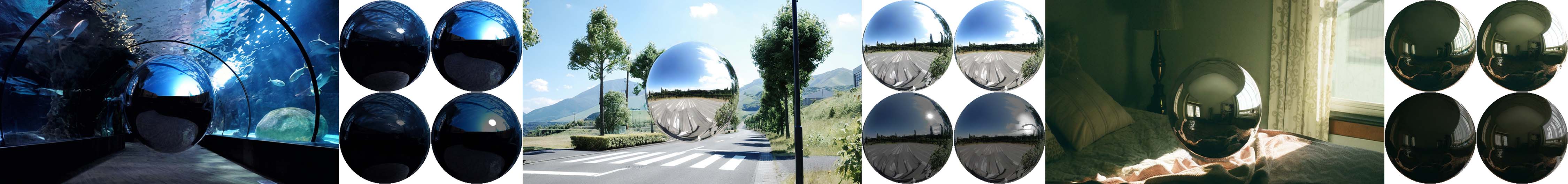} \\
        \includegraphics[width=1.0\textwidth]{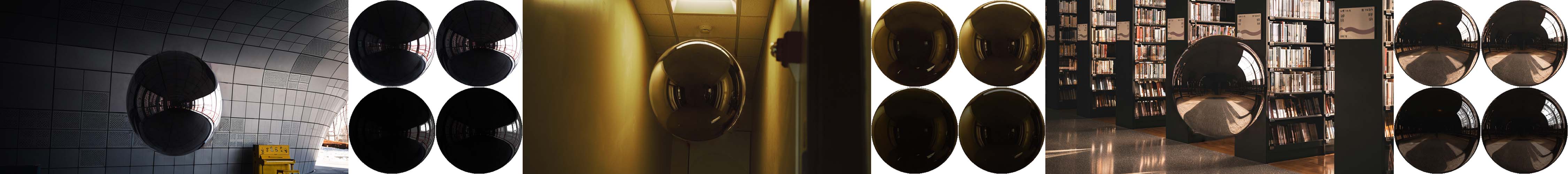}
        
    \end{tabular}

    \caption{Additional qualitative results for in-the-wild scenes. For each input, we show a chrome ball generated from our DiffusionLight and its underexposed version in the left column. The outputs from DiffusionLight-Turbo are presented in the right column.}
    \label{fig:aba_wild_general}
\end{figure*}

\begin{figure*}
    \centering
    \renewcommand{\arraystretch}{0.5} 
    \begin{tabular}{c}
        \includegraphics[width=1.0\textwidth]{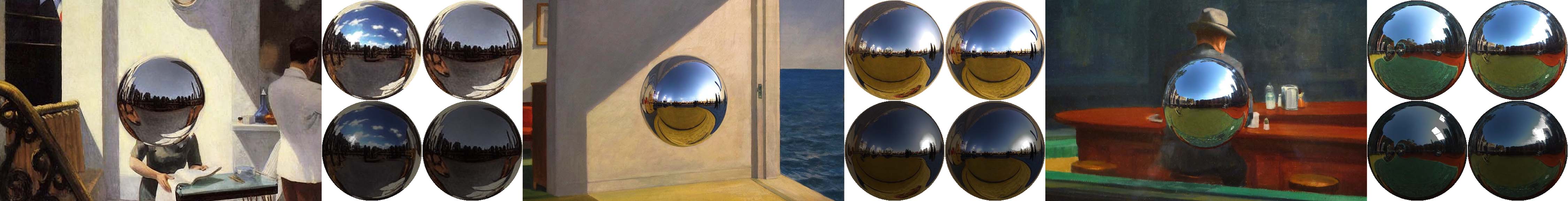} \\
        \includegraphics[width=1.0\textwidth]{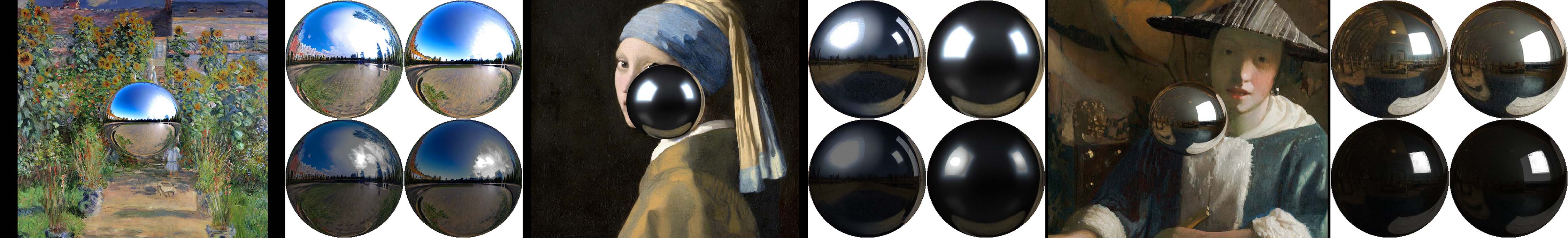} \\
        \includegraphics[width=1.0\textwidth]{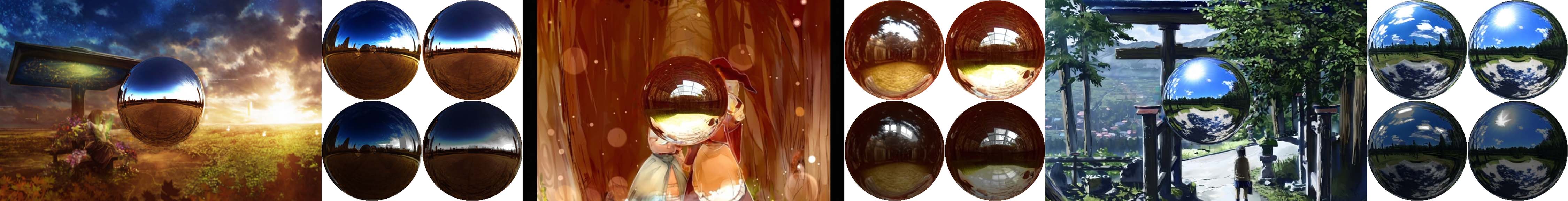} \\

    \end{tabular}
    \caption{Qualitative results for artificial images such as paintings and painting-like Japanese animation-style images. For each input, we show a chrome ball generated from our DiffusionLight and its underexposed version in the left column. The outputs from DiffusionLight-Turbo are presented in the right column. Our proposed methods can still perform reasonably well, albeit with some performance degradation, by leveraging the strong generative prior of SDXL \cite{podell2023sdxl}.}
    \label{fig:aba_wild_painting}
\end{figure*}

We present additional qualitative results for in-the-wild scenes in Figure \ref{fig:aba_wild_general}. Our method produces high-quality chrome balls that harmonize well with diverse scenes and lighting environments, such as a dim hallway under red neon lighting, an underwater tunnel with blue-tinted sunlight, a close-up shot of food, and a bird's-eye view from a tall building. 
Thanks to the strong image priors in diffusion models, our method also performs well on non-photorealistic images that resemble those in the training data. As shown in Figure~\ref{fig:aba_wild_painting}, this includes paintings and painting-like, animation-style images, where visual cues such as shading and shadows are still present.


\section{Spatially-Varying Light Estimation}
\begin{figure*}[ht]
    \centering

    \begin{tabu} to \textwidth {
        @{}
        c@{\hspace{2pt}}
        c@{\hspace{0.5pt}}
        c@{\hspace{0.5pt}}
        c@{\hspace{2pt}}
        c@{\hspace{0.5pt}}
        c@{\hspace{0.5pt}}
        c@{\hspace{2pt}}
        c@{\hspace{0.5pt}}
        c@{\hspace{0.5pt}}
        c@{\hspace{0.5pt}}
        c@{}
    }
        
        \multicolumn{1}{c}{\shortstack{\scriptsize Input image}} & 
        \multicolumn{3}{c}{\shortstack{\scriptsize Prediction}} &
        \multicolumn{3}{c}{\shortstack{\scriptsize Median ball (1\textsuperscript{st} iteration)}} &
        \multicolumn{3}{c}{\shortstack{\scriptsize Median ball (2\textsuperscript{nd} iteration)}} &
        \\

        \noindent\parbox[c]{0.140\textwidth}{\includegraphics[width=0.140\textwidth]{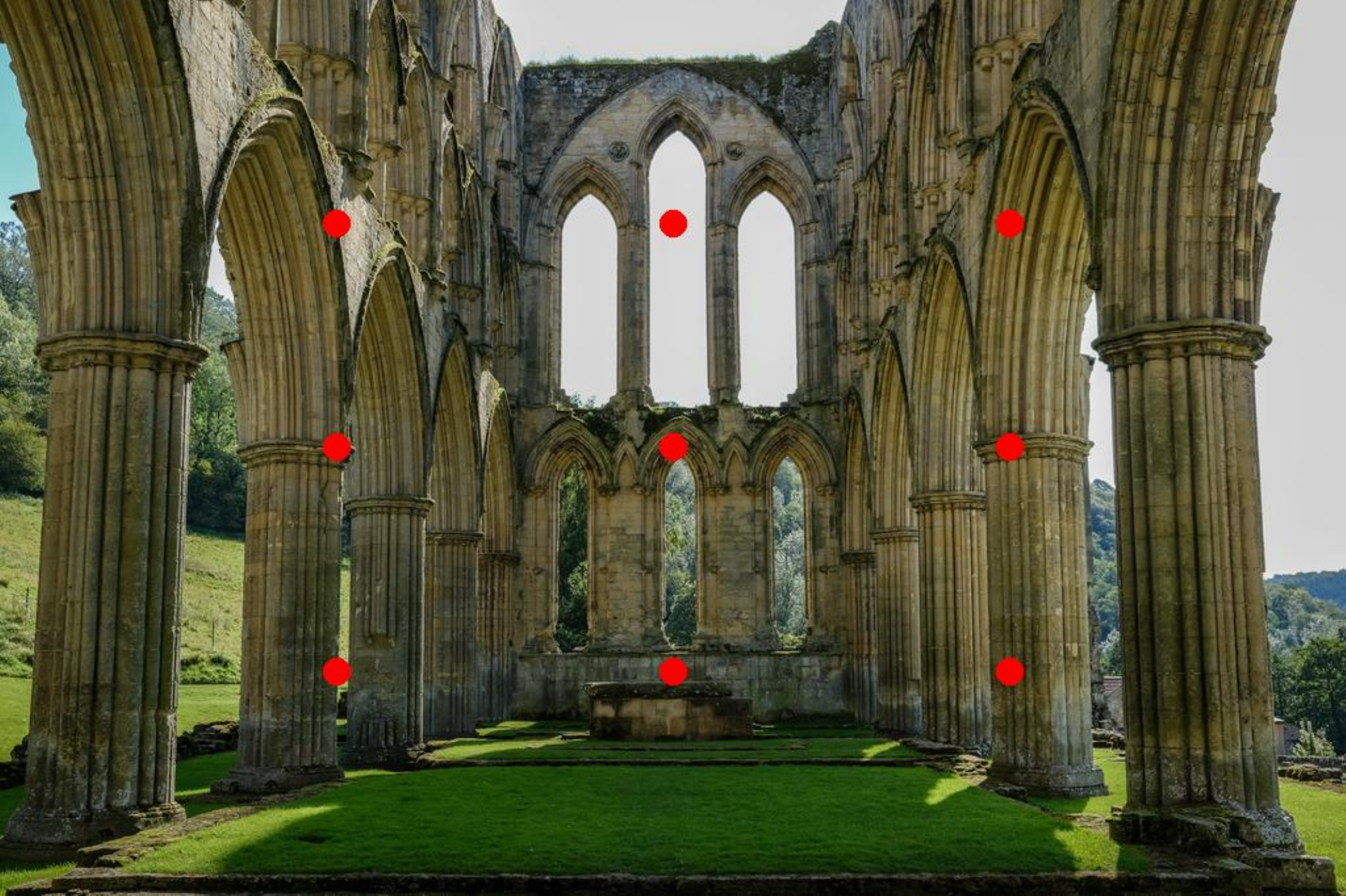}} & 
        \noindent\parbox[c]{0.092\textwidth}{\includegraphics[width=0.092\textwidth]{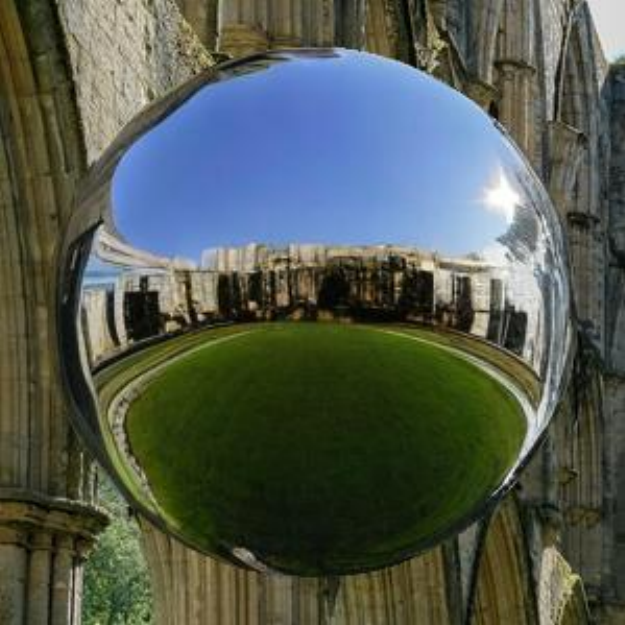}} & 
        \noindent\parbox[c]{0.092\textwidth}{\includegraphics[width=0.092\textwidth]{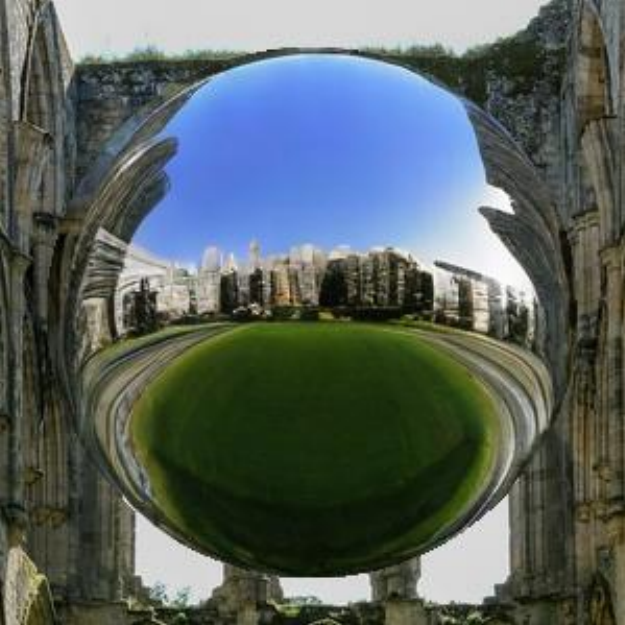}} & 
        \noindent\parbox[c]{0.092\textwidth}{\includegraphics[width=0.092\textwidth]{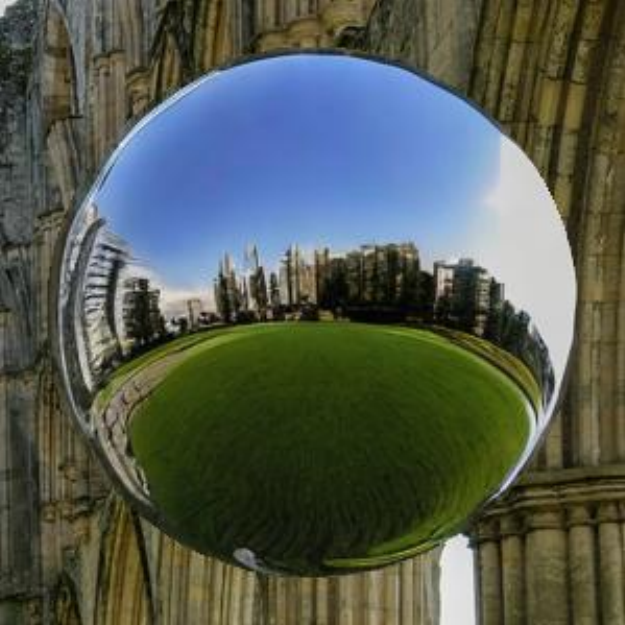}} & 
        \noindent\parbox[c]{0.092\textwidth}{\includegraphics[width=0.092\textwidth]{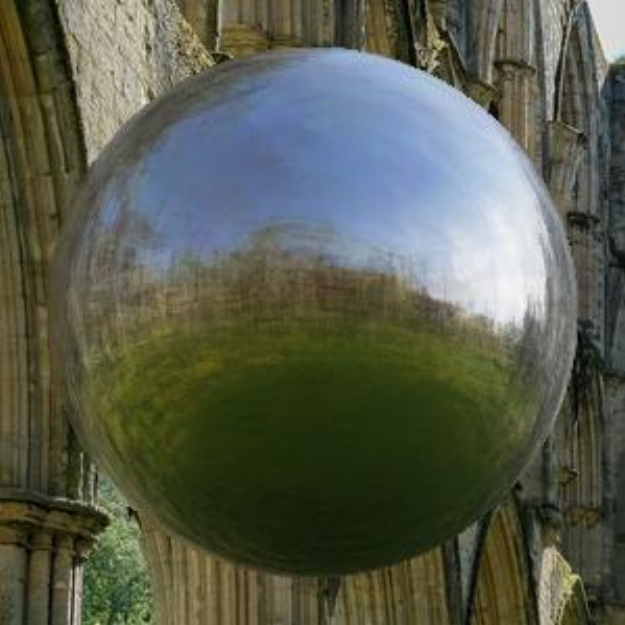}} & 
        \noindent\parbox[c]{0.092\textwidth}{\includegraphics[width=0.092\textwidth]{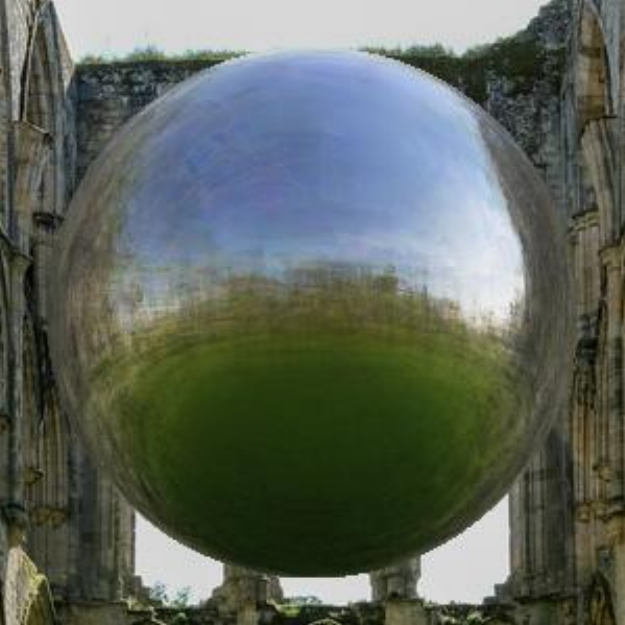}} & 
        \noindent\parbox[c]{0.092\textwidth}{\includegraphics[width=0.092\textwidth]{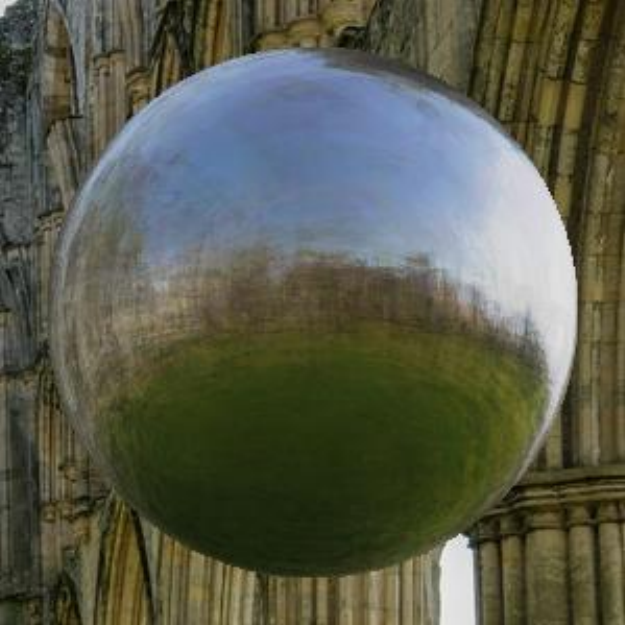}} & 
        \noindent\parbox[c]{0.092\textwidth}{\includegraphics[width=0.092\textwidth]{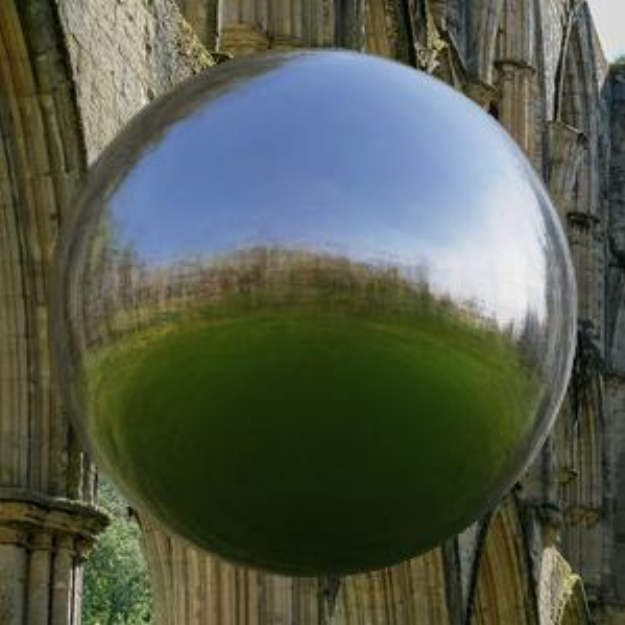}} & 
        \noindent\parbox[c]{0.092\textwidth}{\includegraphics[width=0.092\textwidth]{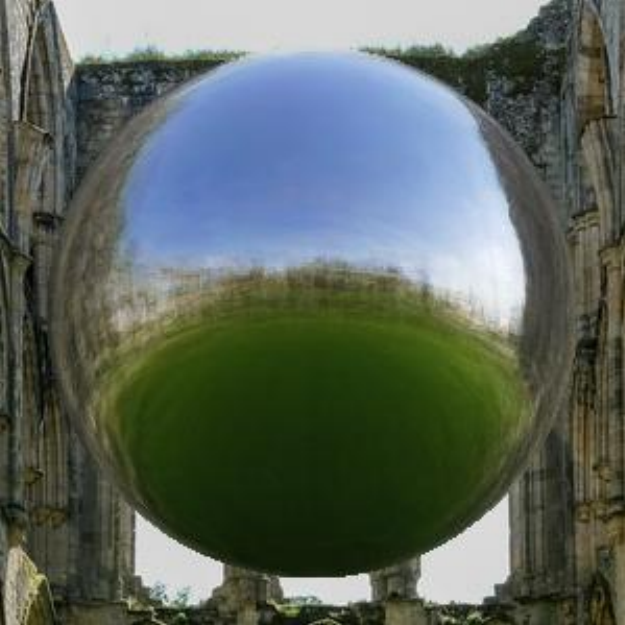}} & 
        \noindent\parbox[c]{0.092\textwidth}{\includegraphics[width=0.092\textwidth]{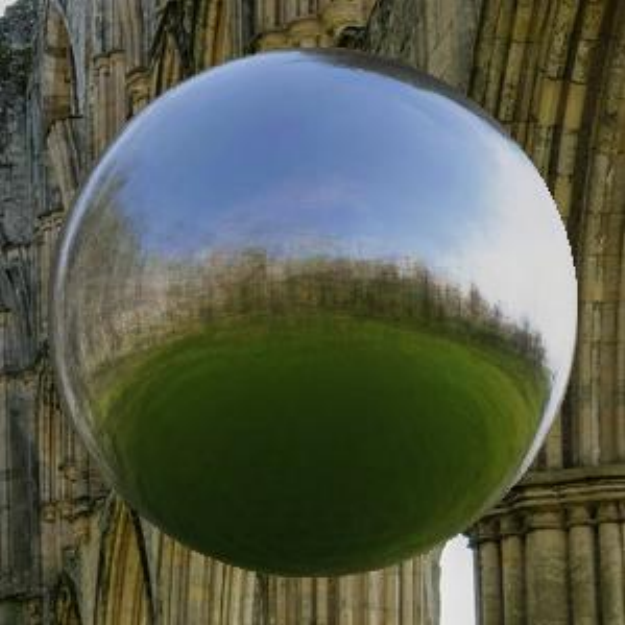}} & 
        
        \\

        & 
        \noindent\parbox[c]{0.092\textwidth}{\includegraphics[width=0.092\textwidth]{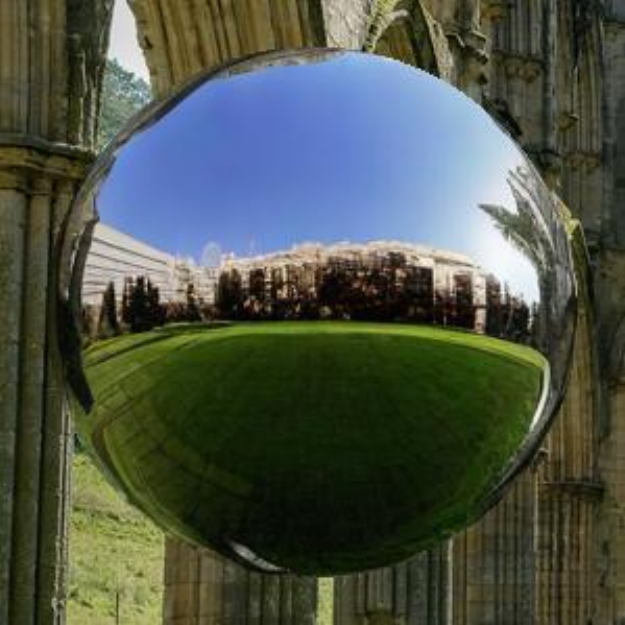}} & 
        \noindent\parbox[c]{0.092\textwidth}{\includegraphics[width=0.092\textwidth]{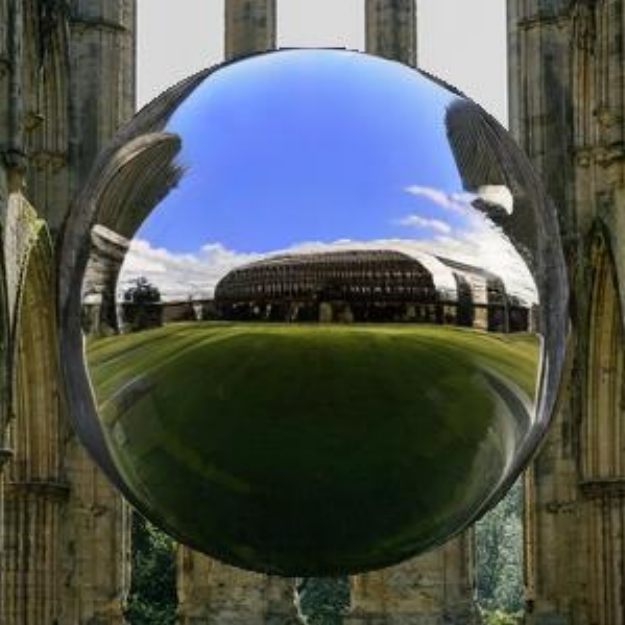}} & 
        \noindent\parbox[c]{0.092\textwidth}{\includegraphics[width=0.092\textwidth]{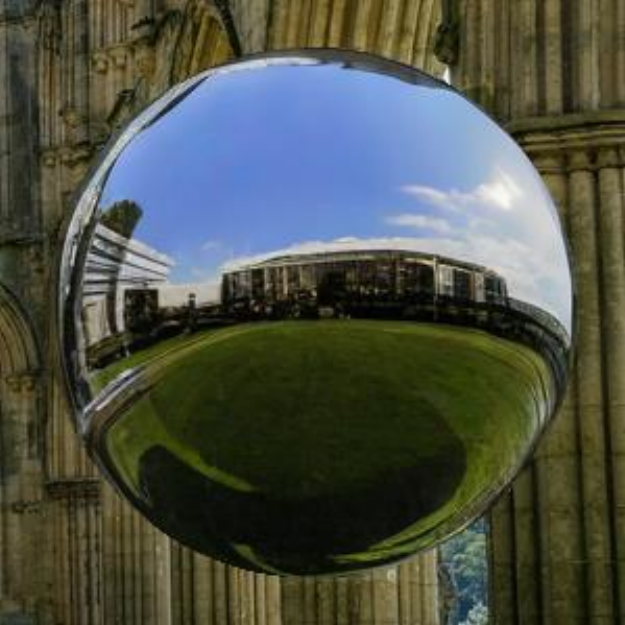}} & 
        \noindent\parbox[c]{0.092\textwidth}{\includegraphics[width=0.092\textwidth]{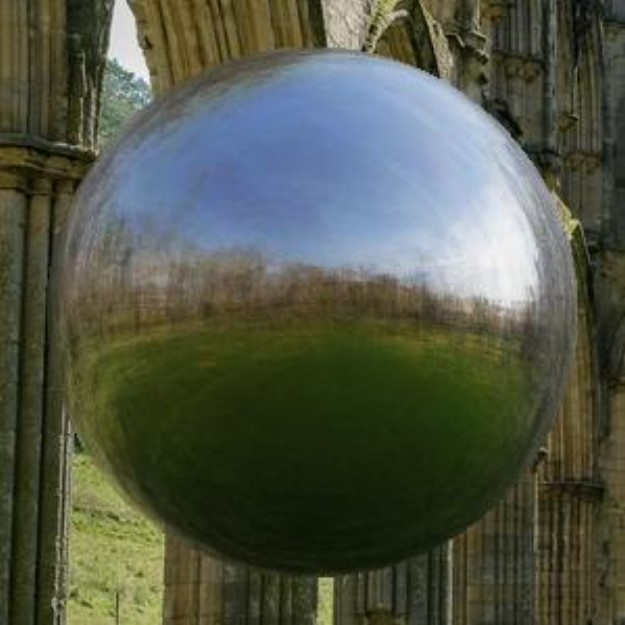}} & 
        \noindent\parbox[c]{0.092\textwidth}{\includegraphics[width=0.092\textwidth]{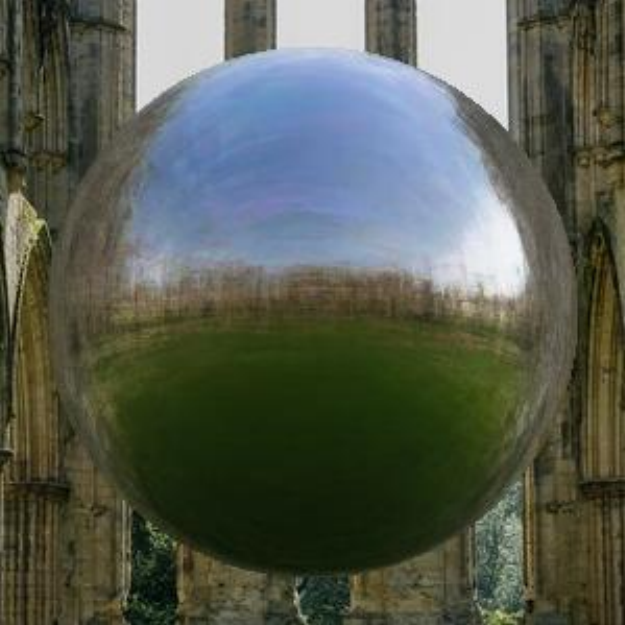}} & 
        \noindent\parbox[c]{0.092\textwidth}{\includegraphics[width=0.092\textwidth]{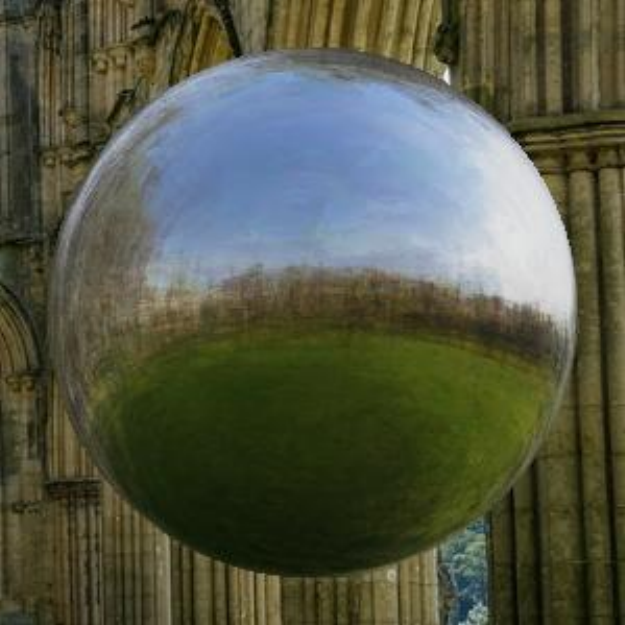}} & 
        \noindent\parbox[c]{0.092\textwidth}{\includegraphics[width=0.092\textwidth]{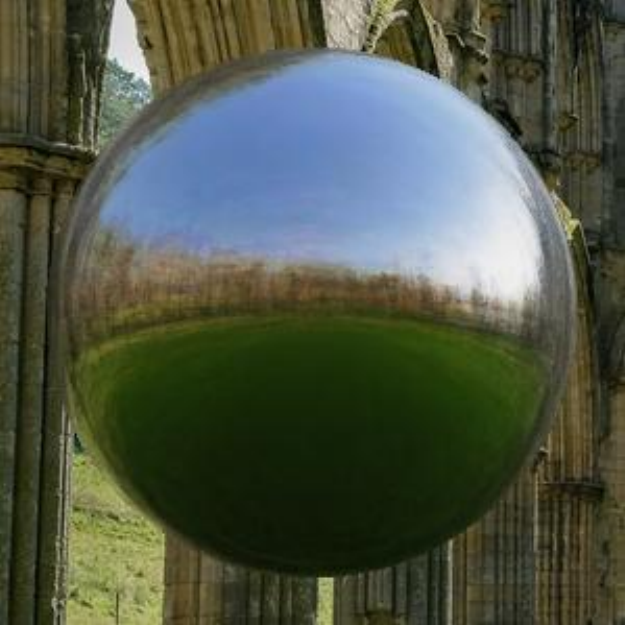}} & 
        \noindent\parbox[c]{0.092\textwidth}{\includegraphics[width=0.092\textwidth]{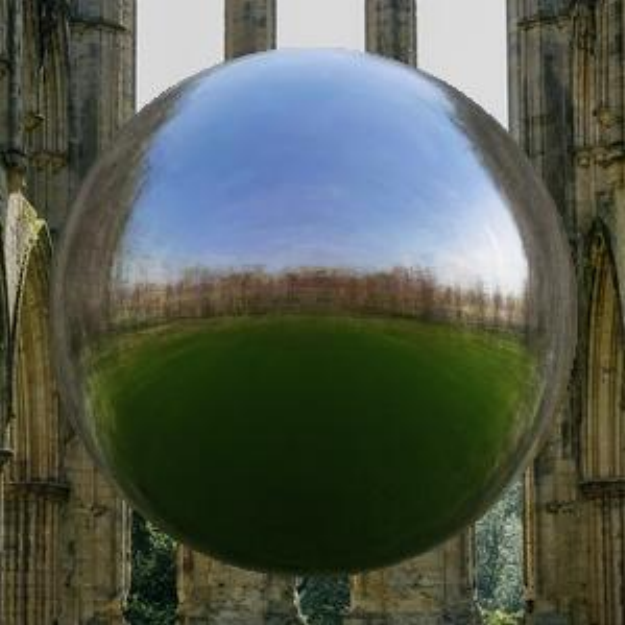}} & 
        \noindent\parbox[c]{0.092\textwidth}{\includegraphics[width=0.092\textwidth]{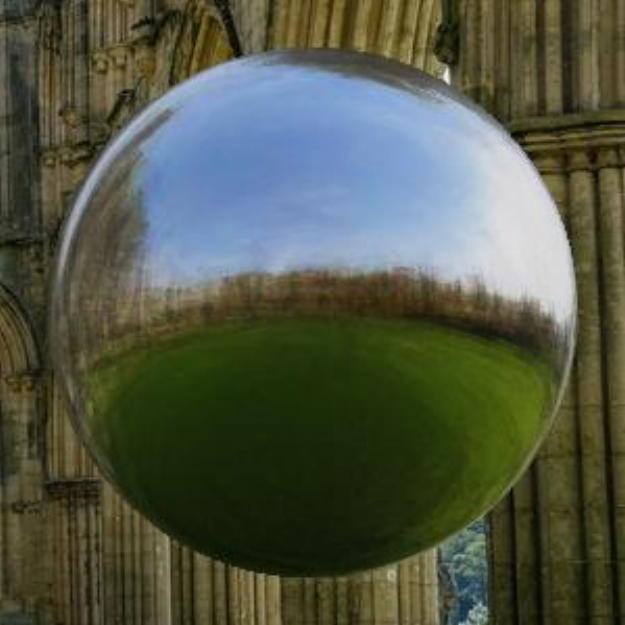}} & 
        
        \\

        & 
        \noindent\parbox[c]{0.092\textwidth}{\includegraphics[width=0.092\textwidth]{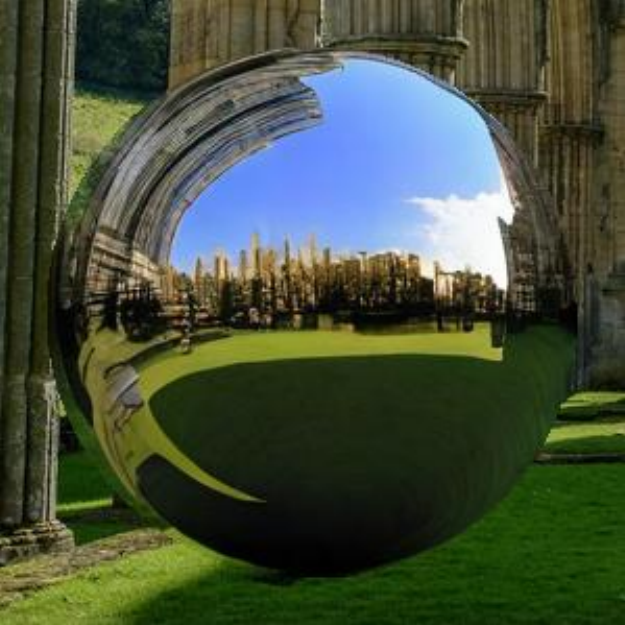}} & 
        \noindent\parbox[c]{0.092\textwidth}{\includegraphics[width=0.092\textwidth]{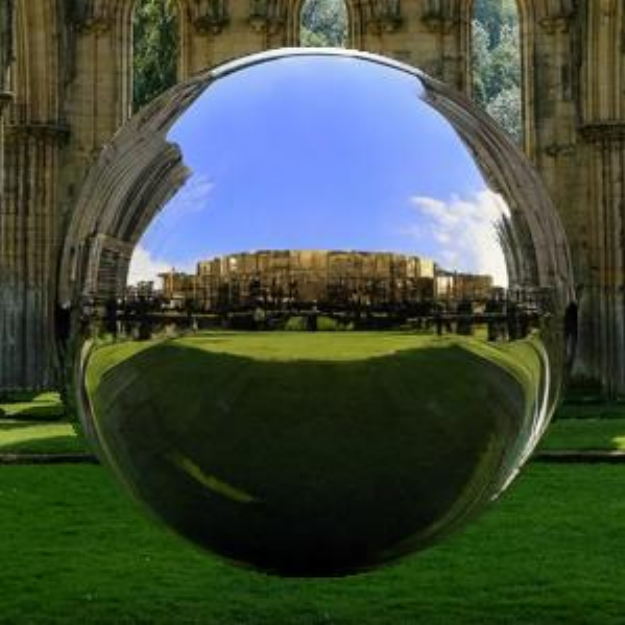}} & 
        \noindent\parbox[c]{0.092\textwidth}{\includegraphics[width=0.092\textwidth]{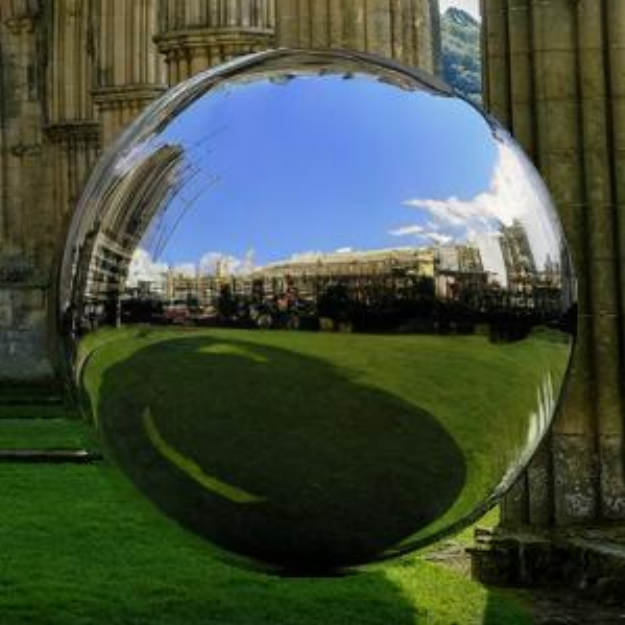}} & 
        \noindent\parbox[c]{0.092\textwidth}{\includegraphics[width=0.092\textwidth]{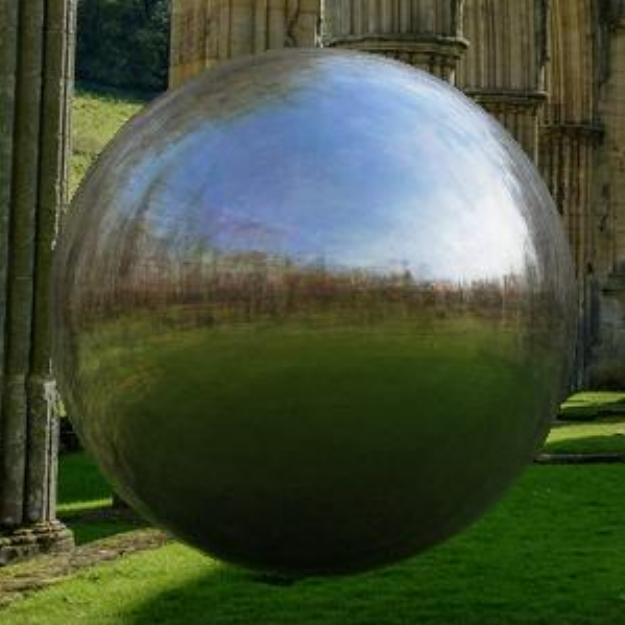}} & 
        \noindent\parbox[c]{0.092\textwidth}{\includegraphics[width=0.092\textwidth]{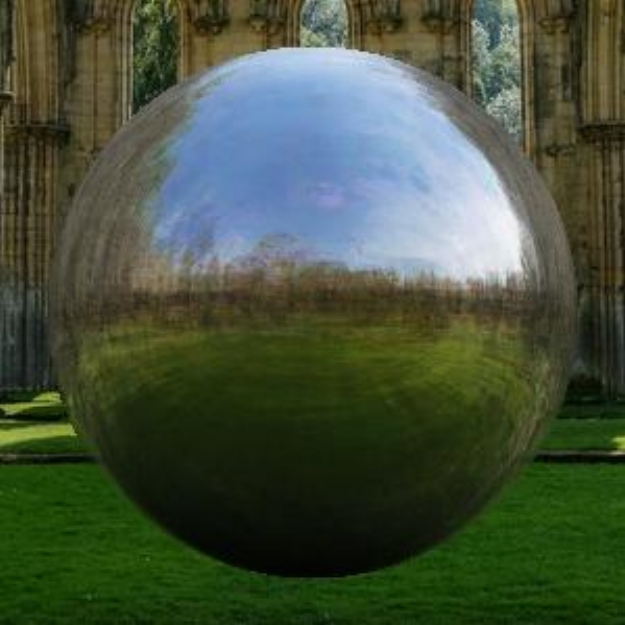}} & 
        \noindent\parbox[c]{0.092\textwidth}{\includegraphics[width=0.092\textwidth]{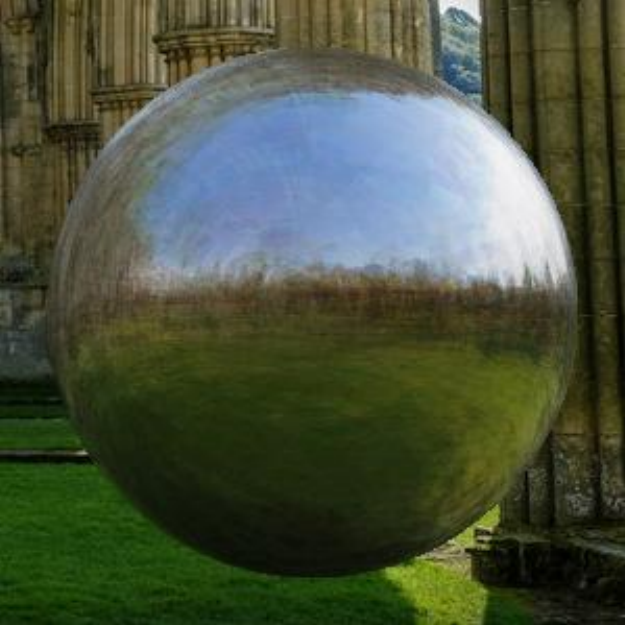}} & 
        \noindent\parbox[c]{0.092\textwidth}{\includegraphics[width=0.092\textwidth]{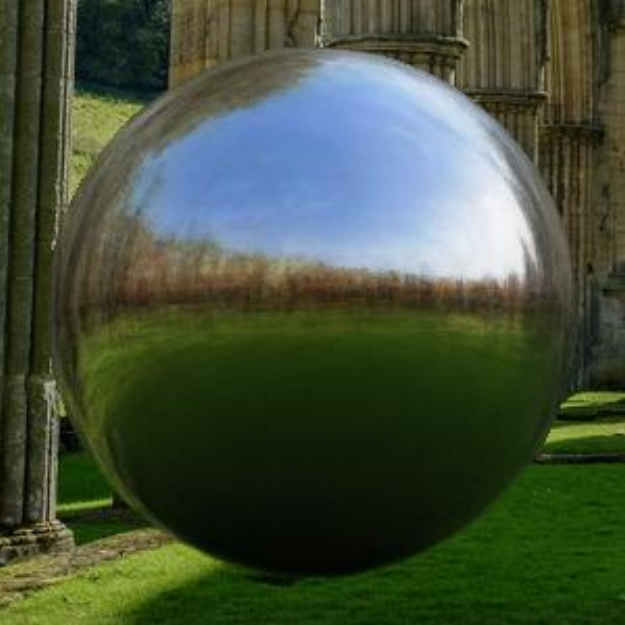}} & 
        \noindent\parbox[c]{0.092\textwidth}{\includegraphics[width=0.092\textwidth]{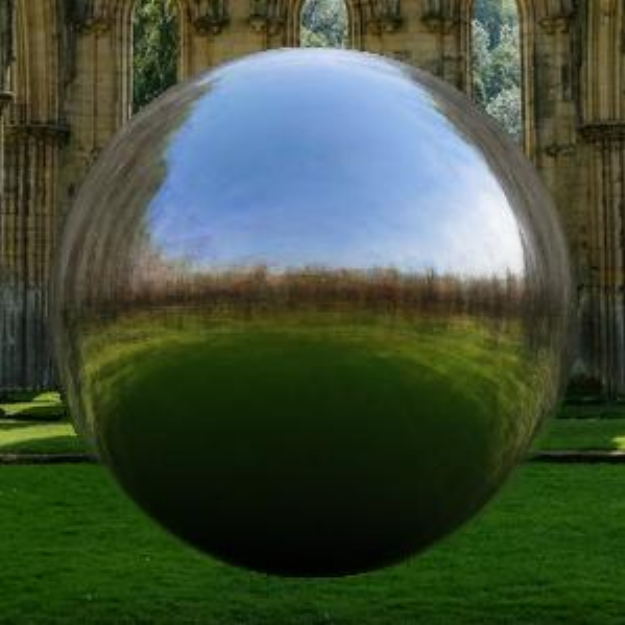}} & 
        \noindent\parbox[c]{0.092\textwidth}{\includegraphics[width=0.092\textwidth]{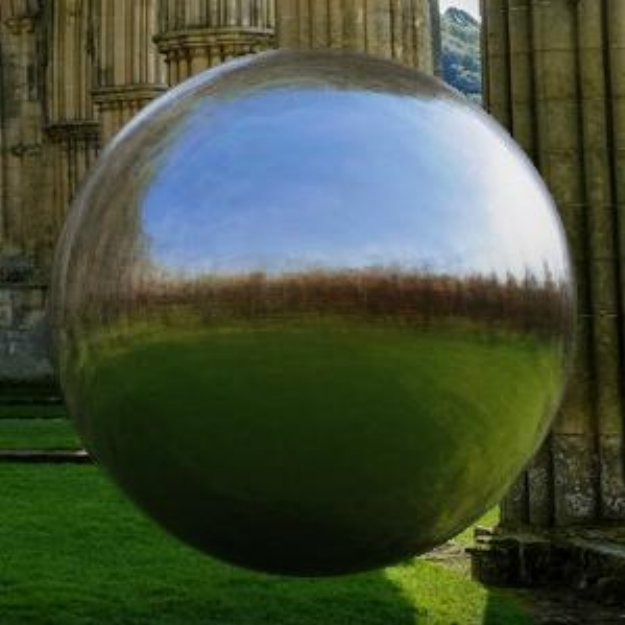}} & 
        
        \\
        
    \end{tabu}

    \smallskip
    \begin{tabu} to \textwidth {
        @{}
        c@{\hspace{2pt}}
        c@{\hspace{0.5pt}}
        c@{\hspace{0.5pt}}
        c@{\hspace{2pt}}
        c@{\hspace{0.5pt}}
        c@{\hspace{0.5pt}}
        c@{\hspace{2pt}}
        c@{\hspace{0.5pt}}
        c@{\hspace{0.5pt}}
        c@{\hspace{0.5pt}}
        c@{}
    }
        

        \noindent\parbox[c]{0.140\textwidth}{\includegraphics[width=0.140\textwidth]{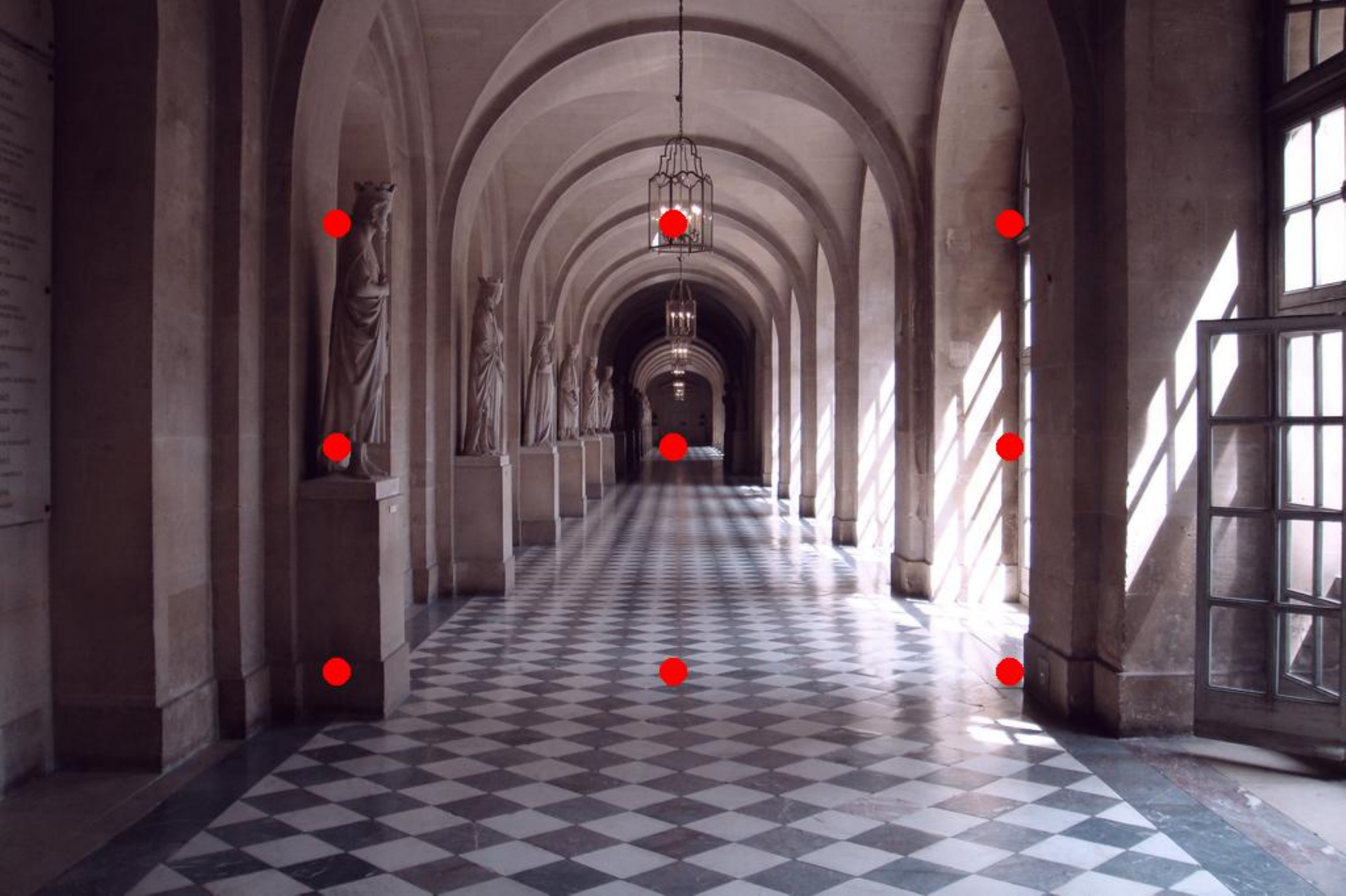}} & 
        \noindent\parbox[c]{0.092\textwidth}{\includegraphics[width=0.092\textwidth]{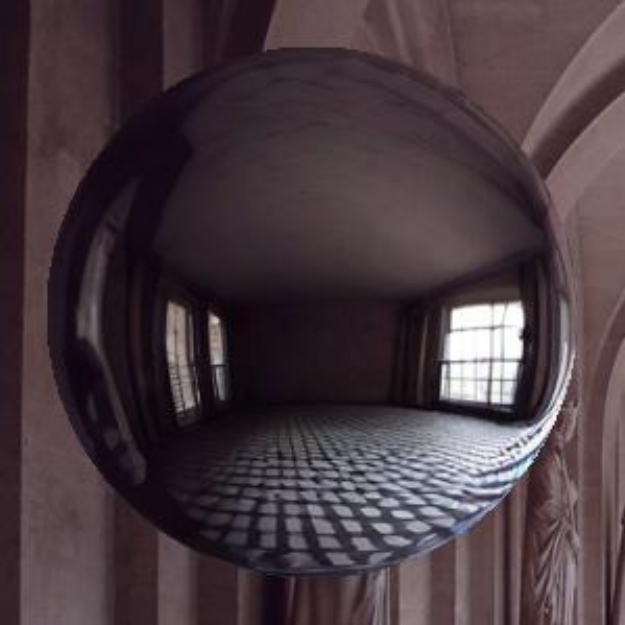}} & 
        \noindent\parbox[c]{0.092\textwidth}{\includegraphics[width=0.092\textwidth]{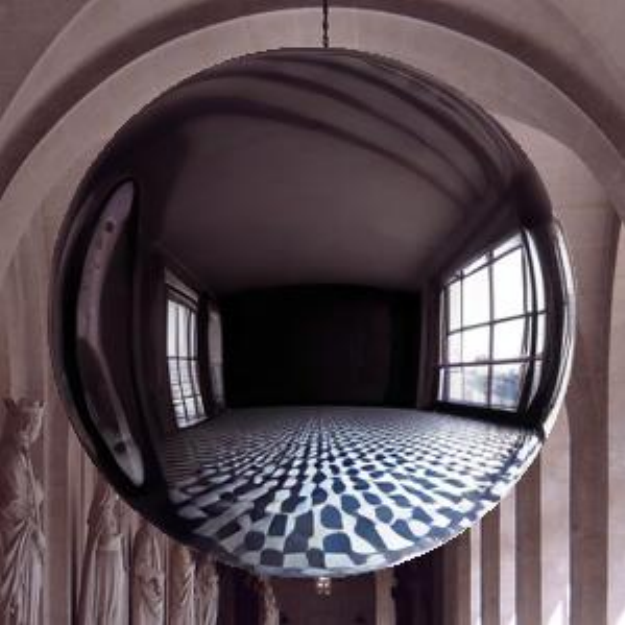}} & 
        \noindent\parbox[c]{0.092\textwidth}{\includegraphics[width=0.092\textwidth]{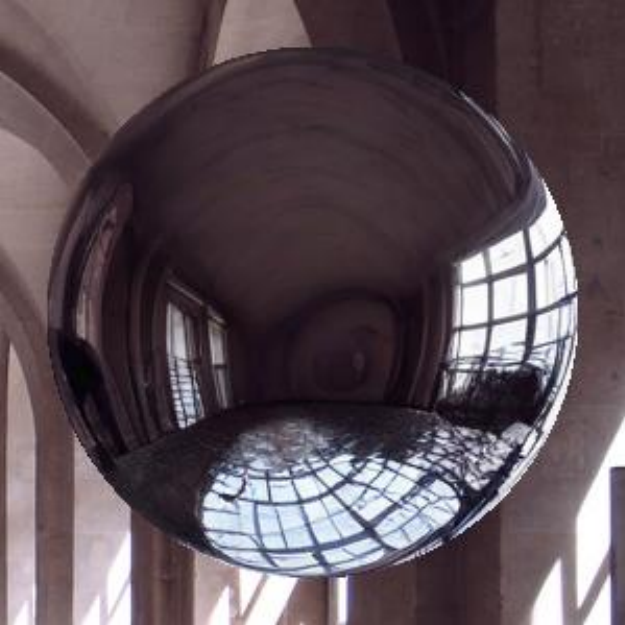}} & 
        \noindent\parbox[c]{0.092\textwidth}{\includegraphics[width=0.092\textwidth]{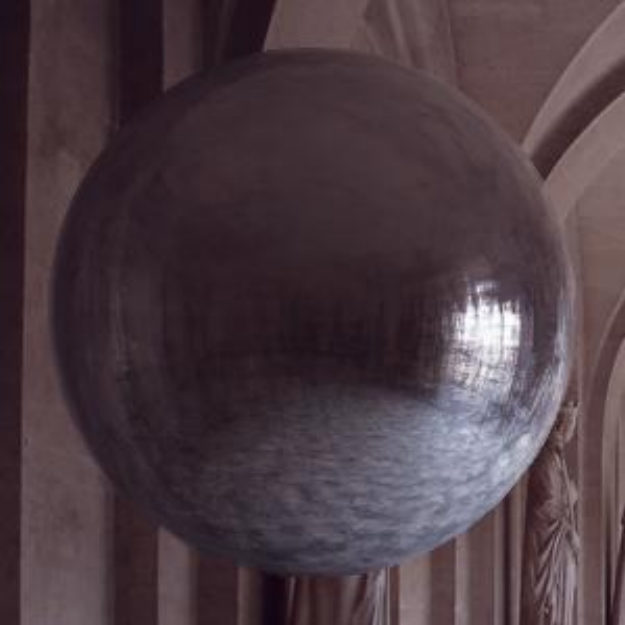}} & 
        \noindent\parbox[c]{0.092\textwidth}{\includegraphics[width=0.092\textwidth]{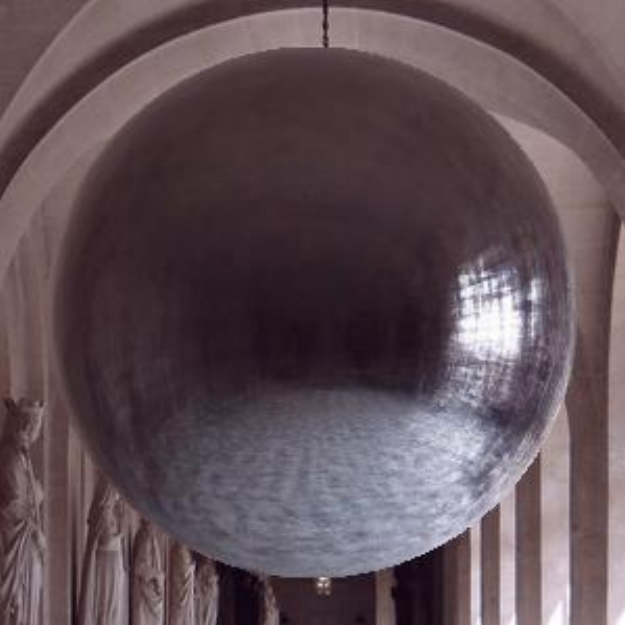}} & 
        \noindent\parbox[c]{0.092\textwidth}{\includegraphics[width=0.092\textwidth]{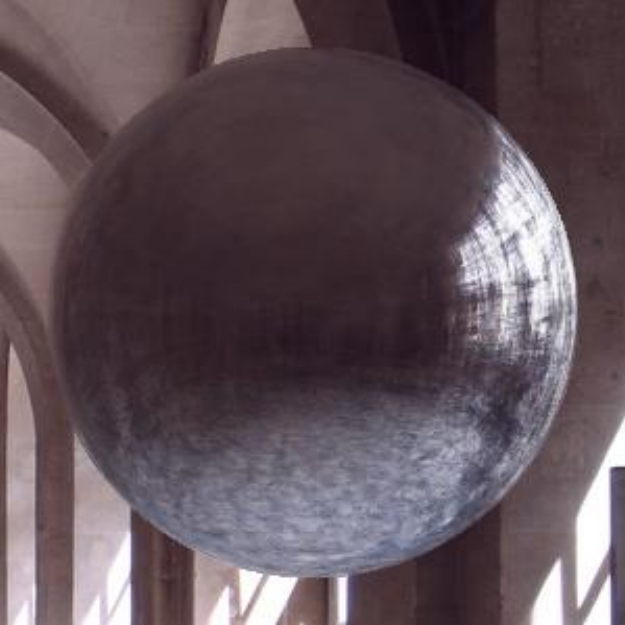}} & 
        \noindent\parbox[c]{0.092\textwidth}{\includegraphics[width=0.092\textwidth]{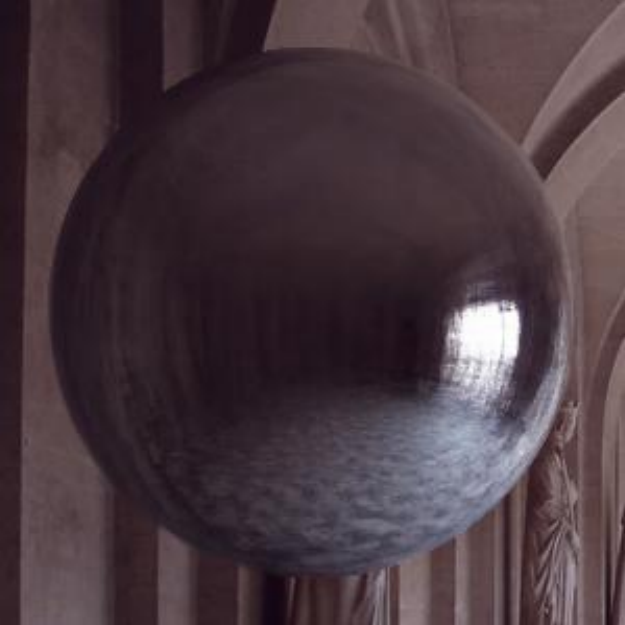}} & 
        \noindent\parbox[c]{0.092\textwidth}{\includegraphics[width=0.092\textwidth]{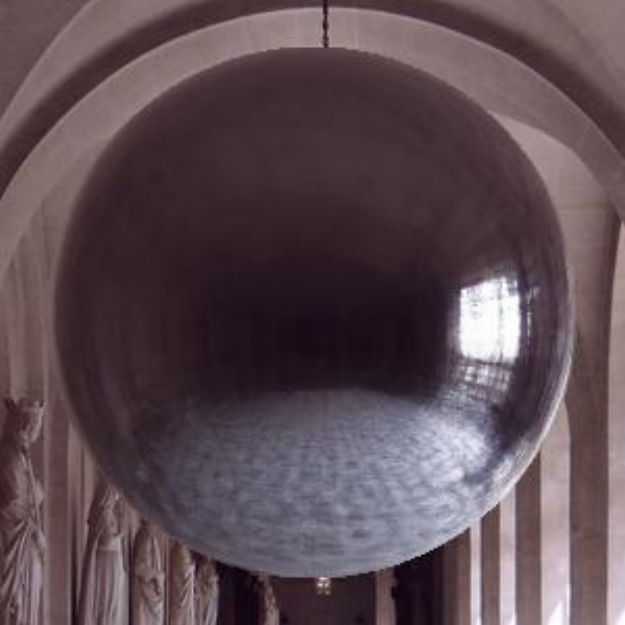}} & 
        \noindent\parbox[c]{0.092\textwidth}{\includegraphics[width=0.092\textwidth]{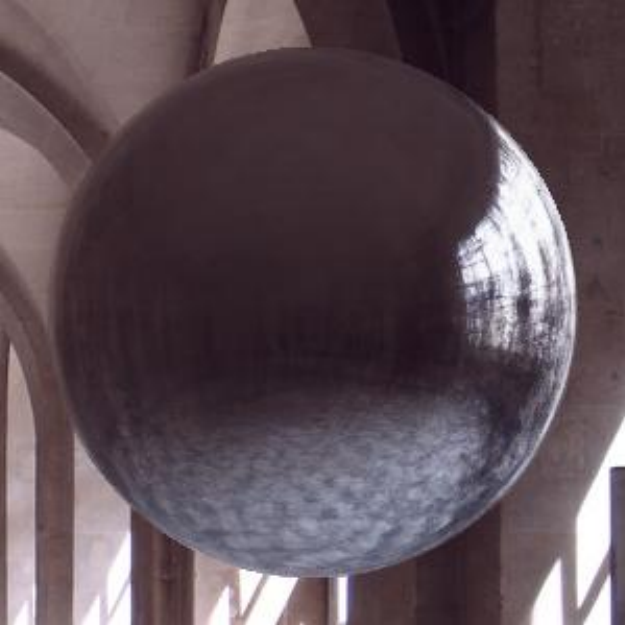}} & 
        
        \\

        & 
        \noindent\parbox[c]{0.092\textwidth}{\includegraphics[width=0.092\textwidth]{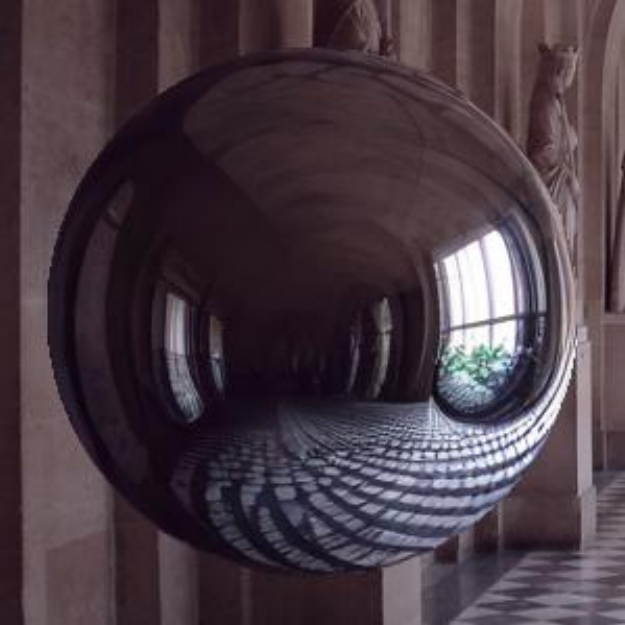}} & 
        \noindent\parbox[c]{0.092\textwidth}{\includegraphics[width=0.092\textwidth]{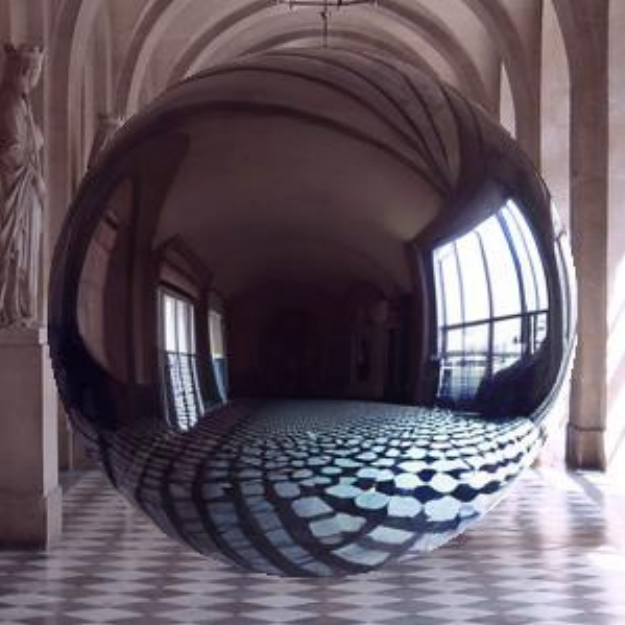}} & 
        \noindent\parbox[c]{0.092\textwidth}{\includegraphics[width=0.092\textwidth]{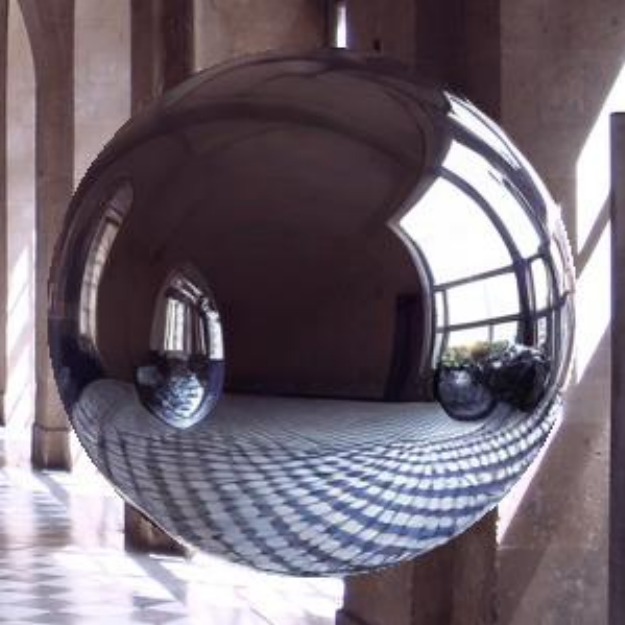}} & 
        \noindent\parbox[c]{0.092\textwidth}{\includegraphics[width=0.092\textwidth]{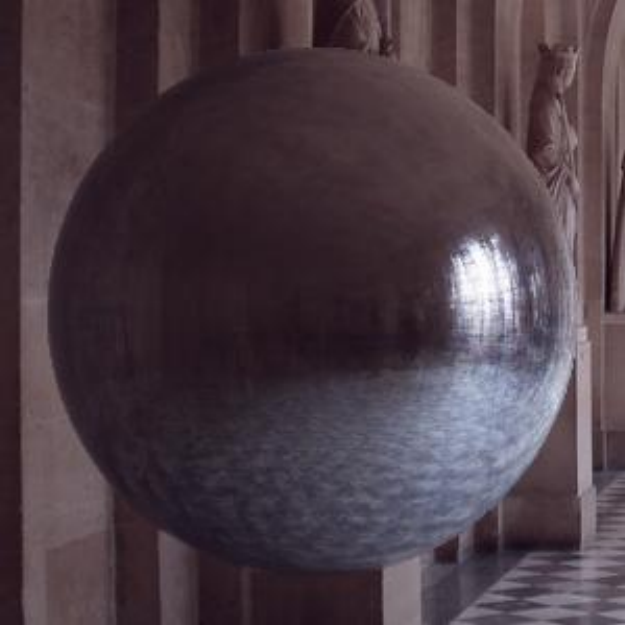}} & 
        \noindent\parbox[c]{0.092\textwidth}{\includegraphics[width=0.092\textwidth]{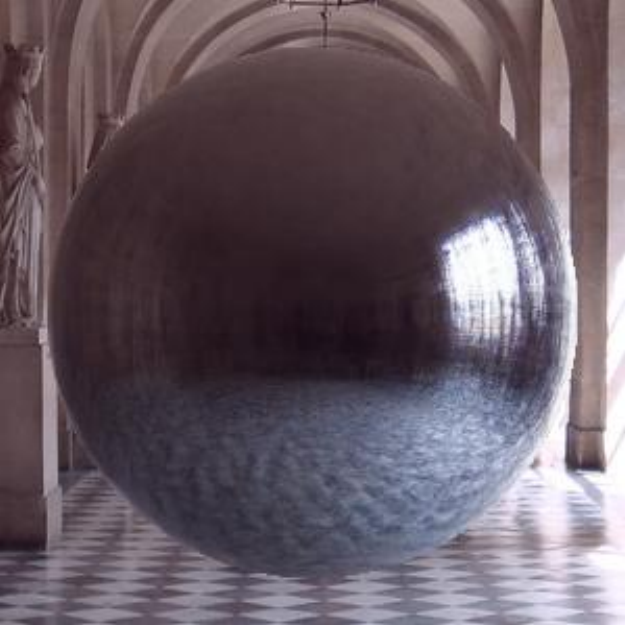}} & 
        \noindent\parbox[c]{0.092\textwidth}{\includegraphics[width=0.092\textwidth]{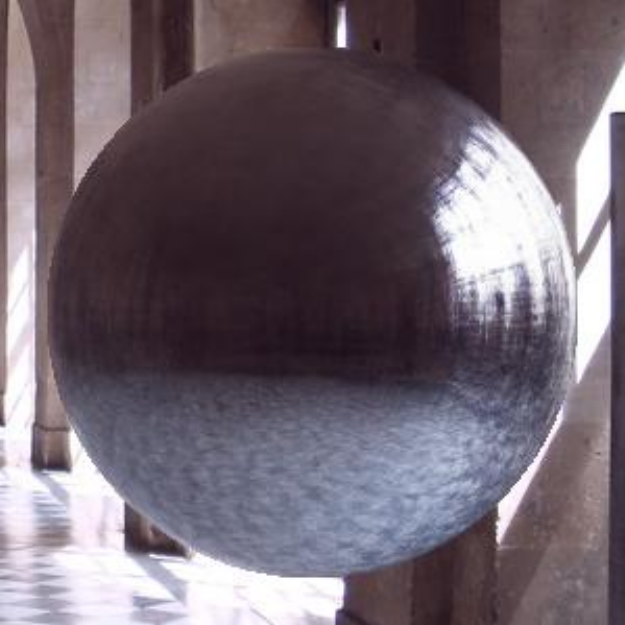}} & 
        \noindent\parbox[c]{0.092\textwidth}{\includegraphics[width=0.092\textwidth]{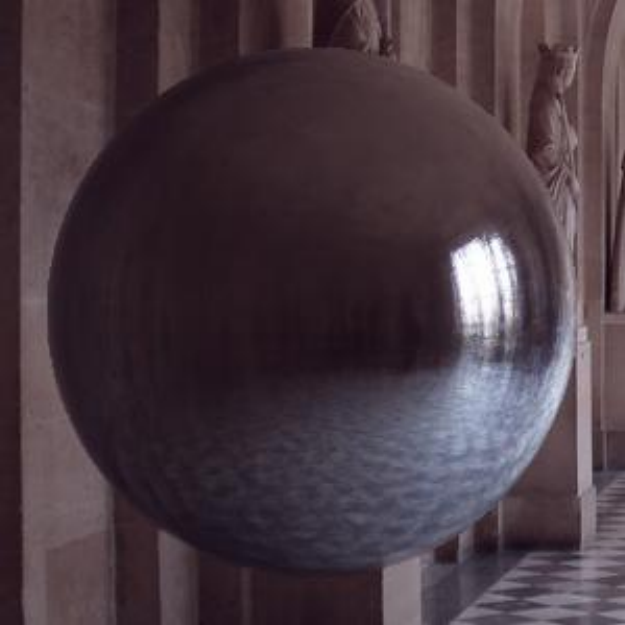}} & 
        \noindent\parbox[c]{0.092\textwidth}{\includegraphics[width=0.092\textwidth]{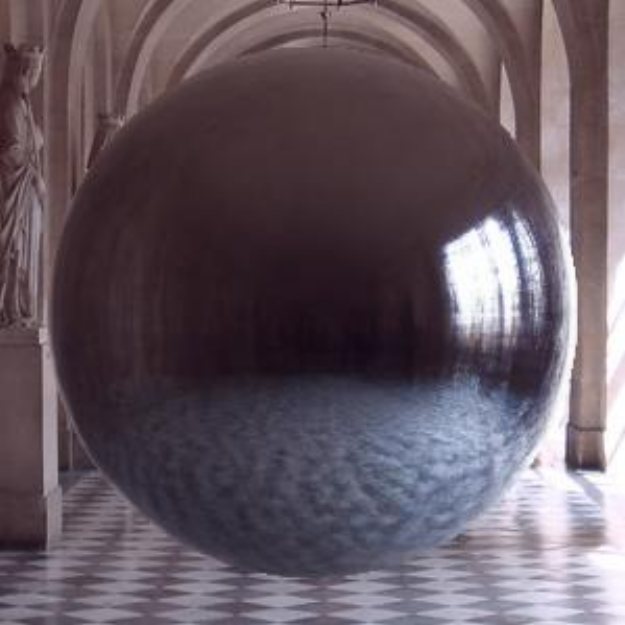}} & 
        \noindent\parbox[c]{0.092\textwidth}{\includegraphics[width=0.092\textwidth]{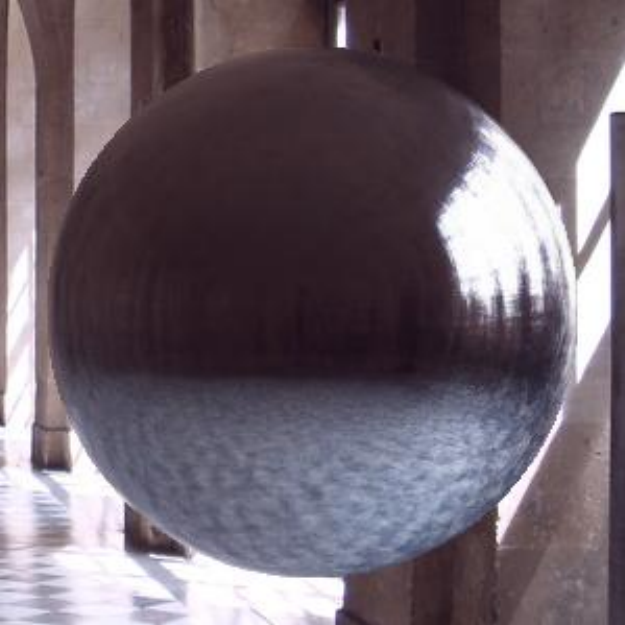}} & 
        
        \\

        & 
        \noindent\parbox[c]{0.092\textwidth}{\includegraphics[width=0.092\textwidth]{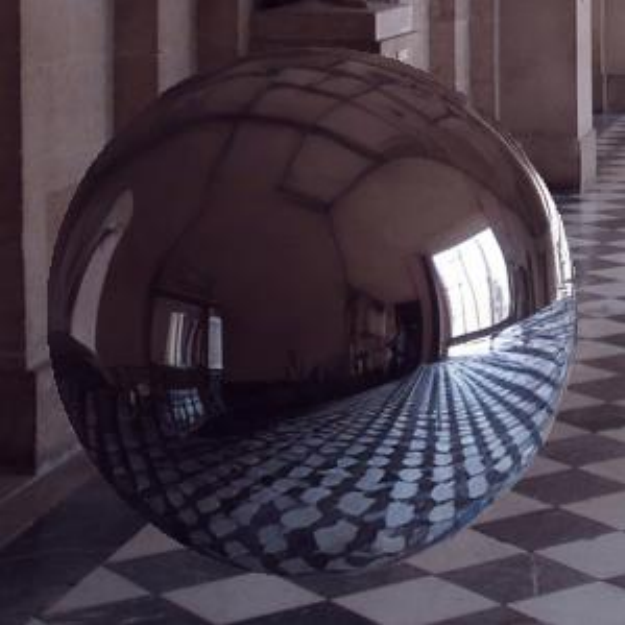}} & 
        \noindent\parbox[c]{0.092\textwidth}{\includegraphics[width=0.092\textwidth]{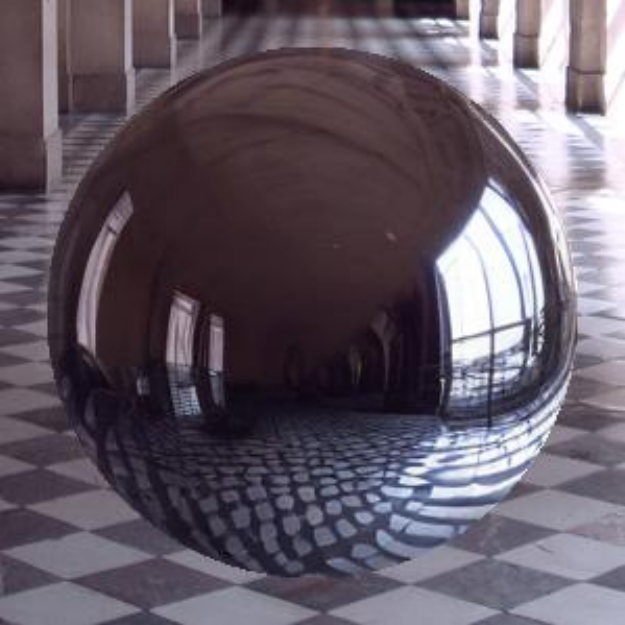}} & 
        \noindent\parbox[c]{0.092\textwidth}{\includegraphics[width=0.092\textwidth]{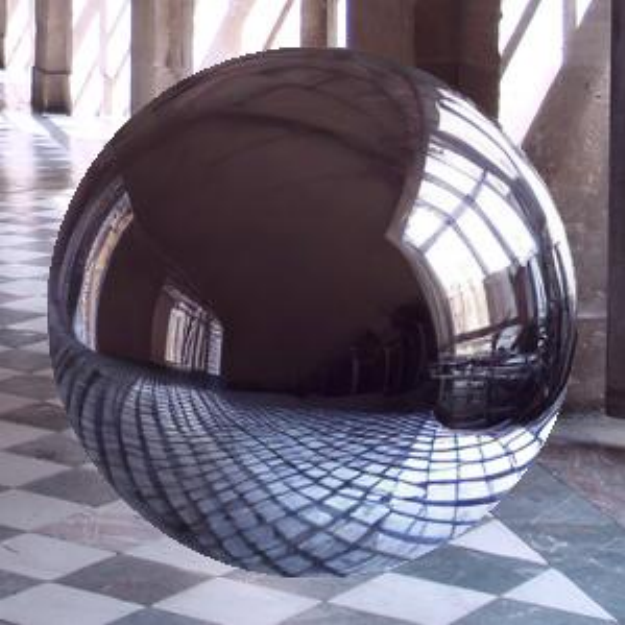}} & 
        \noindent\parbox[c]{0.092\textwidth}{\includegraphics[width=0.092\textwidth]{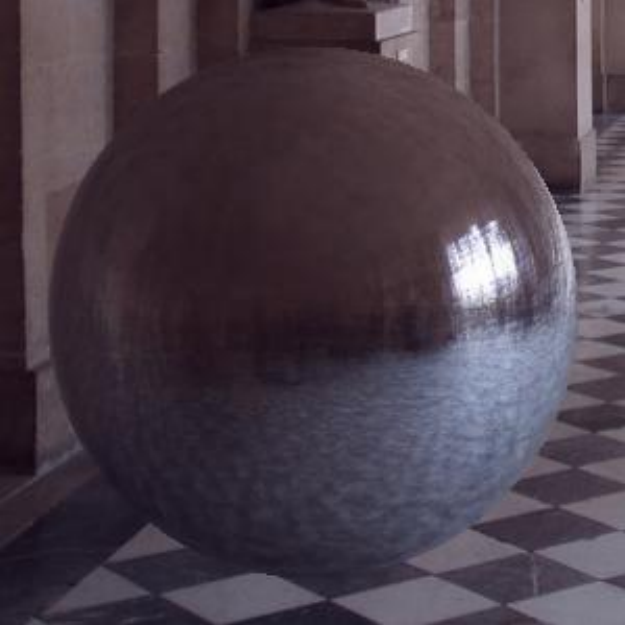}} & 
        \noindent\parbox[c]{0.092\textwidth}{\includegraphics[width=0.092\textwidth]{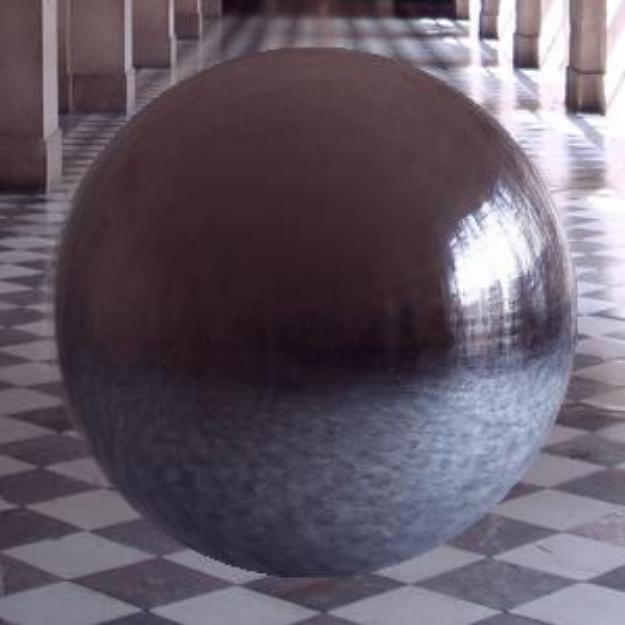}} & 
        \noindent\parbox[c]{0.092\textwidth}{\includegraphics[width=0.092\textwidth]{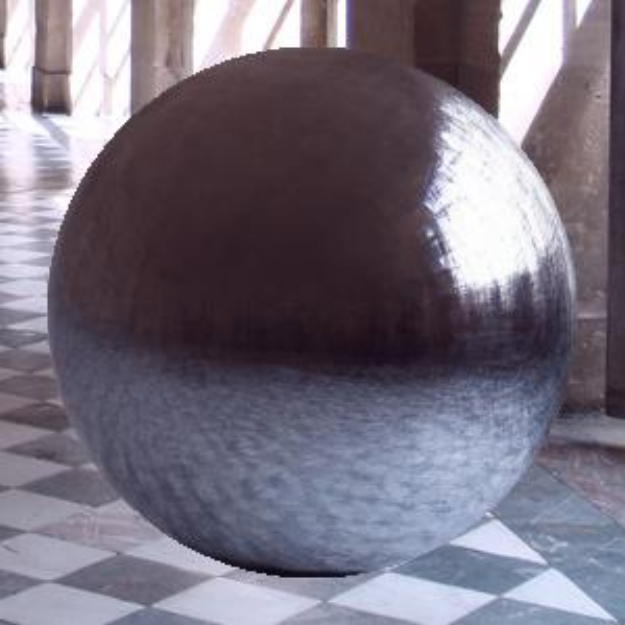}} & 
        \noindent\parbox[c]{0.092\textwidth}{\includegraphics[width=0.092\textwidth]{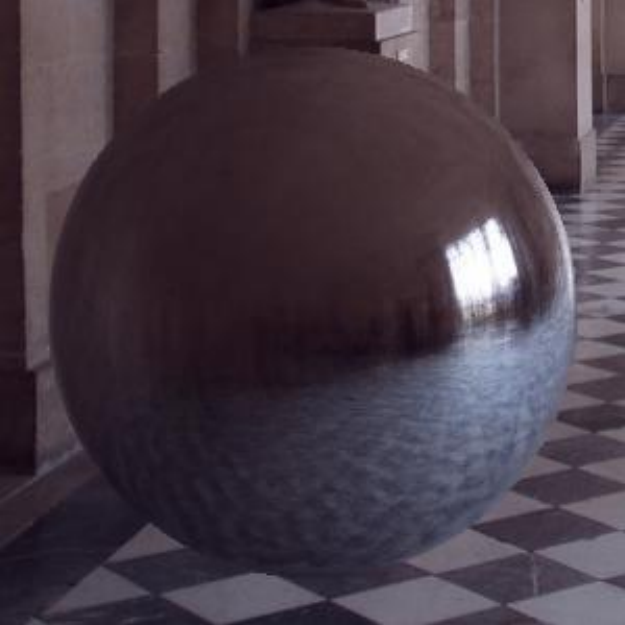}} & 
        \noindent\parbox[c]{0.092\textwidth}{\includegraphics[width=0.092\textwidth]{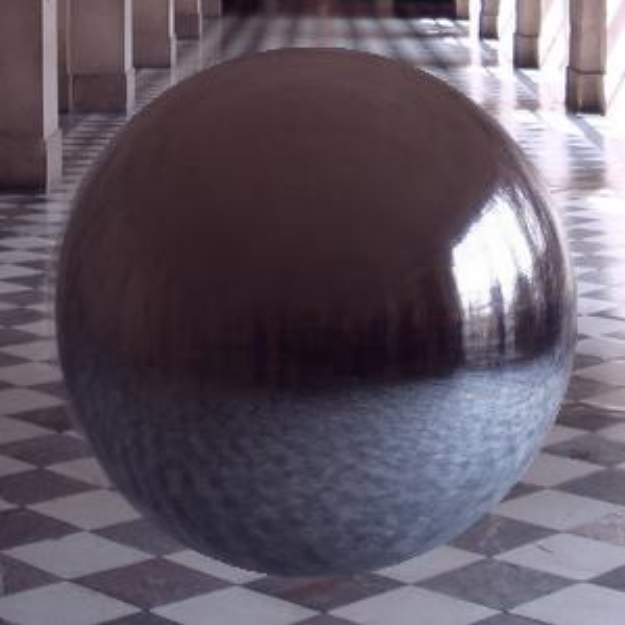}} & 
        \noindent\parbox[c]{0.092\textwidth}{\includegraphics[width=0.092\textwidth]{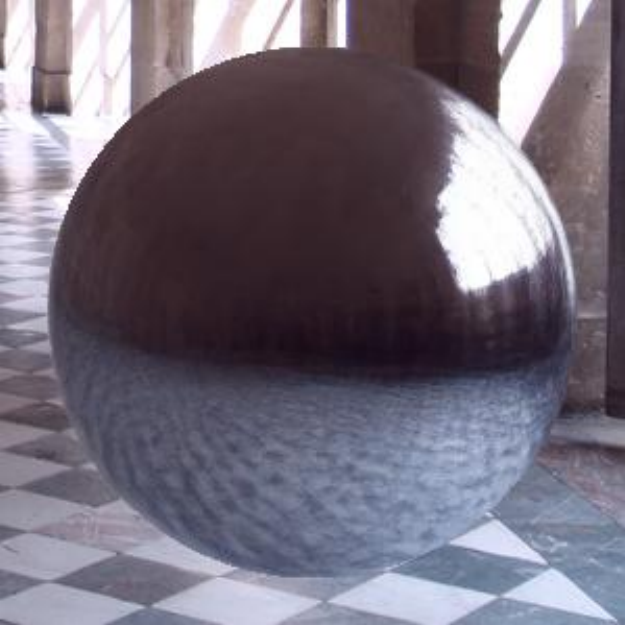}} & 
        
        \\
        
    \end{tabu}

    \smallskip
    \begin{tabu} to \textwidth {
        @{}
        c@{\hspace{2pt}}
        c@{\hspace{0.5pt}}
        c@{\hspace{0.5pt}}
        c@{\hspace{2pt}}
        c@{\hspace{0.5pt}}
        c@{\hspace{0.5pt}}
        c@{\hspace{2pt}}
        c@{\hspace{0.5pt}}
        c@{\hspace{0.5pt}}
        c@{\hspace{0.5pt}}
        c@{}
    }
        

        \noindent\parbox[c]{0.140\textwidth}{\includegraphics[width=0.140\textwidth]{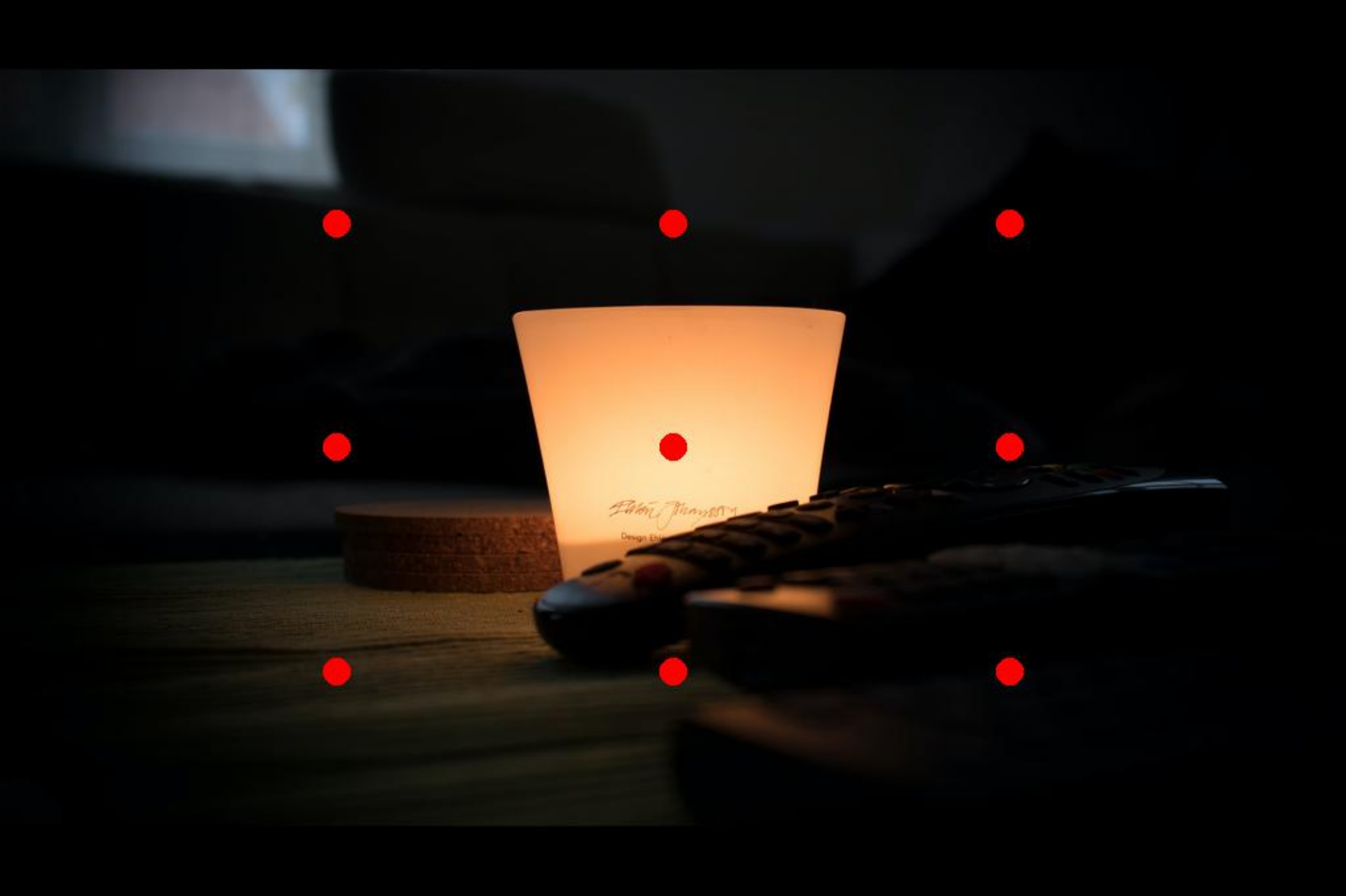}} & 
        \noindent\parbox[c]{0.092\textwidth}{\includegraphics[width=0.092\textwidth]{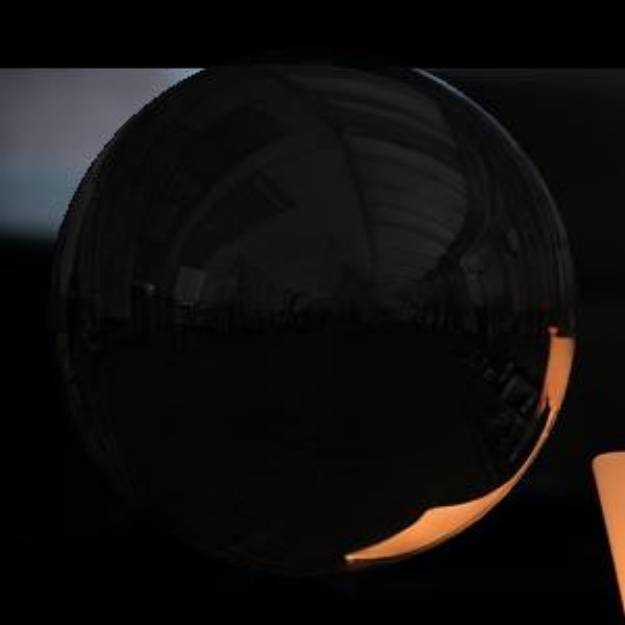}} & 
        \noindent\parbox[c]{0.092\textwidth}{\includegraphics[width=0.092\textwidth]{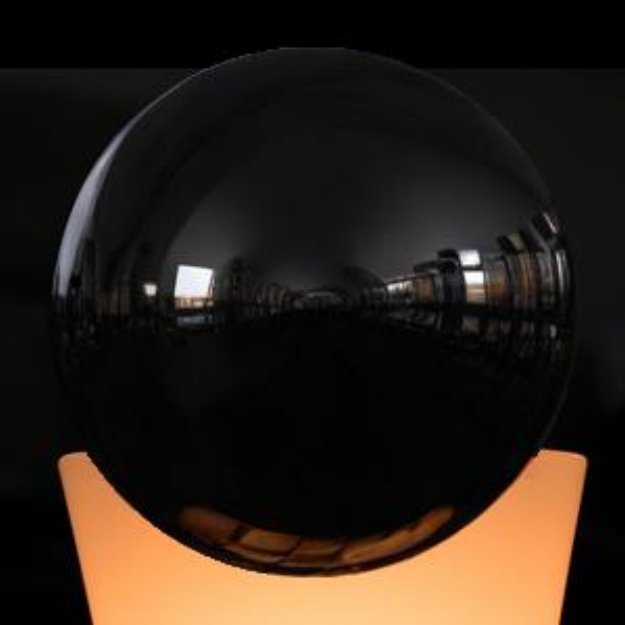}} & 
        \noindent\parbox[c]{0.092\textwidth}{\includegraphics[width=0.092\textwidth]{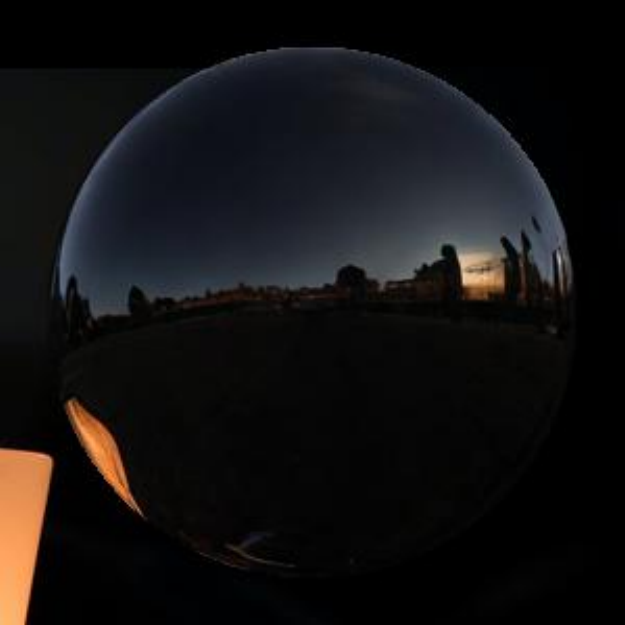}} & 
        \noindent\parbox[c]{0.092\textwidth}{\includegraphics[width=0.092\textwidth]{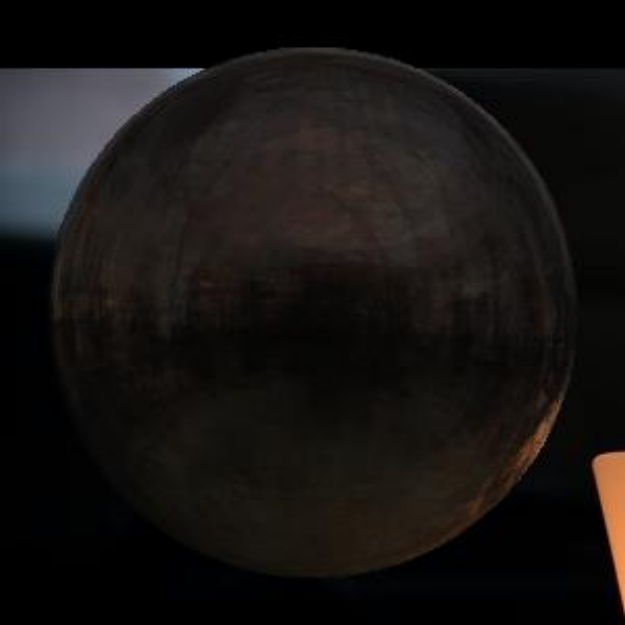}} & 
        \noindent\parbox[c]{0.092\textwidth}{\includegraphics[width=0.092\textwidth]{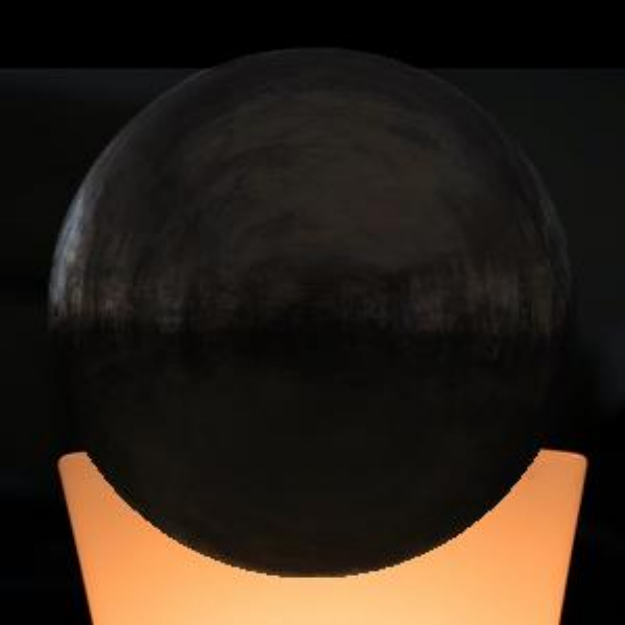}} & 
        \noindent\parbox[c]{0.092\textwidth}{\includegraphics[width=0.092\textwidth]{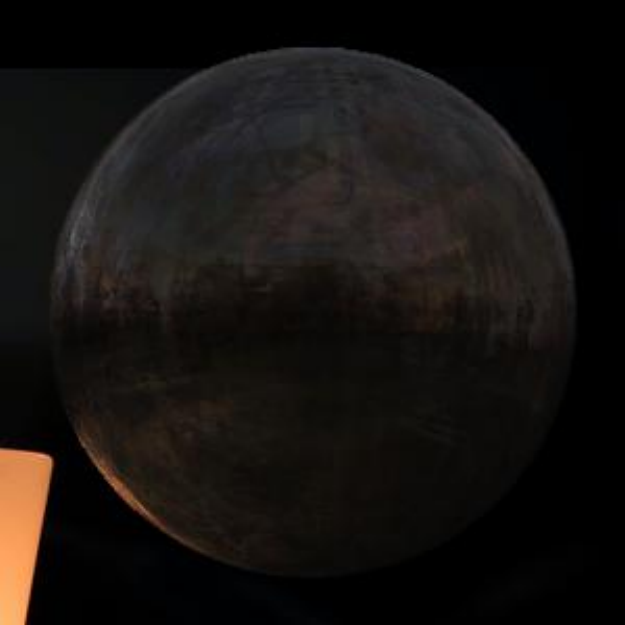}} & 
        \noindent\parbox[c]{0.092\textwidth}{\includegraphics[width=0.092\textwidth]{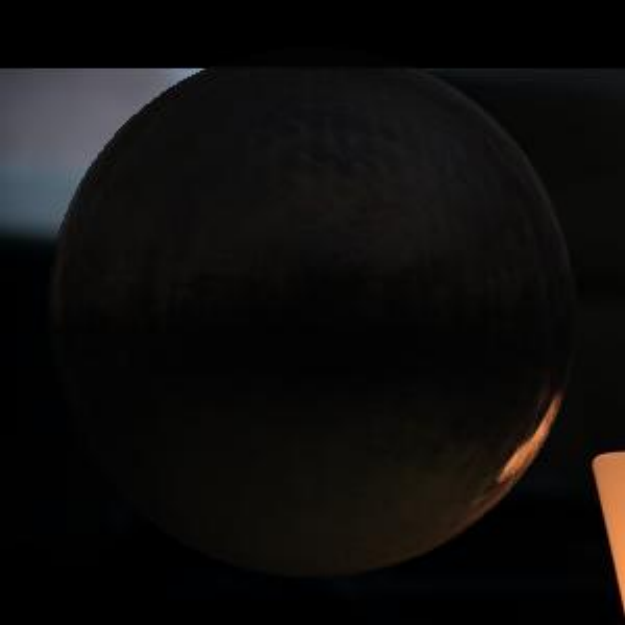}} & 
        \noindent\parbox[c]{0.092\textwidth}{\includegraphics[width=0.092\textwidth]{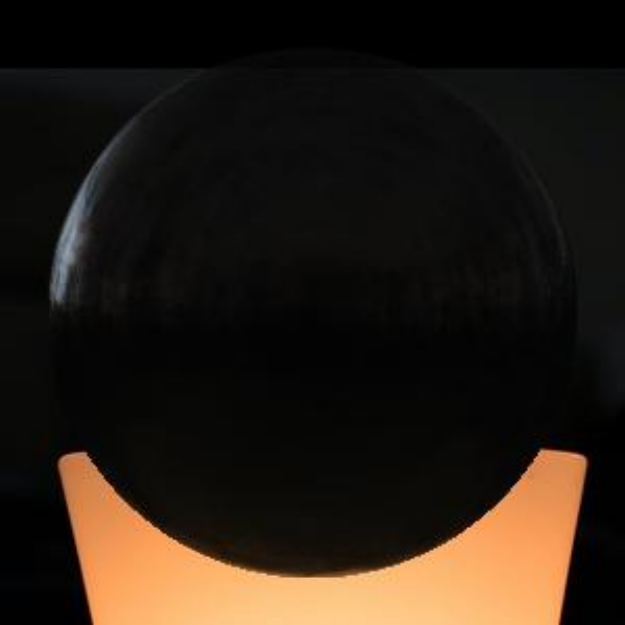}} & 
        \noindent\parbox[c]{0.092\textwidth}{\includegraphics[width=0.092\textwidth]{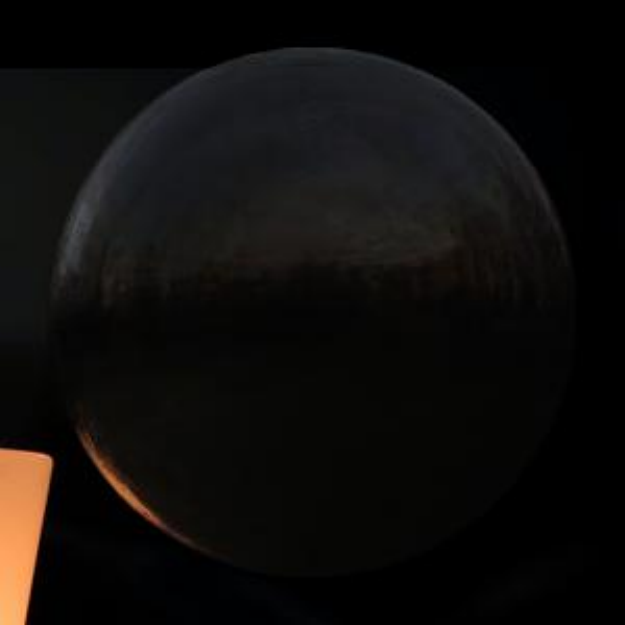}} & 
        
        \\

        & 
        \noindent\parbox[c]{0.092\textwidth}{\includegraphics[width=0.092\textwidth]{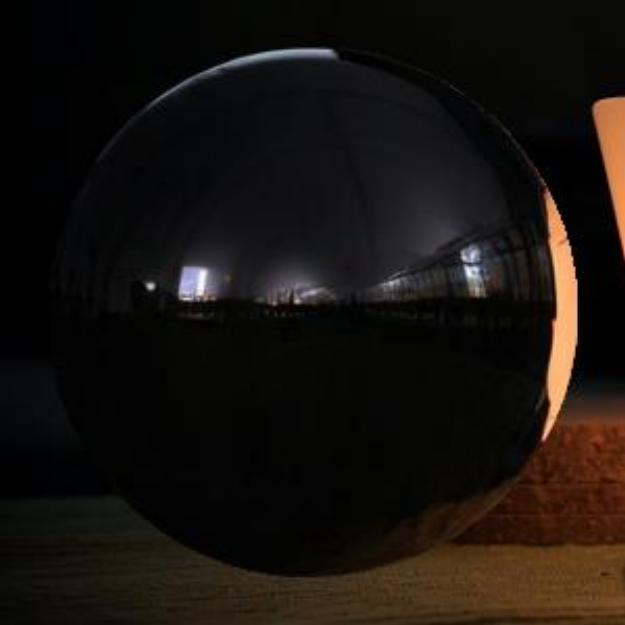}} & 
        \noindent\parbox[c]{0.092\textwidth}{\includegraphics[width=0.092\textwidth]{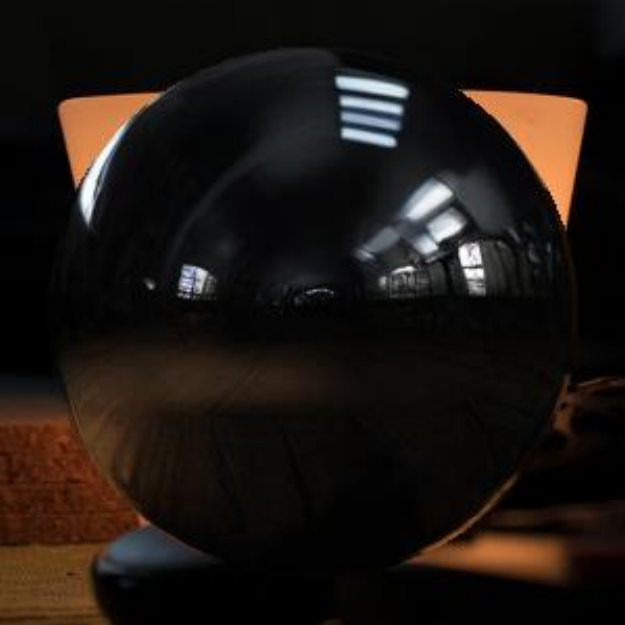}} & 
        \noindent\parbox[c]{0.092\textwidth}{\includegraphics[width=0.092\textwidth]{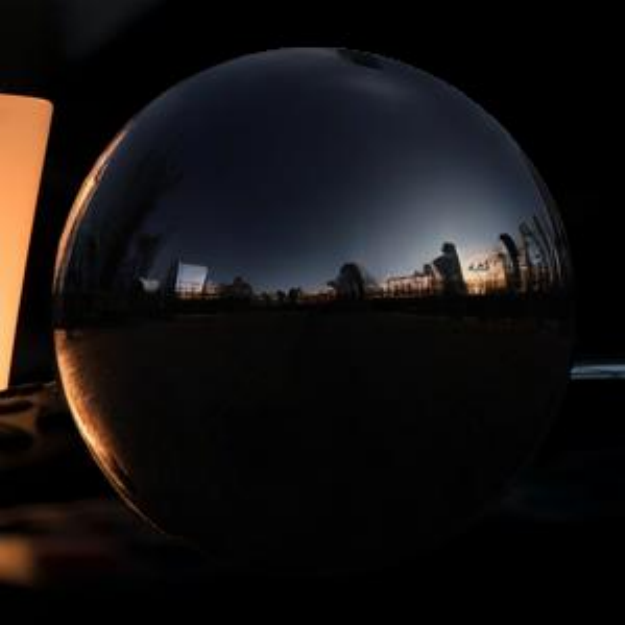}} & 
        \noindent\parbox[c]{0.092\textwidth}{\includegraphics[width=0.092\textwidth]{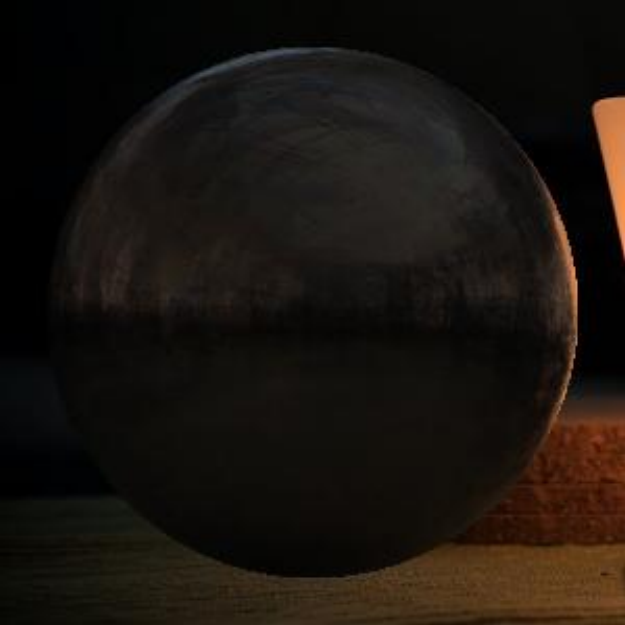}} & 
        \noindent\parbox[c]{0.092\textwidth}{\includegraphics[width=0.092\textwidth]{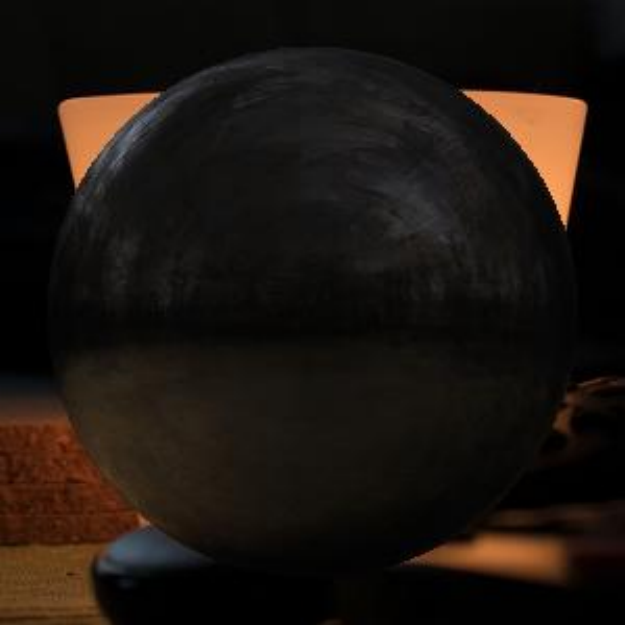}} & 
        \noindent\parbox[c]{0.092\textwidth}{\includegraphics[width=0.092\textwidth]{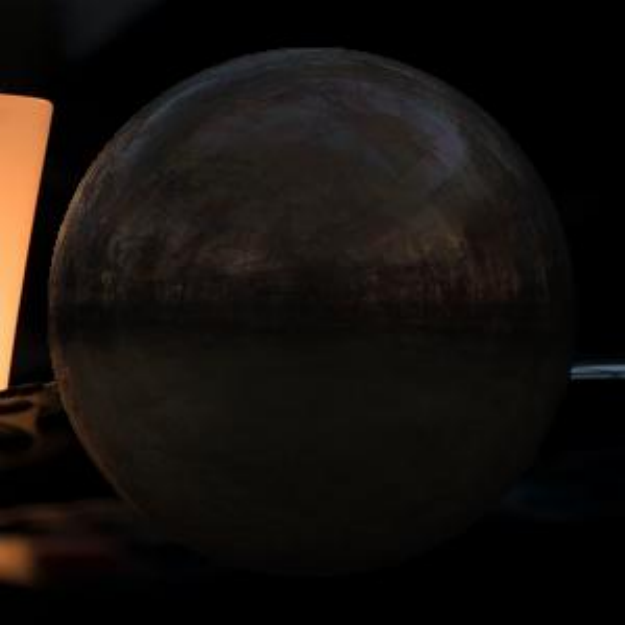}} & 
        \noindent\parbox[c]{0.092\textwidth}{\includegraphics[width=0.092\textwidth]{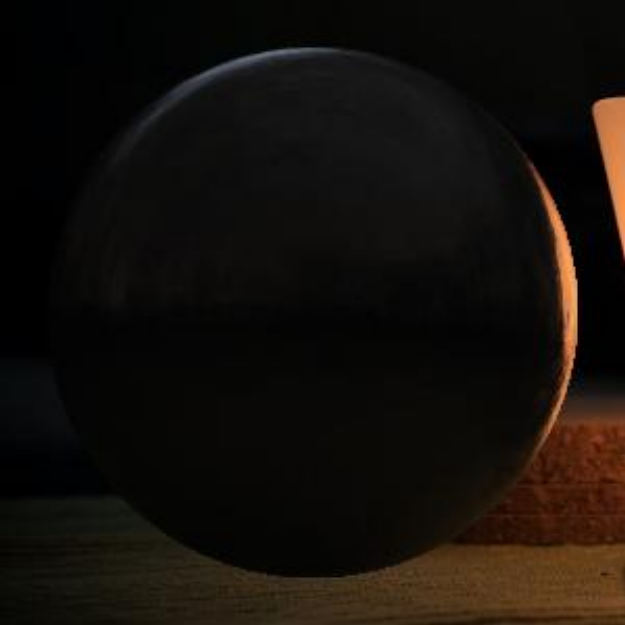}} & 
        \noindent\parbox[c]{0.092\textwidth}{\includegraphics[width=0.092\textwidth]{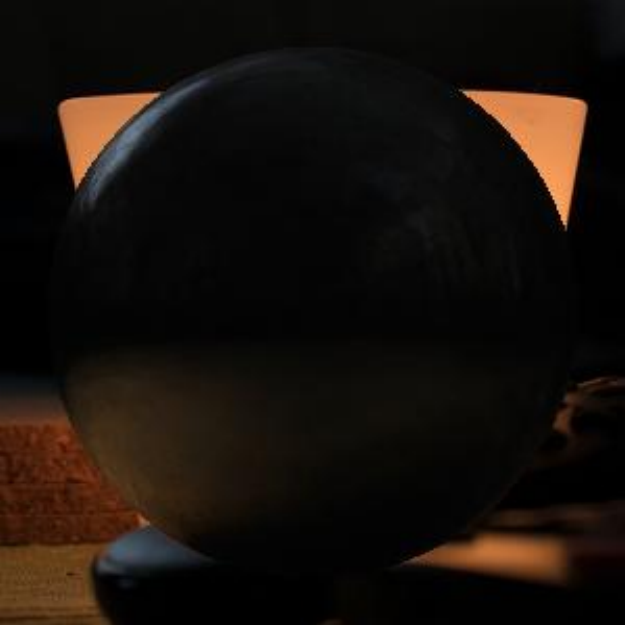}} & 
        \noindent\parbox[c]{0.092\textwidth}{\includegraphics[width=0.092\textwidth]{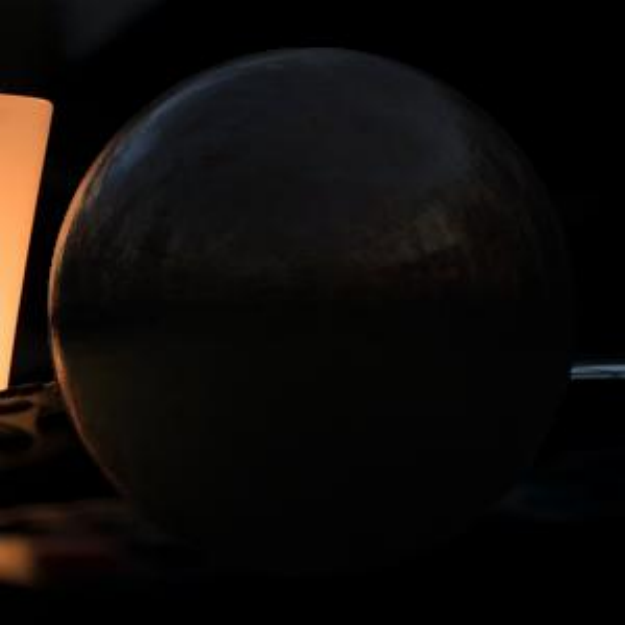}} & 
        
        \\

        & 
        \noindent\parbox[c]{0.092\textwidth}{\includegraphics[width=0.092\textwidth]{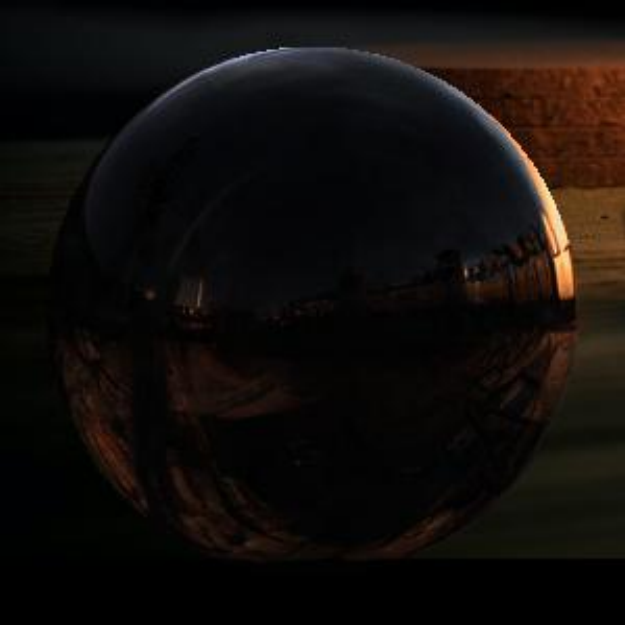}} & 
        \noindent\parbox[c]{0.092\textwidth}{\includegraphics[width=0.092\textwidth]{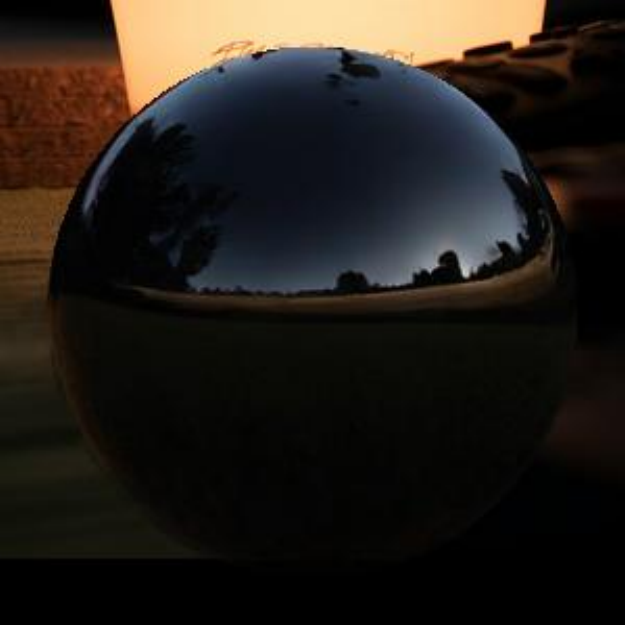}} & 
        \noindent\parbox[c]{0.092\textwidth}{\includegraphics[width=0.092\textwidth]{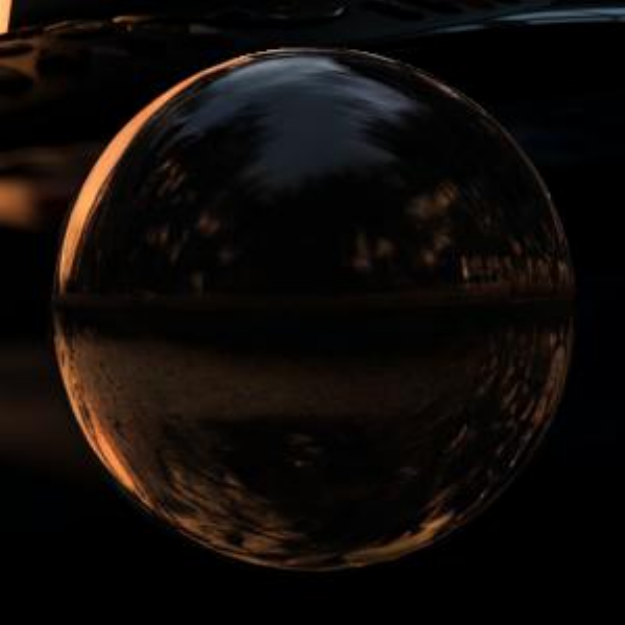}} & 
        \noindent\parbox[c]{0.092\textwidth}{\includegraphics[width=0.092\textwidth]{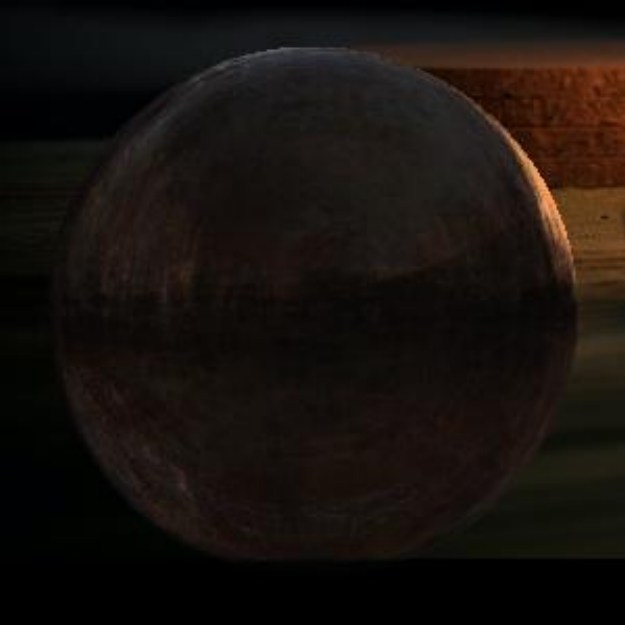}} & 
        \noindent\parbox[c]{0.092\textwidth}{\includegraphics[width=0.092\textwidth]{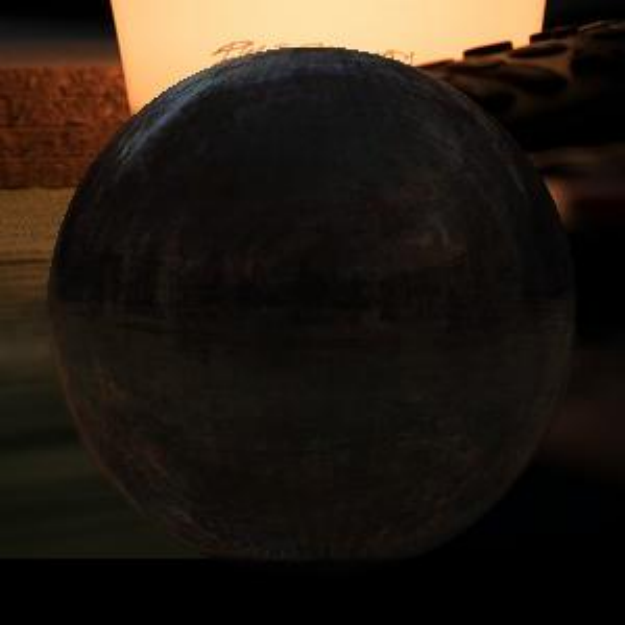}} & 
        \noindent\parbox[c]{0.092\textwidth}{\includegraphics[width=0.092\textwidth]{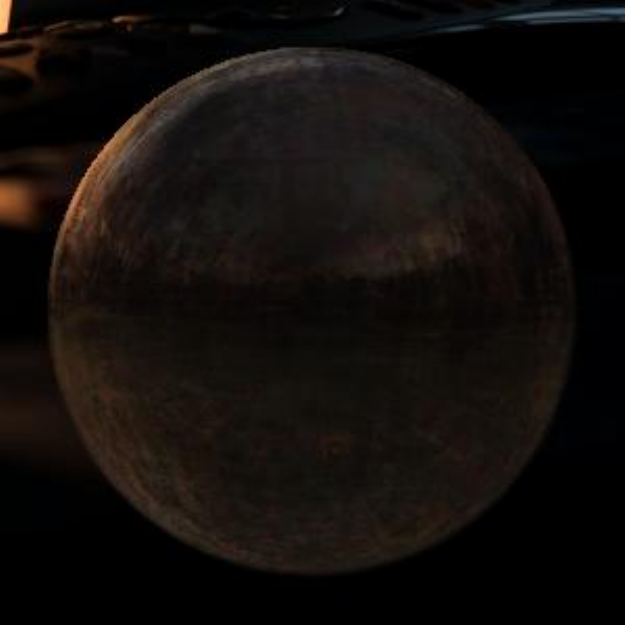}} & 
        \noindent\parbox[c]{0.092\textwidth}{\includegraphics[width=0.092\textwidth]{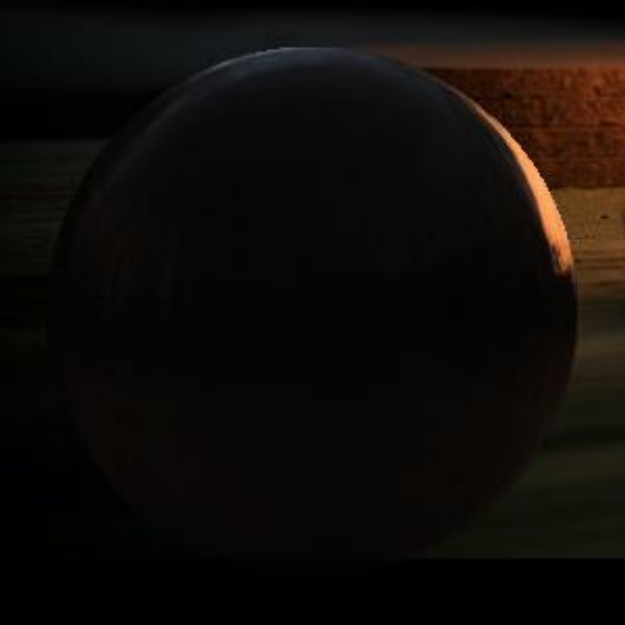}} & 
        \noindent\parbox[c]{0.092\textwidth}{\includegraphics[width=0.092\textwidth]{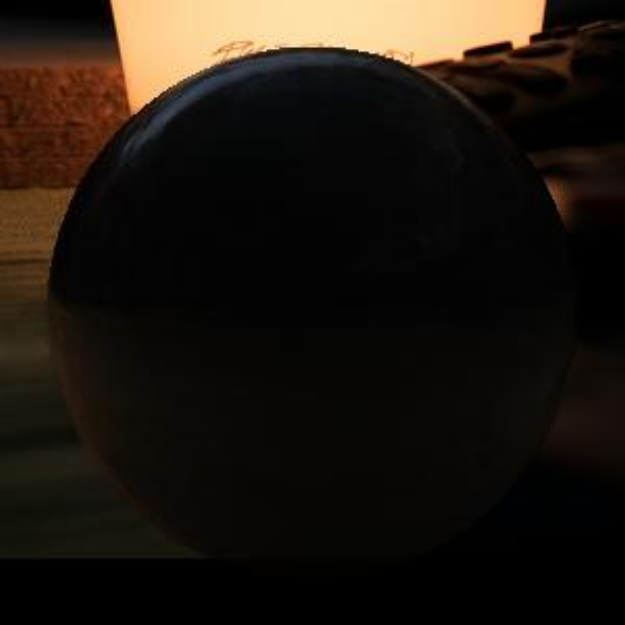}} & 
        \noindent\parbox[c]{0.092\textwidth}{\includegraphics[width=0.092\textwidth]{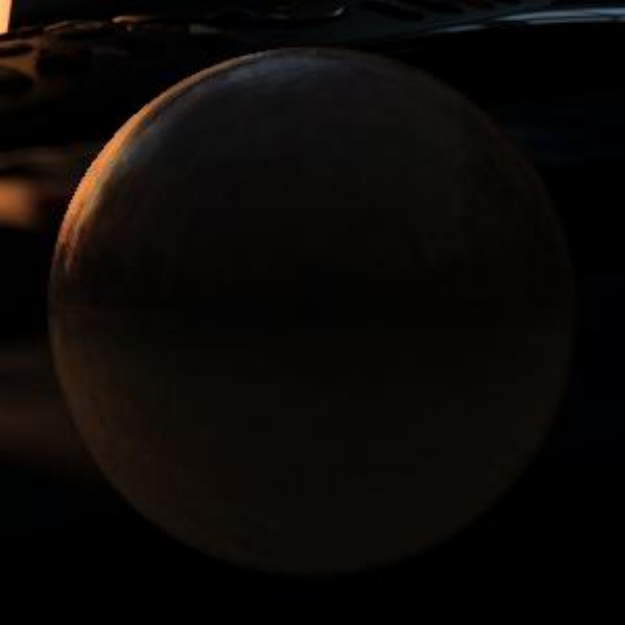}} & 
        
        \\
        
    \end{tabu}

    \caption{We show the spatially varying effects of painting a chrome ball at nine different locations, specified by red dots in the input images.
    For each input image, we present predictions from a random seed and median balls at the 1\textsuperscript{st} and 2\textsuperscript{nd} iterations. The effects can be seen in the changes of the curvature the horizon line, the size of the window, and the position of the light reflected from the lamp. These effects are more apparent in median balls as the number of iterations increases.
    } 
    
    
    
    
    \label{fig:aba_spatial_varying}
\end{figure*}

In this work, we inpaint a chrome ball in the input's center to represent global lighting in the scene and do not model any spatially varying effects by assuming orthographic projection. Nonetheless, our preliminary study suggests that the 
output from our inpainting pipeline does change according to where the chrome ball is placed in the input image, as illustrated in Figure \ref{fig:aba_spatial_varying}. This behavior can be leveraged for spatially-varying light estimation. 
To correctly infer spatially-varying, omnidirectional lighting, one needs to also infer the scene geometry, the depth of the inpainted chrome ball and camera parameters such as the focal length from the input image. These problems are interesting areas for future work.



\section{StyleLight's Score Discrepancies} \label{appendix:stylelight-score-diff}
We used StyleLight's official implementation to produce their scores in Table \ref{tab:indoor_stylelight}. However, there are discrepancies with their reported scores due to unknown implementations of their metrics.
We discussed this with the authors on GitHub\footnote{\url{https://github.com/Wanggcong/StyleLight/issues/9}}, and they clarified that additional masking of black regions and rotation of panoramas were performed before evaluation. Despite implementing these additional steps, we still could not match the scores.
The authors further suggested that we apply a consistent post-processing technique to all baselines for a fair comparison, which resulted in the scores we reported. To ensure transparency, we have made our evaluation code available at \url{https://github.com/DiffusionLight/DiffusionLight-evaluation}.

\tabulinesep=0.1pt

\tabulinesep=0.1pt
\tabulinesep=0.1pt

\end{document}